%% file: main.tex
\documentclass[10pt]{article} % For LaTeX2e
\usepackage[usenames,dvipsnames]{xcolor}
\usepackage[preprint]{tmlr}
\usepackage{hyperref}
\usepackage{url}

\usepackage{adjustbox}
\usepackage{array}
\usepackage{amsmath}
\usepackage{booktabs}
\usepackage{graphicx}
\usepackage{bbding}   % for checkmarks
\usepackage[edges]{forest}
\usepackage{multirow}
\usepackage{wrapfig}
\usepackage{pdflscape}

\title{Unifying the Perspectives of NLP and Software Engineering:\\A Survey on Language Models for Code}

\author{\name Ziyin Zhang \\
      \addr Shanghai Jiao Tong University \email daenerystargaryen@sjtu.edu.cn\\
      Ant Group
      \AND
      \name Chaoyu Chen \\
      \addr Ant Group
      \AND
      \name Bingchang Liu \\
      \addr Ant Group
      \AND
      \name Cong Liao \\
      \addr Ant Group
      \AND
      \name Zi Gong \\
      \addr Ant Group
      \AND
      \name Hang Yu\thanks{Corresponding authors.} \email hyu.hugo@antgroup.com\\
      \addr Ant Group
      \AND
      \name Jianguo Li$^*$ \email lijg.zero@antgroup.com\\
      \addr Ant Group
      \AND
      \name Rui Wang$^*$ \email wangrui12@sjtu.edu.cn\\
      \addr Shanghai Jiao Tong University}

\def\month{MM}  % Insert correct month for camera-ready version
\def\year{YYYY} % Insert correct year for camera-ready version
\def\openreview{\url{https://openreview.net/forum?id=XXXX}} % Insert correct link to OpenReview for camera-ready version

\begin{document}

\maketitle

\begin{abstract}
In this work we systematically review the recent advancements in software engineering with language models, covering 70+ models, 40+ evaluation tasks, 180+ datasets, and 900 related works. Unlike previous works, we integrate software engineering (SE) with natural language processing (NLP) by discussing the perspectives of both sides: SE applies language models for development automation, while NLP adopts SE tasks for language model evaluation. We break down code processing models into general language models represented by the GPT family and specialized models that are specifically pretrained on code, often with tailored objectives. We discuss the relations and differences between these models, and highlight the historical transition of code modeling from statistical models and RNNs to pretrained Transformers and LLMs, which is exactly the same course that had been taken by NLP. We also go beyond programming and review LLMs' application in other software engineering activities including requirement engineering, testing, deployment, and operations in an endeavor to provide a global view of NLP in SE, and identify key challenges and potential future directions in this domain.
We keep the survey open and updated on GitHub at \url{https://github.com/codefuse-ai/Awesome-Code-LLM}.
\end{abstract}

\input{sec_introduction}
\input{sec_evaluation}
\input{sec_background}
\input{sec_general-lm}
\input{sec_special-lm}
\input{sec_code-feature}

\input{sec_eco}
\input{sec_conclusion}

% \subsubsection*{Broader Impact Statement}
% As large language models keep advancing, we expect their application in software engineering to keep growing in the foreseeable future. However, we caution programmers, especially those from the industry, against over-reliance on code generated by AI systems, and urge them to always double-check machine generated code for security risks before any real-world deployment. As a mitigation to such risks, we also call on both the academia and the industry to put more emphasis on security risks in the evaluation of code-related AI systems.

% in many journals. This is optional and at the discretion of the authors. Only add
% this information once your submission is accepted and deanonymized. 

\subsubsection*{Acknowledgments}
This work is supported by Ant Group.
Use unnumbered third level headings for the acknowledgments. All
acknowledgments, including those to funding agencies, go at the end of the paper.
Only add this information once your submission is accepted and deanonymized. 

\bibliography{main}
\bibliographystyle{tmlr}

\input{appendix}

\end{document}

%% file: sec_introduction.tex
\section{Introduction}
Language modeling has advanced remarkably in recent years with the advent of pretrained Transformers~\citep{2017Transformer} such as BERT~\citep{2018BERT} and GPT~\citep{2018GPT}. As large language models (LLMs) scaled to hundreds of billions of parameters and started to display early signs of artificial general intelligence~\citep{2020GPT3,2022PaLM,2023GPT4}, their applications have also transcended text processing. Pioneered by Codex~\citep{2021Codex}, LLMs have achieved impressive results in code processing, giving rise to commercial products such as GitHub Copilot\footnote{\url{https://github.com/features/copilot}} and open-source multi-billion code models such as StarCoder~\citep{2023StarCoder} and Code LLaMA~\citep{2023CodeLLaMA}.

The application of pretrained Transformers in software engineering, however, can be traced back to dates before decoder-only autoregressive models became dominant~\citep{2020CodeBERT,2020CugLM}, and also goes beyond code processing~\citep{re-ana2023-4,UI-2024-1,log2024-1}. However, this domain is yet to witness a comprehensive review. In an attempt to bridge the gap between natural language processing~(NLP) community and software engineering~(SE) community on the topic of language model applications, we undertake a panoramic survey of language models for SE in this work, covering 70+ models, 40+ downstream tasks, 180+ datasets, and 900 related works. Putting the emphasis on code language models, we analyze their development, discuss their differences from general language models, and highlight the integration of code-specific features such as abstract syntax trees or data flows, as well as the latest techniques adapted from NLP.

Related to our work, we are aware of several surveys on similar topics, with three works concurrent to us~\citep{2023survey2,2023survey3,2023survey4}. These works, however, focus either on NLP side~\citep{2022survey1,2022survey2} or SE side~\citep{2023survey,2023survey2,2023survey3,2023survey4}, and do not cover models, tasks, and challenges from the other side. For example, \citet{2022survey1} focus on LLMs for text-to-code generation, while giving little discussion of other applications in software engineering community. \citet{2023survey2} and \citet{2023survey4}, in contrast, comprehensively review works from SE venues such as ASE and ICSE, but cite only a handful of works from deep learning and NLP venues such as ACL, EMNLP, NeurIPS, and ICLR.

\begin{figure}[t]
    \centering
    \input{trees/model}
    \caption{Our taxonomy of pretrained language models for code.}
    \label{fig:taxonomy}
\end{figure}
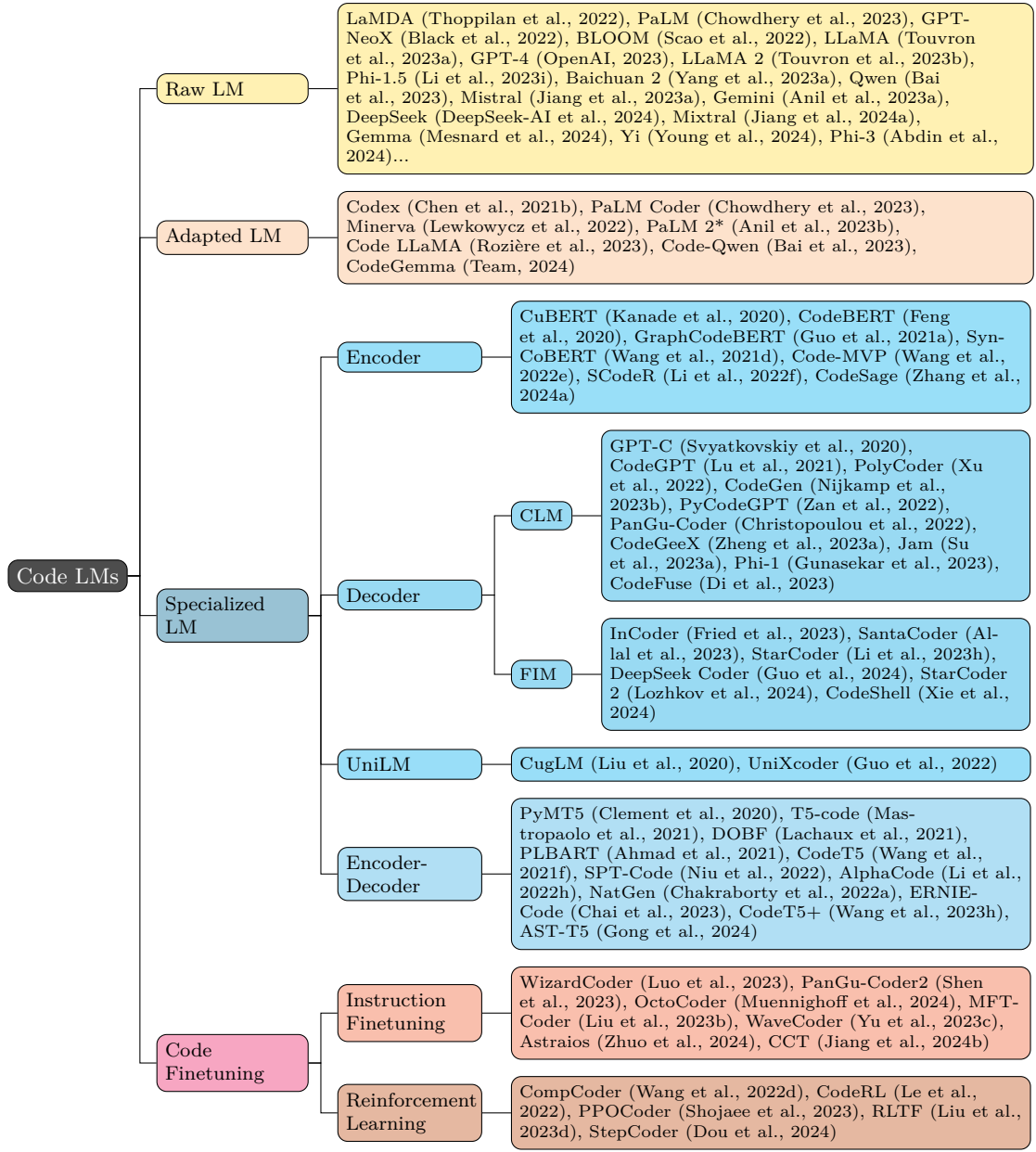

Thus, building on these works, we endeavor to unite the perspectives from both communities, and highlight the relationship between NLP and SE: SE applies language models to various tasks for automation, while NLP adopts tasks from SE to evaluate language models. We observe that advanced topics from language modeling have been recently introduced into code processing, including instruction tuning~\citep{2022Unnatural,2023WizardLM,2023WizardCoder}, infilling objectives~\citep{2022UL2,2023StarCoder,2023CodeLLaMA}, recontemplation of scaling laws~\citep{2022Chinchilla,2023Phi-1,2023Phi-1.5}, architectural improvements~\citep{2019MultiQuery,2021RoPE,2022FlashAttention}, and autonomous agents~\citep{2023ChatDev,2023MetaGPT}, while in return SE requirements, represented by programming~\citep{2021Codex}, are providing real-world testbeds for these technologies and driving the development of LLMs forward into production and deployment. We believe a systematic review of these advancements would benefit both communities. Furthermore, unlike the existing reviews that focus on programming-related tasks, we are also the first to explicitly go beyond programming and provide a global view of LLM applications in the full life cycle of software development, covering distinct stages such as requirement engineering, deployment, and operations.

The rest of this work is organized following the taxonomy presented in Figure~\ref{fig:taxonomy}. In Section~\ref{sec:evaluation} we first contextualize the downstream tasks in software engineering, highlighting the focus on programming related tasks. Then, in Section~\ref{sec:background} we provide the preliminaries of language modeling and Transformer models, and in Section~\ref{sec:general-lm} we discuss the plethora of LLMs that have demonstrated coding ability. In Section~\ref{sec:specialized-lm} we review the specialized and often smaller models by their architecture, with special attention on the recent application of infilling objectives, instruction tuning, reinforcement learning, and engineering improvements. Then, in Section~\ref{sec:code-specific}, we discuss unique features of code that are not available to natural languages but have been utilized to aid code processing. In Section~\ref{sec:eco}, we review the most recent integration between LLMs and software development, before finally concluding this work in Section~\ref{sec:conclusion} and highlighting the current challenges in code processing.

%% file: trees/model.tex
\begin{forest}
for tree={
forked edges,
draw,
rounded corners,
grow=east,
anchor=base west,
anchor=center,
reversed=true,
},
where level=0{font=\small}{},
where level=1{text width=5.5em,font=\footnotesize}{},
where level=2{text width=5.2em,font=\footnotesize}{},
where level=3{text width=1.8em,font=\scriptsize}{},
where level=4{text width=5.5em,font=\scriptsize}{},
[Code LMs, fill=black, fill opacity=0.7, text=white
    [Raw LM, fill=Goldenrod, fill opacity=0.4
        [LaMDA~\citep{2022LaMDA}{,} PaLM~\citep{2022PaLM}{,} GPT-NeoX~\citep{2022GPT-NeoX}{,} 
        BLOOM~\citep{2022BLOOM}{,} LLaMA~\citep{2023LLaMA}{,} GPT-4~\citep{2023GPT4}{,} 
        LLaMA 2~\citep{2023LLaMA2}{,} Phi-1.5~\citep{2023Phi-1.5}{,} Baichuan 2~\citep{2023Baichuan2}{,} 
        Qwen~\citep{2023Qwen}{,} Mistral~\citep{2023Mistral}{,} Gemini~\citep{2023Gemini}{,} 
        DeepSeek~\citep{2024DeepSeek}{,} 
        Mixtral~\citep{2024Mixtral}{,} 
        Gemma~\citep{2024Gemma}{,} 
        Yi~\citep{2024Yi}{,}
        Phi-3~\citep{2024Phi-3}...,
        text width=27.7em,font=\scriptsize, fill=Goldenrod, fill opacity=0.4]
    ]
    [Adapted LM, fill=Apricot, fill opacity=0.4
        [Codex~\citep{2021Codex}{,} PaLM Coder~\citep{2022PaLM}{,} Minerva~\citep{2022Minerva}{,} 
        PaLM 2*~\citep{2023PaLM2}{,} Code LLaMA~\citep{2023CodeLLaMA}{,} Code-Qwen~\citep{2023Qwen}{,}
        CodeGemma~\citep{2024CodeGemma},
        text width=27.7em,font=\scriptsize, fill=Apricot, fill opacity=0.4]
    ]
    [Specialized LM, fill=MidnightBlue, fill opacity=0.4
        [Encoder, fill=ProcessBlue, fill opacity=0.4
            [CuBERT~\citep{2019CuBERT}{,} CodeBERT~\citep{2020CodeBERT}{,} GraphCodeBERT~\citep{2020GraphCodeBERT}{,} 
            SynCoBERT~\citep{2021SynCoBERT}{,} Code-MVP~\citep{2022Code-MVP}{,} SCodeR~\citep{2022SCodeR}{,} 
            CodeSage~\citep{2024CodeSage},
            text width=20.5em, fill=ProcessBlue, fill opacity=0.4]
        ]
        [Decoder, fill=Cerulean, fill opacity=0.4
            [CLM, fill=Cerulean, fill opacity=0.4
                [GPT-C~\citep{2020GPT-C}{,} CodeGPT~\citep{2021CodeXGLUE}{,} PolyCoder~\citep{2022PolyCoder}{,} 
                CodeGen~\citep{2022CodeGen}{,} PyCodeGPT~\citep{2022PyCodeGPT}{,} PanGu-Coder~\citep{2022Pangu-Coder}{,} 
                CodeGeeX~\citep{2023CodeGeeX}{,} Jam~\citep{2023Jam}{,} 
                Phi-1~\citep{2023Phi-1}{,} CodeFuse~\citep{2023CodeFuse13B},
                text width=16.8em, fill=Cerulean, fill opacity=0.4]
            ]
            [FIM, fill=Cerulean, fill opacity=0.4
                [InCoder~\citep{2022InCoder}{,} SantaCoder~\citep{2023SantaCoder}{,} StarCoder~\citep{2023StarCoder}{,} 
                DeepSeek Coder~\citep{2023DeepSeekCoder}{,}
                StarCoder 2~\citep{2024StarCoder2}{,} 
                CodeShell~\citep{2024CodeShell},
                text width=16.8em, fill=Cerulean, fill opacity=0.4]
            ]
        ]
        [UniLM, fill=Cyan, fill opacity=0.4
            [CugLM~\citep{2020CugLM}{,} UniXcoder~\citep{2022UniXcoder},
            text width=20.5em, fill=Cyan, fill opacity=0.4]
        ]
        [Encoder-Decoder, fill=CornflowerBlue, fill opacity=0.4
            [PyMT5~\citep{2020PyMT5}{,} 
            T5-code~\citep{2021T5Code}{,} 
            DOBF~\citep{2021DOBF}{,} 
            PLBART~\citep{2021PLBART}{,} 
            CodeT5~\citep{2021CodeT5}{,} 
            SPT-Code~\citep{2022SPT-Code}{,} 
            AlphaCode~\citep{2022AlphaCode}{,} 
            NatGen~\citep{2022NatGen}{,} 
            ERNIE-Code~\citep{2022ERNIE-Code}{,} 
            CodeT5+~\citep{2023CodeT5+}{,} 
            AST-T5~\citep{2024AST-T5},
            text width=20.5em, fill=CornflowerBlue, fill opacity=0.4]
        ]
    ]
    [Code\\Finetuning, fill=WildStrawberry, fill opacity=0.4
        [Instruction Finetuning, fill=RedOrange, fill opacity=0.4
            [WizardCoder~\citep{2023WizardCoder}{,} PanGu-Coder2~\citep{2023Pangu-Coder2}{,} OctoCoder~\citep{2023OctoPack}{,} 
            MFTCoder~\citep{2023MFTCoder}{,} WaveCoder~\citep{2023WaveCoder}{,} 
            Astraios~\citep{2024Astraios}{,} 
            CCT~\citep{2024CCT},
            text width=20.5em, fill=RedOrange, fill opacity=0.4]
        ]
        [Reinforcement Learning, fill=Bittersweet, fill opacity=0.4
            [CompCoder~\citep{2022CompCoder}{,} CodeRL~\citep{2022CodeRL}{,} PPOCoder~\citep{2023PPOCoder}{,} RLTF~\citep{2023RLTF}{,} 
            StepCoder~\citep{2024StepCoder},
            text width=20.5em, fill=Bittersweet, fill opacity=0.4]
        ]
    ]
]
\end{forest}

%% file: sec_evaluation.tex
\section{Downstream Tasks in Software Engineering}\label{sec:evaluation}
\begin{figure}
    \centering
    \includegraphics[width=1\textwidth]{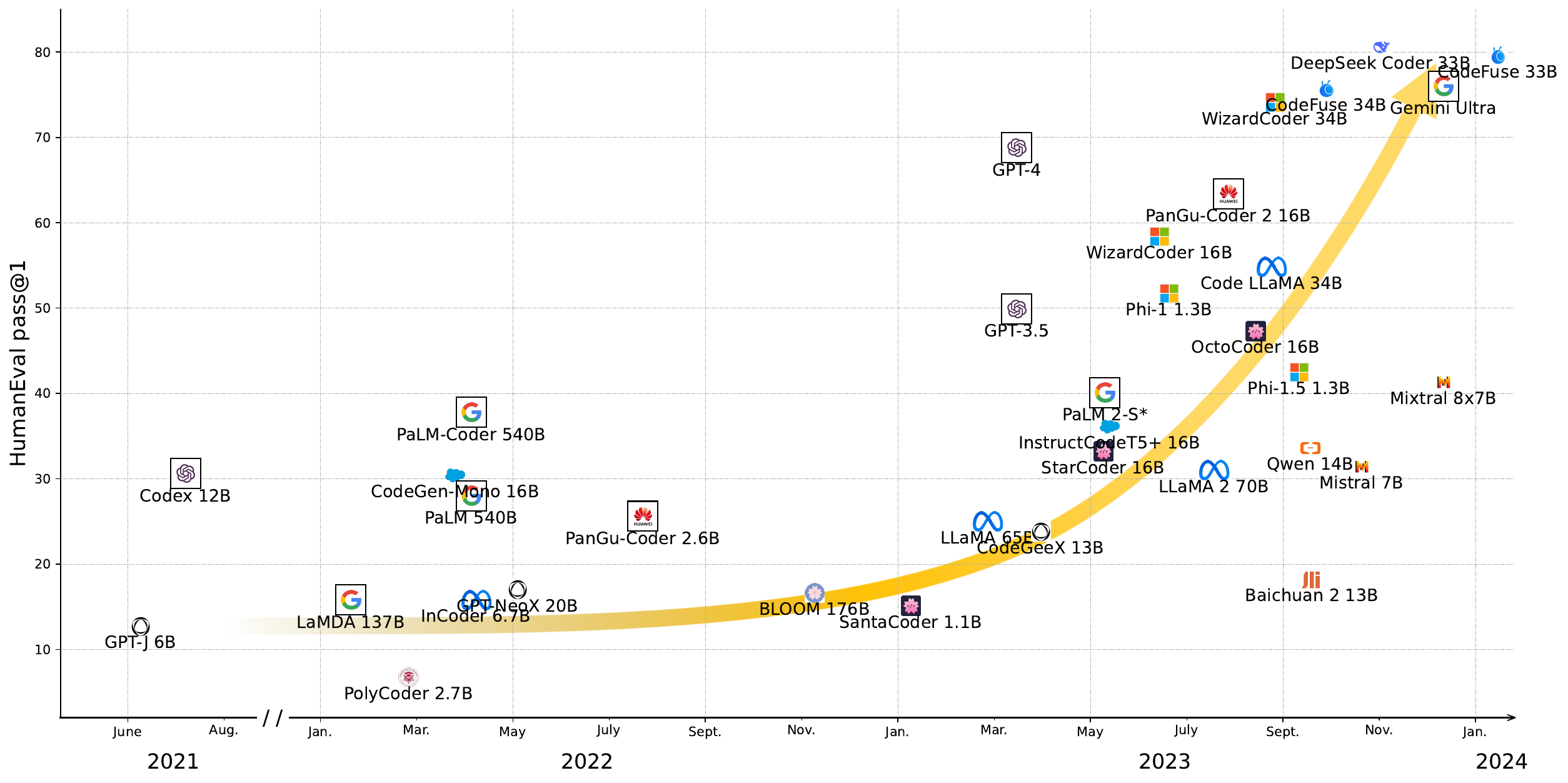}
    \caption{The timeline of code language models' progress on HumanEval.}
    \label{fig:timeline}
\end{figure}

Over the past decade, the SE community has found applications of language models in various downstream tasks. CodeXGLUE~\citep{2021CodeXGLUE} consolidates most of the code-related tasks into a single benchmark, while HumanEval~\citet{2021Codex} brought NL-to-code synthesis into the spotlight in the NLP community, which has since become a standard task for evaluating LLMs (Figure~\ref{fig:timeline}). However, other tasks, especially those not directly related to coding, have remained understudied. 

In this section, we first briefly introduce each of the traditional SE tasks and the application of pretrained language models in them in \ref{sec:evaluation-downstream}, and provide a comprehensive list of related works for each task. Then, we review the evaluation metrics in \ref{sec:evaluation-metrics} and investigate program synthesis in more detail in \ref{sec:evaluation-text2code}. Lastly, we also discuss the latest trend of repository-level coding in \ref{sec:evaluation-repo}. In Appendix~\ref{sec:benchmark} and \ref{sec:performance} respectively, we list benchmarks for each downstream task and language models' performance on them.

\subsection{Downstream Tasks in SE}\label{sec:evaluation-downstream}
The custom in software engineering is to categorize downstream tasks according to their input/output modality, such as NL-to-PL (also referred to as text-to-code) tasks and PL-to-NL (i.e. code-to-text) tasks~\citep{2021CodeXGLUE}. However, as we are dedicated to the full life cycle of software development, we adopt a different taxonomy in this work, and classify downstream tasks according to the stages in software development: requirement engineering~(\ref{sec:tasks-requirement}), development~(\ref{sec:tasks-development}), testing and analysis~(\ref{sec:tasks-testing}), deployment and operations~(\ref{sec:tasks-deployment}), and lastly several novel tasks recently proposed for evaluating LLMs~(\ref{sec:tasks-llm}). We note that this taxonomy is interleaved with the understanding-generation dichotomy in NLP, since each category may contain both understanding and generation tasks, as discussed in \ref{sec:evaluation-downstream-NLP-POV}.

\subsubsection{Requirement Engineering}\label{sec:tasks-requirement}
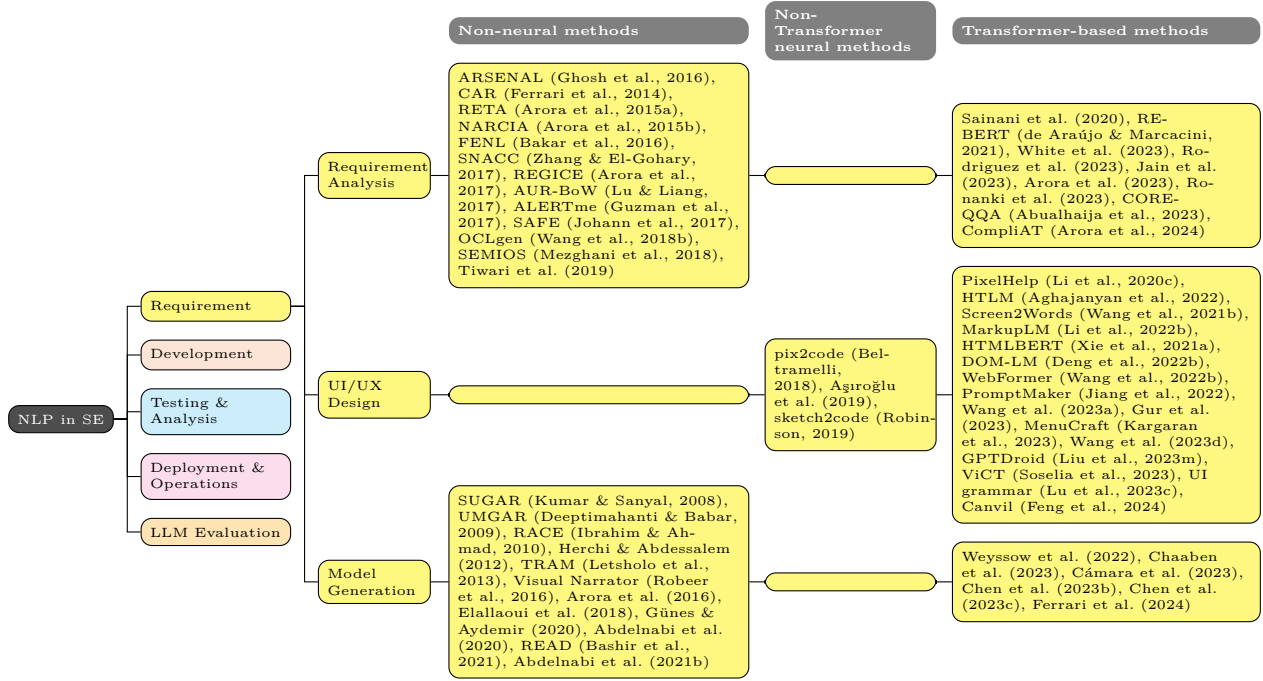
\begin{figure}[!ht]
    \centering
    \adjustbox{width=\textwidth+0.5cm,center}{
        \input{trees/task1-requirement}
    }
    \caption{NLP applications in the \emph{requirement engineering} stage of SE.}
    \label{fig:tasks-requirement}
\end{figure}

Requirement engineering refers to the specification, analysis, and validation of software requirements, software modeling, and UI/UX design. Related works are listed in Figure~\ref{fig:tasks-requirement}.

- \emph{Requirement analysis} refers to the process of extracting, summarizing, classifying, and validating software requirements. Most early works in this field rely on statistical NLP technologies such as POS (Part-of-Speech) tagging and dependency parsing to extract actions from requirements, while more recent works utilize LLMs to directly classify, summarize, or extract requirement clauses. \citet{re-ana-survey-2020-1} provide a fine-grained survey on requirement analysis.

- \emph{UI/UX Design}, short for User Interface and User Experience design, is a fundamental step in software development. While many works on UI design utilize computer vision technologies for layout design~\citep{UI-CV1,UI-CV2,UI-CV3}, we only focus on the intersection of this task and NLP technologies, such as using language models to generate markup languages.

- \emph{Model generation} aims to generate software models in modeling languages such as UML (Unified Modeling Language). \citet{re-model-survey2021-1} and \citet{re-model-survey2022-1} provide comprehensive reviews on this task before the rise of LLMs.

\begin{figure}[!ht]
    \centering
    \adjustbox{width=\textwidth+0.5cm,center}{
        \input{trees/task2-development}
    }
    \caption{NLP applications in the \emph{development} stage of SE.}
    \label{fig:tasks-development}
\end{figure}

\subsubsection{Development}\label{sec:tasks-development}
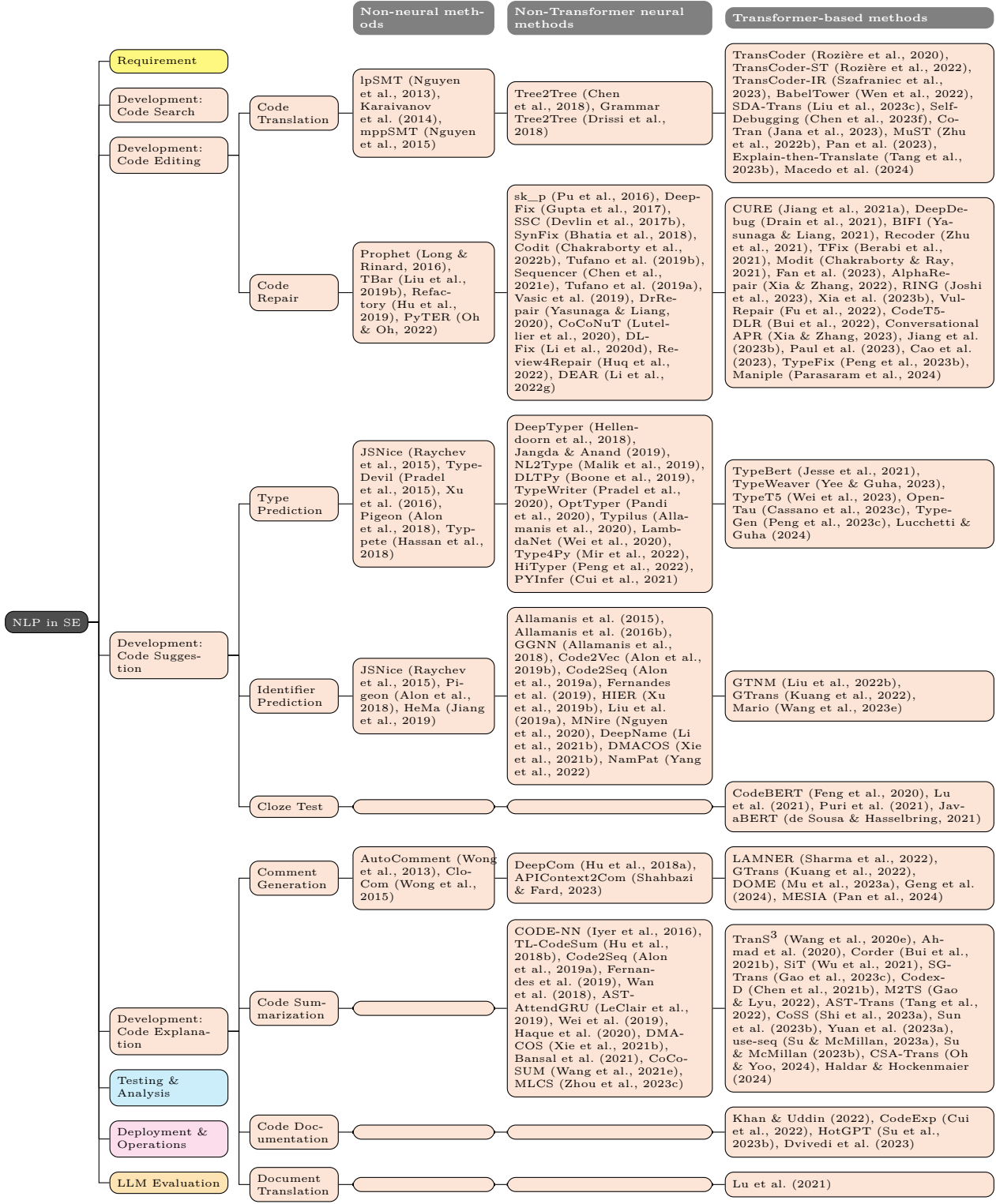
\begin{figure}[!ht]
    \centering
    \adjustbox{width=\textwidth+0.5cm,center}{
        \input{trees/task2-development2}
    }
    \caption{NLP applications in the \emph{development} stage of SE.}
    \label{fig:tasks-development2}
\end{figure}

The development phase is the stage where most programming activities take place, and also the stage where LLM applications are most abundant. We further categorize this stage into different activities such as code search, code generation, code editing, code suggestion, and code explanation. Related works are listed in Figure~\ref{fig:tasks-development} and \ref{fig:tasks-development2}.

\paragraph{Code Search}\phantom{.}

- \emph{NL-to-code search}, also referred to as text-to-code search, aims to retrieve relevant code given natural language queries, or to mine parallel text-code pairs from an unannotated corpus. This task is usually performed by computing a similarity metric between the embedding of query and candidate code, and the contextual embeddings produced by bidirectional language models - such as BERT - has proven to be extremely helpful. \citet{retrieval-survey-2022-1} and \citet{retrieval-survey-2023-1} provide comprehensive reviews on this topic.

- \emph{Code-to-code search} is a similar task where the input is an existing code snippet, often in a different programming language from the target. Code-to-code search can be reformulated as finding clones of the query in the pool of targets, and is thus equivalent to clone detection to some extent.

- \emph{API mining} refers to the process of finding similar APIs in different libraries, potentially in different programming languages. API mining is traditionally tackled by computing similarity metrics between source and target APIs using information retrieval models, but as generative models become ever more capable, it is also worth exploring to directly generate the target API as a sequence-to-sequence task. Another closely related task is \emph{idiom mining}~\citep{idiom2014-1}, where the objective is to discover commonly used code patterns, which exposes the potential need for new APIs~\citep{idiom2021-1}.

\paragraph{Code Generation}\phantom{.}

- \emph{Program synthesis} aims to generate code (usually a function or a method) given a natural language description. This task can be viewed as an updated version of NL-to-code retrieval using generative models instead of retrieval models. Statistical machine translation (SMT) and neural machine translation (NMT) models have been widely adopted for this task, often with enhanced decoders that leverage the unique grammatical rules of programming languages~\citep{syn2017-2,syn2017-3}. Pretrained language models based on Transformer architecture, however, changed the game by directly generating the source code in the autoregressive language modeling style, even without task-specific finetuning~\citep{2021Codex}. We discuss this task in more detail in Section~\ref{sec:evaluation-text2code}.

- \emph{Code completion} aims to complete a piece of code given its prefix, and remains to date one of the most popular applications of code language models in IDEs. This is essentially language modeling applied to code (which we dub ``code modeling''), and related technologies have been progressively introduced: n-gram, RNN, and Transformer. However, due to the structured nature of programming languages, many early works found grammar-aided statistical models to perform better~\citep{completion2016-2,completion2017-1}, and neural models only became dominant after 2018.

- \emph{Code infilling} is another recently proposed task, after fill-in-the-middle pretraining~\citep{2022FIM} became popular. It is a generalization of code completion, where not only the left context but also the right context is given.

- \emph{Text-to-SQL} is a special case of program synthesis, where the model is tasked to generate SQL commands from natural language queries. It has been a topic of special interest in the NLP community (as can be seen from Figure~\ref{fig:tasks-development}), and one hypothesis for this is that SQL's declarative nature and restricted grammar make it easier to optimize compared with imperative programming languages such as C and Python~\citep{SQL-optimizer1,SQL-optimizer2}. We refer to \citet{sql-survey-2022-1,sql-survey-2022-2,sql-survey-2022-3,sql-survey-2023-1} for surveys on this topic.

\paragraph{Code Editing}\phantom{.}

- \emph{Code translation} aims to translate a piece of code (usually a function or method) into another programming language. The relation between code translation and cross-lingual code search is similar to the one between program synthesis and text-to-code retrieval, and SMT/MNT models have also been widely applied to this task. One of the important applications of code translation is migrating old projects written in obsolete languages. However, we are yet to witness such applications at scale in the LLM era, as the context window of even the most powerful language models are quite limited in the face of such projects. \citet{trans-survey-2023-2} provide a short survey on this task from the SE perspective.

- \emph{Program repair}, also known as bug fix, aims to fix a piece of buggy code. Like code translation, it is a traditional sequence-to-sequence generation task, and surveys are abundant on this topic~\citep{fix-survey-2017,fix-survey-2018,fix-survey-2022,fix-survey-2023,fix-survey-2023-2}.

\paragraph{Code Suggestion}\phantom{.}

- \emph{Type prediction} aims to predict the type of variables in dynamic programming languages such as Python and JavaScript. It has been used as a pretraining objective for code language models~\citep{2022Code-MVP}, where it is often simplified as a binary tagging task to predict which tokens in the code are identifiers~\citep{2021SynCoBERT,2021CodeT5}.

- \emph{Identifier prediction} is the task of predicting identifier names in the code. As these names are deemed to contain important semantic information, this task has been utilized for code summarization~\citep{id2016-1}, as well as pretraining code models~\citep{2021CodeT5,2022SPT-Code}. A special case of identifier prediction is \emph{method name prediction}.

- \emph{Cloze test} is a recently proposed task for code understanding, after the rise of BERT-style pretraining. Due to the unique semantics of programming languages, several keywords are often selected for this test, such as \texttt{min} and \texttt{max}~\citep{2020CodeBERT}.

\paragraph{Code Explanation}\phantom{.}

- \emph{Code summarization}, \emph{comment generation}, and \emph{code documentation} all aim to generate explanations for code to facilitate understanding of the code, but with slightly different emphasis. Code summarization generates a natural language summary of the given code, while comment generation generates comments. These two terms are often used interchangeably in the literature, and their intended audience is usually developers. Code documentation, on the other hand, is more structured, includes more detailed information about the code's usage and functions, and is usually intended for a broader audience. \citet{sum-survey-2022} provide a survey on code summarization.

- \emph{Document translation} is the automatic translation of code-related documents. Since models, datasets, and prompting strategies for machine translation are abundant in NLP~\citep{2017Transformer,2021Flores101,2023MAPS}, we do not go into detail about this task.

\subsubsection{Testing and Analysis}\label{sec:tasks-testing}
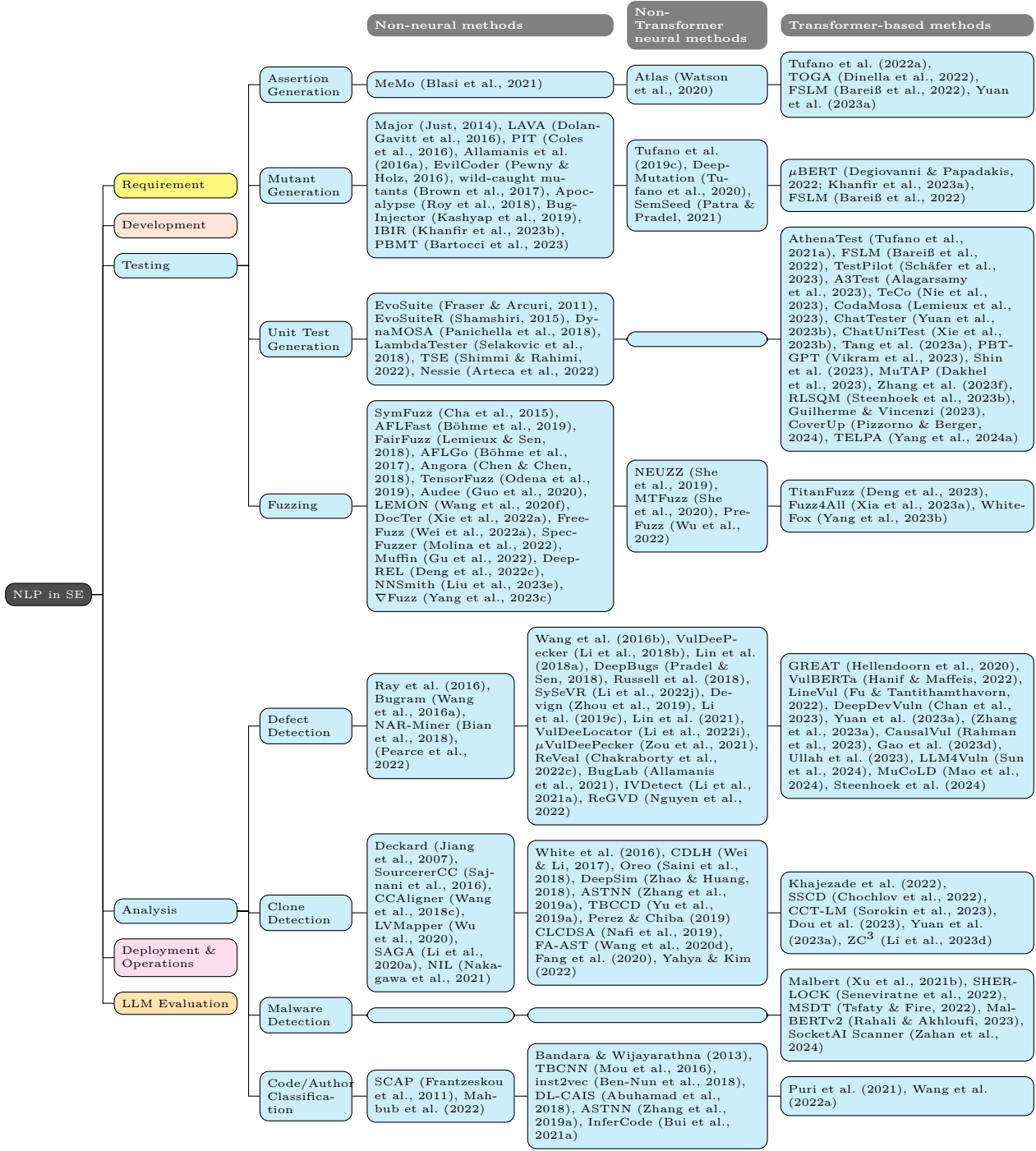
\begin{figure}[!ht]
    \centering
    \adjustbox{width=\textwidth+0.5cm,center}{
        \input{trees/task3-testing}
    }
    \caption{NLP applications in the \emph{testing \& analysis} stage of SE.}
    \label{fig:tasks-testing}
\end{figure}

The third stage in software development is about testing and analyzing the correctness of programs (as well as other properties, such as security). Related works are listed in Figure~\ref{fig:tasks-testing}.

\paragraph{Testing}\phantom{.}

- \emph{Unit test generation} aims to generate unit tests for a given program. Prior to the rise of Codex and other code LLMs, almost all works in this area employed non-neural methods (see Figure~\ref{fig:tasks-testing}). In the age of LLMs, however, this task is ever more important, as research has shown that the current unit tests for evaluating LLMs' program synthesis capability may be insufficient~\citep{2023EvalPlus}. \citet{test-survey-2023} provide a comprehensive survey on software testing with LLMs.

- \emph{Assertion generation} is a subtask of unit testing. Given a program and a partial unit test, this task aims to generate assertions (also known as \emph{oracles} in software engineering) within the unit test. This task has generally went unnoticed by the NLP community, as the program synthesis task used for evaluating LLMs often concern standalone, competition-style methods, for which the simple assertion of the equality between program output and expected answer suffices.

- \emph{Mutant generation} aims to generate mutants of a given program for the purpose of mutation testing, and relates closely to unit test generation and assertion generation. A mutant that is not detected by a given set of unit tests and assertions indicates that either additional test cases or better assertions are required~\citep{unit2011-1}. Recently, masking out tokens in the source code and sampling them from the output of a masked language model has become a common method for this task. \citet{mutant2021-2,mutant2023-3} give empirical comparisons between different mutation methods.

- \emph{Fuzzing} is another software testing task, where the objective is to generate a large set of inputs covering as many corner cases as possible. While many recent works on fuzzing target deep learning libraries, few have utilized language models to conduct this process (see Figure~\ref{fig:tasks-testing}).

\paragraph{Analysis}\phantom{.}

- \emph{Defect detection}, or \emph{vulnerability detection}, predicts whether the input code is buggy or not, and is a standard single-sentence classification task. \citet{defect-survey-2022-1,defect-survey-2022-2,defect-survey-2023-1,defect-survey-2023-2,defect-survey-2024-1} provide surveys on this task.

- \emph{Malware detection} is a similar task to defect detection. However, malware differs from other types of vulnerable code in that the bugs therein are malicious - i.e. they are intentionally injected. We refer to \citet{malware-survey2020-1} and \citet{malware-survey2022-1} for reviews on non-Transformer based methods for this task.

- \emph{Clone detection} predicts whether or not two pieces of code are clones of each other. In software engineering there exist four types of code clones, and the most challenging type to identify is semantic clones, i.e. syntactically dissimilar code that have the same functionality. As this task can be viewed as a two-sentence classification task, BERT-style language models have been widely applied to it. \citet{clone-survey-2020-1} and \citet{clone-survey-2021-1} provide comprehensive reviews on non-deep-learning based methods for this task.

- \emph{Code classification} aims to predict the functionality of a piece of code within a predefined set of labels. A very similar task is \emph{author identification}, which predicts the author of the input code. Both tasks are standard single-sentence classification tasks, and traditional machine learning methods have been widely adopted in them~\citep{author-survey-2019-1}, while pretrained language models have seen almost no application.

\subsubsection{Deployment and Operations}\label{sec:tasks-deployment}
\begin{figure}[!ht]
    \centering
    \adjustbox{width=\textwidth+0.5cm,center}{
        \input{trees/task4-devops}
    }
    \caption{NLP applications in the \emph{deployment \& operations} stage of SE.}
    \label{fig:tasks-devops}
\end{figure}
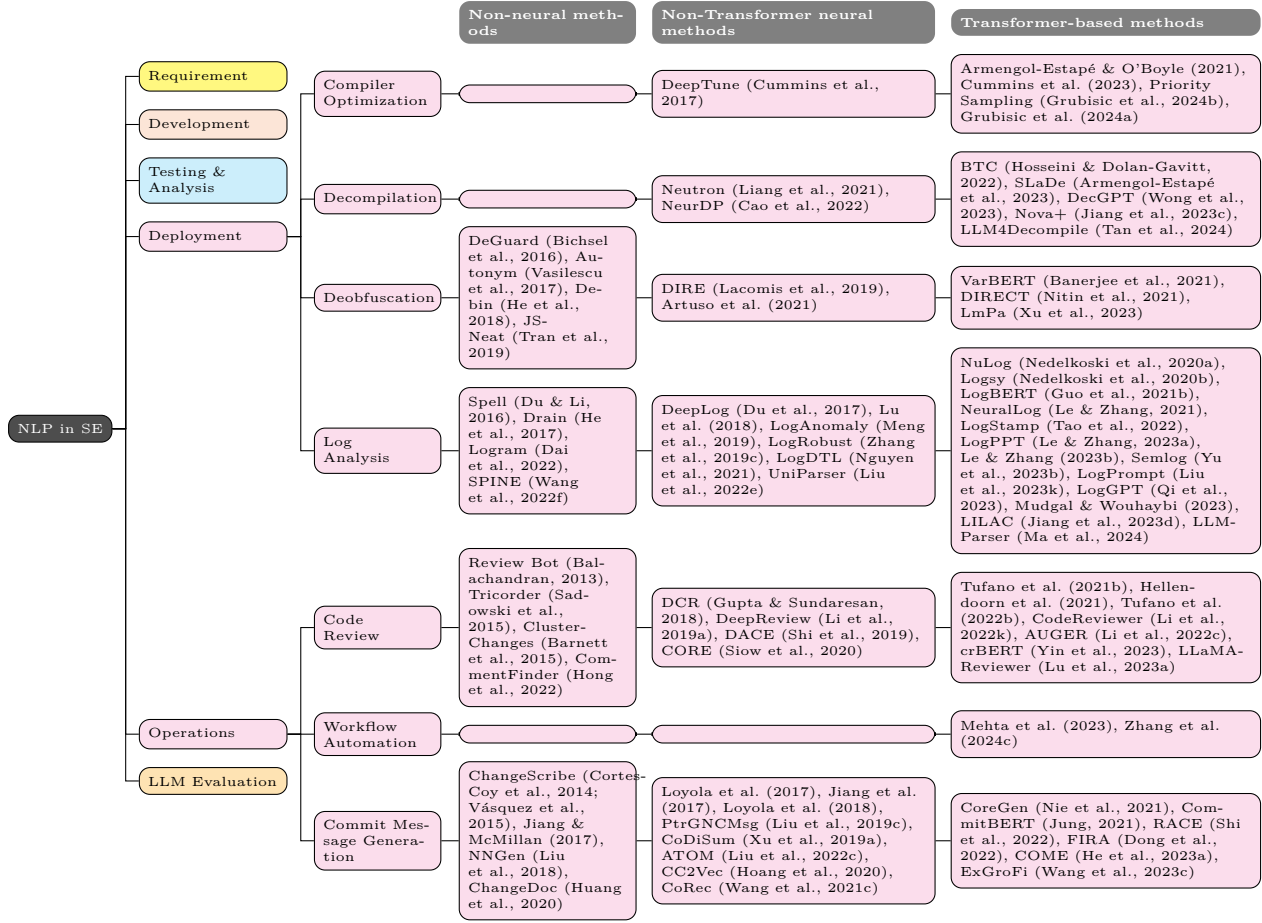

After software passes the testing stage, it is deployed into production, and often continuously maintained or updated. For this stage, we summarize the specific tasks and their related works in Figure~\ref{fig:tasks-devops}.

\paragraph{Deployment}\phantom{.}

\emph{Compiler optimization} is the task of optimizing compiled code to increase its efficiency. Many early works applied reinforcement learning and other machine learning technologies to this task by optimizing task scheduling or pass ordering~\citep{comp-opt-survey2018-1,comp-opt-survey2020-1}, and it is only recently that researchers started to view it as a sequence-to-sequence generation task with the help of powerful LLMs~\citep{comp-opt2023-1}.

- \emph{Decompilation} is the reverse process of compilation, and an important topic in reverse engineering. In this task a model takes a compiled program - represented in assembly code or binary machine code - and aims to output the original source program in high-level languages such as C.

- \emph{Obfuscation} refers to the process of renaming identifiers (e.g. variables, methods, and classes), for example to generic names like \texttt{var\_1, var\_2} or \texttt{x, y}. It is an important technique in virus detection, intellectual property protection, and code size reduction~\citep{ob2000-1,ob2009-1,ob2017-1}. \emph{Deobfuscation} refers to the reverse process, where meaningful identifier names are recovered from obfuscated programs. Obfuscation can be easily achieved statically, but deobfuscation has been a subject of more interest in recent years. It plays a significant role in decompiling, and has also been adopted as a pretraining objective for code language models~\citep{2021DOBF,2021DISCO,ob2022-1}.

- \emph{Log analysis} aims to analyze the system logs produced by software products, for example parsing logs into structured templates or finding anomalies from raw logs. \citet{log-survey-2018} provide a survey on traditional methods for this task up to 2018, and \citet{log-survey-2021-1} give an empirical comparison between neural network based methods. \citet{log-survey-2022-1} also cover more recent methods for log parsing, while \citet{log-survey-2022-2} survey methods for anomaly detection in logs.

\paragraph{Operations}\phantom{.}

- \emph{Code review} aims to automate the process of peer code review, and includes many subtasks, such as review necessity prediction, review comment generation, code refinement, and review decision prediction.

- \emph{Commit message generation} aims to automatically generate commit messages for code changes. This task takes the code before and after change as input, and output the description for the change. This can be viewed as the dual task of program repair, as many code changes and their accompanying commit messages concern bug fixing. \citet{commit-survey-2021} provide a survey on methods and datasets for this task up to 2021.

- \emph{Workflow Automation} is the automation of workflows in Git. Due to the complexity involved in this task, it was only recently brought to attention after the advent of powerful LLMs.

\subsubsection{LLM Evaluation}\label{sec:tasks-llm}
Apart from the previously covered stages in SE, we observe that programming-related evaluation of LLMs has been gaining momentum in both the NLP and the SE community since the release of Codex. Thus, we also list several novel tasks that appeared recently for this purpose. We discuss this topic in more detail in Section~\ref{sec:eco}.

\begin{figure}[!ht]
    \centering
    \adjustbox{width=\textwidth+0.5cm,center}{
        \input{trees/task5-llm}
    }
    \caption{Programming-related evaluation of LLMs.}
    \label{fig:tasks-llm}
\end{figure}
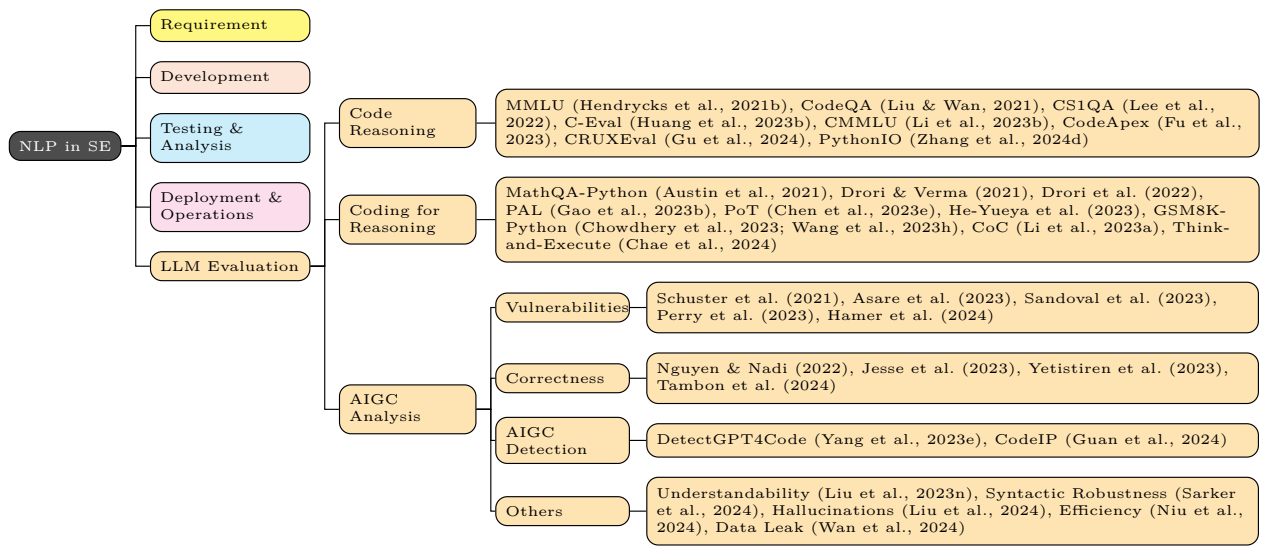

- \emph{Code reasoning} requires a model to reason about code or algorithms, and answer related questions which are written in multiple-choice format or free-form QA format, which may range from conceptual understanding to numerical calculation and complexity analysis. It often comes as a subset of composite evaluation benchmarks such as MMLU~\citep{2020MMLU}. 

- \emph{Coding for reasoning} is a special case of reasoning where LLMs solve complex reasoning problems by generating code that will be executed by external interpreters. This task abstracts the reasoning process from numerical calculations, and is thus of special interest in evaluating LLMs. More recently research has also shown that asking the LLM itself to simulate the execution of generated code can also help its reasoning process.

- \emph{AIGC analysis} aims to analyze AI-generated code\footnote{In the AI community ``AIGC'' usually refers to AI-Generated \emph{Content}. In this work we overload this expression with AI-Generated \emph{Code}.} from various aspects, such as correctness, vulnerabilities, and detection of AI-generated code.

\subsubsection{NLP Point-of-View}\label{sec:evaluation-downstream-NLP-POV}
Unlike software engineering, evaluation tasks in NLP are generally categorized into understanding and generation. The former, represented by GLUE~\citep{2018GLUE} and SuperGLUE~\citep{2019SuperGLUE}, emphasizes the comprehension of input text, and is typically formalized as classification, regression, sequence tagging, or span extraction. The later, on the other hand, involves autoregressive generation of text, such as machine translation and summarization.

Among the previously listed tasks, code synthesis, code translation, code repair, deobfuscation, unit test generation, assertion generation, mutant generation, code summarization, code review, identifier prediction, and commit message geneartion are sequence-to-sequence generation tasks. Formally, each instance of these tasks has a source sequence $\mathbf x$ (e.g. a piece of source code) and a target sequence $\mathbf y$ (e.g. its corresponding summarization), and the language model is tasked to maximize the conditional probability given by \eqref{eq:seq2seq}, where $\theta$ can be either a decoder-only model or an encoder-decoder model. In the former case, $\mathbf x$ and $\mathbf y$ are concatenated. In the later case, $\mathbf x$ is processed by the encoder and $\mathbf y$ is processed by the decoder.

Code completion and code infilling are also generation tasks, but differ from sequence-to-sequence tasks where the input and output are related by different sequences. In these two tasks, the target is a continuation or infill of the input. They correlate closely to the language modeling objectives given in Equation~\eqref{eq:clm} and \eqref{eq:seq2seq}. Similarly, cloze test takes the same form as Equation~\eqref{eq:mlm} but is usually considered an understanding task, as its output is usually a single token and does not involve autoregressive generation.

Defect detection, malware detection, clone detection, code classification, and author identification are sequence classification tasks. In these tasks, a set of labels $\mathcal Y$ is defined over the input, and each instance is assigned a label $y\in\mathcal Y$ (e.g. for defect detection $\mathcal Y=\{0, 1\}$, while for author identification a possible $\mathcal Y$ is \{Alice, Bob, John, others\}). The model is then tasked to maximize
\begin{equation}
    p_\theta(y|\mathbf{x}).
\end{equation}

Type prediction is a token classification task, also known as tagging. In this task, each token $x_i$ is assigned a label $y_i\in\mathcal Y$, with an example $\mathcal Y$ being \{int, float, string, bool, non-identifier, other\}. The model's objective is to maximize
\begin{equation}
    \prod_{i=1}^np_\theta(y_i|\mathbf{x}).
\end{equation}

The last two tasks - code retrieval and code search - also belong to understanding tasks. In these tasks, each source sequence $\mathbf x$ is paired with a positive target sequence $\mathbf y$ and a set of negative targets $\mathbf{\bar{y}}\in\{\mathbf y_1, \cdots, \mathbf y_k\}$. The model's task is to find a similarity metric $s$ such that $s(\mathbf x, \mathbf y)$ is larger than $s(\mathbf x, \mathbf{\bar{y}})$.

\subsubsection{Code LLMs for Low-Resource, Low-Level, and Domain-Specific Languages}\label{sec:evaluation-low-resource}
In NLP, human languages are categorized into high-, middle-, and low-resource languages based on the amount of available data in each language. High-resource languages such as English are extensively studied, while low-resource languages such as Swahili and Yoruba often rely on transfer learning from other languages to improve performance due to data scarcity~\citep{2019XLM-R,2020mT5,2023MELA}.

This phenomenon also exists in code modeling, as most works listed in Figure~\ref{fig:tasks-requirement}-\ref{fig:tasks-llm} target the most popular programming languages such as C, Java, and Python. However, a major difference between NLP and SE is that in NLP, low-resource languages are often spoken by few people or even endangered, while in SE low-resource languages are often designed for specific domains and purposes, and thus have an active user community. Verilog, a hardware description language, is one such example, for which code modeling research is quite abundant~\citep{2022Verilog-1,2023VeriGen,2023RTLLM,2023VerilogEval,2023RTLFixer,2023Verilog-1,2023RTLCoder,2024MEV-LLM}.

From a different perspective, most of these popular languages are imperative languages, while few works have studied NLP and LLM applications in declarative or functional languages, except those concerning SQL and one recent work on Haskell~\citep{2024Haskel}. However, research has shown that declarative languages are more aligned for optimization since they describe the ``what'' and not ``how'' in programming~\citep{SQL-optimizer1,SQL-optimizer2}, thus they are worth more attention in the future research.

Similar to NLP, many works in code modeling have studied the transfer of model capability across languages. \citet{2022Transfer}, for example, investigate the performance of multilingual language models on Ruby (which is one of the six languages in the popular CodeSearchNet pertaining corpus~\citep{2019CodeSearchNet}, but relatively low-resourced compared with the other five), finding that multilingual models have lower performance-to-time ratio compared with monolingual ones. \citet{2023MultiPL-T}, on the other hand, propose MutiPL-T, an approach for translating high-resource training data into low-resource languages for fine-tuning. However, such research is yet scarce compared with the NLP community, but with the recent advent of colossal, multilingual datasets such as The Stack v2~\citep{2024StarCoder2} we expect to see more of it in the future.

\subsection{Evaluation Metrics}\label{sec:evaluation-metrics}
Of the tasks mentioned in Section~\ref{sec:evaluation-downstream}, the understanding tasks are similar in form to natural language understanding tasks~\citep{2018GLUE,2019SuperGLUE} and evaluated likewise by metrics such as accuracy, F1 and Mean Reciprocal Rank (MRR), while short generation tasks such as identifier prediction is also evaluated by accuracy of exact matches. Code-to-text tasks are evaluated with common metrics for text generation such as BLEU~\citep{2002BLEU}.

Evaluation of tasks involving code generation, on the other hand, is more complicated. Most early works evaluate syntactical correctness, i.e. the percentage of generations that can be successfully parsed. \citet{trans2018-1} argue against such metrics and suggest reference match instead, which is the percentage of generations that are exactly the same as the references. \citet{2020CodeBLEU} propose CodeBLUE, a variant of BLEU that takes code syntax and semantics into account by evaluating the overlap of abstract syntax tree (AST) and data flow.

As code generation models became more capable over the years, however, these metrics based on content-overlap have been found to be inadequate~\citep{2020Transcoder,2021APPS,2021MBPP}, since functionally equivalent snippets of code can differ dramatically in their lexical forms. Consequently, researchers have turned their attention to functional correctness. One popular example of such metrics is pass@$k$, proposed by \citet{2019passk} and refined by \citet{2021Codex}, which is an unbiased estimator of the model's chance in passing all unit tests of a program with any of $k$ generated samples. This metric can be generalized to pass$n@k$~\citep{2022AlphaCode}, which limits the number of model submissions to $n$ but allows filtering by unit tests given in the input from $k$ samples.

\subsection{Program Synthesis}\label{sec:evaluation-text2code}
While dozens of evaluation tasks exist in software engineering, they have generally stayed out of the focus of the NLP community until very recently. The only exception is program synthesis, which has become a standard evaluation task for LLMs since the advent of HumanEval in 2021. Looking back at this task, we identify four changes in program synthesis over the years: shift of coding paradigms (from example-based to intention-based), generalization in languages (from domain-specific languages to general-purpose languages), simplification of model architectures (from grammar-guided decoders to general-purpose language models), and application of execution-based feedback.

Many of the early methods for program synthesis are example-based~\citep{syn2013-1}, which means they induce programs from input-output examples, often in domain-specific languages (DSLs) such as FlashFill~\citep{syn2017-1} and Karel\footnote{FlashFill is used in Microsoft Excel for string transformation. Karel is a simple programming language for educational purpose.}~\citep{syn2018-3}, as these languages are usually simple in syntax and structure.

As code generation models became more capable over the years, researchers started to pay attention to program synthesis in general-purpose programming languages as well. Hearthstone~\citep{syn2016-1} and CONCODE~\citep{2018CONCODE} are two of the early datasets, representing Python and Java respectively. Each example in Hearthstone is the description of a card in the game and its corresponding class implementation, while examples in CONCODE are simply Java methods paired with their natural-language documentation crawled from public GitHub repositories. Synthesizing programs from their corresponding natural language descriptions has since then become a standard practice in program synthesis, and has led to some of the most widely used benchmarks, such as HumanEval~\citep{2021Codex}, which has even been translated into multiple languages~\citep{2022MultiPL-E,2023CodeGeeX,2023OctoPack}. Some recent benchmarks use general-purpose languages but focus on specific domains, such as data science~\citep{syn2019-1,2022DS-1000} or Jupyter notebooks~\citep{2019JuICe}, while several math reasoning benchmarks have also been converted to programming tasks, including MathQA-Python~\citep{2019MathQA,2021MBPP} and GSM8K-Python~\citep{2021GSM8K,2022PaLM,2023CodeT5+}.

Many early works argue that simply treating program synthesis as a text generation task does not utilize the underlying syntax of programming languages, and thus often use syntax-enhanced decoders to inject the target syntax as prior knowledge~\citep{syn2017-2}. LLMs, however, have demonstrated that pretrained language models are capable of generating syntactically correct programs without loss of generality. Under this setting, researchers start to \emph{execute} the generated programs and provide feedback to the generation model to inject the prior knowledge of code instead. This has recently led to the popularity of \emph{interactive coding}, which we discuss in more detail in Section~\ref{sec:llm-extention}.

\subsection{Repository-Level Coding}\label{sec:evaluation-repo}
Most tasks discussed in Section~\ref{sec:evaluation-downstream} are limited to a single file or even a single function, as cross-file code modeling poses challenges that are beyond the capability of most existing language models. Recently, however, position interpolation techniques~\citep{2023PI,2023CodeLLaMA,2023YaRN} have extended the context window of LLMs to hundreds of thousands of tokens, making it possible to contextualize coding activities within entire repositories. Several works have studied code completion~\citep{repo2022-1,repo2022-2,repo2023-1,repo2023-2,2024RepoHyper,2024Repoformer} and generation~\citep{2023A3-CodeGen,2024CodeS} leveraging repository-level context, and corresponding benchmarks have been proposed~\citet{2023RepoBench,2023CrossCodeEval,2024EvoCodeBench}. \citet{CodePlan} investigate the more challenging tasks of repository-level API migration and temporal editing, while \citet{2023SWE-bench} introduce a related benchmark, SWE-bench.

%% file: trees/task1-requirement.tex
\begin{forest}
for tree={
forked edges,
draw,
rounded corners,
grow=east,
anchor=base west,
anchor=center,
reversed=true,
l sep=0.2cm,
},
where level=0{font=\tiny}{},
where level=1{text width=4.9em,font=\tiny}{},
where level=2{text width=3.5em,font=\tiny}{},
where level=3{text width=10.5em,font=\tiny}{},
where level=4{text width=5.7em,font=\tiny}{},
where level=5{text width=10.8em,font=\tiny}{},
for tree={
  if level=0{l sep+=0.15cm}{},
  if level=1{l sep+=0.15cm}{},
  if level=2{l sep+=0.03cm}{},
},
[NLP in SE, fill=black, fill opacity=0.7, text=white
    [Requirement, fill=yellow, fill opacity=0.5
        [, draw=none, no edge
            [\textbf{Non-neural methods}, draw=none, no edge, fill=black, fill opacity=0.5, text=white
                [\textbf{Non-Transformer neural methods}, draw=none, no edge, fill=black, fill opacity=0.5, text=white
                    [\textbf{Transformer-based methods}, draw=none, no edge, fill=black, fill opacity=0.5, text=white]
                ]
            ]
        ]
        [Requirement Analysis, fill=yellow, fill opacity=0.5
            [
            ARSENAL~\citep{re-ana2014-1}{,} 
            CAR~\citep{re-ana2014-2}{,} 
            RETA~\citep{re-ana2015-1}{,} 
            NARCIA~\citep{re-ana2015-2}{,} 
            FENL~\citep{re-ana2016-1}{,} 
            SNACC~\citep{re-ana2016-2}{,} 
            REGICE~\citep{re-ana2016-3}{,} 
            AUR-BoW~\citep{re-ana2017-1}{,} 
            ALERTme~\citep{re-ana2017-2}{,} 
            SAFE~\citep{re-ana2017-3}{,} 
            OCLgen~\citep{re-ana2018-1}{,} 
            SEMIOS~\citep{re-ana2018-2}{,} 
            \citet{re-ana2019-1}
            , fill=yellow, fill opacity=0.5
                [
                , fill=yellow, fill opacity=0.5
                    [
                    \citet{re-ana2020-1}{,} 
                    RE-BERT~\citep{re-ana2021-1}{,} 
                    \citet{re-ana2023-1}{,} 
                    \citet{re-ana2023-2}{,} 
                    \citet{re-ana2023-3}{,} 
                    \citet{re-ana2023-4}{,} 
                    \citet{re-ana2023-5}{,} 
                    COREQQA~\citep{re-ana2023-6}{,} 
                    CompliAT~\citep{re-ana2024-1}
                    , fill=yellow, fill opacity=0.5
                    ]
                ]
            ]
        ]
        [UI/UX\\ Design, fill=yellow, fill opacity=0.5
            [
            , fill=yellow, fill opacity=0.5
                [
                pix2code~\citep{UI2017-1}{,} 
                \citet{UI2019-1}{,} 
                sketch2code~\citep{UI2019-2}
                , fill=yellow, fill opacity=0.5
                    [
                    PixelHelp~\citep{UI2020-1}{,} 
                    HTLM~\citep{UI2021-1}{,} 
                    Screen2Words~\citep{UI2021-2}{,} 
                    MarkupLM~\citep{UI2021-3}{,} 
                    HTMLBERT~\citep{UI2021-4}{,} 
                    DOM-LM~\citep{UI2022-1}{,} 
                    WebFormer~\citep{ui2022-2}{,} 
                    PromptMaker~\citep{UI2022-3}{,} 
                    \citet{UI2022-4}{,} 
                    \citet{UI2022-5}{,} 
                    MenuCraft~\citep{UI2023-1}{,} 
                    \citet{UI-2023-2}{,} 
                    GPTDroid~\citep{UI-2023-3}{,} 
                    ViCT~\citep{UI-2023-4}{,} 
                    UI grammar~\citep{UI-2023-5}{,} 
                    Canvil~\citep{UI-2024-1}
                    , fill=yellow, fill opacity=0.5
                    ]
                ]
            ]
        ]
        [Model\\ Generation, fill=yellow, fill opacity=0.5
            [
            SUGAR~\citep{re-model2008-1}{,} 
            UMGAR~\citep{re-model2009-1}{,} 
            RACE~\citep{re-model2010-1}{,} 
            \citet{re-model2012-1}{,} 
            TRAM~\citep{re-model2013-1}{,} 
            Visual Narrator~\citep{re-model2016-1}{,} 
            \citet{re-model2016-2}{,} 
            \citet{re-model2018-1}{,} 
            \citet{re-model2020-1}{,} 
            \citet{re-model2020-2}{,} 
            READ~\citep{re-model2021-1}{,} 
            \citet{re-model2021-3}
            , fill=yellow, fill opacity=0.5
                [
                , fill=yellow, fill opacity=0.5
                    [
                    \citet{re-model2021-2}{,} 
                    \citet{re-model2022-1}{,} 
                    \citet{re-model2023-1}{,} 
                    \citet{re-model2023-2}{,} 
                    \citet{re-model2023-3}{,} 
                    \citet{re-model2024-1}
                    , fill=yellow, fill opacity=0.5
                    ]
                ]
            ]
        ]
    ]
    [Development, fill=Orange, fill opacity=0.2]
    [Testing \&\\ Analysis, fill=cyan, fill opacity=0.2]
    [Deployment \& Operations, fill=VioletRed, fill opacity=0.2]
    [LLM Evaluation, fill=Dandelion, fill opacity=0.4]
]
\end{forest}

% \begin{forest}
% for tree={
% forked edges,
% draw,
% rounded corners,
% grow=east,
% anchor=base west,
% anchor=center,
% reversed=true,
% l sep=0.2cm,
% },
% where level=0{font=\tiny}{},
% where level=1{text width=4.9em,font=\tiny}{},
% where level=2{text width=3.5em,font=\tiny}{},
% where level=3{text width=10.5em,font=\tiny}{},
% where level=4{text width=5.7em,font=\tiny}{},
% where level=5{text width=10.8em,font=\tiny}{},
% [Title, fill=black, fill opacity=0.7, text=white
%     [Subtitle1
%         [, draw=none, no edge
%             [Column A, draw=none, no edge
%                 [Column B, draw=none, no edge
%                     [Column C, draw=none, no edge]
%                 ]
%             ]
%         ]
%         [Subsubtitle 1
%             [
%             Column A1
%                 [
%                 Column B1
%                     [
%                     Column C1
%                     ]
%                 ]
%             ]
%         ]
%         [Subsubtitle 2
%             [
%             Column A2
%                 [
%                 Column B2
%                     [
%                     Column C2
%                     ]
%                 ]
%             ]
%         ]
%         [Subsubtitle 3
%             [
%             Column A3
%                 [
%                 Column B3
%                     [
%                     Column C3
%                     ]
%                 ]
%             ]
%         ]
%     ]
%     [Subtitle2]
% ]
% \end{forest}

%% file: trees/task2-development.tex
\begin{forest}
for tree={
forked edges,
draw,
rounded corners,
grow=east,
anchor=base west,
anchor=center,
reversed=true,
l sep=0.2cm,
},
where level=0{font=\tiny}{},
where level=1{text width=4.9em,font=\tiny}{},
where level=2{text width=3.5em,font=\tiny}{},
where level=3{text width=6.0em,font=\tiny}{},
where level=4{text width=9.0em,font=\tiny}{},
where level=5{text width=12.0em,font=\tiny}{},
for tree={
  if level=0{l sep+=0.15cm}{},
  if level=1{l sep+=0.15cm}{},
  if level=2{l sep+=0.03cm}{},
},
[NLP in SE, fill=black, fill opacity=0.7, text=white
    [Requirement, fill=yellow, fill opacity=0.5]
    [Development:\\Code Search, fill=Orange, fill opacity=0.2
        [, draw=none, no edge
            [\textbf{Non-neural methods}, draw=none, no edge, fill=black, fill opacity=0.5, text=white
                [\textbf{Non-Transformer neural methods}, draw=none, no edge, fill=black, fill opacity=0.5, text=white
                    [\textbf{Transformer-based methods}, draw=none, no edge, fill=black, fill opacity=0.5, text=white]
                ]
            ]
        ]
        [NL-to-Code\\Search, fill=Orange, fill opacity=0.2
            [
            \citet{retrieval2015-1}{,} CodeHow~\citep{retrieval2015-2}{,} 
            RACS~\citep{retrieval2016-1}{,} CodeMatcher~\citep{retrieval2020-0.2}, fill=Orange, fill opacity=0.2
                [
                CODE-NN~\citep{sum2016-1}{,} DeepCS~\citep{retrieval2018-1}{,} 
                \citet{retrieval2018-2}{,} NCS~\citep{retrieval2018-3}{,} 
                UNIF~\citep{retrieval2019-1}{,} HECS~\citep{retrieval2020-0.3}{,} 
                CARLCS-CNN~\citep{retrieval2020-0.7}{,} DGMS~\citep{retrieval2020-1}{,} 
                NJACS~\citep{retrieval2020-2}{,} 
                TabCS~\citep{retrieval2021-1}{,} GraphSearchNet~\citep{retrieval2021-6}{,} 
                TranCS~\citep{retrieval2022-3}, fill=Orange, fill opacity=0.2
                    [
                    TranS$^3$~\citep{2020Trans3}{,} \citet{retrieval2020-0.4}{,} 
                    Corder~\citep{retrieval2020-0.5}{,} SAN-CS~\citep{retrieval2021-1.5}{,} 
                    SST~\citep{retrieval2021-3}{,} 
                    MuCoS~\citep{retrieval2021-4}{,} MEM~\citep{retrieval2021-5}{,} 
                    CDCS~\citep{retrieval2022-1}{,} CodeRetriever~\citep{retrieval2022-2}{,} 
                    CoCoSoDa~\citep{retrieval2022-4}{,} \citet{retrieval2022-5}{,} 
                    CCT-LM~\citep{clone2023-0.5}, fill=Orange, fill opacity=0.2
                    ]
                ]
            ]
        ]
        [Code-to-Code\\Search, fill=Orange, fill opacity=0.2
            [
            FaCoY~\citep{search2018-1}{,} Aroma~\citep{search2018-2}{,} 
            COSAL~\citep{search2021-1}, fill=Orange, fill opacity=0.2
                [, fill=Orange, fill opacity=0.2
                    [
                    Corder~\citep{retrieval2020-0.5}{,} Cosco~\citep{search2023-1}, fill=Orange, fill opacity=0.2
                    ]
                ]
            ]
        ]
        [API\\Mining, fill=Orange, fill opacity=0.2
            [
            RASH~\citep{api2017-4}{,} RAPIM~\citep{api2019-2}, fill=Orange, fill opacity=0.2
                [
                \citet{api2016-1}{,} DeepAM~\citep{api2017-1}{,} 
                JV2CS~\citep{api2017-2}{,} \citet{api2017-3}{,} 
                TL-CodeSum~\citep{sum2018-2}{,} 
                BIKER~\citep{api2018-1}{,} \citet{api2019-1}{,} 
                SAR~\citep{api2019-3}, fill=Orange, fill opacity=0.2
                    [
                    HaPiM~\citep{api2023-1}{,}
                    Code2API~\citep{api2024-1}
                    , fill=Orange, fill opacity=0.2
                    ]
                ]
            ]
        ]
    ]
    [Development:\\Code Generation, fill=Orange, fill opacity=0.2
        [Program\\ Synthesis, fill=Orange, fill opacity=0.2
            [
            Euphony~\citep{syn2018-2}{,} Neo~\citep{syn2018-4}, fill=Orange, fill opacity=0.2
                [
                LPN~\citep{syn2016-1}{,} NSPS~\citep{syn2016-2}{,} 
                DeepCoder~\citep{syn2016-3}{,} RobustFill~\citep{syn2017-1}{,} 
                \citep{syn2017-2}{,} ASN~\citep{syn2017-3}{,} 
                NGDS~\citep{syn2018-1}{,} \citet{syn2018-3}{,} 
                ReCode~\citep{syn2018-5}{,} AutoPandas~\citep{syn2019-1}{,} 
                \citet{sum2019-1}{,} PlotCoder~\citep{2021PlotCoder}{,} 
                \citet{syn2022-1}, fill=Orange, fill opacity=0.2
                    [
                    TreeGen~\citep{syn2019-2}{,} REDCODER~\citep{syn2021-1}{,} 
                    Jigsaw~\citep{syn2021-2}{,} JuPyT5~\citep{2022DSP}{,} 
                    CodeT~\citep{2022CodeT}{,} TiCoder~\citep{2022TiCoder}{,} 
                    AceCoder~\citep{syn2023-1}{,} Self-Debugging~\citep{2023self-debug}{,} 
                    ClarifyGPT~\citep{2023ClarifyGPT}{,} 
                    MANGO~\citep{2024MANGO}, fill=Orange, fill opacity=0.2
                    ]
                ]
            ]
        ]
        [Code\\ Completion, fill=Orange, fill opacity=0.2
            [
            BMN~\citep{completion2009-1}{,} MSE~\citep{completion2012-1}{,} Naturalize~\citep{completion2014-1}{,} 
            Cache LM~\citep{completion2014-3}{,} DeepSyn~\citep{completion2016-1}{,} PHOG~\citep{completion2016-2}{,} 
            Deep3~\citep{completion2016-3}{,} \citet{completion2017-1}, fill=Orange, fill opacity=0.2
                [
                \citet{completion2014-2}{,} \citet{completion2015-1}{,} 
                Pointer Mixture Network~\citep{completion2017-2}{,} \citet{completion2018-1}{,} 
                SLM~\citep{completion2019-1}{,} Pythia~\citep{completion2019-2}{,} NLM~\citep{completion2020-1}, fill=Orange, fill opacity=0.2
                    [
                    IntelliCode~\citep{2020GPT-C}{,} CugLM~\citep{2020CugLM}{,} 
                    eWASH~\citep{2021eWASH}{,} 
                    LongCoder~\citep{2023LongCoder}, fill=Orange, fill opacity=0.2
                    ]
                ]
            ]
        ]
        [Code\\ Infilling, fill=Orange, fill opacity=0.2
            [, fill=Orange, fill opacity=0.2
                [, fill=Orange, fill opacity=0.2
                    [
                    InCoder~\citep{2022InCoder}{,} FIM~\citep{2022FIM}{,} SantaCoder~\citep{2023SantaCoder}{,} 
                    StarCoder~\citep{2023StarCoder}{,} Code LLaMA~\citep{2023CodeLLaMA}, fill=Orange, fill opacity=0.2
                    ]
                ]
            ]
        ]
        [Text-to-SQL, fill=Orange, fill opacity=0.2
            [, fill=Orange, fill opacity=0.2, text width=3em
                [
                Seq2SQL~\citep{sql2017-1}{,} SQLNet~\citep{sql2017-2}{,} 
                \citet{sql2018-1}{,} 
                TypeSQL~\citep{sql2018-2}{,} Coarse2Fine~\citep{sql2018-3}{,} 
                \citet{sql2018-4}{,} SyntaxSQLNet~\citep{sql2018-6}{,} 
                GNN~\citep{sql2019-2}{,} TREQS~\citep{sql-data-2019-1}, fill=Orange, fill opacity=0.2
                    [
                    SQLova~\citep{sql2019-1}{,} IRNet~\citep{sql2019-3}{,} 
                    \citet{sql2019-5}{,} RAT-SQL~\citep{sql2019-7}{,} 
                    Bertrand-DR~\citep{sql2020-1}{,} RYANSQL~\citep{sql2020-2}{,} 
                    TaBERT~\citep{sql2020-2.2}{,} 
                    Photon~\citep{sql2020-2.3}{,} HydraNet~\citep{sql2020-2.4}{,} 
                    GAZP~\citep{sql2020-2.5}{,} GraPPa~\citep{sql2020-2.7}{,} 
                    SmBoP~\citep{sql2020-3}{,} NQG-T5~\citep{sql2020-4}{,} 
                    StruG~\citep{sql2020-4.5}{,} SLSQL~\citep{sql2020-4.7}{,} 
                    GAP~\citep{sql2020-5}{,} \citet{sql2021-0.7}{,} 
                    GP~\citep{sql2021-0.5}{,} 
                    LGESQL~\citep{sql2021-1}{,} Picard~\citep{sql2021-2}{,} 
                    H-NeurSyn~\citep{sql2021-3}{,} 
                    UnifiedSKG~\citep{2022UnifiedSKG}{,} CodexDB~\citep{sql2022-1}{,} 
                    T5QL~\citep{sql2022-3}{,} TKK~\citep{sql2022-4}{,} 
                    Graphix-T5~\citep{sql2023-0.1}{,} 
                    RESDSQL~\citep{sql2023-0.2}{,} \citet{sql2023-0.3}{,} 
                    DIN-SQL~\citep{sql2023-0.4}{,} 
                    \citet{sql2023-0.5}{,} 
                    \citet{sql2023-1}{,} SQL-PaLM~\citep{sql2023-1.5}{,} 
                    \citet{sql2023-1.7}{,} DAIL-SQL~\citep{sql2023-2}
                    % SQL-Encoder~\citep{sql2024-1}
                    , fill=Orange, fill opacity=0.2, text width=15em
                    ]
                ]
            ]
        ]
    ]
    [Development:\\Code Editing, fill=Orange, fill opacity=0.2]
    [Development:\\Code Suggestion, fill=Orange, fill opacity=0.2]
    [Development:\\Code Explanation, fill=Orange, fill opacity=0.2]
    % [Development:\\Code Review, fill=Orange, fill opacity=0.2]
    [Testing \&\\ Analysis, fill=cyan, fill opacity=0.2]
    [Deployment \& Operations, fill=VioletRed, fill opacity=0.2]
    [LLM Evaluation, fill=Dandelion, fill opacity=0.4]
]
\end{forest}

%% file: trees/task2-development2.tex
\begin{forest}
for tree={
forked edges,
draw,
rounded corners,
grow=east,
anchor=base west,
anchor=center,
reversed=true,
l sep=0.2cm,
},
where level=0{font=\tiny}{},
where level=1{text width=4.9em,font=\tiny}{},
where level=2{text width=3.5em,font=\tiny}{},
where level=3{text width=6.0em,font=\tiny}{},
where level=4{text width=9.0em,font=\tiny}{},
where level=5{text width=12.0em,font=\tiny}{},
for tree={
  if level=0{l sep+=0.15cm}{},
  if level=1{l sep+=0.15cm}{},
  if level=2{l sep+=0.03cm}{},
},
[NLP in SE, fill=black, fill opacity=0.7, text=white
    [Requirement, fill=yellow, fill opacity=0.5]
    [Development:\\Code Search, fill=Orange, fill opacity=0.2]
    [Development:\\Code Editing, fill=Orange, fill opacity=0.2
        [, draw=none, no edge
            [\textbf{Non-neural methods}, draw=none, no edge, fill=black, fill opacity=0.5, text=white
                [\textbf{Non-Transformer neural methods}, draw=none, no edge, fill=black, fill opacity=0.5, text=white
                    [\textbf{Transformer-based methods}, draw=none, no edge, fill=black, fill opacity=0.5, text=white]
                ]
            ]
        ]
        [Code\\ Translation, fill=Orange, fill opacity=0.2
            [
            lpSMT~\citep{trans2013-1}{,} \citet{trans2014-1}{,} 
            mppSMT~\citep{trans2015-1}, fill=Orange, fill opacity=0.2
                [
                Tree2Tree~\citep{trans2018-1}{,} Grammar Tree2Tree~\citep{trans2018-2}, fill=Orange, fill opacity=0.2
                    [
                        TransCoder~\citep{2020Transcoder}{,} 
                        TransCoder-ST~\citep{2021Transcoder}{,} 
                        TransCoder-IR~\citep{2022Transcoder}{,} 
                        BabelTower~\citep{trans2022-1}{,} 
                        SDA-Trans~\citep{trans2023-1}{,} 
                        Self-Debugging~\citep{2023self-debug}{,} 
                        CoTran~\citep{trans2023-2}{,} 
                        MuST~\citep{trans-data-2022-2}{,} 
                        \citet{trans2023-3}{,} 
                        Explain-then-Translate~\citep{trans2023-4}{,} 
                        \citet{trans2024-1}, fill=Orange, fill opacity=0.2
                    ]
                ]
            ]
        ]
        [Code\\ Repair, fill=Orange, fill opacity=0.2
            [
            Prophet~\citep{fix2016-1}{,} TBar~\citep{fix2019-1.5}{,} 
            Refactory~\citep{fix-data-2019-1}{,} PyTER~\citep{fix-data-2022-1}, fill=Orange, fill opacity=0.2
                [
                sk\_p~\citep{fix2016-3}{,} 
                DeepFix~\citep{fix2017-1}{,} SSC~\citep{fix2017-2}{,} 
                SynFix~\citep{fix2018-1}{,} 
                Codit~\citep{fix2018-2}{,} \citet{fix2018-3}{,} 
                Sequencer~\citep{fix2018-4}{,} \citet{fix2019-1}{,} 
                \citet{fix2019-2}{,} DrRepair~\citep{fix2020-1}{,} 
                CoCoNuT~\citep{fix2020-2}{,} DLFix~\citep{fix2020-3}{,} 
                Review4Repair~\citep{fix2020-4}{,} 
                DEAR~\citep{fix2022-0.5}, fill=Orange, fill opacity=0.2
                    [
                    CURE~\citep{fix2021-1}{,} 
                    DeepDebug~\citep{fix2021-2}{,} BIFI~\citep{fix2021-3}{,} 
                    Recoder~\citep{fix2021-4}{,} TFix~\citep{fix2021-4.5}{,} 
                    Modit~\citep{fix2021-5}{,} \citet{fix2022-1}{,} 
                    AlphaRepair~\citep{fix2022-2}{,} RING~\citep{fix2022-2.5}{,} 
                    \citet{fix2022-3}{,} VulRepair~\citep{fix2022-3.5}{,} 
                    CodeT5-DLR~\citep{fix2022-4}{,} 
                    Conversational APR~\citep{fix2023-0.5}{,} \citet{fix2023-0.7}{,} 
                    \citet{fix2023-1}{,} \citet{fix2023-2}{,} 
                    TypeFix~\citep{type2023-3.5}{,} 
                    Maniple~\citep{2024Maniple}, fill=Orange, fill opacity=0.2
                    ]
                ]
            ]
        ]
    ]
    [Development:\\Code Suggestion, fill=Orange, fill opacity=0.2
        [Type\\ Prediction, fill=Orange, fill opacity=0.2
            [
            JSNice~\citep{id2015-1}{,} TypeDevil~\citep{type2015-1}{,} 
            \citet{type2016-1}{,} Pigeon~\citep{id2018-1}{,} 
            Typpete~\citep{type2018-1}, fill=Orange, fill opacity=0.2
                [
                DeepTyper~\citep{type2018-2}{,} \citet{type2019-1}{,} 
                NL2Type~\citep{type2019-2}{,} DLTPy~\citep{type2019-3}{,} 
                TypeWriter~\citep{type2019-4}{,} OptTyper~\citep{type2020-1}{,} 
                Typilus~\citep{type2020-2}{,} LambdaNet~\citep{type2020-3}{,} 
                Type4Py~\citep{type2021-1}{,} HiTyper~\citep{type2021-2}{,} 
                PYInfer~\citep{type2021-3}, fill=Orange, fill opacity=0.2
                    [
                    TypeBert~\citep{type2021-4}{,} TypeWeaver~\citep{type2023-1}{,} 
                    TypeT5~\citep{type2023-2}{,} OpenTau~\citep{type2023-3}{,} 
                    TypeGen~\citep{type2023-4}{,} 
                    \citet{type2023-5}
                    , fill=Orange, fill opacity=0.2
                    ]
                ]
            ]
        ]
        [Identifier Prediction, fill=Orange, fill opacity=0.2
            [
            JSNice~\citep{id2015-1}{,} Pigeon~\citep{id2018-1}{,} 
            HeMa~\citep{id2019-3}, fill=Orange, fill opacity=0.2
                [
                \citet{id2015-2}{,} \citet{id2016-1}{,} 
                GGNN~\citep{id2017-1}{,} Code2Vec~\citep{id2018-2}{,} 
                Code2Seq~\citep{id2018-3}{,} \citet{id2018-4}{,} 
                HIER~\citep{id2019-1}{,} \citet{id2019-2}{,} 
                MNire~\citep{id2020-1}{,} DeepName~\citep{id2021-1}{,} 
                DMACOS~\citep{id2021-2}{,} NamPat~\citep{id2022-3}, fill=Orange, fill opacity=0.2
                    [
                    GTNM~\citep{id2022-2}{,} GTrans~\citep{sum2022-3}{,} 
                    Mario~\citep{id2023-1}, fill=Orange, fill opacity=0.2
                    ]
                ]
            ]
        ]
        [Cloze Test, fill=Orange, fill opacity=0.2
            [, fill=Orange, fill opacity=0.2
                [, fill=Orange, fill opacity=0.2
                    [
                    CodeBERT~\citep{2020CodeBERT}{,} 
                    \citet{2021CodeXGLUE}{,} 
                    \citet{2021CodeNet}{,} 
                    JavaBERT~\citep{cloze2021-1}, fill=Orange, fill opacity=0.2
                    ]
                ]
            ]
        ]
    ]
    [Development:\\Code Explanation, fill=Orange, fill opacity=0.2
        [Comment\\Generation, fill=Orange, fill opacity=0.2
            [
            AutoComment~\citep{comment2013-1}{,} 
            CloCom~\citep{sum2015-1}
            , fill=Orange, fill opacity=0.2
                [
                DeepCom~\citep{sum2018-1}{,} 
                APIContext2Com~\citep{comment2023-2}
                , fill=Orange, fill opacity=0.2
                    [
                    LAMNER~\citep{comment2022-1}{,} 
                    GTrans~\citep{sum2022-3}{,} 
                    DOME~\citep{comment2023-1}{,} 
                    \citet{comment2023-3}{,} 
                    MESIA~\citep{comment2024-1}
                    , fill=Orange, fill opacity=0.2
                    ]
                ]
            ]
        ]
        [Code Summarization, fill=Orange, fill opacity=0.2
            [
            , fill=Orange, fill opacity=0.2
                [
                CODE-NN~\citep{sum2016-1}{,} 
                TL-CodeSum~\citep{sum2018-2}{,} Code2Seq~\citep{id2018-3}{,} 
                \citet{id2018-4}{,} \citet{sum2018-3}{,} 
                AST-AttendGRU~\citep{sum2019-0.5}{,} \citet{sum2019-1}{,} 
                \citet{sum2020-1}{,} DMACOS~\citep{id2021-2}{,} 
                \citet{sum2021-0.5}{,} 
                CoCoSUM~\citep{sum2021-1}{,} 
                MLCS~\citep{sum2023-1}, fill=Orange, fill opacity=0.2
                    [
                    TranS$^3$~\citep{2020Trans3}{,} \citet{sum2020-2}{,} 
                    Corder~\citep{retrieval2020-0.5}{,} SiT~\citep{sum2020-3}{,} 
                    SG-Trans~\citep{sum2021-0.7}{,} 
                    Codex-D~\citep{2021Codex}{,} 
                    M2TS~\citep{sum2022-1}{,} AST-Trans~\citep{sum2022-2}{,} 
                    CoSS~\citep{sum2023-2}{,} 
                    \citet{sum2023-3}{,} \citet{defect2023-1}{,} 
                    use-seq~\citep{sum2023-4}{,} 
                    \citet{sum2023-5}{,} 
                    CSA-Trans~\citep{2024CSA-Trans}{,} 
                    \citet{sum2024-1}
                    , fill=Orange, fill opacity=0.2
                    ]
                ]
            ]
        ]
        [Code Documentation, fill=Orange, fill opacity=0.2
            [
            , fill=Orange, fill opacity=0.2
                [
                , fill=Orange, fill opacity=0.2
                    [
                    \citet{doc2022-1}{,} 
                    CodeExp~\citep{doc2022-2}{,} 
                    HotGPT~\citep{doc2023-1}{,} 
                    \citet{doc2023-2}
                    , fill=Orange, fill opacity=0.2
                    ]
                ]
            ]
        ]
        [Document Translation, fill=Orange, fill opacity=0.2
            [, fill=Orange, fill opacity=0.2
                [, fill=Orange, fill opacity=0.2
                    [\citet{2021CodeXGLUE}, fill=Orange, fill opacity=0.2]
                ]
            ]
        ]
    ]
    % [Development:\\Code Review, fill=Orange, fill opacity=0.2]
    [Testing \&\\ Analysis, fill=cyan, fill opacity=0.2]
    [Deployment \& Operations, fill=VioletRed, fill opacity=0.2]
    [LLM Evaluation, fill=Dandelion, fill opacity=0.4]
]
\end{forest}

%% file: trees/task3-testing.tex
\begin{forest}
for tree={
forked edges,
draw,
rounded corners,
grow=east,
anchor=base west,
anchor=center,
reversed=true,
l sep=0.2cm,
},
where level=0{font=\tiny}{},
where level=1{text width=4.9em,font=\tiny}{},
where level=2{text width=3.5em,font=\tiny}{},
where level=3{text width=10.5em,font=\tiny}{},
where level=4{text width=5.7em,font=\tiny}{},
where level=5{text width=10.8em,font=\tiny}{},
for tree={
  if level=0{l sep+=0.15cm}{},
  if level=1{l sep+=0.15cm}{},
  if level=2{l sep+=0.03cm}{},
},
[NLP in SE, fill=black, fill opacity=0.7, text=white
    [Requirement, fill=yellow, fill opacity=0.5]
    [Development, fill=Orange, fill opacity=0.2]
    [Testing, fill=cyan, fill opacity=0.2
        [, draw=none, no edge
            [\textbf{Non-neural methods}, draw=none, no edge, fill=black, fill opacity=0.5, text=white
                [\textbf{Non-Transformer neural methods}, draw=none, no edge, fill=black, fill opacity=0.5, text=white
                    [\textbf{Transformer-based methods}, draw=none, no edge, fill=black, fill opacity=0.5, text=white]
                ]
            ]
        ]
        [Assertion Generation, fill=cyan, fill opacity=0.2
            [
            MeMo~\citep{assert2021-2}, fill=cyan, fill opacity=0.2
                [
                Atlas~\citep{assert2020-1}, fill=cyan, fill opacity=0.2
                    [
                    \citet{assert2020-2}{,} TOGA~\citep{assert2021-1}{,} 
                    FSLM~\citep{unit2022-3}{,} \citet{defect2023-1}, fill=cyan, fill opacity=0.2
                    ]
                ]
            ]
        ]
        [Mutant\\ Generation, fill=cyan, fill opacity=0.2
            [
            Major~\citep{mutant2014-1}{,} LAVA~\citep{mutant2016-1}{,} 
            PIT~\citep{mutant2016-2}{,} \citet{mutant2016-3}{,} 
            EvilCoder~\citep{mutant2016-4}{,} wild-caught mutants~\citep{mutant2017-1}{,} 
            Apocalypse~\citep{mutant2018-1}{,} Bug-Injector~\citep{mutant2019-1}{,} 
            IBIR~\citep{mutant2020-2}{,} PBMT~\citep{mutant2023-2}, fill=cyan, fill opacity=0.2
                [
                \citet{mutant2018-2}{,} DeepMutation~\citep{mutant2020-1}{,} 
                SemSeed~\citep{mutant2021-1}, fill=cyan, fill opacity=0.2
                    [
                    $\mu$BERT~\citep{mutant2022-1,mutant2023-1}{,} 
                    FSLM~\citep{unit2022-3}, fill=cyan, fill opacity=0.2
                    ]
                ]
            ]
        ]
        [Unit Test Generation, fill=cyan, fill opacity=0.2
            [
            EvoSuite~\citep{unit2011-1}{,} EvoSuiteR~\citep{unit2015-1}{,} 
            DynaMOSA~\citep{unit2017-1}{,} LambdaTester~\citep{unit2018-1}{,} 
            TSE~\citep{unit2022-1}{,} Nessie~\citep{unit2022-2}, fill=cyan, fill opacity=0.2
                [, fill=cyan, fill opacity=0.2
                    [
                    AthenaTest~\citep{unit2020-1}{,} FSLM~\citep{unit2022-3}{,} 
                    TestPilot~\citep{unit2023-1}{,} A3Test~\citep{unit2023-2}{,} 
                    TeCo~\citep{unit2023-3}{,} CodaMosa~\citep{unit2023-4}{,} 
                    ChatTester~\citep{unit2023-5}{,} ChatUniTest~\citep{unit2023-6}{,} 
                    \citet{unit2023-7}{,} 
                    PBT-GPT~\citep{2023PBT-GPT}{,} 
                    \citet{unit2023-8}{,} 
                    MuTAP~\citep{2023MuTAP}{,} 
                    \citet{unit2023-9}{,} 
                    RLSQM~\citep{2023RLSQM}{,} 
                    \citet{unit2023-10}{,} 
                    CoverUp~\citep{2024CoverUp}{,} 
                    TELPA~\citep{2024TELPA}
                    , fill=cyan, fill opacity=0.2
                    ]
                ]
            ]
        ]
        [Fuzzing, fill=cyan, fill opacity=0.2
            [
            SymFuzz~\citep{fuzz2015-1}{,} AFLFast~\citep{fuzz2016-1}{,} 
            FairFuzz~\citep{fuzz2017-1}{,} AFLGo~\citep{fuzz2017-2}{,} 
            Angora~\citep{fuzz2018-1}{,} TensorFuzz~\citep{fuzz2018-3}{,} 
            Audee~\citep{fuzz2020-2}{,} LEMON~\citep{fuzz2020-3}{,} 
            DocTer~\citep{fuzz2021-1}{,} FreeFuzz~\citep{fuzz2022-1}{,} 
            SpecFuzzer~\citep{fuzz2022-2}{,} Muffin~\citep{fuzz2022-3}{,} 
            DeepREL~\citep{fuzz2022-5}{,} NNSmith~\citep{fuzz2022-6}{,} 
            $\nabla$Fuzz~\citep{fuzz2023-1}, fill=cyan, fill opacity=0.2
                [
                NEUZZ~\citep{fuzz2018-2}{,} MTFuzz~\citep{fuzz2020-1}{,}
                PreFuzz~\citep{fuzz2022-4}, fill=cyan, fill opacity=0.2
                    [
                    TitanFuzz~\citep{fuzz2022-7}{,} 
                    Fuzz4All~\citep{2023Fuzz4All}{,} 
                    WhiteFox~\citep{fuzz2023-2}, fill=cyan, fill opacity=0.2
                    ]
                ]
            ]
        ]
    ]
    [Analysis, fill=cyan, fill opacity=0.2        
        [Defect\\ Detection, fill=cyan, fill opacity=0.2
            [
            \citet{defect2015-1}{,} Bugram~\citep{defect2016-2}{,} 
            NAR-Miner~\citep{defect2018-4}{,} ~\citep{defect2021-1}
            , fill=cyan, fill opacity=0.2, text width=6em
                [
                \citet{defect2016-1}{,} VulDeePecker~\citep{defect2018-1}{,} 
                \citet{defect-data-2018-1}{,} 
                DeepBugs~\citep{defect2018-2}{,} \citet{defect2018-3}{,} 
                SySeVR~\citep{defect2018-3.5}{,} 
                Devign~\citep{2019Devign}{,} \citet{defect2019-1}{,} 
                \citet{defect-data-2019-2}{,} 
                VulDeeLocator~\citep{defect2020-0.5}{,} $\mu$VulDeePecker~\citep{defect-data-2020-3}{,} 
                ReVeal~\citep{defect2020-1}{,} BugLab~\citep{defect2021-0.3}{,} 
                IVDetect~\citep{defect2021-0.5}{,} 
                ReGVD~\citep{defect2021-2}
                , fill=cyan, fill opacity=0.2, text width=10.2em
                    [
                    GREAT~\citep{fix2019-3}{,} VulBERTa~\citep{defect2022-1}{,} 
                    LineVul~\citep{defect2022-2}{,} 
                    DeepDevVuln~\citep{defect2023-0.5}{,} \citet{defect2023-1}{,} 
                    \citep{defect2023-2}{,} CausalVul~\citep{defect2023-3}{,} 
                    \citet{defect2023-4}{,} 
                    \citet{defect2023-5}{,} 
                    LLM4Vuln~\citep{2024LLM4Vuln}{,} 
                    MuCoLD~\citep{2024MuCoLD}{,} 
                    \citet{defect2024-1}
                    , fill=cyan, fill opacity=0.2
                    ]
                ]
            ]
        ]
        [Clone\\ Detection, fill=cyan, fill opacity=0.2
            [
            Deckard~\citep{clone2007-1}{,} SourcererCC~\citep{clone2015-1}{,} 
            CCAligner~\citep{clone2018-1}{,} LVMapper~\citep{clone2019-3}{,} 
            SAGA~\citep{clone2020-0.5}{,} 
            NIL~\citep{clone2021-1}
            , fill=cyan, fill opacity=0.2, text width=6em
                [
                \citet{clone2016-1}{,} CDLH~\citep{clone2017-1}{,} 
                Oreo~\citep{clone2018-2}{,} DeepSim~\citep{clone2018-3}{,} 
                ASTNN~\citep{clone2019-1}{,} 
                TBCCD~\citep{clone2019-2}{,} \citet{clone2019-2.5}
                CLCDSA~\citep{clone-data-2019-1}{,} 
                FA-AST~\citep{clone2020-1}{,} 
                \citet{clone2020-2}{,} \citet{clone2022-1}
                , fill=cyan, fill opacity=0.2, text width=10.2em
                    [
                    \citet{clone2022-0.5}{,} 
                    SSCD~\citep{clone2022-2}{,} 
                    CCT-LM~\citep{clone2023-0.5}{,} 
                    \citet{clone2023-1}{,} 
                    \citet{defect2023-1}{,} 
                    ZC$^3$~\citep{clone2023-2}
                    , fill=cyan, fill opacity=0.2
                    ]
                ]
            ]
        ]
        [Malware\\ Detection, fill=cyan, fill opacity=0.2
            [
            , fill=cyan, fill opacity=0.2, text width=6em
                [
                , fill=cyan, fill opacity=0.2, text width=10.2em
                    [
                    Malbert~\citep{2021Malbert}{,} 
                    SHERLOCK~\citep{2022SHERLOCK}{,} 
                    MSDT~\citep{2022MSDT}{,} 
                    MalBERTv2~\citep{2023MalBERTv2}{,} 
                    SocketAI Scanner~\citep{2024SocketAIScanner}
                    , fill=cyan, fill opacity=0.2
                    ]
                ]
            ]
        ]
        [Code/Author Classification, fill=cyan, fill opacity=0.2
            [
            SCAP~\citep{author2011-1}{,} \citet{author2022-1}
            , fill=cyan, fill opacity=0.2, text width=6em
                [\citet{author2013-1}{,} 
                TBCNN~\citep{clf2014-1}{,} inst2vec~\citep{clf2018-1}{,} 
                DL-CAIS~\citep{author2018-1}{,} 
                ASTNN~\citep{clone2019-1}{,} InferCode~\citep{2020InferCode}
                , fill=cyan, fill opacity=0.2, text width=10.2em
                    [
                    \citet{2021CodeNet}{,} \citet{retrieval2021-7}, fill=cyan, fill opacity=0.2
                    ]
                ]
            ]
        ]
    ]
    [Deployment \& Operations, fill=VioletRed, fill opacity=0.2]
    [LLM Evaluation, fill=Dandelion, fill opacity=0.4]
]
\end{forest}

%% file: trees/task4-devops.tex
\begin{forest}
for tree={
forked edges,
draw,
rounded corners,
grow=east,
anchor=base west,
anchor=center,
reversed=true,
l sep=0.2cm,
},
where level=0{font=\tiny}{},
where level=1{text width=4.9em,font=\tiny}{},
where level=2{text width=4.1em,font=\tiny}{},
where level=3{text width=6em,font=\tiny}{},
where level=4{text width=10em,font=\tiny}{},
where level=5{text width=11em,font=\tiny}{},
for tree={
  if level=0{l sep+=0.15cm}{},
  if level=1{l sep+=0.15cm}{},
  if level=2{l sep+=0.03cm}{},
},
[NLP in SE, fill=black, fill opacity=0.7, text=white
    [Requirement, fill=yellow, fill opacity=0.5]
    [Development, fill=Orange, fill opacity=0.2]
    [Testing \&\\ Analysis, fill=cyan, fill opacity=0.2]
    [Deployment, fill=VioletRed, fill opacity=0.2
        [, draw=none, no edge
            [\textbf{Non-neural methods}, draw=none, no edge, fill=black, fill opacity=0.5, text=white
                [\textbf{Non-Transformer neural methods}, draw=none, no edge, fill=black, fill opacity=0.5, text=white
                    [\textbf{Transformer-based methods}, draw=none, no edge, fill=black, fill opacity=0.5, text=white]
                ]
            ]
        ]
        [Compiler\\Optimization, fill=VioletRed, fill opacity=0.2
            [, fill=VioletRed, fill opacity=0.2
                [
                DeepTune~\citep{comp-opt2017-1}
                , fill=VioletRed, fill opacity=0.2
                    [
                    \citet{comp-opt2021-1}{,} 
                    \citet{comp-opt2023-1}{,} 
                    Priority Sampling~\citep{comp-opt2024-1}{,} 
                    \citet{comp-opt2024-2}
                    , fill=VioletRed, fill opacity=0.2
                    ]
                ]
            ]
        ]
        [Decompilation, fill=VioletRed, fill opacity=0.2
            [
            , fill=VioletRed, fill opacity=0.2
                [
                Neutron~\citep{2021Neutron}{,} 
                NeurDP~\citep{2023NeurDP}
                , fill=VioletRed, fill opacity=0.2
                    [
                    BTC~\citep{2022BTC}{,} 
                    SLaDe~\citep{2023SLaDe}{,} 
                    DecGPT~\citep{2023DecGPT}{,} 
                    Nova+~\citep{2023Nova+}{,} 
                    LLM4Decompile~\citep{LLM4Decompile}
                    , fill=VioletRed, fill opacity=0.2
                    ]
                ]
            ]
        ]
        [Deobfuscation, fill=VioletRed, fill opacity=0.2
            [
            DeGuard~\citep{ob2016-1}{,} Autonym~\citep{ob2017-1}{,} 
            Debin~\citep{ob2018-1}{,} JSNeat~\citep{ob2019-1}, fill=VioletRed, fill opacity=0.2
                [
                DIRE~\citep{ob2019-2}{,} \citet{ob2019-3}, fill=VioletRed, fill opacity=0.2
                    [
                    VarBERT~\citep{ob2021-1}{,} DIRECT~\citep{ob2021-2}{,} 
                    LmPa~\citep{ob2023-2}, fill=VioletRed, fill opacity=0.2
                    ]
                ]
            ]
        ]
        [Log\\Analysis, fill=VioletRed, fill opacity=0.2
            [Spell~\citep{log2016-1}{,} Drain~\citep{log2017-1}{,} 
            Logram~\citep{log2020-1}{,} SPINE~\citep{log2022-2}
            , fill=VioletRed, fill opacity=0.2
                [
                DeepLog~\citep{log2017-2}{,} 
                \citet{log2018-1}{,} 
                LogAnomaly~\citep{log2019-1}{,} 
                LogRobust~\citep{log2019-2}{,} 
                LogDTL~\citep{log2021-0.5}{,} 
                UniParser~\citep{log2022-1}
                , fill=VioletRed, fill opacity=0.2
                    [
                    NuLog~\citep{log2020-1.5}{,} 
                    Logsy~\citep{log2020-2}{,} 
                    LogBERT~\citep{log2021-0.1}{,} 
                    NeuralLog~\citep{log2021-1}{,} LogStamp~\citep{log2022-3}{,} LogPPT~\citep{log2023-1}{,} 
                    \citet{log2023-2}{,} Semlog~\citep{log2023-3}{,} 
                    LogPrompt~\citep{log2023-4}{,} LogGPT~\citep{log2023-5}{,} 
                    \citet{log2023-5.5}{,} LILAC~\citep{log2023-6}{,}
                    LLMParser~\citep{log2024-1}
                    , fill=VioletRed, fill opacity=0.2
                    ]
                ]
            ]
        ]
    ]
    [Operations, fill=VioletRed, fill opacity=0.2
        [Code\\ Review, fill=VioletRed, fill opacity=0.2
            [
            Review Bot~\citep{review2013-1}{,} Tricorder~\citep{review2015-1}{,} 
            ClusterChanges~\citep{review2015-2}{,} CommentFinder~\citep{review2022-4}, fill=VioletRed, fill opacity=0.2
                [
                DCR~\citep{review2018-1}{,} DeepReview~\citep{review2019-1}{,} 
                DACE~\citep{review2019-2}{,} CORE~\citep{review2019-3}, fill=VioletRed, fill opacity=0.2
                    [
                    \citet{review2021-1}{,} \citet{review2021-2}{,} 
                    \citet{review2022-1}{,} CodeReviewer~\citep{review2022-2}{,} 
                    AUGER~\citep{review2022-3}{,} crBERT~\citep{review2023-1}{,} 
                    LLaMA-Reviewer~\citep{review2023-2}, fill=VioletRed, fill opacity=0.2
                    ]
                ]
            ]
        ]
        [Workflow\\Automation, fill=VioletRed, fill opacity=0.2
            [
            , fill=VioletRed, fill opacity=0.2
                [
                , fill=VioletRed, fill opacity=0.2
                    [
                    \citet{workflow2023-1}{,} 
                    \citet{workflow2024-1}
                    , fill=VioletRed, fill opacity=0.2
                    ]
                ]
            ]
        ]
        [Commit Message Generation, fill=VioletRed, fill opacity=0.2
            [
            ChangeScribe~\citep{commit2014-1,commit2015-1}{,} \citet{commit-data-2017-1}{,} 
            NNGen~\citep{commit2018-1}{,} ChangeDoc~\citep{commit2020-3}
            , fill=VioletRed, fill opacity=0.2
                [
                \citet{commit2017-1}{,} \citet{commit2017-2}{,} 
                \citet{commit2018-2}{,} PtrGNCMsg~\citep{commit2019-1}{,} 
                CoDiSum~\citep{commit2019-2}{,} ATOM~\citep{commit2019-3}{,} 
                CC2Vec~\citep{commit2020-1}{,} CoRec~\citep{commit-data-2021-1}
                , fill=VioletRed, fill opacity=0.2
                    [
                    CoreGen~\citep{commit2020-2}{,} CommitBERT~\citep{commit2021-1}{,} 
                    RACE~\citep{commit2022-1}{,} FIRA~\citep{commit2022-2}{,} 
                    COME~\citep{commit2023-1}{,} ExGroFi~\citep{commit-data-2023-1}
                    , fill=VioletRed, fill opacity=0.2
                    ]
                ]
            ]
        ]
    ]
    [LLM Evaluation, fill=Dandelion, fill opacity=0.4]
]
\end{forest}

%% file: trees/task5-llm.tex
\begin{forest}
for tree={
forked edges,
draw,
rounded corners,
grow=east,
anchor=base west,
anchor=center,
reversed=true,
l sep=0.2cm,
},
where level=0{font=\tiny}{},
where level=1{text width=4.9em,font=\tiny}{},
where level=2{text width=4.1em,font=\tiny}{},
where level=3{text width=26em,font=\tiny}{},
where level=4{text width=20.7em,font=\tiny}{},
for tree={
  if level=0{l sep+=0.15cm}{},
  if level=1{l sep+=0.15cm}{},
  if level=2{l sep+=0.03cm}{},
},
[NLP in SE, fill=black, fill opacity=0.7, text=white
    [Requirement, fill=yellow, fill opacity=0.5]
    [Development, fill=Orange, fill opacity=0.2]
    [Testing \&\\ Analysis, fill=cyan, fill opacity=0.2]
    [Deployment \& Operations, fill=VioletRed, fill opacity=0.2
    ]
    [LLM Evaluation, fill=Dandelion, fill opacity=0.4
        [Code\\ Reasoning, fill=Dandelion, fill opacity=0.4
            [
            MMLU~\citep{2020MMLU}{,} 
            CodeQA~\citep{2021CodeQA}{,} 
            CS1QA~\citep{2022CS1QA}{,} 
            C-Eval~\citep{2023C-Eval}{,} 
            CMMLU~\citep{2023CMMLU}{,} 
            CodeApex~\citep{2023CodeApex}{,}
            CRUXEval~\citep{2024CRUXEval}{,}
            PythonIO~\citep{2024PythonIO}
            , fill=Dandelion, fill opacity=0.4
            ]
        ]
        [Coding for\\ Reasoning, fill=Dandelion, fill opacity=0.4
            [
            MathQA-Python~\citep{2021MBPP}{,} 
            \citet{math2021-0.5}{,} 
            \citet{math2021-1}{,} 
            PAL~\citep{2022PAL}{,} 
            PoT~\citep{2022PoT}{,} 
            \citet{math2023-1}{,} 
            GSM8K-Python~\citep{2022PaLM,2023CodeT5+}{,} 
            CoC~\citep{2023CoC}{,} 
            Think-and-Execute~\citep{2024Think-and-Execute}
            , fill=Dandelion, fill opacity=0.4
            ]
        ]
        [AIGC\\Analysis, fill=Dandelion, fill opacity=0.4
            [Vulnerabilities
            , fill=Dandelion, fill opacity=0.4, text width=4em
                [
                \citet{ana-vul2021-1}{,} 
                \citet{analysis2022-1}{,} 
                \citet{analysis2022-3}{,} 
                \citet{analysis2022-4}{,} 
                \citet{analysis2024-2}
                , fill=Dandelion, fill opacity=0.4
                ]
            ]
            [Correctness
            , fill=Dandelion, fill opacity=0.4, text width=4em
                [
                \citet{analysis2022-2}{,} 
                \citet{analysis2023-1}{,} 
                \citet{analysis2023-2}{,} 
                \citet{analysis2024-1}
                , fill=Dandelion, fill opacity=0.4
                ]
            ]
            [AIGC\\Detection
            , fill=Dandelion, fill opacity=0.4, text width=4em
                [
                DetectGPT4Code~\citep{2023detect}{,} 
                CodeIP~\citep{2024CodeIP}
                , fill=Dandelion, fill opacity=0.4
                ]
            ]
            [Others
            , fill=Dandelion, fill opacity=0.4, text width=4em
                [
                Understandability~\citep{analysis2023-3}{,} 
                Syntactic Robustness~\citep{analysis2024-4}{,} 
                Hallucinations~\citep{analysis2024-3}{,} 
                Efficiency~\citep{analysis2024-5}{,} 
                Data Leak~\citep{ana-leak2024-1}
                , fill=Dandelion, fill opacity=0.4
                ]
            ]
        ]
    ]
]
\end{forest}

%% file: sec_background.tex
\section{Language Modeling Preliminaries}\label{sec:background}
As code is ultimately a subset of natural languages, language models have been extensively used to tackle the tasks listed in Section~\ref{sec:evaluation}. Before diving into the language models themselves, we first briefly review the preliminaries of Transformer-based language modeling in this section following the common choices of training objectives, and also some implementation designs.

\subsection{Causal Language Modeling}
Unidirectional language models (also known as causal language models\footnote{The training objective of such language models is Causal Language Modeling (CLM), but also referred to as Next Token Prediction.}) factor the probability of a sentence into the product of each token's conditional probability with the chain rule. A piece of input text $\mathbf{x} = [x_1,x_2,\cdots,x_n]$ consisting of $n$ tokens is modeled as
\begin{equation}\label{eq:clm}
    P(\mathbf{x}) = \prod_{i=1}^np_\theta(x_i|\mathbf{x}_{1:i-1}),
\end{equation}
\noindent
where $\mathbf{x}_{1:i-1}$ is a shorthand for tokens before $x_i$ in the input, and $\theta$ is the parameters of the model. With Transformer decoders such as GPT~\citep{2018GPT,2019GPT2,2020GPT3} and LLaMA~\citep{2023LLaMA,2023LLaMA2}, the conditional probability in \eqref{eq:clm} is modeled by adding an attention mask to the attention matrix of each Transformer block, ensuring that $x_i$ can only attend to previous tokens. During training, the cross entropy loss on all tokens in the input is calculated in parallel, while at inference time each new token is generated autoregressively. For further details about the Transformer architecture we refer to \citet{2017Transformer}.

\subsection{Masked Language Modeling}
Unlike causal language models, bidirectional language models are trained to acquire a better contextual representation of text rather than generating text autoregressively. In the vanilla Transformer, the encoder part is allowed to attend to a token's left as well as right context for this purpose. BERT~\citep{2018BERT} takes one step further and pretrains only a Transformer encoder. A set $\mathcal{M}$ of randomly chosen tokens in the input are replaced by a special token \texttt{[MASK]} to obtain a noisy input $\mathbf{\hat{x}}$, for example $[\texttt{[CLS]}, x_1, \texttt{[MASK]}, x_3, \texttt{[MASK]}, x_5, \texttt{[EOS]}]$\footnote{Both \texttt{[CLS]} and \texttt{[EOS]} are artificial tokens added to the input text. \texttt{[CLS]} is added at the beginning and its representation is used for sentence classification, while \texttt{[EOS]} indicates end of sentence. The original BERT also uses another special token \texttt{[SEP]}, which is not in common use in LLMs, and we refer to \citet{2018BERT} for details.}, and the model is trained to recover the original tokens by maximizing
\begin{equation}\label{eq:mlm}
    \prod_{m\in\mathcal{M}} p_\theta(m|\mathbf{\hat{x}}).
\end{equation}

While this objective requires the model to have a deep understanding of the input text to reconstruct it, it suffers from low training efficiency, since only a small set of tokens (usually 15\%) are masked (and thus ``trained on''). To address this issue, \citet{2020ELECTRA} propose ELECTRA, which is trained to discriminate whether or not each token in the input has been replaced by a BERT-like model instead, thereby computing loss on all input tokens.

\subsection{Denoising Objectives}
GPT-style causal LM and BERT-style bidirectional LM each has its own strengths and weaknesses. While GPT can be used for autoregressive generation, it lacks a bidirectional representation of input text, and is thus unsuitable for sequence-to-sequence (seq2seq) generation tasks such as translation and summarization. BERT, on the other hand, can produce bidirectional representations, but is pretrained only for mask filling, not generation.

The vanilla Transformer encoder-decoder architecture combines the respective advantages of GPT and BERT. T5~\citep{2019T5} is such a model pretrained with \emph{span corruption}, which can be regarded as a variant of MLM. During pretraining, spans of text in the input are replaced with sentinel tokens, which play the same role as \texttt{[MASK]} in BERT. The noisy input is first processed by the encoder with bidirectional attention, and the masked spans are then generated autoregressively by the decoder. Formally, if $k$ spans are sampled for corruption in input $\mathbf{x}$, the noisy input $\mathbf{\hat{x}}$ is then constructed by replacing each span with a special token \texttt{<extra\_id\_i>}, for $i=1,2,\cdots,k$, and the target $\mathbf y$ is constructed by concatenating all spans prepended with corresponding sentinels: [\texttt{<extra\_id\_1>}, span$_1$, $\cdots$, \texttt{<extra\_id\_k>}, span$_k$]. The model is then trained with a standard seq2seq objective, by maximizing
\begin{equation}\label{eq:seq2seq}
    p_\theta(\mathbf y|\mathbf{\hat{x}}) = \prod_{i=1}^{n_y}p_\theta(y_i|\mathbf{\hat{x}}, \mathbf y_{1:i-1}).
\end{equation}

\citet{2021T5-LM} show that models pretrained with such objectives can be adapted for autoregressive language modeling with extra pretraining using the prefix language modeling objective, i.e. splitting the text into two parts, processing the first part with encoder and generating the second part with decoder.

\citet{2022UL2} argue that span corruption is also closely related to CLM, since one can mask out the whole input text as a single span and train the decoder to generate it autoregressively. Inspired by this relation, they propose UL2, which is the combination of many span corruption objectives that differ in corruption rate and span length. Applying it to both encoder-decoder models and decoder-only models, they find that encoder-decoder models perform better under the same computation budget constraint. Other researches have also found that such encoder-decoder models generally perform better than causal decoder-only models~\citep{20220-shot,2022AlexaTM}.

\subsection{Auxiliary Objectives}
Language modeling objectives, such as previously discussed CLM and MLM, mainly train the model to capture token-level information and are ineffective at modeling document structures. Thus, auxiliary objectives are often added to help the models learn such global information. BERT is pretrained with next sentence prediction (NSP) along with MLM, which is formulated as a binary classification task to predict whether two segments in the input are neighboring in the original corpus. \citet{2019ALBERT} propose a more challenging sentence-order prediction (SOP) task, where the negative samples are constructed by swapping the order of two neighboring sentences instead of sampling a random sentence from other documents.

Relatedly, \citet{2019T5} mix supervised downstream samples such as GLUE~\citep{2018GLUE} into T5's pretraining dataset to conduct multi-task pretraining. However, it is worth noting that since they unify all tasks into a text-to-text format, the training objective is the same for their self-supervised pretraining and supervised downstream tasks, i.e. conditional generation of the target sequence.

\begin{figure}
    \centering
    \includegraphics[width=1\textwidth]{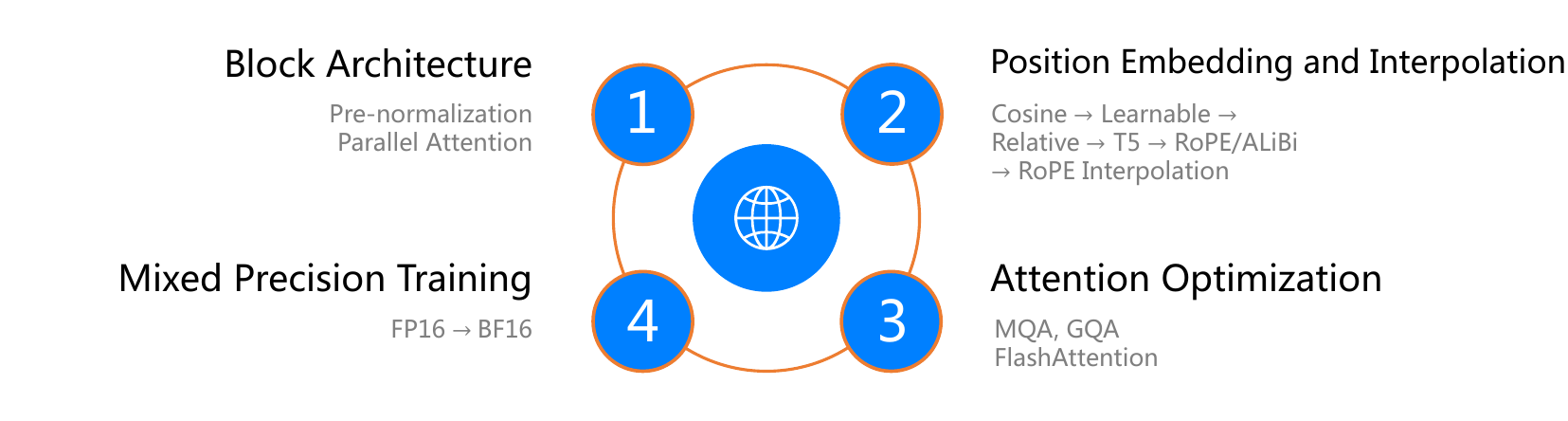}
    \caption{Major implementation changes in LLM over the past few years.}
    \label{fig:background}
\end{figure}

\subsection{Implementation Design}\label{sec:background-design}
While most researches on pretraining language models have focused on designing novel training objectives, low-level implementation of the Transformer architecture itself is also being continuously improved over the years in pursuit of stability, performance, and efficiency, as shown in Figure~\ref{fig:background}. 

The original Transformer block proposed by \citet{2017Transformer} is formulated as
\begin{align}
    h &= \textrm{LN}(\textrm{Attention}(x) + x),\\
    y &= \textrm{LN}(\textrm{FFN}(h) + h),
\end{align}
where $x$ is the layer's input; $y$ is the layer's output; ``Attention'' is the self-attention sublayer; ``FFN'' is the feed-forward sublayer, and ``LN'' is layer normalization~\citep{2016LayerNorm}.

GPT-2~\citep{2019GPT2} moves layer normalization to the input of each Transformer sub-block to stabilize training:
\begin{align}
    h &= \textrm{Attention}(\textrm{LN}(x)) + x,\\
    y &= \textrm{FFN}(\textrm{LN}(h)) + h,
\end{align}
and such pre-norm has since become a standard practice in Transformer decoders. 

GPT-J~\citep{2021GPT-J} modifies the Transformer block to compute FFN sub-layer and self-attention sub-layer in parallel to increase computation throughput:

\begin{equation}
    y = x + \textrm{FFN}(\textrm{LN}(x)) + \textrm{Attention}(\textrm{LN}(x)),
\end{equation}
and \citet{2022PaLM} observes limited performance degradation when applying this design to larger models.

As self-attention alone cannot capture the position information of each input token, the vanilla Transformer adds a non-learnable vector that's dependent on the absolute position in the input sequence to the embedding of each token, which is known as cosine position embedding. Later works such as GPT and BERT use learnable embeddings instead, while T5 adopts relative position embedding~\citep{2018RelativePE}, where the information of relative position between query and key is injected in the computation of self-attention instead. A more recent scheme RoPE~\citep{2021RoPE} multiplies the keys and queries by a position-dependent rotation matrix, and is later shown to enable position interpolation for processing of longer sequences~\citep{2023PI,2023CodeLLaMA,2023YaRN}. Alternatively, \citet{2021ALiBi} propose ALiBi, which directly attenuates the attention scores according to the relative position between key and query. RoPE is popularized by its application in PaLM~\citep{2022PaLM} and GPT-NeoX~\citep{2022GPT-NeoX}, while ALiBi is adopted by BLOOM~\citep{2022BLOOM}. We refer to \citet{2023PESurvey} for a survey on position embedding and interpolation in Transformers.

Apart from position embeddings, another issue in Transformer that has long troubled researchers is the fact that the complexity of self-attention scales quadratically with the input sequence length. Some works such as Reformer~\citep{2020Reformer}, Linformer~\citep{2020Linformer}, Performer~\citep{2020Performer} and cosFormer~\citep{2022cosformer} use approximate attention to reduce this complexity, but they mostly come at the cost of degraded performance~\citep{2020EfficientTransformer,2021TransformerSurvey}. Other works tackle this issue from an engineering point-of-view. MQA~\citep{2019MultiQuery} shares the same set of keys and values across all attention heads to optimize memory-to-arithmetic ratio and significantly improves inference speed at small costs of model performance. Its variant Grouped-Query Attention~(GQA, \citealp{2023GQA}) takes a middle-ground approach by dividing attention heads into groups and sharing the same set of keys/values within each group. Orthogonally, \citet{2022FlashAttention} introduce FlashAttention, which is an exact but improved implementation of self-attention that optimizes IO operations on the accelerating device via tiling to improve memory efficiency.

Another important technique for improving LLMs' training efficiency is mixed precision training~\citep{2017mixed_precision}, which stores model weights and activations in lower precision to save memory consumption. Early works use FP16 format for this low precision, while BF16 has now become more popular~\citep{2022PaLM,2022BLOOM}, as it can represent the same range of floating point numbers as FP32, thus preventing numerical overflowing and underflowing during training.

%% file: sec_general-lm.tex
\section{General Language Models for Code}\label{sec:general-lm}
Since language models scaled to hundreds of billions of parameters~(\citealp{2020GPT3}; \citealp{2022PaLM}), many of them have demonstrated non-trivial coding capability, even if they are not specifically designed or trained for code. Pioneered by Codex, researchers have also found continual pretraining on code to significantly benefit language models' performance on code\footnote{While some works refer to this process as ``finetuning on code", it is still self-supervised in nature. Thus we choose to adopt the term ``extra/additional/continual pretraining" in this work to avoid confusion with supervised in-task finetuning or instruction finetuning.}.

\input{tables/general-model-performance}
% \vspace{-0.42cm}
\subsection{Off-the-Shelf Language Models}\label{sec:off-the-shelf}
Large language models are often pretrained on trillions of tokens following the scaling laws~(\citealp{2020scaling}; \citealp{2022Chinchilla}), and such an amount of text data is often a diverse composite with a non-negligible part of code. The Pile~\citep{2021Pile}, for example, includes 95GB of code crawled from GitHub out of its 800GB raw dataset, while the multilingual pretraining dataset ROOTS~\citep{2023ROOTS} also contains 163GB of code spanning 13 programming languages in its 1.6TB compound. As two of the largest open-source pretraining datasets, they have supported many language models with coding ability. GPT-J~\citep{2021GPT-J}, for example, is reported by \citet{2021Codex} to demonstrate non-trivial performance on HumanEval, while \citet{2022BLOOM} report similar results for GPT-NeoX~\citep{2022GPT-NeoX} and BLOOM. LLaMA~\citep{2023LLaMA}, whose pretraining dataset includes 328GB code from GitHub, achieves 23.7 pass@$1$ performance on HumanEval, and its successor LLaMA 2~\citep{2023LLaMA2}, achieves an even higher score of 29.9. 

Closed-source models, on the other hand, perform generally better. LaMDA~\citep{2022LaMDA} and PaLM~\citep{2022PaLM}, whose pretraining dataset contains 12.5\% and 5\% code respectively, achieve 14.0 and 26.2 pass@$1$ performance on HumanEval, while GPT-4~\citep{2023GPT4} set a staggering record of 67.0 (and an early version is reported by \citet{2023AGI} to be 82) that until recently has remained higher than any specialized models pretrained or instruction-finetuned for code.

More recently, the general trend has been to train smaller models with larger datasets, following the revised scaling law~\citep{2022Chinchilla}. Baichuan 2~\citep{2023Baichuan2}, for example, is a 13B model trained on 2.6T tokens, while Qwen~\citep{2023Qwen} is a 14B model trained on 3T tokens. They achieve 17.1 and 32.3 pass@1 on HumanEval, respectively. \citet{2023Phi-1.5}, however, demonstrate that models as small as 1.3B can acquire coding capability that's comparable to much larger models while also maintaining a reasonable performance on general text processing and even manifesting some emergent abilities~\citep{2022emergent} such as chain-of-though reasoning~\citep{2022CoT}. Their model, Phi-1.5, is trained on 21B tokens of textbook data generated by ChatGPT, and 100B tokens of filtered web data from Stack Overflow and Refined Web~\citep{2023RefinedWeb}, and attains 41.4 pass@$1$ performance on HumanEval. 

Another rising trend is mixture-of-expert models~\citep{2020GShard,2021SwitchTransformer,2021GLaM}. Notably, \citet{2024Mixtral} recently introduce Mixtral 8x7B, where only 13B parameters are activated for each token during inference, but achieving 40.2 on HumanEval and surpassing much larger dense models such as LLaMA 2. Similarly, \citet{2024DeepSeekMoE} present DeepSeekMoE, where only 2.8B parameters are activated in a 16B model, scoring 26.8 on HumanEval.

The exact performance of these models are presented in Table~\ref{tab:humaneval-mbpp}.

\subsection{Language Models with Additional Pretraining on Code}\label{sec:additional-pretraining}
Along with the seminal benchmark HumanEval, \citet{2021Codex} kick-started the age of LLM for code with Codex, which are GPT-3 checkpoints pretrained on 100B additional code tokens and one of the earliest multi-billion models for code. Following their work, other researchers have also specialized their LLMs on code with additional pretraining. \citet{2022PaLM} train PaLM on 7.8B additional code tokens to obtain PaLM-Coder, setting new state-of-the-art on HumanEval and MBPP (Table~\ref{tab:humaneval-mbpp}) that are only broken later by its successor PaLM 2-S*, the smallest version of PaLM 2~\citep{2023PaLM2} further trained on an undisclosed amount of code. Similarly, \citet{2022Minerva} train PaLM on 38.5B tokens of arXiv papers and mathematical content, while \citet{2023CodeLLaMA} train LLaMA 2~\citep{2023LLaMA2} on more than 500B code tokens to acquire Code LLaMA, whose performance on HumanEval surpasses all previous LMs except GPT-4 (Table~\ref{tab:humaneval-mbpp}). \citet{2023MFTCoder} further train Code LLaMA with multi-task finetuning (MFT) to introduce CodeFuse-CodeLLaMA, achieving 74.4 pass@1 on HumanEval and surpassing even the performance of GPT-4 published in \citet{2023GPT4}.

While almost all of these models are Transformer decoders pretrained with CLM, several architectural modifications have been introduced along the way, as we noted in Section~\ref{sec:background-design}. All these models use pre-norm, and GPT-J introduces parallel attention, which is later adopted by PaLM, GPT-NeoX, and Phi-1.5. PaLM introduces MQA and RoPE into LLMs, and RoPE is now employed by most language models, including GPT-NeoX, two generations of LLaMA, Qwen, and the 7B version of Baichuan 2. BLOOM and the 13B version of Baichuan 2, however, use ALiBi for position embeddings, while LLaMA 2 and Code LLaMA adopt GQA instead of MHA or MQA. In Section~\ref{sec:specialized-lm}, we show that specialized models pretrained exclusively on code have also followed these advancements closely.

%% file: tables/general-model-performance.tex
\begin{wraptable}{r}{0.5\linewidth}
    \caption{Pass@$k$ performance of raw language models (top) and language models with extra training on code (bottom) on HumanEval (0-shot) and MBPP (3-shot), ordered chronologically. For Phi-1.5 we consider Phi-1.5-web version, and for Code LLaMA we consider its Python version. $^1$~\citet{2021Codex}; $^2$~\citet{2022PaLM}; $^3$~\citet{2021MBPP}; $^4$~\citet{2022BLOOM}; $^5$~\citet{2023LLaMA}; $^6$~\citet{2023GPT4}; $^7$~\citet{2023AGI}; $^8$~\citet{2023LLaMA2}; $^9$~\citet{2023Phi-1.5}; $^{10}$~\citet{2023Baichuan2}; $^{11}$~\citet{2023Qwen}; $^{12}$~\citet{2023Mistral}; $^{13}$~\citet{2023Gemini}; $^{14}$~\citet{2024DeepSeek}; $^{15}$~\citet{2024Mixtral}; $^{16}$~\citet{2024DeepSeekMoE}; $^{17}$~\citet{2023PaLM2}; $^{18}$~\citet{2023CodeLLaMA}.}
    \label{tab:humaneval-mbpp}
    \centering
    \adjustbox{width=0.5\textwidth,center}{
    \begin{tabular}{lrcccc}
    \toprule
        & Size & \multicolumn{2}{c}{HumanEval (0)} & \multicolumn{2}{c}{MBPP (3)} \\
        & & k=1 & k=100 & k=1 & k=80 \\
    \midrule
        GPT-J$^1$ &6B & 11.6 & 27.7\\
        LaMDA$^{23}$ & 137B & 14.0 & 47.3 & 14.8 & 62.4 \\
        PaLM$^2$ & 540B & 26.2 & 76.2 & 36.8 & 75.0 \\
        GPT-NeoX$^4$ & 20B & 15.4 & 41.2 \\
        BLOOM$^4$ &176B & 15.5 & 55.5\\
        LLaMA$^5$ &65B & 23.7 & 79.3 & 37.7 & 76.8 \\
        GPT-4 & & 67.0$^6$/82$^7$ \\
        LLaMA 2$^8$ & 70B & 29.9 & 89.0 & 45.0 & 81.5 \\
        Phi-1.5$^9$ &1.3B& 41.4 & & 43.5\\
        Baichuan 2$^{10}$ &13B & 17.1 & & 30.2 \\
        Qwen$^{11}$ & 14B & 32.3 & & 40.8 \\
        Mistral$^{12}$ & 7B & 30.5 && 47.5 \\
        Gemini$^{13}$ & Ultra & 74.7 && \\
        DeepSeek$^{14}$ & 67B & 42.7 & & 57.4 \\
        Mixtral$^{15}$ & 8x7B & 40.2 & & 60.7 \\
        DeepSeekMoE$^{16}$ & 16B & 26.8 && 39.2\\
    \midrule
        Codex$^1$ & 12B & 28.8 & 72.3  \\
        PaLM-Coder$^2$ & 540B & 36.0 & 88.4 & 47.0 & 80.8 \\
        PaLM 2*$^{17}$ &S & 37.6 & 88.4 & 50.0 & 86.8 \\
        Code LLaMA$^{18}$ & 34B & 53.7 & 94.7 & 56.2 \\
        Code-Qwen$^{11}$ &14B & 45.1 && 51.4 \\
    \bottomrule
    \end{tabular}
    }
\end{wraptable}

%% file: sec_special-lm.tex
 \section{Specialized Language Models for Code}\label{sec:specialized-lm}
As pretrained Transformers such as GPT and BERT achieved remarkable success in natural language processing, such model architectures, learning paradigms, and training objectives were soon adopted by the software engineering community to produce specialized models for code understanding and generation. In this section, we first review common datasets used for pretraining code language models~(Section~\ref{sec:dataset}), and then dive into the complex family of code LMs by their model architecture: encoder-only models~(Section~\ref{sec:encoder}), encoder-decoder models~(Section~\ref{sec:encoder-decoder}), decoder-only models~(Section~\ref{sec:decoder}), UniLM~(Section~\ref{sec:unilm}), 
\begin{wraptable}{r}{0.5\linewidth}
    \caption{Statistics of several pretraining datasets for code models: size in bytes, number of files, and number of programming languages. In CodeSearchNet each file is a function. For Pile and ROOTS we only consider their code composite. $^a$~\citet{2019CodeSearchNet}; $^b$~\citet{2021Pile}; $^c$~\citet{2022PileDatasheet}; $^d$~\url{https://huggingface.co/datasets/codeparrot/github-code}; $^e$~\citet{2022Stack}; $^f$~\citet{2023ROOTS}; $^g$~\citet{2024StarCoder2}.}
    \label{tab:pretrain-dataset}
    \centering
    % \scalebox{.95}{
    \begin{tabular}{lrrr}
    \toprule
        Dataset & Size (GB) & Files (M) & \# PL \\
    \midrule
        CodeSearchNet$^a$ & 20 & 6.5 & 6\\
        The Pile$^{bc}$ & 95 & 19 & - \\
        CodeParrot$^d$ & 1K & 115 & 30\\
        The Stack$^e$ & 3136 & 317 & 30\\
        ROOTS$^f$ & 163 & 15 & 13\\
        The Stack v2$^g$ & 32K & 3K & 619\\
    \bottomrule
    \end{tabular}
    % }
\end{wraptable}
and diffusion models~(Section~\ref{sec:diffusion}). Lastly, in Section~\ref{sec:instruction} we also illustrate the current trend of applying more recent techniques in NLP, such as instruction tuning~(\citealp{2021FLAN}; \citealp{2021T0}; \citealp{2022FLAN}) and reinforcement learning~\citep{2022InstructGPT} to code processing. An overview of these pretrained models are provided in Table~\ref{tab:models}.

\subsection{Training Dataset for Code}\label{sec:dataset}
While text data for pretraining language models are often crawled from the web and must undergo meticulous and often aggressive preprocessing~\citep{2019T5}, code data come naturally as whole documents from public GitHub repositories. Even better, they come with readily available quality indicators such as the count of stars or forks (although \citet{2023SantaCoder} suggest that star count correlates poorly with downstream performance). As a result, many large-scale code pretraining datasets have been introduced, including CodeSearchNet~\citep{2019CodeSearchNet}, CodeParrot~\citep{2022TransformersBook}, and the Stack~\citep{2022Stack}, totaling 20GB, 50GB and 3TB of code documents respectively (Table~\ref{tab:pretrain-dataset}).

While these datasets are meant for training code models, it should be noted that code is ultimately a special form of natural language, as the vocabulary of most programming languages is a small subset of English. Besides, high-quality code is often interleaved with natural language comments or documentations, which also enables models to acquire certain knowledge of general text representation. In fact, of the 6.5M functions in CodeSearchNet, 2.3M are paired with natural language documentation, allowing models to train explicitly on such bimodal data.

Compared with natural language, another byproduct of scraping code from GitHub is commit histories, which consist of code before commit, code after commit, and a short message describing the commit, which can loosely serve as an instruction for language models. \citet{2023OctoPack} utilize this feature and construct a 2GB dataset CommitPackFT containing 742K samples of instruction data for code, obviating the need of extensive human labor that's required to construct natural language instructions~(\citealp{2021T0}; \citealp{2022SuperNatural}).

Apart from bimodal training and instruction finetuning, another recent trend in constructing code dataset is synthesizing data with powerful models such as ChatGPT. While this method is originally proposed for generating instruction data in natural language~(\citealp{2022Self-Instruct}; \citealp{2022Unnatural}), \citet{2023Phi-1} take one step further and synthesize 1B tokens of Python textbooks and coding exercises to pretrain a 1.3B model, achieving state-of-the-art results on HumanEval that's comparable to much larger models trained on significantly larger datasets.

\input{tables/specialized_models}

\subsection{Encoders}\label{sec:encoder}
Pretrained Transformer encoders such as BERT~\citep{2018BERT}, RoBERTa~\citep{2019RoBERTa}, and ELECTRA~\citep{2020ELECTRA} have attained impressive results on natural language understanding tasks, and these methods were soon introduced into code processing after their advent. \citet{2019CuBERT} replicate the training procedure of BERT on a code corpus to produce CuBERT, showcasing its superior performance over LSTM~\citep{1997LSTM} and non-pretrained Transformers. \citet{2020CodeBERT}, on the other hand, train CodeBERT with MLM and ELECTRA's RTD on CodeSearchNet. They also utilize the explicit text-code pairs in CodeSearchNet, and use them respectively as the first and second segment in BERT's input. When using CodeBERT to initialize the encoder part of a vanilla Transformer for sequence-to-sequence generation tasks such as code summarization, they observe a moderate performance gain over non-pretrained baselines.

Apart from these standard training objectives, many auxiliary objectives specifically designed for code have also been introduced. GraphCodeBERT~\citep{2020GraphCodeBERT} and SynCoBERT~\citep{2021SynCoBERT} both extract graphs from the source code (data flow graph and abstract syntax tree, respectively) and train the models to predict the typological relations between the nodes, while SynCoBERT and Code-MVP~\citep{2022Code-MVP} also add type inference to their pretraining stage in the form of tagging. Another common objective is contrastive learning: SynCoBERT and Code-MVP contrast between different views of the input (such as code, comment, AST, and transformed code), while DISCO~\citep{2021DISCO} constructs positive sample pairs by semantic-preserving transformations such as obfuscation, and negative pairs by injecting artificial bugs.

\subsection{Encoder-Decoders}\label{sec:encoder-decoder}
In NLP, pretrained Transformer encoder-decoders such as T5~\citep{2019T5} and BART~\citep{2019BART} have also left a notable mark in the past few years' advancement in language modeling. T5, for example, unifies all textual tasks into a sequence to sequence format and sets new records on GLUE~\citep{2018GLUE} and SuperGLUE~\citep{2019SuperGLUE}. Compared with encoder-only models, encoder-decoders are naturally more powerful as they can be used for conditional text generation, while their encoder part can always be taken out to perform tasks that require an encoder-only architecture, such as regression~\citep{2022UL2}.

Inspired by these advantages of encoder-decoder architecture, many such models have been proposed for code processing. PyMT5~\citep{2020PyMT5} and \citet{2021T5Code} replicate the pretraining and multi-task finetuning process of T5 on code corpus, while \citet{2021PLBART} introduce PLBART, a BART pretrained on 655GB combined data of Java, Pyhton, and natural language. \citet{2021DOBF} argue that MLM could be too easy a task for programming languages as identifier names often occur multiple times in a single context window, and propose a deobfuscation pretraining objective, where the model is trained to convert obfuscated code back to its original form. Related to this method, we note that meaningful variable names have also been found to have a positive impact on the code generation process of large language models~\citep{2022PoT}.

Building on these early works, \citet{2021CodeT5} propose CodeT5, which is pretrained alternatively with 1) T5's original span corruption; 2) identifier tagging (where each token in the code input is tagged as either identifier or non-identifier); 3) masked identifier prediction (a special form of span corruption where all identifiers are masked); and 4) text-to-code \& code-to-text generation. Its successor, CodeT5+~\citep{2023CodeT5+}, take inspiration from UL2~\citep{2022UL2} and introduce causal language modeling (CLM) into pretraining, along with additional contrastive objectives based on text-code matching. 

AlphaCode~\citep{2022AlphaCode} is also trained with multiple objectives, where the encoder is trained with MLM and the decoder is trained with CLM, with architecture modifications such as shallow-encoder \& deep-decoder, multi-query attention~\citep{2019MultiQuery}, and being much larger than CodeT5 (up to 41B parameters). NatGen~\citep{2022NatGen}, on the other hand, is pretrained with a "naturalization" objective similar to deobfuscation: semantically equivalent but unnatural code is generated by predefined operations such as loop transformation, dead code injection, and variable renaming, and the model is pretrained to translate 
\input{tables/specialized-model-performance}
these unnatural code back to its original form. We note that some of these models are built on previous works. For example, NatGen is initialized with CodeT5, while the largest version of CodeT5+ is initialized from a decoder-only model, CodeGen~\citep{2022CodeGen}.

Apart from these general pretraining objectives, several works have also trained Transformer encoder-decoders with a focus on code translation, which is a natural application of Transformer models in code as the Transformer architecture was originally proposed by \citet{2017Transformer} for machine translation (MT). However, unlike natural languages, where parallel corpus across two or more human languages exist in abundance, there is little parallel data for code. To tackle this issue, \citet{2020Transcoder} propose Transcoder, which first pretrains an encoder with XLM~\citep{2019XLM}, and then initializes a vanilla Transformer with this encoder and continue to pretrain it with Denoising Auto-Encoding~(DAE, \citealp{2019BART}) and back translation~\citep{2015BT}, while its follow-up work~\citep{2022Transcoder} also utilize language-independent intermediate representations to enhance this process, which we discuss in more detail in Section~\ref{sec:code-specific}.

Apart from training data and objectives, these models mostly keep to the original architectures proposed by the NLP community, as shown in Table~\ref{tab:models}. Models based on BART, for example, use post-normalization and learnable absolute position embeddings, while those based on T5 use its simplified relative position embeddings and pre-normalization.

\subsection{Decoders}\label{sec:decoder}
After the monumental debut of GPT-3~\citep{2020GPT3} and the discovery of in-context learning, decoder-only Transformer models have become dominant in language modeling~(\citealp{2021Gopher,2022Chinchilla,2022PaLM,2022BLOOM,2023LLaMA,2023LLaMA2}, \textit{inter alia}). Many models similarly pretrained with CLM have also emerged in code processing, such as GPT-C~\citep{2020GPT-C}, CodeGPT~\citep{2021CodeXGLUE}, PolyCoder~\citep{2022PolyCoder}, CodeGen~\citep{2022CodeGen}, PyCodeGPT~\citep{2022PyCodeGPT}, Pangu-Coder~\citep{2022Pangu-Coder}, CodeGeeX~\citep{2023CodeGeeX}, Jam~\citep{2023Jam}, Phi-1~\citep{2023Phi-1}, and CodeFuse~\citep{2023CodeFuse13B}. Of these models, several alternative training objectives have been experimented with, such as MLM and Masked CLM\footnote{In their paper, MLM is conducted by replacing tokens in the input with \texttt{<mask>} and predicting it from only the left context, while Masked CLM is performed by adding a \texttt{<mask>} in the input and predicting the next token from it. Both tasks do not change the attention mask patterns of the model.} in Pangu-Coder, but are found to underperform compared with CLM-only training. \citet{2022PyCodeGPT} also propose continual training on sketches, where the model learns to first generate a sketch of a program and then the actual code. Notably, \citet{2023Phi-1} present Phi-1, a 1.3B small model trained on a dataset of only 7B tokens consisting of 6B tokens from StackOverflow and 1B synthetic data generated by ChatGPT but achieving 50.6 pass@$1$ on HumanEval and 55.5 pass@$1$ on MBPP, comparable to much larger (both in model size and training data size) models such as Code LLaMA or PaLM 2.

Although \citet{2022Pangu-Coder} report denoising objectives to underperform in decoder-only models, there have been other works that successfully combine denoising or multi-task pretraining with decoder architecture. Incoder~\citep{2022InCoder}, SantaCoder~\citep{2023SantaCoder}, StarCoder~\citep{2023StarCoder}, DeepSeek Coder~\citep{2023DeepSeekCoder}, and CodeShell~\citep{2024CodeShell} are trained with fill-in-the-middle (FIM) objective, also referred to as causal masking by \citet{2022InCoder}, which is essentially span corruption~\citep{2019T5} adopted to decoder-only architecture. One of the visible advantages of these infilling objectives is that they inject the models with the ability to fill in blanks in the middle of input code at inference time, while CLM allows only for autoregressive generation. As Table~\ref{tab:specialized-model} shows, however, these objectives also lead to higher performance on downstream tasks when compared with CLM-only models such as CodeGen, although the exact benefits of infilling training remain controversial~\citep{2023CodeGen2}.

Observing Table~\ref{tab:models}, it is clear that decoder-only models for code have generally followed the practices in NLP more closely, when compared with other model architectures. All these models use pre-normalization, while MQA, RoPE, and parallel attention have also been adopted by several models. Notably, the three most recent models - StarCoder, Phi-1, and CodeFuse - also employ FlashAttention to improve model throughput.

\subsection{UniLMs}\label{sec:unilm}
Following UniLM~\citep{2019UniLM} in NLP, several works in code processing have also pretrained this fourth family of Transformer models on code. CugLM~\citep{2020CugLM} is trained with both CLM and MLM + NSP via alternating attention masks, while UniXcoder is trained with CLM, MLM, Span Corruption (in Prefix LM style) along with auxiliary objectives including contrastive learning and text-code mutual generation. Both two models, however, are relatively small in size, and whether or not this architecture is suitable for code processing is yet to be explored at scale.

\subsection{Diffusion Models}\label{sec:diffusion}
Currently the Transformer architecture dominates text generation, but several works~(\citealp{2022diffusionLM}; \citealp{2022CPD}) have also adopted Diffusion Models~\citep{2020diffusion} from computer vision for text generation. Recently CodeFusion~\citep{2023codefusion} also introduces diffusion models into code modeling, and demonstrates that a 75M diffusion model can outperform StarCoder, CodeT5+, and GPT-3 on three code synthesis datasets.

\subsection{Instruction Finetuning and Reinforcement Learning for Code}\label{sec:instruction}
In natural language processing, training models on a diverse set of tasks with instruction prefix, known as instruction finetuning, has been shown to unlock the ability of cross-task generalization~(\citealp{2022InstructGPT}; \citealp{2022FLAN}; \citealp{2022OPT-IML}). At first, these instruction data samples are manually compiled or crowd-sourced~(\citealp{2021FLAN}; \citealp{2021T0}), but later researches find LLM-generated instructions to be sufficient~(\citealp{2022Self-Instruct}; \citealp{2022Unnatural}).

Following these works in natural language, researchers from the code community have applied instruction tuning to their models as well. \citet{2023CodeT5+} finetune CodeT5+ with 20K instruction data generated by InstructGPT~\citep{2022InstructGPT} to obtain InstructCodeT5+. WizardCoder~\citep{2023WizardCoder} follows the methods of WizardLM~\citep{2023WizardLM} to evolve 20K code Alpaca~\citep{2023alpaca} samples into a 78K dataset and uses it to finetune StarCoder. Pangu-Coder 2~\citep{2023Pangu-Coder2} also uses WizardLM's Evol-Instruct to generate 68K instruction samples from 20K code Alpaca, but also introduces reinforcement learning via Rank Responses to align Test \& Teacher Feedback (RRTF). OctoCoder~\citep{2023OctoPack}, on the other hand, takes a different path and uses Git commit histories as instruction data to finetune StarCoder and CodeGeeX2. More recently, CodeFuse~\citep{2023MFTCoder} also employs multitask-finetuning and explicitly introduces multiple downstream tasks into their instruction data. The performance of these instruction finetuned code models can also be found in Table~\ref{tab:specialized-model}.

In NLP, another technology closely related to instruction finetuning is reinforcement learning from human feedback (RLHF), which has played a significant role in aligning LLMs with human values~(\citealp{2022InstructGPT}; \citealp{2022Anthropic}). The merit of reinforcement learning is that it can incorporate non-differentiable reward signals into training, such as BLEU~\citep{2016RL-seq2seq} and human preference~\citep{2017RL-human}, but the human feedback required in aligning LLMs often involves extensive labor on annotation. In comparison, applying reinforcement learning to code models has a natural advantage, as compilers can be used for automatically generating feedback for code samples produced by language models.

CodeRL~\citep{2022CodeRL} is one such model, which defines four levels of rewards for each generated program (viz. compile error, runtime error, unit test failure, pass) as well as fine-grained token-level reward estimated by a critic model. The actor model, which is an extension of CodeT5, is then trained with REINFORCE algorithm~\citep{1992REINFORCE}. Similarly, CompCoder~\citep{2022CompCoder} and PPOCoder~\citep{2023PPOCoder} train CodeGPT and CodeT5 respectively with proximal policy optimization~\citep{2017PPO}, while RLTF~\citep{2023RLTF} proposes fine-grained feedback based on the error information and location provided by the compiler, as well as adaptive feedback that takes the ratio of passed test cases into account.

%% file: tables/specialized_models.tex
\begin{landscape}
\begin{table}[p]
    \caption{An overview of pretrained code language models' architecture and training details: their base architecture, model size, vocabulary, context length, position embedding, training precision, attention type (MHA, MQA, or GQA), layer normalization type (post-norm or pre-norm), usage of FlashAttention, training initialization, objectives, dataset size (either in disk size, measured by GB/TB, or in number of tokens, measured by B/T), tokens seen during training, supported number of programming languages, and institute. We note that the number of training tokens does not count the training tokens of the model used for initialization, if any. The training objectives are: MLM (Masked Language Modeling), NSP (Next Sentence Prediction), RTD (Replaced Token Detection), IP (Identifier Prediction), CL (Contrastive Learning), SC (Span Corruption), DAE (Denoising Auto-Encoding), Text$\leftrightarrow$Code (text-to-code generation and code-to-text generation), MT (Machine Translation). Missing information (such as AlphaCode's position embedding type) is left as blank.}
    \label{tab:models}
    \centering
    \adjustbox{width=\textwidth+6.45cm,center}{
    % \begin{tabular}{llcrm{3cm}<{\centering}crrrrrr}
    % \begin{tabular}{lm{2cm}m{2cm}lrm{3cm}crr>{\raggedleft\arraybackslash}m{2cm}rr>{\raggedleft\arraybackslash}m{1.7cm}}
    \begin{tabular}{lm{2cm}|m{1.8cm}>{\raggedleft\arraybackslash}m{1.8cm}rrcm{0.9cm}<{\centering}m{0.9cm}<{\centering}m{1cm}<{\centering}m{0.8cm}<{\centering}m{1cm}<{\centering}|ll>{\raggedleft\arraybackslash}m{2.05cm}rr|r}
    \toprule
        Date    	&Model         	&Arch.      &Size      	&Vocab 	&Context   	&PE    &Preci-sion    	&Atten. Type	&Parallel Atten.	&Pre-Norm	&Flash Atten.	&Init. from	&Objectives        	&Dataset   	&Training  	&PL 	&Inst.\\
    \midrule
        2019-12 	&CuBERT        	&BERT      	&350M      	&50K   	&1024	&absolute   &fp32&MHA	&	&	&	&-         	&MLM + NSP         	&9.3B      	&93B       	&1	&Google \\
        2020-02 	&CodeBERT      	&RoBERTa   	&125M      	&50K   	&512	&absolute  	&fp32&MHA	&	&	&	&RoBERTa   	&MLM + RTD         	&20GB      	&105B      	&6	&Microsoft\\
        2020-09 	&GraphCode-BERT 	&RoBERTa   	&125M      	&50K 	&640	&absolute&fp32&MHA	&	&	&	&CodeBERT  	&\small{MLM + Edge Prediction + Node Alignment} 	&20GB 	&131B 	&6	&Microsoft\\
        2021-08 	&SynCoBERT     	&RoBERTa   	&125M      	&50K 	&512	&absolute 	&fp32&MHA	&	&	&	&CodeBERT  	&\small{MLM + IP + AST Edge Prediction + CL} 	&20GB 	&7B 	&6	&Huawei\\
        2021-10 	&DISCO         	&BERT      	&100M      	&20K 	&512	&absolute 	&fp32&MHA	&	&	&	&-         	&\small{MLM + Node Type MLM + CL} 	&1.8GB 	&	&2	&Columbia \& IBM\\
        2022-05 	&Code-MVP      	&RoBERTa 	&125M     	&50K 	&512	&absolute 	&fp32&MHA	&	&	&	&GraphCodeBERT 	&\small{MLM + Type Inference + CL} 	&2GB 	&39B     	&1	&Huawei\\
        2022-10     &SCodeR         &RoBERTa    &125M       &51K    &1024   &absolute   &fp32&MHA   &	&	&	&UniXcoder    & CL & 20GB & &6&Microsoft\\
    \midrule
        2020-05 	&GPT-C         	&GPT-2     	&374M      	&60K   	&1024	&absolute  	&    &MHA	&	&\Checkmark	&	&-         	&CLM               	&11B       	&270B      	&4	&Microsoft\\
        2021-02 	&CodeGPT       	&GPT-2     	&124M      	&50K   	&1024	&absolute  	&fp32&MHA	&	&\Checkmark	&	&GPT-2     	&CLM               	&2GB       	&	&1	&Microsoft \\
        2022-02 	&PolyCoder     	&GPT-NeoX   &160M-2.7B 	&50K   	&2048	&RoPE  	    &fp32&MHA	&	&\Checkmark	&	&-         	&CLM               	&254GB     	&39B       	&12	&CMU\\
        2022-03 	&CodeGen-Multi(Mono)&GPT-J  &350M-16.1B	&50K   	&2048	&RoPE      	&fp16&MHA	&\Checkmark	&\Checkmark	&	&-    	&CLM               	&1.6TB(1.8TB)/ 506B(577B) 	&1T(1.2T) 	&6(1) 	&Salesforce\\
        2022-04 	&InCoder       	&GPT-3     	&6.7B      	&50K   	&2048	&Cosine    	&fp32&MHA	&	&\Checkmark	&	&-         	&Causal Masking    	&204GB     	&52B       	&28	&Meta\\
        2022-06     &PyCodeGPT      &GPT-Neo    &110M       &32K    &1024   &absolute   &fp32&MHA   &   &\Checkmark	&   &-          &CLM                &96GB       &100B       &1  &Microsoft\\
        2022-07 	&PanGu-Coder   	&PanGu-$\alpha$ &317M-2.6B &42K &1024	&absolute  	&    &MHA	&	&\Checkmark	&	&-         	&CLM          	&147GB     	&230B      	&1	&Huawei\\
        2023-01 	&SantaCoder    	&GPT-2     	&1.1B      	&49K   	&2048	&absolute  	&fp32&MQA	&	&\Checkmark	&	&-         	&FIM               	&268GB     	&236B      	&3	&BigCode\\
        2023-03 	&CodeGeeX      	&PanGu-$\alpha$ &13B  	&52K   	&2048	&absolute  	&fp16&MHA	&	&\Checkmark	&	&-         	&CLM               	&158B      	&850B      	&23	&Tsinghua\\
        2023-05 	&StarCoder     	&GPT-2     	&15.5B     	&49K   	&8192	&absolute  	&fp32&MQA	&	&\Checkmark	&\Checkmark	&-  &FIM               	&815GB     	&1T        	&86	&BigCode\\
        2023-05     &Jam            &GPT-2      &350M       &50K    &256    &absolute   &bf16&MHA    &   &\Checkmark &   &-          &CLM                &36GB/20B   &9B         &1  &U.  Notre Dame\\
        2023-06 	&Phi-1         	&GPT-J     	&1.3B      	&51K   	&2048	&RoPE      	&fp16&MHA	&\Checkmark	&\Checkmark	&\Checkmark	&-         	&CLM               	&7B        	&53B       	&1	&Microsoft\\
        2023-10 	&CodeFuse      	&GPT-J     	&350M-13B  	&101K  	&4096	&RoPE      	&fp16&MHA	&\Checkmark	&\Checkmark	&\Checkmark	&-         	&CLM               	&1.6TB / 1T	&	&40+	&Ant Group\\
        2023-10     &CodeShell      &GPT-2      &7B         &70K    &8192   &RoPE       &bf16&GQA    &   & \Checkmark&   &-          &FIM                & 100B          &500B       &   &Peking U.\\
    \midrule
        2020-10 	&PyMT5         	&GPT-2	&374M      	&50K   	&1024+1024 	&absolute	&fp16&MHA	&	&\Checkmark	&	&-         	&SC                	&27GB      	&	&1	&Microsoft\\
        2021-02 	&Mastropaolo et al. &T5 &60M       	&32k   	&512+512   	&T5         &bf16&MHA	&	&\Checkmark	&	&-    	&SC                	&1GB       	&	&1	&USI\\
        2021-02 	&DOBF          	&	 &250M      	&50K   	&512+512   	&absolute	&fp16&MHA	&	&	&	&-         	&\small{MLM + Deobfuscation} 	&45GB      	&	&2	&Meta\\
        2021-03 	&PLBART        	&BART   &140M      	&50K   	&1024+1024 	&absolute  	&fp32&MHA	&	&	&	&-         	&DAE               	&655GB / 71B 	&210B    	&2	&\small{UCLA \& Columbia}\\
        2021-09 	&CodeT5        	&T5        	&60M-220M  	&32K 	&512+256  	&T5 	&fp16&MHA	&	&\Checkmark	&	&-         	&\small{SC + IP + Masked IP + Text$\leftrightarrow$Code}	&$\sim$25GB 	&	&8	&Salesforce\\
        2022-01 	&SPT-Code      	&BART	&262M      	&80K	&512+512	&absolute	&fp16&MHA	&	&	&	&-         	&\small{NSP + SC + Method Name Prediction} 	&20GB 	&	&6	&Nanjing U.\\
        2022-02 	&AlphaCode     	&	&300M-41B  	    &8K    	&1536+768  	&	        &bf16&MQA	&	&	&	&-         	&MLM + CLM         	&715GB     	&967B      	&13	&DeepMind\\
        2022-06 	&NatGen        	&T5        	&220M    &32K   &512+256   	&T5        	&fp16&MHA	&	&\Checkmark	&	&CodeT5    	&Naturalization    	&$\sim$26GB	&14B       	&8	&\small{Columbia \& UCD}\\
        2022-12     &ERNIE-Code     &mT5        &580M     &250K &1024+1024  &T5         &bf16&MHA   &	&\Checkmark	&   &mT5 & \small{SC + Text$\leftrightarrow$Code + MT}   &  & 197B & 6 & Baidu \\
        2023-05 	&CodeT5+       	&T5/GPT-3	&220M-16B &50K 	&2048+2048 	&absolute 	&fp16&MHA	&\Checkmark	&\Checkmark	&	&CodeGen-mono 	&\small{SC + CLM + CL + Text$\leftrightarrow$Code} 	&52B 	&	&9	&Salesforce\\
    \midrule
        2020-12 	&CugLM         	&BERT	&51M       	&50K 	&128	    &absolute 	&fp32&MHA	&	&	&	&-         	&\small{MLM + NSP + CLM} 	&8M        	&1.2B      	&2	&Peking U.\\
        2022-03 	&UniXcoder     	&RoBERTa	&125M      	&51K 	&1024	&absolute 	&fp32&MHA	&	&	&	&-         	&\small{MLM + CLM + SC + CL + Code2Text} 	&20GB+ 	&839B 	&6	&Microsoft\\
    \bottomrule
    \end{tabular}
    }
\end{table}
\end{landscape}

%% file: tables/specialized-model-performance.tex
\begin{wraptable}{r}{0.5\linewidth}
    \caption{Pass@$1$ performance of pretrained code models (top), instruction finetuned code models (middle), in comparison with some of the best general language models (bottom), with models in each category ordered chronologically. The sources of these figures can be found in Section~\ref{sec:encoder-decoder}, Section~\ref{sec:decoder}, and Table~\ref{tab:humaneval-mbpp}.}
    \label{tab:specialized-model}
    \centering
    \adjustbox{width=0.5\textwidth,center}{
    \begin{tabular}{lrcc}
    \toprule
        Model & Size & HumanEval & MBPP \\
    \midrule
        PolyCoder       &   2.7B    & 5.6  & -    \\
        CodeGen-Mono    &   16.1B   & 29.3 & 35.3 \\
        InCoder         &   6.7B    & 15.2 & 19.4 \\
        PyCodeGPT       &   110M    & 8.3  & -    \\
        Pangu-Coder     &   2.6B    & 23.8 & 23.0 \\
        SantaCoder      &   1.1B    & 14.0 & 35.0 \\
        CodeGeeX        &   13B     & 22.9 & 24.4 \\
        StarCoder       &   15.5B   & 33.6 & 52.7 \\
        CodeT5+         &   16B     & 30.9 & -    \\
        Phi-1           &   1.3B    & 50.6 & 55.5 \\
        CodeFuse        &   13B     & 24.8 & -    \\
        DeepSeek Coder  &   33B     & 56.1 & 66.0 \\
    \midrule
        InstructCodeT5+ &   16B     & 35.0 & -    \\
        WizardCoder     &   15.5B   & 57.3 & 51.8 \\
        Pangu-Coder 2   &   15.5B   & 61.6 & -    \\
        OctoCoder       &   15.5B   & 46.2 & -    \\
        CodeFuse        &   34B     & 74.4 & -    \\
        DeepSeek Coder-Instruct &33B& 79.3 & 70.0 \\
    \midrule
        GPT-4           &   -       &67.0/82 & -  \\
        PaLM 2*         &   S       & 37.6 & 50.0 \\
        Code LLaMA      &   34B     & 53.7 & 56.2 \\
        Phi-1.5         &   1.3B    & 41.4 & 43.5 \\
    \bottomrule
    \end{tabular}
    }
\end{wraptable}

%% file: sec_code-feature.tex
\section{Code Features for Language Models}\label{sec:code-specific}
A major difference between programming languages and natural languages is that the former is artificially defined to be precise and unambiguous, and need to be compiled (or interpreted) without error before execution. This allows for a much larger flexibility in designing pretraining objectives on code, beside lexical manipulations such as CLM, MLM, and Span Corruption. A similar trend can be observed in the last years before neural networks were introduced into mainstream NLP literature~(\citealp{2014seq2seq}; \citealp{2014attention}), when researchers in the MT community utilized alternative views of text such as syntactic features to improve the performance of SMT systems~(\citealp{2006Galley}; \citealp{2007Chiang}). These features, however, are not universally applicable or even agreed upon, and often result in highly complicated systems (for example, the size of English part-of-speech tagging's label set may range from dozens to hundreds). 

Programming languages, however, fare much better in these aspects. Each mainstream programming language, such as C, Python, and Java, comes with readily available compiler toolkits that allow for easy and accurate extraction of semantic information such as Abstract Syntax Tree (AST), language-independent Intermediate Representation (IR), and auxiliary information such as type of each token and control/data flow graph (CFG/DFG). Thus, in the context of Transformer-based language modeling for code, many works have incorporated these features into their training procedure.

\subsection{Abstract Syntax Tree and Intermediate Representation}
AST is one of the most common intermediate results of the compiling process, where a program is parsed into a tree of operations and their operands. Before the popularization of Transformer in the code processing community, there had been works such as InferCode~\citep{2020InferCode} that processes these representations with special network architectures like Tree-Based CNN and conducts self-supervised pretraining by predicting subtrees. 

TreeBERT~\citep{2021TreeBERT} is one of the first attempts to take AST into the Transformer-based pretraining-finetuning framework. It's a Transformer encoder-decoder pretrained with Tree MLM and Node Order Prediction, where the encoder takes a set of constituent paths in the AST as input (with each token being a path, which is the concatenation of its nodes' representations) while the decoder takes the code as input. Tree MLM is then performed by masking certain nodes in a path representation and its corresponding code tokens in the decoder input, while Node Order Prediction is accomplished by swapping nodes in a path and predicting it with a \texttt{[CLS]} token similar to BERT.

The method used by TreeBERT, however, is complicated and does not scale well. Later works mostly opt to first process AST into a text sequence and treat it like a normal part of the input. \citet{2021SynCoBERT}, for example, process AST with depth-first traversal and concatenate it with code and comment, and then train SynCoBERT (which, unlike TreeBERT, is actually a BERT-like encoder-only model) with four objectives: 1) MLM; 2) identifier tagging; 3) AST edge prediction (predicting whether there exists an edge between two AST nodes from the dot product of these nodes' representations); and 4) contrastive learning over i) code and AST pairs, as well as ii) text and code-AST pairs. Similarly, SPT-Code~\citep{2022SPT-Code}, a Transformer encoder-decoder, takes the concatenation of code, sequentialized AST, and text as input, and is pretrained with 1) span corruption; 2) code-AST prediction (NSP with one segment being code and one segment being AST); and 3) method name generation, a special form of span corruption where a method name is masked. Different from other works, however, they do not take the docstrings as the text segment in their input, but instead concatenate all method names appearing in the code as a succinct natural language description. Likewise, UniXcoder~\citep{2022UniXcoder} takes flattened AST instead of source code as its input during training.

In the compiling pipeline, AST is usually followed by language-independent intermediate representations, such as LLVM IR~\citep{2004LLVM}. Such features' independence from specific programming languages makes them suitable candidates for translation pivots, as is English in machine translation of low-resource natural languages~\citep{2019pivot}. \citet{2022Transcoder} take advantage of this characteristic and extend Transcoder~\citep{2020Transcoder} with translation language modeling~\citep{2019XLM} over code and IR, as well as IR generation from code. They also investigate other objectives such as IR decompilation (i.e. generating code from IR) and IR pivot (i.e. directly generating code in one language from the IR of another language), both showing promising results.

\subsection{Control Flow and Data Flow}\label{sec:CFGDFG}
While AST and IR have proved to be useful information in certain tasks such as code translation, they are static by nature, just like the source code, and may fail to capture semantic properties of code that are only revealed at runtime~\citep{2019LiGer}. Such semantics, however, are contained in dynamic features such as control flow and data flow. Similar to AST, specialized networks were used to process such information before the rise of pretrained Transformers, such as Message Passing Neural Network used by ProGraML~\citep{2020ProGraML}. Unlike AST, however, even after pretrained Transformers became dominant few works have looked in this direction. 

GraphCodeBERT~\citep{2020GraphCodeBERT} is one of such works, which creates special tokens and position embeddings for variables in the flow graph, and concatenates the variable sequence after text and source code to construct model input, with tailored attention masks on the code and variable segments: tokens from code segment and variable segment can attend to each other if and only if the variable is identified from the code token, and for tokens within the variable segment, $v_i$ is allowed to attend to $v_j$ if there is a direct edge from $v_j$ to $v_i$ in the dataflow. The model is then pretrained with MLM in combination with edge prediction and node alignment, both of which are accomplished by binary classification from the dot product of two tokens' representations (one from code segment and one from variable segment for node alignment, and both from variable segment for edge prediction).

\subsection{Type}
Apart from AST, IR, and data flow, type information has also been used to aid language models in processing code. CugLM~\citep{2020CugLM}, for example, uses type information during finetuning to aid in the prediction of tokens for unidirectional MLM (i.e. MLM with unidirectional attention mask): the type of a masked token is first predicted from the final Transformer layer's representation, and then the token itself is predicted based on both the hidden representation and predicted type. In contrast, both CodeT5~\citep{2021CodeT5} and SynCoBERT~\citep{2021SynCoBERT} include identifier tagging in their pretraining objectives, which can be viewed as coarse-grained type prediction.

\subsection{Program Transformation}
As we have shown in Section~\ref{sec:encoder-decoder}, function-preserving program transformations have proven to be important techniques in pretraining code language models. Obfuscation is one instance of program transformation, and others include loop transformation (for-while), condition transformation (if-switch), dead code injection (e.g. \texttt{if True: pass}), and statement swapping. DOBF~\citep{2021DOBF} and NatGen~\citep{2022NatGen} are two code language models pretrained to recover the original program from such transformations, while \citet{retrieval2021-7} also apply program transformations during the finetuning stage of language models to make them more robust to transformed test samples.

Notably, \citet{2022Code-MVP} integrate many of the aforementioned features into Code-MVP: source code, docstrings, AST, CFG, and transformed source code via identifier renaming, loop exchange, and dead code insertion. The model, initialized from GraphCodeBERT, is then trained with MLM, fine-grained type prediction, and contrastive learning across different views, such as text vs. code, code vs. AST, and code vs. CFG.

%% file: sec_eco.tex
\section{LLMs in Software Development}\label{sec:eco}
As language models set new records on software engineering benchmarks, software engineering technologies are also expanding the boundaries of language models in return, and have subsequently led them into real-world development cycles.

\subsection{LLMs Extended with Coding Tools}\label{sec:llm-extention}
Research in the NLP community has shown that LLMs can learn to use external tools such as calculators, MT systems, and search engines~(\citealp{2022LaMDA}; \citealp{2023Toolformer}). As such, \emph{interpreter} has been used to augment LLMs in complex reasoning tasks. PAL~\citep{2022PAL} and PoT~\citep{2022PoT} both extend Codex with Python interpreters for numerical calculations, while ViperGPT~\citep{2023ViperGPT} extends it further by calling vision APIs to extract information from visual input and answer related questions. 

However, LLMs do not always produce code that can be interpreted and executed, and there has also been a line of works that explore the possibility of emulating the interpreter with LLMs themselves. \citet{2021Scratchpad} trains LLMs to emulate the execution of programs by outputting program states at each step. More recently, \citet{2023CoC} propose Chain-of-Code, where a real interpreter executes the generated code until an error occurs, whereupon the LLM takes over to simulate execution. \citet{2024Think-and-Execute} similarly propose Think-and-Execute framework, where an LLM generates pseudo code and then simulates its execution to solve reasoning tasks.

Apart from alleviating the burden of numerical calculation in abstract reasoning tasks, interpreter (together with unit tests) also provides feedback on the process of code generation itself, allowing for interactive generation and refinement of code. Such works include Self-Edit~\citep{2023Self-Edit}, LeTI~\citep{2023LeTI}, OpenCodeInterpreter~\citep{2024OpenCodeInterpreter}, ProCoder~\citep{2024ProCoder}, Cycle~\citep{2024Cycle}, and SOAP~\citep{2024SOAP}, which run model-generated code against unit tests to provide feedback for further refinement. Alternatively, CodeT~\citep{2022CodeT}, TiCoder~\citep{unit2022-3}, and CONLINE~\citep{2024CONLINE} also utilize the LLM itself to generate unit tests, while Self-Refine~\citep{2023Self-Refine} sends the generated code to an LLM instead of an interpreter for feedback. \citet{2023interpreter} show that OpenAI's interpreter plugin\footnote{\url{https://openai.com/blog/chatgpt-plugins\#code-interpreter}} allows GPT-4 to self-debug, while InterCode~\citep{2023InterCode} provides a benchmark for evaluating interactive coding. In Section~\ref{sec:instruction} we have also shown that the execution results on unit tests serve as natural supervision signals for reinforcement learning on code. 

A topic closely related to tool using in LLM research is \emph{planning} as intelligent agents, which has been shown to enhance LLMs' capability both theoretically and empirically~\citep{2023CoT-theory}. \citet{2023TPTU} find that LLMs can plan to solve complex tasks using external SQL generators and Python generators, while CodePlan~\citep{CodePlan} demonstrates they can perform repository-level coding via adaptive planning.

Another stream of works use LLMs to create multi-agent systems for code generation, such as self-collaboration~\citep{2023self-collaboration}, ChatDev~\citep{2023ChatDev}, MetaGPT~\citep{2023MetaGPT}, LCG~\citep{2024LCG}, MAGIS~\citep{2024MAGIS}, and SoA~\citep{2024SoA}. In these frameworks, multiple LLMs are prompted to play distinct roles such as programmer, reviewer, and manager. These roles interact with each other, breakdown code generation into different phases (e.g. designing, coding, testing, and documenting), and collaborate to complete complex tasks.

\subsection{LLMs Integrated into Software Development}\label{sec:llm-integration}
With the increase in LLMs' interactive coding capability, researchers have also started to integrate them into each and every process of software development.

Auto code completion is one of the earliest applications of language models in software development, as they require only the ability to predict the next token. Even before language models scaled to billions of parameters, there had been integration of completion systems such as Pythia~\citep{completion2019-2} and IntelliCode~\citep{2020GPT-C} into popular IDEs.

Recently, however, the application of code language models have transcended simple code completion. GitHub Copilot is arguably one of the most popular AI code assistants, with diverse features including code generation, vulnerability detection, and license management\footnote{\url{https://github.com/features/copilot}}, while CodeFuse~\citep{2023CodeFuse13B} also integrates code generation, code translation, code commenting, and testcase generation into a single IDE extension. As code language models become larger, however, their client-side deployment and real-time performance also raise new challenges.

As LLMs continue to advance, building applications on top of them is also evolving into a consequential task itself. Many open-source frameworks for such applications have been released, including LangChain\footnote{\url{https://www.langchain.com/}}, AutoGPT\footnote{\url{https://github.com/Significant-Gravitas/AutoGPT}}, and WorkGPT\footnote{\url{https://github.com/team-openpm/workgpt}}. These frameworks provide abstractions over language models for developers, and are actively revolutionizing the entire process of software development even as this survey is being finalized.

\subsection{Analysis of LLM-Generated Code}
As AI code assistants become prevalent, many recent works have also focused on examining AI-generated code from different aspects, including correctness~\citep{analysis2022-2,analysis2023-2}, bugs~\citep{analysis2023-1,analysis2024-1}, vulnerabilities~\citep{analysis2022-1,analysis2022-3,analysis2022-4,analysis2024-2}, syntactic robustness~\citep{analysis2024-4}, efficiency~\citep{analysis2024-5}, and hallucinations~\citep{analysis2024-3}. \citet{analysis2022-1} and \citet{analysis2022-3}'s studies show that AI code assistants do not introduce extra security risks compared with human programmers, and \citet{analysis2024-2} also find that ChatGPT's responses to security-related questions contain less vulnerabilities than answers from StackOverflow. Contrarily, \citet{analysis2022-4} find that users write significantly less secure code when assisted by AI. In terms of code complexity, \citet{analysis2022-2} find GitHub Copilot to generate low-complexity code with no statistically significant differences between languages, whereas \citet{analysis2023-3} find ChatGPT to generate the most complex code in C and the least complex code in Python. Overall, AI code assistants are still in their nascent stage and constantly evolving, and their implications on software development are yet to be investigated systematically.

%% file: sec_conclusion.tex
\section{Conclusion and Challenges}\label{sec:conclusion}
In this work, we systematically reviewed the history of pretrained Transformer language models in code processing and other software engineering tasks. The advancement in code modeling generally follows the history course of NLP, evolving from SMT models, to NMT models, and then to finetuning pretrained Transformers and lastly to few-shot application of LLMs and even autonomous agents in real-world production. Unlike natural languages, the nature of code makes it easy to extract auxiliary information from alternative views, and to utilize interpreter and unit tests for automatic feedback.

With these in mind, we identify several challenges in the current development of code modeling.

- \textbf{More comprehensive and challenging benchmarks}. The widely used HumanEval benchmark plays a key role in the evolution of Code LLMs. However, it is relatively small and its scoreboard has been manipulated to near perfect, and the community is eager for a new standard to evaluate LLMs. HumanEval and other similar benchmarks focus on generating standalone Python functions from well-structured docstrings, which do not reflect real-world user behaviors. In real-world scenarios, software requirements are seldom condensed into a single docstring, and LLMs must learn to communicate effectively for clarifications when uncertain about the requirements~\citep{2024communication}. Also, standalone functions are hardly used in production, where software has complex dependencies within a repository. Thus, next-generation benchmarks should also take such cross-file context into account, and recent benchmarks such as SWE-bench~\citep{2023SWE-bench} represent promising efforts in this direction. Contrary to the models' perfect scores on HumanEval, these repository-level benchmarks expose LLMs' limits in real-world applications, with the most recent SWE-agent resolving only 12.5\% real-world GitHub issues in SWE-bench~\citep{2024SWE-agent}.

% - \textbf{Building code LLM ecosystem for full-life-cycle of software development}. While the academia have witnessed an abundance of code models, most have been deployed in the coding stage as IDE plugins while neglecting other stages in the life-cycle of software development. In Section~\ref{sec:llm-integration} we mentioned several inspiring examples, and we are hoping to see more applications of code LMs throughout the full life-cycle of software development, from requirement engineering to DevOps, eventually leading to full-scale ecosystems like those around PyTorch~\citep{2019torch} and Hugging Face\footnote{\url{https://huggingface.co/}}.

- \textbf{Evaluation and application of LLMs beyond traditional code generation}. As we have pointed out in Section~\ref{sec:evaluation}, current evaluation of LLMs' coding capability in the NLP community is focused on code generation, while overlooking other activities in software engineering, such as software modeling and testing. From an application point-of-view, the most widespread application of LLMs in software engineering is IDE plugins, which provide developers with code suggestions, while other stages in software development that we mentioned in Section~\ref{sec:evaluation} are also largely overlooked. However, it should be also noted that LLMs trained on code are highly specialized to reasoning-heavy tasks such as programming and math, and non-coding tasks in SE - such as requirement elicitation, product management, and marketing - are probably better suited for general-domain models. Beyond the previously mentioned text-based evaluation and applications, the recent advancement of multimodel LLMs also created new opportunities, especially for UI design and other activities that involve visual input, leading to novel tasks such as visually grounded code generation~\citep{2024MMCode,2024Plot2Code}, webpage reverse engineering~\citep{UI-data-2024-1,UI-data-2024-2}, and layout design~\citep{UI-CV3}. 

- \textbf{Acquisition of high-quality data}. With \citet{2023Phi-1} achieving SOTA performance with a 1.3B model trained on textbook data, we believe the selection of training data and utilization of synthetic data will be ever more prominent in the near future, for both self-supervised pretraining and supervised finetuning.

- \textbf{Integration of code features into language models}. As we noted in Section~\ref{sec:CFGDFG}, CFG and DFG are yet to be employed at scale in code language modeling. The few works that do employ data flow make changes to the models' attention masks, which severely limits their cross-task generalization and scaling ability. We believe the seamless integration of such features into textual input is worth researching in the future.

- \textbf{Beyond the imperative programming paradigm}. In Section~\ref{sec:evaluation-low-resource} we have shown that most of the current research on code LLM focus on popular imperative programming languages such as C and Python, while neglecting declarative and functional languages except SQL. However, the paradigm of declarative and functional languages makes them more aligned to natural languages, and thus is worth more research attention in the future.

- \textbf{Alternative model architectures and training objectives}. In Table~\ref{tab:models}, we have shown that many code language models are pretrained with auxiliary objectives specific to code, but these models all belong to the encoder-only or encoder-decoder family, while decoder-only models are yet to be augmented with alternative objectives. Also, as pioneered by \citet{2023codefusion}, we believe diffusion models will find its ground in code modeling in the future.

- \textbf{Safety and ethics issues related to code LLMs}. As language models grow in might, they also raise safety concerns including but not limited to data contamination, toxic or biased generation, personal information leak, and hallucinations. In software development, these models should be deployed with extra caution, as their generated code may contain security risks leading to catastrophic results. Pretraining data is also becoming a sensitive topic of ethics, and \citet{2022Stack} take a meaningful step towards this issue by allowing developers to remove their code from the Stack. As synthetic training data becomes widespread, researchers should also proceed with caution about such practice, as the consequence of training AI models with AI generated data is yet to be investigated at scale.

With the presentation of this survey, we hope to provide a global view of language models' application in software engineering and connect the research from the two communities. We believe the current surge of LLMs will be ultimately transformed into real-world applications, and lead humanity into a brighter future.

%% file: appendix.tex
\newpage

\appendix
\section{Benchmarks for Downstrem Tasks}\label{sec:benchmark}
Table~\ref{tab:benchmark-1}, \ref{tab:benchmark-2}, \ref{tab:benchmark-3}, \ref{tab:benchmark-4}, \ref{tab:benchmark-5}, \ref{tab:benchmark-6} list benchmark datasets for code downstream tasks. For the size of these datasets, we use self-reported number whenever available. 
\begin{table}[h]
    \caption{Benchmarks for text-to-SQL generation.}
    \label{tab:benchmark-1}
    \centering
    % \adjustbox{width=\textwidth+1cm,center}{
    \begin{tabular}{m{2.4cm}llm{4.5cm}r>{\raggedleft\arraybackslash}m{2.7cm}}
    \toprule
        \textbf{Task} & \textbf{Date} & \textbf{Benchmark} & \textbf{Source} & \textbf{Size}\\
    \midrule
        \multirow{23}{2.4cm}{Text-to-SQL}
        & 1990 & ATIS & \citet{sql-data-1990,sql-data-1994} & 11508\\
        & 1996 & GeoQuery & \citet{sql-data-1996} & 877\\
        & 2000 & Restaurants & \citet{sql-data-2000} & 378\\
        & 2014-09 & MAS & \citet{sql-data-2014-1} & 196\\
        & 2017-02 & Yelp & \citet{sql-data-2017-1} & 128\\
        & 2017-02 & IMDb & \citet{sql-data-2017-1} & 131\\
        & 2017-04 & Scholar & \citet{sql-data-2017-2} & 816\\
        & 2017-08 & WikiSQL & \citet{sql2017-1} & 80654\\
        & 2018-06 & Advising & \citet{sql2018-4} & 4570\\
        & 2018-09 & Spider & \citet{sql2018-5} & 10181\\
        & 2019-06 & SParC & \citet{sql2019-4} & 12726\\
        & 2019-07 & MIMICSQL & \citet{sql-data-2019-1} & 10000\\
        & 2019-09 & CoSQL & \citet{sql2019-6} & 15598\\
        & 2020-05 & Criteria-to-SQL & \citet{sql-data-2020-2} & 2003\\
        & 2020-10 & Squall & \citet{sql-data-2020-1} & 11276\\
        & 2020-10 & Spider-Realistic & \citet{sql2020-4.5} & 508\\
        & 2021-06 & Spider-Syn & \citet{sql-data-2021-4} & 8034\\
        & 2021-06 & SEDE & \citet{sql-data-2021-2} & 12023\\
        & 2021-06 & KaggleDBQA & \citet{sql-data-2021-1} & 400\\
        & 2021-09 & Spider-DK & \citet{sql-data-2021-3} & 535\\
        & 2022-05 & Spider-SS & \citet{sql-data-2022-1} & 8034\\
        & 2022-05 & Spider-CG & \citet{sql-data-2022-1} & 45599\\
        & 2023-05 & BIRD & \citet{sql-data-2023-1} & 12751\\
    \bottomrule
    \end{tabular}
    % }
\end{table}

\begin{table}[]
    \caption{Benchmarks for program synthesis. JS is short for JavaScript. $^*$Automatically mined/human-annotated. $^\dagger$These are 1749 prompts for 48 problems. $^\ddagger$10538 prompts for 1420 methods. $^\diamond$Machine-generated/human-annotated prompts.}
    \label{tab:benchmark-2}
    \centering
    \adjustbox{width=\textwidth,center}{
    \begin{tabular}{m{2.4cm}llm{4.5cm}r>{\raggedleft\arraybackslash}m{2.7cm}}
    \toprule
        \textbf{Task} & \textbf{Date} & \textbf{Benchmark} & \textbf{Source} & \textbf{Size} & \textbf{Language}\\
    \midrule
        \multirow{26}{2.4cm}{Program Synthesis}
        & 2018-02 & NL2Bash & \citet{2018NL2Bash} & 9305 & Bash\\
        & 2018-08 & CONCODE & \citet{2018CONCODE} & 104K & Java\\
        & 2019-10 & JuICe & \citet{2019JuICe} & $^*$1.5M/3725 & Python\\
        & 2021-05 & APPS & \citet{2021APPS} & 10000 & Python\\
        & 2021-07 & HumanEval & \citet{2021Codex} & 164 & Python\\
        & 2021-08 & MBPP & \citet{2021MBPP} & 974 & Python\\
        & 2021-08 & MathQA-Python & \citet{2021MBPP} & 23914 & Python\\
        & 2021-08 & PlotCoder & \citet{2021PlotCoder} & 40797 & Python\\
        & 2022-01 & DSP & \citet{2022DSP} & 1119 & Python\\
        \cline{6-6}
        & 2022-02 & CodeContests & \citet{2022AlphaCode} & 13610 & C++, Python, Java\\
        \cline{6-6}
        & 2022-03 & MCoNaLa & \citet{2022MCoNaLa} & 896 & Python\\
        & 2022-06 & AixBench & \citet{2022AixBench} & 336 & Java\\
        & 2022-10 & MBXP & \citet{2022MBXP} & 12.4K & 13\\
        & 2022-10 & Multilingual HumanEval & \citet{2022MBXP} & 1.9K & 12\\
        & 2022-10 & MathQA-X & \citet{2022MBXP} & 5.6K & Python, Java, JS\\
        & 2022-11 & ExeDS & \citet{2022ExeDS} & 534 & Python\\
        & 2022-11 & DS-1000 & \citet{2022DS-1000} & 1000 & Python\\
        & 2022-12 & ODEX & \citet{2022ODEX} & 945 & Python\\
        & 2023-02 & CoderEval & \citet{2023CoderEval} & 460 & Python, Java\\
        & 2023-03 & xCodeEval & \citet{2023xCodeEval} & 5.5M & 11\\
        \cline{6-6}
        & 2023-03 & HumanEval-X & \citet{2023CodeGeeX} & 820 & Python, C++, Java, JS, Go\\
        \cline{6-6}
        & 2023-06 & StudentEval & \citet{2023StudentEval} & $^\dagger$1749 & Python\\
        & 2023-06 & DotPrompts & \citet{repo2023-1.5} & $^\ddagger$10538 & Java\\
        & 2023-09 & CodeApex & \citet{2023CodeApex} & 476 & C++\\
        & 2023-09 & VerilogEval & \citet{2023VerilogEval} & $^\diamond$8645/156 & Verilog\\
        & 2023-11 & ML-Bench & \citet{2023ML-Bench} & 10040 & Bash\\
        \cline{6-6}
        & 2024-01 & ParEval & \citet{2024ParEval} & 420 & C++, HIP, CUDA\\
        \cline{6-6}
        & 2024-04 & MMCode & \citet{2024MMCode} & 3548 & Python \\
        & 2024-04 & PECC & \citet{2024PECC} & 2396 & Python\\
        & 2024-05 & NaturalCodeBench & \citet{2024NaturalCodeBench} & 402 & Python, Java\\
        & 2024-05 & Plot2Code & \citet{2024Plot2Code} & 132 & Python \\
        & 2024-05 & MHPP & \citet{2024MHPP} & 140 & Python \\
    \bottomrule
    \end{tabular}
    }
\end{table}

\begin{table}[]
    \caption{Benchmarks for code translation and program repair. JS is short for JavaScript. $^*$These are pairwise sample counts. For example, HumanEval-X includes 164 programs, each implemented in 5 languages, totaling 164 $\times$ (5 $\times$ 4 / 2) = 1640 translation pairs. $^\dagger$These are code change datasets, and only a subset therein concern bug-fixing.}
    \label{tab:benchmark-3}
    \centering
    \adjustbox{width=\textwidth,center}{
    \begin{tabular}{m{2.4cm}llm{4.5cm}r>{\raggedleft\arraybackslash}m{2.7cm}}
    \toprule
        \textbf{Task} & \textbf{Date} & \textbf{Benchmark} & \textbf{Source} & \textbf{Size} & \textbf{Language}\\
    \midrule
        \multirow{16}{2.4cm}{Code Translation}& 2020-06 & GeeksforGeeks & \citet{2020Transcoder} & 1.4K & C++, Java, Python\\
        & 2021-02 & CodeTrans & \citet{2021CodeXGLUE} & 11.8K & Java, C\#\\
        & 2021-08 & Avatar & \citet{trans-data-2021-1} & 9515 & Java, Python\\
        \cline{6-6}
        & 2022-06 & CoST & \citet{trans-data-2022-2} & $^*$132K & C++, Java, Python, C\#, JS, PHP, C\\
        \cline{6-6}
        & 2022-06 & XLCoST & \citet{trans-data-2022-1} & $^*$567K & C++, Java, Python, C\#, JS, PHP, C\\
        \cline{6-6}
        & 2023-03 & xCodeEval & \citet{2023xCodeEval} & $^*$5.6M & 11\\
        \cline{6-6}
        & 2023-03 & HumanEval-X & \citet{2023CodeGeeX} & $^*$1640 & Python, C++, Java, JS, Go\\
        \cline{6-6}
        & 2023-08 & G-TransEval & \citet{trans-survey-2023-1} & $^*$4000 & C++, Java, C\#, JS, Python\\
        \cline{6-6}
        & 2023-10 & CodeTransOcean & \citet{trans-data-2023-1} & 270.5K & 45\\
    \midrule
        \multirow{31}{2.4cm}{Program Repair}
        & 2014-07 & Defects4J & \citet{fix-data-2014-1} & 357 & Java\\
        & 2015-12 & ManyBugs & \citet{fix-data-2015-1} & 185 & C\\
        & 2015-12 & IntroClass & \citet{fix-data-2015-1} & 998 & C\\
        & 2016-11 & BugAID & \citet{fix-data-2016-1} & 105K & JS\\
        & 2017-02 & DeepFix & \citet{fix2017-1} & 6971 & C\\
        & 2017-05 & Codeflaws & \citet{fix-data-2017-1} & 3902 & C\\
        & 2017-10 & QuixBugs & \citet{fix-data-2017-2} & 80 & Java, Python\\
        & 2018-05 & Bugs.jar & \citet{fix-data-2018-1} & 1158 & Java\\
        & 2018-12 & BFP & \citet{fix2018-2} & 124K & Java\\
        & 2019-01 & Bears & \citet{fix-data-2019-3} & 251 & Java\\
        & 2019-01 & unnamed & \citet{fix2019-1} & $^\dagger$21.8K & Java\\
        & 2019-04 & BugsJS & \citet{fix-data-2019-5} & 453 & JS\\
        & 2019-05 & BugSwarm & \citet{fix-data-2019-4} & 1827/1264 & Java/Python\\
        & 2019-05 & CPatMiner & \citet{fix-data-2019-6} & $^\dagger$17K & Java\\
        & 2019-05 & ManySStuBs4J & \citet{fix-data-2019-2} & 154K & Java\\
        & 2019-11 & Refactory & \citet{fix-data-2019-1} & 1783 & Python\\
        \cline{6-6}
        & 2020-07 & CoCoNut & \citet{fix2020-2} & 24M & Java, JS, C, Python\\
        \cline{6-6}
        & 2020-10 & Review4Repair & \citet{fix2020-4} & 58021 & Java\\
        & 2020-11 & BugsInPy & \citet{fix-data-2020-1} & 493 & Python\\
        & 2021-07 & TFix & \citet{fix2021-4.5} & 105K & JS\\
        & 2021-08 & Megadiff & \citet{fix-data-2021-1} & $^\dagger$663K & Java\\
        & 2022-01 & SSB/TSSB & \citet{fix-data-2022-3} & 9M/3M & Python\\
        & 2022-10 & FixJS & \citet{fix-data-2022-2} & 324K & JS\\
        & 2022-11 & TypeBugs & \citet{fix-data-2022-1} & 93 & Python\\
        & 2023-03 & xCodeEval & \citet{2023xCodeEval} & 4.7M & 11\\
        \cline{6-6}
        & 2023-04 & RunBugRun & \citet{fix-data-2023-1} & 450K & C, C++, Java, Python, JS, Ruby, Go, PHP\\
        \cline{6-6}
        & 2023-08 & HumanEvalPack & \citet{2023OctoPack} & 984 & Python, JS, Go, Java, C++, Rust\\
        \cline{6-6}
        & 2024-01 & DebugBench & \citet{2024DebugBench} & 4253 & C++, Java, Python\\
    \bottomrule
    \end{tabular}
    }
\end{table}

\begin{table}[]
    \caption{Benchmarks for code summarization, code completion, commit message generation, and defect/vulnerability detection. JS is short for JavaScript. $^*$The task of code completion can be evaluated on any source code corpus, so we only list a few widely used benchmarks here. For cross-file code completion please refer to Table~\ref{tab:benchmark-6}. $^\dagger$With/without verb-direct object filter.}
    \label{tab:benchmark-4}
    \centering
    \adjustbox{width=\textwidth,center}{
    \begin{tabular}{m{2.4cm}llm{4.5cm}r>{\raggedleft\arraybackslash}m{2.7cm}}
    \toprule
        \textbf{Task} & \textbf{Date} & \textbf{Benchmark} & \textbf{Source} & \textbf{Size} & \textbf{Language}\\
    \midrule
        \multirow{10}{2.4cm}{Code Summarization}& 2016-08 & CODE-NN & \citet{sum2016-1} & 66K/32K & C\#/SQL\\
        & 2017-07 & unnamed & \citet{sum2017-1} & 150K & Python\\
        & 2018-05 & DeepCom & \citet{sum2018-1} & 588K & Java\\
        & 2018-07 & TL-CodeSum & \citet{sum2018-2} & 411K & Java\\
        & 2018-11 & unnamed & \citet{sum2018-3} & 109K & Python\\
        & 2019-02 & unnamed & \citet{sum2019-0.5} & 2.1M & Java\\
        \cline{6-6}
        & 2019-09 & CodeSearchNet & \citet{2019CodeSearchNet} & 2.3M & Go, JS, Python, PHP, Java, Ruby\\
        \cline{6-6}
        & 2023-08 & HumanEvalPack & \citet{2023OctoPack} & 984 & Python, JS, Go, Java, C++, Rust\\
    \midrule
        \multirow{4}{2.4cm}{$^*$Code Completion}& 2013-05 & GitHub Java Corpus & \citet{completion-data-2013-1} & 2.1M & Java\\
        & 2016-10 & Py150 & \citet{completion2016-3} & 150K & Python\\
        & 2016-10 & JS150 & \citet{completion2016-3} & 150K & JS\\
        & 2023-06 & LCC & \citet{2023LongCoder} & 360K & Python, Java, C\#\\
    \midrule
        \multirow{15}{2.4cm}{Commit Message Generation}
        & 2017-03 & unnamed & \citet{commit-data-2017-1} & 509K & Java\\
        \cline{6-6}
        & 2017-04 & CommitGen & \citet{commit2017-1} & 153K & Python, C++, Java, JS\\
        \cline{6-6}
        & 2017-08 & CommitGen & \citet{commit2017-2} & $^\dagger$32K/75K & Java\\
        & 2018-09 & NNGen & \citet{commit2018-1} & 27K & Java\\
        & 2019-05 & PtrGNCMsg & \citet{commit2019-1} & 64.9K & Java\\
        & 2019-08 & CoDiSum & \citet{commit2019-2} & 90.7K & Java\\
        & 2019-12 & ATOM & \citet{commit2019-3} & 160K & Java\\
        \cline{6-6}
        & 2021-05 & CommitBERT & \citet{commit2021-1} & 346K & Python, PHP, Go, Java, JS, Ruby\\
        \cline{6-6}
        & 2021-07 & MCMD & \citet{commit-survey-2021} & 2.25M & Java, C\#, C++, Python, JS\\
        \cline{6-6}
        & 2021-07 & CoRec & \citet{commit-data-2021-1} & 107K & Java\\
        & 2023-07 & ExGroFi & \citet{commit-data-2023-1} & 19263 & Java\\
        & 2023-08 & CommitChronicle & \citet{commit-data-2023-2} & 10.7M & 20\\
    \midrule
        \multirow{19}{2.4cm}{Defect (Vulnerability) Detection}& 2018-01 & CGD & \citet{defect2018-1} & 62K & C, C++\\
        & 2018-07 & Draper VDISC & \citet{defect2018-3} & 12.8M & C, C++\\
        & 2018-07 & SySeVR & \citet{defect2018-3.5} & 15591 & C, C++\\
        & 2018-04 & unnamed & \citet{defect-data-2018-1} & 32988 & C, C++\\
        & 2019-02 & unnamed & \citet{defect-data-2019-1} & 624 & Java\\
        & 2019-09 & Devign & \citet{2019Devign} & 48687 & C\\
        & 2019-11 & unnamed & \citet{defect-data-2019-2} & 170K & C, C++\\
        & 2019-12 & GREAT & \citet{fix2019-3} & 2.8M & Python\\
        & 2020-01 & MVD & \citet{defect-data-2020-3} & 182K & C, C++\\
        & 2020-02 & unnamed & \citet{defect-data-2020-2} & 1471 & C\\
        & 2020-09 & ReVeal & \citet{defect2020-1} & 18K & C\\
        & 2020-09 & Big-Vul & \citet{defect-data-2020-1} & 265K & C, C++\\
        & 2021-02 & D2A & \citet{defect-data-2021-1} & 1.3M & C, C++\\
        & 2021-05 & PyPIBugs & \citet{defect2021-0.3} & 2374 & Python\\
        & 2021-07 & CVEfixes & \citet{defect-data-2021-2} & 5495 & 27\\
        & 2021-08 & CrossVul & \citet{defect-data-2021-3} & 27476 & 40+\\
        & 2023-04 & DiverseVul & \citet{defect-data-2023-1} & 349K & C, C++\\
        & 2023-06 & VulnPatchPairs & \citet{defect-data-2023-2} & 26K & C\\
        & 2023-11 & VulBench & \citet{defect2023-4} & 455 & C\\
    \bottomrule
    \end{tabular}
    }
\end{table}

\begin{table}[]
    \caption{Benchmarks for code retrieval, code reasoning, type inference, and clone detection/code search,. JS is short for JavaScript. $^*$These benchmarks include a large number of automatically constructed samples, and a small set of human-annotated samples. $^\dagger$These are general-domain reasoning benchmarks, and only a subset therein concern programming, algorithms, and other topics related to computer science. $^\ddagger$These are project counts (or, in the case of \citet{type2023-3}, file counts). \citet{type2023-1} propose to measure project-level type check rate instead of type prediction accuracy for TypeScript. $^\diamond$These are 21K/24K Java/Python methods for 576 programming problems.} 
    \label{tab:benchmark-5}
    \centering
    \adjustbox{width=\textwidth,center}{
    \begin{tabular}{m{2.4cm}llm{4.5cm}r>{\raggedleft\arraybackslash}m{2.7cm}}
    \toprule
        \textbf{Task} & \textbf{Date} & \textbf{Benchmark} & \textbf{Source} & \textbf{Size} & \textbf{Language}\\
    \midrule
        \multirow{12}{2.4cm}{Code Retrieval}& 2018-03 & StaQC  & \citet{retrieval-data-2018-1} & 268K & Python, SQL\\
        & 2018-05 & DeepCS & \citet{ retrieval2018-1} & 16M & Java\\
        & 2018-05 & CoNaLa & \citet{ retrieval2018-2} & $^*$600K/2.9K & Python\\
        & 2019-08 & unnamed & \citet{retrieval-data-2019-1} & 287 & Java\\
        \cline{6-6}
        & 2019-09 & CodeSearchNet & \citet{2019CodeSearchNet} & $^*$2.3M/99 & Go, JS, Python, PHP, Java, Ruby\\
        \cline{6-6}
        & 2020-02 & CosBench & \citet{retrieval-data-2020-1} & 52 & Java\\
        & 2020-08 & SO-DS & \citet{retrieval2020-0.4} & 2.2K & Python\\
        & 2020-10 & FB-Java & \citet{ retrieval2020-1} & 249K & Java\\
        & 2021-02 & AdvTest & \citet{2021CodeXGLUE} & 280K & Python\\
        & 2021-02 & WebQueryTest & \citet{2021CodeXGLUE} & 1K & Python\\
        & 2021-05 & CoSQA & \citet{ retrieval2021-2} & 21K & Python\\
        \cline{6-6}
        & 2024-03 & ProCQA & \citet{2024ProCQA} & 5.2M & C, C++, Java, Python, Ruby, Lisp, JS, C\#, Go, Rust, PHP\\
    \midrule
        \multirow{6}{2.4cm}{Code Reasoning}& 
        2020-09 & MMLU & \citet{2020MMLU} & $^\dagger$15908\\
        & 2021-09 & CodeQA & \citet{2021CodeQA} & 120K/70K & Java/Python\\
        & 2022-10 & CS1QA & \citet{2022CS1QA} & 9237\\
        & 2023-05 & C-Eval & \citet{2023C-Eval} & $^\dagger$13948\\
        & 2023-06 & CMMLU & \citet{2023CMMLU} & $^\dagger$11528\\
        & 2023-09 & CodeApex & \citet{2023CodeApex} & 250 & C++\\
        & 2024-01 & CRUXEval & \citet{2024CRUXEval} & 800 & Python\\
        & 2024-05 & PythonIO & \citet{2024PythonIO} & 2650 & Python\\
    \midrule
        \multirow{9}{2,4cm}{Type Inference}& 2019-12 & TypeWriter OSS & \citet{type2019-4} & 208K & Python\\
        & 2020-04 & Typilus & \citet{type2020-2} & 252K & Python\\
        & 2020-04 & LambdaNet & \citet{type2020-3} & $^\ddagger$300 & TypeScript\\
        & 2021-04 & ManyTypes4Py & \citet{type-data-2021-1} & 869K & Python\\
        & 2022-10 & ManyTypes4TypeScript & \citet{type-data-2022-1} & 9.1M & TypeScript\\
        & 2023-02 & TypeWeaver & \citet{type2023-1} & $^\ddagger$513 & TypeScript\\
        & 2023-03 & BetterTypes4Py & \citet{type2023-2} & 608K & Python\\
        & 2023-03 & InferTypes4Py & \citet{type2023-2} & 4.6K & Python\\
        & 2023-05 & OpenTau & \citet{type2023-3} & $^\ddagger$744 & TypeScript\\
    \midrule
        \multirow{5}{2.4cm}{Clone Detection / Code Search}& 2014-09 & BigCloneBench & \citet{2014BigCloneBench} & 6M & Java\\
        & 2014-09 & POJ-104 & \citet{clf2014-1} & 52K & C, C++\\
        & 2019-05 & unnamed & \citet{clone2019-2.5} & $^\diamond$21K/24K & Java/Python\\
        \cline{6-6}
        & 2019-11 & CLCDSA & \citet{clone-data-2019-1} & 78K & Java, C\#, Python\\
    \bottomrule
    \end{tabular}
    }
\end{table}

\begin{table}[]
    \caption{Benchmarks for unit test/assertion generation, log parsing, and repository level coding. $^*$LogHub (2023) is an annotated subset of LogHub (2018). $^\dagger$Line Completion/API Invocation Completion/Function Completion. $^\ddagger$Retrieval/Completion/Pipeline. $^\star$File count. $^\diamond$Migration/Temporal Edit.}
    \label{tab:benchmark-6}
    \centering
    \adjustbox{width=\textwidth,center}{
    \begin{tabular}{m{2.4cm}llm{4.5cm}r>{\raggedleft\arraybackslash}m{2.7cm}}
    \toprule
        \textbf{Task} & \textbf{Date} & \textbf{Benchmark} & \textbf{Source} & \textbf{Size} & \textbf{Language}\\
    \midrule
        \multirow{3}{2.4cm}{Unit Test / Assertion Generation}
        & 2014-12 & SF110 & \citet{unit-data-2014-1} & 24K & Java\\
        & 2020-09 & Method2Test & \citet{unit2020-1} & 781K & Java\\
        & 2021-08 & ConTest & \citet{assert-data-2021-1} & 365K & Java\\
    \midrule
        \multirow{2}{2.4cm}{Log Parsing}& 2018-11 & LogHub (2018) & \citet{log-survey-2018}; \citet{log-data-2020-1}& 379M\\
        & 2023-08 & LogHub (2023) & \citet{log-data-2023-1} & $^*$50.4M\\
    \midrule
        \multirow{8}{2.4cm}{Repository-Level Coding}& 2023-03 & RepoEval & \citet{repo2023-1} & $^\dagger$1600/1600/373 & Python\\
        & 2023-06 & RepoBench & \citet{2023RepoBench} & $^\ddagger$890K/9M/43K & Python, Java\\
        & 2023-06 & PragmaticCode & \citet{repo2023-1.5} & $^\star$880 & Java\\
        & 2023-06 & Stack-Repo  & \citet{repo2023-2} & 816K & Java\\
        & 2023-09 & CodePlan & \citet{CodePlan} & $^\diamond$645/21 & $^\diamond$C\#/Python\\
        & 2023-10 & SWE-Bench & \citet{2023SWE-bench} & 2294 & Python\\
        \cline{6-6}
        & 2023-10 & CrossCodeEval & \citet{2023CrossCodeEval} & 9928 & Python, Java, TypeScript, C\#\\
        \cline{6-6}
        & 2024-03 & EvoCodeBench & \citet{2024EvoCodeBench} & 275 & Python \\
    \bottomrule
    \end{tabular}
    }
\end{table}

\newpage
\phantom{.}
\newpage
\section{Results on Downstream Benchmarks}\label{sec:performance}
For the benchmarks listed in Appendix~\ref{sec:benchmark}, we also list code language models' reported performance on some of the most widely used ones, including:
\begin{itemize}
    \item NL-code search (Table~\ref{tab:perf-search}): CodeSearchNet~\citep{2019CodeSearchNet}, CosQA~\citep{retrieval2021-2}, and AdvTest~\citep{2021CodeXGLUE}, all measured by Mean Reciprocal Rank (MRR);
    \item Clone detection (Table~\ref{tab:perf-clone}): BigCloneBench~\citep{2014BigCloneBench}, measured by F1, and POJ-104~\citep{clf2014-1}, measured by Mean Average Precision; the train/test splits of both two datasets are provided by \citet{2021CodeXGLUE};
    \item Defect detection (Table~\ref{tab:perf-defect}): Devign~\citep{2019Devign}, measured by accuracy; the train/test split is provided by \citet{2021CodeXGLUE};
    \item Code summarization (Table~\ref{tab:perf-summarization}): CodeSearchNet~\citep{2019CodeSearchNet}, measured by BLEU; the train/test split is provided by \citet{2021CodeXGLUE};
    \item Code translation (Table~\ref{tab:perf-translation1}, \ref{tab:perf-translation2}): CodeTrans~\citep{2021CodeXGLUE}, measured by BLEU and CodeBLEU, and Transcoder~\citep{2020Transcoder}, measured by pass@1;
    \item Code reasoning (Table~\ref{tab:perf-reasoning1}, \ref{tab:perf-reasoning2}): CRUXEval~\citep{2024CRUXEval}, measured by pass@1, and PythonIO, measured by accuracy.
\end{itemize}

\input{tables/search}

\input{tables/clone}

\input{tables/defect}

\input{tables/summarization}

\input{tables/translation}

\input{tables/reasoning}

%% file: tables/search.tex
\begin{table}[h]
    \centering
    \caption{Mean Reciprocal Rank of NL-code search performance on CodeSearchNet, AdvTest, and CosQA.}
    \label{tab:perf-search}
    \adjustbox{width=1\textwidth,center}{
    \begin{tabular}{lccccccc|cc|l}
\toprule
 & Ruby & JS & Go & Python & Java & PHP & Average & AdvTest & CosQA & source \\
\midrule
RoBERTa & 58.7 & 51.7 & 85.0 & 58.7 & 59.9 & 56.0 & 61.7 & 18.3 & 60.3 & \citet{2021CodeXGLUE}$\downarrow$\\
CodeBERT & 67.9 & 62.0 & 88.2 & 67.2 & 67.6 & 62.8 & 69.3 & 27.2 & 65.7 & \\
GraphCodeBERT & 70.3 & 64.4 & 89.7 & 69.2 & 69.1 & 64.9 & 71.3 & 35.2 & 68.4 & \citet{2020GraphCodeBERT}\\
SynCoBERT & 72.2 & 67.7 & 91.3 & 72.4 & 72.3 & 67.8 & 74.0 & 38.1 & 69.6 & \citet{2021SynCoBERT}\\
SPT-Code & 70.1 & 64.1 & 89.5 & 69.9 & 70.0 & 65.1 & 71.5 &  &  & \citet{2022SPT-Code}\\
CodeRetriever & 77.1 & 71.9 & 92.4 & 75.8 & 76.5 & 70.8 & 77.4 & 46.9 & 75.4 & \citet{retrieval2022-2}\\
UniXcoder & 74.0 & 68.4 & 91.5 & 72.0 & 72.6 & 67.6 & 74.4 & 41.3 & 70.1 & \citet{2022UniXcoder}$\downarrow$\\
PLBART & 67.5 & 61.6 & 88.7 & 66.3 & 66.3 & 61.1 & 68.5 & 34.7 & 65.0 & \\
CodeT5$_{base}$ & 71.9 & 65.5 & 88.8 & 69.8 & 68.6 & 64.5 & 71.5 & 39.3 & 67.8 & \\
CoCoSoDa & 81.8 & 76.4 & 92.1 & 75.7 & 76.3 & 70.3 & 78.8 &  &  & \citet{retrieval2022-4}\\
Code-MVP &  &  &  &  &  &  &  & 40.4 & 72.1 & \citet{2022Code-MVP}\\
SCodeR & 77.5 & 72.0 & 92.7 & 74.2 & 74.8 & 69.2 & 76.7 & 45.5 & 74.5 & \citet{2022SCodeR}\\
CodeT5+$_{large}$ & 78.0 & 71.3 & 92.7 & 75.8 & 76.2 & 70.1 & 77.4 & 44.7 & 74.0 & \citet{2023CodeT5+}$\downarrow$\\
CodeGen-multi 350M & 66.0 & 62.2 & 90.0 & 68.6 & 70.1 & 63.9 & 70.1 & 34.8 & 64.8 & \\
\bottomrule
    \end{tabular}
    }
\end{table}

%% file: tables/clone.tex
\begin{table}[]
    \centering
    \caption{Clone detection performance on BigCloneBench (F1) and POJ-104 (Mean Average Precision). Works in the two blocks use different implementations and report different baseline scores.}
    \label{tab:perf-clone}
    % \adjustbox{width=1\textwidth,center}{
    \begin{tabular}{lccl}
\toprule
 & BigCloneBench (F1) & POJ-104 (MAP) & source \\
\midrule
RoBERTa & 94.9 & 79.96 & \citet{2021CodeXGLUE}$\downarrow$\\
CodeBERT & 96.5 & 84.29 & \\
DOBF & 96.5 &  & \citet{2021DOBF}\\
PLBART & 97.2 &  & \citet{2021PLBART}\\
GraphCodeBERT & 97.1 & 85.16 & \citet{2021SynCoBERT}$\downarrow$\\
SynCoBERT & 97.4 & 88.24 & \\
CodeT5$_{base}$ & 97.2 &  & \citet{2021CodeT5}\\
\midrule
RoBERTa & 91.3 & 76.67 & \citet{2022UniXcoder}\\
CodeBERT & 94.1 & 82.67 & \citet{2020GraphCodeBERT}$\downarrow$\\
GraphCodeBERT & 95.0 & 85.16 & \\
DISCO & 94.4 & 83.32 & \citet{2021DISCO}\\
PLBART & 93.6 & 86.27 & \citet{2022UniXcoder}$\downarrow$\\
CodeT5$_{base}$ & 95.0 & 88.65 & \\
UniXcoder & 95.2 & 90.52 & \\
SCodeR & 95.3 & 92.45 & \citet{2022SCodeR}\\
\bottomrule
    \end{tabular}
    % }
\end{table}

%% file: tables/defect.tex
\begin{table}[]
    \centering
    \caption{Accuracy of defect detection on Devign. pt: pretrained. ft: finetuned.}
    \label{tab:perf-defect}
    % \adjustbox{width=1\textwidth,center}{
    \begin{tabular}{lccccl}
\toprule
 & Accuracy & pt & ft & prompt & source \\
\midrule
Transformer & 61.6 &  & \Checkmark & - & \citet{2021PLBART}\\
RoBERTa & 61.0 & \Checkmark & \Checkmark & - & \citet{2021CodeXGLUE}$\downarrow$\\
CodeBERT & 62.1 & \Checkmark & \Checkmark & - & \\
PLBART & 63.2 & \Checkmark & \Checkmark & - & \citet{2021PLBART}\\
GraphCodeBERT & 63.2 & \Checkmark & \Checkmark & - & \citet{2021SynCoBERT}$\downarrow$\\
SynCoBERT & 64.5 & \Checkmark & \Checkmark & - & \\
CodeT5$_{base}$ & 65.8 & \Checkmark & \Checkmark & - & \citet{2021CodeT5}\\
DISCO & 64.4 & \Checkmark & \Checkmark & - & \citet{2021DISCO}\\
VulBERTa & 64.8 & \Checkmark & \Checkmark & - & \citet{defect2022-1}\\
Instruct-CodeGen 16B & 47.8 & \Checkmark &  & 0-shot & \citet{defect2023-1}$\downarrow$\\
CodeAlpaca 7B & 51.9 & \Checkmark &  & 0-shot & \\
Vicuna 7B & 54.0 & \Checkmark &  & 0-shot & \\
WizardCoder 15B & 54.4 & \Checkmark &  & 0-shot & \\
\bottomrule
    \end{tabular}
    % }
\end{table}

%% file: tables/summarization.tex
\begin{table}[]
    \centering
    \caption{BLEU scores of code summarization on CodeSearchNet, using the CodeXGLUE split (top) and the original split (bottom). pt: pretrained. ft: finetuned.}
    \label{tab:perf-summarization}
    \adjustbox{width=1\textwidth,center}{
    \begin{tabular}{lccccccc|cccl}
\toprule
 & Ruby & JS & Go & Python & Java & PHP & Average & pt & ft & prompt & source \\
\midrule
Transformer & 11.18 & 11.59 & 16.38 & 15.81 & 16.26 & 22.12 & 15.56 &  & \Checkmark & - & \citet{2021CodeXGLUE}$\downarrow$\\
RoBERTa & 11.17 & 11.90 & 17.72 & 18.14 & 16.47 & 24.02 & 16.57 & \Checkmark & \Checkmark & - & \\
CodeBERT & 12.16 & 14.90 & 18.07 & 19.06 & 17.65 & 25.16 & 17.83 & \Checkmark & \Checkmark & - & \\
PLBART & 14.11 & 15.56 & 18.91 & 19.30 & 18.45 & 23.58 & 18.32 & \Checkmark & \Checkmark & - & \citet{2021PLBART}\\
CodeT5$_{base}$ & 15.69 & 16.24 & 19.76 & 20.36 & 20.46 & 26.09 & 19.77 & \Checkmark & \Checkmark & - & \citet{2021CodeT5}\\
UniXcoder & 14.87 & 15.85 & 19.07 & 19.13 & 20.31 & 26.54 & 19.30 & \Checkmark & \Checkmark & - & \citet{2022UniXcoder}\\
InCoder &  &  &  & 18.27 &  &  &  & \Checkmark &  & 0-shot & \citet{2022InCoder}\\
NatGen & 15.38 & 16.00 & 19.43 & 20.09 & 20.38 & 26.00 & 19.55 & \Checkmark & \Checkmark & - & \citet{2022NatGen}\\
CodeT5+$_{large}$ & 15.63 & 17.93 & 19.64 & 20.47 & 20.83 & 26.39 & 20.15 & \Checkmark & \Checkmark & - & \citet{2023CodeT5+}\\
StarCoder &  &  &  & 21.99 &  &  &  & \Checkmark &  & 0-shot & \citet{2023StarCoder}\\
ChatGPT &  &  &  & 10.28 &  &  &  & \Checkmark &  & 0-shot & \citet{sum2023-3}\\
PaLM 2 &  &  &  & 19.23 &  &  &  & \Checkmark &  & few-shot & \citet{sum2024-1}$\downarrow$\\
LLaMA 2 70B &  &  &  & 22.41 &  &  &  & \Checkmark &  & few-shot & \\
\midrule
Mastropaolo et al. & 7.8 & 9.0 & 15.3 & 9.7 & 13.5 & 18.4 & 12.3 & \Checkmark & \Checkmark & - & \citet{2022SPT-Code}$\downarrow$\\
CugLM & 7.4 & 10.0 & 17.0 & 11.4 & 13.2 & 17.8 & 12.8 & \Checkmark & \Checkmark & - & \\
TreeBERT & 7.4 & 10.3 & 17.1 & 11.1 & 13.8 & 18.0 & 12.9 & \Checkmark & \Checkmark & - & \\
CodeBERT & 8.3 & 10.0 & 16.0 & 10.7 & 13.6 & 20.1 & 13.1 & \Checkmark & \Checkmark & - & \\
GraphCodeBERT & 8.3 & 10.9 & 16.5 & 11.0 & 14.5 & 20.0 & 13.5 & \Checkmark & \Checkmark & - & \\
SPT-Code & 8.4 & 12.8 & 18.8 & 12.8 & 16.8 & 20.4 & 15.0 & \Checkmark & \Checkmark & - & \\
\bottomrule
    \end{tabular}
    }
\end{table}

%% file: tables/translation.tex
\begin{table}[]
    \centering
    \caption{BLEU and CodeBLEU scores of code translation on CodeTrans.}
    \label{tab:perf-translation1}
    % \adjustbox{width=1\textwidth,center}{
    \begin{tabular}{lccccl}
\toprule
 & \multicolumn{2}{c}{Java$\to$C\#} & \multicolumn{2}{c}{C\#$\to$Java} & \multirow{2}{*}{source}\\
 & BLEU & CodeBLEU & BLEU & CodeBLEU & \\
\midrule
Transformer & 55.8 & 63.7 & 50.5 & 61.6 & \citet{2021CodeXGLUE}$\downarrow$\\
CodeBERT & 79.9 & 85.1 & 72.1 & 79.4 & \\
GraphCodeBERT & 80.6 &  & 72.6 &  & \citet{2020GraphCodeBERT}\\
PLBART & 83.0 & 87.9 & 78.3 & 85.3 & \citet{2021PLBART}\\
SynCoBERT & 80.8 & 84.8 & 76.5 & 82.2 & \citet{2021SynCoBERT}\\
CodeT5$_{base}$ & 84.0 & 87.8 & 79.9 & 84.4 & \citet{2021CodeT5}\\
NatGen &  & 88.1 &  & 85.2 & \citet{2022NatGen}\\
\bottomrule
    \end{tabular}
    % }
\end{table}

\begin{table}[]
    \centering
    \caption{Pass@1 scores of code translation on TransCoder evaluation set.}
    \label{tab:perf-translation2}
    \adjustbox{width=1\textwidth,center}{
    \begin{tabular}{lccccccl}
\toprule
 & C++ $\to$ Java & C++ $\to$ Py & Java $\to$ C++ & Java $\to$ Py & Py $\to$ C++ & Py $\to$ Java & source \\
\midrule
TransCoder & 60.9 & 44.5 & 80.9 & 35.0 & 32.2 & 24.7 & \citet{2020Transcoder}\\
DOBF &  &  &  & 40.6 &  & 46.6 & \citet{2021DOBF}\\
TransCoder-ST & 66.7 & 61.1 & 84.1 & 67.8 & 52.2 & 56.7 & \citet{2021Transcoder}\\
TransCoder-IR & 62.9 &  & 74.5 &  &  &  & \citet{2022Transcoder}\\
LaMDA &  & 30.2 &  &  &  &  & \citet{2022PaLM}$\downarrow$\\
PaLM &  & 51.8 &  &  &  &  & \\
PaLM Coder &  & 55.1 &  &  &  &  & \\
StarCoder (self-debug) &  & 70.0 (76.6) &  &  &  &  & \citet{2023self-debug}$\downarrow$\\
Codex (self-debug) &  & 80.4 (92.5) &  &  &  &  & \\
GPT-3.5 (self-debug) &  & 89.1 (92.7) &  &  &  &  & \\
GPT-4 (self-debug) &  & 77.3 (90.4) &  &  &  &  & \\
\bottomrule
    \end{tabular}
    }
\end{table}

%% file: tables/reasoning.tex
\begin{table}[]
    \centering
    \caption{Pass@1 scores of code reasoning on CRUXEval input and output prediction. DeepSeek-Coder-V2 is a Mixture-of-Experts model with a total of 236B parameters, where 21B are activated for each token.}
    \label{tab:perf-reasoning1}
    % \adjustbox{width=1\textwidth,center}{
    \begin{tabular}{lrccl}
\toprule
& size & input & output & source \\
\midrule
Phi-1 & 1.3B & 13.9 & 23.3 & \citet{2024CRUXEval}$\downarrow$ \\
Phi-1.5 & 1.3B & 24.1 & 27.1 \\
Mistral & 7B & 36.0 & 31.7 \\
StarCoder$_{Base}$ & 16B & 31.6 & 33.3 \\
CodeLLaMA (CoT) & 34B & 46.5 (50.4) & 41.1 (46.0) \\
DeepSeek-Coder$_{Instruct}$ & 33B & 47.4 & 44.0 \\
GPT-3.5 (CoT) &  & 49.2 (49.1) & 50.0 (63.3) \\
GPT-4 (CoT) & - & 67.1 (74.8) & 63.4 (81.9) \\
StarCoder2 & 16B & 48.1 & 47.1 & \citet{2024StarCoder2} \\
Codestral + CoT & 22B & 48.0 & 60.6 & \citet{2024DeepSeekCoder-V2}$\downarrow$ \\
LLaMA-3$_{Instruct}$ + CoT & 70B & 61.1 & 64.3 \\
DeepSeek-Coder-V2$_{Instruct}$ + CoT & 236B (21B) & 70.0 & 75.1 \\
Gemini-1.5-Pro + CoT & - & 67.0 & 77.5 \\
Claude-3-Opus + CoT & - & 73.4 & 82.0 \\
GPT-4o + CoT & - & 77.4 & 88.7 \\
\bottomrule
    \end{tabular}
    % }
\end{table}

\begin{table}[]
    \centering
    \caption{Accuracy of code reasoning performance on PythonIO, extracted from \citet{2024PythonIO}.}
    \label{tab:perf-reasoning2}
    % \adjustbox{width=1\textwidth,center}{
    \begin{tabular}{lrc}
\toprule
& size & accuracy \\
\midrule
Phi-2 & 2.7B & 29.8   \\
Phi-3 & 3.8B & 38.2  \\
ChatGLM3$_{Base}$ & 6B & 27.7  \\
Gemma & 7B & 30.5  \\
Mistral & 7B & 31.7  \\
Qwen1.5 & 7B & 32.8 \\
LLaMA-3$_{Instruct}$ & 8B & 39.0  \\
Flan-T5 & 11B & 29.3  \\
LLaMA-2 & 13B & 26.6  \\
Qwen1.5 & 14B & 40.9  \\
LLaMA-2 & 70B & 38.2  \\
LLaMA-3$_{Instruct}$ & 70B & 70.1  \\
Qwen1.5 & 72B & 50.4  \\
\bottomrule
    \end{tabular}
    % }
\end{table}

%% file: main.bbl
\begin{thebibliography}{902}
\providecommand{\natexlab}[1]{#1}
\providecommand{\url}[1]{\texttt{#1}}
\expandafter\ifx\csname urlstyle\endcsname\relax
  \providecommand{\doi}[1]{doi: #1}\else
  \providecommand{\doi}{doi: \begingroup \urlstyle{rm}\Url}\fi

\bibitem[Abdelnabi et~al.(2020)Abdelnabi, Maatuk, Abdelaziz, and Elakeili]{re-model2020-2}
Esra~A. Abdelnabi, Abdelsalam~M. Maatuk, Tawfig~M. Abdelaziz, and Salwa~M. Elakeili.
\newblock Generating uml class diagram using nlp techniques and heuristic rules.
\newblock In \emph{2020 20th International Conference on Sciences and Techniques of Automatic Control and Computer Engineering (STA)}, pp.\  277--282, 2020.
\newblock \doi{10.1109/STA50679.2020.9329301}.

\bibitem[Abdelnabi et~al.(2021{\natexlab{a}})Abdelnabi, Maatuk, and Hagal]{re-model-survey2021-1}
Esra~A. Abdelnabi, Abdelsalam~M. Maatuk, and Mohammed Hagal.
\newblock Generating uml class diagram from natural language requirements: A survey of approaches and techniques.
\newblock In \emph{2021 IEEE 1st International Maghreb Meeting of the Conference on Sciences and Techniques of Automatic Control and Computer Engineering MI-STA}, pp.\  288--293, 2021{\natexlab{a}}.
\newblock \doi{10.1109/MI-STA52233.2021.9464433}.

\bibitem[Abdelnabi et~al.(2021{\natexlab{b}})Abdelnabi, Maatuk, and Hagal]{re-model2021-3}
Esra~A. Abdelnabi, Abdelsalam~M. Maatuk, and Mohammed Hagal.
\newblock Generating uml class diagram from natural language requirements: A survey of approaches and techniques.
\newblock In \emph{2021 IEEE 1st International Maghreb Meeting of the Conference on Sciences and Techniques of Automatic Control and Computer Engineering MI-STA}, pp.\  288--293, 2021{\natexlab{b}}.
\newblock \doi{10.1109/MI-STA52233.2021.9464433}.

\bibitem[Abdin et~al.(2024)Abdin, Jacobs, Awan, Aneja, Awadallah, Awadalla, Bach, Bahree, Bakhtiari, Behl, Benhaim, Bilenko, Bjorck, Bubeck, Cai, Mendes, Chen, Chaudhary, Chopra, Giorno, de~Rosa, Dixon, Eldan, Iter, Garg, Goswami, Gunasekar, Haider, Hao, Hewett, Huynh, Javaheripi, Jin, Kauffmann, Karampatziakis, Kim, Khademi, Kurilenko, Lee, Lee, Li, Liang, Liu, Lin, Lin, Madan, Mitra, Modi, Nguyen, Norick, Patra, Perez{-}Becker, Portet, Pryzant, Qin, Radmilac, Rosset, Roy, Ruwase, Saarikivi, Saied, Salim, Santacroce, Shah, Shang, Sharma, Song, Tanaka, Wang, Ward, Wang, Witte, Wyatt, Xu, Xu, Yadav, Yang, Yang, Yu, Zhang, Zhang, Zhang, Zhang, Zhang, Zhang, Zhang, and Zhou]{2024Phi-3}
Marah~I Abdin, Sam~Ade Jacobs, Ammar~Ahmad Awan, Jyoti Aneja, Ahmed Awadallah, Hany Awadalla, Nguyen Bach, Amit Bahree, Arash Bakhtiari, Harkirat~S. Behl, Alon Benhaim, Misha Bilenko, Johan Bjorck, S{\'{e}}bastien Bubeck, Martin Cai, Caio C{\'{e}}sar~Teodoro Mendes, Weizhu Chen, Vishrav Chaudhary, Parul Chopra, Allie~Del Giorno, Gustavo de~Rosa, Matthew Dixon, Ronen Eldan, Dan Iter, Amit Garg, Abhishek Goswami, Suriya Gunasekar, Emman Haider, Junheng Hao, Russell~J. Hewett, Jamie Huynh, Mojan Javaheripi, Xin Jin, Piero Kauffmann, Nikos Karampatziakis, Dongwoo Kim, Mahoud Khademi, Lev Kurilenko, James~R. Lee, Yin~Tat Lee, Yuanzhi Li, Chen Liang, Weishung Liu, Eric Lin, Zeqi Lin, Piyush Madan, Arindam Mitra, Hardik Modi, Anh Nguyen, Brandon Norick, Barun Patra, Daniel Perez{-}Becker, Thomas Portet, Reid Pryzant, Heyang Qin, Marko Radmilac, Corby Rosset, Sambudha Roy, Olatunji Ruwase, Olli Saarikivi, Amin Saied, Adil Salim, Michael Santacroce, Shital Shah, Ning Shang, Hiteshi Sharma, Xia Song, Masahiro Tanaka,
  Xin Wang, Rachel Ward, Guanhua Wang, Philipp Witte, Michael Wyatt, Can Xu, Jiahang Xu, Sonali Yadav, Fan Yang, Ziyi Yang, Donghan Yu, Chengruidong Zhang, Cyril Zhang, Jianwen Zhang, Li~Lyna Zhang, Yi~Zhang, Yue Zhang, Yunan Zhang, and Xiren Zhou.
\newblock Phi-3 technical report: {A} highly capable language model locally on your phone.
\newblock \emph{CoRR}, abs/2404.14219, 2024.
\newblock \doi{10.48550/ARXIV.2404.14219}.
\newblock URL \url{https://doi.org/10.48550/arXiv.2404.14219}.

\bibitem[Abualhaija et~al.(2023)Abualhaija, Ceci, and Briand]{re-ana2023-6}
Sallam Abualhaija, Marcello Ceci, and Lionel~C. Briand.
\newblock Legal requirements analysis.
\newblock \emph{CoRR}, abs/2311.13871, 2023.
\newblock \doi{10.48550/ARXIV.2311.13871}.
\newblock URL \url{https://doi.org/10.48550/arXiv.2311.13871}.

\bibitem[Abuhamad et~al.(2018)Abuhamad, AbuHmed, Mohaisen, and Nyang]{author2018-1}
Mohammed Abuhamad, Tamer AbuHmed, Aziz Mohaisen, and DaeHun Nyang.
\newblock Large-scale and language-oblivious code authorship identification.
\newblock In David Lie, Mohammad Mannan, Michael Backes, and XiaoFeng Wang (eds.), \emph{Proceedings of the 2018 {ACM} {SIGSAC} Conference on Computer and Communications Security, {CCS} 2018, Toronto, ON, Canada, October 15-19, 2018}, pp.\  101--114. {ACM}, 2018.
\newblock \doi{10.1145/3243734.3243738}.
\newblock URL \url{https://doi.org/10.1145/3243734.3243738}.

\bibitem[Agashe et~al.(2019)Agashe, Iyer, and Zettlemoyer]{2019JuICe}
Rajas Agashe, Srinivasan Iyer, and Luke Zettlemoyer.
\newblock Juice: {A} large scale distantly supervised dataset for open domain context-based code generation.
\newblock In Kentaro Inui, Jing Jiang, Vincent Ng, and Xiaojun Wan (eds.), \emph{Proceedings of the 2019 Conference on Empirical Methods in Natural Language Processing and the 9th International Joint Conference on Natural Language Processing, {EMNLP-IJCNLP} 2019, Hong Kong, China, November 3-7, 2019}, pp.\  5435--5445. Association for Computational Linguistics, 2019.
\newblock \doi{10.18653/V1/D19-1546}.
\newblock URL \url{https://doi.org/10.18653/v1/D19-1546}.

\bibitem[Aghajanyan et~al.(2022)Aghajanyan, Okhonko, Lewis, Joshi, Xu, Ghosh, and Zettlemoyer]{UI2021-1}
Armen Aghajanyan, Dmytro Okhonko, Mike Lewis, Mandar Joshi, Hu~Xu, Gargi Ghosh, and Luke Zettlemoyer.
\newblock {HTLM:} hyper-text pre-training and prompting of language models.
\newblock In \emph{The Tenth International Conference on Learning Representations, {ICLR} 2022, Virtual Event, April 25-29, 2022}. OpenReview.net, 2022.
\newblock URL \url{https://openreview.net/forum?id=P-pPW1nxf1r}.

\bibitem[Agrawal et~al.(2023)Agrawal, Kanade, Goyal, Lahiri, and Rajamani]{repo2023-1.5}
Lakshya~A Agrawal, Aditya Kanade, Navin Goyal, Shuvendu~K. Lahiri, and Sriram~K. Rajamani.
\newblock Guiding language models of code with global context using monitors.
\newblock \emph{CoRR}, abs/2306.10763, 2023.
\newblock \doi{10.48550/ARXIV.2306.10763}.
\newblock URL \url{https://doi.org/10.48550/arXiv.2306.10763}.

\bibitem[Ahmad et~al.(2020)Ahmad, Chakraborty, Ray, and Chang]{sum2020-2}
Wasi~Uddin Ahmad, Saikat Chakraborty, Baishakhi Ray, and Kai{-}Wei Chang.
\newblock A transformer-based approach for source code summarization.
\newblock In Dan Jurafsky, Joyce Chai, Natalie Schluter, and Joel~R. Tetreault (eds.), \emph{Proceedings of the 58th Annual Meeting of the Association for Computational Linguistics, {ACL} 2020, Online, July 5-10, 2020}, pp.\  4998--5007. Association for Computational Linguistics, 2020.
\newblock \doi{10.18653/v1/2020.acl-main.449}.
\newblock URL \url{https://doi.org/10.18653/v1/2020.acl-main.449}.

\bibitem[Ahmad et~al.(2021)Ahmad, Chakraborty, Ray, and Chang]{2021PLBART}
Wasi~Uddin Ahmad, Saikat Chakraborty, Baishakhi Ray, and Kai{-}Wei Chang.
\newblock Unified pre-training for program understanding and generation.
\newblock In Kristina Toutanova, Anna Rumshisky, Luke Zettlemoyer, Dilek Hakkani{-}T{\"{u}}r, Iz~Beltagy, Steven Bethard, Ryan Cotterell, Tanmoy Chakraborty, and Yichao Zhou (eds.), \emph{Proceedings of the 2021 Conference of the North American Chapter of the Association for Computational Linguistics: Human Language Technologies, {NAACL-HLT} 2021, Online, June 6-11, 2021}, pp.\  2655--2668. Association for Computational Linguistics, 2021.
\newblock \doi{10.18653/v1/2021.naacl-main.211}.
\newblock URL \url{https://doi.org/10.18653/v1/2021.naacl-main.211}.

\bibitem[Ahmad et~al.(2023)Ahmad, Tushar, Chakraborty, and Chang]{trans-data-2021-1}
Wasi~Uddin Ahmad, Md~Golam~Rahman Tushar, Saikat Chakraborty, and Kai{-}Wei Chang.
\newblock {AVATAR:} {A} parallel corpus for java-python program translation.
\newblock In Anna Rogers, Jordan~L. Boyd{-}Graber, and Naoaki Okazaki (eds.), \emph{Findings of the Association for Computational Linguistics: {ACL} 2023, Toronto, Canada, July 9-14, 2023}, pp.\  2268--2281. Association for Computational Linguistics, 2023.
\newblock \doi{10.18653/V1/2023.FINDINGS-ACL.143}.
\newblock URL \url{https://doi.org/10.18653/v1/2023.findings-acl.143}.

\bibitem[Ahmed et~al.(2022)Ahmed, Ahmed, and Eisty]{re-model-survey2022-1}
Sharif Ahmed, Arif Ahmed, and Nasir~U. Eisty.
\newblock Automatic transformation of natural to unified modeling language: {A} systematic review.
\newblock In Juyeon Jo, Yeong{-}Tae Song, Lin Deng, and Junghwan~John Rhee (eds.), \emph{20th {IEEE/ACIS} International Conference on Software Engineering Research, Management and Applications, {SERA} 2022, Las Vegas, NV, USA, May 25-27, 2022}, pp.\  112--119. {IEEE}, 2022.
\newblock \doi{10.1109/SERA54885.2022.9806783}.
\newblock URL \url{https://doi.org/10.1109/SERA54885.2022.9806783}.

\bibitem[Ainslie et~al.(2023)Ainslie, Lee{-}Thorp, de~Jong, Zemlyanskiy, Lebr{\'{o}}n, and Sanghai]{2023GQA}
Joshua Ainslie, James Lee{-}Thorp, Michiel de~Jong, Yury Zemlyanskiy, Federico Lebr{\'{o}}n, and Sumit Sanghai.
\newblock {GQA:} training generalized multi-query transformer models from multi-head checkpoints.
\newblock In Houda Bouamor, Juan Pino, and Kalika Bali (eds.), \emph{Proceedings of the 2023 Conference on Empirical Methods in Natural Language Processing, {EMNLP} 2023, Singapore, December 6-10, 2023}, pp.\  4895--4901. Association for Computational Linguistics, 2023.
\newblock URL \url{https://aclanthology.org/2023.emnlp-main.298}.

\bibitem[Alagarsamy et~al.(2023)Alagarsamy, Tantithamthavorn, and Aleti]{unit2023-2}
Saranya Alagarsamy, Chakkrit Tantithamthavorn, and Aldeida Aleti.
\newblock A3test: Assertion-augmented automated test case generation.
\newblock \emph{CoRR}, abs/2302.10352, 2023.
\newblock \doi{10.48550/ARXIV.2302.10352}.
\newblock URL \url{https://doi.org/10.48550/arXiv.2302.10352}.

\bibitem[Allal et~al.(2023)Allal, Li, Kocetkov, Mou, Akiki, Ferrandis, Muennighoff, Mishra, Gu, Dey, Umapathi, Anderson, Zi, Lamy{-}Poirier, Schoelkopf, Troshin, Abulkhanov, Romero, Lappert, Toni, del R{\'{\i}}o, Liu, Bose, Bhattacharyya, Zhuo, Yu, Villegas, Zocca, Mangrulkar, Lansky, Nguyen, Contractor, Villa, Li, Bahdanau, Jernite, Hughes, Fried, Guha, de~Vries, and von Werra]{2023SantaCoder}
Loubna~Ben Allal, Raymond Li, Denis Kocetkov, Chenghao Mou, Christopher Akiki, Carlos~Mu{\~{n}}oz Ferrandis, Niklas Muennighoff, Mayank Mishra, Alex Gu, Manan Dey, Logesh~Kumar Umapathi, Carolyn~Jane Anderson, Yangtian Zi, Joel Lamy{-}Poirier, Hailey Schoelkopf, Sergey Troshin, Dmitry Abulkhanov, Manuel Romero, Michael Lappert, Francesco~De Toni, Bernardo~Garc{\'{\i}}a del R{\'{\i}}o, Qian Liu, Shamik Bose, Urvashi Bhattacharyya, Terry~Yue Zhuo, Ian Yu, Paulo Villegas, Marco Zocca, Sourab Mangrulkar, David Lansky, Huu Nguyen, Danish Contractor, Luis Villa, Jia Li, Dzmitry Bahdanau, Yacine Jernite, Sean Hughes, Daniel Fried, Arjun Guha, Harm de~Vries, and Leandro von Werra.
\newblock Santacoder: don't reach for the stars!
\newblock \emph{CoRR}, abs/2301.03988, 2023.
\newblock \doi{10.48550/arXiv.2301.03988}.
\newblock URL \url{https://doi.org/10.48550/arXiv.2301.03988}.

\bibitem[Allamanis \& Sutton(2013)Allamanis and Sutton]{completion-data-2013-1}
Miltiadis Allamanis and Charles Sutton.
\newblock Mining source code repositories at massive scale using language modeling.
\newblock In Thomas Zimmermann, Massimiliano~Di Penta, and Sunghun Kim (eds.), \emph{Proceedings of the 10th Working Conference on Mining Software Repositories, {MSR} '13, San Francisco, CA, USA, May 18-19, 2013}, pp.\  207--216. {IEEE} Computer Society, 2013.
\newblock \doi{10.1109/MSR.2013.6624029}.
\newblock URL \url{https://doi.org/10.1109/MSR.2013.6624029}.

\bibitem[Allamanis \& Sutton(2014)Allamanis and Sutton]{idiom2014-1}
Miltiadis Allamanis and Charles Sutton.
\newblock Mining idioms from source code.
\newblock In Shing{-}Chi Cheung, Alessandro Orso, and Margaret{-}Anne~D. Storey (eds.), \emph{Proceedings of the 22nd {ACM} {SIGSOFT} International Symposium on Foundations of Software Engineering, (FSE-22), Hong Kong, China, November 16 - 22, 2014}, pp.\  472--483. {ACM}, 2014.
\newblock \doi{10.1145/2635868.2635901}.
\newblock URL \url{https://doi.org/10.1145/2635868.2635901}.

\bibitem[Allamanis et~al.(2014)Allamanis, Barr, Bird, and Sutton]{completion2014-1}
Miltiadis Allamanis, Earl~T. Barr, Christian Bird, and Charles Sutton.
\newblock Learning natural coding conventions.
\newblock In Shing{-}Chi Cheung, Alessandro Orso, and Margaret{-}Anne~D. Storey (eds.), \emph{Proceedings of the 22nd {ACM} {SIGSOFT} International Symposium on Foundations of Software Engineering, (FSE-22), Hong Kong, China, November 16 - 22, 2014}, pp.\  281--293. {ACM}, 2014.
\newblock \doi{10.1145/2635868.2635883}.
\newblock URL \url{https://doi.org/10.1145/2635868.2635883}.

\bibitem[Allamanis et~al.(2015)Allamanis, Barr, Bird, and Sutton]{id2015-2}
Miltiadis Allamanis, Earl~T. Barr, Christian Bird, and Charles Sutton.
\newblock Suggesting accurate method and class names.
\newblock In Elisabetta~Di Nitto, Mark Harman, and Patrick Heymans (eds.), \emph{Proceedings of the 2015 10th Joint Meeting on Foundations of Software Engineering, {ESEC/FSE} 2015, Bergamo, Italy, August 30 - September 4, 2015}, pp.\  38--49. {ACM}, 2015.
\newblock \doi{10.1145/2786805.2786849}.
\newblock URL \url{https://doi.org/10.1145/2786805.2786849}.

\bibitem[Allamanis et~al.(2016{\natexlab{a}})Allamanis, Barr, Just, and Sutton]{mutant2016-3}
Miltiadis Allamanis, Earl~T. Barr, Ren{\'{e}} Just, and Charles Sutton.
\newblock Tailored mutants fit bugs better.
\newblock \emph{CoRR}, abs/1611.02516, 2016{\natexlab{a}}.
\newblock URL \url{http://arxiv.org/abs/1611.02516}.

\bibitem[Allamanis et~al.(2016{\natexlab{b}})Allamanis, Peng, and Sutton]{id2016-1}
Miltiadis Allamanis, Hao Peng, and Charles Sutton.
\newblock A convolutional attention network for extreme summarization of source code.
\newblock In Maria{-}Florina Balcan and Kilian~Q. Weinberger (eds.), \emph{Proceedings of the 33nd International Conference on Machine Learning, {ICML} 2016, New York City, NY, USA, June 19-24, 2016}, volume~48 of \emph{{JMLR} Workshop and Conference Proceedings}, pp.\  2091--2100. JMLR.org, 2016{\natexlab{b}}.
\newblock URL \url{http://proceedings.mlr.press/v48/allamanis16.html}.

\bibitem[Allamanis et~al.(2018)Allamanis, Brockschmidt, and Khademi]{id2017-1}
Miltiadis Allamanis, Marc Brockschmidt, and Mahmoud Khademi.
\newblock Learning to represent programs with graphs.
\newblock In \emph{6th International Conference on Learning Representations, {ICLR} 2018, Vancouver, BC, Canada, April 30 - May 3, 2018, Conference Track Proceedings}. OpenReview.net, 2018.
\newblock URL \url{https://openreview.net/forum?id=BJOFETxR-}.

\bibitem[Allamanis et~al.(2020)Allamanis, Barr, Ducousso, and Gao]{type2020-2}
Miltiadis Allamanis, Earl~T. Barr, Soline Ducousso, and Zheng Gao.
\newblock Typilus: neural type hints.
\newblock In Alastair~F. Donaldson and Emina Torlak (eds.), \emph{Proceedings of the 41st {ACM} {SIGPLAN} International Conference on Programming Language Design and Implementation, {PLDI} 2020, London, UK, June 15-20, 2020}, pp.\  91--105. {ACM}, 2020.
\newblock \doi{10.1145/3385412.3385997}.
\newblock URL \url{https://doi.org/10.1145/3385412.3385997}.

\bibitem[Allamanis et~al.(2021)Allamanis, Jackson{-}Flux, and Brockschmidt]{defect2021-0.3}
Miltiadis Allamanis, Henry Jackson{-}Flux, and Marc Brockschmidt.
\newblock Self-supervised bug detection and repair.
\newblock In Marc'Aurelio Ranzato, Alina Beygelzimer, Yann~N. Dauphin, Percy Liang, and Jennifer~Wortman Vaughan (eds.), \emph{Advances in Neural Information Processing Systems 34: Annual Conference on Neural Information Processing Systems 2021, NeurIPS 2021, December 6-14, 2021, virtual}, pp.\  27865--27876, 2021.
\newblock URL \url{https://proceedings.neurips.cc/paper/2021/hash/ea96efc03b9a050d895110db8c4af057-Abstract.html}.

\bibitem[Alon et~al.(2018)Alon, Zilberstein, Levy, and Yahav]{id2018-1}
Uri Alon, Meital Zilberstein, Omer Levy, and Eran Yahav.
\newblock A general path-based representation for predicting program properties.
\newblock In Jeffrey~S. Foster and Dan Grossman (eds.), \emph{Proceedings of the 39th {ACM} {SIGPLAN} Conference on Programming Language Design and Implementation, {PLDI} 2018, Philadelphia, PA, USA, June 18-22, 2018}, pp.\  404--419. {ACM}, 2018.
\newblock \doi{10.1145/3192366.3192412}.
\newblock URL \url{https://doi.org/10.1145/3192366.3192412}.

\bibitem[Alon et~al.(2019{\natexlab{a}})Alon, Brody, Levy, and Yahav]{id2018-3}
Uri Alon, Shaked Brody, Omer Levy, and Eran Yahav.
\newblock code2seq: Generating sequences from structured representations of code.
\newblock In \emph{7th International Conference on Learning Representations, {ICLR} 2019, New Orleans, LA, USA, May 6-9, 2019}. OpenReview.net, 2019{\natexlab{a}}.
\newblock URL \url{https://openreview.net/forum?id=H1gKYo09tX}.

\bibitem[Alon et~al.(2019{\natexlab{b}})Alon, Zilberstein, Levy, and Yahav]{id2018-2}
Uri Alon, Meital Zilberstein, Omer Levy, and Eran Yahav.
\newblock code2vec: learning distributed representations of code.
\newblock \emph{Proc. {ACM} Program. Lang.}, 3\penalty0 ({POPL}):\penalty0 40:1--40:29, 2019{\natexlab{b}}.
\newblock \doi{10.1145/3290353}.
\newblock URL \url{https://doi.org/10.1145/3290353}.

\bibitem[Alon et~al.(2020)Alon, Sadaka, Levy, and Yahav]{completion2019-1}
Uri Alon, Roy Sadaka, Omer Levy, and Eran Yahav.
\newblock Structural language models of code.
\newblock In \emph{Proceedings of the 37th International Conference on Machine Learning, {ICML} 2020, 13-18 July 2020, Virtual Event}, volume 119 of \emph{Proceedings of Machine Learning Research}, pp.\  245--256. {PMLR}, 2020.
\newblock URL \url{http://proceedings.mlr.press/v119/alon20a.html}.

\bibitem[Alrubaye et~al.(2020)Alrubaye, Mkaouer, Khokhlov, Reznik, Ouni, and Mcgoff]{api2019-2}
Hussein Alrubaye, Mohamed~Wiem Mkaouer, Igor Khokhlov, Leon Reznik, Ali Ouni, and Jason Mcgoff.
\newblock Learning to recommend third-party library migration opportunities at the {API} level.
\newblock \emph{Appl. Soft Comput.}, 90:\penalty0 106140, 2020.
\newblock \doi{10.1016/J.ASOC.2020.106140}.
\newblock URL \url{https://doi.org/10.1016/j.asoc.2020.106140}.

\bibitem[Amini et~al.(2019)Amini, Gabriel, Lin, Koncel{-}Kedziorski, Choi, and Hajishirzi]{2019MathQA}
Aida Amini, Saadia Gabriel, Shanchuan Lin, Rik Koncel{-}Kedziorski, Yejin Choi, and Hannaneh Hajishirzi.
\newblock Mathqa: Towards interpretable math word problem solving with operation-based formalisms.
\newblock In Jill Burstein, Christy Doran, and Thamar Solorio (eds.), \emph{Proceedings of the 2019 Conference of the North American Chapter of the Association for Computational Linguistics: Human Language Technologies, {NAACL-HLT} 2019, Minneapolis, MN, USA, June 2-7, 2019, Volume 1 (Long and Short Papers)}, pp.\  2357--2367. Association for Computational Linguistics, 2019.
\newblock \doi{10.18653/v1/n19-1245}.
\newblock URL \url{https://doi.org/10.18653/v1/n19-1245}.

\bibitem[Anil et~al.(2023{\natexlab{a}})Anil, Borgeaud, Wu, Alayrac, Yu, Soricut, Schalkwyk, Dai, Hauth, Millican, Silver, Petrov, Johnson, Antonoglou, Schrittwieser, Glaese, Chen, Pitler, Lillicrap, Lazaridou, Firat, Molloy, Isard, Barham, Hennigan, Lee, Viola, Reynolds, Xu, Doherty, Collins, Meyer, Rutherford, Moreira, Ayoub, Goel, Tucker, Piqueras, Krikun, Barr, Savinov, Danihelka, Roelofs, White, Andreassen, von Glehn, Yagati, Kazemi, Gonzalez, Khalman, Sygnowski, and et~al.]{2023Gemini}
Rohan Anil, Sebastian Borgeaud, Yonghui Wu, Jean{-}Baptiste Alayrac, Jiahui Yu, Radu Soricut, Johan Schalkwyk, Andrew~M. Dai, Anja Hauth, Katie Millican, David Silver, Slav Petrov, Melvin Johnson, Ioannis Antonoglou, Julian Schrittwieser, Amelia Glaese, Jilin Chen, Emily Pitler, Timothy~P. Lillicrap, Angeliki Lazaridou, Orhan Firat, James Molloy, Michael Isard, Paul~Ronald Barham, Tom Hennigan, Benjamin Lee, Fabio Viola, Malcolm Reynolds, Yuanzhong Xu, Ryan Doherty, Eli Collins, Clemens Meyer, Eliza Rutherford, Erica Moreira, Kareem Ayoub, Megha Goel, George Tucker, Enrique Piqueras, Maxim Krikun, Iain Barr, Nikolay Savinov, Ivo Danihelka, Becca Roelofs, Ana{\"{\i}}s White, Anders Andreassen, Tamara von Glehn, Lakshman Yagati, Mehran Kazemi, Lucas Gonzalez, Misha Khalman, Jakub Sygnowski, and et~al.
\newblock Gemini: {A} family of highly capable multimodal models.
\newblock \emph{CoRR}, abs/2312.11805, 2023{\natexlab{a}}.
\newblock \doi{10.48550/ARXIV.2312.11805}.
\newblock URL \url{https://doi.org/10.48550/arXiv.2312.11805}.

\bibitem[Anil et~al.(2023{\natexlab{b}})Anil, Dai, Firat, Johnson, Lepikhin, Passos, Shakeri, Taropa, Bailey, Chen, Chu, Clark, Shafey, Huang, Meier{-}Hellstern, Mishra, Moreira, Omernick, Robinson, Ruder, Tay, Xiao, Xu, Zhang, {\'{A}}brego, Ahn, Austin, Barham, Botha, Bradbury, Brahma, Brooks, Catasta, Cheng, Cherry, Choquette{-}Choo, Chowdhery, Crepy, Dave, Dehghani, Dev, Devlin, D{\'{\i}}az, Du, Dyer, Feinberg, Feng, Fienber, Freitag, Garcia, Gehrmann, Gonzalez, and et~al.]{2023PaLM2}
Rohan Anil, Andrew~M. Dai, Orhan Firat, Melvin Johnson, Dmitry Lepikhin, Alexandre Passos, Siamak Shakeri, Emanuel Taropa, Paige Bailey, Zhifeng Chen, Eric Chu, Jonathan~H. Clark, Laurent~El Shafey, Yanping Huang, Kathy Meier{-}Hellstern, Gaurav Mishra, Erica Moreira, Mark Omernick, Kevin Robinson, Sebastian Ruder, Yi~Tay, Kefan Xiao, Yuanzhong Xu, Yujing Zhang, Gustavo~Hern{\'{a}}ndez {\'{A}}brego, Junwhan Ahn, Jacob Austin, Paul Barham, Jan~A. Botha, James Bradbury, Siddhartha Brahma, Kevin Brooks, Michele Catasta, Yong Cheng, Colin Cherry, Christopher~A. Choquette{-}Choo, Aakanksha Chowdhery, Cl{\'{e}}ment Crepy, Shachi Dave, Mostafa Dehghani, Sunipa Dev, Jacob Devlin, Mark D{\'{\i}}az, Nan Du, Ethan Dyer, Vladimir Feinberg, Fangxiaoyu Feng, Vlad Fienber, Markus Freitag, Xavier Garcia, Sebastian Gehrmann, Lucas Gonzalez, and et~al.
\newblock Palm 2 technical report.
\newblock \emph{CoRR}, abs/2305.10403, 2023{\natexlab{b}}.
\newblock \doi{10.48550/arXiv.2305.10403}.
\newblock URL \url{https://doi.org/10.48550/arXiv.2305.10403}.

\bibitem[Arcadinho et~al.(2022)Arcadinho, Apar{\'{\i}}cio, Veiga, and Alegria]{sql2022-3}
Samuel Arcadinho, David Apar{\'{\i}}cio, Hugo Veiga, and Ant{\'{o}}nio Alegria.
\newblock {T5QL:} taming language models for {SQL} generation.
\newblock \emph{CoRR}, abs/2209.10254, 2022.
\newblock \doi{10.48550/ARXIV.2209.10254}.
\newblock URL \url{https://doi.org/10.48550/arXiv.2209.10254}.

\bibitem[Armengol{-}Estap{\'{e}} \& O'Boyle(2021)Armengol{-}Estap{\'{e}} and O'Boyle]{comp-opt2021-1}
Jordi Armengol{-}Estap{\'{e}} and Michael F.~P. O'Boyle.
\newblock Learning {C} to x86 translation: An experiment in neural compilation.
\newblock \emph{CoRR}, abs/2108.07639, 2021.
\newblock URL \url{https://arxiv.org/abs/2108.07639}.

\bibitem[Armengol{-}Estap{\'{e}} et~al.(2023)Armengol{-}Estap{\'{e}}, Woodruff, Cummins, and O'Boyle]{2023SLaDe}
Jordi Armengol{-}Estap{\'{e}}, Jackson Woodruff, Chris Cummins, and Michael F.~P. O'Boyle.
\newblock Slade: {A} portable small language model decompiler for optimized assembler.
\newblock \emph{CoRR}, abs/2305.12520, 2023.
\newblock \doi{10.48550/ARXIV.2305.12520}.
\newblock URL \url{https://doi.org/10.48550/arXiv.2305.12520}.

\bibitem[Arora et~al.(2015{\natexlab{a}})Arora, Sabetzadeh, Briand, and Zimmer]{re-ana2015-1}
Chetan Arora, Mehrdad Sabetzadeh, Lionel~C. Briand, and Frank Zimmer.
\newblock Automated checking of conformance to requirements templates using natural language processing.
\newblock \emph{{IEEE} Trans. Software Eng.}, 41\penalty0 (10):\penalty0 944--968, 2015{\natexlab{a}}.
\newblock \doi{10.1109/TSE.2015.2428709}.
\newblock URL \url{https://doi.org/10.1109/TSE.2015.2428709}.

\bibitem[Arora et~al.(2015{\natexlab{b}})Arora, Sabetzadeh, Goknil, Briand, and Zimmer]{re-ana2015-2}
Chetan Arora, Mehrdad Sabetzadeh, Arda Goknil, Lionel~C. Briand, and Frank Zimmer.
\newblock Change impact analysis for natural language requirements: An {NLP} approach.
\newblock In Didar Zowghi, Vincenzo Gervasi, and Daniel Amyot (eds.), \emph{23rd {IEEE} International Requirements Engineering Conference, {RE} 2015, Ottawa, ON, Canada, August 24-28, 2015}, pp.\  6--15. {IEEE} Computer Society, 2015{\natexlab{b}}.
\newblock \doi{10.1109/RE.2015.7320403}.
\newblock URL \url{https://doi.org/10.1109/RE.2015.7320403}.

\bibitem[Arora et~al.(2016)Arora, Sabetzadeh, Briand, and Zimmer]{re-model2016-2}
Chetan Arora, Mehrdad Sabetzadeh, Lionel~C. Briand, and Frank Zimmer.
\newblock Extracting domain models from natural-language requirements: approach and industrial evaluation.
\newblock In Benoit Baudry and Beno{\^{\i}}t Combemale (eds.), \emph{Proceedings of the {ACM/IEEE} 19th International Conference on Model Driven Engineering Languages and Systems, Saint-Malo, France, October 2-7, 2016}, pp.\  250--260. {ACM}, 2016.
\newblock URL \url{http://dl.acm.org/citation.cfm?id=2976769}.

\bibitem[Arora et~al.(2017)Arora, Sabetzadeh, Briand, and Zimmer]{re-ana2016-3}
Chetan Arora, Mehrdad Sabetzadeh, Lionel~C. Briand, and Frank Zimmer.
\newblock Automated extraction and clustering of requirements glossary terms.
\newblock \emph{{IEEE} Trans. Software Eng.}, 43\penalty0 (10):\penalty0 918--945, 2017.
\newblock \doi{10.1109/TSE.2016.2635134}.
\newblock URL \url{https://doi.org/10.1109/TSE.2016.2635134}.

\bibitem[Arora et~al.(2023)Arora, Grundy, and Abdelrazek]{re-ana2023-4}
Chetan Arora, John Grundy, and Mohamed Abdelrazek.
\newblock Advancing requirements engineering through generative {AI:} assessing the role of llms.
\newblock \emph{CoRR}, abs/2310.13976, 2023.
\newblock \doi{10.48550/ARXIV.2310.13976}.
\newblock URL \url{https://doi.org/10.48550/arXiv.2310.13976}.

\bibitem[Arora et~al.(2024)Arora, Grundy, Puli, and Layton]{re-ana2024-1}
Chetan Arora, John Grundy, Louise Puli, and Natasha Layton.
\newblock Towards standards-compliant assistive technology product specifications via llms.
\newblock \emph{CoRR}, abs/2404.03122, 2024.
\newblock \doi{10.48550/ARXIV.2404.03122}.
\newblock URL \url{https://doi.org/10.48550/arXiv.2404.03122}.

\bibitem[Arteca et~al.(2022)Arteca, Harner, Pradel, and Tip]{unit2022-2}
Ellen Arteca, Sebastian Harner, Michael Pradel, and Frank Tip.
\newblock Nessie: Automatically testing javascript apis with asynchronous callbacks.
\newblock In \emph{44th {IEEE/ACM} 44th International Conference on Software Engineering, {ICSE} 2022, Pittsburgh, PA, USA, May 25-27, 2022}, pp.\  1494--1505. {ACM}, 2022.
\newblock \doi{10.1145/3510003.3510106}.
\newblock URL \url{https://doi.org/10.1145/3510003.3510106}.

\bibitem[Artuso et~al.(2021)Artuso, Luna, Massarelli, and Querzoni]{ob2019-3}
Fiorella Artuso, Giuseppe Antonio~Di Luna, Luca Massarelli, and Leonardo Querzoni.
\newblock In nomine function: Naming functions in stripped binaries with neural networks, 2021.

\bibitem[Asare et~al.(2023)Asare, Nagappan, and Asokan]{analysis2022-1}
Owura Asare, Meiyappan Nagappan, and N.~Asokan.
\newblock Is github's copilot as bad as humans at introducing vulnerabilities in code?
\newblock \emph{Empir. Softw. Eng.}, 28\penalty0 (6):\penalty0 129, 2023.
\newblock \doi{10.1007/S10664-023-10380-1}.
\newblock URL \url{https://doi.org/10.1007/s10664-023-10380-1}.

\bibitem[Athiwaratkun et~al.(2023)Athiwaratkun, Gouda, Wang, Li, Tian, Tan, Ahmad, Wang, Sun, Shang, Gonugondla, Ding, Kumar, Fulton, Farahani, Jain, Giaquinto, Qian, Ramanathan, and Nallapati]{2022MBXP}
Ben Athiwaratkun, Sanjay~Krishna Gouda, Zijian Wang, Xiaopeng Li, Yuchen Tian, Ming Tan, Wasi~Uddin Ahmad, Shiqi Wang, Qing Sun, Mingyue Shang, Sujan~Kumar Gonugondla, Hantian Ding, Varun Kumar, Nathan Fulton, Arash Farahani, Siddhartha Jain, Robert Giaquinto, Haifeng Qian, Murali~Krishna Ramanathan, and Ramesh Nallapati.
\newblock Multi-lingual evaluation of code generation models.
\newblock In \emph{The Eleventh International Conference on Learning Representations, {ICLR} 2023, Kigali, Rwanda, May 1-5, 2023}. OpenReview.net, 2023.
\newblock URL \url{https://openreview.net/pdf?id=Bo7eeXm6An8}.

\bibitem[Austin et~al.(2021)Austin, Odena, Nye, Bosma, Michalewski, Dohan, Jiang, Cai, Terry, Le, and Sutton]{2021MBPP}
Jacob Austin, Augustus Odena, Maxwell~I. Nye, Maarten Bosma, Henryk Michalewski, David Dohan, Ellen Jiang, Carrie~J. Cai, Michael Terry, Quoc~V. Le, and Charles Sutton.
\newblock Program synthesis with large language models.
\newblock \emph{CoRR}, abs/2108.07732, 2021.
\newblock URL \url{https://arxiv.org/abs/2108.07732}.

\bibitem[Aşıroğlu et~al.(2019)Aşıroğlu, Mete, Yıldız, Nalçakan, Sezen, Dağtekin, and Ensari]{UI2019-1}
Batuhan Aşıroğlu, Büşta~Rümeysa Mete, Eyyüp Yıldız, Yağız Nalçakan, Alper Sezen, Mustafa Dağtekin, and Tolga Ensari.
\newblock Automatic html code generation from mock-up images using machine learning techniques.
\newblock In \emph{2019 Scientific Meeting on Electrical-Electronics \& Biomedical Engineering and Computer Science (EBBT)}, pp.\  1--4, 2019.
\newblock \doi{10.1109/EBBT.2019.8741736}.

\bibitem[Ba et~al.(2016)Ba, Kiros, and Hinton]{2016LayerNorm}
Lei~Jimmy Ba, Jamie~Ryan Kiros, and Geoffrey~E. Hinton.
\newblock Layer normalization.
\newblock \emph{CoRR}, abs/1607.06450, 2016.
\newblock URL \url{http://arxiv.org/abs/1607.06450}.

\bibitem[Babe et~al.(2023)Babe, Nguyen, Zi, Guha, Feldman, and Anderson]{2023StudentEval}
Hannah~McLean Babe, Sydney Nguyen, Yangtian Zi, Arjun Guha, Molly~Q. Feldman, and Carolyn~Jane Anderson.
\newblock Studenteval: {A} benchmark of student-written prompts for large language models of code.
\newblock \emph{CoRR}, abs/2306.04556, 2023.
\newblock \doi{10.48550/ARXIV.2306.04556}.
\newblock URL \url{https://doi.org/10.48550/arXiv.2306.04556}.

\bibitem[Bahdanau et~al.(2015)Bahdanau, Cho, and Bengio]{2014attention}
Dzmitry Bahdanau, Kyunghyun Cho, and Yoshua Bengio.
\newblock Neural machine translation by jointly learning to align and translate.
\newblock In Yoshua Bengio and Yann LeCun (eds.), \emph{3rd International Conference on Learning Representations, {ICLR} 2015, San Diego, CA, USA, May 7-9, 2015, Conference Track Proceedings}, 2015.
\newblock URL \url{http://arxiv.org/abs/1409.0473}.

\bibitem[Bahdanau et~al.(2017)Bahdanau, Brakel, Xu, Goyal, Lowe, Pineau, Courville, and Bengio]{2016RL-seq2seq}
Dzmitry Bahdanau, Philemon Brakel, Kelvin Xu, Anirudh Goyal, Ryan Lowe, Joelle Pineau, Aaron~C. Courville, and Yoshua Bengio.
\newblock An actor-critic algorithm for sequence prediction.
\newblock In \emph{5th International Conference on Learning Representations, {ICLR} 2017, Toulon, France, April 24-26, 2017, Conference Track Proceedings}. OpenReview.net, 2017.
\newblock URL \url{https://openreview.net/forum?id=SJDaqqveg}.

\bibitem[Bai et~al.(2023)Bai, Bai, Chu, Cui, Dang, Deng, Fan, Ge, Han, Huang, Hui, Ji, Li, Lin, Lin, Liu, Liu, Lu, Lu, Ma, Men, Ren, Ren, Tan, Tan, Tu, Wang, Wang, Wang, Wu, Xu, Xu, Yang, Yang, Yang, Yang, Yao, Yu, Yuan, Yuan, Zhang, Zhang, Zhang, Zhang, Zhou, Zhou, Zhou, and Zhu]{2023Qwen}
Jinze Bai, Shuai Bai, Yunfei Chu, Zeyu Cui, Kai Dang, Xiaodong Deng, Yang Fan, Wenbin Ge, Yu~Han, Fei Huang, Binyuan Hui, Luo Ji, Mei Li, Junyang Lin, Runji Lin, Dayiheng Liu, Gao Liu, Chengqiang Lu, Keming Lu, Jianxin Ma, Rui Men, Xingzhang Ren, Xuancheng Ren, Chuanqi Tan, Sinan Tan, Jianhong Tu, Peng Wang, Shijie Wang, Wei Wang, Shengguang Wu, Benfeng Xu, Jin Xu, An~Yang, Hao Yang, Jian Yang, Shusheng Yang, Yang Yao, Bowen Yu, Hongyi Yuan, Zheng Yuan, Jianwei Zhang, Xingxuan Zhang, Yichang Zhang, Zhenru Zhang, Chang Zhou, Jingren Zhou, Xiaohuan Zhou, and Tianhang Zhu.
\newblock Qwen technical report.
\newblock \emph{CoRR}, abs/2309.16609, 2023.
\newblock \doi{10.48550/arXiv.2309.16609}.
\newblock URL \url{https://doi.org/10.48550/arXiv.2309.16609}.

\bibitem[Bai et~al.(2022)Bai, Jones, Ndousse, Askell, Chen, DasSarma, Drain, Fort, Ganguli, Henighan, Joseph, Kadavath, Kernion, Conerly, Showk, Elhage, Hatfield{-}Dodds, Hernandez, Hume, Johnston, Kravec, Lovitt, Nanda, Olsson, Amodei, Brown, Clark, McCandlish, Olah, Mann, and Kaplan]{2022Anthropic}
Yuntao Bai, Andy Jones, Kamal Ndousse, Amanda Askell, Anna Chen, Nova DasSarma, Dawn Drain, Stanislav Fort, Deep Ganguli, Tom Henighan, Nicholas Joseph, Saurav Kadavath, Jackson Kernion, Tom Conerly, Sheer~El Showk, Nelson Elhage, Zac Hatfield{-}Dodds, Danny Hernandez, Tristan Hume, Scott Johnston, Shauna Kravec, Liane Lovitt, Neel Nanda, Catherine Olsson, Dario Amodei, Tom~B. Brown, Jack Clark, Sam McCandlish, Chris Olah, Benjamin Mann, and Jared Kaplan.
\newblock Training a helpful and harmless assistant with reinforcement learning from human feedback.
\newblock \emph{CoRR}, abs/2204.05862, 2022.
\newblock \doi{10.48550/ARXIV.2204.05862}.
\newblock URL \url{https://doi.org/10.48550/arXiv.2204.05862}.

\bibitem[Bairi et~al.(2023)Bairi, Sonwane, Kanade, C, Iyer, Parthasarathy, Rajamani, Ashok, and Shet]{CodePlan}
Ramakrishna Bairi, Atharv Sonwane, Aditya Kanade, Vageesh~D. C, Arun Iyer, Suresh Parthasarathy, Sriram~K. Rajamani, Balasubramanyan Ashok, and Shashank Shet.
\newblock Codeplan: Repository-level coding using llms and planning.
\newblock \emph{CoRR}, abs/2309.12499, 2023.
\newblock \doi{10.48550/ARXIV.2309.12499}.
\newblock URL \url{https://doi.org/10.48550/arXiv.2309.12499}.

\bibitem[Bakar et~al.(2016)Bakar, Kasirun, Salleh, and Jalab]{re-ana2016-1}
Noor~Hasrina Bakar, Zarinah~Mohd Kasirun, Norsaremah Salleh, and Hamid~Abdullah Jalab.
\newblock Extracting features from online software reviews to aid requirements reuse.
\newblock \emph{Appl. Soft Comput.}, 49:\penalty0 1297--1315, 2016.
\newblock \doi{10.1016/J.ASOC.2016.07.048}.
\newblock URL \url{https://doi.org/10.1016/j.asoc.2016.07.048}.

\bibitem[Balachandran(2013)]{review2013-1}
Vipin Balachandran.
\newblock Reducing human effort and improving quality in peer code reviews using automatic static analysis and reviewer recommendation.
\newblock In David Notkin, Betty H.~C. Cheng, and Klaus Pohl (eds.), \emph{35th International Conference on Software Engineering, {ICSE} '13, San Francisco, CA, USA, May 18-26, 2013}, pp.\  931--940. {IEEE} Computer Society, 2013.
\newblock \doi{10.1109/ICSE.2013.6606642}.
\newblock URL \url{https://doi.org/10.1109/ICSE.2013.6606642}.

\bibitem[Balog et~al.(2017)Balog, Gaunt, Brockschmidt, Nowozin, and Tarlow]{syn2016-3}
Matej Balog, Alexander~L. Gaunt, Marc Brockschmidt, Sebastian Nowozin, and Daniel Tarlow.
\newblock Deepcoder: Learning to write programs.
\newblock In \emph{5th International Conference on Learning Representations, {ICLR} 2017, Toulon, France, April 24-26, 2017, Conference Track Proceedings}. OpenReview.net, 2017.
\newblock URL \url{https://openreview.net/forum?id=ByldLrqlx}.

\bibitem[Bandara \& Wijayarathna(2013)Bandara and Wijayarathna]{author2013-1}
Upul Bandara and Gamini Wijayarathna.
\newblock Deep neural networks for source code author identification.
\newblock In Minho Lee, Akira Hirose, Zeng{-}Guang Hou, and Rhee~Man Kil (eds.), \emph{Neural Information Processing - 20th International Conference, {ICONIP} 2013, Daegu, Korea, November 3-7, 2013. Proceedings, Part {II}}, volume 8227 of \emph{Lecture Notes in Computer Science}, pp.\  368--375. Springer, 2013.
\newblock \doi{10.1007/978-3-642-42042-9\_46}.
\newblock URL \url{https://doi.org/10.1007/978-3-642-42042-9\_46}.

\bibitem[Banerjee et~al.(2021)Banerjee, Pal, Wang, and Baral]{ob2021-1}
Pratyay Banerjee, Kuntal~Kumar Pal, Fish Wang, and Chitta Baral.
\newblock Variable name recovery in decompiled binary code using constrained masked language modeling.
\newblock \emph{CoRR}, abs/2103.12801, 2021.
\newblock URL \url{https://arxiv.org/abs/2103.12801}.

\bibitem[Bansal et~al.(2021)Bansal, Haque, and McMillan]{sum2021-0.5}
Aakash Bansal, Sakib Haque, and Collin McMillan.
\newblock Project-level encoding for neural source code summarization of subroutines.
\newblock In \emph{29th {IEEE/ACM} International Conference on Program Comprehension, {ICPC} 2021, Madrid, Spain, May 20-21, 2021}, pp.\  253--264. {IEEE}, 2021.
\newblock \doi{10.1109/ICPC52881.2021.00032}.
\newblock URL \url{https://doi.org/10.1109/ICPC52881.2021.00032}.

\bibitem[Barei{\ss} et~al.(2022)Barei{\ss}, Souza, d'Amorim, and Pradel]{unit2022-3}
Patrick Barei{\ss}, Beatriz Souza, Marcelo d'Amorim, and Michael Pradel.
\newblock Code generation tools (almost) for free? {A} study of few-shot, pre-trained language models on code.
\newblock \emph{CoRR}, abs/2206.01335, 2022.
\newblock \doi{10.48550/arXiv.2206.01335}.
\newblock URL \url{https://doi.org/10.48550/arXiv.2206.01335}.

\bibitem[Barnett et~al.(2015)Barnett, Bird, Brunet, and Lahiri]{review2015-2}
Mike Barnett, Christian Bird, Jo{\~{a}}o Brunet, and Shuvendu~K. Lahiri.
\newblock Helping developers help themselves: Automatic decomposition of code review changesets.
\newblock In Antonia Bertolino, Gerardo Canfora, and Sebastian~G. Elbaum (eds.), \emph{37th {IEEE/ACM} International Conference on Software Engineering, {ICSE} 2015, Florence, Italy, May 16-24, 2015, Volume 1}, pp.\  134--144. {IEEE} Computer Society, 2015.
\newblock \doi{10.1109/ICSE.2015.35}.
\newblock URL \url{https://doi.org/10.1109/ICSE.2015.35}.

\bibitem[Barone \& Sennrich(2017)Barone and Sennrich]{sum2017-1}
Antonio Valerio~Miceli Barone and Rico Sennrich.
\newblock A parallel corpus of python functions and documentation strings for automated code documentation and code generation.
\newblock In Greg Kondrak and Taro Watanabe (eds.), \emph{Proceedings of the Eighth International Joint Conference on Natural Language Processing, {IJCNLP} 2017, Taipei, Taiwan, November 27 - December 1, 2017, Volume 2: Short Papers}, pp.\  314--319. Asian Federation of Natural Language Processing, 2017.
\newblock URL \url{https://aclanthology.org/I17-2053/}.

\bibitem[Bartocci et~al.(2023)Bartocci, Mariani, Nickovic, and Yadav]{mutant2023-2}
Ezio Bartocci, Leonardo Mariani, Dejan Nickovic, and Drishti Yadav.
\newblock Property-based mutation testing.
\newblock In \emph{{IEEE} Conference on Software Testing, Verification and Validation, {ICST} 2023, Dublin, Ireland, April 16-20, 2023}, pp.\  222--233. {IEEE}, 2023.
\newblock \doi{10.1109/ICST57152.2023.00029}.
\newblock URL \url{https://doi.org/10.1109/ICST57152.2023.00029}.

\bibitem[Bashir et~al.(2021)Bashir, Bilal, Liaqat, Marjani, Malik, and Ali]{re-model2021-1}
Nauman Bashir, Muhammad Bilal, Misbah Liaqat, Mohsen Marjani, Nadia Malik, and Mohsin Ali.
\newblock Modeling class diagram using nlp in object-oriented designing.
\newblock In \emph{2021 National Computing Colleges Conference (NCCC)}, pp.\  1--6, 2021.
\newblock \doi{10.1109/NCCC49330.2021.9428817}.

\bibitem[Bavarian et~al.(2022)Bavarian, Jun, Tezak, Schulman, McLeavey, Tworek, and Chen]{2022FIM}
Mohammad Bavarian, Heewoo Jun, Nikolas Tezak, John Schulman, Christine McLeavey, Jerry Tworek, and Mark Chen.
\newblock Efficient training of language models to fill in the middle.
\newblock \emph{CoRR}, abs/2207.14255, 2022.
\newblock \doi{10.48550/arXiv.2207.14255}.
\newblock URL \url{https://doi.org/10.48550/arXiv.2207.14255}.

\bibitem[Bavishi et~al.(2019)Bavishi, Lemieux, Fox, Sen, and Stoica]{syn2019-1}
Rohan Bavishi, Caroline Lemieux, Roy Fox, Koushik Sen, and Ion Stoica.
\newblock Autopandas: neural-backed generators for program synthesis.
\newblock \emph{Proc. {ACM} Program. Lang.}, 3\penalty0 ({OOPSLA}):\penalty0 168:1--168:27, 2019.
\newblock \doi{10.1145/3360594}.
\newblock URL \url{https://doi.org/10.1145/3360594}.

\bibitem[Beltramelli(2018)]{UI2017-1}
Tony Beltramelli.
\newblock pix2code: Generating code from a graphical user interface screenshot.
\newblock In \emph{Proceedings of the {ACM} {SIGCHI} Symposium on Engineering Interactive Computing Systems, {EICS} 2018, Paris, France, June 19-22, 2018}, pp.\  3:1--3:6. {ACM}, 2018.
\newblock \doi{10.1145/3220134.3220135}.
\newblock URL \url{https://doi.org/10.1145/3220134.3220135}.

\bibitem[Ben{-}Nun et~al.(2018)Ben{-}Nun, Jakobovits, and Hoefler]{clf2018-1}
Tal Ben{-}Nun, Alice~Shoshana Jakobovits, and Torsten Hoefler.
\newblock Neural code comprehension: {A} learnable representation of code semantics.
\newblock In Samy Bengio, Hanna~M. Wallach, Hugo Larochelle, Kristen Grauman, Nicol{\`{o}} Cesa{-}Bianchi, and Roman Garnett (eds.), \emph{Advances in Neural Information Processing Systems 31: Annual Conference on Neural Information Processing Systems 2018, NeurIPS 2018, December 3-8, 2018, Montr{\'{e}}al, Canada}, pp.\  3589--3601, 2018.
\newblock URL \url{https://proceedings.neurips.cc/paper/2018/hash/17c3433fecc21b57000debdf7ad5c930-Abstract.html}.

\bibitem[Berabi et~al.(2021)Berabi, He, Raychev, and Vechev]{fix2021-4.5}
Berkay Berabi, Jingxuan He, Veselin Raychev, and Martin~T. Vechev.
\newblock Tfix: Learning to fix coding errors with a text-to-text transformer.
\newblock In Marina Meila and Tong Zhang (eds.), \emph{Proceedings of the 38th International Conference on Machine Learning, {ICML} 2021, 18-24 July 2021, Virtual Event}, volume 139 of \emph{Proceedings of Machine Learning Research}, pp.\  780--791. {PMLR}, 2021.
\newblock URL \url{http://proceedings.mlr.press/v139/berabi21a.html}.

\bibitem[Bhandari et~al.(2021)Bhandari, Naseer, and Moonen]{defect-data-2021-2}
Guru~Prasad Bhandari, Amara Naseer, and Leon Moonen.
\newblock Cvefixes: automated collection of vulnerabilities and their fixes from open-source software.
\newblock In Shane McIntosh, Xin Xia, and Sousuke Amasaki (eds.), \emph{{PROMISE} '21: 17th International Conference on Predictive Models and Data Analytics in Software Engineering, Athens Greece, August 19-20, 2021}, pp.\  30--39. {ACM}, 2021.
\newblock \doi{10.1145/3475960.3475985}.
\newblock URL \url{https://doi.org/10.1145/3475960.3475985}.

\bibitem[Bhatia et~al.(2018)Bhatia, Kohli, and Singh]{fix2018-1}
Sahil Bhatia, Pushmeet Kohli, and Rishabh Singh.
\newblock Neuro-symbolic program corrector for introductory programming assignments.
\newblock In Michel Chaudron, Ivica Crnkovic, Marsha Chechik, and Mark Harman (eds.), \emph{Proceedings of the 40th International Conference on Software Engineering, {ICSE} 2018, Gothenburg, Sweden, May 27 - June 03, 2018}, pp.\  60--70. {ACM}, 2018.
\newblock \doi{10.1145/3180155.3180219}.
\newblock URL \url{https://doi.org/10.1145/3180155.3180219}.

\bibitem[Bi et~al.(2023)Bi, Huang, Liu, and Wang]{defect-survey-2023-1}
Yingzhou Bi, Jiangtao Huang, Penghui Liu, and Lianmei Wang.
\newblock Benchmarking software vulnerability detection techniques: {A} survey.
\newblock \emph{CoRR}, abs/2303.16362, 2023.
\newblock \doi{10.48550/ARXIV.2303.16362}.
\newblock URL \url{https://doi.org/10.48550/arXiv.2303.16362}.

\bibitem[Bi et~al.(2024)Bi, Wan, Wang, Zhang, Guan, Lu, Zhang, Sui, Shi, and Jin]{2024ProCoder}
Zhangqian Bi, Yao Wan, Zheng Wang, Hongyu Zhang, Batu Guan, Fangxin Lu, Zili Zhang, Yulei Sui, Xuanhua Shi, and Hai Jin.
\newblock Iterative refinement of project-level code context for precise code generation with compiler feedback.
\newblock \emph{CoRR}, abs/2403.16792, 2024.
\newblock \doi{10.48550/ARXIV.2403.16792}.
\newblock URL \url{https://doi.org/10.48550/arXiv.2403.16792}.

\bibitem[Bian et~al.(2018)Bian, Liang, Shi, Huang, and Cai]{defect2018-4}
Pan Bian, Bin Liang, Wenchang Shi, Jianjun Huang, and Yan Cai.
\newblock Nar-miner: discovering negative association rules from code for bug detection.
\newblock In Gary~T. Leavens, Alessandro Garcia, and Corina~S. Pasareanu (eds.), \emph{Proceedings of the 2018 {ACM} Joint Meeting on European Software Engineering Conference and Symposium on the Foundations of Software Engineering, {ESEC/SIGSOFT} {FSE} 2018, Lake Buena Vista, FL, USA, November 04-09, 2018}, pp.\  411--422. {ACM}, 2018.
\newblock \doi{10.1145/3236024.3236032}.
\newblock URL \url{https://doi.org/10.1145/3236024.3236032}.

\bibitem[Bichsel et~al.(2016)Bichsel, Raychev, Tsankov, and Vechev]{ob2016-1}
Benjamin Bichsel, Veselin Raychev, Petar Tsankov, and Martin~T. Vechev.
\newblock Statistical deobfuscation of android applications.
\newblock In Edgar~R. Weippl, Stefan Katzenbeisser, Christopher Kruegel, Andrew~C. Myers, and Shai Halevi (eds.), \emph{Proceedings of the 2016 {ACM} {SIGSAC} Conference on Computer and Communications Security, Vienna, Austria, October 24-28, 2016}, pp.\  343--355. {ACM}, 2016.
\newblock \doi{10.1145/2976749.2978422}.
\newblock URL \url{https://doi.org/10.1145/2976749.2978422}.

\bibitem[Biderman et~al.(2022)Biderman, Bicheno, and Gao]{2022PileDatasheet}
Stella Biderman, Kieran Bicheno, and Leo Gao.
\newblock Datasheet for the pile.
\newblock \emph{CoRR}, abs/2201.07311, 2022.
\newblock URL \url{https://arxiv.org/abs/2201.07311}.

\bibitem[Bielik et~al.(2016)Bielik, Raychev, and Vechev]{completion2016-2}
Pavol Bielik, Veselin Raychev, and Martin~T. Vechev.
\newblock {PHOG:} probabilistic model for code.
\newblock In Maria{-}Florina Balcan and Kilian~Q. Weinberger (eds.), \emph{Proceedings of the 33nd International Conference on Machine Learning, {ICML} 2016, New York City, NY, USA, June 19-24, 2016}, volume~48 of \emph{{JMLR} Workshop and Conference Proceedings}, pp.\  2933--2942. JMLR.org, 2016.
\newblock URL \url{http://proceedings.mlr.press/v48/bielik16.html}.

\bibitem[Black et~al.(2022)Black, Biderman, Hallahan, Anthony, Gao, Golding, He, Leahy, McDonell, Phang, Pieler, Prashanth, Purohit, Reynolds, Tow, Wang, and Weinbach]{2022GPT-NeoX}
Sidney Black, Stella Biderman, Eric Hallahan, Quentin Anthony, Leo Gao, Laurence Golding, Horace He, Connor Leahy, Kyle McDonell, Jason Phang, Michael Pieler, Usvsn~Sai Prashanth, Shivanshu Purohit, Laria Reynolds, Jonathan Tow, Ben Wang, and Samuel Weinbach.
\newblock {GPT}-{N}eo{X}-20{B}: An open-source autoregressive language model.
\newblock In Angela Fan, Suzana Ilic, Thomas Wolf, and Matthias Gall{\'e} (eds.), \emph{Proceedings of BigScience Episode {\#}5 -- Workshop on Challenges {\&} Perspectives in Creating Large Language Models}, pp.\  95--136, virtual+Dublin, May 2022. Association for Computational Linguistics.
\newblock \doi{10.18653/v1/2022.bigscience-1.9}.
\newblock URL \url{https://aclanthology.org/2022.bigscience-1.9}.

\bibitem[Blasi et~al.(2021)Blasi, Gorla, Ernst, Pezz{\`{e}}, and Carzaniga]{assert2021-2}
Arianna Blasi, Alessandra Gorla, Michael~D. Ernst, Mauro Pezz{\`{e}}, and Antonio Carzaniga.
\newblock Memo: Automatically identifying metamorphic relations in javadoc comments for test automation.
\newblock \emph{J. Syst. Softw.}, 181:\penalty0 111041, 2021.
\newblock \doi{10.1016/j.jss.2021.111041}.
\newblock URL \url{https://doi.org/10.1016/j.jss.2021.111041}.

\bibitem[Bogin et~al.(2019)Bogin, Berant, and Gardner]{sql2019-2}
Ben Bogin, Jonathan Berant, and Matt Gardner.
\newblock Representing schema structure with graph neural networks for text-to-sql parsing.
\newblock In Anna Korhonen, David~R. Traum, and Llu{\'{\i}}s M{\`{a}}rquez (eds.), \emph{Proceedings of the 57th Conference of the Association for Computational Linguistics, {ACL} 2019, Florence, Italy, July 28- August 2, 2019, Volume 1: Long Papers}, pp.\  4560--4565. Association for Computational Linguistics, 2019.
\newblock \doi{10.18653/v1/p19-1448}.
\newblock URL \url{https://doi.org/10.18653/v1/p19-1448}.

\bibitem[B{\"{o}}hme et~al.(2017)B{\"{o}}hme, Pham, Nguyen, and Roychoudhury]{fuzz2017-2}
Marcel B{\"{o}}hme, Van{-}Thuan Pham, Manh{-}Dung Nguyen, and Abhik Roychoudhury.
\newblock Directed greybox fuzzing.
\newblock In Bhavani Thuraisingham, David Evans, Tal Malkin, and Dongyan Xu (eds.), \emph{Proceedings of the 2017 {ACM} {SIGSAC} Conference on Computer and Communications Security, {CCS} 2017, Dallas, TX, USA, October 30 - November 03, 2017}, pp.\  2329--2344. {ACM}, 2017.
\newblock \doi{10.1145/3133956.3134020}.
\newblock URL \url{https://doi.org/10.1145/3133956.3134020}.

\bibitem[B{\"{o}}hme et~al.(2019)B{\"{o}}hme, Pham, and Roychoudhury]{fuzz2016-1}
Marcel B{\"{o}}hme, Van{-}Thuan Pham, and Abhik Roychoudhury.
\newblock Coverage-based greybox fuzzing as markov chain.
\newblock \emph{{IEEE} Trans. Software Eng.}, 45\penalty0 (5):\penalty0 489--506, 2019.
\newblock \doi{10.1109/TSE.2017.2785841}.
\newblock URL \url{https://doi.org/10.1109/TSE.2017.2785841}.

\bibitem[Boone et~al.(2019)Boone, de~Bruin, Langerak, and Stelmach]{type2019-3}
Casper Boone, Niels de~Bruin, Arjan Langerak, and Fabian Stelmach.
\newblock Dltpy: Deep learning type inference of python function signatures using natural language context.
\newblock \emph{CoRR}, abs/1912.00680, 2019.
\newblock URL \url{http://arxiv.org/abs/1912.00680}.

\bibitem[Brown et~al.(2017)Brown, Vaughn, Liblit, and Reps]{mutant2017-1}
David~Bingham Brown, Michael Vaughn, Ben Liblit, and Thomas~W. Reps.
\newblock The care and feeding of wild-caught mutants.
\newblock In Eric Bodden, Wilhelm Sch{\"{a}}fer, Arie van Deursen, and Andrea Zisman (eds.), \emph{Proceedings of the 2017 11th Joint Meeting on Foundations of Software Engineering, {ESEC/FSE} 2017, Paderborn, Germany, September 4-8, 2017}, pp.\  511--522. {ACM}, 2017.
\newblock \doi{10.1145/3106237.3106280}.
\newblock URL \url{https://doi.org/10.1145/3106237.3106280}.

\bibitem[Brown et~al.(2020)Brown, Mann, Ryder, Subbiah, Kaplan, Dhariwal, Neelakantan, Shyam, Sastry, Askell, Agarwal, Herbert{-}Voss, Krueger, Henighan, Child, Ramesh, Ziegler, Wu, Winter, Hesse, Chen, Sigler, Litwin, Gray, Chess, Clark, Berner, McCandlish, Radford, Sutskever, and Amodei]{2020GPT3}
Tom~B. Brown, Benjamin Mann, Nick Ryder, Melanie Subbiah, Jared Kaplan, Prafulla Dhariwal, Arvind Neelakantan, Pranav Shyam, Girish Sastry, Amanda Askell, Sandhini Agarwal, Ariel Herbert{-}Voss, Gretchen Krueger, Tom Henighan, Rewon Child, Aditya Ramesh, Daniel~M. Ziegler, Jeffrey Wu, Clemens Winter, Christopher Hesse, Mark Chen, Eric Sigler, Mateusz Litwin, Scott Gray, Benjamin Chess, Jack Clark, Christopher Berner, Sam McCandlish, Alec Radford, Ilya Sutskever, and Dario Amodei.
\newblock Language models are few-shot learners.
\newblock In Hugo Larochelle, Marc'Aurelio Ranzato, Raia Hadsell, Maria{-}Florina Balcan, and Hsuan{-}Tien Lin (eds.), \emph{Advances in Neural Information Processing Systems 33: Annual Conference on Neural Information Processing Systems 2020, NeurIPS 2020, December 6-12, 2020, virtual}, 2020.
\newblock URL \url{https://proceedings.neurips.cc/paper/2020/hash/1457c0d6bfcb4967418bfb8ac142f64a-Abstract.html}.

\bibitem[Bruch et~al.(2009)Bruch, Monperrus, and Mezini]{completion2009-1}
Marcel Bruch, Martin Monperrus, and Mira Mezini.
\newblock Learning from examples to improve code completion systems.
\newblock In Hans van Vliet and Val{\'{e}}rie Issarny (eds.), \emph{Proceedings of the 7th joint meeting of the European Software Engineering Conference and the {ACM} {SIGSOFT} International Symposium on Foundations of Software Engineering, 2009, Amsterdam, The Netherlands, August 24-28, 2009}, pp.\  213--222. {ACM}, 2009.
\newblock \doi{10.1145/1595696.1595728}.
\newblock URL \url{https://doi.org/10.1145/1595696.1595728}.

\bibitem[Bubeck et~al.(2023)Bubeck, Chandrasekaran, Eldan, Gehrke, Horvitz, Kamar, Lee, Lee, Li, Lundberg, Nori, Palangi, Ribeiro, and Zhang]{2023AGI}
S{\'{e}}bastien Bubeck, Varun Chandrasekaran, Ronen Eldan, Johannes Gehrke, Eric Horvitz, Ece Kamar, Peter Lee, Yin~Tat Lee, Yuanzhi Li, Scott~M. Lundberg, Harsha Nori, Hamid Palangi, Marco~T{\'{u}}lio Ribeiro, and Yi~Zhang.
\newblock Sparks of artificial general intelligence: Early experiments with {GPT-4}.
\newblock \emph{CoRR}, abs/2303.12712, 2023.
\newblock \doi{10.48550/arXiv.2303.12712}.
\newblock URL \url{https://doi.org/10.48550/arXiv.2303.12712}.

\bibitem[Bui et~al.(2022)Bui, Wang, and Hoi]{fix2022-4}
Nghi Bui, Yue Wang, and Steven C.~H. Hoi.
\newblock Detect-localize-repair: {A} unified framework for learning to debug with codet5.
\newblock In Yoav Goldberg, Zornitsa Kozareva, and Yue Zhang (eds.), \emph{Findings of the Association for Computational Linguistics: {EMNLP} 2022, Abu Dhabi, United Arab Emirates, December 7-11, 2022}, pp.\  812--823. Association for Computational Linguistics, 2022.
\newblock \doi{10.18653/V1/2022.FINDINGS-EMNLP.57}.
\newblock URL \url{https://doi.org/10.18653/v1/2022.findings-emnlp.57}.

\bibitem[Bui et~al.(2019)Bui, Yu, and Jiang]{api2019-3}
Nghi D.~Q. Bui, Yijun Yu, and Lingxiao Jiang.
\newblock {SAR:} learning cross-language {API} mappings with little knowledge.
\newblock In Marlon Dumas, Dietmar Pfahl, Sven Apel, and Alessandra Russo (eds.), \emph{Proceedings of the {ACM} Joint Meeting on European Software Engineering Conference and Symposium on the Foundations of Software Engineering, {ESEC/SIGSOFT} {FSE} 2019, Tallinn, Estonia, August 26-30, 2019}, pp.\  796--806. {ACM}, 2019.
\newblock \doi{10.1145/3338906.3338924}.
\newblock URL \url{https://doi.org/10.1145/3338906.3338924}.

\bibitem[Bui et~al.(2021{\natexlab{a}})Bui, Yu, and Jiang]{2020InferCode}
Nghi D.~Q. Bui, Yijun Yu, and Lingxiao Jiang.
\newblock Infercode: Self-supervised learning of code representations by predicting subtrees.
\newblock In \emph{43rd {IEEE/ACM} International Conference on Software Engineering, {ICSE} 2021, Madrid, Spain, 22-30 May 2021}, pp.\  1186--1197. {IEEE}, 2021{\natexlab{a}}.
\newblock \doi{10.1109/ICSE43902.2021.00109}.
\newblock URL \url{https://doi.org/10.1109/ICSE43902.2021.00109}.

\bibitem[Bui et~al.(2021{\natexlab{b}})Bui, Yu, and Jiang]{retrieval2020-0.5}
Nghi D.~Q. Bui, Yijun Yu, and Lingxiao Jiang.
\newblock Self-supervised contrastive learning for code retrieval and summarization via semantic-preserving transformations.
\newblock In Fernando Diaz, Chirag Shah, Torsten Suel, Pablo Castells, Rosie Jones, and Tetsuya Sakai (eds.), \emph{{SIGIR} '21: The 44th International {ACM} {SIGIR} Conference on Research and Development in Information Retrieval, Virtual Event, Canada, July 11-15, 2021}, pp.\  511--521. {ACM}, 2021{\natexlab{b}}.
\newblock \doi{10.1145/3404835.3462840}.
\newblock URL \url{https://doi.org/10.1145/3404835.3462840}.

\bibitem[Bunel et~al.(2018)Bunel, Hausknecht, Devlin, Singh, and Kohli]{syn2018-3}
Rudy Bunel, Matthew~J. Hausknecht, Jacob Devlin, Rishabh Singh, and Pushmeet Kohli.
\newblock Leveraging grammar and reinforcement learning for neural program synthesis.
\newblock In \emph{6th International Conference on Learning Representations, {ICLR} 2018, Vancouver, BC, Canada, April 30 - May 3, 2018, Conference Track Proceedings}. OpenReview.net, 2018.
\newblock URL \url{https://openreview.net/forum?id=H1Xw62kRZ}.

\bibitem[C{\'{a}}mara et~al.(2023)C{\'{a}}mara, Troya, Burgue{\~{n}}o, and Vallecillo]{re-model2023-1}
Javier C{\'{a}}mara, Javier Troya, Lola Burgue{\~{n}}o, and Antonio Vallecillo.
\newblock On the assessment of generative {AI} in modeling tasks: an experience report with chatgpt and {UML}.
\newblock \emph{Softw. Syst. Model.}, 22\penalty0 (3):\penalty0 781--793, 2023.
\newblock \doi{10.1007/S10270-023-01105-5}.
\newblock URL \url{https://doi.org/10.1007/s10270-023-01105-5}.

\bibitem[Cambronero et~al.(2019)Cambronero, Li, Kim, Sen, and Chandra]{retrieval2019-1}
Jos{\'{e}} Cambronero, Hongyu Li, Seohyun Kim, Koushik Sen, and Satish Chandra.
\newblock When deep learning met code search.
\newblock In Marlon Dumas, Dietmar Pfahl, Sven Apel, and Alessandra Russo (eds.), \emph{Proceedings of the {ACM} Joint Meeting on European Software Engineering Conference and Symposium on the Foundations of Software Engineering, {ESEC/SIGSOFT} {FSE} 2019, Tallinn, Estonia, August 26-30, 2019}, pp.\  964--974. {ACM}, 2019.
\newblock \doi{10.1145/3338906.3340458}.
\newblock URL \url{https://doi.org/10.1145/3338906.3340458}.

\bibitem[Cao et~al.(2023)Cao, Li, Wen, and Cheung]{fix2023-2}
Jialun Cao, Meiziniu Li, Ming Wen, and Shing{-}Chi Cheung.
\newblock A study on prompt design, advantages and limitations of chatgpt for deep learning program repair.
\newblock \emph{CoRR}, abs/2304.08191, 2023.
\newblock \doi{10.48550/ARXIV.2304.08191}.
\newblock URL \url{https://doi.org/10.48550/arXiv.2304.08191}.

\bibitem[Cao et~al.(2021)Cao, Chen, Chen, Zhao, Zhu, and Yu]{sql2021-1}
Ruisheng Cao, Lu~Chen, Zhi Chen, Yanbin Zhao, Su~Zhu, and Kai Yu.
\newblock {LGESQL:} line graph enhanced text-to-sql model with mixed local and non-local relations.
\newblock In Chengqing Zong, Fei Xia, Wenjie Li, and Roberto Navigli (eds.), \emph{Proceedings of the 59th Annual Meeting of the Association for Computational Linguistics and the 11th International Joint Conference on Natural Language Processing, {ACL/IJCNLP} 2021, (Volume 1: Long Papers), Virtual Event, August 1-6, 2021}, pp.\  2541--2555. Association for Computational Linguistics, 2021.
\newblock \doi{10.18653/v1/2021.acl-long.198}.
\newblock URL \url{https://doi.org/10.18653/v1/2021.acl-long.198}.

\bibitem[Cao et~al.(2022)Cao, Liang, Chen, and Hu]{2023NeurDP}
Ying Cao, Ruigang Liang, Kai Chen, and Peiwei Hu.
\newblock Boosting neural networks to decompile optimized binaries.
\newblock In \emph{Annual Computer Security Applications Conference, {ACSAC} 2022, Austin, TX, USA, December 5-9, 2022}, pp.\  508--518. {ACM}, 2022.
\newblock \doi{10.1145/3564625.3567998}.
\newblock URL \url{https://doi.org/10.1145/3564625.3567998}.

\bibitem[Cassano et~al.(2023{\natexlab{a}})Cassano, Gouwar, Lucchetti, Schlesinger, Anderson, Greenberg, Jangda, and Guha]{2023MultiPL-T}
Federico Cassano, John Gouwar, Francesca Lucchetti, Claire Schlesinger, Carolyn~Jane Anderson, Michael Greenberg, Abhinav Jangda, and Arjun Guha.
\newblock Knowledge transfer from high-resource to low-resource programming languages for code llms.
\newblock \emph{CoRR}, abs/2308.09895, 2023{\natexlab{a}}.
\newblock \doi{10.48550/ARXIV.2308.09895}.
\newblock URL \url{https://doi.org/10.48550/arXiv.2308.09895}.

\bibitem[Cassano et~al.(2023{\natexlab{b}})Cassano, Gouwar, Nguyen, Nguyen, Phipps-Costin, Pinckney, Yee, Zi, Anderson, Feldman, et~al.]{2022MultiPL-E}
Federico Cassano, John Gouwar, Daniel Nguyen, Sydney Nguyen, Luna Phipps-Costin, Donald Pinckney, Ming-Ho Yee, Yangtian Zi, Carolyn~Jane Anderson, Molly~Q Feldman, et~al.
\newblock Multipl-e: a scalable and polyglot approach to benchmarking neural code generation.
\newblock \emph{IEEE Transactions on Software Engineering}, 2023{\natexlab{b}}.

\bibitem[Cassano et~al.(2023{\natexlab{c}})Cassano, Yee, Shinn, Guha, and Holtzen]{type2023-3}
Federico Cassano, Ming{-}Ho Yee, Noah Shinn, Arjun Guha, and Steven Holtzen.
\newblock Type prediction with program decomposition and fill-in-the-type training.
\newblock \emph{CoRR}, abs/2305.17145, 2023{\natexlab{c}}.
\newblock \doi{10.48550/ARXIV.2305.17145}.
\newblock URL \url{https://doi.org/10.48550/arXiv.2305.17145}.

\bibitem[Cha et~al.(2015)Cha, Woo, and Brumley]{fuzz2015-1}
Sang~Kil Cha, Maverick Woo, and David Brumley.
\newblock Program-adaptive mutational fuzzing.
\newblock In \emph{2015 {IEEE} Symposium on Security and Privacy, {SP} 2015, San Jose, CA, USA, May 17-21, 2015}, pp.\  725--741. {IEEE} Computer Society, 2015.
\newblock \doi{10.1109/SP.2015.50}.
\newblock URL \url{https://doi.org/10.1109/SP.2015.50}.

\bibitem[Chaaben et~al.(2023)Chaaben, Burgue{\~{n}}o, and Sahraoui]{re-model2022-1}
Meriem~Ben Chaaben, Lola Burgue{\~{n}}o, and Houari~A. Sahraoui.
\newblock Towards using few-shot prompt learning for automating model completion.
\newblock In \emph{45th {IEEE/ACM} International Conference on Software Engineering: New Ideas and Emerging Results, NIER@ICSE, Melbourne, Australia, May 14-20, 2023}, pp.\  7--12. {IEEE}, 2023.
\newblock \doi{10.1109/ICSE-NIER58687.2023.00008}.
\newblock URL \url{https://doi.org/10.1109/ICSE-NIER58687.2023.00008}.

\bibitem[Chae et~al.(2024)Chae, Kim, Kim, iunn Ong, woo Kwak, Kim, Kim, Kwon, Chung, Yu, and Yeo]{2024Think-and-Execute}
Hyungjoo Chae, Yeonghyeon Kim, Seungone Kim, Kai~Tzu iunn Ong, Beong woo Kwak, Moohyeon Kim, Seonghwan Kim, Taeyoon Kwon, Jiwan Chung, Youngjae Yu, and Jinyoung Yeo.
\newblock Language models as compilers: Simulating pseudocode execution improves algorithmic reasoning in language models.
\newblock \emph{CoRR}, abs/2404.02575, 2024.
\newblock \doi{10.48550/ARXIV.2404.02575}.
\newblock URL \url{https://doi.org/10.48550/arXiv.2404.02575}.

\bibitem[Chai et~al.(2023)Chai, Wang, Pang, Sun, Tian, and Wu]{2022ERNIE-Code}
Yekun Chai, Shuohuan Wang, Chao Pang, Yu~Sun, Hao Tian, and Hua Wu.
\newblock Ernie-code: Beyond english-centric cross-lingual pretraining for programming languages.
\newblock In Anna Rogers, Jordan~L. Boyd{-}Graber, and Naoaki Okazaki (eds.), \emph{Findings of the Association for Computational Linguistics: {ACL} 2023, Toronto, Canada, July 9-14, 2023}, pp.\  10628--10650. Association for Computational Linguistics, 2023.
\newblock \doi{10.18653/V1/2023.FINDINGS-ACL.676}.
\newblock URL \url{https://doi.org/10.18653/v1/2023.findings-acl.676}.

\bibitem[Chai et~al.(2022)Chai, Zhang, Shen, and Gu]{retrieval2022-1}
Yitian Chai, Hongyu Zhang, Beijun Shen, and Xiaodong Gu.
\newblock Cross-domain deep code search with meta learning.
\newblock In \emph{44th {IEEE/ACM} 44th International Conference on Software Engineering, {ICSE} 2022, Pittsburgh, PA, USA, May 25-27, 2022}, pp.\  487--498. {ACM}, 2022.
\newblock \doi{10.1145/3510003.3510125}.
\newblock URL \url{https://doi.org/10.1145/3510003.3510125}.

\bibitem[Chakraborty \& Ray(2021)Chakraborty and Ray]{fix2021-5}
Saikat Chakraborty and Baishakhi Ray.
\newblock On multi-modal learning of editing source code.
\newblock In \emph{36th {IEEE/ACM} International Conference on Automated Software Engineering, {ASE} 2021, Melbourne, Australia, November 15-19, 2021}, pp.\  443--455. {IEEE}, 2021.
\newblock \doi{10.1109/ASE51524.2021.9678559}.
\newblock URL \url{https://doi.org/10.1109/ASE51524.2021.9678559}.

\bibitem[Chakraborty et~al.(2022{\natexlab{a}})Chakraborty, Ahmed, Ding, Devanbu, and Ray]{2022NatGen}
Saikat Chakraborty, Toufique Ahmed, Yangruibo Ding, Premkumar~T. Devanbu, and Baishakhi Ray.
\newblock Natgen: generative pre-training by "naturalizing" source code.
\newblock In Abhik Roychoudhury, Cristian Cadar, and Miryung Kim (eds.), \emph{Proceedings of the 30th {ACM} Joint European Software Engineering Conference and Symposium on the Foundations of Software Engineering, {ESEC/FSE} 2022, Singapore, Singapore, November 14-18, 2022}, pp.\  18--30. {ACM}, 2022{\natexlab{a}}.
\newblock \doi{10.1145/3540250.3549162}.
\newblock URL \url{https://doi.org/10.1145/3540250.3549162}.

\bibitem[Chakraborty et~al.(2022{\natexlab{b}})Chakraborty, Ding, Allamanis, and Ray]{fix2018-2}
Saikat Chakraborty, Yangruibo Ding, Miltiadis Allamanis, and Baishakhi Ray.
\newblock {CODIT:} code editing with tree-based neural models.
\newblock \emph{{IEEE} Trans. Software Eng.}, 48\penalty0 (4):\penalty0 1385--1399, 2022{\natexlab{b}}.
\newblock \doi{10.1109/TSE.2020.3020502}.
\newblock URL \url{https://doi.org/10.1109/TSE.2020.3020502}.

\bibitem[Chakraborty et~al.(2022{\natexlab{c}})Chakraborty, Krishna, Ding, and Ray]{defect2020-1}
Saikat Chakraborty, Rahul Krishna, Yangruibo Ding, and Baishakhi Ray.
\newblock Deep learning based vulnerability detection: Are we there yet?
\newblock \emph{{IEEE} Trans. Software Eng.}, 48\penalty0 (9):\penalty0 3280--3296, 2022{\natexlab{c}}.
\newblock \doi{10.1109/TSE.2021.3087402}.
\newblock URL \url{https://doi.org/10.1109/TSE.2021.3087402}.

\bibitem[Chan et~al.(2023)Chan, Kharkar, Moghaddam, Mohylevskyy, Helyar, Kamal, Elkamhawy, and Sundaresan]{defect2023-0.5}
Aaron Chan, Anant Kharkar, Roshanak~Zilouchian Moghaddam, Yevhen Mohylevskyy, Alec Helyar, Eslam Kamal, Mohamed Elkamhawy, and Neel Sundaresan.
\newblock Transformer-based vulnerability detection in code at edittime: Zero-shot, few-shot, or fine-tuning?
\newblock \emph{CoRR}, abs/2306.01754, 2023.
\newblock \doi{10.48550/ARXIV.2306.01754}.
\newblock URL \url{https://doi.org/10.48550/arXiv.2306.01754}.

\bibitem[Chandel et~al.(2022)Chandel, Clement, Serrato, and Sundaresan]{2022DSP}
Shubham Chandel, Colin~B. Clement, Guillermo Serrato, and Neel Sundaresan.
\newblock Training and evaluating a jupyter notebook data science assistant.
\newblock \emph{CoRR}, abs/2201.12901, 2022.
\newblock URL \url{https://arxiv.org/abs/2201.12901}.

\bibitem[Chang \& Fosler{-}Lussier(2023)Chang and Fosler{-}Lussier]{sql2023-0.5}
Shuaichen Chang and Eric Fosler{-}Lussier.
\newblock How to prompt llms for text-to-sql: {A} study in zero-shot, single-domain, and cross-domain settings.
\newblock \emph{CoRR}, abs/2305.11853, 2023.
\newblock \doi{10.48550/ARXIV.2305.11853}.
\newblock URL \url{https://doi.org/10.48550/arXiv.2305.11853}.

\bibitem[Chen et~al.(2023{\natexlab{a}})Chen, Zhang, Nguyen, Zan, Lin, Lou, and Chen]{2022CodeT}
Bei Chen, Fengji Zhang, Anh Nguyen, Daoguang Zan, Zeqi Lin, Jian{-}Guang Lou, and Weizhu Chen.
\newblock Codet: Code generation with generated tests.
\newblock In \emph{The Eleventh International Conference on Learning Representations, {ICLR} 2023, Kigali, Rwanda, May 1-5, 2023}. OpenReview.net, 2023{\natexlab{a}}.
\newblock URL \url{https://openreview.net/pdf?id=ktrw68Cmu9c}.

\bibitem[Chen et~al.(2023{\natexlab{b}})Chen, Chen, Hassani, Yang, Amyot, Lessard, Mussbacher, Sabetzadeh, and Varr{\'{o}}]{re-model2023-2}
Boqi Chen, Kua Chen, Shabnam Hassani, Yujing Yang, Daniel Amyot, Lysanne Lessard, Gunter Mussbacher, Mehrdad Sabetzadeh, and D{\'{a}}niel Varr{\'{o}}.
\newblock On the use of {GPT-4} for creating goal models: An exploratory study.
\newblock In Kurt Schneider, Fabiano Dalpiaz, and Jennifer Horkoff (eds.), \emph{31st {IEEE} International Requirements Engineering Conference, {RE} 2023 - Workshops, Hannover, Germany, September 4-5, 2023}, pp.\  262--271. {IEEE}, 2023{\natexlab{b}}.
\newblock \doi{10.1109/REW57809.2023.00052}.
\newblock URL \url{https://doi.org/10.1109/REW57809.2023.00052}.

\bibitem[Chen et~al.(2021{\natexlab{a}})Chen, Xing, Liu, and Xiong]{api2019-1}
Chunyang Chen, Zhenchang Xing, Yang Liu, and Kent Ong~Long Xiong.
\newblock Mining likely analogical apis across third-party libraries via large-scale unsupervised {API} semantics embedding.
\newblock \emph{{IEEE} Trans. Software Eng.}, 47\penalty0 (3):\penalty0 432--447, 2021{\natexlab{a}}.
\newblock \doi{10.1109/TSE.2019.2896123}.
\newblock URL \url{https://doi.org/10.1109/TSE.2019.2896123}.

\bibitem[Chen et~al.(2022)Chen, Fard, Lo, and Bryksin]{2022Transfer}
Fuxiang Chen, Fatemeh~H. Fard, David Lo, and Timofey Bryksin.
\newblock On the transferability of pre-trained language models for low-resource programming languages.
\newblock In Ayushi Rastogi, Rosalia Tufano, Gabriele Bavota, Venera Arnaoudova, and Sonia Haiduc (eds.), \emph{Proceedings of the 30th {IEEE/ACM} International Conference on Program Comprehension, {ICPC} 2022, Virtual Event, May 16-17, 2022}, pp.\  401--412. {ACM}, 2022.
\newblock \doi{10.1145/3524610.3527917}.
\newblock URL \url{https://doi.org/10.1145/3524610.3527917}.

\bibitem[Chen et~al.(2023{\natexlab{c}})Chen, Yang, Chen, L{\'{o}}pez, Mussbacher, and Varr{\'{o}}]{re-model2023-3}
Kua Chen, Yujing Yang, Boqi Chen, Jos{\'{e}} Antonio~Hern{\'{a}}ndez L{\'{o}}pez, Gunter Mussbacher, and D{\'{a}}niel Varr{\'{o}}.
\newblock Automated domain modeling with large language models: {A} comparative study.
\newblock In \emph{26th {ACM/IEEE} International Conference on Model Driven Engineering Languages and Systems, {MODELS} 2023, V{\"{a}}ster{\aa}s, Sweden, October 1-6, 2023}, pp.\  162--172. {IEEE}, 2023{\natexlab{c}}.
\newblock \doi{10.1109/MODELS58315.2023.00037}.
\newblock URL \url{https://doi.org/10.1109/MODELS58315.2023.00037}.

\bibitem[Chen et~al.(2021{\natexlab{b}})Chen, Tworek, Jun, Yuan, de~Oliveira~Pinto, Kaplan, Edwards, Burda, Joseph, Brockman, Ray, Puri, Krueger, Petrov, Khlaaf, Sastry, Mishkin, Chan, Gray, Ryder, Pavlov, Power, Kaiser, Bavarian, Winter, Tillet, Such, Cummings, Plappert, Chantzis, Barnes, Herbert{-}Voss, Guss, Nichol, Paino, Tezak, Tang, Babuschkin, Balaji, Jain, Saunders, Hesse, Carr, Leike, Achiam, Misra, Morikawa, Radford, Knight, Brundage, Murati, Mayer, Welinder, McGrew, Amodei, McCandlish, Sutskever, and Zaremba]{2021Codex}
Mark Chen, Jerry Tworek, Heewoo Jun, Qiming Yuan, Henrique~Pond{\'{e}} de~Oliveira~Pinto, Jared Kaplan, Harrison Edwards, Yuri Burda, Nicholas Joseph, Greg Brockman, Alex Ray, Raul Puri, Gretchen Krueger, Michael Petrov, Heidy Khlaaf, Girish Sastry, Pamela Mishkin, Brooke Chan, Scott Gray, Nick Ryder, Mikhail Pavlov, Alethea Power, Lukasz Kaiser, Mohammad Bavarian, Clemens Winter, Philippe Tillet, Felipe~Petroski Such, Dave Cummings, Matthias Plappert, Fotios Chantzis, Elizabeth Barnes, Ariel Herbert{-}Voss, William~Hebgen Guss, Alex Nichol, Alex Paino, Nikolas Tezak, Jie Tang, Igor Babuschkin, Suchir Balaji, Shantanu Jain, William Saunders, Christopher Hesse, Andrew~N. Carr, Jan Leike, Joshua Achiam, Vedant Misra, Evan Morikawa, Alec Radford, Matthew Knight, Miles Brundage, Mira Murati, Katie Mayer, Peter Welinder, Bob McGrew, Dario Amodei, Sam McCandlish, Ilya Sutskever, and Wojciech Zaremba.
\newblock Evaluating large language models trained on code.
\newblock \emph{CoRR}, abs/2107.03374, 2021{\natexlab{b}}.
\newblock URL \url{https://arxiv.org/abs/2107.03374}.

\bibitem[Chen \& Chen(2018)Chen and Chen]{fuzz2018-1}
Peng Chen and Hao Chen.
\newblock Angora: Efficient fuzzing by principled search.
\newblock In \emph{2018 {IEEE} Symposium on Security and Privacy, {SP} 2018, Proceedings, 21-23 May 2018, San Francisco, California, {USA}}, pp.\  711--725. {IEEE} Computer Society, 2018.
\newblock \doi{10.1109/SP.2018.00046}.
\newblock URL \url{https://doi.org/10.1109/SP.2018.00046}.

\bibitem[Chen et~al.(2023{\natexlab{d}})Chen, Wong, Chen, and Tian]{2023PI}
Shouyuan Chen, Sherman Wong, Liangjian Chen, and Yuandong Tian.
\newblock Extending context window of large language models via positional interpolation.
\newblock \emph{CoRR}, abs/2306.15595, 2023{\natexlab{d}}.
\newblock \doi{10.48550/ARXIV.2306.15595}.
\newblock URL \url{https://doi.org/10.48550/arXiv.2306.15595}.

\bibitem[Chen et~al.(2023{\natexlab{e}})Chen, Ma, Wang, and Cohen]{2022PoT}
Wenhu Chen, Xueguang Ma, Xinyi Wang, and William~W. Cohen.
\newblock Program of thoughts prompting: Disentangling computation from reasoning for numerical reasoning tasks.
\newblock \emph{Transactions on Machine Learning Research}, 2023{\natexlab{e}}.
\newblock ISSN 2835-8856.
\newblock URL \url{https://openreview.net/forum?id=YfZ4ZPt8zd}.

\bibitem[Chen et~al.(2018)Chen, Liu, and Song]{trans2018-1}
Xinyun Chen, Chang Liu, and Dawn Song.
\newblock Tree-to-tree neural networks for program translation.
\newblock In Samy Bengio, Hanna~M. Wallach, Hugo Larochelle, Kristen Grauman, Nicol{\`{o}} Cesa{-}Bianchi, and Roman Garnett (eds.), \emph{Advances in Neural Information Processing Systems 31: Annual Conference on Neural Information Processing Systems 2018, NeurIPS 2018, December 3-8, 2018, Montr{\'{e}}al, Canada}, pp.\  2552--2562, 2018.
\newblock URL \url{https://proceedings.neurips.cc/paper/2018/hash/d759175de8ea5b1d9a2660e45554894f-Abstract.html}.

\bibitem[Chen et~al.(2021{\natexlab{c}})Chen, Gong, Cheung, and Song]{2021PlotCoder}
Xinyun Chen, Linyuan Gong, Alvin Cheung, and Dawn Song.
\newblock Plotcoder: Hierarchical decoding for synthesizing visualization code in programmatic context.
\newblock In Chengqing Zong, Fei Xia, Wenjie Li, and Roberto Navigli (eds.), \emph{Proceedings of the 59th Annual Meeting of the Association for Computational Linguistics and the 11th International Joint Conference on Natural Language Processing, {ACL/IJCNLP} 2021, (Volume 1: Long Papers), Virtual Event, August 1-6, 2021}, pp.\  2169--2181. Association for Computational Linguistics, 2021{\natexlab{c}}.
\newblock \doi{10.18653/V1/2021.ACL-LONG.169}.
\newblock URL \url{https://doi.org/10.18653/v1/2021.acl-long.169}.

\bibitem[Chen et~al.(2023{\natexlab{f}})Chen, Lin, Sch{\"{a}}rli, and Zhou]{2023self-debug}
Xinyun Chen, Maxwell Lin, Nathanael Sch{\"{a}}rli, and Denny Zhou.
\newblock Teaching large language models to self-debug.
\newblock \emph{CoRR}, abs/2304.05128, 2023{\natexlab{f}}.
\newblock \doi{10.48550/arXiv.2304.05128}.
\newblock URL \url{https://doi.org/10.48550/arXiv.2304.05128}.

\bibitem[Chen et~al.(2024)Chen, Liu, Meng, Chen, Xu, and Zhou]{2024MANGO}
Yijie Chen, Yijin Liu, Fandong Meng, Yufeng Chen, Jinan Xu, and Jie Zhou.
\newblock Comments as natural logic pivots: Improve code generation via comment perspective.
\newblock \emph{CoRR}, abs/2404.07549, 2024.
\newblock \doi{10.48550/ARXIV.2404.07549}.
\newblock URL \url{https://doi.org/10.48550/arXiv.2404.07549}.

\bibitem[Chen et~al.(2023{\natexlab{g}})Chen, Ding, Alowain, Chen, and Wagner]{defect-data-2023-1}
Yizheng Chen, Zhoujie Ding, Lamya Alowain, Xinyun Chen, and David~A. Wagner.
\newblock Diversevul: {A} new vulnerable source code dataset for deep learning based vulnerability detection.
\newblock In \emph{Proceedings of the 26th International Symposium on Research in Attacks, Intrusions and Defenses, {RAID} 2023, Hong Kong, China, October 16-18, 2023}, pp.\  654--668. {ACM}, 2023{\natexlab{g}}.
\newblock \doi{10.1145/3607199.3607242}.
\newblock URL \url{https://doi.org/10.1145/3607199.3607242}.

\bibitem[Chen et~al.(2021{\natexlab{d}})Chen, Liu, Gu, Su, and Lyu]{log-survey-2021-1}
Zhuangbin Chen, Jinyang Liu, Wenwei Gu, Yuxin Su, and Michael~R. Lyu.
\newblock Experience report: Deep learning-based system log analysis for anomaly detection.
\newblock \emph{CoRR}, abs/2107.05908, 2021{\natexlab{d}}.
\newblock URL \url{https://arxiv.org/abs/2107.05908}.

\bibitem[Chen et~al.(2021{\natexlab{e}})Chen, Kommrusch, Tufano, Pouchet, Poshyvanyk, and Monperrus]{fix2018-4}
Zimin Chen, Steve Kommrusch, Michele Tufano, Louis{-}No{\"{e}}l Pouchet, Denys Poshyvanyk, and Martin Monperrus.
\newblock Sequencer: Sequence-to-sequence learning for end-to-end program repair.
\newblock \emph{{IEEE} Trans. Software Eng.}, 47\penalty0 (9):\penalty0 1943--1959, 2021{\natexlab{e}}.
\newblock \doi{10.1109/TSE.2019.2940179}.
\newblock URL \url{https://doi.org/10.1109/TSE.2019.2940179}.

\bibitem[Cheng et~al.(2023)Cheng, Huang, Li, and Li]{UI-CV1}
Chin{-}Yi Cheng, Forrest Huang, Gang Li, and Yang Li.
\newblock Play: Parametrically conditioned layout generation using latent diffusion.
\newblock In Andreas Krause, Emma Brunskill, Kyunghyun Cho, Barbara Engelhardt, Sivan Sabato, and Jonathan Scarlett (eds.), \emph{International Conference on Machine Learning, {ICML} 2023, 23-29 July 2023, Honolulu, Hawaii, {USA}}, volume 202 of \emph{Proceedings of Machine Learning Research}, pp.\  5449--5471. {PMLR}, 2023.
\newblock URL \url{https://proceedings.mlr.press/v202/cheng23b.html}.

\bibitem[Chiang(2007)]{2007Chiang}
David Chiang.
\newblock Hierarchical phrase-based translation.
\newblock \emph{Comput. Linguistics}, 33\penalty0 (2):\penalty0 201--228, 2007.
\newblock \doi{10.1162/coli.2007.33.2.201}.
\newblock URL \url{https://doi.org/10.1162/coli.2007.33.2.201}.

\bibitem[Chochlov et~al.(2022)Chochlov, Ahmed, Patten, Lu, Hou, Gregg, and Buckley]{clone2022-2}
Muslim Chochlov, Gul~Aftab Ahmed, James~Vincent Patten, Guoxian Lu, Wei Hou, David Gregg, and Jim Buckley.
\newblock Using a nearest-neighbour, bert-based approach for scalable clone detection.
\newblock In \emph{{IEEE} International Conference on Software Maintenance and Evolution, {ICSME} 2022, Limassol, Cyprus, October 3-7, 2022}, pp.\  582--591. {IEEE}, 2022.
\newblock \doi{10.1109/ICSME55016.2022.00080}.
\newblock URL \url{https://doi.org/10.1109/ICSME55016.2022.00080}.

\bibitem[Choi et~al.(2021)Choi, Shin, Kim, and Shin]{sql2020-2}
DongHyun Choi, Myeongcheol Shin, EungGyun Kim, and Dong~Ryeol Shin.
\newblock {RYANSQL:} recursively applying sketch-based slot fillings for complex text-to-sql in cross-domain databases.
\newblock \emph{Comput. Linguistics}, 47\penalty0 (2):\penalty0 309--332, 2021.
\newblock \doi{10.1162/coli\_a\_00403}.
\newblock URL \url{https://doi.org/10.1162/coli\_a\_00403}.

\bibitem[Choromanski et~al.(2021)Choromanski, Likhosherstov, Dohan, Song, Gane, Sarl{\'{o}}s, Hawkins, Davis, Mohiuddin, Kaiser, Belanger, Colwell, and Weller]{2020Performer}
Krzysztof~Marcin Choromanski, Valerii Likhosherstov, David Dohan, Xingyou Song, Andreea Gane, Tam{\'{a}}s Sarl{\'{o}}s, Peter Hawkins, Jared~Quincy Davis, Afroz Mohiuddin, Lukasz Kaiser, David~Benjamin Belanger, Lucy~J. Colwell, and Adrian Weller.
\newblock Rethinking attention with performers.
\newblock In \emph{9th International Conference on Learning Representations, {ICLR} 2021, Virtual Event, Austria, May 3-7, 2021}. OpenReview.net, 2021.
\newblock URL \url{https://openreview.net/forum?id=Ua6zuk0WRH}.

\bibitem[Chowdhery et~al.(2023)Chowdhery, Narang, Devlin, Bosma, Mishra, Roberts, Barham, Chung, Sutton, Gehrmann, Schuh, Shi, Tsvyashchenko, Maynez, Rao, Barnes, Tay, Shazeer, Prabhakaran, Reif, Du, Hutchinson, Pope, Bradbury, Austin, Isard, Gur{-}Ari, Yin, Duke, Levskaya, Ghemawat, Dev, Michalewski, Garcia, Misra, Robinson, Fedus, Zhou, Ippolito, Luan, Lim, Zoph, Spiridonov, Sepassi, Dohan, Agrawal, Omernick, Dai, Pillai, Pellat, Lewkowycz, Moreira, Child, Polozov, Lee, Zhou, Wang, Saeta, Diaz, Firat, Catasta, Wei, Meier{-}Hellstern, Eck, Dean, Petrov, and Fiedel]{2022PaLM}
Aakanksha Chowdhery, Sharan Narang, Jacob Devlin, Maarten Bosma, Gaurav Mishra, Adam Roberts, Paul Barham, Hyung~Won Chung, Charles Sutton, Sebastian Gehrmann, Parker Schuh, Kensen Shi, Sasha Tsvyashchenko, Joshua Maynez, Abhishek Rao, Parker Barnes, Yi~Tay, Noam Shazeer, Vinodkumar Prabhakaran, Emily Reif, Nan Du, Ben Hutchinson, Reiner Pope, James Bradbury, Jacob Austin, Michael Isard, Guy Gur{-}Ari, Pengcheng Yin, Toju Duke, Anselm Levskaya, Sanjay Ghemawat, Sunipa Dev, Henryk Michalewski, Xavier Garcia, Vedant Misra, Kevin Robinson, Liam Fedus, Denny Zhou, Daphne Ippolito, David Luan, Hyeontaek Lim, Barret Zoph, Alexander Spiridonov, Ryan Sepassi, David Dohan, Shivani Agrawal, Mark Omernick, Andrew~M. Dai, Thanumalayan~Sankaranarayana Pillai, Marie Pellat, Aitor Lewkowycz, Erica Moreira, Rewon Child, Oleksandr Polozov, Katherine Lee, Zongwei Zhou, Xuezhi Wang, Brennan Saeta, Mark Diaz, Orhan Firat, Michele Catasta, Jason Wei, Kathy Meier{-}Hellstern, Douglas Eck, Jeff Dean, Slav Petrov, and Noah Fiedel.
\newblock Palm: Scaling language modeling with pathways.
\newblock \emph{J. Mach. Learn. Res.}, 24:\penalty0 240:1--240:113, 2023.
\newblock URL \url{http://jmlr.org/papers/v24/22-1144.html}.

\bibitem[Christiano et~al.(2017)Christiano, Leike, Brown, Martic, Legg, and Amodei]{2017RL-human}
Paul~F. Christiano, Jan Leike, Tom~B. Brown, Miljan Martic, Shane Legg, and Dario Amodei.
\newblock Deep reinforcement learning from human preferences.
\newblock In Isabelle Guyon, Ulrike von Luxburg, Samy Bengio, Hanna~M. Wallach, Rob Fergus, S.~V.~N. Vishwanathan, and Roman Garnett (eds.), \emph{Advances in Neural Information Processing Systems 30: Annual Conference on Neural Information Processing Systems 2017, December 4-9, 2017, Long Beach, CA, {USA}}, pp.\  4299--4307, 2017.
\newblock URL \url{https://proceedings.neurips.cc/paper/2017/hash/d5e2c0adad503c91f91df240d0cd4e49-Abstract.html}.

\bibitem[Christopoulou et~al.(2022)Christopoulou, Lampouras, Gritta, Zhang, Guo, Li, Zhang, Xiao, Shen, Li, Yu, Yan, Zhou, Wang, Ma, Iacobacci, Wang, Liang, Wei, Jiang, Wang, and Liu]{2022Pangu-Coder}
Fenia Christopoulou, Gerasimos Lampouras, Milan Gritta, Guchun Zhang, Yinpeng Guo, Zhongqi Li, Qi~Zhang, Meng Xiao, Bo~Shen, Lin Li, Hao Yu, Li~Yan, Pingyi Zhou, Xin Wang, Yuchi Ma, Ignacio Iacobacci, Yasheng Wang, Guangtai Liang, Jiansheng Wei, Xin Jiang, Qianxiang Wang, and Qun Liu.
\newblock Pangu-coder: Program synthesis with function-level language modeling.
\newblock \emph{CoRR}, abs/2207.11280, 2022.
\newblock \doi{10.48550/arXiv.2207.11280}.
\newblock URL \url{https://doi.org/10.48550/arXiv.2207.11280}.

\bibitem[Chung et~al.(2022)Chung, Hou, Longpre, Zoph, Tay, Fedus, Li, Wang, Dehghani, Brahma, Webson, Gu, Dai, Suzgun, Chen, Chowdhery, Narang, Mishra, Yu, Zhao, Huang, Dai, Yu, Petrov, Chi, Dean, Devlin, Roberts, Zhou, Le, and Wei]{2022FLAN}
Hyung~Won Chung, Le~Hou, Shayne Longpre, Barret Zoph, Yi~Tay, William Fedus, Eric Li, Xuezhi Wang, Mostafa Dehghani, Siddhartha Brahma, Albert Webson, Shixiang~Shane Gu, Zhuyun Dai, Mirac Suzgun, Xinyun Chen, Aakanksha Chowdhery, Sharan Narang, Gaurav Mishra, Adams Yu, Vincent~Y. Zhao, Yanping Huang, Andrew~M. Dai, Hongkun Yu, Slav Petrov, Ed~H. Chi, Jeff Dean, Jacob Devlin, Adam Roberts, Denny Zhou, Quoc~V. Le, and Jason Wei.
\newblock Scaling instruction-finetuned language models.
\newblock \emph{CoRR}, abs/2210.11416, 2022.
\newblock \doi{10.48550/arXiv.2210.11416}.
\newblock URL \url{https://doi.org/10.48550/arXiv.2210.11416}.

\bibitem[Clark et~al.(2020)Clark, Luong, Le, and Manning]{2020ELECTRA}
Kevin Clark, Minh{-}Thang Luong, Quoc~V. Le, and Christopher~D. Manning.
\newblock {ELECTRA:} pre-training text encoders as discriminators rather than generators.
\newblock In \emph{8th International Conference on Learning Representations, {ICLR} 2020, Addis Ababa, Ethiopia, April 26-30, 2020}. OpenReview.net, 2020.
\newblock URL \url{https://openreview.net/forum?id=r1xMH1BtvB}.

\bibitem[Clement et~al.(2020)Clement, Drain, Timcheck, Svyatkovskiy, and Sundaresan]{2020PyMT5}
Colin~B. Clement, Dawn Drain, Jonathan Timcheck, Alexey Svyatkovskiy, and Neel Sundaresan.
\newblock Pymt5: multi-mode translation of natural language and python code with transformers.
\newblock In Bonnie Webber, Trevor Cohn, Yulan He, and Yang Liu (eds.), \emph{Proceedings of the 2020 Conference on Empirical Methods in Natural Language Processing, {EMNLP} 2020, Online, November 16-20, 2020}, pp.\  9052--9065. Association for Computational Linguistics, 2020.
\newblock \doi{10.18653/v1/2020.emnlp-main.728}.
\newblock URL \url{https://doi.org/10.18653/v1/2020.emnlp-main.728}.

\bibitem[Clement et~al.(2021)Clement, Lu, Liu, Tufano, Drain, Duan, Sundaresan, and Svyatkovskiy]{2021eWASH}
Colin~B. Clement, Shuai Lu, Xiaoyu Liu, Michele Tufano, Dawn Drain, Nan Duan, Neel Sundaresan, and Alexey Svyatkovskiy.
\newblock Long-range modeling of source code files with ewash: Extended window access by syntax hierarchy.
\newblock In Marie{-}Francine Moens, Xuanjing Huang, Lucia Specia, and Scott~Wen{-}tau Yih (eds.), \emph{Proceedings of the 2021 Conference on Empirical Methods in Natural Language Processing, {EMNLP} 2021, Virtual Event / Punta Cana, Dominican Republic, 7-11 November, 2021}, pp.\  4713--4722. Association for Computational Linguistics, 2021.
\newblock \doi{10.18653/V1/2021.EMNLP-MAIN.387}.
\newblock URL \url{https://doi.org/10.18653/v1/2021.emnlp-main.387}.

\bibitem[Cobbe et~al.(2021)Cobbe, Kosaraju, Bavarian, Hilton, Nakano, Hesse, and Schulman]{2021GSM8K}
Karl Cobbe, Vineet Kosaraju, Mohammad Bavarian, Jacob Hilton, Reiichiro Nakano, Christopher Hesse, and John Schulman.
\newblock Training verifiers to solve math word problems.
\newblock \emph{CoRR}, abs/2110.14168, 2021.
\newblock URL \url{https://arxiv.org/abs/2110.14168}.

\bibitem[Coles et~al.(2016)Coles, Laurent, Henard, Papadakis, and Ventresque]{mutant2016-2}
Henry Coles, Thomas Laurent, Christopher Henard, Mike Papadakis, and Anthony Ventresque.
\newblock {PIT:} a practical mutation testing tool for java (demo).
\newblock In Andreas Zeller and Abhik Roychoudhury (eds.), \emph{Proceedings of the 25th International Symposium on Software Testing and Analysis, {ISSTA} 2016, Saarbr{\"{u}}cken, Germany, July 18-20, 2016}, pp.\  449--452. {ACM}, 2016.
\newblock \doi{10.1145/2931037.2948707}.
\newblock URL \url{https://doi.org/10.1145/2931037.2948707}.

\bibitem[Collberg \& Thomborson(2002)Collberg and Thomborson]{ob2000-1}
{Christian S.} Collberg and Clark Thomborson.
\newblock Watermarking, tamper-proofing, and obfuscation - tools for software protection.
\newblock \emph{IEEE Transactions on Software Engineering}, 28\penalty0 (8):\penalty0 735--746, August 2002.
\newblock ISSN 0098-5589.
\newblock \doi{10.1109/TSE.2002.1027797}.
\newblock Funding Information: The authors are grateful for the extensive and insightful comments of two anonymous referees. This material is based upon work supported by the US National Science Foundation under Grant No. 0073483.

\bibitem[Conneau \& Lample(2019)Conneau and Lample]{2019XLM}
Alexis Conneau and Guillaume Lample.
\newblock Cross-lingual language model pretraining.
\newblock In Hanna~M. Wallach, Hugo Larochelle, Alina Beygelzimer, Florence d'Alch{\'{e}}{-}Buc, Emily~B. Fox, and Roman Garnett (eds.), \emph{Advances in Neural Information Processing Systems 32: Annual Conference on Neural Information Processing Systems 2019, NeurIPS 2019, December 8-14, 2019, Vancouver, BC, Canada}, pp.\  7057--7067, 2019.
\newblock URL \url{https://proceedings.neurips.cc/paper/2019/hash/c04c19c2c2474dbf5f7ac4372c5b9af1-Abstract.html}.

\bibitem[Conneau et~al.(2020)Conneau, Khandelwal, Goyal, Chaudhary, Wenzek, Guzm{\'{a}}n, Grave, Ott, Zettlemoyer, and Stoyanov]{2019XLM-R}
Alexis Conneau, Kartikay Khandelwal, Naman Goyal, Vishrav Chaudhary, Guillaume Wenzek, Francisco Guzm{\'{a}}n, Edouard Grave, Myle Ott, Luke Zettlemoyer, and Veselin Stoyanov.
\newblock Unsupervised cross-lingual representation learning at scale.
\newblock In Dan Jurafsky, Joyce Chai, Natalie Schluter, and Joel~R. Tetreault (eds.), \emph{Proceedings of the 58th Annual Meeting of the Association for Computational Linguistics, {ACL} 2020, Online, July 5-10, 2020}, pp.\  8440--8451. Association for Computational Linguistics, 2020.
\newblock \doi{10.18653/v1/2020.acl-main.747}.
\newblock URL \url{https://doi.org/10.18653/v1/2020.acl-main.747}.

\bibitem[Cortes{-}Coy et~al.(2014)Cortes{-}Coy, V{\'{a}}squez, Aponte, and Poshyvanyk]{commit2014-1}
Luis~Fernando Cortes{-}Coy, Mario~Linares V{\'{a}}squez, Jairo Aponte, and Denys Poshyvanyk.
\newblock On automatically generating commit messages via summarization of source code changes.
\newblock In \emph{14th {IEEE} International Working Conference on Source Code Analysis and Manipulation, {SCAM} 2014, Victoria, BC, Canada, September 28-29, 2014}, pp.\  275--284. {IEEE} Computer Society, 2014.
\newblock \doi{10.1109/SCAM.2014.14}.
\newblock URL \url{https://doi.org/10.1109/SCAM.2014.14}.

\bibitem[Csuvik \& Vid{\'{a}}cs(2022)Csuvik and Vid{\'{a}}cs]{fix-data-2022-2}
Viktor Csuvik and L{\'{a}}szl{\'{o}} Vid{\'{a}}cs.
\newblock Fixjs: {A} dataset of bug-fixing javascript commits.
\newblock In \emph{19th {IEEE/ACM} International Conference on Mining Software Repositories, {MSR} 2022, Pittsburgh, PA, USA, May 23-24, 2022}, pp.\  712--716. {ACM}, 2022.
\newblock \doi{10.1145/3524842.3528480}.
\newblock URL \url{https://doi.org/10.1145/3524842.3528480}.

\bibitem[Cui et~al.(2022)Cui, Wang, Huang, Inala, Mytkowicz, Wang, Gao, and Duan]{doc2022-2}
Haotian Cui, Chenglong Wang, Junjie Huang, Jeevana~Priya Inala, Todd Mytkowicz, Bo~Wang, Jianfeng Gao, and Nan Duan.
\newblock Codeexp: Explanatory code document generation.
\newblock In Yoav Goldberg, Zornitsa Kozareva, and Yue Zhang (eds.), \emph{Findings of the Association for Computational Linguistics: {EMNLP} 2022, Abu Dhabi, United Arab Emirates, December 7-11, 2022}, pp.\  2342--2354. Association for Computational Linguistics, 2022.
\newblock \doi{10.18653/V1/2022.FINDINGS-EMNLP.174}.
\newblock URL \url{https://doi.org/10.18653/v1/2022.findings-emnlp.174}.

\bibitem[Cui et~al.(2021)Cui, Zhao, Dai, Wang, Huang, and Huang]{type2021-3}
Siwei Cui, Gang Zhao, Zeyu Dai, Luochao Wang, Ruihong Huang, and Jeff Huang.
\newblock Pyinfer: Deep learning semantic type inference for python variables.
\newblock \emph{CoRR}, abs/2106.14316, 2021.
\newblock URL \url{https://arxiv.org/abs/2106.14316}.

\bibitem[Cummins et~al.(2017)Cummins, Petoumenos, Wang, and Leather]{comp-opt2017-1}
Chris Cummins, Pavlos Petoumenos, Zheng Wang, and Hugh Leather.
\newblock End-to-end deep learning of optimization heuristics.
\newblock In \emph{26th International Conference on Parallel Architectures and Compilation Techniques, {PACT} 2017, Portland, OR, USA, September 9-13, 2017}, pp.\  219--232. {IEEE} Computer Society, 2017.
\newblock \doi{10.1109/PACT.2017.24}.
\newblock URL \url{https://doi.org/10.1109/PACT.2017.24}.

\bibitem[Cummins et~al.(2021)Cummins, Fisches, Ben-Nun, Hoefler, O'Boyle, and Leather]{2020ProGraML}
Chris Cummins, Zacharias~V. Fisches, Tal Ben-Nun, Torsten Hoefler, Michael F~P O'Boyle, and Hugh Leather.
\newblock Programl: A graph-based program representation for data flow analysis and compiler optimizations.
\newblock In Marina Meila and Tong Zhang (eds.), \emph{Proceedings of the 38th International Conference on Machine Learning}, volume 139 of \emph{Proceedings of Machine Learning Research}, pp.\  2244--2253. PMLR, 18--24 Jul 2021.
\newblock URL \url{https://proceedings.mlr.press/v139/cummins21a.html}.

\bibitem[Cummins et~al.(2023)Cummins, Seeker, Grubisic, Elhoushi, Liang, Rozi{\`{e}}re, Gehring, Gloeckle, Hazelwood, Synnaeve, and Leather]{comp-opt2023-1}
Chris Cummins, Volker Seeker, Dejan Grubisic, Mostafa Elhoushi, Youwei Liang, Baptiste Rozi{\`{e}}re, Jonas Gehring, Fabian Gloeckle, Kim~M. Hazelwood, Gabriel Synnaeve, and Hugh Leather.
\newblock Large language models for compiler optimization.
\newblock \emph{CoRR}, abs/2309.07062, 2023.
\newblock \doi{10.48550/ARXIV.2309.07062}.
\newblock URL \url{https://doi.org/10.48550/arXiv.2309.07062}.

\bibitem[Dahl et~al.(1994)Dahl, Bates, Brown, Fisher, Hunicke{-}Smith, Pallett, Pao, Rudnicky, and Shriberg]{sql-data-1994}
Deborah~A. Dahl, Madeleine Bates, Michael Brown, William~M. Fisher, Kate Hunicke{-}Smith, David~S. Pallett, Christine Pao, Alexander~I. Rudnicky, and Elizabeth Shriberg.
\newblock Expanding the scope of the {ATIS} task: The {ATIS-3} corpus.
\newblock In \emph{Human Language Technology, Proceedings of a Workshop held at Plainsboro, New Jerey, USA, March 8-11, 1994}. Morgan Kaufmann, 1994.
\newblock URL \url{https://aclanthology.org/H94-1010/}.

\bibitem[Dai et~al.(2024{\natexlab{a}})Dai, Deng, Zhao, Xu, Gao, Chen, Li, Zeng, Yu, Wu, Xie, Li, Huang, Luo, Ruan, Sui, and Liang]{2024DeepSeekMoE}
Damai Dai, Chengqi Deng, Chenggang Zhao, R.~X. Xu, Huazuo Gao, Deli Chen, Jiashi Li, Wangding Zeng, Xingkai Yu, Y.~Wu, Zhenda Xie, Y.~K. Li, Panpan Huang, Fuli Luo, Chong Ruan, Zhifang Sui, and Wenfeng Liang.
\newblock Deepseekmoe: Towards ultimate expert specialization in mixture-of-experts language models.
\newblock \emph{CoRR}, abs/2401.06066, 2024{\natexlab{a}}.
\newblock \doi{10.48550/ARXIV.2401.06066}.
\newblock URL \url{https://doi.org/10.48550/arXiv.2401.06066}.

\bibitem[Dai et~al.(2022)Dai, Li, Chen, Shang, and Chen]{log2020-1}
Hetong Dai, Heng Li, Che{-}Shao Chen, Weiyi Shang, and Tse{-}Hsun Chen.
\newblock Logram: Efficient log parsing using {\textdollar}n{\textdollar}n-gram dictionaries.
\newblock \emph{{IEEE} Trans. Software Eng.}, 48\penalty0 (3):\penalty0 879--892, 2022.
\newblock \doi{10.1109/TSE.2020.3007554}.
\newblock URL \url{https://doi.org/10.1109/TSE.2020.3007554}.

\bibitem[Dai et~al.(2024{\natexlab{b}})Dai, Lu, Feng, Ruan, Cheng, Tan, and Guo]{2024MHPP}
Jianbo Dai, Jianqiao Lu, Yunlong Feng, Rongju Ruan, Ming Cheng, Haochen Tan, and Zhijiang Guo.
\newblock Mhpp: Exploring the capabilities and limitations of language models beyond basic code generation.
\newblock 2024{\natexlab{b}}.
\newblock URL \url{https://doi.org/10.48550/arXiv.2405.11430}.

\bibitem[Dakhel et~al.(2023)Dakhel, Nikanjam, Majdinasab, Khomh, and Desmarais]{2023MuTAP}
Arghavan~Moradi Dakhel, Amin Nikanjam, Vahid Majdinasab, Foutse Khomh, and Michel~C. Desmarais.
\newblock Effective test generation using pre-trained large language models and mutation testing.
\newblock \emph{CoRR}, abs/2308.16557, 2023.
\newblock \doi{10.48550/ARXIV.2308.16557}.
\newblock URL \url{https://doi.org/10.48550/arXiv.2308.16557}.

\bibitem[Dao et~al.(2022)Dao, Fu, Ermon, Rudra, and R{\'{e}}]{2022FlashAttention}
Tri Dao, Daniel~Y. Fu, Stefano Ermon, Atri Rudra, and Christopher R{\'{e}}.
\newblock Flashattention: Fast and memory-efficient exact attention with io-awareness.
\newblock In \emph{NeurIPS}, 2022.
\newblock URL \url{http://papers.nips.cc/paper\_files/paper/2022/hash/67d57c32e20fd0a7a302cb81d36e40d5-Abstract-Conference.html}.

\bibitem[de~Ara{\'{u}}jo \& Marcacini(2021)de~Ara{\'{u}}jo and Marcacini]{re-ana2021-1}
Adailton~Ferreira de~Ara{\'{u}}jo and Ricardo~Marcondes Marcacini.
\newblock {RE-BERT:} automatic extraction of software requirements from app reviews using {BERT} language model.
\newblock In Chih{-}Cheng Hung, Jiman Hong, Alessio Bechini, and Eunjee Song (eds.), \emph{{SAC} '21: The 36th {ACM/SIGAPP} Symposium on Applied Computing, Virtual Event, Republic of Korea, March 22-26, 2021}, pp.\  1321--1327. {ACM}, 2021.
\newblock \doi{10.1145/3412841.3442006}.
\newblock URL \url{https://doi.org/10.1145/3412841.3442006}.

\bibitem[de~Sousa \& Hasselbring(2021)de~Sousa and Hasselbring]{cloze2021-1}
Nelson~Tavares de~Sousa and Wilhelm Hasselbring.
\newblock Javabert: Training a transformer-based model for the java programming language.
\newblock In \emph{36th {IEEE/ACM} International Conference on Automated Software Engineering, {ASE} 2021 - Workshops, Melbourne, Australia, November 15-19, 2021}, pp.\  90--95. {IEEE}, 2021.
\newblock \doi{10.1109/ASEW52652.2021.00028}.
\newblock URL \url{https://doi.org/10.1109/ASEW52652.2021.00028}.

\bibitem[DeepSeek-AI et~al.(2024)DeepSeek-AI, Bi, Chen, Chen, Chen, Dai, Deng, Ding, Dong, Du, Fu, Gao, Gao, Gao, Ge, Guan, Guo, Guo, Hao, Hao, He, Hu, Huang, Li, Li, Li, Li, Li, Liang, Lin, Liu, Liu, Liu, Liu, Liu, Liu, Lu, Lu, Luo, Ma, Nie, Pei, Piao, Qiu, Qu, Ren, Ren, Ruan, Sha, Shao, Song, Su, Sun, Sun, Tang, Wang, Wang, Wang, Wang, Wang, Wu, Wu, Xie, Xie, Xie, Xiong, Xu, Xu, Xu, Yang, You, Yu, Yu, Zhang, Zhang, Zhang, Zhang, Zhang, Zhang, Zhang, Zhang, Zhao, Zhao, Zhou, Zhou, Zhu, and Zou]{2024DeepSeek}
DeepSeek-AI, Xiao Bi, Deli Chen, Guanting Chen, Shanhuang Chen, Damai Dai, Chengqi Deng, Honghui Ding, Kai Dong, Qiushi Du, Zhe Fu, Huazuo Gao, Kaige Gao, Wenjun Gao, Ruiqi Ge, Kang Guan, Daya Guo, Jianzhong Guo, Guangbo Hao, Zhewen Hao, Ying He, Wenjie Hu, Panpan Huang, Erhang Li, Guowei Li, Jiashi Li, Yao Li, Y.~K. Li, Wenfeng Liang, Fangyun Lin, A.~X. Liu, Bo~Liu, Wen Liu, Xiaodong Liu, Xin Liu, Yiyuan Liu, Haoyu Lu, Shanghao Lu, Fuli Luo, Shirong Ma, Xiaotao Nie, Tian Pei, Yishi Piao, Junjie Qiu, Hui Qu, Tongzheng Ren, Zehui Ren, Chong Ruan, Zhangli Sha, Zhihong Shao, Junxiao Song, Xuecheng Su, Jingxiang Sun, Yaofeng Sun, Minghui Tang, Bingxuan Wang, Peiyi Wang, Shiyu Wang, Yaohui Wang, Yongji Wang, Tong Wu, Y.~Wu, Xin Xie, Zhenda Xie, Ziwei Xie, Yiliang Xiong, Hanwei Xu, R.~X. Xu, Yanhong Xu, Dejian Yang, Yuxiang You, Shuiping Yu, Xingkai Yu, B.~Zhang, Haowei Zhang, Lecong Zhang, Liyue Zhang, Mingchuan Zhang, Minghua Zhang, Wentao Zhang, Yichao Zhang, Chenggang Zhao, Yao Zhao, Shangyan Zhou, Shunfeng
  Zhou, Qihao Zhu, and Yuheng Zou.
\newblock Deepseek llm: Scaling open-source language models with longtermism.
\newblock \emph{CoRR}, abs/2401.02954, 2024.
\newblock \doi{10.48550/ARXIV.2401.02954}.
\newblock URL \url{https://doi.org/10.48550/arXiv.2401.02954}.

\bibitem[Deeptimahanti \& Babar(2009)Deeptimahanti and Babar]{re-model2009-1}
Deva~Kumar Deeptimahanti and Muhammad~Ali Babar.
\newblock An automated tool for generating {UML} models from natural language requirements.
\newblock In \emph{{ASE} 2009, 24th {IEEE/ACM} International Conference on Automated Software Engineering, Auckland, New Zealand, November 16-20, 2009}, pp.\  680--682. {IEEE} Computer Society, 2009.
\newblock \doi{10.1109/ASE.2009.48}.
\newblock URL \url{https://doi.org/10.1109/ASE.2009.48}.

\bibitem[Degiovanni \& Papadakis(2022)Degiovanni and Papadakis]{mutant2022-1}
Renzo Degiovanni and Mike Papadakis.
\newblock {\(\mathrm{\mu}\)}bert: Mutation testing using pre-trained language models.
\newblock In \emph{15th {IEEE} International Conference on Software Testing, Verification and Validation Workshops {ICST} Workshops 2022, Valencia, Spain, April 4-13, 2022}, pp.\  160--169. {IEEE}, 2022.
\newblock \doi{10.1109/ICSTW55395.2022.00039}.
\newblock URL \url{https://doi.org/10.1109/ICSTW55395.2022.00039}.

\bibitem[Deng et~al.(2022{\natexlab{a}})Deng, Chen, and Zhang]{sql-survey-2022-2}
Naihao Deng, Yulong Chen, and Yue Zhang.
\newblock Recent advances in text-to-sql: {A} survey of what we have and what we expect.
\newblock In Nicoletta Calzolari, Chu{-}Ren Huang, Hansaem Kim, James Pustejovsky, Leo Wanner, Key{-}Sun Choi, Pum{-}Mo Ryu, Hsin{-}Hsi Chen, Lucia Donatelli, Heng Ji, Sadao Kurohashi, Patrizia Paggio, Nianwen Xue, Seokhwan Kim, Younggyun Hahm, Zhong He, Tony~Kyungil Lee, Enrico Santus, Francis Bond, and Seung{-}Hoon Na (eds.), \emph{Proceedings of the 29th International Conference on Computational Linguistics, {COLING} 2022, Gyeongju, Republic of Korea, October 12-17, 2022}, pp.\  2166--2187. International Committee on Computational Linguistics, 2022{\natexlab{a}}.
\newblock URL \url{https://aclanthology.org/2022.coling-1.190}.

\bibitem[Deng et~al.(2021)Deng, Awadallah, Meek, Polozov, Sun, and Richardson]{sql2020-4.5}
Xiang Deng, Ahmed~Hassan Awadallah, Christopher Meek, Oleksandr Polozov, Huan Sun, and Matthew Richardson.
\newblock Structure-grounded pretraining for text-to-sql.
\newblock In Kristina Toutanova, Anna Rumshisky, Luke Zettlemoyer, Dilek Hakkani{-}T{\"{u}}r, Iz~Beltagy, Steven Bethard, Ryan Cotterell, Tanmoy Chakraborty, and Yichao Zhou (eds.), \emph{Proceedings of the 2021 Conference of the North American Chapter of the Association for Computational Linguistics: Human Language Technologies, {NAACL-HLT} 2021, Online, June 6-11, 2021}, pp.\  1337--1350. Association for Computational Linguistics, 2021.
\newblock \doi{10.18653/V1/2021.NAACL-MAIN.105}.
\newblock URL \url{https://doi.org/10.18653/v1/2021.naacl-main.105}.

\bibitem[Deng et~al.(2022{\natexlab{b}})Deng, Shiralkar, Lockard, Huang, and Sun]{UI2022-1}
Xiang Deng, Prashant Shiralkar, Colin Lockard, Binxuan Huang, and Huan Sun.
\newblock {DOM-LM:} learning generalizable representations for {HTML} documents.
\newblock \emph{CoRR}, abs/2201.10608, 2022{\natexlab{b}}.
\newblock URL \url{https://arxiv.org/abs/2201.10608}.

\bibitem[Deng et~al.(2022{\natexlab{c}})Deng, Yang, Wei, and Zhang]{fuzz2022-5}
Yinlin Deng, Chenyuan Yang, Anjiang Wei, and Lingming Zhang.
\newblock Fuzzing deep-learning libraries via automated relational {API} inference.
\newblock In Abhik Roychoudhury, Cristian Cadar, and Miryung Kim (eds.), \emph{Proceedings of the 30th {ACM} Joint European Software Engineering Conference and Symposium on the Foundations of Software Engineering, {ESEC/FSE} 2022, Singapore, Singapore, November 14-18, 2022}, pp.\  44--56. {ACM}, 2022{\natexlab{c}}.
\newblock \doi{10.1145/3540250.3549085}.
\newblock URL \url{https://doi.org/10.1145/3540250.3549085}.

\bibitem[Deng et~al.(2023)Deng, Xia, Peng, Yang, and Zhang]{fuzz2022-7}
Yinlin Deng, Chunqiu~Steven Xia, Haoran Peng, Chenyuan Yang, and Lingming Zhang.
\newblock Large language models are zero-shot fuzzers: Fuzzing deep-learning libraries via large language models.
\newblock In Ren{\'{e}} Just and Gordon Fraser (eds.), \emph{Proceedings of the 32nd {ACM} {SIGSOFT} International Symposium on Software Testing and Analysis, {ISSTA} 2023, Seattle, WA, USA, July 17-21, 2023}, pp.\  423--435. {ACM}, 2023.
\newblock \doi{10.1145/3597926.3598067}.
\newblock URL \url{https://doi.org/10.1145/3597926.3598067}.

\bibitem[Devlin et~al.(2017{\natexlab{a}})Devlin, Uesato, Bhupatiraju, Singh, Mohamed, and Kohli]{syn2017-1}
Jacob Devlin, Jonathan Uesato, Surya Bhupatiraju, Rishabh Singh, Abdel{-}rahman Mohamed, and Pushmeet Kohli.
\newblock Robustfill: Neural program learning under noisy {I/O}.
\newblock In Doina Precup and Yee~Whye Teh (eds.), \emph{Proceedings of the 34th International Conference on Machine Learning, {ICML} 2017, Sydney, NSW, Australia, 6-11 August 2017}, volume~70 of \emph{Proceedings of Machine Learning Research}, pp.\  990--998. {PMLR}, 2017{\natexlab{a}}.
\newblock URL \url{http://proceedings.mlr.press/v70/devlin17a.html}.

\bibitem[Devlin et~al.(2017{\natexlab{b}})Devlin, Uesato, Singh, and Kohli]{fix2017-2}
Jacob Devlin, Jonathan Uesato, Rishabh Singh, and Pushmeet Kohli.
\newblock Semantic code repair using neuro-symbolic transformation networks.
\newblock \emph{CoRR}, abs/1710.11054, 2017{\natexlab{b}}.
\newblock URL \url{http://arxiv.org/abs/1710.11054}.

\bibitem[Devlin et~al.(2019)Devlin, Chang, Lee, and Toutanova]{2018BERT}
Jacob Devlin, Ming{-}Wei Chang, Kenton Lee, and Kristina Toutanova.
\newblock {BERT:} pre-training of deep bidirectional transformers for language understanding.
\newblock In Jill Burstein, Christy Doran, and Thamar Solorio (eds.), \emph{Proceedings of the 2019 Conference of the North American Chapter of the Association for Computational Linguistics: Human Language Technologies, {NAACL-HLT} 2019, Minneapolis, MN, USA, June 2-7, 2019, Volume 1 (Long and Short Papers)}, pp.\  4171--4186. Association for Computational Linguistics, 2019.
\newblock \doi{10.18653/v1/n19-1423}.
\newblock URL \url{https://doi.org/10.18653/v1/n19-1423}.

\bibitem[Di et~al.(2023)Di, Li, Yu, Jiang, Cai, Cao, Chen, Chen, Chen, Chen, Fan, Gong, Gong, Hu, Guo, Lei, Li, Li, Liang, Liao, Liu, Liu, Liu, Lu, Shen, Wang, Wang, Wang, Xu, Yang, Ye, Zhang, Zhang, Zhao, Zheng, Zhou, Zhu, and Zhu]{2023CodeFuse13B}
Peng Di, Jianguo Li, Hang Yu, Wei Jiang, Wenting Cai, Yang Cao, Chaoyu Chen, Dajun Chen, Hongwei Chen, Liang Chen, Gang Fan, Jie Gong, Zi~Gong, Wen Hu, Tingting Guo, Zhichao Lei, Ting Li, Zheng Li, Ming Liang, Cong Liao, Bingchang Liu, Jiachen Liu, Zhiwei Liu, Shaojun Lu, Min Shen, Guangpei Wang, Huan Wang, Zhi Wang, Zhaogui Xu, Jiawei Yang, Qing Ye, Gehao Zhang, Yu~Zhang, Zelin Zhao, Xunjin Zheng, Hailian Zhou, Lifu Zhu, and Xianying Zhu.
\newblock Codefuse-13b: {A} pretrained multi-lingual code large language model.
\newblock \emph{CoRR}, abs/2310.06266, 2023.
\newblock \doi{10.48550/ARXIV.2310.06266}.
\newblock URL \url{https://doi.org/10.48550/arXiv.2310.06266}.

\bibitem[Dinella et~al.(2022)Dinella, Ryan, Mytkowicz, and Lahiri]{assert2021-1}
Elizabeth Dinella, Gabriel Ryan, Todd Mytkowicz, and Shuvendu~K. Lahiri.
\newblock {TOGA:} {A} neural method for test oracle generation.
\newblock In \emph{44th {IEEE/ACM} 44th International Conference on Software Engineering, {ICSE} 2022, Pittsburgh, PA, USA, May 25-27, 2022}, pp.\  2130--2141. {ACM}, 2022.
\newblock \doi{10.1145/3510003.3510141}.
\newblock URL \url{https://doi.org/10.1145/3510003.3510141}.

\bibitem[Ding et~al.(2022{\natexlab{a}})Ding, Buratti, Pujar, Morari, Ray, and Chakraborty]{2021DISCO}
Yangruibo Ding, Luca Buratti, Saurabh Pujar, Alessandro Morari, Baishakhi Ray, and Saikat Chakraborty.
\newblock Towards learning (dis)-similarity of source code from program contrasts.
\newblock In Smaranda Muresan, Preslav Nakov, and Aline Villavicencio (eds.), \emph{Proceedings of the 60th Annual Meeting of the Association for Computational Linguistics (Volume 1: Long Papers), {ACL} 2022, Dublin, Ireland, May 22-27, 2022}, pp.\  6300--6312. Association for Computational Linguistics, 2022{\natexlab{a}}.
\newblock \doi{10.18653/v1/2022.acl-long.436}.
\newblock URL \url{https://doi.org/10.18653/v1/2022.acl-long.436}.

\bibitem[Ding et~al.(2022{\natexlab{b}})Ding, Wang, Ahmad, Ramanathan, Nallapati, Bhatia, Roth, and Xiang]{repo2022-2}
Yangruibo Ding, Zijian Wang, Wasi~Uddin Ahmad, Murali~Krishna Ramanathan, Ramesh Nallapati, Parminder Bhatia, Dan Roth, and Bing Xiang.
\newblock Cocomic: Code completion by jointly modeling in-file and cross-file context.
\newblock \emph{CoRR}, abs/2212.10007, 2022{\natexlab{b}}.
\newblock \doi{10.48550/ARXIV.2212.10007}.
\newblock URL \url{https://doi.org/10.48550/arXiv.2212.10007}.

\bibitem[Ding et~al.(2023)Ding, Wang, Ahmad, Ding, Tan, Jain, Ramanathan, Nallapati, Bhatia, Roth, and Xiang]{2023CrossCodeEval}
Yangruibo Ding, Zijian Wang, Wasi~Uddin Ahmad, Hantian Ding, Ming Tan, Nihal Jain, Murali~Krishna Ramanathan, Ramesh Nallapati, Parminder Bhatia, Dan Roth, and Bing Xiang.
\newblock Crosscodeeval: {A} diverse and multilingual benchmark for cross-file code completion.
\newblock \emph{CoRR}, abs/2310.11248, 2023.
\newblock \doi{10.48550/ARXIV.2310.11248}.
\newblock URL \url{https://doi.org/10.48550/arXiv.2310.11248}.

\bibitem[Ding et~al.(2024)Ding, Min, Kaiser, and Ray]{2024Cycle}
Yangruibo Ding, Marcus~J. Min, Gail~E. Kaiser, and Baishakhi Ray.
\newblock {CYCLE:} learning to self-refine the code generation.
\newblock \emph{CoRR}, abs/2403.18746, 2024.
\newblock \doi{10.48550/ARXIV.2403.18746}.
\newblock URL \url{https://doi.org/10.48550/arXiv.2403.18746}.

\bibitem[Dolan{-}Gavitt et~al.(2016)Dolan{-}Gavitt, Hulin, Kirda, Leek, Mambretti, Robertson, Ulrich, and Whelan]{mutant2016-1}
Brendan Dolan{-}Gavitt, Patrick Hulin, Engin Kirda, Tim Leek, Andrea Mambretti, William~K. Robertson, Frederick Ulrich, and Ryan Whelan.
\newblock {LAVA:} large-scale automated vulnerability addition.
\newblock In \emph{{IEEE} Symposium on Security and Privacy, {SP} 2016, San Jose, CA, USA, May 22-26, 2016}, pp.\  110--121. {IEEE} Computer Society, 2016.
\newblock \doi{10.1109/SP.2016.15}.
\newblock URL \url{https://doi.org/10.1109/SP.2016.15}.

\bibitem[Dong et~al.(2022)Dong, Lou, Zhu, Sun, Li, Zhang, and Hao]{commit2022-2}
Jinhao Dong, Yiling Lou, Qihao Zhu, Zeyu Sun, Zhilin Li, Wenjie Zhang, and Dan Hao.
\newblock {FIRA:} fine-grained graph-based code change representation for automated commit message generation.
\newblock In \emph{44th {IEEE/ACM} 44th International Conference on Software Engineering, {ICSE} 2022, Pittsburgh, PA, USA, May 25-27, 2022}, pp.\  970--981. {ACM}, 2022.
\newblock \doi{10.1145/3510003.3510069}.
\newblock URL \url{https://doi.org/10.1145/3510003.3510069}.

\bibitem[Dong \& Lapata(2018)Dong and Lapata]{sql2018-3}
Li~Dong and Mirella Lapata.
\newblock Coarse-to-fine decoding for neural semantic parsing.
\newblock In Iryna Gurevych and Yusuke Miyao (eds.), \emph{Proceedings of the 56th Annual Meeting of the Association for Computational Linguistics, {ACL} 2018, Melbourne, Australia, July 15-20, 2018, Volume 1: Long Papers}, pp.\  731--742. Association for Computational Linguistics, 2018.
\newblock \doi{10.18653/v1/P18-1068}.
\newblock URL \url{https://aclanthology.org/P18-1068/}.

\bibitem[Dong et~al.(2019)Dong, Yang, Wang, Wei, Liu, Wang, Gao, Zhou, and Hon]{2019UniLM}
Li~Dong, Nan Yang, Wenhui Wang, Furu Wei, Xiaodong Liu, Yu~Wang, Jianfeng Gao, Ming Zhou, and Hsiao{-}Wuen Hon.
\newblock Unified language model pre-training for natural language understanding and generation.
\newblock In Hanna~M. Wallach, Hugo Larochelle, Alina Beygelzimer, Florence d'Alch{\'{e}}{-}Buc, Emily~B. Fox, and Roman Garnett (eds.), \emph{Advances in Neural Information Processing Systems 32: Annual Conference on Neural Information Processing Systems 2019, NeurIPS 2019, December 8-14, 2019, Vancouver, BC, Canada}, pp.\  13042--13054, 2019.
\newblock URL \url{https://proceedings.neurips.cc/paper/2019/hash/c20bb2d9a50d5ac1f713f8b34d9aac5a-Abstract.html}.

\bibitem[Dong et~al.(2023)Dong, Jiang, Jin, and Li]{2023self-collaboration}
Yihong Dong, Xue Jiang, Zhi Jin, and Ge~Li.
\newblock Self-collaboration code generation via chatgpt.
\newblock \emph{CoRR}, abs/2304.07590, 2023.
\newblock \doi{10.48550/ARXIV.2304.07590}.
\newblock URL \url{https://doi.org/10.48550/arXiv.2304.07590}.

\bibitem[Dou et~al.(2023)Dou, Shan, Jia, Deng, Xi, He, Wu, Gui, Liu, and Huang]{clone2023-1}
Shihan Dou, Junjie Shan, Haoxiang Jia, Wenhao Deng, Zhiheng Xi, Wei He, Yueming Wu, Tao Gui, Yang Liu, and Xuanjing Huang.
\newblock Towards understanding the capability of large language models on code clone detection: {A} survey.
\newblock \emph{CoRR}, abs/2308.01191, 2023.
\newblock \doi{10.48550/arXiv.2308.01191}.
\newblock URL \url{https://doi.org/10.48550/arXiv.2308.01191}.

\bibitem[Dou et~al.(2024)Dou, Liu, Jia, Xiong, Zhou, Shen, Shan, Huang, Wang, Fan, Xi, Zhou, Ji, Zheng, Zhang, Huang, and Gui]{2024StepCoder}
Shihan Dou, Yan Liu, Haoxiang Jia, Limao Xiong, Enyu Zhou, Wei Shen, Junjie Shan, Caishuang Huang, Xiao Wang, Xiaoran Fan, Zhiheng Xi, Yuhao Zhou, Tao Ji, Rui Zheng, Qi~Zhang, Xuanjing Huang, and Tao Gui.
\newblock Stepcoder: Improve code generation with reinforcement learning from compiler feedback.
\newblock \emph{CoRR}, abs/2402.01391, 2024.
\newblock \doi{10.48550/ARXIV.2402.01391}.
\newblock URL \url{https://doi.org/10.48550/arXiv.2402.01391}.

\bibitem[Drain et~al.(2021)Drain, Clement, Serrato, and Sundaresan]{fix2021-2}
Dawn Drain, Colin~B. Clement, Guillermo Serrato, and Neel Sundaresan.
\newblock Deepdebug: Fixing python bugs using stack traces, backtranslation, and code skeletons.
\newblock \emph{CoRR}, abs/2105.09352, 2021.
\newblock URL \url{https://arxiv.org/abs/2105.09352}.

\bibitem[Drissi et~al.(2018)Drissi, Watkins, Khant, Ojha, Segura, Segev, Weiner, and Keller]{trans2018-2}
Mehdi Drissi, Olivia Watkins, Aditya Khant, Vivaswat Ojha, Pedro~Sandoval Segura, Rakia Segev, Eric Weiner, and Robert Keller.
\newblock Program language translation using a grammar-driven tree-to-tree model.
\newblock \emph{CoRR}, abs/1807.01784, 2018.
\newblock URL \url{http://arxiv.org/abs/1807.01784}.

\bibitem[Drori \& Verma(2021)Drori and Verma]{math2021-0.5}
Iddo Drori and Nakul Verma.
\newblock Solving linear algebra by program synthesis.
\newblock \emph{CoRR}, abs/2111.08171, 2021.
\newblock URL \url{https://arxiv.org/abs/2111.08171}.

\bibitem[Drori et~al.(2022)Drori, Zhang, Shuttleworth, Tang, Lu, Ke, Liu, Chen, Tran, Cheng, Wang, Singh, Patti, Lynch, Shporer, Verma, Wu, and Strang]{math2021-1}
Iddo Drori, Sarah Zhang, Reece Shuttleworth, Leonard Tang, Albert Lu, Elizabeth Ke, Kevin Liu, Linda Chen, Sunny Tran, Newman Cheng, Roman Wang, Nikhil Singh, Taylor~L. Patti, Jayson Lynch, Avi Shporer, Nakul Verma, Eugene Wu, and Gilbert Strang.
\newblock A neural network solves, explains, and generates university math problems by program synthesis and few-shot learning at human level.
\newblock \emph{Proceedings of the National Academy of Sciences (PNAS)}, 119\penalty0 (32), 2022.

\bibitem[Du et~al.(2021)Du, Shi, Wang, Shi, Han, and Zhang]{retrieval2021-4}
Lun Du, Xiaozhou Shi, Yanlin Wang, Ensheng Shi, Shi Han, and Dongmei Zhang.
\newblock Is a single model enough? mucos: {A} multi-model ensemble learning approach for semantic code search.
\newblock In Gianluca Demartini, Guido Zuccon, J.~Shane Culpepper, Zi~Huang, and Hanghang Tong (eds.), \emph{{CIKM} '21: The 30th {ACM} International Conference on Information and Knowledge Management, Virtual Event, Queensland, Australia, November 1 - 5, 2021}, pp.\  2994--2998. {ACM}, 2021.
\newblock \doi{10.1145/3459637.3482127}.
\newblock URL \url{https://doi.org/10.1145/3459637.3482127}.

\bibitem[Du \& Li(2016)Du and Li]{log2016-1}
Min Du and Feifei Li.
\newblock Spell: Streaming parsing of system event logs.
\newblock In Francesco Bonchi, Josep Domingo{-}Ferrer, Ricardo Baeza{-}Yates, Zhi{-}Hua Zhou, and Xindong Wu (eds.), \emph{{IEEE} 16th International Conference on Data Mining, {ICDM} 2016, December 12-15, 2016, Barcelona, Spain}, pp.\  859--864. {IEEE} Computer Society, 2016.
\newblock \doi{10.1109/ICDM.2016.0103}.
\newblock URL \url{https://doi.org/10.1109/ICDM.2016.0103}.

\bibitem[Du et~al.(2017)Du, Li, Zheng, and Srikumar]{log2017-2}
Min Du, Feifei Li, Guineng Zheng, and Vivek Srikumar.
\newblock Deeplog: Anomaly detection and diagnosis from system logs through deep learning.
\newblock In Bhavani Thuraisingham, David Evans, Tal Malkin, and Dongyan Xu (eds.), \emph{Proceedings of the 2017 {ACM} {SIGSAC} Conference on Computer and Communications Security, {CCS} 2017, Dallas, TX, USA, October 30 - November 03, 2017}, pp.\  1285--1298. {ACM}, 2017.
\newblock \doi{10.1145/3133956.3134015}.
\newblock URL \url{https://doi.org/10.1145/3133956.3134015}.

\bibitem[Du et~al.(2022)Du, Huang, Dai, Tong, Lepikhin, Xu, Krikun, Zhou, Yu, Firat, Zoph, Fedus, Bosma, Zhou, Wang, Wang, Webster, Pellat, Robinson, Meier{-}Hellstern, Duke, Dixon, Zhang, Le, Wu, Chen, and Cui]{2021GLaM}
Nan Du, Yanping Huang, Andrew~M. Dai, Simon Tong, Dmitry Lepikhin, Yuanzhong Xu, Maxim Krikun, Yanqi Zhou, Adams~Wei Yu, Orhan Firat, Barret Zoph, Liam Fedus, Maarten~P. Bosma, Zongwei Zhou, Tao Wang, Yu~Emma Wang, Kellie Webster, Marie Pellat, Kevin Robinson, Kathleen~S. Meier{-}Hellstern, Toju Duke, Lucas Dixon, Kun Zhang, Quoc~V. Le, Yonghui Wu, Zhifeng Chen, and Claire Cui.
\newblock Glam: Efficient scaling of language models with mixture-of-experts.
\newblock In Kamalika Chaudhuri, Stefanie Jegelka, Le~Song, Csaba Szepesv{\'{a}}ri, Gang Niu, and Sivan Sabato (eds.), \emph{International Conference on Machine Learning, {ICML} 2022, 17-23 July 2022, Baltimore, Maryland, {USA}}, volume 162 of \emph{Proceedings of Machine Learning Research}, pp.\  5547--5569. {PMLR}, 2022.
\newblock URL \url{https://proceedings.mlr.press/v162/du22c.html}.

\bibitem[Dvivedi et~al.(2023)Dvivedi, Vijay, Pujari, Lodh, and Kumar]{doc2023-2}
Shubhang~Shekhar Dvivedi, Vyshnav Vijay, Sai Leela~Rahul Pujari, Shoumik Lodh, and Dhruv Kumar.
\newblock A comparative analysis of large language models for code documentation generation.
\newblock \emph{CoRR}, abs/2312.10349, 2023.
\newblock \doi{10.48550/ARXIV.2312.10349}.
\newblock URL \url{https://doi.org/10.48550/arXiv.2312.10349}.

\bibitem[Elallaoui et~al.(2018)Elallaoui, Nafil, and Touahni]{re-model2018-1}
Meryem Elallaoui, Khalid Nafil, and Raja Touahni.
\newblock Automatic transformation of user stories into {UML} use case diagrams using {NLP} techniques.
\newblock In Elhadi~M. Shakshuki and Ansar{-}Ul{-}Haque Yasar (eds.), \emph{The 9th International Conference on Ambient Systems, Networks and Technologies {(ANT} 2018) / The 8th International Conference on Sustainable Energy Information Technology {(SEIT} 2018) / Affiliated Workshops, May 8-11, 2018, Porto, Portugal}, volume 130 of \emph{Procedia Computer Science}, pp.\  42--49. Elsevier, 2018.
\newblock \doi{10.1016/J.PROCS.2018.04.010}.
\newblock URL \url{https://doi.org/10.1016/j.procs.2018.04.010}.

\bibitem[Eliseeva et~al.(2023)Eliseeva, Sokolov, Bogomolov, Golubev, Dig, and Bryksin]{commit-data-2023-2}
Aleksandra Eliseeva, Yaroslav Sokolov, Egor Bogomolov, Yaroslav Golubev, Danny Dig, and Timofey Bryksin.
\newblock From commit message generation to history-aware commit message completion.
\newblock In \emph{38th {IEEE/ACM} International Conference on Automated Software Engineering, {ASE} 2023, Luxembourg, September 11-15, 2023}, pp.\  723--735. {IEEE}, 2023.
\newblock \doi{10.1109/ASE56229.2023.00078}.
\newblock URL \url{https://doi.org/10.1109/ASE56229.2023.00078}.

\bibitem[Fan et~al.(2020)Fan, Li, Wang, and Nguyen]{defect-data-2020-1}
Jiahao Fan, Yi~Li, Shaohua Wang, and Tien~N. Nguyen.
\newblock A {C/C++} code vulnerability dataset with code changes and {CVE} summaries.
\newblock In Sunghun Kim, Georgios Gousios, Sarah Nadi, and Joseph Hejderup (eds.), \emph{{MSR} '20: 17th International Conference on Mining Software Repositories, Seoul, Republic of Korea, 29-30 June, 2020}, pp.\  508--512. {ACM}, 2020.
\newblock \doi{10.1145/3379597.3387501}.
\newblock URL \url{https://doi.org/10.1145/3379597.3387501}.

\bibitem[Fan et~al.(2023)Fan, Gao, Mirchev, Roychoudhury, and Tan]{fix2022-1}
Zhiyu Fan, Xiang Gao, Martin Mirchev, Abhik Roychoudhury, and Shin~Hwei Tan.
\newblock Automated repair of programs from large language models.
\newblock In \emph{45th {IEEE/ACM} International Conference on Software Engineering, {ICSE} 2023, Melbourne, Australia, May 14-20, 2023}, pp.\  1469--1481. {IEEE}, 2023.
\newblock \doi{10.1109/ICSE48619.2023.00128}.
\newblock URL \url{https://doi.org/10.1109/ICSE48619.2023.00128}.

\bibitem[Fang et~al.(2020)Fang, Liu, Shi, Huang, and Shi]{clone2020-2}
Chunrong Fang, Zixi Liu, Yangyang Shi, Jeff Huang, and Qingkai Shi.
\newblock Functional code clone detection with syntax and semantics fusion learning.
\newblock In Sarfraz Khurshid and Corina~S. Pasareanu (eds.), \emph{{ISSTA} '20: 29th {ACM} {SIGSOFT} International Symposium on Software Testing and Analysis, Virtual Event, USA, July 18-22, 2020}, pp.\  516--527. {ACM}, 2020.
\newblock \doi{10.1145/3395363.3397362}.
\newblock URL \url{https://doi.org/10.1145/3395363.3397362}.

\bibitem[Fang et~al.(2021)Fang, Tan, Zhang, and Liu]{retrieval2021-1.5}
Sen Fang, Youshuai Tan, Tao Zhang, and Yepang Liu.
\newblock Self-attention networks for code search.
\newblock \emph{Inf. Softw. Technol.}, 134:\penalty0 106542, 2021.
\newblock \doi{10.1016/J.INFSOF.2021.106542}.
\newblock URL \url{https://doi.org/10.1016/j.infsof.2021.106542}.

\bibitem[Fedus et~al.(2022)Fedus, Zoph, and Shazeer]{2021SwitchTransformer}
William Fedus, Barret Zoph, and Noam Shazeer.
\newblock Switch transformers: Scaling to trillion parameter models with simple and efficient sparsity.
\newblock \emph{J. Mach. Learn. Res.}, 23:\penalty0 120:1--120:39, 2022.
\newblock URL \url{http://jmlr.org/papers/v23/21-0998.html}.

\bibitem[Feng et~al.(2023{\natexlab{a}})Feng, Zhang, Gu, Ye, He, and Wang]{2023CoT-theory}
Guhao Feng, Bohang Zhang, Yuntian Gu, Haotian Ye, Di~He, and Liwei Wang.
\newblock Towards revealing the mystery behind chain of thought: a theoretical perspective.
\newblock \emph{CoRR}, abs/2305.15408, 2023{\natexlab{a}}.
\newblock \doi{10.48550/ARXIV.2305.15408}.
\newblock URL \url{https://doi.org/10.48550/arXiv.2305.15408}.

\bibitem[Feng et~al.(2024)Feng, Liao, Xiao, Vaughan, Zhang, and McDonald]{UI-2024-1}
K.~J.~Kevin Feng, Q.~Vera Liao, Ziang Xiao, Jennifer~Wortman Vaughan, Amy~X. Zhang, and David~W. McDonald.
\newblock Canvil: Designerly adaptation for llm-powered user experiences.
\newblock \emph{CoRR}, abs/2401.09051, 2024.
\newblock \doi{10.48550/ARXIV.2401.09051}.
\newblock URL \url{https://doi.org/10.48550/arXiv.2401.09051}.

\bibitem[Feng et~al.(2023{\natexlab{b}})Feng, Zhu, Fu, Jampani, Akula, He, Basu, Wang, and Wang]{UI-CV3}
Weixi Feng, Wanrong Zhu, Tsu{-}Jui Fu, Varun Jampani, Arjun~R. Akula, Xuehai He, Sugato Basu, Xin~Eric Wang, and William~Yang Wang.
\newblock Layoutgpt: Compositional visual planning and generation with large language models.
\newblock In Alice Oh, Tristan Naumann, Amir Globerson, Kate Saenko, Moritz Hardt, and Sergey Levine (eds.), \emph{Advances in Neural Information Processing Systems 36: Annual Conference on Neural Information Processing Systems 2023, NeurIPS 2023, New Orleans, LA, USA, December 10 - 16, 2023}, 2023{\natexlab{b}}.
\newblock URL \url{http://papers.nips.cc/paper\_files/paper/2023/hash/3a7f9e485845dac27423375c934cb4db-Abstract-Conference.html}.

\bibitem[Feng et~al.(2018)Feng, Martins, Bastani, and Dillig]{syn2018-4}
Yu~Feng, Ruben Martins, Osbert Bastani, and Isil Dillig.
\newblock Program synthesis using conflict-driven learning.
\newblock In Jeffrey~S. Foster and Dan Grossman (eds.), \emph{Proceedings of the 39th {ACM} {SIGPLAN} Conference on Programming Language Design and Implementation, {PLDI} 2018, Philadelphia, PA, USA, June 18-22, 2018}, pp.\  420--435. {ACM}, 2018.
\newblock \doi{10.1145/3192366.3192382}.
\newblock URL \url{https://doi.org/10.1145/3192366.3192382}.

\bibitem[Feng et~al.(2020)Feng, Guo, Tang, Duan, Feng, Gong, Shou, Qin, Liu, Jiang, and Zhou]{2020CodeBERT}
Zhangyin Feng, Daya Guo, Duyu Tang, Nan Duan, Xiaocheng Feng, Ming Gong, Linjun Shou, Bing Qin, Ting Liu, Daxin Jiang, and Ming Zhou.
\newblock Codebert: {A} pre-trained model for programming and natural languages.
\newblock In Trevor Cohn, Yulan He, and Yang Liu (eds.), \emph{Findings of the Association for Computational Linguistics: {EMNLP} 2020, Online Event, 16-20 November 2020}, volume {EMNLP} 2020 of \emph{Findings of {ACL}}, pp.\  1536--1547. Association for Computational Linguistics, 2020.
\newblock \doi{10.18653/v1/2020.findings-emnlp.139}.
\newblock URL \url{https://doi.org/10.18653/v1/2020.findings-emnlp.139}.

\bibitem[Fernandes et~al.(2019)Fernandes, Allamanis, and Brockschmidt]{id2018-4}
Patrick Fernandes, Miltiadis Allamanis, and Marc Brockschmidt.
\newblock Structured neural summarization.
\newblock In \emph{7th International Conference on Learning Representations, {ICLR} 2019, New Orleans, LA, USA, May 6-9, 2019}. OpenReview.net, 2019.
\newblock URL \url{https://openreview.net/forum?id=H1ersoRqtm}.

\bibitem[Ferrari et~al.(2014)Ferrari, Dell'Orletta, Spagnolo, and Gnesi]{re-ana2014-2}
Alessio Ferrari, Felice Dell'Orletta, Giorgio~Oronzo Spagnolo, and Stefania Gnesi.
\newblock Measuring and improving the completeness of natural language requirements.
\newblock In Camille Salinesi and Inge van~de Weerd (eds.), \emph{Requirements Engineering: Foundation for Software Quality - 20th International Working Conference, {REFSQ} 2014, Essen, Germany, April 7-10, 2014. Proceedings}, volume 8396 of \emph{Lecture Notes in Computer Science}, pp.\  23--38. Springer, 2014.
\newblock \doi{10.1007/978-3-319-05843-6\_3}.
\newblock URL \url{https://doi.org/10.1007/978-3-319-05843-6\_3}.

\bibitem[Ferrari et~al.(2024)Ferrari, Abualhaija, and Arora]{re-model2024-1}
Alessio Ferrari, Sallam Abualhaija, and Chetan Arora.
\newblock Model generation from requirements with llms: an exploratory study.
\newblock \emph{CoRR}, abs/2404.06371, 2024.
\newblock \doi{10.48550/ARXIV.2404.06371}.
\newblock URL \url{https://doi.org/10.48550/arXiv.2404.06371}.

\bibitem[Finegan{-}Dollak et~al.(2018)Finegan{-}Dollak, Kummerfeld, Zhang, Ramanathan, Sadasivam, Zhang, and Radev]{sql2018-4}
Catherine Finegan{-}Dollak, Jonathan~K. Kummerfeld, Li~Zhang, Karthik Ramanathan, Sesh Sadasivam, Rui Zhang, and Dragomir~R. Radev.
\newblock Improving text-to-sql evaluation methodology.
\newblock In Iryna Gurevych and Yusuke Miyao (eds.), \emph{Proceedings of the 56th Annual Meeting of the Association for Computational Linguistics, {ACL} 2018, Melbourne, Australia, July 15-20, 2018, Volume 1: Long Papers}, pp.\  351--360. Association for Computational Linguistics, 2018.
\newblock \doi{10.18653/v1/P18-1033}.
\newblock URL \url{https://aclanthology.org/P18-1033/}.

\bibitem[Frantzeskou et~al.(2011)Frantzeskou, MacDonell, Stamatatos, Georgiou, and Gritzalis]{author2011-1}
Georgia Frantzeskou, Stephen~G. MacDonell, Efstathios Stamatatos, Stelios Georgiou, and Stefanos Gritzalis.
\newblock The significance of user-defined identifiers in java source code authorship identification.
\newblock \emph{Comput. Syst. Sci. Eng.}, 26\penalty0 (2), 2011.

\bibitem[Fraser \& Arcuri(2011)Fraser and Arcuri]{unit2011-1}
Gordon Fraser and Andrea Arcuri.
\newblock Evosuite: automatic test suite generation for object-oriented software.
\newblock In Tibor Gyim{\'{o}}thy and Andreas Zeller (eds.), \emph{SIGSOFT/FSE'11 19th {ACM} {SIGSOFT} Symposium on the Foundations of Software Engineering {(FSE-19)} and ESEC'11: 13th European Software Engineering Conference (ESEC-13), Szeged, Hungary, September 5-9, 2011}, pp.\  416--419. {ACM}, 2011.
\newblock \doi{10.1145/2025113.2025179}.
\newblock URL \url{https://doi.org/10.1145/2025113.2025179}.

\bibitem[Fraser \& Arcuri(2014)Fraser and Arcuri]{unit-data-2014-1}
Gordon Fraser and Andrea Arcuri.
\newblock A large-scale evaluation of automated unit test generation using evosuite.
\newblock \emph{{ACM} Trans. Softw. Eng. Methodol.}, 24\penalty0 (2):\penalty0 8:1--8:42, 2014.
\newblock \doi{10.1145/2685612}.
\newblock URL \url{https://doi.org/10.1145/2685612}.

\bibitem[Fried et~al.(2023)Fried, Aghajanyan, Lin, Wang, Wallace, Shi, Zhong, Yih, Zettlemoyer, and Lewis]{2022InCoder}
Daniel Fried, Armen Aghajanyan, Jessy Lin, Sida Wang, Eric Wallace, Freda Shi, Ruiqi Zhong, Scott Yih, Luke Zettlemoyer, and Mike Lewis.
\newblock Incoder: {A} generative model for code infilling and synthesis.
\newblock In \emph{The Eleventh International Conference on Learning Representations, {ICLR} 2023, Kigali, Rwanda, May 1-5, 2023}. OpenReview.net, 2023.
\newblock URL \url{https://openreview.net/pdf?id=hQwb-lbM6EL}.

\bibitem[Fu et~al.(2023)Fu, Chai, Luo, Du, Zhang, Fan, Lei, Rui, Lin, Fang, Liu, Wang, Qi, Zhang, Zhang, and Yu]{2023CodeApex}
Lingyue Fu, Huacan Chai, Shuang Luo, Kounianhua Du, Weiming Zhang, Longteng Fan, Jiayi Lei, Renting Rui, Jianghao Lin, Yuchen Fang, Yifan Liu, Jingkuan Wang, Siyuan Qi, Kangning Zhang, Weinan Zhang, and Yong Yu.
\newblock Codeapex: {A} bilingual programming evaluation benchmark for large language models.
\newblock \emph{CoRR}, abs/2309.01940, 2023.
\newblock \doi{10.48550/arXiv.2309.01940}.
\newblock URL \url{https://doi.org/10.48550/arXiv.2309.01940}.

\bibitem[Fu \& Tantithamthavorn(2022)Fu and Tantithamthavorn]{defect2022-2}
Michael Fu and Chakkrit Tantithamthavorn.
\newblock Linevul: {A} transformer-based line-level vulnerability prediction.
\newblock In \emph{19th {IEEE/ACM} International Conference on Mining Software Repositories, {MSR} 2022, Pittsburgh, PA, USA, May 23-24, 2022}, pp.\  608--620. {ACM}, 2022.
\newblock \doi{10.1145/3524842.3528452}.
\newblock URL \url{https://doi.org/10.1145/3524842.3528452}.

\bibitem[Fu et~al.(2022)Fu, Tantithamthavorn, Le, Nguyen, and Phung]{fix2022-3.5}
Michael Fu, Chakkrit Tantithamthavorn, Trung Le, Van Nguyen, and Dinh~Q. Phung.
\newblock Vulrepair: a t5-based automated software vulnerability repair.
\newblock In Abhik Roychoudhury, Cristian Cadar, and Miryung Kim (eds.), \emph{Proceedings of the 30th {ACM} Joint European Software Engineering Conference and Symposium on the Foundations of Software Engineering, {ESEC/FSE} 2022, Singapore, Singapore, November 14-18, 2022}, pp.\  935--947. {ACM}, 2022.
\newblock \doi{10.1145/3540250.3549098}.
\newblock URL \url{https://doi.org/10.1145/3540250.3549098}.

\bibitem[Galley et~al.(2006)Galley, Graehl, Knight, Marcu, DeNeefe, Wang, and Thayer]{2006Galley}
Michel Galley, Jonathan Graehl, Kevin Knight, Daniel Marcu, Steve DeNeefe, Wei Wang, and Ignacio Thayer.
\newblock Scalable inference and training of context-rich syntactic translation models.
\newblock In Nicoletta Calzolari, Claire Cardie, and Pierre Isabelle (eds.), \emph{{ACL} 2006, 21st International Conference on Computational Linguistics and 44th Annual Meeting of the Association for Computational Linguistics, Proceedings of the Conference, Sydney, Australia, 17-21 July 2006}. The Association for Computer Linguistics, 2006.
\newblock \doi{10.3115/1220175.1220296}.
\newblock URL \url{https://aclanthology.org/P06-1121/}.

\bibitem[Gan et~al.(2021{\natexlab{a}})Gan, Chen, Huang, Purver, Woodward, Xie, and Huang]{sql-data-2021-4}
Yujian Gan, Xinyun Chen, Qiuping Huang, Matthew Purver, John~R. Woodward, Jinxia Xie, and Pengsheng Huang.
\newblock Towards robustness of text-to-sql models against synonym substitution.
\newblock In Chengqing Zong, Fei Xia, Wenjie Li, and Roberto Navigli (eds.), \emph{Proceedings of the 59th Annual Meeting of the Association for Computational Linguistics and the 11th International Joint Conference on Natural Language Processing, {ACL/IJCNLP} 2021, (Volume 1: Long Papers), Virtual Event, August 1-6, 2021}, pp.\  2505--2515. Association for Computational Linguistics, 2021{\natexlab{a}}.
\newblock \doi{10.18653/V1/2021.ACL-LONG.195}.
\newblock URL \url{https://doi.org/10.18653/v1/2021.acl-long.195}.

\bibitem[Gan et~al.(2021{\natexlab{b}})Gan, Chen, and Purver]{sql-data-2021-3}
Yujian Gan, Xinyun Chen, and Matthew Purver.
\newblock Exploring underexplored limitations of cross-domain text-to-sql generalization.
\newblock In Marie{-}Francine Moens, Xuanjing Huang, Lucia Specia, and Scott~Wen{-}tau Yih (eds.), \emph{Proceedings of the 2021 Conference on Empirical Methods in Natural Language Processing, {EMNLP} 2021, Virtual Event / Punta Cana, Dominican Republic, 7-11 November, 2021}, pp.\  8926--8931. Association for Computational Linguistics, 2021{\natexlab{b}}.
\newblock \doi{10.18653/V1/2021.EMNLP-MAIN.702}.
\newblock URL \url{https://doi.org/10.18653/v1/2021.emnlp-main.702}.

\bibitem[Gan et~al.(2022)Gan, Chen, Huang, and Purver]{sql-data-2022-1}
Yujian Gan, Xinyun Chen, Qiuping Huang, and Matthew Purver.
\newblock Measuring and improving compositional generalization in text-to-sql via component alignment.
\newblock In Marine Carpuat, Marie{-}Catherine de~Marneffe, and Iv{\'{a}}n Vladimir~Meza Ru{\'{\i}}z (eds.), \emph{Findings of the Association for Computational Linguistics: {NAACL} 2022, Seattle, WA, United States, July 10-15, 2022}, pp.\  831--843. Association for Computational Linguistics, 2022.
\newblock \doi{10.18653/V1/2022.FINDINGS-NAACL.62}.
\newblock URL \url{https://doi.org/10.18653/v1/2022.findings-naacl.62}.

\bibitem[Gao et~al.(2022)Gao, Li, Zhang, Lam, Li, Huang, Si, and Li]{sql2022-4}
Chang Gao, Bowen Li, Wenxuan Zhang, Wai Lam, Binhua Li, Fei Huang, Luo Si, and Yongbin Li.
\newblock Towards generalizable and robust text-to-sql parsing.
\newblock In Yoav Goldberg, Zornitsa Kozareva, and Yue Zhang (eds.), \emph{Findings of the Association for Computational Linguistics: {EMNLP} 2022, Abu Dhabi, United Arab Emirates, December 7-11, 2022}, pp.\  2113--2125. Association for Computational Linguistics, 2022.
\newblock \doi{10.18653/V1/2022.FINDINGS-EMNLP.155}.
\newblock URL \url{https://doi.org/10.18653/v1/2022.findings-emnlp.155}.

\bibitem[Gao et~al.(2023{\natexlab{a}})Gao, Wang, Li, Sun, Qian, Ding, and Zhou]{sql2023-2}
Dawei Gao, Haibin Wang, Yaliang Li, Xiuyu Sun, Yichen Qian, Bolin Ding, and Jingren Zhou.
\newblock Text-to-sql empowered by large language models: {A} benchmark evaluation.
\newblock \emph{CoRR}, abs/2308.15363, 2023{\natexlab{a}}.
\newblock \doi{10.48550/ARXIV.2308.15363}.
\newblock URL \url{https://doi.org/10.48550/arXiv.2308.15363}.

\bibitem[Gao et~al.(2021)Gao, Biderman, Black, Golding, Hoppe, Foster, Phang, He, Thite, Nabeshima, Presser, and Leahy]{2021Pile}
Leo Gao, Stella Biderman, Sid Black, Laurence Golding, Travis Hoppe, Charles Foster, Jason Phang, Horace He, Anish Thite, Noa Nabeshima, Shawn Presser, and Connor Leahy.
\newblock The pile: An 800gb dataset of diverse text for language modeling.
\newblock \emph{CoRR}, abs/2101.00027, 2021.
\newblock URL \url{https://arxiv.org/abs/2101.00027}.

\bibitem[Gao et~al.(2023{\natexlab{b}})Gao, Madaan, Zhou, Alon, Liu, Yang, Callan, and Neubig]{2022PAL}
Luyu Gao, Aman Madaan, Shuyan Zhou, Uri Alon, Pengfei Liu, Yiming Yang, Jamie Callan, and Graham Neubig.
\newblock {PAL:} program-aided language models.
\newblock In Andreas Krause, Emma Brunskill, Kyunghyun Cho, Barbara Engelhardt, Sivan Sabato, and Jonathan Scarlett (eds.), \emph{International Conference on Machine Learning, {ICML} 2023, 23-29 July 2023, Honolulu, Hawaii, {USA}}, volume 202 of \emph{Proceedings of Machine Learning Research}, pp.\  10764--10799. {PMLR}, 2023{\natexlab{b}}.
\newblock URL \url{https://proceedings.mlr.press/v202/gao23f.html}.

\bibitem[Gao et~al.(2023{\natexlab{c}})Gao, Gao, He, Zeng, Nie, Xia, and Lyu]{sum2021-0.7}
Shuzheng Gao, Cuiyun Gao, Yulan He, Jichuan Zeng, Lunyiu Nie, Xin Xia, and Michael~R. Lyu.
\newblock Code structure-guided transformer for source code summarization.
\newblock \emph{{ACM} Trans. Softw. Eng. Methodol.}, 32\penalty0 (1):\penalty0 23:1--23:32, 2023{\natexlab{c}}.
\newblock \doi{10.1145/3522674}.
\newblock URL \url{https://doi.org/10.1145/3522674}.

\bibitem[Gao \& Lyu(2022)Gao and Lyu]{sum2022-1}
Yuexiu Gao and Chen Lyu.
\newblock {M2TS:} multi-scale multi-modal approach based on transformer for source code summarization.
\newblock In Ayushi Rastogi, Rosalia Tufano, Gabriele Bavota, Venera Arnaoudova, and Sonia Haiduc (eds.), \emph{Proceedings of the 30th {IEEE/ACM} International Conference on Program Comprehension, {ICPC} 2022, Virtual Event, May 16-17, 2022}, pp.\  24--35. {ACM}, 2022.
\newblock \doi{10.1145/3524610.3527907}.
\newblock URL \url{https://doi.org/10.1145/3524610.3527907}.

\bibitem[Gao et~al.(2023{\natexlab{d}})Gao, Wang, Zhou, Zhu, and Zhang]{defect2023-4}
Zeyu Gao, Hao Wang, Yuchen Zhou, Wenyu Zhu, and Chao Zhang.
\newblock How far have we gone in vulnerability detection using large language models.
\newblock \emph{CoRR}, abs/2311.12420, 2023{\natexlab{d}}.
\newblock \doi{10.48550/ARXIV.2311.12420}.
\newblock URL \url{https://doi.org/10.48550/arXiv.2311.12420}.

\bibitem[Gazzola et~al.(2018)Gazzola, Micucci, and Mariani]{fix-survey-2017}
Luca Gazzola, Daniela Micucci, and Leonardo Mariani.
\newblock Automatic software repair: a survey.
\newblock In Michel Chaudron, Ivica Crnkovic, Marsha Chechik, and Mark Harman (eds.), \emph{Proceedings of the 40th International Conference on Software Engineering, {ICSE} 2018, Gothenburg, Sweden, May 27 - June 03, 2018}, pp.\  1219. {ACM}, 2018.
\newblock \doi{10.1145/3180155.3182526}.
\newblock URL \url{https://doi.org/10.1145/3180155.3182526}.

\bibitem[Geng et~al.(2024)Geng, Wang, Dong, Wang, Li, Jin, Mao, and Liao]{comment2023-3}
Mingyang Geng, Shangwen Wang, Dezun Dong, Haotian Wang, Ge~Li, Zhi Jin, Xiaoguang Mao, and Xiangke Liao.
\newblock Large language models are few-shot summarizers: Multi-intent comment generation via in-context learning.
\newblock In \emph{Proceedings of the 46th {IEEE/ACM} International Conference on Software Engineering, {ICSE} 2024, Lisbon, Portugal, April 14-20, 2024}, pp.\  39:1--39:13. {ACM}, 2024.
\newblock \doi{10.1145/3597503.3608134}.
\newblock URL \url{https://doi.org/10.1145/3597503.3608134}.

\bibitem[Ghosh et~al.(2016)Ghosh, Elenius, Li, Lincoln, Shankar, and Steiner]{re-ana2014-1}
Shalini Ghosh, Daniel Elenius, Wenchao Li, Patrick Lincoln, Natarajan Shankar, and Wilfried Steiner.
\newblock {ARSENAL:} automatic requirements specification extraction from natural language.
\newblock In Sanjai Rayadurgam and Oksana Tkachuk (eds.), \emph{{NASA} Formal Methods - 8th International Symposium, {NFM} 2016, Minneapolis, MN, USA, June 7-9, 2016, Proceedings}, volume 9690 of \emph{Lecture Notes in Computer Science}, pp.\  41--46. Springer, 2016.
\newblock \doi{10.1007/978-3-319-40648-0\_4}.
\newblock URL \url{https://doi.org/10.1007/978-3-319-40648-0\_4}.

\bibitem[Gong et~al.(2024)Gong, Elhoushi, and Cheung]{2024AST-T5}
Linyuan Gong, Mostafa Elhoushi, and Alvin Cheung.
\newblock {AST-T5:} structure-aware pretraining for code generation and understanding.
\newblock \emph{CoRR}, abs/2401.03003, 2024.
\newblock \doi{10.48550/ARXIV.2401.03003}.
\newblock URL \url{https://doi.org/10.48550/arXiv.2401.03003}.

\bibitem[Gopinath \& Sethuraman(2023)Gopinath and Sethuraman]{malware-survey2022-1}
M.~Gopinath and Sibi~Chakkaravarthy Sethuraman.
\newblock A comprehensive survey on deep learning based malware detection techniques.
\newblock \emph{Comput. Sci. Rev.}, 47:\penalty0 100529, 2023.
\newblock \doi{10.1016/J.COSREV.2022.100529}.
\newblock URL \url{https://doi.org/10.1016/j.cosrev.2022.100529}.

\bibitem[Goues et~al.(2015)Goues, Holtschulte, Smith, Brun, Devanbu, Forrest, and Weimer]{fix-data-2015-1}
Claire~Le Goues, Neal~J. Holtschulte, Edward~K. Smith, Yuriy Brun, Premkumar~T. Devanbu, Stephanie Forrest, and Westley Weimer.
\newblock The manybugs and introclass benchmarks for automated repair of {C} programs.
\newblock \emph{{IEEE} Trans. Software Eng.}, 41\penalty0 (12):\penalty0 1236--1256, 2015.
\newblock \doi{10.1109/TSE.2015.2454513}.
\newblock URL \url{https://doi.org/10.1109/TSE.2015.2454513}.

\bibitem[Goyal et~al.(2022)Goyal, Gao, Chaudhary, Chen, Wenzek, Ju, Krishnan, Ranzato, Guzm{\'{a}}n, and Fan]{2021Flores101}
Naman Goyal, Cynthia Gao, Vishrav Chaudhary, Peng{-}Jen Chen, Guillaume Wenzek, Da~Ju, Sanjana Krishnan, Marc'Aurelio Ranzato, Francisco Guzm{\'{a}}n, and Angela Fan.
\newblock The flores-101 evaluation benchmark for low-resource and multilingual machine translation.
\newblock \emph{Trans. Assoc. Comput. Linguistics}, 10:\penalty0 522--538, 2022.
\newblock \doi{10.1162/TACL\_A\_00474}.
\newblock URL \url{https://doi.org/10.1162/tacl\_a\_00474}.

\bibitem[Grazia \& Pradel(2023)Grazia and Pradel]{retrieval-survey-2022-1}
Luca~Di Grazia and Michael Pradel.
\newblock Code search: {A} survey of techniques for finding code.
\newblock \emph{{ACM} Comput. Surv.}, 55\penalty0 (11):\penalty0 220:1--220:31, 2023.
\newblock \doi{10.1145/3565971}.
\newblock URL \url{https://doi.org/10.1145/3565971}.

\bibitem[Grubisic et~al.(2024{\natexlab{a}})Grubisic, Cummins, Seeker, and Leather]{comp-opt2024-2}
Dejan Grubisic, Chris Cummins, Volker Seeker, and Hugh Leather.
\newblock Compiler generated feedback for large language models.
\newblock \emph{CoRR}, abs/2403.14714, 2024{\natexlab{a}}.
\newblock \doi{10.48550/ARXIV.2403.14714}.
\newblock URL \url{https://doi.org/10.48550/arXiv.2403.14714}.

\bibitem[Grubisic et~al.(2024{\natexlab{b}})Grubisic, Seeker, Synnaeve, Leather, Mellor{-}Crummey, and Cummins]{comp-opt2024-1}
Dejan Grubisic, Volker Seeker, Gabriel Synnaeve, Hugh Leather, John~M. Mellor{-}Crummey, and Chris Cummins.
\newblock Priority sampling of large language models for compilers.
\newblock In \emph{Proceedings of the 4th Workshop on Machine Learning and Systems, EuroMLSys 2024, Athens, Greece, 22 April 2024}, pp.\  91--97. {ACM}, 2024{\natexlab{b}}.
\newblock \doi{10.1145/3642970.3655831}.
\newblock URL \url{https://doi.org/10.1145/3642970.3655831}.

\bibitem[Gu et~al.(2024)Gu, Rozière, Leather, Solar-Lezama, Synnaeve, and Wang]{2024CRUXEval}
Alex Gu, Baptiste Rozière, Hugh Leather, Armando Solar-Lezama, Gabriel Synnaeve, and Sida~I. Wang.
\newblock Cruxeval: A benchmark for code reasoning, understanding and execution.
\newblock \emph{CoRR}, abs/2401.03065, 2024.
\newblock \doi{10.48550/ARXIV.2401.03065}.
\newblock URL \url{https://doi.org/10.48550/arXiv.2401.03065}.

\bibitem[Gu et~al.(2021)Gu, Chen, and Monperrus]{retrieval2021-3}
Jian Gu, Zimin Chen, and Martin Monperrus.
\newblock Multimodal representation for neural code search.
\newblock In \emph{{IEEE} International Conference on Software Maintenance and Evolution, {ICSME} 2021, Luxembourg, September 27 - October 1, 2021}, pp.\  483--494. {IEEE}, 2021.
\newblock \doi{10.1109/ICSME52107.2021.00049}.
\newblock URL \url{https://doi.org/10.1109/ICSME52107.2021.00049}.

\bibitem[Gu et~al.(2022)Gu, Luo, Zhou, and Wang]{fuzz2022-3}
Jiazhen Gu, Xuchuan Luo, Yangfan Zhou, and Xin Wang.
\newblock Muffin: Testing deep learning libraries via neural architecture fuzzing.
\newblock In \emph{44th {IEEE/ACM} 44th International Conference on Software Engineering, {ICSE} 2022, Pittsburgh, PA, USA, May 25-27, 2022}, pp.\  1418--1430. {ACM}, 2022.
\newblock \doi{10.1145/3510003.3510092}.
\newblock URL \url{https://doi.org/10.1145/3510003.3510092}.

\bibitem[Gu et~al.(2017)Gu, Zhang, Zhang, and Kim]{api2017-1}
Xiaodong Gu, Hongyu Zhang, Dongmei Zhang, and Sunghun Kim.
\newblock Deepam: Migrate apis with multi-modal sequence to sequence learning.
\newblock In Carles Sierra (ed.), \emph{Proceedings of the Twenty-Sixth International Joint Conference on Artificial Intelligence, {IJCAI} 2017, Melbourne, Australia, August 19-25, 2017}, pp.\  3675--3681. ijcai.org, 2017.
\newblock \doi{10.24963/IJCAI.2017/514}.
\newblock URL \url{https://doi.org/10.24963/ijcai.2017/514}.

\bibitem[Gu et~al.(2018)Gu, Zhang, and Kim]{retrieval2018-1}
Xiaodong Gu, Hongyu Zhang, and Sunghun Kim.
\newblock Deep code search.
\newblock In Michel Chaudron, Ivica Crnkovic, Marsha Chechik, and Mark Harman (eds.), \emph{Proceedings of the 40th International Conference on Software Engineering, {ICSE} 2018, Gothenburg, Sweden, May 27 - June 03, 2018}, pp.\  933--944. {ACM}, 2018.
\newblock \doi{10.1145/3180155.3180167}.
\newblock URL \url{https://doi.org/10.1145/3180155.3180167}.

\bibitem[Guan et~al.(2024)Guan, Wan, Bi, Wang, Zhang, Sui, Zhou, and Sun]{2024CodeIP}
Batu Guan, Yao Wan, Zhangqian Bi, Zheng Wang, Hongyu Zhang, Yulei Sui, Pan Zhou, and Lichao Sun.
\newblock Codeip: A grammar-guided multi-bit watermark for large language models of code.
\newblock \emph{CoRR}, abs/2404.15639, 2024.
\newblock \doi{10.48550/ARXIV.2404.15639}.
\newblock URL \url{https://doi.org/10.48550/arXiv.2404.15639}.

\bibitem[Guilherme \& Vincenzi(2023)Guilherme and Vincenzi]{unit2023-10}
Vitor Guilherme and Auri Vincenzi.
\newblock An initial investigation of chatgpt unit test generation capability.
\newblock In Awdren~L. Font{\~{a}}o, D{\'{e}}bora M.~B. Paiva, Hudson Borges, Maria~Istela Cagnin, Patr{\'{\i}}cia~Gomes Fernandes, Vanessa Borges, Silvana~M. Melo, Vinicius H.~S. Durelli, and Edna~Dias Canedo (eds.), \emph{8th Brazilian Symposium on Systematic and Automated Software Testing, {SAST} 2023, Campo Grande, MS, Brazil, September 25-29, 2023}, pp.\  15--24. {ACM}, 2023.
\newblock \doi{10.1145/3624032.3624035}.
\newblock URL \url{https://doi.org/10.1145/3624032.3624035}.

\bibitem[Gunasekar et~al.(2023)Gunasekar, Zhang, Aneja, Mendes, Giorno, Gopi, Javaheripi, Kauffmann, de~Rosa, Saarikivi, Salim, Shah, Behl, Wang, Bubeck, Eldan, Kalai, Lee, and Li]{2023Phi-1}
Suriya Gunasekar, Yi~Zhang, Jyoti Aneja, Caio C{\'{e}}sar~Teodoro Mendes, Allie~Del Giorno, Sivakanth Gopi, Mojan Javaheripi, Piero Kauffmann, Gustavo de~Rosa, Olli Saarikivi, Adil Salim, Shital Shah, Harkirat~Singh Behl, Xin Wang, S{\'{e}}bastien Bubeck, Ronen Eldan, Adam~Tauman Kalai, Yin~Tat Lee, and Yuanzhi Li.
\newblock Textbooks are all you need.
\newblock \emph{CoRR}, abs/2306.11644, 2023.
\newblock \doi{10.48550/arXiv.2306.11644}.
\newblock URL \url{https://doi.org/10.48550/arXiv.2306.11644}.

\bibitem[G{\"{u}}nes \& Aydemir(2020)G{\"{u}}nes and Aydemir]{re-model2020-1}
Tug{\c{c}}e G{\"{u}}nes and Fatma~Basak Aydemir.
\newblock Automated goal model extraction from user stories using {NLP}.
\newblock In Travis~D. Breaux, Andrea Zisman, Samuel Fricker, and Martin Glinz (eds.), \emph{28th {IEEE} International Requirements Engineering Conference, {RE} 2020, Zurich, Switzerland, August 31 - September 4, 2020}, pp.\  382--387. {IEEE}, 2020.
\newblock \doi{10.1109/RE48521.2020.00052}.
\newblock URL \url{https://doi.org/10.1109/RE48521.2020.00052}.

\bibitem[Guo et~al.(2023{\natexlab{a}})Guo, Tian, Tang, Li, Wen, Wang, and Wang]{sql2023-1.7}
Chunxi Guo, Zhiliang Tian, Jintao Tang, Shasha Li, Zhihua Wen, Kaixuan Wang, and Ting Wang.
\newblock Retrieval-augmented gpt-3.5-based text-to-sql framework with sample-aware prompting and dynamic revision chain.
\newblock In Biao Luo, Long Cheng, Zheng{-}Guang Wu, Hongyi Li, and Chaojie Li (eds.), \emph{Neural Information Processing - 30th International Conference, {ICONIP} 2023, Changsha, China, November 20-23, 2023, Proceedings, Part {VI}}, volume 14452 of \emph{Lecture Notes in Computer Science}, pp.\  341--356. Springer, 2023{\natexlab{a}}.
\newblock \doi{10.1007/978-981-99-8076-5\_25}.
\newblock URL \url{https://doi.org/10.1007/978-981-99-8076-5\_25}.

\bibitem[Guo et~al.(2021{\natexlab{a}})Guo, Ren, Lu, Feng, Tang, Liu, Zhou, Duan, Svyatkovskiy, Fu, Tufano, Deng, Clement, Drain, Sundaresan, Yin, Jiang, and Zhou]{2020GraphCodeBERT}
Daya Guo, Shuo Ren, Shuai Lu, Zhangyin Feng, Duyu Tang, Shujie Liu, Long Zhou, Nan Duan, Alexey Svyatkovskiy, Shengyu Fu, Michele Tufano, Shao~Kun Deng, Colin~B. Clement, Dawn Drain, Neel Sundaresan, Jian Yin, Daxin Jiang, and Ming Zhou.
\newblock Graphcodebert: Pre-training code representations with data flow.
\newblock In \emph{9th International Conference on Learning Representations, {ICLR} 2021, Virtual Event, Austria, May 3-7, 2021}. OpenReview.net, 2021{\natexlab{a}}.
\newblock URL \url{https://openreview.net/forum?id=jLoC4ez43PZ}.

\bibitem[Guo et~al.(2022)Guo, Lu, Duan, Wang, Zhou, and Yin]{2022UniXcoder}
Daya Guo, Shuai Lu, Nan Duan, Yanlin Wang, Ming Zhou, and Jian Yin.
\newblock Unixcoder: Unified cross-modal pre-training for code representation.
\newblock In Smaranda Muresan, Preslav Nakov, and Aline Villavicencio (eds.), \emph{Proceedings of the 60th Annual Meeting of the Association for Computational Linguistics (Volume 1: Long Papers), {ACL} 2022, Dublin, Ireland, May 22-27, 2022}, pp.\  7212--7225. Association for Computational Linguistics, 2022.
\newblock \doi{10.18653/v1/2022.acl-long.499}.
\newblock URL \url{https://doi.org/10.18653/v1/2022.acl-long.499}.

\bibitem[Guo et~al.(2023{\natexlab{b}})Guo, Xu, Duan, Yin, and McAuley]{2023LongCoder}
Daya Guo, Canwen Xu, Nan Duan, Jian Yin, and Julian~J. McAuley.
\newblock Longcoder: {A} long-range pre-trained language model for code completion.
\newblock In Andreas Krause, Emma Brunskill, Kyunghyun Cho, Barbara Engelhardt, Sivan Sabato, and Jonathan Scarlett (eds.), \emph{International Conference on Machine Learning, {ICML} 2023, 23-29 July 2023, Honolulu, Hawaii, {USA}}, volume 202 of \emph{Proceedings of Machine Learning Research}, pp.\  12098--12107. {PMLR}, 2023{\natexlab{b}}.
\newblock URL \url{https://proceedings.mlr.press/v202/guo23j.html}.

\bibitem[Guo et~al.(2024)Guo, Zhu, Yang, Xie, Dong, Zhang, Chen, Bi, Wu, Li, Luo, Xiong, and Liang]{2023DeepSeekCoder}
Daya Guo, Qihao Zhu, Dejian Yang, Zhenda Xie, Kai Dong, Wentao Zhang, Guanting Chen, Xiao Bi, Y.~Wu, Y.~K. Li, Fuli Luo, Yingfei Xiong, and Wenfeng Liang.
\newblock Deepseek-coder: When the large language model meets programming - the rise of code intelligence.
\newblock \emph{CoRR}, abs/2401.14196, 2024.
\newblock \doi{10.48550/ARXIV.2401.14196}.
\newblock URL \url{https://doi.org/10.48550/arXiv.2401.14196}.

\bibitem[Guo et~al.(2021{\natexlab{b}})Guo, Yuan, and Wu]{log2021-0.1}
Haixuan Guo, Shuhan Yuan, and Xintao Wu.
\newblock Logbert: Log anomaly detection via {BERT}.
\newblock In \emph{International Joint Conference on Neural Networks, {IJCNN} 2021, Shenzhen, China, July 18-22, 2021}, pp.\  1--8. {IEEE}, 2021{\natexlab{b}}.
\newblock \doi{10.1109/IJCNN52387.2021.9534113}.
\newblock URL \url{https://doi.org/10.1109/IJCNN52387.2021.9534113}.

\bibitem[Guo et~al.(2019)Guo, Zhan, Gao, Xiao, Lou, Liu, and Zhang]{sql2019-3}
Jiaqi Guo, Zecheng Zhan, Yan Gao, Yan Xiao, Jian{-}Guang Lou, Ting Liu, and Dongmei Zhang.
\newblock Towards complex text-to-sql in cross-domain database with intermediate representation.
\newblock In Anna Korhonen, David~R. Traum, and Llu{\'{\i}}s M{\`{a}}rquez (eds.), \emph{Proceedings of the 57th Conference of the Association for Computational Linguistics, {ACL} 2019, Florence, Italy, July 28- August 2, 2019, Volume 1: Long Papers}, pp.\  4524--4535. Association for Computational Linguistics, 2019.
\newblock \doi{10.18653/v1/p19-1444}.
\newblock URL \url{https://doi.org/10.18653/v1/p19-1444}.

\bibitem[Guo et~al.(2020)Guo, Xie, Li, Zhang, Liu, Li, and Shen]{fuzz2020-2}
Qianyu Guo, Xiaofei Xie, Yi~Li, Xiaoyu Zhang, Yang Liu, Xiaohong Li, and Chao Shen.
\newblock Audee: Automated testing for deep learning frameworks.
\newblock In \emph{35th {IEEE/ACM} International Conference on Automated Software Engineering, {ASE} 2020, Melbourne, Australia, September 21-25, 2020}, pp.\  486--498. {IEEE}, 2020.
\newblock \doi{10.1145/3324884.3416571}.
\newblock URL \url{https://doi.org/10.1145/3324884.3416571}.

\bibitem[Gupta \& Sundaresan(2018)Gupta and Sundaresan]{review2018-1}
Anshul Gupta and Neel Sundaresan.
\newblock Intelligent code reviews using deep learning, 2018.

\bibitem[Gupta et~al.(2017)Gupta, Pal, Kanade, and Shevade]{fix2017-1}
Rahul Gupta, Soham Pal, Aditya Kanade, and Shirish~K. Shevade.
\newblock Deepfix: Fixing common {C} language errors by deep learning.
\newblock In Satinder Singh and Shaul Markovitch (eds.), \emph{Proceedings of the Thirty-First {AAAI} Conference on Artificial Intelligence, February 4-9, 2017, San Francisco, California, {USA}}, pp.\  1345--1351. {AAAI} Press, 2017.
\newblock \doi{10.1609/aaai.v31i1.10742}.
\newblock URL \url{https://doi.org/10.1609/aaai.v31i1.10742}.

\bibitem[Gur et~al.(2023)Gur, Nachum, Miao, Safdari, Huang, Chowdhery, Narang, Fiedel, and Faust]{UI2022-5}
Izzeddin Gur, Ofir Nachum, Yingjie Miao, Mustafa Safdari, Austin Huang, Aakanksha Chowdhery, Sharan Narang, Noah Fiedel, and Aleksandra Faust.
\newblock Understanding {HTML} with large language models.
\newblock In Houda Bouamor, Juan Pino, and Kalika Bali (eds.), \emph{Findings of the Association for Computational Linguistics: {EMNLP} 2023, Singapore, December 6-10, 2023}, pp.\  2803--2821. Association for Computational Linguistics, 2023.
\newblock \doi{10.18653/V1/2023.FINDINGS-EMNLP.185}.
\newblock URL \url{https://doi.org/10.18653/v1/2023.findings-emnlp.185}.

\bibitem[Guzman et~al.(2017)Guzman, Ibrahim, and Glinz]{re-ana2017-2}
Emitza Guzman, Mohamed Ibrahim, and Martin Glinz.
\newblock A little bird told me: Mining tweets for requirements and software evolution.
\newblock In Ana Moreira, Jo{\~{a}}o Ara{\'{u}}jo, Jane Hayes, and Barbara Paech (eds.), \emph{25th {IEEE} International Requirements Engineering Conference, {RE} 2017, Lisbon, Portugal, September 4-8, 2017}, pp.\  11--20. {IEEE} Computer Society, 2017.
\newblock \doi{10.1109/RE.2017.88}.
\newblock URL \url{https://doi.org/10.1109/RE.2017.88}.

\bibitem[Gyimesi et~al.(2019)Gyimesi, Vancsics, Stocco, Mazinanian, Besz{\'{e}}des, Ferenc, and Mesbah]{fix-data-2019-5}
P{\'{e}}ter Gyimesi, B{\'{e}}la Vancsics, Andrea Stocco, Davood Mazinanian, {\'{A}}rp{\'{a}}d Besz{\'{e}}des, Rudolf Ferenc, and Ali Mesbah.
\newblock Bugsjs: a benchmark of javascript bugs.
\newblock In \emph{12th {IEEE} Conference on Software Testing, Validation and Verification, {ICST} 2019, Xi'an, China, April 22-27, 2019}, pp.\  90--101. {IEEE}, 2019.
\newblock \doi{10.1109/ICST.2019.00019}.
\newblock URL \url{https://doi.org/10.1109/ICST.2019.00019}.

\bibitem[Haldar \& Hockenmaier(2024)Haldar and Hockenmaier]{sum2024-1}
Rajarshi Haldar and Julia Hockenmaier.
\newblock Analyzing the performance of large language models on code summarization.
\newblock \emph{CoRR}, abs/2404.08018, 2024.
\newblock \doi{10.48550/ARXIV.2404.08018}.
\newblock URL \url{https://doi.org/10.48550/arXiv.2404.08018}.

\bibitem[Haller et~al.(2024)Haller, Golde, and Akbik]{2024PECC}
Patrick Haller, Jonas Golde, and Alan Akbik.
\newblock {PECC:} problem extraction and coding challenges.
\newblock In Nicoletta Calzolari, Min{-}Yen Kan, V{\'{e}}ronique Hoste, Alessandro Lenci, Sakriani Sakti, and Nianwen Xue (eds.), \emph{Proceedings of the 2024 Joint International Conference on Computational Linguistics, Language Resources and Evaluation, {LREC/COLING} 2024, 20-25 May, 2024, Torino, Italy}, pp.\  12690--12699. {ELRA} and {ICCL}, 2024.
\newblock URL \url{https://aclanthology.org/2024.lrec-main.1111}.

\bibitem[Hamer et~al.(2024)Hamer, d'Amorim, and Williams]{analysis2024-2}
Sivana Hamer, Marcelo d'Amorim, and Laurie Williams.
\newblock Just another copy and paste? comparing the security vulnerabilities of chatgpt generated code and stackoverflow answers.
\newblock \emph{CoRR}, abs/2403.15600, 2024.
\newblock \doi{10.48550/ARXIV.2403.15600}.
\newblock URL \url{https://doi.org/10.48550/arXiv.2403.15600}.

\bibitem[Hanam et~al.(2016)Hanam, Brito, and Mesbah]{fix-data-2016-1}
Quinn Hanam, Fernando Santos De~Mattos Brito, and Ali Mesbah.
\newblock Discovering bug patterns in javascript.
\newblock In Thomas Zimmermann, Jane Cleland{-}Huang, and Zhendong Su (eds.), \emph{Proceedings of the 24th {ACM} {SIGSOFT} International Symposium on Foundations of Software Engineering, {FSE} 2016, Seattle, WA, USA, November 13-18, 2016}, pp.\  144--156. {ACM}, 2016.
\newblock \doi{10.1145/2950290.2950308}.
\newblock URL \url{https://doi.org/10.1145/2950290.2950308}.

\bibitem[Hanif \& Maffeis(2022)Hanif and Maffeis]{defect2022-1}
Hazim Hanif and Sergio Maffeis.
\newblock Vulberta: Simplified source code pre-training for vulnerability detection.
\newblock In \emph{International Joint Conference on Neural Networks, {IJCNN} 2022, Padua, Italy, July 18-23, 2022}, pp.\  1--8. {IEEE}, 2022.
\newblock \doi{10.1109/IJCNN55064.2022.9892280}.
\newblock URL \url{https://doi.org/10.1109/IJCNN55064.2022.9892280}.

\bibitem[Hao et~al.(2022)Hao, Li, Liu, Miao, Zong, Jiang, Liu, and Wei]{2022AixBench}
Yiyang Hao, Ge~Li, Yongqiang Liu, Xiaowei Miao, He~Zong, Siyuan Jiang, Yang Liu, and He~Wei.
\newblock Aixbench: {A} code generation benchmark dataset.
\newblock \emph{CoRR}, abs/2206.13179, 2022.
\newblock \doi{10.48550/arXiv.2206.13179}.
\newblock URL \url{https://doi.org/10.48550/arXiv.2206.13179}.

\bibitem[Haque et~al.(2020)Haque, LeClair, Wu, and McMillan]{sum2020-1}
Sakib Haque, Alexander LeClair, Lingfei Wu, and Collin McMillan.
\newblock Improved automatic summarization of subroutines via attention to file context.
\newblock In Sunghun Kim, Georgios Gousios, Sarah Nadi, and Joseph Hejderup (eds.), \emph{{MSR} '20: 17th International Conference on Mining Software Repositories, Seoul, Republic of Korea, 29-30 June, 2020}, pp.\  300--310. {ACM}, 2020.
\newblock \doi{10.1145/3379597.3387449}.
\newblock URL \url{https://doi.org/10.1145/3379597.3387449}.

\bibitem[Harzevili et~al.(2023)Harzevili, Belle, Wang, Wang, Jiang, and Nagappan]{defect-survey-2023-2}
Nima~Shiri Harzevili, Alvine~Boaye Belle, Junjie Wang, Song Wang, Zhen~Ming Jiang, and Nachiappan Nagappan.
\newblock A survey on automated software vulnerability detection using machine learning and deep learning.
\newblock \emph{CoRR}, abs/2306.11673, 2023.
\newblock \doi{10.48550/ARXIV.2306.11673}.
\newblock URL \url{https://doi.org/10.48550/arXiv.2306.11673}.

\bibitem[Hassan et~al.(2018)Hassan, Urban, Eilers, and M{\"{u}}ller]{type2018-1}
Mostafa Hassan, Caterina Urban, Marco Eilers, and Peter M{\"{u}}ller.
\newblock Maxsmt-based type inference for python 3.
\newblock In Hana Chockler and Georg Weissenbacher (eds.), \emph{Computer Aided Verification - 30th International Conference, {CAV} 2018, Held as Part of the Federated Logic Conference, FloC 2018, Oxford, UK, July 14-17, 2018, Proceedings, Part {II}}, volume 10982 of \emph{Lecture Notes in Computer Science}, pp.\  12--19. Springer, 2018.
\newblock \doi{10.1007/978-3-319-96142-2\_2}.
\newblock URL \url{https://doi.org/10.1007/978-3-319-96142-2\_2}.

\bibitem[Hayati et~al.(2018)Hayati, Olivier, Avvaru, Yin, Tomasic, and Neubig]{syn2018-5}
Shirley~Anugrah Hayati, Rapha{\"{e}}l Olivier, Pravalika Avvaru, Pengcheng Yin, Anthony Tomasic, and Graham Neubig.
\newblock Retrieval-based neural code generation.
\newblock In Ellen Riloff, David Chiang, Julia Hockenmaier, and Jun'ichi Tsujii (eds.), \emph{Proceedings of the 2018 Conference on Empirical Methods in Natural Language Processing, Brussels, Belgium, October 31 - November 4, 2018}, pp.\  925--930. Association for Computational Linguistics, 2018.
\newblock \doi{10.18653/V1/D18-1111}.
\newblock URL \url{https://doi.org/10.18653/v1/d18-1111}.

\bibitem[Hazoom et~al.(2021)Hazoom, Malik, and Bogin]{sql-data-2021-2}
Moshe Hazoom, Vibhor Malik, and Ben Bogin.
\newblock Text-to-sql in the wild: {A} naturally-occurring dataset based on stack exchange data.
\newblock \emph{CoRR}, abs/2106.05006, 2021.
\newblock URL \url{https://arxiv.org/abs/2106.05006}.

\bibitem[He et~al.(2018)He, Ivanov, Tsankov, Raychev, and Vechev]{ob2018-1}
Jingxuan He, Pesho Ivanov, Petar Tsankov, Veselin Raychev, and Martin~T. Vechev.
\newblock Debin: Predicting debug information in stripped binaries.
\newblock In David Lie, Mohammad Mannan, Michael Backes, and XiaoFeng Wang (eds.), \emph{Proceedings of the 2018 {ACM} {SIGSAC} Conference on Computer and Communications Security, {CCS} 2018, Toronto, ON, Canada, October 15-19, 2018}, pp.\  1667--1680. {ACM}, 2018.
\newblock \doi{10.1145/3243734.3243866}.
\newblock URL \url{https://doi.org/10.1145/3243734.3243866}.

\bibitem[He et~al.(2017)He, Zhu, Zheng, and Lyu]{log2017-1}
Pinjia He, Jieming Zhu, Zibin Zheng, and Michael~R. Lyu.
\newblock Drain: An online log parsing approach with fixed depth tree.
\newblock In Ilkay Altintas and Shiping Chen (eds.), \emph{2017 {IEEE} International Conference on Web Services, {ICWS} 2017, Honolulu, HI, USA, June 25-30, 2017}, pp.\  33--40. {IEEE}, 2017.
\newblock \doi{10.1109/ICWS.2017.13}.
\newblock URL \url{https://doi.org/10.1109/ICWS.2017.13}.

\bibitem[He et~al.(2020)He, Zhu, He, and Lyu]{log-data-2020-1}
Shilin He, Jieming Zhu, Pinjia He, and Michael~R. Lyu.
\newblock Loghub: {A} large collection of system log datasets towards automated log analytics.
\newblock \emph{CoRR}, abs/2008.06448, 2020.
\newblock URL \url{https://arxiv.org/abs/2008.06448}.

\bibitem[He et~al.(2024)He, Zou, Lin, Zhou, Han, Yuan, and Zhang]{2024CONLINE}
Xinyi He, Jiaru Zou, Yun Lin, Mengyu Zhou, Shi Han, Zejian Yuan, and Dongmei Zhang.
\newblock {CONLINE:} complex code generation and refinement with online searching and correctness testing.
\newblock \emph{CoRR}, abs/2403.13583, 2024.
\newblock \doi{10.48550/ARXIV.2403.13583}.
\newblock URL \url{https://doi.org/10.48550/arXiv.2403.13583}.

\bibitem[He et~al.(2023{\natexlab{a}})He, Wang, Wang, Zhang, Zhang, and Li]{commit2023-1}
Yichen He, Liran Wang, Kaiyi Wang, Yupeng Zhang, Hang Zhang, and Zhoujun Li.
\newblock {COME:} commit message generation with modification embedding.
\newblock In Ren{\'{e}} Just and Gordon Fraser (eds.), \emph{Proceedings of the 32nd {ACM} {SIGSOFT} International Symposium on Software Testing and Analysis, {ISSTA} 2023, Seattle, WA, USA, July 17-21, 2023}, pp.\  792--803. {ACM}, 2023{\natexlab{a}}.
\newblock \doi{10.1145/3597926.3598096}.
\newblock URL \url{https://doi.org/10.1145/3597926.3598096}.

\bibitem[He et~al.(2023{\natexlab{b}})He, Liang, Jiao, Zhang, Yang, Wang, Tu, Shi, and Wang]{2023MAPS}
Zhiwei He, Tian Liang, Wenxiang Jiao, Zhuosheng Zhang, Yujiu Yang, Rui Wang, Zhaopeng Tu, Shuming Shi, and Xing Wang.
\newblock Exploring human-like translation strategy with large language models.
\newblock \emph{CoRR}, abs/2305.04118, 2023{\natexlab{b}}.
\newblock \doi{10.48550/ARXIV.2305.04118}.
\newblock URL \url{https://doi.org/10.48550/arXiv.2305.04118}.

\bibitem[He{-}Yueya et~al.(2023)He{-}Yueya, Poesia, Wang, and Goodman]{math2023-1}
Joy He{-}Yueya, Gabriel Poesia, Rose~E. Wang, and Noah~D. Goodman.
\newblock Solving math word problems by combining language models with symbolic solvers.
\newblock \emph{CoRR}, abs/2304.09102, 2023.
\newblock \doi{10.48550/ARXIV.2304.09102}.
\newblock URL \url{https://doi.org/10.48550/arXiv.2304.09102}.

\bibitem[Hellendoorn \& Devanbu(2017)Hellendoorn and Devanbu]{completion2017-1}
Vincent~J. Hellendoorn and Premkumar~T. Devanbu.
\newblock Are deep neural networks the best choice for modeling source code?
\newblock In Eric Bodden, Wilhelm Sch{\"{a}}fer, Arie van Deursen, and Andrea Zisman (eds.), \emph{Proceedings of the 2017 11th Joint Meeting on Foundations of Software Engineering, {ESEC/FSE} 2017, Paderborn, Germany, September 4-8, 2017}, pp.\  763--773. {ACM}, 2017.
\newblock \doi{10.1145/3106237.3106290}.
\newblock URL \url{https://doi.org/10.1145/3106237.3106290}.

\bibitem[Hellendoorn et~al.(2018)Hellendoorn, Bird, Barr, and Allamanis]{type2018-2}
Vincent~J. Hellendoorn, Christian Bird, Earl~T. Barr, and Miltiadis Allamanis.
\newblock Deep learning type inference.
\newblock In Gary~T. Leavens, Alessandro Garcia, and Corina~S. Pasareanu (eds.), \emph{Proceedings of the 2018 {ACM} Joint Meeting on European Software Engineering Conference and Symposium on the Foundations of Software Engineering, {ESEC/SIGSOFT} {FSE} 2018, Lake Buena Vista, FL, USA, November 04-09, 2018}, pp.\  152--162. {ACM}, 2018.
\newblock \doi{10.1145/3236024.3236051}.
\newblock URL \url{https://doi.org/10.1145/3236024.3236051}.

\bibitem[Hellendoorn et~al.(2020)Hellendoorn, Sutton, Singh, Maniatis, and Bieber]{fix2019-3}
Vincent~J. Hellendoorn, Charles Sutton, Rishabh Singh, Petros Maniatis, and David Bieber.
\newblock Global relational models of source code.
\newblock In \emph{8th International Conference on Learning Representations, {ICLR} 2020, Addis Ababa, Ethiopia, April 26-30, 2020}. OpenReview.net, 2020.
\newblock URL \url{https://openreview.net/forum?id=B1lnbRNtwr}.

\bibitem[Hellendoorn et~al.(2021)Hellendoorn, Tsay, Mukherjee, and Hirzel]{review2021-2}
Vincent~J. Hellendoorn, Jason Tsay, Manisha Mukherjee, and Martin Hirzel.
\newblock Towards automating code review at scale.
\newblock In Diomidis Spinellis, Georgios Gousios, Marsha Chechik, and Massimiliano~Di Penta (eds.), \emph{{ESEC/FSE} '21: 29th {ACM} Joint European Software Engineering Conference and Symposium on the Foundations of Software Engineering, Athens, Greece, August 23-28, 2021}, pp.\  1479--1482. {ACM}, 2021.
\newblock \doi{10.1145/3468264.3473134}.
\newblock URL \url{https://doi.org/10.1145/3468264.3473134}.

\bibitem[Hemphill et~al.(1990)Hemphill, Godfrey, and Doddington]{sql-data-1990}
Charles~T. Hemphill, John~J. Godfrey, and George~R. Doddington.
\newblock The {ATIS} spoken language systems pilot corpus.
\newblock In \emph{Speech and Natural Language: Proceedings of a Workshop Held at Hidden Valley, Pennsylvania, USA, June 24-27, 1990}. Morgan Kaufmann, 1990.
\newblock URL \url{https://aclanthology.org/H90-1021/}.

\bibitem[Hendrycks et~al.(2021{\natexlab{a}})Hendrycks, Basart, Kadavath, Mazeika, Arora, Guo, Burns, Puranik, He, Song, and Steinhardt]{2021APPS}
Dan Hendrycks, Steven Basart, Saurav Kadavath, Mantas Mazeika, Akul Arora, Ethan Guo, Collin Burns, Samir Puranik, Horace He, Dawn Song, and Jacob Steinhardt.
\newblock Measuring coding challenge competence with {APPS}.
\newblock In Joaquin Vanschoren and Sai{-}Kit Yeung (eds.), \emph{Proceedings of the Neural Information Processing Systems Track on Datasets and Benchmarks 1, NeurIPS Datasets and Benchmarks 2021, December 2021, virtual}, 2021{\natexlab{a}}.
\newblock URL \url{https://datasets-benchmarks-proceedings.neurips.cc/paper/2021/hash/c24cd76e1ce41366a4bbe8a49b02a028-Abstract-round2.html}.

\bibitem[Hendrycks et~al.(2021{\natexlab{b}})Hendrycks, Burns, Basart, Zou, Mazeika, Song, and Steinhardt]{2020MMLU}
Dan Hendrycks, Collin Burns, Steven Basart, Andy Zou, Mantas Mazeika, Dawn Song, and Jacob Steinhardt.
\newblock Measuring massive multitask language understanding.
\newblock In \emph{9th International Conference on Learning Representations, {ICLR} 2021, Virtual Event, Austria, May 3-7, 2021}. OpenReview.net, 2021{\natexlab{b}}.
\newblock URL \url{https://openreview.net/forum?id=d7KBjmI3GmQ}.

\bibitem[Herchi \& Abdessalem(2012)Herchi and Abdessalem]{re-model2012-1}
Hatem Herchi and Wahiba~Ben Abdessalem.
\newblock From user requirements to {UML} class diagram.
\newblock \emph{CoRR}, abs/1211.0713, 2012.
\newblock URL \url{http://arxiv.org/abs/1211.0713}.

\bibitem[Heyman \& Cutsem(2020)Heyman and Cutsem]{retrieval2020-0.4}
Geert Heyman and Tom~Van Cutsem.
\newblock Neural code search revisited: Enhancing code snippet retrieval through natural language intent.
\newblock \emph{CoRR}, abs/2008.12193, 2020.
\newblock URL \url{https://arxiv.org/abs/2008.12193}.

\bibitem[Hindle et~al.(2012)Hindle, Barr, Su, Gabel, and Devanbu]{completion2012-1}
Abram Hindle, Earl~T. Barr, Zhendong Su, Mark Gabel, and Premkumar~T. Devanbu.
\newblock On the naturalness of software.
\newblock In Martin Glinz, Gail~C. Murphy, and Mauro Pezz{\`{e}} (eds.), \emph{34th International Conference on Software Engineering, {ICSE} 2012, June 2-9, 2012, Zurich, Switzerland}, pp.\  837--847. {IEEE} Computer Society, 2012.
\newblock \doi{10.1109/ICSE.2012.6227135}.
\newblock URL \url{https://doi.org/10.1109/ICSE.2012.6227135}.

\bibitem[Ho et~al.(2020)Ho, Jain, and Abbeel]{2020diffusion}
Jonathan Ho, Ajay Jain, and Pieter Abbeel.
\newblock Denoising diffusion probabilistic models.
\newblock In Hugo Larochelle, Marc'Aurelio Ranzato, Raia Hadsell, Maria{-}Florina Balcan, and Hsuan{-}Tien Lin (eds.), \emph{Advances in Neural Information Processing Systems 33: Annual Conference on Neural Information Processing Systems 2020, NeurIPS 2020, December 6-12, 2020, virtual}, 2020.
\newblock URL \url{https://proceedings.neurips.cc/paper/2020/hash/4c5bcfec8584af0d967f1ab10179ca4b-Abstract.html}.

\bibitem[Hoang et~al.(2020)Hoang, Kang, Lo, and Lawall]{commit2020-1}
Thong Hoang, Hong~Jin Kang, David Lo, and Julia Lawall.
\newblock Cc2vec: distributed representations of code changes.
\newblock In Gregg Rothermel and Doo{-}Hwan Bae (eds.), \emph{{ICSE} '20: 42nd International Conference on Software Engineering, Seoul, South Korea, 27 June - 19 July, 2020}, pp.\  518--529. {ACM}, 2020.
\newblock \doi{10.1145/3377811.3380361}.
\newblock URL \url{https://doi.org/10.1145/3377811.3380361}.

\bibitem[Hochreiter \& Schmidhuber(1997)Hochreiter and Schmidhuber]{1997LSTM}
Sepp Hochreiter and J{\"{u}}rgen Schmidhuber.
\newblock Long short-term memory.
\newblock \emph{Neural Comput.}, 9\penalty0 (8):\penalty0 1735--1780, 1997.
\newblock \doi{10.1162/neco.1997.9.8.1735}.
\newblock URL \url{https://doi.org/10.1162/neco.1997.9.8.1735}.

\bibitem[Hoffmann et~al.(2022)Hoffmann, Borgeaud, Mensch, Buchatskaya, Cai, Rutherford, de~Las~Casas, Hendricks, Welbl, Clark, Hennigan, Noland, Millican, van~den Driessche, Damoc, Guy, Osindero, Simonyan, Elsen, Vinyals, Rae, and Sifre]{2022Chinchilla}
Jordan Hoffmann, Sebastian Borgeaud, Arthur Mensch, Elena Buchatskaya, Trevor Cai, Eliza Rutherford, Diego de~Las~Casas, Lisa~Anne Hendricks, Johannes Welbl, Aidan Clark, Tom Hennigan, Eric Noland, Katherine Millican, George van~den Driessche, Bogdan Damoc, Aurelia Guy, Simon Osindero, Karen Simonyan, Erich Elsen, Oriol Vinyals, Jack~W. Rae, and Laurent Sifre.
\newblock An empirical analysis of compute-optimal large language model training.
\newblock In Sanmi Koyejo, S.~Mohamed, A.~Agarwal, Danielle Belgrave, K.~Cho, and A.~Oh (eds.), \emph{Advances in Neural Information Processing Systems 35: Annual Conference on Neural Information Processing Systems 2022, NeurIPS 2022, New Orleans, LA, USA, November 28 - December 9, 2022}, 2022.
\newblock URL \url{http://papers.nips.cc/paper\_files/paper/2022/hash/c1e2faff6f588870935f114ebe04a3e5-Abstract-Conference.html}.

\bibitem[Hong et~al.(2023)Hong, Zheng, Chen, Cheng, Wang, Zhang, Wang, Yau, Lin, Zhou, Ran, Xiao, and Wu]{2023MetaGPT}
Sirui Hong, Xiawu Zheng, Jonathan Chen, Yuheng Cheng, Jinlin Wang, Ceyao Zhang, Zili Wang, Steven Ka~Shing Yau, Zijuan Lin, Liyang Zhou, Chenyu Ran, Lingfeng Xiao, and Chenglin Wu.
\newblock Metagpt: Meta programming for multi-agent collaborative framework.
\newblock \emph{CoRR}, abs/2308.00352, 2023.
\newblock \doi{10.48550/ARXIV.2308.00352}.
\newblock URL \url{https://doi.org/10.48550/arXiv.2308.00352}.

\bibitem[Hong et~al.(2022)Hong, Tantithamthavorn, Thongtanunam, and Aleti]{review2022-4}
Yang Hong, Chakkrit Tantithamthavorn, Patanamon Thongtanunam, and Aldeida Aleti.
\newblock Commentfinder: a simpler, faster, more accurate code review comments recommendation.
\newblock In Abhik Roychoudhury, Cristian Cadar, and Miryung Kim (eds.), \emph{Proceedings of the 30th {ACM} Joint European Software Engineering Conference and Symposium on the Foundations of Software Engineering, {ESEC/FSE} 2022, Singapore, Singapore, November 14-18, 2022}, pp.\  507--519. {ACM}, 2022.
\newblock \doi{10.1145/3540250.3549119}.
\newblock URL \url{https://doi.org/10.1145/3540250.3549119}.

\bibitem[Honovich et~al.(2023)Honovich, Scialom, Levy, and Schick]{2022Unnatural}
Or~Honovich, Thomas Scialom, Omer Levy, and Timo Schick.
\newblock Unnatural instructions: Tuning language models with (almost) no human labor.
\newblock In Anna Rogers, Jordan~L. Boyd{-}Graber, and Naoaki Okazaki (eds.), \emph{Proceedings of the 61st Annual Meeting of the Association for Computational Linguistics (Volume 1: Long Papers), {ACL} 2023, Toronto, Canada, July 9-14, 2023}, pp.\  14409--14428. Association for Computational Linguistics, 2023.
\newblock \doi{10.18653/v1/2023.acl-long.806}.
\newblock URL \url{https://doi.org/10.18653/v1/2023.acl-long.806}.

\bibitem[Hosseini \& Dolan{-}Gavitt(2022)Hosseini and Dolan{-}Gavitt]{2022BTC}
Iman Hosseini and Brendan Dolan{-}Gavitt.
\newblock Beyond the {C:} retargetable decompilation using neural machine translation.
\newblock \emph{CoRR}, abs/2212.08950, 2022.
\newblock \doi{10.48550/ARXIV.2212.08950}.
\newblock URL \url{https://doi.org/10.48550/arXiv.2212.08950}.

\bibitem[Hou et~al.(2023)Hou, Zhao, Liu, Yang, Wang, Li, Luo, Lo, Grundy, and Wang]{2023survey2}
Xinyi Hou, Yanjie Zhao, Yue Liu, Zhou Yang, Kailong Wang, Li~Li, Xiapu Luo, David Lo, John~C. Grundy, and Haoyu Wang.
\newblock Large language models for software engineering: {A} systematic literature review.
\newblock \emph{CoRR}, abs/2308.10620, 2023.
\newblock \doi{10.48550/ARXIV.2308.10620}.
\newblock URL \url{https://doi.org/10.48550/arXiv.2308.10620}.

\bibitem[Hu et~al.(2020)Hu, Peng, Zhang, Xie, and Yuan]{retrieval2020-2}
Gang Hu, Min Peng, Yihan Zhang, Qianqian Xie, and Mengting Yuan.
\newblock Neural joint attention code search over structure embeddings for software q{\&}a sites.
\newblock \emph{J. Syst. Softw.}, 170:\penalty0 110773, 2020.
\newblock \doi{10.1016/J.JSS.2020.110773}.
\newblock URL \url{https://doi.org/10.1016/j.jss.2020.110773}.

\bibitem[Hu et~al.(2018{\natexlab{a}})Hu, Li, Xia, Lo, and Jin]{sum2018-1}
Xing Hu, Ge~Li, Xin Xia, David Lo, and Zhi Jin.
\newblock Deep code comment generation.
\newblock In Foutse Khomh, Chanchal~K. Roy, and Janet Siegmund (eds.), \emph{Proceedings of the 26th Conference on Program Comprehension, {ICPC} 2018, Gothenburg, Sweden, May 27-28, 2018}, pp.\  200--210. {ACM}, 2018{\natexlab{a}}.
\newblock \doi{10.1145/3196321.3196334}.
\newblock URL \url{https://doi.org/10.1145/3196321.3196334}.

\bibitem[Hu et~al.(2018{\natexlab{b}})Hu, Li, Xia, Lo, Lu, and Jin]{sum2018-2}
Xing Hu, Ge~Li, Xin Xia, David Lo, Shuai Lu, and Zhi Jin.
\newblock Summarizing source code with transferred {API} knowledge.
\newblock In J{\'{e}}r{\^{o}}me Lang (ed.), \emph{Proceedings of the Twenty-Seventh International Joint Conference on Artificial Intelligence, {IJCAI} 2018, July 13-19, 2018, Stockholm, Sweden}, pp.\  2269--2275. ijcai.org, 2018{\natexlab{b}}.
\newblock \doi{10.24963/ijcai.2018/314}.
\newblock URL \url{https://doi.org/10.24963/ijcai.2018/314}.

\bibitem[Hu et~al.(2019)Hu, Ahmed, Mechtaev, Leong, and Roychoudhury]{fix-data-2019-1}
Yang Hu, Umair~Z. Ahmed, Sergey Mechtaev, Ben Leong, and Abhik Roychoudhury.
\newblock Re-factoring based program repair applied to programming assignments.
\newblock In \emph{34th {IEEE/ACM} International Conference on Automated Software Engineering, {ASE} 2019, San Diego, CA, USA, November 11-15, 2019}, pp.\  388--398. {IEEE}, 2019.
\newblock \doi{10.1109/ASE.2019.00044}.
\newblock URL \url{https://doi.org/10.1109/ASE.2019.00044}.

\bibitem[Huang et~al.(2022{\natexlab{a}})Huang, Zhang, Hu, Zhang, Jin, Li, Du, Guo, and Chen]{syn2022-1}
Di~Huang, Rui Zhang, Xing Hu, Xishan Zhang, Pengwei Jin, Nan Li, Zidong Du, Qi~Guo, and Yunji Chen.
\newblock Neural program synthesis with query.
\newblock In \emph{The Tenth International Conference on Learning Representations, {ICLR} 2022, Virtual Event, April 25-29, 2022}. OpenReview.net, 2022{\natexlab{a}}.
\newblock URL \url{https://openreview.net/forum?id=NyJ2KIN8P17}.

\bibitem[Huang et~al.(2024)Huang, Dai, Weng, Wu, Qing, Zhang, Cui, and Guo]{2024SOAP}
Dong Huang, Jianbo Dai, Han Weng, Puzhen Wu, Yuhao Qing, Jie~M Zhang, Heming Cui, and Zhijiang Guo.
\newblock Soap: Enhancing efficiency of generated code via self-optimization.
\newblock 2024.
\newblock URL \url{https://doi.org/10.48550/arXiv.2405.15189}.

\bibitem[Huang et~al.(2021)Huang, Tang, Shou, Gong, Xu, Jiang, Zhou, and Duan]{retrieval2021-2}
Junjie Huang, Duyu Tang, Linjun Shou, Ming Gong, Ke~Xu, Daxin Jiang, Ming Zhou, and Nan Duan.
\newblock Cosqa: 20, 000+ web queries for code search and question answering.
\newblock In Chengqing Zong, Fei Xia, Wenjie Li, and Roberto Navigli (eds.), \emph{Proceedings of the 59th Annual Meeting of the Association for Computational Linguistics and the 11th International Joint Conference on Natural Language Processing, {ACL/IJCNLP} 2021, (Volume 1: Long Papers), Virtual Event, August 1-6, 2021}, pp.\  5690--5700. Association for Computational Linguistics, 2021.
\newblock \doi{10.18653/v1/2021.acl-long.442}.
\newblock URL \url{https://doi.org/10.18653/v1/2021.acl-long.442}.

\bibitem[Huang et~al.(2022{\natexlab{b}})Huang, Wang, Zhang, Yan, Cui, Inala, Clement, Duan, and Gao]{2022ExeDS}
Junjie Huang, Chenglong Wang, Jipeng Zhang, Cong Yan, Haotian Cui, Jeevana~Priya Inala, Colin~B. Clement, Nan Duan, and Jianfeng Gao.
\newblock Execution-based evaluation for data science code generation models.
\newblock \emph{CoRR}, abs/2211.09374, 2022{\natexlab{b}}.
\newblock \doi{10.48550/ARXIV.2211.09374}.
\newblock URL \url{https://doi.org/10.48550/arXiv.2211.09374}.

\bibitem[Huang et~al.(2023{\natexlab{a}})Huang, Xu, Yang, Sun, Li, Yan, and Zhang]{fix-survey-2023-2}
Kai Huang, Zhengzi Xu, Su~Yang, Hongyu Sun, Xuejun Li, Zheng Yan, and Yuqing Zhang.
\newblock A survey on automated program repair techniques.
\newblock \emph{CoRR}, abs/2303.18184, 2023{\natexlab{a}}.
\newblock \doi{10.48550/ARXIV.2303.18184}.
\newblock URL \url{https://doi.org/10.48550/arXiv.2303.18184}.

\bibitem[Huang et~al.(2018)Huang, Xia, Xing, Lo, and Wang]{api2018-1}
Qiao Huang, Xin Xia, Zhenchang Xing, David Lo, and Xinyu Wang.
\newblock {API} method recommendation without worrying about the task-api knowledge gap.
\newblock In Marianne Huchard, Christian K{\"{a}}stner, and Gordon Fraser (eds.), \emph{Proceedings of the 33rd {ACM/IEEE} International Conference on Automated Software Engineering, {ASE} 2018, Montpellier, France, September 3-7, 2018}, pp.\  293--304. {ACM}, 2018.
\newblock \doi{10.1145/3238147.3238191}.
\newblock URL \url{https://doi.org/10.1145/3238147.3238191}.

\bibitem[Huang et~al.(2020)Huang, Jia, Zhou, Chen, Zheng, and Tang]{commit2020-3}
Yuan Huang, Nan Jia, Hao{-}Jie Zhou, Xiangping Chen, Zibin Zheng, and Mingdong Tang.
\newblock Learning human-written commit messages to document code changes.
\newblock \emph{J. Comput. Sci. Technol.}, 35\penalty0 (6):\penalty0 1258--1277, 2020.
\newblock \doi{10.1007/S11390-020-0496-0}.
\newblock URL \url{https://doi.org/10.1007/s11390-020-0496-0}.

\bibitem[Huang et~al.(2023{\natexlab{b}})Huang, Bai, Zhu, Zhang, Zhang, Su, Liu, Lv, Zhang, Lei, Fu, Sun, and He]{2023C-Eval}
Yuzhen Huang, Yuzhuo Bai, Zhihao Zhu, Junlei Zhang, Jinghan Zhang, Tangjun Su, Junteng Liu, Chuancheng Lv, Yikai Zhang, Jiayi Lei, Yao Fu, Maosong Sun, and Junxian He.
\newblock C-eval: {A} multi-level multi-discipline chinese evaluation suite for foundation models.
\newblock \emph{CoRR}, abs/2305.08322, 2023{\natexlab{b}}.
\newblock \doi{10.48550/arXiv.2305.08322}.
\newblock URL \url{https://doi.org/10.48550/arXiv.2305.08322}.

\bibitem[Huq et~al.(2022)Huq, Hasan, Haque, Mahbub, Iqbal, and Ahmed]{fix2020-4}
Faria Huq, Masum Hasan, Md. Mahim~Anjum Haque, Sazan Mahbub, Anindya Iqbal, and Toufique Ahmed.
\newblock Review4repair: Code review aided automatic program repairing.
\newblock \emph{Inf. Softw. Technol.}, 143:\penalty0 106765, 2022.
\newblock \doi{10.1016/J.INFSOF.2021.106765}.
\newblock URL \url{https://doi.org/10.1016/j.infsof.2021.106765}.

\bibitem[Husain et~al.(2019)Husain, Wu, Gazit, Allamanis, and Brockschmidt]{2019CodeSearchNet}
Hamel Husain, Ho{-}Hsiang Wu, Tiferet Gazit, Miltiadis Allamanis, and Marc Brockschmidt.
\newblock Codesearchnet challenge: Evaluating the state of semantic code search.
\newblock \emph{CoRR}, abs/1909.09436, 2019.
\newblock URL \url{http://arxiv.org/abs/1909.09436}.

\bibitem[Hwang et~al.(2019)Hwang, Yim, Park, and Seo]{sql2019-1}
Wonseok Hwang, Jinyeung Yim, Seunghyun Park, and Minjoon Seo.
\newblock A comprehensive exploration on wikisql with table-aware word contextualization.
\newblock \emph{CoRR}, abs/1902.01069, 2019.
\newblock URL \url{http://arxiv.org/abs/1902.01069}.

\bibitem[Ibrahim \& Ahmad(2010)Ibrahim and Ahmad]{re-model2010-1}
Mohd Ibrahim and Rodina Ahmad.
\newblock Class diagram extraction from textual requirements using natural language processing (nlp) techniques.
\newblock In \emph{2010 Second International Conference on Computer Research and Development}, pp.\  200--204, 2010.
\newblock \doi{10.1109/ICCRD.2010.71}.

\bibitem[Ishibashi \& Nishimura(2024)Ishibashi and Nishimura]{2024SoA}
Yoichi Ishibashi and Yoshimasa Nishimura.
\newblock Self-organized agents: A llm multi-agent framework toward ultra large-scale code generation and optimization.
\newblock \emph{CoRR}, abs/2404.02183, 2024.
\newblock \doi{10.48550/ARXIV.2404.02183}.
\newblock URL \url{https://doi.org/10.48550/arXiv.2404.02183}.

\bibitem[Iyer et~al.(2016)Iyer, Konstas, Cheung, and Zettlemoyer]{sum2016-1}
Srinivasan Iyer, Ioannis Konstas, Alvin Cheung, and Luke Zettlemoyer.
\newblock Summarizing source code using a neural attention model.
\newblock In \emph{Proceedings of the 54th Annual Meeting of the Association for Computational Linguistics, {ACL} 2016, August 7-12, 2016, Berlin, Germany, Volume 1: Long Papers}. The Association for Computer Linguistics, 2016.
\newblock \doi{10.18653/v1/p16-1195}.
\newblock URL \url{https://doi.org/10.18653/v1/p16-1195}.

\bibitem[Iyer et~al.(2017)Iyer, Konstas, Cheung, Krishnamurthy, and Zettlemoyer]{sql-data-2017-2}
Srinivasan Iyer, Ioannis Konstas, Alvin Cheung, Jayant Krishnamurthy, and Luke Zettlemoyer.
\newblock Learning a neural semantic parser from user feedback.
\newblock In Regina Barzilay and Min{-}Yen Kan (eds.), \emph{Proceedings of the 55th Annual Meeting of the Association for Computational Linguistics, {ACL} 2017, Vancouver, Canada, July 30 - August 4, Volume 1: Long Papers}, pp.\  963--973. Association for Computational Linguistics, 2017.
\newblock \doi{10.18653/V1/P17-1089}.
\newblock URL \url{https://doi.org/10.18653/v1/P17-1089}.

\bibitem[Iyer et~al.(2018)Iyer, Konstas, Cheung, and Zettlemoyer]{2018CONCODE}
Srinivasan Iyer, Ioannis Konstas, Alvin Cheung, and Luke Zettlemoyer.
\newblock Mapping language to code in programmatic context.
\newblock In Ellen Riloff, David Chiang, Julia Hockenmaier, and Jun'ichi Tsujii (eds.), \emph{Proceedings of the 2018 Conference on Empirical Methods in Natural Language Processing, Brussels, Belgium, October 31 - November 4, 2018}, pp.\  1643--1652. Association for Computational Linguistics, 2018.
\newblock \doi{10.18653/v1/d18-1192}.
\newblock URL \url{https://doi.org/10.18653/v1/d18-1192}.

\bibitem[Iyer et~al.(2022)Iyer, Lin, Pasunuru, Mihaylov, Simig, Yu, Shuster, Wang, Liu, Koura, Li, O'Horo, Pereyra, Wang, Dewan, Celikyilmaz, Zettlemoyer, and Stoyanov]{2022OPT-IML}
Srinivasan Iyer, Xi~Victoria Lin, Ramakanth Pasunuru, Todor Mihaylov, Daniel Simig, Ping Yu, Kurt Shuster, Tianlu Wang, Qing Liu, Punit~Singh Koura, Xian Li, Brian O'Horo, Gabriel Pereyra, Jeff Wang, Christopher Dewan, Asli Celikyilmaz, Luke Zettlemoyer, and Ves Stoyanov.
\newblock {OPT-IML:} scaling language model instruction meta learning through the lens of generalization.
\newblock \emph{CoRR}, abs/2212.12017, 2022.
\newblock \doi{10.48550/arXiv.2212.12017}.
\newblock URL \url{https://doi.org/10.48550/arXiv.2212.12017}.

\bibitem[Jain et~al.(2023)Jain, Anish, Singh, and Ghaisas]{re-ana2023-3}
Chirag Jain, Preethu~Rose Anish, Amrita Singh, and Smita Ghaisas.
\newblock A transformer-based approach for abstractive summarization of requirements from obligations in software engineering contracts.
\newblock In Kurt Schneider, Fabiano Dalpiaz, and Jennifer Horkoff (eds.), \emph{31st {IEEE} International Requirements Engineering Conference, {RE} 2023, Hannover, Germany, September 4-8, 2023}, pp.\  169--179. {IEEE}, 2023.
\newblock \doi{10.1109/RE57278.2023.00025}.
\newblock URL \url{https://doi.org/10.1109/RE57278.2023.00025}.

\bibitem[Jain et~al.(2022)Jain, Vaidyanath, Iyer, Natarajan, Parthasarathy, Rajamani, and Sharma]{syn2021-2}
Naman Jain, Skanda Vaidyanath, Arun~Shankar Iyer, Nagarajan Natarajan, Suresh Parthasarathy, Sriram~K. Rajamani, and Rahul Sharma.
\newblock Jigsaw: Large language models meet program synthesis.
\newblock In \emph{44th {IEEE/ACM} 44th International Conference on Software Engineering, {ICSE} 2022, Pittsburgh, PA, USA, May 25-27, 2022}, pp.\  1219--1231. {ACM}, 2022.
\newblock \doi{10.1145/3510003.3510203}.
\newblock URL \url{https://doi.org/10.1145/3510003.3510203}.

\bibitem[Jana et~al.(2023)Jana, Jha, Ju, Kishore, Mahajan, and Ganesh]{trans2023-2}
Prithwish Jana, Piyush Jha, Haoyang Ju, Gautham Kishore, Aryan Mahajan, and Vijay Ganesh.
\newblock Attention, compilation, and solver-based symbolic analysis are all you need.
\newblock \emph{CoRR}, abs/2306.06755, 2023.
\newblock \doi{10.48550/ARXIV.2306.06755}.
\newblock URL \url{https://doi.org/10.48550/arXiv.2306.06755}.

\bibitem[Jangda \& Anand(2019)Jangda and Anand]{type2019-1}
Abhinav Jangda and Gaurav Anand.
\newblock Predicting variable types in dynamically typed programming languages.
\newblock \emph{CoRR}, abs/1901.05138, 2019.
\newblock URL \url{http://arxiv.org/abs/1901.05138}.

\bibitem[Jesse \& Devanbu(2022)Jesse and Devanbu]{type-data-2022-1}
Kevin Jesse and Premkumar~T. Devanbu.
\newblock Manytypes4typescript: {A} comprehensive typescript dataset for sequence-based type inference.
\newblock In \emph{19th {IEEE/ACM} International Conference on Mining Software Repositories, {MSR} 2022, Pittsburgh, PA, USA, May 23-24, 2022}, pp.\  294--298. {ACM}, 2022.
\newblock \doi{10.1145/3524842.3528507}.
\newblock URL \url{https://doi.org/10.1145/3524842.3528507}.

\bibitem[Jesse et~al.(2021)Jesse, Devanbu, and Ahmed]{type2021-4}
Kevin Jesse, Premkumar~T. Devanbu, and Toufique Ahmed.
\newblock Learning type annotation: is big data enough?
\newblock In Diomidis Spinellis, Georgios Gousios, Marsha Chechik, and Massimiliano~Di Penta (eds.), \emph{{ESEC/FSE} '21: 29th {ACM} Joint European Software Engineering Conference and Symposium on the Foundations of Software Engineering, Athens, Greece, August 23-28, 2021}, pp.\  1483--1486. {ACM}, 2021.
\newblock \doi{10.1145/3468264.3473135}.
\newblock URL \url{https://doi.org/10.1145/3468264.3473135}.

\bibitem[Jesse et~al.(2023)Jesse, Ahmed, Devanbu, and Morgan]{analysis2023-1}
Kevin Jesse, Toufique Ahmed, Premkumar~T. Devanbu, and Emily Morgan.
\newblock Large language models and simple, stupid bugs.
\newblock In \emph{20th {IEEE/ACM} International Conference on Mining Software Repositories, {MSR} 2023, Melbourne, Australia, May 15-16, 2023}, pp.\  563--575. {IEEE}, 2023.
\newblock \doi{10.1109/MSR59073.2023.00082}.
\newblock URL \url{https://doi.org/10.1109/MSR59073.2023.00082}.

\bibitem[Jiang et~al.(2023{\natexlab{a}})Jiang, Sablayrolles, Mensch, Bamford, Chaplot, de~Las~Casas, Bressand, Lengyel, Lample, Saulnier, Lavaud, Lachaux, Stock, Scao, Lavril, Wang, Lacroix, and Sayed]{2023Mistral}
Albert~Q. Jiang, Alexandre Sablayrolles, Arthur Mensch, Chris Bamford, Devendra~Singh Chaplot, Diego de~Las~Casas, Florian Bressand, Gianna Lengyel, Guillaume Lample, Lucile Saulnier, L{\'{e}}lio~Renard Lavaud, Marie{-}Anne Lachaux, Pierre Stock, Teven~Le Scao, Thibaut Lavril, Thomas Wang, Timoth{\'{e}}e Lacroix, and William~El Sayed.
\newblock Mistral 7b.
\newblock \emph{CoRR}, abs/2310.06825, 2023{\natexlab{a}}.
\newblock \doi{10.48550/ARXIV.2310.06825}.
\newblock URL \url{https://doi.org/10.48550/arXiv.2310.06825}.

\bibitem[Jiang et~al.(2024{\natexlab{a}})Jiang, Sablayrolles, Roux, Mensch, Savary, Bamford, Chaplot, de~las Casas, Hanna, Bressand, Lengyel, Bour, Lample, Lavaud, Saulnier, Lachaux, Stock, Subramanian, Yang, Antoniak, Scao, Gervet, Lavril, Wang, Lacroix, and Sayed]{2024Mixtral}
Albert~Q. Jiang, Alexandre Sablayrolles, Antoine Roux, Arthur Mensch, Blanche Savary, Chris Bamford, Devendra~Singh Chaplot, Diego de~las Casas, Emma~Bou Hanna, Florian Bressand, Gianna Lengyel, Guillaume Bour, Guillaume Lample, Lélio~Renard Lavaud, Lucile Saulnier, Marie-Anne Lachaux, Pierre Stock, Sandeep Subramanian, Sophia Yang, Szymon Antoniak, Teven~Le Scao, Théophile Gervet, Thibaut Lavril, Thomas Wang, Timothée Lacroix, and William~El Sayed.
\newblock Mixtral of experts.
\newblock \emph{CoRR}, abs/2401.04088, 2024{\natexlab{a}}.
\newblock \doi{10.48550/ARXIV.2401.04088}.
\newblock URL \url{https://doi.org/10.48550/arXiv.2401.04088}.

\bibitem[Jiang et~al.(2022)Jiang, Olson, Toh, Molina, Donsbach, Terry, and Cai]{UI2022-3}
Ellen Jiang, Kristen Olson, Edwin Toh, Alejandra Molina, Aaron Donsbach, Michael Terry, and Carrie~J. Cai.
\newblock Promptmaker: Prompt-based prototyping with large language models.
\newblock In Simone D.~J. Barbosa, Cliff Lampe, Caroline Appert, and David~A. Shamma (eds.), \emph{{CHI} '22: {CHI} Conference on Human Factors in Computing Systems, New Orleans, LA, USA, 29 April 2022 - 5 May 2022, Extended Abstracts}, pp.\  35:1--35:8. {ACM}, 2022.
\newblock \doi{10.1145/3491101.3503564}.
\newblock URL \url{https://doi.org/10.1145/3491101.3503564}.

\bibitem[Jiang et~al.(2019)Jiang, Liu, and Jiang]{id2019-3}
Lin Jiang, Hui Liu, and He~Jiang.
\newblock Machine learning based recommendation of method names: How far are we.
\newblock In \emph{34th {IEEE/ACM} International Conference on Automated Software Engineering, {ASE} 2019, San Diego, CA, USA, November 11-15, 2019}, pp.\  602--614. {IEEE}, 2019.
\newblock \doi{10.1109/ASE.2019.00062}.
\newblock URL \url{https://doi.org/10.1109/ASE.2019.00062}.

\bibitem[Jiang et~al.(2007)Jiang, Misherghi, Su, and Glondu]{clone2007-1}
Lingxiao Jiang, Ghassan Misherghi, Zhendong Su, and St{\'{e}}phane Glondu.
\newblock {DECKARD:} scalable and accurate tree-based detection of code clones.
\newblock In \emph{29th International Conference on Software Engineering {(ICSE} 2007), Minneapolis, MN, USA, May 20-26, 2007}, pp.\  96--105. {IEEE} Computer Society, 2007.
\newblock \doi{10.1109/ICSE.2007.30}.
\newblock URL \url{https://doi.org/10.1109/ICSE.2007.30}.

\bibitem[Jiang et~al.(2021{\natexlab{a}})Jiang, Lutellier, and Tan]{fix2021-1}
Nan Jiang, Thibaud Lutellier, and Lin Tan.
\newblock {CURE:} code-aware neural machine translation for automatic program repair.
\newblock In \emph{43rd {IEEE/ACM} International Conference on Software Engineering, {ICSE} 2021, Madrid, Spain, 22-30 May 2021}, pp.\  1161--1173. {IEEE}, 2021{\natexlab{a}}.
\newblock \doi{10.1109/ICSE43902.2021.00107}.
\newblock URL \url{https://doi.org/10.1109/ICSE43902.2021.00107}.

\bibitem[Jiang et~al.(2023{\natexlab{b}})Jiang, Liu, Lutellier, and Tan]{fix2023-0.7}
Nan Jiang, Kevin Liu, Thibaud Lutellier, and Lin Tan.
\newblock Impact of code language models on automated program repair.
\newblock In \emph{45th {IEEE/ACM} International Conference on Software Engineering, {ICSE} 2023, Melbourne, Australia, May 14-20, 2023}, pp.\  1430--1442. {IEEE}, 2023{\natexlab{b}}.
\newblock \doi{10.1109/ICSE48619.2023.00125}.
\newblock URL \url{https://doi.org/10.1109/ICSE48619.2023.00125}.

\bibitem[Jiang et~al.(2023{\natexlab{c}})Jiang, Wang, Liu, Xu, Tan, and Zhang]{2023Nova+}
Nan Jiang, Chengxiao Wang, Kevin Liu, Xiangzhe Xu, Lin Tan, and Xiangyu Zhang.
\newblock Nova\({}^{\mbox{+}}\): Generative language models for binaries.
\newblock \emph{CoRR}, abs/2311.13721, 2023{\natexlab{c}}.
\newblock \doi{10.48550/ARXIV.2311.13721}.
\newblock URL \url{https://doi.org/10.48550/arXiv.2311.13721}.

\bibitem[Jiang \& McMillan(2017)Jiang and McMillan]{commit-data-2017-1}
Siyuan Jiang and Collin McMillan.
\newblock Towards automatic generation of short summaries of commits.
\newblock In Giuseppe Scanniello, David Lo, and Alexander Serebrenik (eds.), \emph{Proceedings of the 25th International Conference on Program Comprehension, {ICPC} 2017, Buenos Aires, Argentina, May 22-23, 2017}, pp.\  320--323. {IEEE} Computer Society, 2017.
\newblock \doi{10.1109/ICPC.2017.12}.
\newblock URL \url{https://doi.org/10.1109/ICPC.2017.12}.

\bibitem[Jiang et~al.(2017)Jiang, Armaly, and McMillan]{commit2017-2}
Siyuan Jiang, Ameer Armaly, and Collin McMillan.
\newblock Automatically generating commit messages from diffs using neural machine translation.
\newblock In Grigore Rosu, Massimiliano~Di Penta, and Tien~N. Nguyen (eds.), \emph{Proceedings of the 32nd {IEEE/ACM} International Conference on Automated Software Engineering, {ASE} 2017, Urbana, IL, USA, October 30 - November 03, 2017}, pp.\  135--146. {IEEE} Computer Society, 2017.
\newblock \doi{10.1109/ASE.2017.8115626}.
\newblock URL \url{https://doi.org/10.1109/ASE.2017.8115626}.

\bibitem[Jiang et~al.(2021{\natexlab{b}})Jiang, Zheng, Lyu, Li, and Lyu]{2021TreeBERT}
Xue Jiang, Zhuoran Zheng, Chen Lyu, Liang Li, and Lei Lyu.
\newblock Treebert: {A} tree-based pre-trained model for programming language.
\newblock In Cassio~P. de~Campos, Marloes~H. Maathuis, and Erik Quaeghebeur (eds.), \emph{Proceedings of the Thirty-Seventh Conference on Uncertainty in Artificial Intelligence, {UAI} 2021, Virtual Event, 27-30 July 2021}, volume 161 of \emph{Proceedings of Machine Learning Research}, pp.\  54--63. {AUAI} Press, 2021{\natexlab{b}}.
\newblock URL \url{https://proceedings.mlr.press/v161/jiang21a.html}.

\bibitem[Jiang et~al.(2024{\natexlab{b}})Jiang, He, Zhuang, and Wu]{2024CCT}
Yufan Jiang, Qiaozhi He, Xiaomin Zhuang, and Zhihua Wu.
\newblock Code comparison tuning for code large language models.
\newblock \emph{CoRR}, abs/2403.19121, 2024{\natexlab{b}}.
\newblock \doi{10.48550/ARXIV.2403.19121}.
\newblock URL \url{https://doi.org/10.48550/arXiv.2403.19121}.

\bibitem[Jiang et~al.(2023{\natexlab{d}})Jiang, Liu, Chen, Li, Huang, Huo, He, Gu, and Lyu]{log2023-6}
Zhihan Jiang, Jinyang Liu, Zhuangbin Chen, Yichen Li, Junjie Huang, Yintong Huo, Pinjia He, Jiazhen Gu, and Michael~R. Lyu.
\newblock Lilac: Log parsing using llms with adaptive parsing cache.
\newblock \emph{CoRR}, abs/2310.01796, 2023{\natexlab{d}}.
\newblock \doi{10.48550/ARXIV.2310.01796}.
\newblock URL \url{https://doi.org/10.48550/arXiv.2310.01796}.

\bibitem[Jiang et~al.(2023{\natexlab{e}})Jiang, Liu, Huang, Li, Huo, Gu, Chen, Zhu, and Lyu]{log-data-2023-1}
Zhihan Jiang, Jinyang Liu, Junjie Huang, Yichen Li, Yintong Huo, Jiazhen Gu, Zhuangbin Chen, Jieming Zhu, and Michael~R. Lyu.
\newblock A large-scale benchmark for log parsing.
\newblock \emph{CoRR}, abs/2308.10828, 2023{\natexlab{e}}.
\newblock \doi{10.48550/ARXIV.2308.10828}.
\newblock URL \url{https://doi.org/10.48550/arXiv.2308.10828}.

\bibitem[Jiao et~al.(2023)Jiao, Yu, Li, Qiu, Gu, and Shen]{trans-survey-2023-1}
Mingsheng Jiao, Tingrui Yu, Xuan Li, Guanjie Qiu, Xiaodong Gu, and Beijun Shen.
\newblock On the evaluation of neural code translation: Taxonomy and benchmark.
\newblock \emph{CoRR}, abs/2308.08961, 2023.
\newblock \doi{10.48550/ARXIV.2308.08961}.
\newblock URL \url{https://doi.org/10.48550/arXiv.2308.08961}.

\bibitem[Jimenez et~al.(2023)Jimenez, Yang, Wettig, Yao, Pei, Press, and Narasimhan]{2023SWE-bench}
Carlos~E. Jimenez, John Yang, Alexander Wettig, Shunyu Yao, Kexin Pei, Ofir Press, and Karthik Narasimhan.
\newblock Swe-bench: Can language models resolve real-world github issues?
\newblock \emph{CoRR}, abs/2310.06770, 2023.
\newblock \doi{10.48550/ARXIV.2310.06770}.
\newblock URL \url{https://doi.org/10.48550/arXiv.2310.06770}.

\bibitem[Johann et~al.(2017)Johann, Stanik, B., and Maalej]{re-ana2017-3}
Timo Johann, Christoph Stanik, Alireza M.~Alizadeh B., and Walid Maalej.
\newblock {SAFE:} {A} simple approach for feature extraction from app descriptions and app reviews.
\newblock In Ana Moreira, Jo{\~{a}}o Ara{\'{u}}jo, Jane Hayes, and Barbara Paech (eds.), \emph{25th {IEEE} International Requirements Engineering Conference, {RE} 2017, Lisbon, Portugal, September 4-8, 2017}, pp.\  21--30. {IEEE} Computer Society, 2017.
\newblock \doi{10.1109/RE.2017.71}.
\newblock URL \url{https://doi.org/10.1109/RE.2017.71}.

\bibitem[Joshi et~al.(2023)Joshi, S{\'{a}}nchez, Gulwani, Le, Verbruggen, and Radicek]{fix2022-2.5}
Harshit Joshi, Jos{\'{e}} Pablo~Cambronero S{\'{a}}nchez, Sumit Gulwani, Vu~Le, Gust Verbruggen, and Ivan Radicek.
\newblock Repair is nearly generation: Multilingual program repair with llms.
\newblock In Brian Williams, Yiling Chen, and Jennifer Neville (eds.), \emph{Thirty-Seventh {AAAI} Conference on Artificial Intelligence, {AAAI} 2023, Thirty-Fifth Conference on Innovative Applications of Artificial Intelligence, {IAAI} 2023, Thirteenth Symposium on Educational Advances in Artificial Intelligence, {EAAI} 2023, Washington, DC, USA, February 7-14, 2023}, pp.\  5131--5140. {AAAI} Press, 2023.
\newblock \doi{10.1609/AAAI.V37I4.25642}.
\newblock URL \url{https://doi.org/10.1609/aaai.v37i4.25642}.

\bibitem[Jung(2021)]{commit2021-1}
Tae{-}Hwan Jung.
\newblock Commitbert: Commit message generation using pre-trained programming language model.
\newblock \emph{CoRR}, abs/2105.14242, 2021.
\newblock URL \url{https://arxiv.org/abs/2105.14242}.

\bibitem[Just(2014)]{mutant2014-1}
Ren{\'{e}} Just.
\newblock The major mutation framework: efficient and scalable mutation analysis for java.
\newblock In Corina~S. Pasareanu and Darko Marinov (eds.), \emph{International Symposium on Software Testing and Analysis, {ISSTA} '14, San Jose, CA, {USA} - July 21 - 26, 2014}, pp.\  433--436. {ACM}, 2014.
\newblock \doi{10.1145/2610384.2628053}.
\newblock URL \url{https://doi.org/10.1145/2610384.2628053}.

\bibitem[Just et~al.(2014)Just, Jalali, and Ernst]{fix-data-2014-1}
Ren{\'{e}} Just, Darioush Jalali, and Michael~D. Ernst.
\newblock Defects4j: a database of existing faults to enable controlled testing studies for java programs.
\newblock In Corina~S. Pasareanu and Darko Marinov (eds.), \emph{International Symposium on Software Testing and Analysis, {ISSTA} '14, San Jose, CA, {USA} - July 21 - 26, 2014}, pp.\  437--440. {ACM}, 2014.
\newblock \doi{10.1145/2610384.2628055}.
\newblock URL \url{https://doi.org/10.1145/2610384.2628055}.

\bibitem[Kalgutkar et~al.(2019)Kalgutkar, Kaur, Gonzalez, Stakhanova, and Matyukhina]{author-survey-2019-1}
Vaibhavi Kalgutkar, Ratinder Kaur, Hugo Gonzalez, Natalia Stakhanova, and Alina Matyukhina.
\newblock Code authorship attribution: Methods and challenges.
\newblock \emph{{ACM} Comput. Surv.}, 52\penalty0 (1):\penalty0 3:1--3:36, 2019.
\newblock \doi{10.1145/3292577}.
\newblock URL \url{https://doi.org/10.1145/3292577}.

\bibitem[Kalyan et~al.(2018)Kalyan, Mohta, Polozov, Batra, Jain, and Gulwani]{syn2018-1}
Ashwin Kalyan, Abhishek Mohta, Oleksandr Polozov, Dhruv Batra, Prateek Jain, and Sumit Gulwani.
\newblock Neural-guided deductive search for real-time program synthesis from examples.
\newblock In \emph{6th International Conference on Learning Representations, {ICLR} 2018, Vancouver, BC, Canada, April 30 - May 3, 2018, Conference Track Proceedings}. OpenReview.net, 2018.
\newblock URL \url{https://openreview.net/forum?id=rywDjg-RW}.

\bibitem[Kanade et~al.(2020)Kanade, Maniatis, Balakrishnan, and Shi]{2019CuBERT}
Aditya Kanade, Petros Maniatis, Gogul Balakrishnan, and Kensen Shi.
\newblock Learning and evaluating contextual embedding of source code.
\newblock In \emph{Proceedings of the 37th International Conference on Machine Learning, {ICML} 2020, 13-18 July 2020, Virtual Event}, volume 119 of \emph{Proceedings of Machine Learning Research}, pp.\  5110--5121. {PMLR}, 2020.
\newblock URL \url{http://proceedings.mlr.press/v119/kanade20a.html}.

\bibitem[Kaplan et~al.(2020)Kaplan, McCandlish, Henighan, Brown, Chess, Child, Gray, Radford, Wu, and Amodei]{2020scaling}
Jared Kaplan, Sam McCandlish, Tom Henighan, Tom~B. Brown, Benjamin Chess, Rewon Child, Scott Gray, Alec Radford, Jeffrey Wu, and Dario Amodei.
\newblock Scaling laws for neural language models.
\newblock \emph{CoRR}, abs/2001.08361, 2020.
\newblock URL \url{https://arxiv.org/abs/2001.08361}.

\bibitem[Karaivanov et~al.(2014)Karaivanov, Raychev, and Vechev]{trans2014-1}
Svetoslav Karaivanov, Veselin Raychev, and Martin~T. Vechev.
\newblock Phrase-based statistical translation of programming languages.
\newblock In Andrew~P. Black, Shriram Krishnamurthi, Bernd Bruegge, and Joseph~N. Ruskiewicz (eds.), \emph{Onward! 2014, Proceedings of the 2014 {ACM} International Symposium on New Ideas, New Paradigms, and Reflections on Programming {\&} Software, part of {SPLASH} '14, Portland, OR, USA, October 20-24, 2014}, pp.\  173--184. {ACM}, 2014.
\newblock \doi{10.1145/2661136.2661148}.
\newblock URL \url{https://doi.org/10.1145/2661136.2661148}.

\bibitem[Karampatsis \& Sutton(2020)Karampatsis and Sutton]{fix-data-2019-2}
Rafael{-}Michael Karampatsis and Charles Sutton.
\newblock How often do single-statement bugs occur?: The manysstubs4j dataset.
\newblock In Sunghun Kim, Georgios Gousios, Sarah Nadi, and Joseph Hejderup (eds.), \emph{{MSR} '20: 17th International Conference on Mining Software Repositories, Seoul, Republic of Korea, 29-30 June, 2020}, pp.\  573--577. {ACM}, 2020.
\newblock \doi{10.1145/3379597.3387491}.
\newblock URL \url{https://doi.org/10.1145/3379597.3387491}.

\bibitem[Karampatsis et~al.(2020)Karampatsis, Babii, Robbes, Sutton, and Janes]{completion2020-1}
Rafael{-}Michael Karampatsis, Hlib Babii, Romain Robbes, Charles Sutton, and Andrea Janes.
\newblock Big code != big vocabulary: open-vocabulary models for source code.
\newblock In Gregg Rothermel and Doo{-}Hwan Bae (eds.), \emph{{ICSE} '20: 42nd International Conference on Software Engineering, Seoul, South Korea, 27 June - 19 July, 2020}, pp.\  1073--1085. {ACM}, 2020.
\newblock \doi{10.1145/3377811.3380342}.
\newblock URL \url{https://doi.org/10.1145/3377811.3380342}.

\bibitem[Kargaran et~al.(2023)Kargaran, Nikeghbal, Heydarnoori, and Sch{\"{u}}tze]{UI2023-1}
Amir~Hossein Kargaran, Nafiseh Nikeghbal, Abbas Heydarnoori, and Hinrich Sch{\"{u}}tze.
\newblock Menucraft: Interactive menu system design with large language models.
\newblock \emph{CoRR}, abs/2303.04496, 2023.
\newblock \doi{10.48550/ARXIV.2303.04496}.
\newblock URL \url{https://doi.org/10.48550/arXiv.2303.04496}.

\bibitem[Kashyap et~al.(2019)Kashyap, Ruchti, Kot, Turetsky, Swords, Pan, Henry, Melski, and Schulte]{mutant2019-1}
Vineeth Kashyap, Jason Ruchti, Lucja Kot, Emma Turetsky, Rebecca Swords, Shih~An Pan, Julien Henry, David Melski, and Eric~M. Schulte.
\newblock Automated customized bug-benchmark generation.
\newblock In \emph{19th International Working Conference on Source Code Analysis and Manipulation, {SCAM} 2019, Cleveland, OH, USA, September 30 - October 1, 2019}, pp.\  103--114. {IEEE}, 2019.
\newblock \doi{10.1109/SCAM.2019.00020}.
\newblock URL \url{https://doi.org/10.1109/SCAM.2019.00020}.

\bibitem[Katsogiannis{-}Meimarakis \& Koutrika(2023)Katsogiannis{-}Meimarakis and Koutrika]{sql-survey-2023-1}
George Katsogiannis{-}Meimarakis and Georgia Koutrika.
\newblock A survey on deep learning approaches for text-to-sql.
\newblock \emph{{VLDB} J.}, 32\penalty0 (4):\penalty0 905--936, 2023.
\newblock \doi{10.1007/S00778-022-00776-8}.
\newblock URL \url{https://doi.org/10.1007/s00778-022-00776-8}.

\bibitem[Kelkar et~al.(2020)Kelkar, Relan, Bhardwaj, Vaichal, and Relan]{sql2020-1}
Amol Kelkar, Rohan Relan, Vaishali Bhardwaj, Saurabh Vaichal, and Peter Relan.
\newblock Bertrand-dr: Improving text-to-sql using a discriminative re-ranker.
\newblock \emph{CoRR}, abs/2002.00557, 2020.
\newblock URL \url{https://arxiv.org/abs/2002.00557}.

\bibitem[Khajezade et~al.(2022)Khajezade, Fard, and Shehata]{clone2022-0.5}
Mohamad Khajezade, Fatemeh~Hendijani Fard, and Mohamed~S. Shehata.
\newblock Evaluating few shot and contrastive learning methods for code clone detection.
\newblock \emph{CoRR}, abs/2204.07501, 2022.
\newblock \doi{10.48550/ARXIV.2204.07501}.
\newblock URL \url{https://doi.org/10.48550/arXiv.2204.07501}.

\bibitem[Khan \& Uddin(2022)Khan and Uddin]{doc2022-1}
Junaed~Younus Khan and Gias Uddin.
\newblock Automatic code documentation generation using {GPT-3}.
\newblock In \emph{37th {IEEE/ACM} International Conference on Automated Software Engineering, {ASE} 2022, Rochester, MI, USA, October 10-14, 2022}, pp.\  174:1--174:6. {ACM}, 2022.
\newblock \doi{10.1145/3551349.3559548}.
\newblock URL \url{https://doi.org/10.1145/3551349.3559548}.

\bibitem[Khan et~al.(2023)Khan, Bari, Do, Wang, Parvez, and Joty]{2023xCodeEval}
Mohammad Abdullah~Matin Khan, M.~Saiful Bari, Xuan~Long Do, Weishi Wang, Md.~Rizwan Parvez, and Shafiq~R. Joty.
\newblock xcodeeval: {A} large scale multilingual multitask benchmark for code understanding, generation, translation and retrieval.
\newblock \emph{CoRR}, abs/2303.03004, 2023.
\newblock \doi{10.48550/ARXIV.2303.03004}.
\newblock URL \url{https://doi.org/10.48550/arXiv.2303.03004}.

\bibitem[Khanfir et~al.(2023{\natexlab{a}})Khanfir, Degiovanni, Papadakis, and Traon]{mutant2023-1}
Ahmed Khanfir, Renzo Degiovanni, Mike Papadakis, and Yves~Le Traon.
\newblock Efficient mutation testing via pre-trained language models.
\newblock \emph{CoRR}, abs/2301.03543, 2023{\natexlab{a}}.
\newblock \doi{10.48550/arXiv.2301.03543}.
\newblock URL \url{https://doi.org/10.48550/arXiv.2301.03543}.

\bibitem[Khanfir et~al.(2023{\natexlab{b}})Khanfir, Koyuncu, Papadakis, Cordy, Bissyand{\'{e}}, Klein, and Traon]{mutant2020-2}
Ahmed Khanfir, Anil Koyuncu, Mike Papadakis, Maxime Cordy, Tegawend{\'{e}}~F. Bissyand{\'{e}}, Jacques Klein, and Yves~Le Traon.
\newblock ibir: Bug-report-driven fault injection.
\newblock \emph{{ACM} Trans. Softw. Eng. Methodol.}, 32\penalty0 (2):\penalty0 33:1--33:31, 2023{\natexlab{b}}.
\newblock \doi{10.1145/3542946}.
\newblock URL \url{https://doi.org/10.1145/3542946}.

\bibitem[Kim et~al.(2018)Kim, Kim, Bissyand{\'{e}}, Choi, Li, Klein, and Traon]{search2018-1}
Kisub Kim, Dongsun Kim, Tegawend{\'{e}}~F. Bissyand{\'{e}}, Eunjong Choi, Li~Li, Jacques Klein, and Yves~Le Traon.
\newblock Facoy: a code-to-code search engine.
\newblock In Michel Chaudron, Ivica Crnkovic, Marsha Chechik, and Mark Harman (eds.), \emph{Proceedings of the 40th International Conference on Software Engineering, {ICSE} 2018, Gothenburg, Sweden, May 27 - June 03, 2018}, pp.\  946--957. {ACM}, 2018.
\newblock \doi{10.1145/3180155.3180187}.
\newblock URL \url{https://doi.org/10.1145/3180155.3180187}.

\bibitem[Kitaev et~al.(2020)Kitaev, Kaiser, and Levskaya]{2020Reformer}
Nikita Kitaev, Lukasz Kaiser, and Anselm Levskaya.
\newblock Reformer: The efficient transformer.
\newblock In \emph{8th International Conference on Learning Representations, {ICLR} 2020, Addis Ababa, Ethiopia, April 26-30, 2020}. OpenReview.net, 2020.
\newblock URL \url{https://openreview.net/forum?id=rkgNKkHtvB}.

\bibitem[Kocetkov et~al.(2023)Kocetkov, Li, allal, LI, Mou, Jernite, Mitchell, Ferrandis, Hughes, Wolf, Bahdanau, Werra, and de~Vries]{2022Stack}
Denis Kocetkov, Raymond Li, Loubna~Ben allal, Jia LI, Chenghao Mou, Yacine Jernite, Margaret Mitchell, Carlos~Mu{\~n}oz Ferrandis, Sean Hughes, Thomas Wolf, Dzmitry Bahdanau, Leandro~Von Werra, and Harm de~Vries.
\newblock The stack: 3 {TB} of permissively licensed source code.
\newblock \emph{Transactions on Machine Learning Research}, 2023.
\newblock ISSN 2835-8856.
\newblock URL \url{https://openreview.net/forum?id=pxpbTdUEpD}.

\bibitem[Kuang et~al.(2022)Kuang, Zhou, and Yang]{sum2022-3}
Li~Kuang, Cong Zhou, and Xiaoxian Yang.
\newblock Code comment generation based on graph neural network enhanced transformer model for code understanding in open-source software ecosystems.
\newblock \emph{Autom. Softw. Eng.}, 29\penalty0 (2):\penalty0 43, 2022.
\newblock \doi{10.1007/S10515-022-00341-1}.
\newblock URL \url{https://doi.org/10.1007/s10515-022-00341-1}.

\bibitem[Kulal et~al.(2019)Kulal, Pasupat, Chandra, Lee, Padon, Aiken, and Liang]{2019passk}
Sumith Kulal, Panupong Pasupat, Kartik Chandra, Mina Lee, Oded Padon, Alex Aiken, and Percy Liang.
\newblock Spoc: Search-based pseudocode to code.
\newblock In Hanna~M. Wallach, Hugo Larochelle, Alina Beygelzimer, Florence d'Alch{\'{e}}{-}Buc, Emily~B. Fox, and Roman Garnett (eds.), \emph{Advances in Neural Information Processing Systems 32: Annual Conference on Neural Information Processing Systems 2019, NeurIPS 2019, December 8-14, 2019, Vancouver, BC, Canada}, pp.\  11883--11894, 2019.
\newblock URL \url{https://proceedings.neurips.cc/paper/2019/hash/7298332f04ac004a0ca44cc69ecf6f6b-Abstract.html}.

\bibitem[Kulkarni et~al.(2023)Kulkarni, Druga, Chang, Fiannaca, Cai, and Terry]{UI-CV2}
Chinmay Kulkarni, Stefania Druga, Minsuk Chang, Alex Fiannaca, Carrie~J. Cai, and Michael Terry.
\newblock A word is worth a thousand pictures: Prompts as {AI} design material.
\newblock \emph{CoRR}, abs/2303.12647, 2023.
\newblock \doi{10.48550/ARXIV.2303.12647}.
\newblock URL \url{https://doi.org/10.48550/arXiv.2303.12647}.

\bibitem[Kumar et~al.(2022)Kumar, Nagarkar, Nalhe, and Vijayakumar]{sql-survey-2022-1}
Ayush Kumar, Parth Nagarkar, Prabhav Nalhe, and Sanjeev Vijayakumar.
\newblock Deep learning driven natural languages text to {SQL} query conversion: {A} survey.
\newblock \emph{CoRR}, abs/2208.04415, 2022.
\newblock \doi{10.48550/ARXIV.2208.04415}.
\newblock URL \url{https://doi.org/10.48550/arXiv.2208.04415}.

\bibitem[Kumar \& Sanyal(2008)Kumar and Sanyal]{re-model2008-1}
Deeptimahanti~Deva Kumar and Ratna Sanyal.
\newblock Static uml model generator from analysis of requirements (sugar).
\newblock In \emph{2008 Advanced Software Engineering and Its Applications}, pp.\  77--84, 2008.
\newblock \doi{10.1109/ASEA.2008.25}.

\bibitem[Lachaux et~al.(2021)Lachaux, Rozi{\`{e}}re, Szafraniec, and Lample]{2021DOBF}
Marie{-}Anne Lachaux, Baptiste Rozi{\`{e}}re, Marc Szafraniec, and Guillaume Lample.
\newblock {DOBF:} {A} deobfuscation pre-training objective for programming languages.
\newblock In Marc'Aurelio Ranzato, Alina Beygelzimer, Yann~N. Dauphin, Percy Liang, and Jennifer~Wortman Vaughan (eds.), \emph{Advances in Neural Information Processing Systems 34: Annual Conference on Neural Information Processing Systems 2021, NeurIPS 2021, December 6-14, 2021, virtual}, pp.\  14967--14979, 2021.
\newblock URL \url{https://proceedings.neurips.cc/paper/2021/hash/7d6548bdc0082aacc950ed35e91fcccb-Abstract.html}.

\bibitem[Lacomis et~al.(2019)Lacomis, Yin, Schwartz, Allamanis, Goues, Neubig, and Vasilescu]{ob2019-2}
Jeremy Lacomis, Pengcheng Yin, Edward~J. Schwartz, Miltiadis Allamanis, Claire~Le Goues, Graham Neubig, and Bogdan Vasilescu.
\newblock {DIRE:} {A} neural approach to decompiled identifier naming.
\newblock In \emph{34th {IEEE/ACM} International Conference on Automated Software Engineering, {ASE} 2019, San Diego, CA, USA, November 11-15, 2019}, pp.\  628--639. {IEEE}, 2019.
\newblock \doi{10.1109/ASE.2019.00064}.
\newblock URL \url{https://doi.org/10.1109/ASE.2019.00064}.

\bibitem[Lahiri et~al.(2022)Lahiri, Naik, Sakkas, Choudhury, von Veh, Musuvathi, Inala, Wang, and Gao]{2022TiCoder}
Shuvendu~K. Lahiri, Aaditya Naik, Georgios Sakkas, Piali Choudhury, Curtis von Veh, Madanlal Musuvathi, Jeevana~Priya Inala, Chenglong Wang, and Jianfeng Gao.
\newblock Interactive code generation via test-driven user-intent formalization.
\newblock \emph{CoRR}, abs/2208.05950, 2022.
\newblock \doi{10.48550/arXiv.2208.05950}.
\newblock URL \url{https://doi.org/10.48550/arXiv.2208.05950}.

\bibitem[Lai et~al.(2023)Lai, Li, Wang, Zhang, Zhong, Zettlemoyer, Yih, Fried, Wang, and Yu]{2022DS-1000}
Yuhang Lai, Chengxi Li, Yiming Wang, Tianyi Zhang, Ruiqi Zhong, Luke Zettlemoyer, Wen{-}Tau Yih, Daniel Fried, Sida~I. Wang, and Tao Yu.
\newblock {DS-1000:} {A} natural and reliable benchmark for data science code generation.
\newblock In Andreas Krause, Emma Brunskill, Kyunghyun Cho, Barbara Engelhardt, Sivan Sabato, and Jonathan Scarlett (eds.), \emph{International Conference on Machine Learning, {ICML} 2023, 23-29 July 2023, Honolulu, Hawaii, {USA}}, volume 202 of \emph{Proceedings of Machine Learning Research}, pp.\  18319--18345. {PMLR}, 2023.
\newblock URL \url{https://proceedings.mlr.press/v202/lai23b.html}.

\bibitem[Lan et~al.(2020)Lan, Chen, Goodman, Gimpel, Sharma, and Soricut]{2019ALBERT}
Zhenzhong Lan, Mingda Chen, Sebastian Goodman, Kevin Gimpel, Piyush Sharma, and Radu Soricut.
\newblock {ALBERT:} {A} lite {BERT} for self-supervised learning of language representations.
\newblock In \emph{8th International Conference on Learning Representations, {ICLR} 2020, Addis Ababa, Ethiopia, April 26-30, 2020}. OpenReview.net, 2020.
\newblock URL \url{https://openreview.net/forum?id=H1eA7AEtvS}.

\bibitem[Landauer et~al.(2022)Landauer, Onder, Skopik, and Wurzenberger]{log-survey-2022-2}
Max Landauer, Sebastian Onder, Florian Skopik, and Markus Wurzenberger.
\newblock Deep learning for anomaly detection in log data: {A} survey.
\newblock \emph{CoRR}, abs/2207.03820, 2022.
\newblock \doi{10.48550/ARXIV.2207.03820}.
\newblock URL \url{https://doi.org/10.48550/arXiv.2207.03820}.

\bibitem[Lattner \& Adve(2004)Lattner and Adve]{2004LLVM}
Chris Lattner and Vikram~S. Adve.
\newblock {LLVM:} {A} compilation framework for lifelong program analysis {\&} transformation.
\newblock In \emph{2nd {IEEE} / {ACM} International Symposium on Code Generation and Optimization {(CGO} 2004), 20-24 March 2004, San Jose, CA, {USA}}, pp.\  75--88. {IEEE} Computer Society, 2004.
\newblock \doi{10.1109/CGO.2004.1281665}.
\newblock URL \url{https://doi.org/10.1109/CGO.2004.1281665}.

\bibitem[Lauren{\c{c}}on et~al.(2022)Lauren{\c{c}}on, Saulnier, Wang, Akiki, del Moral, Scao, von Werra, Mou, Ponferrada, Nguyen, Frohberg, Sasko, Lhoest, McMillan{-}Major, Dupont, Biderman, Rogers, Allal, Toni, Pistilli, Nguyen, Nikpoor, Masoud, Colombo, de~la Rosa, Villegas, Thrush, Longpre, Nagel, Weber, Mu{\~{n}}oz, Zhu, van Strien, Alyafeai, Almubarak, Vu, Gonzalez{-}Dios, Soroa, Lo, Dey, Suarez, Gokaslan, Bose, Adelani, Phan, Tran, Yu, Pai, Chim, Lepercq, Ilic, Mitchell, Luccioni, and Jernite]{2023ROOTS}
Hugo Lauren{\c{c}}on, Lucile Saulnier, Thomas Wang, Christopher Akiki, Albert~Villanova del Moral, Teven~Le Scao, Leandro von Werra, Chenghao Mou, Eduardo~Gonz{\'{a}}lez Ponferrada, Huu Nguyen, J{\"{o}}rg Frohberg, Mario Sasko, Quentin Lhoest, Angelina McMillan{-}Major, G{\'{e}}rard Dupont, Stella Biderman, Anna Rogers, Loubna~Ben Allal, Francesco~De Toni, Giada Pistilli, Olivier Nguyen, Somaieh Nikpoor, Maraim Masoud, Pierre Colombo, Javier de~la Rosa, Paulo Villegas, Tristan Thrush, Shayne Longpre, Sebastian Nagel, Leon Weber, Manuel Mu{\~{n}}oz, Jian Zhu, Daniel van Strien, Zaid Alyafeai, Khalid Almubarak, Minh~Chien Vu, Itziar Gonzalez{-}Dios, Aitor Soroa, Kyle Lo, Manan Dey, Pedro~Ortiz Suarez, Aaron Gokaslan, Shamik Bose, David~Ifeoluwa Adelani, Long Phan, Hieu Tran, Ian Yu, Suhas Pai, Jenny Chim, Violette Lepercq, Suzana Ilic, Margaret Mitchell, Alexandra~Sasha Luccioni, and Yacine Jernite.
\newblock The bigscience {ROOTS} corpus: {A} 1.6tb composite multilingual dataset.
\newblock In \emph{NeurIPS}, 2022.
\newblock URL \url{http://papers.nips.cc/paper\_files/paper/2022/hash/ce9e92e3de2372a4b93353eb7f3dc0bd-Abstract-Datasets\_and\_Benchmarks.html}.

\bibitem[Lauren{\c{c}}on et~al.(2024)Lauren{\c{c}}on, Tronchon, and Sanh]{UI-data-2024-2}
Hugo Lauren{\c{c}}on, L{\'{e}}o Tronchon, and Victor Sanh.
\newblock Unlocking the conversion of web screenshots into {HTML} code with the websight dataset.
\newblock \emph{CoRR}, abs/2403.09029, 2024.
\newblock \doi{10.48550/ARXIV.2403.09029}.
\newblock URL \url{https://doi.org/10.48550/arXiv.2403.09029}.

\bibitem[Le et~al.(2022)Le, Wang, Gotmare, Savarese, and Hoi]{2022CodeRL}
Hung Le, Yue Wang, Akhilesh~Deepak Gotmare, Silvio Savarese, and Steven~Chu{-}Hong Hoi.
\newblock Coderl: Mastering code generation through pretrained models and deep reinforcement learning.
\newblock In \emph{NeurIPS}, 2022.
\newblock URL \url{http://papers.nips.cc/paper\_files/paper/2022/hash/8636419dea1aa9fbd25fc4248e702da4-Abstract-Conference.html}.

\bibitem[Le \& Zhang(2021)Le and Zhang]{log2021-1}
Van{-}Hoang Le and Hongyu Zhang.
\newblock Log-based anomaly detection without log parsing.
\newblock In \emph{36th {IEEE/ACM} International Conference on Automated Software Engineering, {ASE} 2021, Melbourne, Australia, November 15-19, 2021}, pp.\  492--504. {IEEE}, 2021.
\newblock \doi{10.1109/ASE51524.2021.9678773}.
\newblock URL \url{https://doi.org/10.1109/ASE51524.2021.9678773}.

\bibitem[Le \& Zhang(2023{\natexlab{a}})Le and Zhang]{log2023-1}
Van{-}Hoang Le and Hongyu Zhang.
\newblock Log parsing with prompt-based few-shot learning.
\newblock In \emph{45th {IEEE/ACM} International Conference on Software Engineering, {ICSE} 2023, Melbourne, Australia, May 14-20, 2023}, pp.\  2438--2449. {IEEE}, 2023{\natexlab{a}}.
\newblock \doi{10.1109/ICSE48619.2023.00204}.
\newblock URL \url{https://doi.org/10.1109/ICSE48619.2023.00204}.

\bibitem[Le \& Zhang(2023{\natexlab{b}})Le and Zhang]{log2023-2}
Van{-}Hoang Le and Hongyu Zhang.
\newblock Log parsing: How far can chatgpt go?
\newblock In \emph{38th {IEEE/ACM} International Conference on Automated Software Engineering, {ASE} 2023, Luxembourg, September 11-15, 2023}, pp.\  1699--1704. {IEEE}, 2023{\natexlab{b}}.
\newblock \doi{10.1109/ASE56229.2023.00206}.
\newblock URL \url{https://doi.org/10.1109/ASE56229.2023.00206}.

\bibitem[Leather \& Cummins(2020)Leather and Cummins]{comp-opt-survey2020-1}
Hugh Leather and Chris Cummins.
\newblock Machine learning in compilers: Past, present and future.
\newblock In \emph{Forum for Specification and Design Languages, {FDL} 2020, Kiel, Germany, September 15-17, 2020}, pp.\  1--8. {IEEE}, 2020.
\newblock \doi{10.1109/FDL50818.2020.9232934}.
\newblock URL \url{https://doi.org/10.1109/FDL50818.2020.9232934}.

\bibitem[LeClair et~al.(2019)LeClair, Jiang, and McMillan]{sum2019-0.5}
Alexander LeClair, Siyuan Jiang, and Collin McMillan.
\newblock A neural model for generating natural language summaries of program subroutines.
\newblock In Joanne~M. Atlee, Tevfik Bultan, and Jon Whittle (eds.), \emph{Proceedings of the 41st International Conference on Software Engineering, {ICSE} 2019, Montreal, QC, Canada, May 25-31, 2019}, pp.\  795--806. {IEEE} / {ACM}, 2019.
\newblock \doi{10.1109/ICSE.2019.00087}.
\newblock URL \url{https://doi.org/10.1109/ICSE.2019.00087}.

\bibitem[Lee et~al.(2022)Lee, Seonwoo, and Oh]{2022CS1QA}
Changyoon Lee, Yeon Seonwoo, and Alice Oh.
\newblock {CS1QA:} {A} dataset for assisting code-based question answering in an introductory programming course.
\newblock In Marine Carpuat, Marie{-}Catherine de~Marneffe, and Iv{\'{a}}n Vladimir~Meza Ru{\'{\i}}z (eds.), \emph{Proceedings of the 2022 Conference of the North American Chapter of the Association for Computational Linguistics: Human Language Technologies, {NAACL} 2022, Seattle, WA, United States, July 10-15, 2022}, pp.\  2026--2040. Association for Computational Linguistics, 2022.
\newblock \doi{10.18653/V1/2022.NAACL-MAIN.148}.
\newblock URL \url{https://doi.org/10.18653/v1/2022.naacl-main.148}.

\bibitem[Lee et~al.(2021)Lee, Polozov, and Richardson]{sql-data-2021-1}
Chia{-}Hsuan Lee, Oleksandr Polozov, and Matthew Richardson.
\newblock Kaggledbqa: Realistic evaluation of text-to-sql parsers.
\newblock In Chengqing Zong, Fei Xia, Wenjie Li, and Roberto Navigli (eds.), \emph{Proceedings of the 59th Annual Meeting of the Association for Computational Linguistics and the 11th International Joint Conference on Natural Language Processing, {ACL/IJCNLP} 2021, (Volume 1: Long Papers), Virtual Event, August 1-6, 2021}, pp.\  2261--2273. Association for Computational Linguistics, 2021.
\newblock \doi{10.18653/V1/2021.ACL-LONG.176}.
\newblock URL \url{https://doi.org/10.18653/v1/2021.acl-long.176}.

\bibitem[Lee et~al.(2018)Lee, Heo, Alur, and Naik]{syn2018-2}
Woosuk Lee, Kihong Heo, Rajeev Alur, and Mayur Naik.
\newblock Accelerating search-based program synthesis using learned probabilistic models.
\newblock In Jeffrey~S. Foster and Dan Grossman (eds.), \emph{Proceedings of the 39th {ACM} {SIGPLAN} Conference on Programming Language Design and Implementation, {PLDI} 2018, Philadelphia, PA, USA, June 18-22, 2018}, pp.\  436--449. {ACM}, 2018.
\newblock \doi{10.1145/3192366.3192410}.
\newblock URL \url{https://doi.org/10.1145/3192366.3192410}.

\bibitem[Lei et~al.(2020)Lei, Wang, Ma, Gan, Lu, Kan, and Chua]{sql2020-4.7}
Wenqiang Lei, Weixin Wang, Zhixin Ma, Tian Gan, Wei Lu, Min{-}Yen Kan, and Tat{-}Seng Chua.
\newblock Re-examining the role of schema linking in text-to-sql.
\newblock In Bonnie Webber, Trevor Cohn, Yulan He, and Yang Liu (eds.), \emph{Proceedings of the 2020 Conference on Empirical Methods in Natural Language Processing, {EMNLP} 2020, Online, November 16-20, 2020}, pp.\  6943--6954. Association for Computational Linguistics, 2020.
\newblock \doi{10.18653/V1/2020.EMNLP-MAIN.564}.
\newblock URL \url{https://doi.org/10.18653/v1/2020.emnlp-main.564}.

\bibitem[Lemieux \& Sen(2018)Lemieux and Sen]{fuzz2017-1}
Caroline Lemieux and Koushik Sen.
\newblock Fairfuzz: a targeted mutation strategy for increasing greybox fuzz testing coverage.
\newblock In Marianne Huchard, Christian K{\"{a}}stner, and Gordon Fraser (eds.), \emph{Proceedings of the 33rd {ACM/IEEE} International Conference on Automated Software Engineering, {ASE} 2018, Montpellier, France, September 3-7, 2018}, pp.\  475--485. {ACM}, 2018.
\newblock \doi{10.1145/3238147.3238176}.
\newblock URL \url{https://doi.org/10.1145/3238147.3238176}.

\bibitem[Lemieux et~al.(2023)Lemieux, Inala, Lahiri, and Sen]{unit2023-4}
Caroline Lemieux, Jeevana~Priya Inala, Shuvendu~K. Lahiri, and Siddhartha Sen.
\newblock Codamosa: Escaping coverage plateaus in test generation with pre-trained large language models.
\newblock In \emph{45th {IEEE/ACM} International Conference on Software Engineering, {ICSE} 2023, Melbourne, Australia, May 14-20, 2023}, pp.\  919--931. {IEEE}, 2023.
\newblock \doi{10.1109/ICSE48619.2023.00085}.
\newblock URL \url{https://doi.org/10.1109/ICSE48619.2023.00085}.

\bibitem[Leng et~al.(2019)Leng, Tan, Qin, Li, and Liu]{2019pivot}
Yichong Leng, Xu~Tan, Tao Qin, Xiang{-}Yang Li, and Tie{-}Yan Liu.
\newblock Unsupervised pivot translation for distant languages.
\newblock In Anna Korhonen, David~R. Traum, and Llu{\'{\i}}s M{\`{a}}rquez (eds.), \emph{Proceedings of the 57th Conference of the Association for Computational Linguistics, {ACL} 2019, Florence, Italy, July 28- August 2, 2019, Volume 1: Long Papers}, pp.\  175--183. Association for Computational Linguistics, 2019.
\newblock \doi{10.18653/v1/p19-1017}.
\newblock URL \url{https://doi.org/10.18653/v1/p19-1017}.

\bibitem[Lepikhin et~al.(2021)Lepikhin, Lee, Xu, Chen, Firat, Huang, Krikun, Shazeer, and Chen]{2020GShard}
Dmitry Lepikhin, HyoukJoong Lee, Yuanzhong Xu, Dehao Chen, Orhan Firat, Yanping Huang, Maxim Krikun, Noam Shazeer, and Zhifeng Chen.
\newblock Gshard: Scaling giant models with conditional computation and automatic sharding.
\newblock In \emph{9th International Conference on Learning Representations, {ICLR} 2021, Virtual Event, Austria, May 3-7, 2021}. OpenReview.net, 2021.
\newblock URL \url{https://openreview.net/forum?id=qrwe7XHTmYb}.

\bibitem[Lester et~al.(2021)Lester, Al{-}Rfou, and Constant]{2021T5-LM}
Brian Lester, Rami Al{-}Rfou, and Noah Constant.
\newblock The power of scale for parameter-efficient prompt tuning.
\newblock In Marie{-}Francine Moens, Xuanjing Huang, Lucia Specia, and Scott~Wen{-}tau Yih (eds.), \emph{Proceedings of the 2021 Conference on Empirical Methods in Natural Language Processing, {EMNLP} 2021, Virtual Event / Punta Cana, Dominican Republic, 7-11 November, 2021}, pp.\  3045--3059. Association for Computational Linguistics, 2021.
\newblock \doi{10.18653/v1/2021.emnlp-main.243}.
\newblock URL \url{https://doi.org/10.18653/v1/2021.emnlp-main.243}.

\bibitem[Letsholo et~al.(2013)Letsholo, Zhao, and Chioasca]{re-model2013-1}
Keletso Letsholo, Liping Zhao, and Erol{-}Valeriu Chioasca.
\newblock {TRAM:} {A} tool for transforming textual requirements into analysis models.
\newblock In Ewen Denney, Tevfik Bultan, and Andreas Zeller (eds.), \emph{2013 28th {IEEE/ACM} International Conference on Automated Software Engineering, {ASE} 2013, Silicon Valley, CA, USA, November 11-15, 2013}, pp.\  738--741. {IEEE}, 2013.
\newblock \doi{10.1109/ASE.2013.6693146}.
\newblock URL \url{https://doi.org/10.1109/ASE.2013.6693146}.

\bibitem[Lewis et~al.(2020)Lewis, Liu, Goyal, Ghazvininejad, Mohamed, Levy, Stoyanov, and Zettlemoyer]{2019BART}
Mike Lewis, Yinhan Liu, Naman Goyal, Marjan Ghazvininejad, Abdelrahman Mohamed, Omer Levy, Veselin Stoyanov, and Luke Zettlemoyer.
\newblock {BART:} denoising sequence-to-sequence pre-training for natural language generation, translation, and comprehension.
\newblock In Dan Jurafsky, Joyce Chai, Natalie Schluter, and Joel~R. Tetreault (eds.), \emph{Proceedings of the 58th Annual Meeting of the Association for Computational Linguistics, {ACL} 2020, Online, July 5-10, 2020}, pp.\  7871--7880. Association for Computational Linguistics, 2020.
\newblock \doi{10.18653/v1/2020.acl-main.703}.
\newblock URL \url{https://doi.org/10.18653/v1/2020.acl-main.703}.

\bibitem[Lewkowycz et~al.(2022)Lewkowycz, Andreassen, Dohan, Dyer, Michalewski, Ramasesh, Slone, Anil, Schlag, Gutman{-}Solo, Wu, Neyshabur, Gur{-}Ari, and Misra]{2022Minerva}
Aitor Lewkowycz, Anders Andreassen, David Dohan, Ethan Dyer, Henryk Michalewski, Vinay~V. Ramasesh, Ambrose Slone, Cem Anil, Imanol Schlag, Theo Gutman{-}Solo, Yuhuai Wu, Behnam Neyshabur, Guy Gur{-}Ari, and Vedant Misra.
\newblock Solving quantitative reasoning problems with language models.
\newblock In \emph{NeurIPS}, 2022.
\newblock URL \url{http://papers.nips.cc/paper\_files/paper/2022/hash/18abbeef8cfe9203fdf9053c9c4fe191-Abstract-Conference.html}.

\bibitem[Li et~al.(2023{\natexlab{a}})Li, Liang, Zeng, Chen, Hausman, Sadigh, Levine, Fei{-}Fei, Xia, and Ichter]{2023CoC}
Chengshu Li, Jacky Liang, Andy Zeng, Xinyun Chen, Karol Hausman, Dorsa Sadigh, Sergey Levine, Li~Fei{-}Fei, Fei Xia, and Brian Ichter.
\newblock Chain of code: Reasoning with a language model-augmented code emulator.
\newblock \emph{CoRR}, abs/2312.04474, 2023{\natexlab{a}}.
\newblock \doi{10.48550/ARXIV.2312.04474}.
\newblock URL \url{https://doi.org/10.48550/arXiv.2312.04474}.

\bibitem[Li \& Jagadish(2014)Li and Jagadish]{sql-data-2014-1}
Fei Li and H.~V. Jagadish.
\newblock Constructing an interactive natural language interface for relational databases.
\newblock \emph{Proc. {VLDB} Endow.}, 8\penalty0 (1):\penalty0 73--84, 2014.
\newblock \doi{10.14778/2735461.2735468}.
\newblock URL \url{http://www.vldb.org/pvldb/vol8/p73-li.pdf}.

\bibitem[Li et~al.(2020{\natexlab{a}})Li, Wu, Roy, Sun, Peng, Zhan, Hu, and Ma]{clone2020-0.5}
Guanhua Li, Yijian Wu, Chanchal~K. Roy, Jun Sun, Xin Peng, Nanjie Zhan, Bin Hu, and Jingyi Ma.
\newblock {SAGA:} efficient and large-scale detection of near-miss clones with {GPU} acceleration.
\newblock In Kostas Kontogiannis, Foutse Khomh, Alexander Chatzigeorgiou, Marios{-}Eleftherios Fokaefs, and Minghui Zhou (eds.), \emph{27th {IEEE} International Conference on Software Analysis, Evolution and Reengineering, {SANER} 2020, London, ON, Canada, February 18-21, 2020}, pp.\  272--283. {IEEE}, 2020{\natexlab{a}}.
\newblock \doi{10.1109/SANER48275.2020.9054832}.
\newblock URL \url{https://doi.org/10.1109/SANER48275.2020.9054832}.

\bibitem[Li et~al.(2022{\natexlab{a}})Li, Miao, Leung, Huang, Huang, Zhang, and Wang]{retrieval2022-5}
Haochen Li, Chunyan Miao, Cyril Leung, Yanxian Huang, Yuan Huang, Hongyu Zhang, and Yanlin Wang.
\newblock Exploring representation-level augmentation for code search.
\newblock In Yoav Goldberg, Zornitsa Kozareva, and Yue Zhang (eds.), \emph{Proceedings of the 2022 Conference on Empirical Methods in Natural Language Processing, {EMNLP} 2022, Abu Dhabi, United Arab Emirates, December 7-11, 2022}, pp.\  4924--4936. Association for Computational Linguistics, 2022{\natexlab{a}}.
\newblock \doi{10.18653/v1/2022.emnlp-main.327}.
\newblock URL \url{https://doi.org/10.18653/v1/2022.emnlp-main.327}.

\bibitem[Li et~al.(2023{\natexlab{b}})Li, Zhang, Koto, Yang, Zhao, Gong, Duan, and Baldwin]{2023CMMLU}
Haonan Li, Yixuan Zhang, Fajri Koto, Yifei Yang, Hai Zhao, Yeyun Gong, Nan Duan, and Timothy Baldwin.
\newblock {CMMLU:} measuring massive multitask language understanding in chinese.
\newblock \emph{CoRR}, abs/2306.09212, 2023{\natexlab{b}}.
\newblock \doi{10.48550/ARXIV.2306.09212}.
\newblock URL \url{https://doi.org/10.48550/arXiv.2306.09212}.

\bibitem[Li et~al.(2023{\natexlab{c}})Li, Zhang, Li, and Chen]{sql2023-0.2}
Haoyang Li, Jing Zhang, Cuiping Li, and Hong Chen.
\newblock {RESDSQL:} decoupling schema linking and skeleton parsing for text-to-sql.
\newblock In Brian Williams, Yiling Chen, and Jennifer Neville (eds.), \emph{Thirty-Seventh {AAAI} Conference on Artificial Intelligence, {AAAI} 2023, Thirty-Fifth Conference on Innovative Applications of Artificial Intelligence, {IAAI} 2023, Thirteenth Symposium on Educational Advances in Artificial Intelligence, {EAAI} 2023, Washington, DC, USA, February 7-14, 2023}, pp.\  13067--13075. {AAAI} Press, 2023{\natexlab{c}}.
\newblock \doi{10.1609/AAAI.V37I11.26535}.
\newblock URL \url{https://doi.org/10.1609/aaai.v37i11.26535}.

\bibitem[Li et~al.(2019{\natexlab{a}})Li, Shi, Thung, Huo, Xu, Li, and Lo]{review2019-1}
Heng{-}Yi Li, Shu{-}Ting Shi, Ferdian Thung, Xuan Huo, Bowen Xu, Ming Li, and David Lo.
\newblock Deepreview: Automatic code review using deep multi-instance learning.
\newblock In Qiang Yang, Zhi{-}Hua Zhou, Zhiguo Gong, Min{-}Ling Zhang, and Sheng{-}Jun Huang (eds.), \emph{Advances in Knowledge Discovery and Data Mining - 23rd Pacific-Asia Conference, {PAKDD} 2019, Macau, China, April 14-17, 2019, Proceedings, Part {II}}, volume 11440 of \emph{Lecture Notes in Computer Science}, pp.\  318--330. Springer, 2019{\natexlab{a}}.
\newblock \doi{10.1007/978-3-030-16145-3\_25}.
\newblock URL \url{https://doi.org/10.1007/978-3-030-16145-3\_25}.

\bibitem[Li et~al.(2019{\natexlab{b}})Li, Kim, and Chandra]{retrieval-data-2019-1}
Hongyu Li, Seohyun Kim, and Satish Chandra.
\newblock Neural code search evaluation dataset.
\newblock \emph{CoRR}, abs/1908.09804, 2019{\natexlab{b}}.
\newblock URL \url{http://arxiv.org/abs/1908.09804}.

\bibitem[Li et~al.(2023{\natexlab{d}})Li, Tao, Jin, Liu, Li, and Li]{clone2023-2}
Jia Li, Chongyang Tao, Zhi Jin, Fang Liu, Jia~Allen Li, and Ge~Li.
\newblock {ZC3:} zero-shot cross-language code clone detection.
\newblock \emph{CoRR}, abs/2308.13754, 2023{\natexlab{d}}.
\newblock \doi{10.48550/ARXIV.2308.13754}.
\newblock URL \url{https://doi.org/10.48550/arXiv.2308.13754}.

\bibitem[Li et~al.(2023{\natexlab{e}})Li, Zhao, Li, Li, and Jin]{syn2023-1}
Jia Li, Yunfei Zhao, Yongmin Li, Ge~Li, and Zhi Jin.
\newblock Acecoder: Utilizing existing code to enhance code generation, 2023{\natexlab{e}}.

\bibitem[Li et~al.(2024{\natexlab{a}})Li, Li, Zhang, Dong, and Jin]{2024EvoCodeBench}
Jia Li, Ge~Li, Xuanming Zhang, Yihong Dong, and Zhi Jin.
\newblock Evocodebench: An evolving code generation benchmark aligned with real-world code repositories.
\newblock \emph{CoRR}, abs/2404.00599, 2024{\natexlab{a}}.
\newblock \doi{10.48550/ARXIV.2404.00599}.
\newblock URL \url{https://doi.org/10.48550/arXiv.2404.00599}.

\bibitem[Li et~al.(2018{\natexlab{a}})Li, Wang, Lyu, and King]{completion2017-2}
Jian Li, Yue Wang, Michael~R. Lyu, and Irwin King.
\newblock Code completion with neural attention and pointer networks.
\newblock In J{\'{e}}r{\^{o}}me Lang (ed.), \emph{Proceedings of the Twenty-Seventh International Joint Conference on Artificial Intelligence, {IJCAI} 2018, July 13-19, 2018, Stockholm, Sweden}, pp.\  4159--4165. ijcai.org, 2018{\natexlab{a}}.
\newblock \doi{10.24963/ijcai.2018/578}.
\newblock URL \url{https://doi.org/10.24963/ijcai.2018/578}.

\bibitem[Li et~al.(2023{\natexlab{f}})Li, Hui, Cheng, Qin, Ma, Huo, Huang, Du, Si, and Li]{sql2023-0.1}
Jinyang Li, Binyuan Hui, Reynold Cheng, Bowen Qin, Chenhao Ma, Nan Huo, Fei Huang, Wenyu Du, Luo Si, and Yongbin Li.
\newblock Graphix-t5: Mixing pre-trained transformers with graph-aware layers for text-to-sql parsing.
\newblock In Brian Williams, Yiling Chen, and Jennifer Neville (eds.), \emph{Thirty-Seventh {AAAI} Conference on Artificial Intelligence, {AAAI} 2023, Thirty-Fifth Conference on Innovative Applications of Artificial Intelligence, {IAAI} 2023, Thirteenth Symposium on Educational Advances in Artificial Intelligence, {EAAI} 2023, Washington, DC, USA, February 7-14, 2023}, pp.\  13076--13084. {AAAI} Press, 2023{\natexlab{f}}.
\newblock \doi{10.1609/AAAI.V37I11.26536}.
\newblock URL \url{https://doi.org/10.1609/aaai.v37i11.26536}.

\bibitem[Li et~al.(2023{\natexlab{g}})Li, Hui, Qu, Li, Yang, Li, Wang, Qin, Cao, Geng, Huo, Zhou, Ma, Li, Chang, Huang, Cheng, and Li]{sql-data-2023-1}
Jinyang Li, Binyuan Hui, Ge~Qu, Binhua Li, Jiaxi Yang, Bowen Li, Bailin Wang, Bowen Qin, Rongyu Cao, Ruiying Geng, Nan Huo, Xuanhe Zhou, Chenhao Ma, Guoliang Li, Kevin~Chen{-}Chuan Chang, Fei Huang, Reynold Cheng, and Yongbin Li.
\newblock Can {LLM} already serve as {A} database interface? {A} big bench for large-scale database grounded text-to-sqls.
\newblock \emph{CoRR}, abs/2305.03111, 2023{\natexlab{g}}.
\newblock \doi{10.48550/ARXIV.2305.03111}.
\newblock URL \url{https://doi.org/10.48550/arXiv.2305.03111}.

\bibitem[Li et~al.(2022{\natexlab{b}})Li, Xu, Cui, and Wei]{UI2021-3}
Junlong Li, Yiheng Xu, Lei Cui, and Furu Wei.
\newblock Markuplm: Pre-training of text and markup language for visually rich document understanding.
\newblock In Smaranda Muresan, Preslav Nakov, and Aline Villavicencio (eds.), \emph{Proceedings of the 60th Annual Meeting of the Association for Computational Linguistics (Volume 1: Long Papers), {ACL} 2022, Dublin, Ireland, May 22-27, 2022}, pp.\  6078--6087. Association for Computational Linguistics, 2022{\natexlab{b}}.
\newblock \doi{10.18653/V1/2022.ACL-LONG.420}.
\newblock URL \url{https://doi.org/10.18653/v1/2022.acl-long.420}.

\bibitem[Li et~al.(2024{\natexlab{b}})Li, Tian, Hu, Luo, and Ma]{2024MMCode}
Kaixin Li, Yuchen Tian, Qisheng Hu, Ziyang Luo, and Jing Ma.
\newblock Mmcode: Evaluating multi-modal code large language models with visually rich programming problems.
\newblock \emph{CoRR}, abs/2404.09486, 2024{\natexlab{b}}.
\newblock \doi{10.48550/ARXIV.2404.09486}.
\newblock URL \url{https://doi.org/10.48550/arXiv.2404.09486}.

\bibitem[Li et~al.(2022{\natexlab{c}})Li, Yang, Jiang, Yan, Luo, Hua, Liang, and Zuo]{review2022-3}
Lingwei Li, Li~Yang, Huaxi Jiang, Jun Yan, Tiejian Luo, Zihan Hua, Geng Liang, and Chun Zuo.
\newblock {AUGER:} automatically generating review comments with pre-training models.
\newblock In Abhik Roychoudhury, Cristian Cadar, and Miryung Kim (eds.), \emph{Proceedings of the 30th {ACM} Joint European Software Engineering Conference and Symposium on the Foundations of Software Engineering, {ESEC/FSE} 2022, Singapore, Singapore, November 14-18, 2022}, pp.\  1009--1021. {ACM}, 2022{\natexlab{c}}.
\newblock \doi{10.1145/3540250.3549099}.
\newblock URL \url{https://doi.org/10.1145/3540250.3549099}.

\bibitem[Li et~al.(2023{\natexlab{h}})Li, allal, Zi, Muennighoff, Kocetkov, Mou, Marone, Akiki, LI, Chim, Liu, Zheltonozhskii, Zhuo, Wang, Dehaene, Lamy-Poirier, Monteiro, Gontier, Yee, Umapathi, Zhu, Lipkin, Oblokulov, Wang, Murthy, Stillerman, Patel, Abulkhanov, Zocca, Dey, Zhang, Bhattacharyya, Yu, Luccioni, Villegas, Zhdanov, Lee, Timor, Ding, Schlesinger, Schoelkopf, Ebert, Dao, Mishra, Gu, Anderson, Dolan-Gavitt, Contractor, Reddy, Fried, Bahdanau, Jernite, Ferrandis, Hughes, Wolf, Guha, Werra, and de~Vries]{2023StarCoder}
Raymond Li, Loubna~Ben allal, Yangtian Zi, Niklas Muennighoff, Denis Kocetkov, Chenghao Mou, Marc Marone, Christopher Akiki, Jia LI, Jenny Chim, Qian Liu, Evgenii Zheltonozhskii, Terry~Yue Zhuo, Thomas Wang, Olivier Dehaene, Joel Lamy-Poirier, Joao Monteiro, Nicolas Gontier, Ming-Ho Yee, Logesh~Kumar Umapathi, Jian Zhu, Ben Lipkin, Muhtasham Oblokulov, Zhiruo Wang, Rudra Murthy, Jason~T Stillerman, Siva~Sankalp Patel, Dmitry Abulkhanov, Marco Zocca, Manan Dey, Zhihan Zhang, Urvashi Bhattacharyya, Wenhao Yu, Sasha Luccioni, Paulo Villegas, Fedor Zhdanov, Tony Lee, Nadav Timor, Jennifer Ding, Claire~S Schlesinger, Hailey Schoelkopf, Jan Ebert, Tri Dao, Mayank Mishra, Alex Gu, Carolyn~Jane Anderson, Brendan Dolan-Gavitt, Danish Contractor, Siva Reddy, Daniel Fried, Dzmitry Bahdanau, Yacine Jernite, Carlos~Mu{\~n}oz Ferrandis, Sean Hughes, Thomas Wolf, Arjun Guha, Leandro~Von Werra, and Harm de~Vries.
\newblock Starcoder: may the source be with you!
\newblock \emph{Transactions on Machine Learning Research}, 2023{\natexlab{h}}.
\newblock ISSN 2835-8856.
\newblock URL \url{https://openreview.net/forum?id=KoFOg41haE}.
\newblock Reproducibility Certification.

\bibitem[Li et~al.(2020{\natexlab{b}})Li, Hu, and Peng]{retrieval2020-0.3}
Ruitong Li, Gang Hu, and Min Peng.
\newblock Hierarchical embedding for code search in software q{\&}a sites.
\newblock In \emph{2020 International Joint Conference on Neural Networks, {IJCNN} 2020, Glasgow, United Kingdom, July 19-24, 2020}, pp.\  1--10. {IEEE}, 2020{\natexlab{b}}.
\newblock \doi{10.1109/IJCNN48605.2020.9207101}.
\newblock URL \url{https://doi.org/10.1109/IJCNN48605.2020.9207101}.

\bibitem[Li et~al.(2022{\natexlab{d}})Li, Thickstun, Gulrajani, Liang, and Hashimoto]{2022diffusionLM}
Xiang Li, John Thickstun, Ishaan Gulrajani, Percy Liang, and Tatsunori~B. Hashimoto.
\newblock Diffusion-lm improves controllable text generation.
\newblock In \emph{NeurIPS}, 2022{\natexlab{d}}.
\newblock URL \url{http://papers.nips.cc/paper\_files/paper/2022/hash/1be5bc25d50895ee656b8c2d9eb89d6a-Abstract-Conference.html}.

\bibitem[Li et~al.(2022{\natexlab{e}})Li, Gong, Shen, Qiu, Zhang, Yao, Qi, Jiang, Chen, and Duan]{retrieval2022-2}
Xiaonan Li, Yeyun Gong, Yelong Shen, Xipeng Qiu, Hang Zhang, Bolun Yao, Weizhen Qi, Daxin Jiang, Weizhu Chen, and Nan Duan.
\newblock Coderetriever: {A} large scale contrastive pre-training method for code search.
\newblock In Yoav Goldberg, Zornitsa Kozareva, and Yue Zhang (eds.), \emph{Proceedings of the 2022 Conference on Empirical Methods in Natural Language Processing, {EMNLP} 2022, Abu Dhabi, United Arab Emirates, December 7-11, 2022}, pp.\  2898--2910. Association for Computational Linguistics, 2022{\natexlab{e}}.
\newblock \doi{10.18653/v1/2022.emnlp-main.187}.
\newblock URL \url{https://doi.org/10.18653/v1/2022.emnlp-main.187}.

\bibitem[Li et~al.(2022{\natexlab{f}})Li, Guo, Gong, Lin, Shen, Qiu, Jiang, Chen, and Duan]{2022SCodeR}
Xiaonan Li, Daya Guo, Yeyun Gong, Yun Lin, Yelong Shen, Xipeng Qiu, Daxin Jiang, Weizhu Chen, and Nan Duan.
\newblock Soft-labeled contrastive pre-training for function-level code representation.
\newblock In Yoav Goldberg, Zornitsa Kozareva, and Yue Zhang (eds.), \emph{Findings of the Association for Computational Linguistics: {EMNLP} 2022, Abu Dhabi, United Arab Emirates, December 7-11, 2022}, pp.\  118--129. Association for Computational Linguistics, 2022{\natexlab{f}}.
\newblock \doi{10.18653/v1/2022.findings-emnlp.9}.
\newblock URL \url{https://doi.org/10.18653/v1/2022.findings-emnlp.9}.

\bibitem[Li et~al.(2016)Li, Wang, Wang, Yan, Xie, and Mei]{retrieval2016-1}
Xuan Li, Zerui Wang, Qianxiang Wang, Shoumeng Yan, Tao Xie, and Hong Mei.
\newblock Relationship-aware code search for javascript frameworks.
\newblock In Thomas Zimmermann, Jane Cleland{-}Huang, and Zhendong Su (eds.), \emph{Proceedings of the 24th {ACM} {SIGSOFT} International Symposium on Foundations of Software Engineering, {FSE} 2016, Seattle, WA, USA, November 13-18, 2016}, pp.\  690--701. {ACM}, 2016.
\newblock \doi{10.1145/2950290.2950341}.
\newblock URL \url{https://doi.org/10.1145/2950290.2950341}.

\bibitem[Li et~al.(2020{\natexlab{c}})Li, He, Zhou, Zhang, and Baldridge]{UI2020-1}
Yang Li, Jiacong He, Xin Zhou, Yuan Zhang, and Jason Baldridge.
\newblock Mapping natural language instructions to mobile {UI} action sequences.
\newblock In Dan Jurafsky, Joyce Chai, Natalie Schluter, and Joel~R. Tetreault (eds.), \emph{Proceedings of the 58th Annual Meeting of the Association for Computational Linguistics, {ACL} 2020, Online, July 5-10, 2020}, pp.\  8198--8210. Association for Computational Linguistics, 2020{\natexlab{c}}.
\newblock \doi{10.18653/V1/2020.ACL-MAIN.729}.
\newblock URL \url{https://doi.org/10.18653/v1/2020.acl-main.729}.

\bibitem[Li et~al.(2019{\natexlab{c}})Li, Wang, Nguyen, and Nguyen]{defect2019-1}
Yi~Li, Shaohua Wang, Tien~N. Nguyen, and Son~Van Nguyen.
\newblock Improving bug detection via context-based code representation learning and attention-based neural networks.
\newblock \emph{Proc. {ACM} Program. Lang.}, 3\penalty0 ({OOPSLA}):\penalty0 162:1--162:30, 2019{\natexlab{c}}.
\newblock \doi{10.1145/3360588}.
\newblock URL \url{https://doi.org/10.1145/3360588}.

\bibitem[Li et~al.(2020{\natexlab{d}})Li, Wang, and Nguyen]{fix2020-3}
Yi~Li, Shaohua Wang, and Tien~N. Nguyen.
\newblock Dlfix: context-based code transformation learning for automated program repair.
\newblock In Gregg Rothermel and Doo{-}Hwan Bae (eds.), \emph{{ICSE} '20: 42nd International Conference on Software Engineering, Seoul, South Korea, 27 June - 19 July, 2020}, pp.\  602--614. {ACM}, 2020{\natexlab{d}}.
\newblock \doi{10.1145/3377811.3380345}.
\newblock URL \url{https://doi.org/10.1145/3377811.3380345}.

\bibitem[Li et~al.(2021{\natexlab{a}})Li, Wang, and Nguyen]{defect2021-0.5}
Yi~Li, Shaohua Wang, and Tien~N. Nguyen.
\newblock Vulnerability detection with fine-grained interpretations.
\newblock In Diomidis Spinellis, Georgios Gousios, Marsha Chechik, and Massimiliano~Di Penta (eds.), \emph{{ESEC/FSE} '21: 29th {ACM} Joint European Software Engineering Conference and Symposium on the Foundations of Software Engineering, Athens, Greece, August 23-28, 2021}, pp.\  292--303. {ACM}, 2021{\natexlab{a}}.
\newblock \doi{10.1145/3468264.3468597}.
\newblock URL \url{https://doi.org/10.1145/3468264.3468597}.

\bibitem[Li et~al.(2021{\natexlab{b}})Li, Wang, and Nguyen]{id2021-1}
Yi~Li, Shaohua Wang, and Tien~N. Nguyen.
\newblock A context-based automated approach for method name consistency checking and suggestion.
\newblock In \emph{43rd {IEEE/ACM} International Conference on Software Engineering, {ICSE} 2021, Madrid, Spain, 22-30 May 2021}, pp.\  574--586. {IEEE}, 2021{\natexlab{b}}.
\newblock \doi{10.1109/ICSE43902.2021.00060}.
\newblock URL \url{https://doi.org/10.1109/ICSE43902.2021.00060}.

\bibitem[Li et~al.(2022{\natexlab{g}})Li, Wang, and Nguyen]{fix2022-0.5}
Yi~Li, Shaohua Wang, and Tien~N. Nguyen.
\newblock {DEAR:} {A} novel deep learning-based approach for automated program repair.
\newblock In \emph{44th {IEEE/ACM} 44th International Conference on Software Engineering, {ICSE} 2022, Pittsburgh, PA, USA, May 25-27, 2022}, pp.\  511--523. {ACM}, 2022{\natexlab{g}}.
\newblock \doi{10.1145/3510003.3510177}.
\newblock URL \url{https://doi.org/10.1145/3510003.3510177}.

\bibitem[Li et~al.(2023{\natexlab{i}})Li, Bubeck, Eldan, Giorno, Gunasekar, and Lee]{2023Phi-1.5}
Yuanzhi Li, S{\'{e}}bastien Bubeck, Ronen Eldan, Allie~Del Giorno, Suriya Gunasekar, and Yin~Tat Lee.
\newblock Textbooks are all you need {II:} phi-1.5 technical report.
\newblock \emph{CoRR}, abs/2309.05463, 2023{\natexlab{i}}.
\newblock \doi{10.48550/arXiv.2309.05463}.
\newblock URL \url{https://doi.org/10.48550/arXiv.2309.05463}.

\bibitem[Li et~al.(2022{\natexlab{h}})Li, Choi, Chung, Kushman, Schrittwieser, Leblond, Eccles, Keeling, Gimeno, Lago, Hubert, Choy, de~Masson~d’Autume, Babuschkin, Chen, Huang, Welbl, Gowal, Cherepanov, Molloy, Mankowitz, Robson, Kohli, de~Freitas, Kavukcuoglu, and Vinyals]{2022AlphaCode}
Yujia Li, David Choi, Junyoung Chung, Nate Kushman, Julian Schrittwieser, Rémi Leblond, Tom Eccles, James Keeling, Felix Gimeno, Agustin~Dal Lago, Thomas Hubert, Peter Choy, Cyprien de~Masson~d’Autume, Igor Babuschkin, Xinyun Chen, Po-Sen Huang, Johannes Welbl, Sven Gowal, Alexey Cherepanov, James Molloy, Daniel~J. Mankowitz, Esme~Sutherland Robson, Pushmeet Kohli, Nando de~Freitas, Koray Kavukcuoglu, and Oriol Vinyals.
\newblock Competition-level code generation with alphacode.
\newblock \emph{Science}, 378\penalty0 (6624):\penalty0 1092--1097, 2022{\natexlab{h}}.
\newblock \doi{10.1126/science.abq1158}.
\newblock URL \url{https://www.science.org/doi/abs/10.1126/science.abq1158}.

\bibitem[Li et~al.(2024{\natexlab{c}})Li, Zhang, Yin, Ouyang, and Rong]{2024ProCQA}
Zehan Li, Jianfei Zhang, Chuantao Yin, Yuanxin Ouyang, and Wenge Rong.
\newblock Procqa: {A} large-scale community-based programming question answering dataset for code search.
\newblock \emph{CoRR}, abs/2403.16702, 2024{\natexlab{c}}.
\newblock \doi{10.48550/ARXIV.2403.16702}.
\newblock URL \url{https://doi.org/10.48550/arXiv.2403.16702}.

\bibitem[Li et~al.(2018{\natexlab{b}})Li, Zou, Xu, Ou, Jin, Wang, Deng, and Zhong]{defect2018-1}
Zhen Li, Deqing Zou, Shouhuai Xu, Xinyu Ou, Hai Jin, Sujuan Wang, Zhijun Deng, and Yuyi Zhong.
\newblock Vuldeepecker: {A} deep learning-based system for vulnerability detection.
\newblock In \emph{25th Annual Network and Distributed System Security Symposium, {NDSS} 2018, San Diego, California, USA, February 18-21, 2018}. The Internet Society, 2018{\natexlab{b}}.
\newblock URL \url{https://www.ndss-symposium.org/wp-content/uploads/2018/02/ndss2018\_03A-2\_Li\_paper.pdf}.

\bibitem[Li et~al.(2022{\natexlab{i}})Li, Zou, Xu, Chen, Zhu, and Jin]{defect2020-0.5}
Zhen Li, Deqing Zou, Shouhuai Xu, Zhaoxuan Chen, Yawei Zhu, and Hai Jin.
\newblock Vuldeelocator: {A} deep learning-based fine-grained vulnerability detector.
\newblock \emph{{IEEE} Trans. Dependable Secur. Comput.}, 19\penalty0 (4):\penalty0 2821--2837, 2022{\natexlab{i}}.
\newblock \doi{10.1109/TDSC.2021.3076142}.
\newblock URL \url{https://doi.org/10.1109/TDSC.2021.3076142}.

\bibitem[Li et~al.(2022{\natexlab{j}})Li, Zou, Xu, Jin, Zhu, and Chen]{defect2018-3.5}
Zhen Li, Deqing Zou, Shouhuai Xu, Hai Jin, Yawei Zhu, and Zhaoxuan Chen.
\newblock Sysevr: {A} framework for using deep learning to detect software vulnerabilities.
\newblock \emph{{IEEE} Trans. Dependable Secur. Comput.}, 19\penalty0 (4):\penalty0 2244--2258, 2022{\natexlab{j}}.
\newblock \doi{10.1109/TDSC.2021.3051525}.
\newblock URL \url{https://doi.org/10.1109/TDSC.2021.3051525}.

\bibitem[Li et~al.(2022{\natexlab{k}})Li, Lu, Guo, Duan, Jannu, Jenks, Majumder, Green, Svyatkovskiy, Fu, and Sundaresan]{review2022-2}
Zhiyu Li, Shuai Lu, Daya Guo, Nan Duan, Shailesh Jannu, Grant Jenks, Deep Majumder, Jared Green, Alexey Svyatkovskiy, Shengyu Fu, and Neel Sundaresan.
\newblock Automating code review activities by large-scale pre-training.
\newblock In Abhik Roychoudhury, Cristian Cadar, and Miryung Kim (eds.), \emph{Proceedings of the 30th {ACM} Joint European Software Engineering Conference and Symposium on the Foundations of Software Engineering, {ESEC/FSE} 2022, Singapore, Singapore, November 14-18, 2022}, pp.\  1035--1047. {ACM}, 2022{\natexlab{k}}.
\newblock \doi{10.1145/3540250.3549081}.
\newblock URL \url{https://doi.org/10.1145/3540250.3549081}.

\bibitem[Liang et~al.(2021)Liang, Cao, Hu, and Chen]{2021Neutron}
Ruigang Liang, Ying Cao, Peiwei Hu, and Kai Chen.
\newblock Neutron: an attention-based neural decompiler.
\newblock \emph{Cybersecur.}, 4\penalty0 (1):\penalty0 5, 2021.
\newblock \doi{10.1186/S42400-021-00070-0}.
\newblock URL \url{https://doi.org/10.1186/s42400-021-00070-0}.

\bibitem[Liao et~al.(2023)Liao, Pan, Sun, Ren, Huang, Xing, Jin, and Li]{2023A3-CodeGen}
Dianshu Liao, Shidong Pan, Xiaoyu Sun, Xiaoxue Ren, Qing Huang, Zhenchang Xing, Huan Jin, and Qinying Li.
\newblock A$^3$-codgen: A repository-level code generation framework for code reuse with local-aware, global-aware, and third-party-library-aware.
\newblock \emph{CoRR}, abs/2312.05772, 2023.
\newblock \doi{10.48550/ARXIV.2312.05772}.
\newblock URL \url{https://doi.org/10.48550/arXiv.2312.05772}.

\bibitem[Lin et~al.(2017)Lin, Koppel, Chen, and Solar{-}Lezama]{fix-data-2017-2}
Derrick Lin, James Koppel, Angela Chen, and Armando Solar{-}Lezama.
\newblock Quixbugs: a multi-lingual program repair benchmark set based on the quixey challenge.
\newblock In Gail~C. Murphy (ed.), \emph{Proceedings Companion of the 2017 {ACM} {SIGPLAN} International Conference on Systems, Programming, Languages, and Applications: Software for Humanity, {SPLASH} 2017, Vancouver, BC, Canada, October 23 - 27, 2017}, pp.\  55--56. {ACM}, 2017.
\newblock \doi{10.1145/3135932.3135941}.
\newblock URL \url{https://doi.org/10.1145/3135932.3135941}.

\bibitem[Lin et~al.(2024)Lin, Kim, and Chen]{2024LCG}
Feng Lin, Dong~Jae Kim, and Tse{-}Husn Chen.
\newblock When llm-based code generation meets the software development process.
\newblock \emph{CoRR}, abs/2403.15852, 2024.
\newblock \doi{10.48550/ARXIV.2403.15852}.
\newblock URL \url{https://doi.org/10.48550/arXiv.2403.15852}.

\bibitem[Lin et~al.(2018{\natexlab{a}})Lin, Zhang, Luo, Pan, Xiang, de~Vel, and Montague]{defect-data-2018-1}
Guanjun Lin, Jun Zhang, Wei Luo, Lei Pan, Yang Xiang, Olivier~Y. de~Vel, and Paul Montague.
\newblock Cross-project transfer representation learning for vulnerable function discovery.
\newblock \emph{{IEEE} Trans. Ind. Informatics}, 14\penalty0 (7):\penalty0 3289--3297, 2018{\natexlab{a}}.
\newblock \doi{10.1109/TII.2018.2821768}.
\newblock URL \url{https://doi.org/10.1109/TII.2018.2821768}.

\bibitem[Lin et~al.(2019)Lin, Xiao, Zhang, and Xiang]{defect-data-2020-2}
Guanjun Lin, Wei Xiao, Jun Zhang, and Yang Xiang.
\newblock Deep learning-based vulnerable function detection: {A} benchmark.
\newblock In Jianying Zhou, Xiapu Luo, Qingni Shen, and Zhen Xu (eds.), \emph{Information and Communications Security - 21st International Conference, {ICICS} 2019, Beijing, China, December 15-17, 2019, Revised Selected Papers}, volume 11999 of \emph{Lecture Notes in Computer Science}, pp.\  219--232. Springer, 2019.
\newblock \doi{10.1007/978-3-030-41579-2\_13}.
\newblock URL \url{https://doi.org/10.1007/978-3-030-41579-2\_13}.

\bibitem[Lin et~al.(2021)Lin, Zhang, Luo, Pan, de~Vel, Montague, and Xiang]{defect-data-2019-2}
Guanjun Lin, Jun Zhang, Wei Luo, Lei Pan, Olivier~Y. de~Vel, Paul Montague, and Yang Xiang.
\newblock Software vulnerability discovery via learning multi-domain knowledge bases.
\newblock \emph{{IEEE} Trans. Dependable Secur. Comput.}, 18\penalty0 (5):\penalty0 2469--2485, 2021.
\newblock \doi{10.1109/TDSC.2019.2954088}.
\newblock URL \url{https://doi.org/10.1109/TDSC.2019.2954088}.

\bibitem[Lin et~al.(2022)Lin, Wang, Liu, and Qiu]{2021TransformerSurvey}
Tianyang Lin, Yuxin Wang, Xiangyang Liu, and Xipeng Qiu.
\newblock A survey of transformers.
\newblock \emph{{AI} Open}, 3:\penalty0 111--132, 2022.
\newblock \doi{10.1016/J.AIOPEN.2022.10.001}.
\newblock URL \url{https://doi.org/10.1016/j.aiopen.2022.10.001}.

\bibitem[Lin et~al.(2018{\natexlab{b}})Lin, Wang, Zettlemoyer, and Ernst]{2018NL2Bash}
Xi~Victoria Lin, Chenglong Wang, Luke Zettlemoyer, and Michael~D. Ernst.
\newblock Nl2bash: {A} corpus and semantic parser for natural language interface to the linux operating system.
\newblock In Nicoletta Calzolari, Khalid Choukri, Christopher Cieri, Thierry Declerck, Sara Goggi, K{\^{o}}iti Hasida, Hitoshi Isahara, Bente Maegaard, Joseph Mariani, H{\'{e}}l{\`{e}}ne Mazo, Asunci{\'{o}}n Moreno, Jan Odijk, Stelios Piperidis, and Takenobu Tokunaga (eds.), \emph{Proceedings of the Eleventh International Conference on Language Resources and Evaluation, {LREC} 2018, Miyazaki, Japan, May 7-12, 2018}. European Language Resources Association {(ELRA)}, 2018{\natexlab{b}}.
\newblock URL \url{http://www.lrec-conf.org/proceedings/lrec2018/summaries/1021.html}.

\bibitem[Lin et~al.(2023)Lin, Gong, Shen, Wu, Fan, Lin, Duan, and Chen]{2022CPD}
Zhenghao Lin, Yeyun Gong, Yelong Shen, Tong Wu, Zhihao Fan, Chen Lin, Nan Duan, and Weizhu Chen.
\newblock Text generation with diffusion language models: {A} pre-training approach with continuous paragraph denoise.
\newblock In Andreas Krause, Emma Brunskill, Kyunghyun Cho, Barbara Engelhardt, Sivan Sabato, and Jonathan Scarlett (eds.), \emph{International Conference on Machine Learning, {ICML} 2023, 23-29 July 2023, Honolulu, Hawaii, {USA}}, volume 202 of \emph{Proceedings of Machine Learning Research}, pp.\  21051--21064. {PMLR}, 2023.
\newblock URL \url{https://proceedings.mlr.press/v202/lin23d.html}.

\bibitem[Ling et~al.(2016)Ling, Blunsom, Grefenstette, Hermann, Kocisk{\'{y}}, Wang, and Senior]{syn2016-1}
Wang Ling, Phil Blunsom, Edward Grefenstette, Karl~Moritz Hermann, Tom{\'{a}}s Kocisk{\'{y}}, Fumin Wang, and Andrew~W. Senior.
\newblock Latent predictor networks for code generation.
\newblock In \emph{Proceedings of the 54th Annual Meeting of the Association for Computational Linguistics, {ACL} 2016, August 7-12, 2016, Berlin, Germany, Volume 1: Long Papers}. The Association for Computer Linguistics, 2016.
\newblock \doi{10.18653/V1/P16-1057}.
\newblock URL \url{https://doi.org/10.18653/v1/p16-1057}.

\bibitem[Ling et~al.(2021)Ling, Wu, Wang, Pan, Ma, Xu, Liu, Wu, and Ji]{retrieval2020-1}
Xiang Ling, Lingfei Wu, Saizhuo Wang, Gaoning Pan, Tengfei Ma, Fangli Xu, Alex~X. Liu, Chunming Wu, and Shouling Ji.
\newblock Deep graph matching and searching for semantic code retrieval.
\newblock \emph{{ACM} Trans. Knowl. Discov. Data}, 15\penalty0 (5):\penalty0 88:1--88:21, 2021.
\newblock \doi{10.1145/3447571}.
\newblock URL \url{https://doi.org/10.1145/3447571}.

\bibitem[Liu et~al.(2023{\natexlab{a}})Liu, Hu, Wen, and Yu]{sql2023-0.3}
Aiwei Liu, Xuming Hu, Lijie Wen, and Philip~S. Yu.
\newblock A comprehensive evaluation of chatgpt's zero-shot text-to-sql capability.
\newblock \emph{CoRR}, abs/2303.13547, 2023{\natexlab{a}}.
\newblock \doi{10.48550/ARXIV.2303.13547}.
\newblock URL \url{https://doi.org/10.48550/arXiv.2303.13547}.

\bibitem[Liu et~al.(2023{\natexlab{b}})Liu, Chen, Liao, Gong, Wang, Lei, Liang, Chen, Shen, Zhou, Yu, and Li]{2023MFTCoder}
Bingchang Liu, Chaoyu Chen, Cong Liao, Zi~Gong, Huan Wang, Zhichao Lei, Ming Liang, Dajun Chen, Min Shen, Hailian Zhou, Hang Yu, and Jianguo Li.
\newblock Mftcoder: Boosting code llms with multitask fine-tuning.
\newblock \emph{CoRR}, abs/2311.02303, 2023{\natexlab{b}}.
\newblock \doi{10.48550/ARXIV.2311.02303}.
\newblock URL \url{https://doi.org/10.48550/arXiv.2311.02303}.

\bibitem[Liu et~al.(2022{\natexlab{a}})Liu, Xia, Lo, Liu, Hassan, and Li]{retrieval2020-0.2}
Chao Liu, Xin Xia, David Lo, Zhiwei Liu, Ahmed~E. Hassan, and Shanping Li.
\newblock Codematcher: Searching code based on sequential semantics of important query words.
\newblock \emph{{ACM} Trans. Softw. Eng. Methodol.}, 31\penalty0 (1):\penalty0 12:1--12:37, 2022{\natexlab{a}}.
\newblock \doi{10.1145/3465403}.
\newblock URL \url{https://doi.org/10.1145/3465403}.

\bibitem[Liu \& Wan(2021)Liu and Wan]{2021CodeQA}
Chenxiao Liu and Xiaojun Wan.
\newblock Codeqa: {A} question answering dataset for source code comprehension.
\newblock In Marie{-}Francine Moens, Xuanjing Huang, Lucia Specia, and Scott~Wen{-}tau Yih (eds.), \emph{Findings of the Association for Computational Linguistics: {EMNLP} 2021, Virtual Event / Punta Cana, Dominican Republic, 16-20 November, 2021}, pp.\  2618--2632. Association for Computational Linguistics, 2021.
\newblock \doi{10.18653/V1/2021.FINDINGS-EMNLP.223}.
\newblock URL \url{https://doi.org/10.18653/v1/2021.findings-emnlp.223}.

\bibitem[Liu et~al.(2020)Liu, Li, Zhao, and Jin]{2020CugLM}
Fang Liu, Ge~Li, Yunfei Zhao, and Zhi Jin.
\newblock Multi-task learning based pre-trained language model for code completion.
\newblock In \emph{35th {IEEE/ACM} International Conference on Automated Software Engineering, {ASE} 2020, Melbourne, Australia, September 21-25, 2020}, pp.\  473--485. {IEEE}, 2020.
\newblock \doi{10.1145/3324884.3416591}.
\newblock URL \url{https://doi.org/10.1145/3324884.3416591}.

\bibitem[Liu et~al.(2022{\natexlab{b}})Liu, Li, Fu, Lu, Hao, and Jin]{id2022-2}
Fang Liu, Ge~Li, Zhiyi Fu, Shuai Lu, Yiyang Hao, and Zhi Jin.
\newblock Learning to recommend method names with global context.
\newblock In \emph{44th {IEEE/ACM} 44th International Conference on Software Engineering, {ICSE} 2022, Pittsburgh, PA, USA, May 25-27, 2022}, pp.\  1294--1306. {ACM}, 2022{\natexlab{b}}.
\newblock \doi{10.1145/3510003.3510154}.
\newblock URL \url{https://doi.org/10.1145/3510003.3510154}.

\bibitem[Liu et~al.(2023{\natexlab{c}})Liu, Li, and Zhang]{trans2023-1}
Fang Liu, Jia Li, and Li~Zhang.
\newblock Syntax and domain aware model for unsupervised program translation.
\newblock In \emph{45th {IEEE/ACM} International Conference on Software Engineering, {ICSE} 2023, Melbourne, Australia, May 14-20, 2023}, pp.\  755--767. {IEEE}, 2023{\natexlab{c}}.
\newblock \doi{10.1109/ICSE48619.2023.00072}.
\newblock URL \url{https://doi.org/10.1109/ICSE48619.2023.00072}.

\bibitem[Liu et~al.(2024)Liu, Liu, Shi, Huang, Wang, Yang, and Zhang]{analysis2024-3}
Fang Liu, Yang Liu, Lin Shi, Houkun Huang, Ruifeng Wang, Zhen Yang, and Li~Zhang.
\newblock Exploring and evaluating hallucinations in llm-powered code generation.
\newblock \emph{CoRR}, abs/2404.00971, 2024.
\newblock \doi{10.48550/ARXIV.2404.00971}.
\newblock URL \url{https://doi.org/10.48550/arXiv.2404.00971}.

\bibitem[Liu et~al.(2023{\natexlab{d}})Liu, Zhu, Xiao, Fu, Han, Yang, and Ye]{2023RLTF}
Jiate Liu, Yiqin Zhu, Kaiwen Xiao, Qiang Fu, Xiao Han, Wei Yang, and Deheng Ye.
\newblock {RLTF:} reinforcement learning from unit test feedback.
\newblock \emph{CoRR}, abs/2307.04349, 2023{\natexlab{d}}.
\newblock \doi{10.48550/ARXIV.2307.04349}.
\newblock URL \url{https://doi.org/10.48550/arXiv.2307.04349}.

\bibitem[Liu et~al.(2023{\natexlab{e}})Liu, Lin, Ruffy, Tan, Li, Panda, and Zhang]{fuzz2022-6}
Jiawei Liu, Jinkun Lin, Fabian Ruffy, Cheng Tan, Jinyang Li, Aurojit Panda, and Lingming Zhang.
\newblock Nnsmith: Generating diverse and valid test cases for deep learning compilers.
\newblock In Tor~M. Aamodt, Natalie D.~Enright Jerger, and Michael~M. Swift (eds.), \emph{Proceedings of the 28th {ACM} International Conference on Architectural Support for Programming Languages and Operating Systems, Volume 2, {ASPLOS} 2023, Vancouver, BC, Canada, March 25-29, 2023}, pp.\  530--543. {ACM}, 2023{\natexlab{e}}.
\newblock \doi{10.1145/3575693.3575707}.
\newblock URL \url{https://doi.org/10.1145/3575693.3575707}.

\bibitem[Liu et~al.(2023{\natexlab{f}})Liu, Xia, Wang, and Zhang]{2023EvalPlus}
Jiawei Liu, Chunqiu~Steven Xia, Yuyao Wang, and Lingming Zhang.
\newblock Is your code generated by chatgpt really correct? rigorous evaluation of large language models for code generation.
\newblock \emph{CoRR}, abs/2305.01210, 2023{\natexlab{f}}.
\newblock \doi{10.48550/arXiv.2305.01210}.
\newblock URL \url{https://doi.org/10.48550/arXiv.2305.01210}.

\bibitem[Liu et~al.(2019{\natexlab{a}})Liu, Kim, Bissyand{\'{e}}, Kim, Kim, Koyuncu, Kim, and Traon]{id2019-2}
Kui Liu, Dongsun Kim, Tegawend{\'{e}}~F. Bissyand{\'{e}}, Tae{-}young Kim, Kisub Kim, Anil Koyuncu, Suntae Kim, and Yves~Le Traon.
\newblock Learning to spot and refactor inconsistent method names.
\newblock In Joanne~M. Atlee, Tevfik Bultan, and Jon Whittle (eds.), \emph{Proceedings of the 41st International Conference on Software Engineering, {ICSE} 2019, Montreal, QC, Canada, May 25-31, 2019}, pp.\  1--12. {IEEE} / {ACM}, 2019{\natexlab{a}}.
\newblock \doi{10.1109/ICSE.2019.00019}.
\newblock URL \url{https://doi.org/10.1109/ICSE.2019.00019}.

\bibitem[Liu et~al.(2019{\natexlab{b}})Liu, Koyuncu, Kim, and Bissyand{\'{e}}]{fix2019-1.5}
Kui Liu, Anil Koyuncu, Dongsun Kim, and Tegawend{\'{e}}~F. Bissyand{\'{e}}.
\newblock Tbar: revisiting template-based automated program repair.
\newblock In Dongmei Zhang and Anders M{\o}ller (eds.), \emph{Proceedings of the 28th {ACM} {SIGSOFT} International Symposium on Software Testing and Analysis, {ISSTA} 2019, Beijing, China, July 15-19, 2019}, pp.\  31--42. {ACM}, 2019{\natexlab{b}}.
\newblock \doi{10.1145/3293882.3330577}.
\newblock URL \url{https://doi.org/10.1145/3293882.3330577}.

\bibitem[Liu et~al.(2023{\natexlab{g}})Liu, Pinckney, Khailany, and Ren]{2023VerilogEval}
Mingjie Liu, Nathaniel~Ross Pinckney, Brucek Khailany, and Haoxing Ren.
\newblock Verilogeval: Evaluating large language models for verilog code generation.
\newblock \emph{CoRR}, abs/2309.07544, 2023{\natexlab{g}}.
\newblock \doi{10.48550/ARXIV.2309.07544}.
\newblock URL \url{https://doi.org/10.48550/arXiv.2309.07544}.

\bibitem[Liu et~al.(2019{\natexlab{c}})Liu, Liu, Zhu, Fan, Du, and Qian]{commit2019-1}
Qin Liu, Zihe Liu, Hongming Zhu, Hongfei Fan, Bowen Du, and Yu~Qian.
\newblock Generating commit messages from diffs using pointer-generator network.
\newblock In Margaret{-}Anne~D. Storey, Bram Adams, and Sonia Haiduc (eds.), \emph{Proceedings of the 16th International Conference on Mining Software Repositories, {MSR} 2019, 26-27 May 2019, Montreal, Canada}, pp.\  299--309. {IEEE} / {ACM}, 2019{\natexlab{c}}.
\newblock \doi{10.1109/MSR.2019.00056}.
\newblock URL \url{https://doi.org/10.1109/MSR.2019.00056}.

\bibitem[Liu et~al.(2023{\natexlab{h}})Liu, Fang, Lu, Zhang, Zhang, and Xie]{2023RTLCoder}
Shang Liu, Wenji Fang, Yao Lu, Qijun Zhang, Hongce Zhang, and Zhiyao Xie.
\newblock Rtlcoder: Outperforming {GPT-3.5} in design {RTL} generation with our open-source dataset and lightweight solution.
\newblock \emph{CoRR}, abs/2312.08617, 2023{\natexlab{h}}.
\newblock \doi{10.48550/ARXIV.2312.08617}.
\newblock URL \url{https://doi.org/10.48550/arXiv.2312.08617}.

\bibitem[Liu et~al.(2022{\natexlab{c}})Liu, Gao, Chen, Nie, and Liu]{commit2019-3}
Shangqing Liu, Cuiyun Gao, Sen Chen, Lun~Yiu Nie, and Yang Liu.
\newblock {ATOM:} commit message generation based on abstract syntax tree and hybrid ranking.
\newblock \emph{{IEEE} Trans. Software Eng.}, 48\penalty0 (5):\penalty0 1800--1817, 2022{\natexlab{c}}.
\newblock \doi{10.1109/TSE.2020.3038681}.
\newblock URL \url{https://doi.org/10.1109/TSE.2020.3038681}.

\bibitem[Liu et~al.(2023{\natexlab{i}})Liu, Xie, Siow, Ma, Meng, and Liu]{retrieval2021-6}
Shangqing Liu, Xiaofei Xie, Jing~Kai Siow, Lei Ma, Guozhu Meng, and Yang Liu.
\newblock Graphsearchnet: Enhancing gnns via capturing global dependencies for semantic code search.
\newblock \emph{{IEEE} Trans. Software Eng.}, 49\penalty0 (4):\penalty0 2839--2855, 2023{\natexlab{i}}.
\newblock \doi{10.1109/TSE.2022.3233901}.
\newblock URL \url{https://doi.org/10.1109/TSE.2022.3233901}.

\bibitem[Liu et~al.(2023{\natexlab{j}})Liu, Xu, and McAuley]{2023RepoBench}
Tianyang Liu, Canwen Xu, and Julian~J. McAuley.
\newblock Repobench: Benchmarking repository-level code auto-completion systems.
\newblock \emph{CoRR}, abs/2306.03091, 2023{\natexlab{j}}.
\newblock \doi{10.48550/ARXIV.2306.03091}.
\newblock URL \url{https://doi.org/10.48550/arXiv.2306.03091}.

\bibitem[Liu et~al.(2022{\natexlab{d}})Liu, Jang, Sundaresan, Allamanis, and Svyatkovskiy]{ob2022-1}
Xiaoyu Liu, Jinu Jang, Neel Sundaresan, Miltiadis Allamanis, and Alexey Svyatkovskiy.
\newblock Adaptivepaste: Code adaptation through learning semantics-aware variable usage representations.
\newblock \emph{CoRR}, abs/2205.11023, 2022{\natexlab{d}}.
\newblock \doi{10.48550/arXiv.2205.11023}.
\newblock URL \url{https://doi.org/10.48550/arXiv.2205.11023}.

\bibitem[Liu et~al.(2023{\natexlab{k}})Liu, Tao, Meng, Wang, Ma, Zhao, Chen, Yang, Jiang, and Chen]{log2023-4}
Yilun Liu, Shimin Tao, Weibin Meng, Jingyu Wang, Wenbing Ma, Yanqing Zhao, Yuhang Chen, Hao Yang, Yanfei Jiang, and Xun Chen.
\newblock Logprompt: Prompt engineering towards zero-shot and interpretable log analysis.
\newblock \emph{CoRR}, abs/2308.07610, 2023{\natexlab{k}}.
\newblock \doi{10.48550/ARXIV.2308.07610}.
\newblock URL \url{https://doi.org/10.48550/arXiv.2308.07610}.

\bibitem[Liu et~al.(2019{\natexlab{d}})Liu, Ott, Goyal, Du, Joshi, Chen, Levy, Lewis, Zettlemoyer, and Stoyanov]{2019RoBERTa}
Yinhan Liu, Myle Ott, Naman Goyal, Jingfei Du, Mandar Joshi, Danqi Chen, Omer Levy, Mike Lewis, Luke Zettlemoyer, and Veselin Stoyanov.
\newblock Roberta: {A} robustly optimized {BERT} pretraining approach.
\newblock \emph{CoRR}, abs/1907.11692, 2019{\natexlab{d}}.
\newblock URL \url{http://arxiv.org/abs/1907.11692}.

\bibitem[Liu et~al.(2022{\natexlab{e}})Liu, Zhang, He, Zhang, Li, Kang, Xu, Ma, Lin, Dang, Rajmohan, and Zhang]{log2022-1}
Yudong Liu, Xu~Zhang, Shilin He, Hongyu Zhang, Liqun Li, Yu~Kang, Yong Xu, Minghua Ma, Qingwei Lin, Yingnong Dang, Saravan Rajmohan, and Dongmei Zhang.
\newblock Uniparser: {A} unified log parser for heterogeneous log data.
\newblock In Fr{\'{e}}d{\'{e}}rique Laforest, Rapha{\"{e}}l Troncy, Elena Simperl, Deepak Agarwal, Aristides Gionis, Ivan Herman, and Lionel M{\'{e}}dini (eds.), \emph{{WWW} '22: The {ACM} Web Conference 2022, Virtual Event, Lyon, France, April 25 - 29, 2022}, pp.\  1893--1901. {ACM}, 2022{\natexlab{e}}.
\newblock \doi{10.1145/3485447.3511993}.
\newblock URL \url{https://doi.org/10.1145/3485447.3511993}.

\bibitem[Liu et~al.(2023{\natexlab{l}})Liu, Tang, Cai, Lu, Zhang, Shao, Deng, Hu, Yang, An, Huang, Si, Chen, Zhao, Li, Chen, Zong, Wang, Liu, Jiang, Chang, Qin, Zhou, Zhao, Cohan, and Gerstein]{2023ML-Bench}
Yuliang Liu, Xiangru Tang, Zefan Cai, Junjie Lu, Yichi Zhang, Yanjun Shao, Zexuan Deng, Helan Hu, Zengxian Yang, Kaikai An, Ruijun Huang, Shuzheng Si, Sheng Chen, Haozhe Zhao, Zhengliang Li, Liang Chen, Yiming Zong, Yan Wang, Tianyu Liu, Zhiwei Jiang, Baobao Chang, Yujia Qin, Wangchunshu Zhou, Yilun Zhao, Arman Cohan, and Mark Gerstein.
\newblock Ml-bench: Large language models leverage open-source libraries for machine learning tasks.
\newblock \emph{CoRR}, abs/2311.09835, 2023{\natexlab{l}}.
\newblock \doi{10.48550/ARXIV.2311.09835}.
\newblock URL \url{https://doi.org/10.48550/arXiv.2311.09835}.

\bibitem[Liu et~al.(2023{\natexlab{m}})Liu, Chen, Wang, Chen, Wu, Che, Wang, and Wang]{UI-2023-3}
Zhe Liu, Chunyang Chen, Junjie Wang, Mengzhuo Chen, Boyu Wu, Xing Che, Dandan Wang, and Qing Wang.
\newblock Chatting with {GPT-3} for zero-shot human-like mobile automated {GUI} testing.
\newblock \emph{CoRR}, abs/2305.09434, 2023{\natexlab{m}}.
\newblock \doi{10.48550/ARXIV.2305.09434}.
\newblock URL \url{https://doi.org/10.48550/arXiv.2305.09434}.

\bibitem[Liu et~al.(2023{\natexlab{n}})Liu, Tang, Luo, Zhou, and Zhang]{analysis2023-3}
Zhijie Liu, Yutian Tang, Xiapu Luo, Yuming Zhou, and Liang~Feng Zhang.
\newblock No need to lift a finger anymore? assessing the quality of code generation by chatgpt.
\newblock \emph{CoRR}, abs/2308.04838, 2023{\natexlab{n}}.
\newblock \doi{10.48550/ARXIV.2308.04838}.
\newblock URL \url{https://doi.org/10.48550/arXiv.2308.04838}.

\bibitem[Liu et~al.(2018)Liu, Xia, Hassan, Lo, Xing, and Wang]{commit2018-1}
Zhongxin Liu, Xin Xia, Ahmed~E. Hassan, David Lo, Zhenchang Xing, and Xinyu Wang.
\newblock Neural-machine-translation-based commit message generation: how far are we?
\newblock In Marianne Huchard, Christian K{\"{a}}stner, and Gordon Fraser (eds.), \emph{Proceedings of the 33rd {ACM/IEEE} International Conference on Automated Software Engineering, {ASE} 2018, Montpellier, France, September 3-7, 2018}, pp.\  373--384. {ACM}, 2018.
\newblock \doi{10.1145/3238147.3238190}.
\newblock URL \url{https://doi.org/10.1145/3238147.3238190}.

\bibitem[Long \& Rinard(2016)Long and Rinard]{fix2016-1}
Fan Long and Martin~C. Rinard.
\newblock Automatic patch generation by learning correct code.
\newblock In Rastislav Bod{\'{\i}}k and Rupak Majumdar (eds.), \emph{Proceedings of the 43rd Annual {ACM} {SIGPLAN-SIGACT} Symposium on Principles of Programming Languages, {POPL} 2016, St. Petersburg, FL, USA, January 20 - 22, 2016}, pp.\  298--312. {ACM}, 2016.
\newblock \doi{10.1145/2837614.2837617}.
\newblock URL \url{https://doi.org/10.1145/2837614.2837617}.

\bibitem[Loyola et~al.(2017)Loyola, Marrese{-}Taylor, and Matsuo]{commit2017-1}
Pablo Loyola, Edison Marrese{-}Taylor, and Yutaka Matsuo.
\newblock A neural architecture for generating natural language descriptions from source code changes.
\newblock In Regina Barzilay and Min{-}Yen Kan (eds.), \emph{Proceedings of the 55th Annual Meeting of the Association for Computational Linguistics, {ACL} 2017, Vancouver, Canada, July 30 - August 4, Volume 2: Short Papers}, pp.\  287--292. Association for Computational Linguistics, 2017.
\newblock \doi{10.18653/V1/P17-2045}.
\newblock URL \url{https://doi.org/10.18653/v1/P17-2045}.

\bibitem[Loyola et~al.(2018)Loyola, Marrese{-}Taylor, Balazs, Matsuo, and Satoh]{commit2018-2}
Pablo Loyola, Edison Marrese{-}Taylor, Jorge~A. Balazs, Yutaka Matsuo, and Fumiko Satoh.
\newblock Content aware source code change description generation.
\newblock In Emiel Krahmer, Albert Gatt, and Martijn Goudbeek (eds.), \emph{Proceedings of the 11th International Conference on Natural Language Generation, Tilburg University, The Netherlands, November 5-8, 2018}, pp.\  119--128. Association for Computational Linguistics, 2018.
\newblock \doi{10.18653/V1/W18-6513}.
\newblock URL \url{https://doi.org/10.18653/v1/w18-6513}.

\bibitem[Lozhkov et~al.(2024)Lozhkov, Li, Allal, Cassano, Lamy{-}Poirier, Tazi, Tang, Pykhtar, Liu, Wei, Liu, Tian, Kocetkov, Zucker, Belkada, Wang, Liu, Abulkhanov, Paul, Li, Li, Risdal, Li, Zhu, Zhuo, Zheltonozhskii, Dade, Yu, Krau{\ss}, Jain, Su, He, Dey, Abati, Chai, Muennighoff, Tang, Oblokulov, Akiki, Marone, Mou, Mishra, Gu, Hui, Dao, Zebaze, Dehaene, Patry, Xu, McAuley, Hu, Scholak, Paquet, Robinson, Anderson, Chapados, and et~al.]{2024StarCoder2}
Anton Lozhkov, Raymond Li, Loubna~Ben Allal, Federico Cassano, Joel Lamy{-}Poirier, Nouamane Tazi, Ao~Tang, Dmytro Pykhtar, Jiawei Liu, Yuxiang Wei, Tianyang Liu, Max Tian, Denis Kocetkov, Arthur Zucker, Younes Belkada, Zijian Wang, Qian Liu, Dmitry Abulkhanov, Indraneil Paul, Zhuang Li, Wen{-}Ding Li, Megan Risdal, Jia Li, Jian Zhu, Terry~Yue Zhuo, Evgenii Zheltonozhskii, Nii Osae~Osae Dade, Wenhao Yu, Lucas Krau{\ss}, Naman Jain, Yixuan Su, Xuanli He, Manan Dey, Edoardo Abati, Yekun Chai, Niklas Muennighoff, Xiangru Tang, Muhtasham Oblokulov, Christopher Akiki, Marc Marone, Chenghao Mou, Mayank Mishra, Alex Gu, Binyuan Hui, Tri Dao, Armel Zebaze, Olivier Dehaene, Nicolas Patry, Canwen Xu, Julian~J. McAuley, Han Hu, Torsten Scholak, S{\'{e}}bastien Paquet, Jennifer Robinson, Carolyn~Jane Anderson, Nicolas Chapados, and et~al.
\newblock Starcoder 2 and the stack v2: The next generation.
\newblock \emph{CoRR}, abs/2402.19173, 2024.
\newblock \doi{10.48550/ARXIV.2402.19173}.
\newblock URL \url{https://doi.org/10.48550/arXiv.2402.19173}.

\bibitem[Lu et~al.(2023{\natexlab{a}})Lu, Yu, Li, Yang, and Zuo]{review2023-2}
Junyi Lu, Lei Yu, Xiaojia Li, Li~Yang, and Chun Zuo.
\newblock Llama-reviewer: Advancing code review automation with large language models through parameter-efficient fine-tuning (practical experience report).
\newblock \emph{CoRR}, abs/2308.11148, 2023{\natexlab{a}}.
\newblock \doi{10.48550/arXiv.2308.11148}.
\newblock URL \url{https://doi.org/10.48550/arXiv.2308.11148}.

\bibitem[Lu et~al.(2015)Lu, Sun, Wang, Lo, and Duan]{retrieval2015-1}
Meili Lu, Xiaobing Sun, Shaowei Wang, David Lo, and Yucong Duan.
\newblock Query expansion via wordnet for effective code search.
\newblock In Yann{-}Ga{\"{e}}l Gu{\'{e}}h{\'{e}}neuc, Bram Adams, and Alexander Serebrenik (eds.), \emph{22nd {IEEE} International Conference on Software Analysis, Evolution, and Reengineering, {SANER} 2015, Montreal, QC, Canada, March 2-6, 2015}, pp.\  545--549. {IEEE} Computer Society, 2015.
\newblock \doi{10.1109/SANER.2015.7081874}.
\newblock URL \url{https://doi.org/10.1109/SANER.2015.7081874}.

\bibitem[Lu \& Liang(2017)Lu and Liang]{re-ana2017-1}
Mengmeng Lu and Peng Liang.
\newblock Automatic classification of non-functional requirements from augmented app user reviews.
\newblock In Emilia Mendes, Steve Counsell, and Kai Petersen (eds.), \emph{Proceedings of the 21st International Conference on Evaluation and Assessment in Software Engineering, {EASE} 2017, Karlskrona, Sweden, June 15-16, 2017}, pp.\  344--353. {ACM}, 2017.
\newblock \doi{10.1145/3084226.3084241}.
\newblock URL \url{https://doi.org/10.1145/3084226.3084241}.

\bibitem[Lu et~al.(2021)Lu, Guo, Ren, Huang, Svyatkovskiy, Blanco, Clement, Drain, Jiang, Tang, Li, Zhou, Shou, Zhou, Tufano, Gong, Zhou, Duan, Sundaresan, Deng, Fu, and Liu]{2021CodeXGLUE}
Shuai Lu, Daya Guo, Shuo Ren, Junjie Huang, Alexey Svyatkovskiy, Ambrosio Blanco, Colin~B. Clement, Dawn Drain, Daxin Jiang, Duyu Tang, Ge~Li, Lidong Zhou, Linjun Shou, Long Zhou, Michele Tufano, Ming Gong, Ming Zhou, Nan Duan, Neel Sundaresan, Shao~Kun Deng, Shengyu Fu, and Shujie Liu.
\newblock Codexglue: {A} machine learning benchmark dataset for code understanding and generation.
\newblock In Joaquin Vanschoren and Sai{-}Kit Yeung (eds.), \emph{Proceedings of the Neural Information Processing Systems Track on Datasets and Benchmarks 1, NeurIPS Datasets and Benchmarks 2021, December 2021, virtual}, 2021.
\newblock URL \url{https://datasets-benchmarks-proceedings.neurips.cc/paper/2021/hash/c16a5320fa475530d9583c34fd356ef5-Abstract-round1.html}.

\bibitem[Lu et~al.(2018)Lu, Wei, Li, and Wang]{log2018-1}
Siyang Lu, Xiang Wei, Yandong Li, and Liqiang Wang.
\newblock Detecting anomaly in big data system logs using convolutional neural network.
\newblock In \emph{2018 {IEEE} 16th Intl Conf on Dependable, Autonomic and Secure Computing, 16th Intl Conf on Pervasive Intelligence and Computing, 4th Intl Conf on Big Data Intelligence and Computing and Cyber Science and Technology Congress, DASC/PiCom/DataCom/CyberSciTech 2018, Athens, Greece, August 12-15, 2018}, pp.\  151--158. {IEEE} Computer Society, 2018.
\newblock \doi{10.1109/DASC/PICOM/DATACOM/CYBERSCITEC.2018.00037}.
\newblock URL \url{https://doi.org/10.1109/DASC/PiCom/DataCom/CyberSciTec.2018.00037}.

\bibitem[Lu et~al.(2017)Lu, Li, Zhao, Wen, and Jin]{api2017-3}
Yangyang Lu, Ge~Li, Zelong Zhao, Linfeng Wen, and Zhi Jin.
\newblock Learning to infer {API} mappings from {API} documents.
\newblock In Gang Li, Yong Ge, Zili Zhang, Zhi Jin, and Michael Blumenstein (eds.), \emph{Knowledge Science, Engineering and Management - 10th International Conference, {KSEM} 2017, Melbourne, VIC, Australia, August 19-20, 2017, Proceedings}, volume 10412 of \emph{Lecture Notes in Computer Science}, pp.\  237--248. Springer, 2017.
\newblock \doi{10.1007/978-3-319-63558-3\_20}.
\newblock URL \url{https://doi.org/10.1007/978-3-319-63558-3\_20}.

\bibitem[Lu et~al.(2023{\natexlab{b}})Lu, Liu, Zhang, and Xie]{2023RTLLM}
Yao Lu, Shang Liu, Qijun Zhang, and Zhiyao Xie.
\newblock {RTLLM:} an open-source benchmark for design {RTL} generation with large language model.
\newblock \emph{CoRR}, abs/2308.05345, 2023{\natexlab{b}}.
\newblock \doi{10.48550/ARXIV.2308.05345}.
\newblock URL \url{https://doi.org/10.48550/arXiv.2308.05345}.

\bibitem[Lu et~al.(2023{\natexlab{c}})Lu, Tong, Zhao, Zhang, and Li]{UI-2023-5}
Yuwen Lu, Ziang Tong, Qinyi Zhao, Chengzhi Zhang, and Toby~Jia{-}Jun Li.
\newblock {UI} layout generation with llms guided by {UI} grammar.
\newblock \emph{CoRR}, abs/2310.15455, 2023{\natexlab{c}}.
\newblock \doi{10.48550/ARXIV.2310.15455}.
\newblock URL \url{https://doi.org/10.48550/arXiv.2310.15455}.

\bibitem[Luan et~al.(2019)Luan, Yang, Barnaby, Sen, and Chandra]{search2018-2}
Sifei Luan, Di~Yang, Celeste Barnaby, Koushik Sen, and Satish Chandra.
\newblock Aroma: code recommendation via structural code search.
\newblock \emph{Proc. {ACM} Program. Lang.}, 3\penalty0 ({OOPSLA}):\penalty0 152:1--152:28, 2019.
\newblock \doi{10.1145/3360578}.
\newblock URL \url{https://doi.org/10.1145/3360578}.

\bibitem[Lucchetti \& Guha(2024)Lucchetti and Guha]{type2023-5}
Francesca Lucchetti and Arjun Guha.
\newblock Activation steering for robust type prediction in codellms.
\newblock \emph{CoRR}, abs/2404.01903, 2024.
\newblock \doi{10.48550/ARXIV.2404.01903}.
\newblock URL \url{https://doi.org/10.48550/arXiv.2404.01903}.

\bibitem[Luo et~al.(2023)Luo, Xu, Zhao, Sun, Geng, Hu, Tao, Ma, Lin, and Jiang]{2023WizardCoder}
Ziyang Luo, Can Xu, Pu~Zhao, Qingfeng Sun, Xiubo Geng, Wenxiang Hu, Chongyang Tao, Jing Ma, Qingwei Lin, and Daxin Jiang.
\newblock Wizardcoder: Empowering code large language models with evol-instruct.
\newblock \emph{CoRR}, abs/2306.08568, 2023.
\newblock \doi{10.48550/arXiv.2306.08568}.
\newblock URL \url{https://doi.org/10.48550/arXiv.2306.08568}.

\bibitem[Lutellier et~al.(2020)Lutellier, Pham, Pang, Li, Wei, and Tan]{fix2020-2}
Thibaud Lutellier, Hung~Viet Pham, Lawrence Pang, Yitong Li, Moshi Wei, and Lin Tan.
\newblock Coconut: combining context-aware neural translation models using ensemble for program repair.
\newblock In Sarfraz Khurshid and Corina~S. Pasareanu (eds.), \emph{{ISSTA} '20: 29th {ACM} {SIGSOFT} International Symposium on Software Testing and Analysis, Virtual Event, USA, July 18-22, 2020}, pp.\  101--114. {ACM}, 2020.
\newblock \doi{10.1145/3395363.3397369}.
\newblock URL \url{https://doi.org/10.1145/3395363.3397369}.

\bibitem[Lv et~al.(2015)Lv, Zhang, Lou, Wang, Zhang, and Zhao]{retrieval2015-2}
Fei Lv, Hongyu Zhang, Jian{-}Guang Lou, Shaowei Wang, Dongmei Zhang, and Jianjun Zhao.
\newblock Codehow: Effective code search based on {API} understanding and extended boolean model {(E)}.
\newblock In Myra~B. Cohen, Lars Grunske, and Michael Whalen (eds.), \emph{30th {IEEE/ACM} International Conference on Automated Software Engineering, {ASE} 2015, Lincoln, NE, USA, November 9-13, 2015}, pp.\  260--270. {IEEE} Computer Society, 2015.
\newblock \doi{10.1109/ASE.2015.42}.
\newblock URL \url{https://doi.org/10.1109/ASE.2015.42}.

\bibitem[Lyu et~al.(2020)Lyu, Chakrabarti, Hathi, Kundu, Zhang, and Chen]{sql2020-2.4}
Qin Lyu, Kaushik Chakrabarti, Shobhit Hathi, Souvik Kundu, Jianwen Zhang, and Zheng Chen.
\newblock Hybrid ranking network for text-to-sql.
\newblock \emph{CoRR}, abs/2008.04759, 2020.
\newblock URL \url{https://arxiv.org/abs/2008.04759}.

\bibitem[Ma et~al.(2024)Ma, Chen, Kim, Chen, and Wang]{log2024-1}
Zeyang Ma, An~Ran Chen, Dong~Jae Kim, Tse{-}Hsun Chen, and Shaowei Wang.
\newblock Llmparser: An exploratory study on using large language models for log parsing.
\newblock In \emph{Proceedings of the 46th {IEEE/ACM} International Conference on Software Engineering, {ICSE} 2024, Lisbon, Portugal, April 14-20, 2024}, pp.\  99:1--99:13. {ACM}, 2024.
\newblock \doi{10.1145/3597503.3639150}.
\newblock URL \url{https://doi.org/10.1145/3597503.3639150}.

\bibitem[Macedo et~al.(2024)Macedo, Tian, C{\^{o}}go, and Adams]{trans2024-1}
Marcos Macedo, Yuan Tian, Filipe~Roseiro C{\^{o}}go, and Bram Adams.
\newblock Exploring the impact of the output format on the evaluation of large language models for code translation.
\newblock \emph{CoRR}, abs/2403.17214, 2024.
\newblock \doi{10.48550/ARXIV.2403.17214}.
\newblock URL \url{https://doi.org/10.48550/arXiv.2403.17214}.

\bibitem[Madaan et~al.(2023)Madaan, Tandon, Gupta, Hallinan, Gao, Wiegreffe, Alon, Dziri, Prabhumoye, Yang, Gupta, Majumder, Hermann, Welleck, Yazdanbakhsh, and Clark]{2023Self-Refine}
Aman Madaan, Niket Tandon, Prakhar Gupta, Skyler Hallinan, Luyu Gao, Sarah Wiegreffe, Uri Alon, Nouha Dziri, Shrimai Prabhumoye, Yiming Yang, Shashank Gupta, Bodhisattwa~Prasad Majumder, Katherine Hermann, Sean Welleck, Amir Yazdanbakhsh, and Peter Clark.
\newblock Self-refine: Iterative refinement with self-feedback.
\newblock In Alice Oh, Tristan Naumann, Amir Globerson, Kate Saenko, Moritz Hardt, and Sergey Levine (eds.), \emph{Advances in Neural Information Processing Systems 36: Annual Conference on Neural Information Processing Systems 2023, NeurIPS 2023, New Orleans, LA, USA, December 10 - 16, 2023}, 2023.
\newblock URL \url{http://papers.nips.cc/paper\_files/paper/2023/hash/91edff07232fb1b55a505a9e9f6c0ff3-Abstract-Conference.html}.

\bibitem[Madeiral et~al.(2019)Madeiral, Urli, de~Almeida~Maia, and Monperrus]{fix-data-2019-3}
Fernanda Madeiral, Simon Urli, Marcelo de~Almeida~Maia, and Martin Monperrus.
\newblock {BEARS:} an extensible java bug benchmark for automatic program repair studies.
\newblock In Xinyu Wang, David Lo, and Emad Shihab (eds.), \emph{26th {IEEE} International Conference on Software Analysis, Evolution and Reengineering, {SANER} 2019, Hangzhou, China, February 24-27, 2019}, pp.\  468--478. {IEEE}, 2019.
\newblock \doi{10.1109/SANER.2019.8667991}.
\newblock URL \url{https://doi.org/10.1109/SANER.2019.8667991}.

\bibitem[Mahbub et~al.(2022)Mahbub, Oishie, and Haque]{author2022-1}
Parvez Mahbub, Naz~Zarreen Oishie, and S.~M.~Rafizul Haque.
\newblock Authorship identification of source code segments written by multiple authors using stacking ensemble method.
\newblock \emph{CoRR}, abs/2212.05610, 2022.
\newblock \doi{10.48550/arXiv.2212.05610}.
\newblock URL \url{https://doi.org/10.48550/arXiv.2212.05610}.

\bibitem[Mai et~al.(2024)Mai, Gao, Hu, Bao, Liu, and Sun]{api2024-1}
Yubo Mai, Zhipeng Gao, Xing Hu, Lingfeng Bao, Yu~Liu, and Jianling Sun.
\newblock Are human rules necessary? generating reusable apis with cot reasoning and in-context learning.
\newblock 2024.
\newblock URL \url{https://doi.org/10.48550/arXiv.2405.03509}.

\bibitem[Malik et~al.(2019)Malik, Patra, and Pradel]{type2019-2}
Rabee~Sohail Malik, Jibesh Patra, and Michael Pradel.
\newblock Nl2type: inferring javascript function types from natural language information.
\newblock In Joanne~M. Atlee, Tevfik Bultan, and Jon Whittle (eds.), \emph{Proceedings of the 41st International Conference on Software Engineering, {ICSE} 2019, Montreal, QC, Canada, May 25-31, 2019}, pp.\  304--315. {IEEE} / {ACM}, 2019.
\newblock \doi{10.1109/ICSE.2019.00045}.
\newblock URL \url{https://doi.org/10.1109/ICSE.2019.00045}.

\bibitem[Malyala et~al.(2023)Malyala, Zhou, Ray, and Chakraborty]{trans-survey-2023-2}
Aniketh Malyala, Katelyn Zhou, Baishakhi Ray, and Saikat Chakraborty.
\newblock On ml-based program translation: Perils and promises.
\newblock In \emph{45th {IEEE/ACM} International Conference on Software Engineering: New Ideas and Emerging Results, NIER@ICSE, Melbourne, Australia, May 14-20, 2023}, pp.\  60--65. {IEEE}, 2023.
\newblock \doi{10.1109/ICSE-NIER58687.2023.00017}.
\newblock URL \url{https://doi.org/10.1109/ICSE-NIER58687.2023.00017}.

\bibitem[Mao et~al.(2024)Mao, Li, Li, and Tei]{2024MuCoLD}
Zhenyu Mao, Jialong Li, Munan Li, and Kenji Tei.
\newblock Multi-role consensus through llms discussions for vulnerability detection.
\newblock \emph{CoRR}, abs/2403.14274, 2024.
\newblock \doi{10.48550/ARXIV.2403.14274}.
\newblock URL \url{https://doi.org/10.48550/arXiv.2403.14274}.

\bibitem[Marcus et~al.(2019)Marcus, Negi, Mao, Zhang, Alizadeh, Kraska, Papaemmanouil, and Tatbul]{SQL-optimizer1}
Ryan Marcus, Parimarjan Negi, Hongzi Mao, Chi Zhang, Mohammad Alizadeh, Tim Kraska, Olga Papaemmanouil, and Nesime Tatbul.
\newblock Neo: {A} learned query optimizer.
\newblock \emph{Proc. {VLDB} Endow.}, 12\penalty0 (11):\penalty0 1705--1718, 2019.
\newblock \doi{10.14778/3342263.3342644}.
\newblock URL \url{http://www.vldb.org/pvldb/vol12/p1705-marcus.pdf}.

\bibitem[Marcus et~al.(2020)Marcus, Negi, Mao, Tatbul, Alizadeh, and Kraska]{SQL-optimizer2}
Ryan Marcus, Parimarjan Negi, Hongzi Mao, Nesime Tatbul, Mohammad Alizadeh, and Tim Kraska.
\newblock Bao: Learning to steer query optimizers.
\newblock \emph{CoRR}, abs/2004.03814, 2020.
\newblock URL \url{https://arxiv.org/abs/2004.03814}.

\bibitem[Mastropaolo et~al.(2021)Mastropaolo, Scalabrino, Cooper, Nader{-}Palacio, Poshyvanyk, Oliveto, and Bavota]{2021T5Code}
Antonio Mastropaolo, Simone Scalabrino, Nathan Cooper, David Nader{-}Palacio, Denys Poshyvanyk, Rocco Oliveto, and Gabriele Bavota.
\newblock Studying the usage of text-to-text transfer transformer to support code-related tasks.
\newblock In \emph{43rd {IEEE/ACM} International Conference on Software Engineering, {ICSE} 2021, Madrid, Spain, 22-30 May 2021}, pp.\  336--347. {IEEE}, 2021.
\newblock \doi{10.1109/ICSE43902.2021.00041}.
\newblock URL \url{https://doi.org/10.1109/ICSE43902.2021.00041}.

\bibitem[Mathew \& Stolee(2021)Mathew and Stolee]{search2021-1}
George Mathew and Kathryn~T. Stolee.
\newblock Cross-language code search using static and dynamic analyses.
\newblock In Diomidis Spinellis, Georgios Gousios, Marsha Chechik, and Massimiliano~Di Penta (eds.), \emph{{ESEC/FSE} '21: 29th {ACM} Joint European Software Engineering Conference and Symposium on the Foundations of Software Engineering, Athens, Greece, August 23-28, 2021}, pp.\  205--217. {ACM}, 2021.
\newblock \doi{10.1145/3468264.3468538}.
\newblock URL \url{https://doi.org/10.1145/3468264.3468538}.

\bibitem[Mehta et~al.(2023)Mehta, Rawool, Gujar, and Xu]{workflow2023-1}
Deep Mehta, Kartik Rawool, Subodh Gujar, and Bowen Xu.
\newblock Automated devops pipeline generation for code repositories using large language models.
\newblock \emph{CoRR}, abs/2312.13225, 2023.
\newblock \doi{10.48550/ARXIV.2312.13225}.
\newblock URL \url{https://doi.org/10.48550/arXiv.2312.13225}.

\bibitem[Meng et~al.(2019)Meng, Liu, Zhu, Zhang, Pei, Liu, Chen, Zhang, Tao, Sun, and Zhou]{log2019-1}
Weibin Meng, Ying Liu, Yichen Zhu, Shenglin Zhang, Dan Pei, Yuqing Liu, Yihao Chen, Ruizhi Zhang, Shimin Tao, Pei Sun, and Rong Zhou.
\newblock Loganomaly: Unsupervised detection of sequential and quantitative anomalies in unstructured logs.
\newblock In Sarit Kraus (ed.), \emph{Proceedings of the Twenty-Eighth International Joint Conference on Artificial Intelligence, {IJCAI} 2019, Macao, China, August 10-16, 2019}, pp.\  4739--4745. ijcai.org, 2019.
\newblock \doi{10.24963/IJCAI.2019/658}.
\newblock URL \url{https://doi.org/10.24963/ijcai.2019/658}.

\bibitem[Menon et~al.(2013)Menon, Tamuz, Gulwani, Lampson, and Kalai]{syn2013-1}
Aditya~Krishna Menon, Omer Tamuz, Sumit Gulwani, Butler~W. Lampson, and Adam Kalai.
\newblock A machine learning framework for programming by example.
\newblock In \emph{Proceedings of the 30th International Conference on Machine Learning, {ICML} 2013, Atlanta, GA, USA, 16-21 June 2013}, volume~28 of \emph{{JMLR} Workshop and Conference Proceedings}, pp.\  187--195. JMLR.org, 2013.
\newblock URL \url{http://proceedings.mlr.press/v28/menon13.html}.

\bibitem[Mesnard et~al.(2024)Mesnard, Hardin, Dadashi, Bhupatiraju, Pathak, Sifre, Rivi{\`{e}}re, Kale, Love, Tafti, Hussenot, Chowdhery, Roberts, Barua, Botev, Castro{-}Ros, Slone, H{\'{e}}liou, Tacchetti, Bulanova, Paterson, Tsai, Shahriari, Lan, Choquette{-}Choo, Crepy, Cer, Ippolito, Reid, Buchatskaya, Ni, Noland, Yan, Tucker, Muraru, Rozhdestvenskiy, Michalewski, Tenney, Grishchenko, Austin, Keeling, Labanowski, Lespiau, Stanway, Brennan, Chen, Ferret, Chiu, and et~al.]{2024Gemma}
Thomas Mesnard, Cassidy Hardin, Robert Dadashi, Surya Bhupatiraju, Shreya Pathak, Laurent Sifre, Morgane Rivi{\`{e}}re, Mihir~Sanjay Kale, Juliette Love, Pouya Tafti, L{\'{e}}onard Hussenot, Aakanksha Chowdhery, Adam Roberts, Aditya Barua, Alex Botev, Alex Castro{-}Ros, Ambrose Slone, Am{\'{e}}lie H{\'{e}}liou, Andrea Tacchetti, Anna Bulanova, Antonia Paterson, Beth Tsai, Bobak Shahriari, Charline~Le Lan, Christopher~A. Choquette{-}Choo, Cl{\'{e}}ment Crepy, Daniel Cer, Daphne Ippolito, David Reid, Elena Buchatskaya, Eric Ni, Eric Noland, Geng Yan, George Tucker, George{-}Christian Muraru, Grigory Rozhdestvenskiy, Henryk Michalewski, Ian Tenney, Ivan Grishchenko, Jacob Austin, James Keeling, Jane Labanowski, Jean{-}Baptiste Lespiau, Jeff Stanway, Jenny Brennan, Jeremy Chen, Johan Ferret, Justin Chiu, and et~al.
\newblock Gemma: Open models based on gemini research and technology.
\newblock \emph{CoRR}, abs/2403.08295, 2024.
\newblock \doi{10.48550/ARXIV.2403.08295}.
\newblock URL \url{https://doi.org/10.48550/arXiv.2403.08295}.

\bibitem[Mezghani et~al.(2018)Mezghani, Kang, and S{\`{e}}des]{re-ana2018-2}
Manel Mezghani, Juyeon Kang, and Florence S{\`{e}}des.
\newblock Industrial requirements classification for redundancy and inconsistency detection in {SEMIOS}.
\newblock In Guenther Ruhe, Walid Maalej, and Daniel Amyot (eds.), \emph{26th {IEEE} International Requirements Engineering Conference, {RE} 2018, Banff, AB, Canada, August 20-24, 2018}, pp.\  297--303. {IEEE} Computer Society, 2018.
\newblock \doi{10.1109/RE.2018.00037}.
\newblock URL \url{https://doi.org/10.1109/RE.2018.00037}.

\bibitem[Micikevicius et~al.(2018)Micikevicius, Narang, Alben, Diamos, Elsen, Garc{\'{\i}}a, Ginsburg, Houston, Kuchaiev, Venkatesh, and Wu]{2017mixed_precision}
Paulius Micikevicius, Sharan Narang, Jonah Alben, Gregory~F. Diamos, Erich Elsen, David Garc{\'{\i}}a, Boris Ginsburg, Michael Houston, Oleksii Kuchaiev, Ganesh Venkatesh, and Hao Wu.
\newblock Mixed precision training.
\newblock In \emph{6th International Conference on Learning Representations, {ICLR} 2018, Vancouver, BC, Canada, April 30 - May 3, 2018, Conference Track Proceedings}. OpenReview.net, 2018.
\newblock URL \url{https://openreview.net/forum?id=r1gs9JgRZ}.

\bibitem[Mir et~al.(2021)Mir, Latoskinas, and Gousios]{type-data-2021-1}
Amir~M. Mir, Evaldas Latoskinas, and Georgios Gousios.
\newblock Manytypes4py: {A} benchmark python dataset for machine learning-based type inference.
\newblock In \emph{18th {IEEE/ACM} International Conference on Mining Software Repositories, {MSR} 2021, Madrid, Spain, May 17-19, 2021}, pp.\  585--589. {IEEE}, 2021.
\newblock \doi{10.1109/MSR52588.2021.00079}.
\newblock URL \url{https://doi.org/10.1109/MSR52588.2021.00079}.

\bibitem[Mir et~al.(2022)Mir, Latoskinas, Proksch, and Gousios]{type2021-1}
Amir~M. Mir, Evaldas Latoskinas, Sebastian Proksch, and Georgios Gousios.
\newblock Type4py: Practical deep similarity learning-based type inference for python.
\newblock In \emph{44th {IEEE/ACM} 44th International Conference on Software Engineering, {ICSE} 2022, Pittsburgh, PA, USA, May 25-27, 2022}, pp.\  2241--2252. {ACM}, 2022.
\newblock \doi{10.1145/3510003.3510124}.
\newblock URL \url{https://doi.org/10.1145/3510003.3510124}.

\bibitem[Molina et~al.(2022)Molina, d'Amorim, and Aguirre]{fuzz2022-2}
Facundo Molina, Marcelo d'Amorim, and Nazareno Aguirre.
\newblock Fuzzing class specifications.
\newblock In \emph{44th {IEEE/ACM} 44th International Conference on Software Engineering, {ICSE} 2022, Pittsburgh, PA, USA, May 25-27, 2022}, pp.\  1008--1020. {ACM}, 2022.
\newblock \doi{10.1145/3510003.3510120}.
\newblock URL \url{https://doi.org/10.1145/3510003.3510120}.

\bibitem[Monperrus(2018)]{fix-survey-2018}
Martin Monperrus.
\newblock Automatic software repair: {A} bibliography.
\newblock \emph{{ACM} Comput. Surv.}, 51\penalty0 (1):\penalty0 17:1--17:24, 2018.
\newblock \doi{10.1145/3105906}.
\newblock URL \url{https://doi.org/10.1145/3105906}.

\bibitem[Monperrus et~al.(2021)Monperrus, Martinez, Ye, Madeiral, Durieux, and Yu]{fix-data-2021-1}
Martin Monperrus, Matias Martinez, He~Ye, Fernanda Madeiral, Thomas Durieux, and Zhongxing Yu.
\newblock Megadiff: {A} dataset of 600k java source code changes categorized by diff size.
\newblock \emph{CoRR}, abs/2108.04631, 2021.
\newblock URL \url{https://arxiv.org/abs/2108.04631}.

\bibitem[Mou et~al.(2016)Mou, Li, Zhang, Wang, and Jin]{clf2014-1}
Lili Mou, Ge~Li, Lu~Zhang, Tao Wang, and Zhi Jin.
\newblock Convolutional neural networks over tree structures for programming language processing.
\newblock In Dale Schuurmans and Michael~P. Wellman (eds.), \emph{Proceedings of the Thirtieth {AAAI} Conference on Artificial Intelligence, February 12-17, 2016, Phoenix, Arizona, {USA}}, pp.\  1287--1293. {AAAI} Press, 2016.
\newblock \doi{10.1609/aaai.v30i1.10139}.
\newblock URL \url{https://doi.org/10.1609/aaai.v30i1.10139}.

\bibitem[Mu et~al.(2023{\natexlab{a}})Mu, Chen, Shi, Wang, and Wang]{comment2023-1}
Fangwen Mu, Xiao Chen, Lin Shi, Song Wang, and Qing Wang.
\newblock Developer-intent driven code comment generation.
\newblock In \emph{45th {IEEE/ACM} International Conference on Software Engineering, {ICSE} 2023, Melbourne, Australia, May 14-20, 2023}, pp.\  768--780. {IEEE}, 2023{\natexlab{a}}.
\newblock \doi{10.1109/ICSE48619.2023.00073}.
\newblock URL \url{https://doi.org/10.1109/ICSE48619.2023.00073}.

\bibitem[Mu et~al.(2023{\natexlab{b}})Mu, Shi, Wang, Yu, Zhang, Wang, Liu, and Wang]{2023ClarifyGPT}
Fangwen Mu, Lin Shi, Song Wang, Zhuohao Yu, Binquan Zhang, Chenxue Wang, Shichao Liu, and Qing Wang.
\newblock Clarifygpt: Empowering llm-based code generation with intention clarification.
\newblock \emph{CoRR}, abs/2310.10996, 2023{\natexlab{b}}.
\newblock \doi{10.48550/ARXIV.2310.10996}.
\newblock URL \url{https://doi.org/10.48550/arXiv.2310.10996}.

\bibitem[Mudgal \& Wouhaybi(2023)Mudgal and Wouhaybi]{log2023-5.5}
Priyanka Mudgal and Rita~H. Wouhaybi.
\newblock An assessment of chatgpt on log data.
\newblock \emph{CoRR}, abs/2309.07938, 2023.
\newblock \doi{10.48550/ARXIV.2309.07938}.
\newblock URL \url{https://doi.org/10.48550/arXiv.2309.07938}.

\bibitem[Muennighoff et~al.(2024)Muennighoff, Liu, Zebaze, Zheng, Hui, Zhuo, Singh, Tang, Werra, and Longpre]{2023OctoPack}
Niklas Muennighoff, Qian Liu, Armel~Randy Zebaze, Qinkai Zheng, Binyuan Hui, Terry~Yue Zhuo, Swayam Singh, Xiangru Tang, Leandro~Von Werra, and Shayne Longpre.
\newblock Octopack: Instruction tuning code large language models.
\newblock In \emph{The Twelfth International Conference on Learning Representations}, 2024.
\newblock URL \url{https://openreview.net/forum?id=mw1PWNSWZP}.

\bibitem[Murad et~al.(2010)Murad, Shirazi, Zikria, and Ikram]{ob2009-1}
Khurram Murad, Syed~Noor{-}ul{-}Hassan Shirazi, Yousaf~Bin Zikria, and Nassar Ikram.
\newblock Evading virus detection using code obfuscation.
\newblock In Tai{-}Hoon Kim, Young{-}Hoon Lee, Byeong~Ho Kang, and Dominik Slezak (eds.), \emph{Future Generation Information Technology - Second International Conference, {FGIT} 2010, Jeju Island, Korea, December 13-15, 2010. Proceedings}, volume 6485 of \emph{Lecture Notes in Computer Science}, pp.\  394--401. Springer, 2010.
\newblock \doi{10.1007/978-3-642-17569-5\_39}.
\newblock URL \url{https://doi.org/10.1007/978-3-642-17569-5\_39}.

\bibitem[Nadimi \& Zheng(2024)Nadimi and Zheng]{2024MEV-LLM}
Bardia Nadimi and Hao Zheng.
\newblock A multi-expert large language model architecture for verilog code generation.
\newblock \emph{CoRR}, abs/2404.08029, 2024.
\newblock \doi{10.48550/ARXIV.2404.08029}.
\newblock URL \url{https://doi.org/10.48550/arXiv.2404.08029}.

\bibitem[Nafi et~al.(2019)Nafi, Kar, Roy, Roy, and Schneider]{clone-data-2019-1}
Kawser~Wazed Nafi, Tonny~Shekha Kar, Banani Roy, Chanchal~K. Roy, and Kevin~A. Schneider.
\newblock {CLCDSA:} cross language code clone detection using syntactical features and {API} documentation.
\newblock In \emph{34th {IEEE/ACM} International Conference on Automated Software Engineering, {ASE} 2019, San Diego, CA, USA, November 11-15, 2019}, pp.\  1026--1037. {IEEE}, 2019.
\newblock \doi{10.1109/ASE.2019.00099}.
\newblock URL \url{https://doi.org/10.1109/ASE.2019.00099}.

\bibitem[Nakagawa et~al.(2021)Nakagawa, Higo, and Kusumoto]{clone2021-1}
Tasuku Nakagawa, Yoshiki Higo, and Shinji Kusumoto.
\newblock {NIL:} large-scale detection of large-variance clones.
\newblock In Diomidis Spinellis, Georgios Gousios, Marsha Chechik, and Massimiliano~Di Penta (eds.), \emph{{ESEC/FSE} '21: 29th {ACM} Joint European Software Engineering Conference and Symposium on the Foundations of Software Engineering, Athens, Greece, August 23-28, 2021}, pp.\  830--841. {ACM}, 2021.
\newblock \doi{10.1145/3468264.3468564}.
\newblock URL \url{https://doi.org/10.1145/3468264.3468564}.

\bibitem[Nan et~al.(2023)Nan, Zhao, Zou, Ri, Tae, Zhang, Cohan, and Radev]{sql2023-1}
Linyong Nan, Yilun Zhao, Weijin Zou, Narutatsu Ri, Jaesung Tae, Ellen Zhang, Arman Cohan, and Dragomir Radev.
\newblock Enhancing few-shot text-to-sql capabilities of large language models: {A} study on prompt design strategies.
\newblock \emph{CoRR}, abs/2305.12586, 2023.
\newblock \doi{10.48550/arXiv.2305.12586}.
\newblock URL \url{https://doi.org/10.48550/arXiv.2305.12586}.

\bibitem[Nedelkoski et~al.(2020{\natexlab{a}})Nedelkoski, Bogatinovski, Acker, Cardoso, and Kao]{log2020-1.5}
Sasho Nedelkoski, Jasmin Bogatinovski, Alexander Acker, Jorge Cardoso, and Odej Kao.
\newblock Self-supervised log parsing.
\newblock In Yuxiao Dong, Dunja Mladenic, and Craig Saunders (eds.), \emph{Machine Learning and Knowledge Discovery in Databases: Applied Data Science Track - European Conference, {ECML} {PKDD} 2020, Ghent, Belgium, September 14-18, 2020, Proceedings, Part {IV}}, volume 12460 of \emph{Lecture Notes in Computer Science}, pp.\  122--138. Springer, 2020{\natexlab{a}}.
\newblock \doi{10.1007/978-3-030-67667-4\_8}.
\newblock URL \url{https://doi.org/10.1007/978-3-030-67667-4\_8}.

\bibitem[Nedelkoski et~al.(2020{\natexlab{b}})Nedelkoski, Bogatinovski, Acker, Cardoso, and Kao]{log2020-2}
Sasho Nedelkoski, Jasmin Bogatinovski, Alexander Acker, Jorge Cardoso, and Odej Kao.
\newblock Self-attentive classification-based anomaly detection in unstructured logs.
\newblock In Claudia Plant, Haixun Wang, Alfredo Cuzzocrea, Carlo Zaniolo, and Xindong Wu (eds.), \emph{20th {IEEE} International Conference on Data Mining, {ICDM} 2020, Sorrento, Italy, November 17-20, 2020}, pp.\  1196--1201. {IEEE}, 2020{\natexlab{b}}.
\newblock \doi{10.1109/ICDM50108.2020.00148}.
\newblock URL \url{https://doi.org/10.1109/ICDM50108.2020.00148}.

\bibitem[Nguyen et~al.(2013)Nguyen, Nguyen, and Nguyen]{trans2013-1}
Anh~Tuan Nguyen, Tung~Thanh Nguyen, and Tien~N. Nguyen.
\newblock Lexical statistical machine translation for language migration.
\newblock In Bertrand Meyer, Luciano Baresi, and Mira Mezini (eds.), \emph{Joint Meeting of the European Software Engineering Conference and the {ACM} {SIGSOFT} Symposium on the Foundations of Software Engineering, ESEC/FSE'13, Saint Petersburg, Russian Federation, August 18-26, 2013}, pp.\  651--654. {ACM}, 2013.
\newblock \doi{10.1145/2491411.2494584}.
\newblock URL \url{https://doi.org/10.1145/2491411.2494584}.

\bibitem[Nguyen et~al.(2015)Nguyen, Nguyen, and Nguyen]{trans2015-1}
Anh~Tuan Nguyen, Tung~Thanh Nguyen, and Tien~N. Nguyen.
\newblock Divide-and-conquer approach for multi-phase statistical migration for source code {(T)}.
\newblock In Myra~B. Cohen, Lars Grunske, and Michael Whalen (eds.), \emph{30th {IEEE/ACM} International Conference on Automated Software Engineering, {ASE} 2015, Lincoln, NE, USA, November 9-13, 2015}, pp.\  585--596. {IEEE} Computer Society, 2015.
\newblock \doi{10.1109/ASE.2015.74}.
\newblock URL \url{https://doi.org/10.1109/ASE.2015.74}.

\bibitem[Nguyen et~al.(2019)Nguyen, Nguyen, Dig, Nguyen, Tran, and Hilton]{fix-data-2019-6}
Hoan~Anh Nguyen, Tien~N. Nguyen, Danny Dig, Son Nguyen, Hieu Tran, and Michael Hilton.
\newblock Graph-based mining of in-the-wild, fine-grained, semantic code change patterns.
\newblock In Joanne~M. Atlee, Tevfik Bultan, and Jon Whittle (eds.), \emph{Proceedings of the 41st International Conference on Software Engineering, {ICSE} 2019, Montreal, QC, Canada, May 25-31, 2019}, pp.\  819--830. {IEEE} / {ACM}, 2019.
\newblock \doi{10.1109/ICSE.2019.00089}.
\newblock URL \url{https://doi.org/10.1109/ICSE.2019.00089}.

\bibitem[Nguyen \& Nadi(2022)Nguyen and Nadi]{analysis2022-2}
Nhan Nguyen and Sarah Nadi.
\newblock An empirical evaluation of github copilot's code suggestions.
\newblock In \emph{19th {IEEE/ACM} International Conference on Mining Software Repositories, {MSR} 2022, Pittsburgh, PA, USA, May 23-24, 2022}, pp.\  1--5. {ACM}, 2022.
\newblock \doi{10.1145/3524842.3528470}.
\newblock URL \url{https://doi.org/10.1145/3524842.3528470}.

\bibitem[Nguyen et~al.(2020)Nguyen, Phan, Le, and Nguyen]{id2020-1}
Son Nguyen, Hung Phan, Trinh Le, and Tien~N. Nguyen.
\newblock Suggesting natural method names to check name consistencies.
\newblock In Gregg Rothermel and Doo{-}Hwan Bae (eds.), \emph{{ICSE} '20: 42nd International Conference on Software Engineering, Seoul, South Korea, 27 June - 19 July, 2020}, pp.\  1372--1384. {ACM}, 2020.
\newblock \doi{10.1145/3377811.3380926}.
\newblock URL \url{https://doi.org/10.1145/3377811.3380926}.

\bibitem[Nguyen et~al.(2021)Nguyen, Kobayashi, and Fukuda]{log2021-0.5}
Thieu Nguyen, Satoru Kobayashi, and Kensuke Fukuda.
\newblock Logdtl: Network log template generation with deep transfer learning.
\newblock In Toufik Ahmed, Olivier Festor, Yacine Ghamri{-}Doudane, Joon{-}Myung Kang, Alberto E.~Schaeffer Filho, Abdelkader Lahmadi, and Edmundo R.~M. Madeira (eds.), \emph{17th {IFIP/IEEE} International Symposium on Integrated Network Management, {IM} 2021, Bordeaux, France, May 17-21, 2021}, pp.\  848--853. {IEEE}, 2021.
\newblock URL \url{https://ieeexplore.ieee.org/document/9464068}.

\bibitem[Nguyen et~al.(2016)Nguyen, Nguyen, and Nguyen]{api2016-1}
Trong~Duc Nguyen, Anh~Tuan Nguyen, and Tien~N. Nguyen.
\newblock Mapping {API} elements for code migration with vector representations.
\newblock In Laura~K. Dillon, Willem Visser, and Laurie~A. Williams (eds.), \emph{Proceedings of the 38th International Conference on Software Engineering, {ICSE} 2016, Austin, TX, USA, May 14-22, 2016 - Companion Volume}, pp.\  756--758. {ACM}, 2016.
\newblock \doi{10.1145/2889160.2892661}.
\newblock URL \url{https://doi.org/10.1145/2889160.2892661}.

\bibitem[Nguyen et~al.(2022)Nguyen, Nguyen, Nguyen, Le, Tran, and Phung]{defect2021-2}
Van{-}Anh Nguyen, Dai~Quoc Nguyen, Van Nguyen, Trung Le, Quan~Hung Tran, and Dinh Phung.
\newblock Regvd: Revisiting graph neural networks for vulnerability detection.
\newblock In \emph{44th {IEEE/ACM} International Conference on Software Engineering: Companion Proceedings, {ICSE} Companion 2022, Pittsburgh, PA, USA, May 22-24, 2022}, pp.\  178--182. {ACM/IEEE}, 2022.
\newblock \doi{10.1145/3510454.3516865}.
\newblock URL \url{https://doi.org/10.1145/3510454.3516865}.

\bibitem[Nichols et~al.(2024)Nichols, Davis, Xie, Rajaram, and Bhatele]{2024ParEval}
Daniel Nichols, Joshua~Hoke Davis, Zhaojun Xie, Arjun Rajaram, and Abhinav Bhatele.
\newblock Can large language models write parallel code?
\newblock \emph{CoRR}, abs/2401.12554, 2024.
\newblock \doi{10.48550/ARXIV.2401.12554}.
\newblock URL \url{https://doi.org/10.48550/arXiv.2401.12554}.

\bibitem[Nie et~al.(2021)Nie, Gao, Zhong, Lam, Liu, and Xu]{commit2020-2}
Lun~Yiu Nie, Cuiyun Gao, Zhicong Zhong, Wai Lam, Yang Liu, and Zenglin Xu.
\newblock Coregen: Contextualized code representation learning for commit message generation.
\newblock \emph{Neurocomputing}, 459:\penalty0 97--107, 2021.
\newblock \doi{10.1016/J.NEUCOM.2021.05.039}.
\newblock URL \url{https://doi.org/10.1016/j.neucom.2021.05.039}.

\bibitem[Nie et~al.(2023)Nie, Banerjee, Li, Mooney, and Gligoric]{unit2023-3}
Pengyu Nie, Rahul Banerjee, Junyi~Jessy Li, Raymond~J. Mooney, and Milos Gligoric.
\newblock Learning deep semantics for test completion.
\newblock In \emph{45th {IEEE/ACM} International Conference on Software Engineering, {ICSE} 2023, Melbourne, Australia, May 14-20, 2023}, pp.\  2111--2123. {IEEE}, 2023.
\newblock \doi{10.1109/ICSE48619.2023.00178}.
\newblock URL \url{https://doi.org/10.1109/ICSE48619.2023.00178}.

\bibitem[Nijkamp et~al.(2023{\natexlab{a}})Nijkamp, Hayashi, Xiong, Savarese, and Zhou]{2023CodeGen2}
Erik Nijkamp, Hiroaki Hayashi, Caiming Xiong, Silvio Savarese, and Yingbo Zhou.
\newblock Codegen2: Lessons for training llms on programming and natural languages.
\newblock \emph{CoRR}, abs/2305.02309, 2023{\natexlab{a}}.
\newblock \doi{10.48550/ARXIV.2305.02309}.
\newblock URL \url{https://doi.org/10.48550/arXiv.2305.02309}.

\bibitem[Nijkamp et~al.(2023{\natexlab{b}})Nijkamp, Pang, Hayashi, Tu, Wang, Zhou, Savarese, and Xiong]{2022CodeGen}
Erik Nijkamp, Bo~Pang, Hiroaki Hayashi, Lifu Tu, Huan Wang, Yingbo Zhou, Silvio Savarese, and Caiming Xiong.
\newblock Codegen: An open large language model for code with multi-turn program synthesis.
\newblock In \emph{The Eleventh International Conference on Learning Representations, {ICLR} 2023, Kigali, Rwanda, May 1-5, 2023}. OpenReview.net, 2023{\natexlab{b}}.
\newblock URL \url{https://openreview.net/pdf?id=iaYcJKpY2B\_}.

\bibitem[Nikitopoulos et~al.(2021)Nikitopoulos, Dritsa, Louridas, and Mitropoulos]{defect-data-2021-3}
Georgios Nikitopoulos, Konstantina Dritsa, Panos Louridas, and Dimitris Mitropoulos.
\newblock Crossvul: a cross-language vulnerability dataset with commit data.
\newblock In Diomidis Spinellis, Georgios Gousios, Marsha Chechik, and Massimiliano~Di Penta (eds.), \emph{{ESEC/FSE} '21: 29th {ACM} Joint European Software Engineering Conference and Symposium on the Foundations of Software Engineering, Athens, Greece, August 23-28, 2021}, pp.\  1565--1569. {ACM}, 2021.
\newblock \doi{10.1145/3468264.3473122}.
\newblock URL \url{https://doi.org/10.1145/3468264.3473122}.

\bibitem[Nitin et~al.(2021)Nitin, Saieva, Ray, and Kaiser]{ob2021-2}
Vikram Nitin, Anthony Saieva, Baishakhi Ray, and Gail Kaiser.
\newblock {DIRECT} : A transformer-based model for decompiled identifier renaming.
\newblock In Royi Lachmy, Ziyu Yao, Greg Durrett, Milos Gligoric, Junyi~Jessy Li, Ray Mooney, Graham Neubig, Yu~Su, Huan Sun, and Reut Tsarfaty (eds.), \emph{Proceedings of the 1st Workshop on Natural Language Processing for Programming (NLP4Prog 2021)}, pp.\  48--57, Online, August 2021. Association for Computational Linguistics.
\newblock \doi{10.18653/v1/2021.nlp4prog-1.6}.
\newblock URL \url{https://aclanthology.org/2021.nlp4prog-1.6}.

\bibitem[Niu et~al.(2022)Niu, Li, Ng, Ge, Huang, and Luo]{2022SPT-Code}
Changan Niu, Chuanyi Li, Vincent Ng, Jidong Ge, Liguo Huang, and Bin Luo.
\newblock Spt-code: Sequence-to-sequence pre-training for learning source code representations.
\newblock In \emph{44th {IEEE/ACM} 44th International Conference on Software Engineering, {ICSE} 2022, Pittsburgh, PA, USA, May 25-27, 2022}, pp.\  1--13. {ACM}, 2022.
\newblock \doi{10.1145/3510003.3510096}.
\newblock URL \url{https://doi.org/10.1145/3510003.3510096}.

\bibitem[Niu et~al.(2023)Niu, Li, Ng, Chen, Ge, and Luo]{2023survey}
Changan Niu, Chuanyi Li, Vincent Ng, Dongxiao Chen, Jidong Ge, and Bin Luo.
\newblock An empirical comparison of pre-trained models of source code.
\newblock In \emph{45th {IEEE/ACM} International Conference on Software Engineering, {ICSE} 2023, Melbourne, Australia, May 14-20, 2023}, pp.\  2136--2148. {IEEE}, 2023.
\newblock \doi{10.1109/ICSE48619.2023.00180}.
\newblock URL \url{https://doi.org/10.1109/ICSE48619.2023.00180}.

\bibitem[Niu et~al.(2024)Niu, Zhang, Li, Luo, and Ng]{analysis2024-5}
Changan Niu, Ting Zhang, Chuanyi Li, Bin Luo, and Vincent Ng.
\newblock On evaluating the efficiency of source code generated by llms.
\newblock \emph{CoRR}, abs/2404.06041, 2024.
\newblock \doi{10.48550/ARXIV.2404.06041}.
\newblock URL \url{https://doi.org/10.48550/arXiv.2404.06041}.

\bibitem[Nong et~al.(2023)Nong, Sharma, Hamou{-}Lhadj, Luo, and Cai]{defect-survey-2022-1}
Yu~Nong, Rainy Sharma, Abdelwahab Hamou{-}Lhadj, Xiapu Luo, and Haipeng Cai.
\newblock Open science in software engineering: {A} study on deep learning-based vulnerability detection.
\newblock \emph{{IEEE} Trans. Software Eng.}, 49\penalty0 (4):\penalty0 1983--2005, 2023.
\newblock \doi{10.1109/TSE.2022.3207149}.
\newblock URL \url{https://doi.org/10.1109/TSE.2022.3207149}.

\bibitem[Nye et~al.(2021)Nye, Andreassen, Gur{-}Ari, Michalewski, Austin, Bieber, Dohan, Lewkowycz, Bosma, Luan, Sutton, and Odena]{2021Scratchpad}
Maxwell~I. Nye, Anders~Johan Andreassen, Guy Gur{-}Ari, Henryk Michalewski, Jacob Austin, David Bieber, David Dohan, Aitor Lewkowycz, Maarten Bosma, David Luan, Charles Sutton, and Augustus Odena.
\newblock Show your work: Scratchpads for intermediate computation with language models.
\newblock \emph{CoRR}, abs/2112.00114, 2021.
\newblock URL \url{https://arxiv.org/abs/2112.00114}.

\bibitem[Odena et~al.(2019)Odena, Olsson, Andersen, and Goodfellow]{fuzz2018-3}
Augustus Odena, Catherine Olsson, David~G. Andersen, and Ian~J. Goodfellow.
\newblock Tensorfuzz: Debugging neural networks with coverage-guided fuzzing.
\newblock In Kamalika Chaudhuri and Ruslan Salakhutdinov (eds.), \emph{Proceedings of the 36th International Conference on Machine Learning, {ICML} 2019, 9-15 June 2019, Long Beach, California, {USA}}, volume~97 of \emph{Proceedings of Machine Learning Research}, pp.\  4901--4911. {PMLR}, 2019.
\newblock URL \url{http://proceedings.mlr.press/v97/odena19a.html}.

\bibitem[Oh \& Yoo(2024)Oh and Yoo]{2024CSA-Trans}
Saeyoon Oh and Shin Yoo.
\newblock Csa-trans: Code structure aware transformer for ast.
\newblock \emph{CoRR}, abs/2404.05767, 2024.
\newblock \doi{10.48550/ARXIV.2404.05767}.
\newblock URL \url{https://doi.org/10.48550/arXiv.2404.05767}.

\bibitem[Oh \& Oh(2022)Oh and Oh]{fix-data-2022-1}
Wonseok Oh and Hakjoo Oh.
\newblock Pyter: effective program repair for python type errors.
\newblock In Abhik Roychoudhury, Cristian Cadar, and Miryung Kim (eds.), \emph{Proceedings of the 30th {ACM} Joint European Software Engineering Conference and Symposium on the Foundations of Software Engineering, {ESEC/FSE} 2022, Singapore, Singapore, November 14-18, 2022}, pp.\  922--934. {ACM}, 2022.
\newblock \doi{10.1145/3540250.3549130}.
\newblock URL \url{https://doi.org/10.1145/3540250.3549130}.

\bibitem[Ojdanic et~al.(2021)Ojdanic, Garg, Khanfir, Degiovanni, Papadakis, and Traon]{mutant2021-2}
Milos Ojdanic, Aayush Garg, Ahmed Khanfir, Renzo Degiovanni, Mike Papadakis, and Yves~Le Traon.
\newblock Syntactic vs. semantic similarity of artificial and real faults in mutation testing studies.
\newblock \emph{CoRR}, abs/2112.14508, 2021.
\newblock URL \url{https://arxiv.org/abs/2112.14508}.

\bibitem[Ojdanic et~al.(2023)Ojdanic, Khanfir, Garg, Degiovanni, Papadakis, and Traon]{mutant2023-3}
Milos Ojdanic, Ahmed Khanfir, Aayush Garg, Renzo Degiovanni, Mike Papadakis, and Yves~Le Traon.
\newblock On comparing mutation testing tools through learning-based mutant selection.
\newblock In \emph{{IEEE/ACM} International Conference on Automation of Software Test, {AST} 2023, Melbourne, Australia, May 15-16, 2023}, pp.\  35--46. {IEEE}, 2023.
\newblock \doi{10.1109/AST58925.2023.00008}.
\newblock URL \url{https://doi.org/10.1109/AST58925.2023.00008}.

\bibitem[OpenAI(2023)]{2023GPT4}
OpenAI.
\newblock {GPT-4} technical report.
\newblock \emph{CoRR}, abs/2303.08774, 2023.
\newblock \doi{10.48550/arXiv.2303.08774}.
\newblock URL \url{https://doi.org/10.48550/arXiv.2303.08774}.

\bibitem[Ouyang et~al.(2022)Ouyang, Wu, Jiang, Almeida, Wainwright, Mishkin, Zhang, Agarwal, Slama, Ray, Schulman, Hilton, Kelton, Miller, Simens, Askell, Welinder, Christiano, Leike, and Lowe]{2022InstructGPT}
Long Ouyang, Jeffrey Wu, Xu~Jiang, Diogo Almeida, Carroll~L. Wainwright, Pamela Mishkin, Chong Zhang, Sandhini Agarwal, Katarina Slama, Alex Ray, John Schulman, Jacob Hilton, Fraser Kelton, Luke Miller, Maddie Simens, Amanda Askell, Peter Welinder, Paul~F. Christiano, Jan Leike, and Ryan Lowe.
\newblock Training language models to follow instructions with human feedback.
\newblock In \emph{NeurIPS}, 2022.
\newblock URL \url{http://papers.nips.cc/paper\_files/paper/2022/hash/b1efde53be364a73914f58805a001731-Abstract-Conference.html}.

\bibitem[Pan et~al.(2023)Pan, Ibrahimzada, Krishna, Sankar, Wassi, Merler, Sobolev, Pavuluri, Sinha, and Jabbarvand]{trans2023-3}
Rangeet Pan, Ali~Reza Ibrahimzada, Rahul Krishna, Divya Sankar, Lambert~Pouguem Wassi, Michele Merler, Boris Sobolev, Raju Pavuluri, Saurabh Sinha, and Reyhaneh Jabbarvand.
\newblock Understanding the effectiveness of large language models in code translation.
\newblock \emph{CoRR}, abs/2308.03109, 2023.
\newblock \doi{10.48550/ARXIV.2308.03109}.
\newblock URL \url{https://doi.org/10.48550/arXiv.2308.03109}.

\bibitem[Pan et~al.(2024)Pan, Liu, Zou, Xie, and Xie]{comment2024-1}
Xinglu Pan, Chenxiao Liu, Yanzhen Zou, Tao Xie, and Bing Xie.
\newblock {MESIA:} understanding and leveraging supplementary nature of method-level comments for automatic comment generation.
\newblock \emph{CoRR}, abs/2403.17357, 2024.
\newblock \doi{10.48550/ARXIV.2403.17357}.
\newblock URL \url{https://doi.org/10.48550/arXiv.2403.17357}.

\bibitem[Pandi et~al.(2020)Pandi, Barr, Gordon, and Sutton]{type2020-1}
Irene~Vlassi Pandi, Earl~T. Barr, Andrew~D. Gordon, and Charles Sutton.
\newblock Opttyper: Probabilistic type inference by optimising logical and natural constraints.
\newblock \emph{CoRR}, abs/2004.00348, 2020.
\newblock URL \url{https://arxiv.org/abs/2004.00348}.

\bibitem[Panichella et~al.(2018)Panichella, Kifetew, and Tonella]{unit2017-1}
Annibale Panichella, Fitsum~Meshesha Kifetew, and Paolo Tonella.
\newblock Automated test case generation as a many-objective optimisation problem with dynamic selection of the targets.
\newblock \emph{{IEEE} Trans. Software Eng.}, 44\penalty0 (2):\penalty0 122--158, 2018.
\newblock \doi{10.1109/TSE.2017.2663435}.
\newblock URL \url{https://doi.org/10.1109/TSE.2017.2663435}.

\bibitem[Papineni et~al.(2002)Papineni, Roukos, Ward, and Zhu]{2002BLEU}
Kishore Papineni, Salim Roukos, Todd Ward, and Wei{-}Jing Zhu.
\newblock Bleu: a method for automatic evaluation of machine translation.
\newblock In \emph{Proceedings of the 40th Annual Meeting of the Association for Computational Linguistics, July 6-12, 2002, Philadelphia, PA, {USA}}, pp.\  311--318. {ACL}, 2002.
\newblock \doi{10.3115/1073083.1073135}.
\newblock URL \url{https://aclanthology.org/P02-1040/}.

\bibitem[Parasaram et~al.(2024)Parasaram, Yan, Yang, Flahy, Qudsi, Ziaber, Barr, and Mechtaev]{2024Maniple}
Nikhil Parasaram, Huijie Yan, Boyu Yang, Zineb Flahy, Abriele Qudsi, Damian Ziaber, Earl Barr, and Sergey Mechtaev.
\newblock The fact selection problem in llm-based program repair.
\newblock \emph{CoRR}, abs/2404.05520, 2024.
\newblock \doi{10.48550/ARXIV.2404.05520}.
\newblock URL \url{https://doi.org/10.48550/arXiv.2404.05520}.

\bibitem[Parisotto et~al.(2017)Parisotto, Mohamed, Singh, Li, Zhou, and Kohli]{syn2016-2}
Emilio Parisotto, Abdel{-}rahman Mohamed, Rishabh Singh, Lihong Li, Dengyong Zhou, and Pushmeet Kohli.
\newblock Neuro-symbolic program synthesis.
\newblock In \emph{5th International Conference on Learning Representations, {ICLR} 2017, Toulon, France, April 24-26, 2017, Conference Track Proceedings}. OpenReview.net, 2017.
\newblock URL \url{https://openreview.net/forum?id=rJ0JwFcex}.

\bibitem[Parvez et~al.(2018)Parvez, Chakraborty, Ray, and Chang]{completion2018-1}
Md.~Rizwan Parvez, Saikat Chakraborty, Baishakhi Ray, and Kai{-}Wei Chang.
\newblock Building language models for text with named entities.
\newblock In Iryna Gurevych and Yusuke Miyao (eds.), \emph{Proceedings of the 56th Annual Meeting of the Association for Computational Linguistics, {ACL} 2018, Melbourne, Australia, July 15-20, 2018, Volume 1: Long Papers}, pp.\  2373--2383. Association for Computational Linguistics, 2018.
\newblock \doi{10.18653/v1/P18-1221}.
\newblock URL \url{https://aclanthology.org/P18-1221/}.

\bibitem[Parvez et~al.(2021)Parvez, Ahmad, Chakraborty, Ray, and Chang]{syn2021-1}
Md.~Rizwan Parvez, Wasi~Uddin Ahmad, Saikat Chakraborty, Baishakhi Ray, and Kai{-}Wei Chang.
\newblock Retrieval augmented code generation and summarization.
\newblock In Marie{-}Francine Moens, Xuanjing Huang, Lucia Specia, and Scott~Wen{-}tau Yih (eds.), \emph{Findings of the Association for Computational Linguistics: {EMNLP} 2021, Virtual Event / Punta Cana, Dominican Republic, 16-20 November, 2021}, pp.\  2719--2734. Association for Computational Linguistics, 2021.
\newblock \doi{10.18653/V1/2021.FINDINGS-EMNLP.232}.
\newblock URL \url{https://doi.org/10.18653/v1/2021.findings-emnlp.232}.

\bibitem[Patra \& Pradel(2021)Patra and Pradel]{mutant2021-1}
Jibesh Patra and Michael Pradel.
\newblock Semantic bug seeding: a learning-based approach for creating realistic bugs.
\newblock In Diomidis Spinellis, Georgios Gousios, Marsha Chechik, and Massimiliano~Di Penta (eds.), \emph{{ESEC/FSE} '21: 29th {ACM} Joint European Software Engineering Conference and Symposium on the Foundations of Software Engineering, Athens, Greece, August 23-28, 2021}, pp.\  906--918. {ACM}, 2021.
\newblock \doi{10.1145/3468264.3468623}.
\newblock URL \url{https://doi.org/10.1145/3468264.3468623}.

\bibitem[Paul et~al.(2023)Paul, Hossain, Hasan, and Iqbal]{fix2023-1}
Rishov Paul, Md.~Mohib Hossain, Masum Hasan, and Anindya Iqbal.
\newblock Automated program repair based on code review: How do pre-trained transformer models perform?
\newblock \emph{CoRR}, abs/2304.07840, 2023.
\newblock \doi{10.48550/ARXIV.2304.07840}.
\newblock URL \url{https://doi.org/10.48550/arXiv.2304.07840}.

\bibitem[Pearce et~al.(2022)Pearce, Ahmad, Tan, Dolan{-}Gavitt, and Karri]{defect2021-1}
Hammond Pearce, Baleegh Ahmad, Benjamin Tan, Brendan Dolan{-}Gavitt, and Ramesh Karri.
\newblock Asleep at the keyboard? assessing the security of github copilot's code contributions.
\newblock In \emph{43rd {IEEE} Symposium on Security and Privacy, {SP} 2022, San Francisco, CA, USA, May 22-26, 2022}, pp.\  754--768. {IEEE}, 2022.
\newblock \doi{10.1109/SP46214.2022.9833571}.
\newblock URL \url{https://doi.org/10.1109/SP46214.2022.9833571}.

\bibitem[Penedo et~al.(2023)Penedo, Malartic, Hesslow, Cojocaru, Cappelli, Alobeidli, Pannier, Almazrouei, and Launay]{2023RefinedWeb}
Guilherme Penedo, Quentin Malartic, Daniel Hesslow, Ruxandra Cojocaru, Alessandro Cappelli, Hamza Alobeidli, Baptiste Pannier, Ebtesam Almazrouei, and Julien Launay.
\newblock The refinedweb dataset for falcon {LLM:} outperforming curated corpora with web data, and web data only.
\newblock \emph{CoRR}, abs/2306.01116, 2023.
\newblock \doi{10.48550/arXiv.2306.01116}.
\newblock URL \url{https://doi.org/10.48550/arXiv.2306.01116}.

\bibitem[Peng et~al.(2023{\natexlab{a}})Peng, Quesnelle, Fan, and Shippole]{2023YaRN}
Bowen Peng, Jeffrey Quesnelle, Honglu Fan, and Enrico Shippole.
\newblock Yarn: Efficient context window extension of large language models.
\newblock \emph{CoRR}, abs/2309.00071, 2023{\natexlab{a}}.
\newblock \doi{10.48550/ARXIV.2309.00071}.
\newblock URL \url{https://doi.org/10.48550/arXiv.2309.00071}.

\bibitem[Peng et~al.(2022)Peng, Gao, Li, Gao, Lo, Zhang, and Lyu]{type2021-2}
Yun Peng, Cuiyun Gao, Zongjie Li, Bowei Gao, David Lo, Qirun Zhang, and Michael~R. Lyu.
\newblock Static inference meets deep learning: {A} hybrid type inference approach for python.
\newblock In \emph{44th {IEEE/ACM} 44th International Conference on Software Engineering, {ICSE} 2022, Pittsburgh, PA, USA, May 25-27, 2022}, pp.\  2019--2030. {ACM}, 2022.
\newblock \doi{10.1145/3510003.3510038}.
\newblock URL \url{https://doi.org/10.1145/3510003.3510038}.

\bibitem[Peng et~al.(2023{\natexlab{b}})Peng, Gao, Gao, Huo, and Lyu]{type2023-3.5}
Yun Peng, Shuzheng Gao, Cuiyun Gao, Yintong Huo, and Michael~R. Lyu.
\newblock Domain knowledge matters: Improving prompts with fix templates for repairing python type errors.
\newblock \emph{CoRR}, abs/2306.01394, 2023{\natexlab{b}}.
\newblock \doi{10.48550/ARXIV.2306.01394}.
\newblock URL \url{https://doi.org/10.48550/arXiv.2306.01394}.

\bibitem[Peng et~al.(2023{\natexlab{c}})Peng, Wang, Wang, Gao, and Lyu]{type2023-4}
Yun Peng, Chaozheng Wang, Wenxuan Wang, Cuiyun Gao, and Michael~R. Lyu.
\newblock Generative type inference for python.
\newblock \emph{CoRR}, abs/2307.09163, 2023{\natexlab{c}}.
\newblock \doi{10.48550/ARXIV.2307.09163}.
\newblock URL \url{https://doi.org/10.48550/arXiv.2307.09163}.

\bibitem[Perez \& Chiba(2019)Perez and Chiba]{clone2019-2.5}
Daniel Perez and Shigeru Chiba.
\newblock Cross-language clone detection by learning over abstract syntax trees.
\newblock In Margaret{-}Anne~D. Storey, Bram Adams, and Sonia Haiduc (eds.), \emph{Proceedings of the 16th International Conference on Mining Software Repositories, {MSR} 2019, 26-27 May 2019, Montreal, Canada}, pp.\  518--528. {IEEE} / {ACM}, 2019.
\newblock \doi{10.1109/MSR.2019.00078}.
\newblock URL \url{https://doi.org/10.1109/MSR.2019.00078}.

\bibitem[Perry et~al.(2023)Perry, Srivastava, Kumar, and Boneh]{analysis2022-4}
Neil Perry, Megha Srivastava, Deepak Kumar, and Dan Boneh.
\newblock Do users write more insecure code with {AI} assistants?
\newblock In Weizhi Meng, Christian~Damsgaard Jensen, Cas Cremers, and Engin Kirda (eds.), \emph{Proceedings of the 2023 {ACM} {SIGSAC} Conference on Computer and Communications Security, {CCS} 2023, Copenhagen, Denmark, November 26-30, 2023}, pp.\  2785--2799. {ACM}, 2023.
\newblock \doi{10.1145/3576915.3623157}.
\newblock URL \url{https://doi.org/10.1145/3576915.3623157}.

\bibitem[Pewny \& Holz(2016)Pewny and Holz]{mutant2016-4}
Jannik Pewny and Thorsten Holz.
\newblock Evilcoder: automated bug insertion.
\newblock In Stephen Schwab, William~K. Robertson, and Davide Balzarotti (eds.), \emph{Proceedings of the 32nd Annual Conference on Computer Security Applications, {ACSAC} 2016, Los Angeles, CA, USA, December 5-9, 2016}, pp.\  214--225. {ACM}, 2016.
\newblock URL \url{http://dl.acm.org/citation.cfm?id=2991103}.

\bibitem[Phan et~al.(2017)Phan, Nguyen, Nguyen, and Nguyen]{api2017-2}
Hung~Dang Phan, Anh~Tuan Nguyen, Trong~Duc Nguyen, and Tien~N. Nguyen.
\newblock Statistical migration of {API} usages.
\newblock In Sebasti{\'{a}}n Uchitel, Alessandro Orso, and Martin~P. Robillard (eds.), \emph{Proceedings of the 39th International Conference on Software Engineering, {ICSE} 2017, Buenos Aires, Argentina, May 20-28, 2017 - Companion Volume}, pp.\  47--50. {IEEE} Computer Society, 2017.
\newblock \doi{10.1109/ICSE-C.2017.17}.
\newblock URL \url{https://doi.org/10.1109/ICSE-C.2017.17}.

\bibitem[Phan et~al.(2024)Phan, Phan, Nguyen, and Bui]{2024RepoHyper}
Huy~N. Phan, Hoang~N. Phan, Tien~N. Nguyen, and Nghi D.~Q. Bui.
\newblock Repohyper: Better context retrieval is all you need for repository-level code completion.
\newblock \emph{CoRR}, abs/2403.06095, 2024.
\newblock \doi{10.48550/ARXIV.2403.06095}.
\newblock URL \url{https://doi.org/10.48550/arXiv.2403.06095}.

\bibitem[Pizzorno \& Berger(2024)Pizzorno and Berger]{2024CoverUp}
Juan~Altmayer Pizzorno and Emery~D. Berger.
\newblock Coverup: Coverage-guided llm-based test generation.
\newblock \emph{CoRR}, abs/2403.16218, 2024.
\newblock \doi{10.48550/ARXIV.2403.16218}.
\newblock URL \url{https://doi.org/10.48550/arXiv.2403.16218}.

\bibitem[Ponta et~al.(2019)Ponta, Plate, Sabetta, Bezzi, and Dangremont]{defect-data-2019-1}
Serena~Elisa Ponta, Henrik Plate, Antonino Sabetta, Michele Bezzi, and C{\'{e}}dric Dangremont.
\newblock A manually-curated dataset of fixes to vulnerabilities of open-source software.
\newblock In Margaret{-}Anne~D. Storey, Bram Adams, and Sonia Haiduc (eds.), \emph{Proceedings of the 16th International Conference on Mining Software Repositories, {MSR} 2019, 26-27 May 2019, Montreal, Canada}, pp.\  383--387. {IEEE} / {ACM}, 2019.
\newblock \doi{10.1109/MSR.2019.00064}.
\newblock URL \url{https://doi.org/10.1109/MSR.2019.00064}.

\bibitem[Pourreza \& Rafiei(2023)Pourreza and Rafiei]{sql2023-0.4}
Mohammadreza Pourreza and Davood Rafiei.
\newblock {DIN-SQL:} decomposed in-context learning of text-to-sql with self-correction.
\newblock \emph{CoRR}, abs/2304.11015, 2023.
\newblock \doi{10.48550/ARXIV.2304.11015}.
\newblock URL \url{https://doi.org/10.48550/arXiv.2304.11015}.

\bibitem[Pradel \& Sen(2018)Pradel and Sen]{defect2018-2}
Michael Pradel and Koushik Sen.
\newblock Deepbugs: a learning approach to name-based bug detection.
\newblock \emph{Proc. {ACM} Program. Lang.}, 2\penalty0 ({OOPSLA}):\penalty0 147:1--147:25, 2018.
\newblock \doi{10.1145/3276517}.
\newblock URL \url{https://doi.org/10.1145/3276517}.

\bibitem[Pradel et~al.(2015)Pradel, Schuh, and Sen]{type2015-1}
Michael Pradel, Parker Schuh, and Koushik Sen.
\newblock Typedevil: Dynamic type inconsistency analysis for javascript.
\newblock In Antonia Bertolino, Gerardo Canfora, and Sebastian~G. Elbaum (eds.), \emph{37th {IEEE/ACM} International Conference on Software Engineering, {ICSE} 2015, Florence, Italy, May 16-24, 2015, Volume 1}, pp.\  314--324. {IEEE} Computer Society, 2015.
\newblock \doi{10.1109/ICSE.2015.51}.
\newblock URL \url{https://doi.org/10.1109/ICSE.2015.51}.

\bibitem[Pradel et~al.(2020)Pradel, Gousios, Liu, and Chandra]{type2019-4}
Michael Pradel, Georgios Gousios, Jason Liu, and Satish Chandra.
\newblock Typewriter: neural type prediction with search-based validation.
\newblock In Prem Devanbu, Myra~B. Cohen, and Thomas Zimmermann (eds.), \emph{{ESEC/FSE} '20: 28th {ACM} Joint European Software Engineering Conference and Symposium on the Foundations of Software Engineering, Virtual Event, USA, November 8-13, 2020}, pp.\  209--220. {ACM}, 2020.
\newblock \doi{10.1145/3368089.3409715}.
\newblock URL \url{https://doi.org/10.1145/3368089.3409715}.

\bibitem[Prenner \& Robbes(2023)Prenner and Robbes]{fix-data-2023-1}
Julian~Aron Prenner and Romain Robbes.
\newblock Runbugrun - an executable dataset for automated program repair.
\newblock \emph{CoRR}, abs/2304.01102, 2023.
\newblock \doi{10.48550/ARXIV.2304.01102}.
\newblock URL \url{https://doi.org/10.48550/arXiv.2304.01102}.

\bibitem[Press et~al.(2022)Press, Smith, and Lewis]{2021ALiBi}
Ofir Press, Noah~A. Smith, and Mike Lewis.
\newblock Train short, test long: Attention with linear biases enables input length extrapolation.
\newblock In \emph{The Tenth International Conference on Learning Representations, {ICLR} 2022, Virtual Event, April 25-29, 2022}. OpenReview.net, 2022.
\newblock URL \url{https://openreview.net/forum?id=R8sQPpGCv0}.

\bibitem[Pu et~al.(2016)Pu, Narasimhan, Solar{-}Lezama, and Barzilay]{fix2016-3}
Yewen Pu, Karthik Narasimhan, Armando Solar{-}Lezama, and Regina Barzilay.
\newblock sk{\_}p: a neural program corrector for moocs.
\newblock In Eelco Visser (ed.), \emph{Companion Proceedings of the 2016 {ACM} {SIGPLAN} International Conference on Systems, Programming, Languages and Applications: Software for Humanity, {SPLASH} 2016, Amsterdam, Netherlands, October 30 - November 4, 2016}, pp.\  39--40. {ACM}, 2016.
\newblock \doi{10.1145/2984043.2989222}.
\newblock URL \url{https://doi.org/10.1145/2984043.2989222}.

\bibitem[Puri et~al.(2021)Puri, Kung, Janssen, Zhang, Domeniconi, Zolotov, Dolby, Chen, Choudhury, Decker, Thost, Buratti, Pujar, Ramji, Finkler, Malaika, and Reiss]{2021CodeNet}
Ruchir Puri, David~S. Kung, Geert Janssen, Wei Zhang, Giacomo Domeniconi, Vladimir Zolotov, Julian Dolby, Jie Chen, Mihir~R. Choudhury, Lindsey Decker, Veronika Thost, Luca Buratti, Saurabh Pujar, Shyam Ramji, Ulrich Finkler, Susan Malaika, and Frederick Reiss.
\newblock Codenet: {A} large-scale {AI} for code dataset for learning a diversity of coding tasks.
\newblock In Joaquin Vanschoren and Sai{-}Kit Yeung (eds.), \emph{Proceedings of the Neural Information Processing Systems Track on Datasets and Benchmarks 1, NeurIPS Datasets and Benchmarks 2021, December 2021, virtual}, 2021.
\newblock URL \url{https://datasets-benchmarks-proceedings.neurips.cc/paper/2021/hash/a5bfc9e07964f8dddeb95fc584cd965d-Abstract-round2.html}.

\bibitem[Qi et~al.(2023)Qi, Huang, Luan, Fung, Yang, and Qian]{log2023-5}
Jiaxing Qi, Shaohan Huang, Zhongzhi Luan, Carol~J. Fung, Hailong Yang, and Depei Qian.
\newblock Loggpt: Exploring chatgpt for log-based anomaly detection.
\newblock \emph{CoRR}, abs/2309.01189, 2023.
\newblock \doi{10.48550/ARXIV.2309.01189}.
\newblock URL \url{https://doi.org/10.48550/arXiv.2309.01189}.

\bibitem[Qian et~al.(2023)Qian, Cong, Yang, Chen, Su, Xu, Liu, and Sun]{2023ChatDev}
Chen Qian, Xin Cong, Cheng Yang, Weize Chen, Yusheng Su, Juyuan Xu, Zhiyuan Liu, and Maosong Sun.
\newblock Communicative agents for software development.
\newblock \emph{CoRR}, abs/2307.07924, 2023.
\newblock \doi{10.48550/ARXIV.2307.07924}.
\newblock URL \url{https://doi.org/10.48550/arXiv.2307.07924}.

\bibitem[Qin et~al.(2022{\natexlab{a}})Qin, Hui, Wang, Yang, Li, Li, Geng, Cao, Sun, Si, Huang, and Li]{sql-survey-2022-3}
Bowen Qin, Binyuan Hui, Lihan Wang, Min Yang, Jinyang Li, Binhua Li, Ruiying Geng, Rongyu Cao, Jian Sun, Luo Si, Fei Huang, and Yongbin Li.
\newblock A survey on text-to-sql parsing: Concepts, methods, and future directions.
\newblock \emph{CoRR}, abs/2208.13629, 2022{\natexlab{a}}.
\newblock \doi{10.48550/ARXIV.2208.13629}.
\newblock URL \url{https://doi.org/10.48550/arXiv.2208.13629}.

\bibitem[Qin et~al.(2022{\natexlab{b}})Qin, Sun, Deng, Li, Wei, Lv, Yan, Kong, and Zhong]{2022cosformer}
Zhen Qin, Weixuan Sun, Hui Deng, Dongxu Li, Yunshen Wei, Baohong Lv, Junjie Yan, Lingpeng Kong, and Yiran Zhong.
\newblock cosformer: Rethinking softmax in attention.
\newblock In \emph{The Tenth International Conference on Learning Representations, {ICLR} 2022, Virtual Event, April 25-29, 2022}. OpenReview.net, 2022{\natexlab{b}}.
\newblock URL \url{https://openreview.net/forum?id=Bl8CQrx2Up4}.

\bibitem[Rabinovich et~al.(2017)Rabinovich, Stern, and Klein]{syn2017-3}
Maxim Rabinovich, Mitchell Stern, and Dan Klein.
\newblock Abstract syntax networks for code generation and semantic parsing.
\newblock In Regina Barzilay and Min{-}Yen Kan (eds.), \emph{Proceedings of the 55th Annual Meeting of the Association for Computational Linguistics, {ACL} 2017, Vancouver, Canada, July 30 - August 4, Volume 1: Long Papers}, pp.\  1139--1149. Association for Computational Linguistics, 2017.
\newblock \doi{10.18653/V1/P17-1105}.
\newblock URL \url{https://doi.org/10.18653/v1/P17-1105}.

\bibitem[Radford et~al.(2018)Radford, Narasimhan, Salimans, and Sutskever]{2018GPT}
Alec Radford, Karthik Narasimhan, Tim Salimans, and Ilya Sutskever.
\newblock Improving language understanding by generative pre-training, 2018.

\bibitem[Radford et~al.(2019)Radford, Wu, Child, Luan, Amodei, Sutskever, et~al.]{2019GPT2}
Alec Radford, Jeffrey Wu, Rewon Child, David Luan, Dario Amodei, Ilya Sutskever, et~al.
\newblock Language models are unsupervised multitask learners.
\newblock \emph{OpenAI blog}, 1\penalty0 (8):\penalty0 9, 2019.

\bibitem[Rae et~al.(2021)Rae, Borgeaud, Cai, Millican, Hoffmann, Song, Aslanides, Henderson, Ring, Young, Rutherford, Hennigan, Menick, Cassirer, Powell, van~den Driessche, Hendricks, Rauh, Huang, Glaese, Welbl, Dathathri, Huang, Uesato, Mellor, Higgins, Creswell, McAleese, Wu, Elsen, Jayakumar, Buchatskaya, Budden, Sutherland, Simonyan, Paganini, Sifre, Martens, Li, Kuncoro, Nematzadeh, Gribovskaya, Donato, Lazaridou, Mensch, Lespiau, Tsimpoukelli, Grigorev, Fritz, Sottiaux, Pajarskas, Pohlen, Gong, Toyama, de~Masson~d'Autume, Li, Terzi, Mikulik, Babuschkin, Clark, de~Las~Casas, Guy, Jones, Bradbury, Johnson, Hechtman, Weidinger, Gabriel, Isaac, Lockhart, Osindero, Rimell, Dyer, Vinyals, Ayoub, Stanway, Bennett, Hassabis, Kavukcuoglu, and Irving]{2021Gopher}
Jack~W. Rae, Sebastian Borgeaud, Trevor Cai, Katie Millican, Jordan Hoffmann, H.~Francis Song, John Aslanides, Sarah Henderson, Roman Ring, Susannah Young, Eliza Rutherford, Tom Hennigan, Jacob Menick, Albin Cassirer, Richard Powell, George van~den Driessche, Lisa~Anne Hendricks, Maribeth Rauh, Po{-}Sen Huang, Amelia Glaese, Johannes Welbl, Sumanth Dathathri, Saffron Huang, Jonathan Uesato, John Mellor, Irina Higgins, Antonia Creswell, Nat McAleese, Amy Wu, Erich Elsen, Siddhant~M. Jayakumar, Elena Buchatskaya, David Budden, Esme Sutherland, Karen Simonyan, Michela Paganini, Laurent Sifre, Lena Martens, Xiang~Lorraine Li, Adhiguna Kuncoro, Aida Nematzadeh, Elena Gribovskaya, Domenic Donato, Angeliki Lazaridou, Arthur Mensch, Jean{-}Baptiste Lespiau, Maria Tsimpoukelli, Nikolai Grigorev, Doug Fritz, Thibault Sottiaux, Mantas Pajarskas, Toby Pohlen, Zhitao Gong, Daniel Toyama, Cyprien de~Masson~d'Autume, Yujia Li, Tayfun Terzi, Vladimir Mikulik, Igor Babuschkin, Aidan Clark, Diego de~Las~Casas, Aurelia Guy,
  Chris Jones, James Bradbury, Matthew~J. Johnson, Blake~A. Hechtman, Laura Weidinger, Iason Gabriel, William~S. Isaac, Edward Lockhart, Simon Osindero, Laura Rimell, Chris Dyer, Oriol Vinyals, Kareem Ayoub, Jeff Stanway, Lorrayne Bennett, Demis Hassabis, Koray Kavukcuoglu, and Geoffrey Irving.
\newblock Scaling language models: Methods, analysis {\&} insights from training gopher.
\newblock \emph{CoRR}, abs/2112.11446, 2021.
\newblock URL \url{https://arxiv.org/abs/2112.11446}.

\bibitem[Raffel et~al.(2020)Raffel, Shazeer, Roberts, Lee, Narang, Matena, Zhou, Li, and Liu]{2019T5}
Colin Raffel, Noam Shazeer, Adam Roberts, Katherine Lee, Sharan Narang, Michael Matena, Yanqi Zhou, Wei Li, and Peter~J. Liu.
\newblock Exploring the limits of transfer learning with a unified text-to-text transformer.
\newblock \emph{J. Mach. Learn. Res.}, 21:\penalty0 140:1--140:67, 2020.
\newblock URL \url{http://jmlr.org/papers/v21/20-074.html}.

\bibitem[Rahali \& Akhloufi(2023)Rahali and Akhloufi]{2023MalBERTv2}
Abir Rahali and Moulay~A. Akhloufi.
\newblock Malbertv2: Code aware bert-based model for malware identification.
\newblock \emph{Big Data Cogn. Comput.}, 7\penalty0 (2):\penalty0 60, 2023.
\newblock \doi{10.3390/BDCC7020060}.
\newblock URL \url{https://doi.org/10.3390/bdcc7020060}.

\bibitem[Rahman et~al.(2023)Rahman, Ceka, Mao, Chakraborty, Ray, and Le]{defect2023-3}
Md~Mahbubur Rahman, Ira Ceka, Chengzhi Mao, Saikat Chakraborty, Baishakhi Ray, and Wei Le.
\newblock Towards causal deep learning for vulnerability detection.
\newblock \emph{CoRR}, abs/2310.07958, 2023.
\newblock \doi{10.48550/ARXIV.2310.07958}.
\newblock URL \url{https://doi.org/10.48550/arXiv.2310.07958}.

\bibitem[Ray et~al.(2016)Ray, Hellendoorn, Godhane, Tu, Bacchelli, and Devanbu]{defect2015-1}
Baishakhi Ray, Vincent~J. Hellendoorn, Saheel Godhane, Zhaopeng Tu, Alberto Bacchelli, and Premkumar~T. Devanbu.
\newblock On the "naturalness" of buggy code.
\newblock In Laura~K. Dillon, Willem Visser, and Laurie~A. Williams (eds.), \emph{Proceedings of the 38th International Conference on Software Engineering, {ICSE} 2016, Austin, TX, USA, May 14-22, 2016}, pp.\  428--439. {ACM}, 2016.
\newblock \doi{10.1145/2884781.2884848}.
\newblock URL \url{https://doi.org/10.1145/2884781.2884848}.

\bibitem[Raychev et~al.(2014)Raychev, Vechev, and Yahav]{completion2014-2}
Veselin Raychev, Martin~T. Vechev, and Eran Yahav.
\newblock Code completion with statistical language models.
\newblock In Michael F.~P. O'Boyle and Keshav Pingali (eds.), \emph{{ACM} {SIGPLAN} Conference on Programming Language Design and Implementation, {PLDI} '14, Edinburgh, United Kingdom - June 09 - 11, 2014}, pp.\  419--428. {ACM}, 2014.
\newblock \doi{10.1145/2594291.2594321}.
\newblock URL \url{https://doi.org/10.1145/2594291.2594321}.

\bibitem[Raychev et~al.(2015)Raychev, Vechev, and Krause]{id2015-1}
Veselin Raychev, Martin~T. Vechev, and Andreas Krause.
\newblock Predicting program properties from "big code".
\newblock In Sriram~K. Rajamani and David Walker (eds.), \emph{Proceedings of the 42nd Annual {ACM} {SIGPLAN-SIGACT} Symposium on Principles of Programming Languages, {POPL} 2015, Mumbai, India, January 15-17, 2015}, pp.\  111--124. {ACM}, 2015.
\newblock \doi{10.1145/2676726.2677009}.
\newblock URL \url{https://doi.org/10.1145/2676726.2677009}.

\bibitem[Raychev et~al.(2016{\natexlab{a}})Raychev, Bielik, and Vechev]{completion2016-3}
Veselin Raychev, Pavol Bielik, and Martin~T. Vechev.
\newblock Probabilistic model for code with decision trees.
\newblock In Eelco Visser and Yannis Smaragdakis (eds.), \emph{Proceedings of the 2016 {ACM} {SIGPLAN} International Conference on Object-Oriented Programming, Systems, Languages, and Applications, {OOPSLA} 2016, part of {SPLASH} 2016, Amsterdam, The Netherlands, October 30 - November 4, 2016}, pp.\  731--747. {ACM}, 2016{\natexlab{a}}.
\newblock \doi{10.1145/2983990.2984041}.
\newblock URL \url{https://doi.org/10.1145/2983990.2984041}.

\bibitem[Raychev et~al.(2016{\natexlab{b}})Raychev, Bielik, Vechev, and Krause]{completion2016-1}
Veselin Raychev, Pavol Bielik, Martin~T. Vechev, and Andreas Krause.
\newblock Learning programs from noisy data.
\newblock In Rastislav Bod{\'{\i}}k and Rupak Majumdar (eds.), \emph{Proceedings of the 43rd Annual {ACM} {SIGPLAN-SIGACT} Symposium on Principles of Programming Languages, {POPL} 2016, St. Petersburg, FL, USA, January 20 - 22, 2016}, pp.\  761--774. {ACM}, 2016{\natexlab{b}}.
\newblock \doi{10.1145/2837614.2837671}.
\newblock URL \url{https://doi.org/10.1145/2837614.2837671}.

\bibitem[Ren et~al.(2020)Ren, Guo, Lu, Zhou, Liu, Tang, Sundaresan, Zhou, Blanco, and Ma]{2020CodeBLEU}
Shuo Ren, Daya Guo, Shuai Lu, Long Zhou, Shujie Liu, Duyu Tang, Neel Sundaresan, Ming Zhou, Ambrosio Blanco, and Shuai Ma.
\newblock Codebleu: a method for automatic evaluation of code synthesis.
\newblock \emph{CoRR}, abs/2009.10297, 2020.
\newblock URL \url{https://arxiv.org/abs/2009.10297}.

\bibitem[Richter \& Wehrheim(2022)Richter and Wehrheim]{fix-data-2022-3}
Cedric Richter and Heike Wehrheim.
\newblock {TSSB-3M:} mining single statement bugs at massive scale.
\newblock In \emph{19th {IEEE/ACM} International Conference on Mining Software Repositories, {MSR} 2022, Pittsburgh, PA, USA, May 23-24, 2022}, pp.\  418--422. {ACM}, 2022.
\newblock \doi{10.1145/3524842.3528505}.
\newblock URL \url{https://doi.org/10.1145/3524842.3528505}.

\bibitem[Risse \& B{\"{o}}hme(2023)Risse and B{\"{o}}hme]{defect-data-2023-2}
Niklas Risse and Marcel B{\"{o}}hme.
\newblock Limits of machine learning for automatic vulnerability detection.
\newblock \emph{CoRR}, abs/2306.17193, 2023.
\newblock \doi{10.48550/ARXIV.2306.17193}.
\newblock URL \url{https://doi.org/10.48550/arXiv.2306.17193}.

\bibitem[Robeer et~al.(2016)Robeer, Lucassen, van~der Werf, Dalpiaz, and Brinkkemper]{re-model2016-1}
Marcel Robeer, Garm Lucassen, Jan Martijn E.~M. van~der Werf, Fabiano Dalpiaz, and Sjaak Brinkkemper.
\newblock Automated extraction of conceptual models from user stories via {NLP}.
\newblock In \emph{24th {IEEE} International Requirements Engineering Conference, {RE} 2016, Beijing, China, September 12-16, 2016}, pp.\  196--205. {IEEE} Computer Society, 2016.
\newblock \doi{10.1109/RE.2016.40}.
\newblock URL \url{https://doi.org/10.1109/RE.2016.40}.

\bibitem[Robinson(2019)]{UI2019-2}
Alex Robinson.
\newblock Sketch2code: Generating a website from a paper mockup.
\newblock \emph{CoRR}, abs/1905.13750, 2019.
\newblock URL \url{http://arxiv.org/abs/1905.13750}.

\bibitem[Rodriguez et~al.(2023)Rodriguez, Dearstyne, and Cleland{-}Huang]{re-ana2023-2}
Alberto~D. Rodriguez, Katherine~R. Dearstyne, and Jane Cleland{-}Huang.
\newblock Prompts matter: Insights and strategies for prompt engineering in automated software traceability.
\newblock In Kurt Schneider, Fabiano Dalpiaz, and Jennifer Horkoff (eds.), \emph{31st {IEEE} International Requirements Engineering Conference, {RE} 2023 - Workshops, Hannover, Germany, September 4-5, 2023}, pp.\  455--464. {IEEE}, 2023.
\newblock \doi{10.1109/REW57809.2023.00087}.
\newblock URL \url{https://doi.org/10.1109/REW57809.2023.00087}.

\bibitem[Ronanki et~al.(2023)Ronanki, Daniel, Horkoff, and Berger]{re-ana2023-5}
Krishna Ronanki, Beatriz~Cabrero Daniel, Jennifer Horkoff, and Christian Berger.
\newblock Requirements engineering using generative {AI:} prompts and prompting patterns.
\newblock \emph{CoRR}, abs/2311.03832, 2023.
\newblock \doi{10.48550/ARXIV.2311.03832}.
\newblock URL \url{https://doi.org/10.48550/arXiv.2311.03832}.

\bibitem[Roy et~al.(2018)Roy, Pandey, Dolan{-}Gavitt, and Hu]{mutant2018-1}
Subhajit Roy, Awanish Pandey, Brendan Dolan{-}Gavitt, and Yu~Hu.
\newblock Bug synthesis: challenging bug-finding tools with deep faults.
\newblock In Gary~T. Leavens, Alessandro Garcia, and Corina~S. Pasareanu (eds.), \emph{Proceedings of the 2018 {ACM} Joint Meeting on European Software Engineering Conference and Symposium on the Foundations of Software Engineering, {ESEC/SIGSOFT} {FSE} 2018, Lake Buena Vista, FL, USA, November 04-09, 2018}, pp.\  224--234. {ACM}, 2018.
\newblock \doi{10.1145/3236024.3236084}.
\newblock URL \url{https://doi.org/10.1145/3236024.3236084}.

\bibitem[Rozi{\`{e}}re et~al.(2020)Rozi{\`{e}}re, Lachaux, Chanussot, and Lample]{2020Transcoder}
Baptiste Rozi{\`{e}}re, Marie{-}Anne Lachaux, Lowik Chanussot, and Guillaume Lample.
\newblock Unsupervised translation of programming languages.
\newblock In Hugo Larochelle, Marc'Aurelio Ranzato, Raia Hadsell, Maria{-}Florina Balcan, and Hsuan{-}Tien Lin (eds.), \emph{Advances in Neural Information Processing Systems 33: Annual Conference on Neural Information Processing Systems 2020, NeurIPS 2020, December 6-12, 2020, virtual}, 2020.
\newblock URL \url{https://proceedings.neurips.cc/paper/2020/hash/ed23fbf18c2cd35f8c7f8de44f85c08d-Abstract.html}.

\bibitem[Rozi{\`{e}}re et~al.(2022)Rozi{\`{e}}re, Zhang, Charton, Harman, Synnaeve, and Lample]{2021Transcoder}
Baptiste Rozi{\`{e}}re, Jie Zhang, Fran{\c{c}}ois Charton, Mark Harman, Gabriel Synnaeve, and Guillaume Lample.
\newblock Leveraging automated unit tests for unsupervised code translation.
\newblock In \emph{The Tenth International Conference on Learning Representations, {ICLR} 2022, Virtual Event, April 25-29, 2022}. OpenReview.net, 2022.
\newblock URL \url{https://openreview.net/forum?id=cmt-6KtR4c4}.

\bibitem[Rozi{\`{e}}re et~al.(2023)Rozi{\`{e}}re, Gehring, Gloeckle, Sootla, Gat, Tan, Adi, Liu, Remez, Rapin, Kozhevnikov, Evtimov, Bitton, Bhatt, Canton{-}Ferrer, Grattafiori, Xiong, D{\'{e}}fossez, Copet, Azhar, Touvron, Martin, Usunier, Scialom, and Synnaeve]{2023CodeLLaMA}
Baptiste Rozi{\`{e}}re, Jonas Gehring, Fabian Gloeckle, Sten Sootla, Itai Gat, Xiaoqing~Ellen Tan, Yossi Adi, Jingyu Liu, Tal Remez, J{\'{e}}r{\'{e}}my Rapin, Artyom Kozhevnikov, Ivan Evtimov, Joanna Bitton, Manish Bhatt, Cristian Canton{-}Ferrer, Aaron Grattafiori, Wenhan Xiong, Alexandre D{\'{e}}fossez, Jade Copet, Faisal Azhar, Hugo Touvron, Louis Martin, Nicolas Usunier, Thomas Scialom, and Gabriel Synnaeve.
\newblock Code llama: Open foundation models for code.
\newblock \emph{CoRR}, abs/2308.12950, 2023.
\newblock \doi{10.48550/arXiv.2308.12950}.
\newblock URL \url{https://doi.org/10.48550/arXiv.2308.12950}.

\bibitem[Ruan et~al.(2023)Ruan, Chen, Zhang, Xu, Bao, Du, Shi, Mao, Zeng, and Zhao]{2023TPTU}
Jingqing Ruan, Yihong Chen, Bin Zhang, Zhiwei Xu, Tianpeng Bao, Guoqing Du, Shiwei Shi, Hangyu Mao, Xingyu Zeng, and Rui Zhao.
\newblock {TPTU:} task planning and tool usage of large language model-based {AI} agents.
\newblock \emph{CoRR}, abs/2308.03427, 2023.
\newblock \doi{10.48550/ARXIV.2308.03427}.
\newblock URL \url{https://doi.org/10.48550/arXiv.2308.03427}.

\bibitem[Rubin \& Berant(2021)Rubin and Berant]{sql2020-3}
Ohad Rubin and Jonathan Berant.
\newblock Smbop: Semi-autoregressive bottom-up semantic parsing.
\newblock In Kristina Toutanova, Anna Rumshisky, Luke Zettlemoyer, Dilek Hakkani{-}T{\"{u}}r, Iz~Beltagy, Steven Bethard, Ryan Cotterell, Tanmoy Chakraborty, and Yichao Zhou (eds.), \emph{Proceedings of the 2021 Conference of the North American Chapter of the Association for Computational Linguistics: Human Language Technologies, {NAACL-HLT} 2021, Online, June 6-11, 2021}, pp.\  311--324. Association for Computational Linguistics, 2021.
\newblock \doi{10.18653/v1/2021.naacl-main.29}.
\newblock URL \url{https://doi.org/10.18653/v1/2021.naacl-main.29}.

\bibitem[Russell et~al.(2018)Russell, Kim, Hamilton, Lazovich, Harer, Ozdemir, Ellingwood, and McConley]{defect2018-3}
Rebecca~L. Russell, Louis~Y. Kim, Lei~H. Hamilton, Tomo Lazovich, Jacob Harer, Onur Ozdemir, Paul~M. Ellingwood, and Marc~W. McConley.
\newblock Automated vulnerability detection in source code using deep representation learning.
\newblock In M.~Arif Wani, Mehmed~M. Kantardzic, Moamar~Sayed Mouchaweh, Jo{\~{a}}o Gama, and Edwin Lughofer (eds.), \emph{17th {IEEE} International Conference on Machine Learning and Applications, {ICMLA} 2018, Orlando, FL, USA, December 17-20, 2018}, pp.\  757--762. {IEEE}, 2018.
\newblock \doi{10.1109/ICMLA.2018.00120}.
\newblock URL \url{https://doi.org/10.1109/ICMLA.2018.00120}.

\bibitem[Sachdev et~al.(2018)Sachdev, Li, Luan, Kim, Sen, and Chandra]{retrieval2018-3}
Saksham Sachdev, Hongyu Li, Sifei Luan, Seohyun Kim, Koushik Sen, and Satish Chandra.
\newblock Retrieval on source code: a neural code search.
\newblock In Justin Gottschlich and Alvin Cheung (eds.), \emph{Proceedings of the 2nd {ACM} {SIGPLAN} International Workshop on Machine Learning and Programming Languages, MAPL@PLDI 2018, Philadelphia, PA, USA, June 18-22, 2018}, pp.\  31--41. {ACM}, 2018.
\newblock \doi{10.1145/3211346.3211353}.
\newblock URL \url{https://doi.org/10.1145/3211346.3211353}.

\bibitem[Sadowski et~al.(2015)Sadowski, van Gogh, Jaspan, S{\"{o}}derberg, and Winter]{review2015-1}
Caitlin Sadowski, Jeffrey van Gogh, Ciera Jaspan, Emma S{\"{o}}derberg, and Collin Winter.
\newblock Tricorder: Building a program analysis ecosystem.
\newblock In Antonia Bertolino, Gerardo Canfora, and Sebastian~G. Elbaum (eds.), \emph{37th {IEEE/ACM} International Conference on Software Engineering, {ICSE} 2015, Florence, Italy, May 16-24, 2015, Volume 1}, pp.\  598--608. {IEEE} Computer Society, 2015.
\newblock \doi{10.1109/ICSE.2015.76}.
\newblock URL \url{https://doi.org/10.1109/ICSE.2015.76}.

\bibitem[Saha et~al.(2018)Saha, Lyu, Lam, Yoshida, and Prasad]{fix-data-2018-1}
Ripon~K. Saha, Yingjun Lyu, Wing Lam, Hiroaki Yoshida, and Mukul~R. Prasad.
\newblock Bugs.jar: a large-scale, diverse dataset of real-world java bugs.
\newblock In Andy Zaidman, Yasutaka Kamei, and Emily Hill (eds.), \emph{Proceedings of the 15th International Conference on Mining Software Repositories, {MSR} 2018, Gothenburg, Sweden, May 28-29, 2018}, pp.\  10--13. {ACM}, 2018.
\newblock \doi{10.1145/3196398.3196473}.
\newblock URL \url{https://doi.org/10.1145/3196398.3196473}.

\bibitem[Sahin \& Bahtiyar(2020)Sahin and Bahtiyar]{malware-survey2020-1}
Muhammet Sahin and Serif Bahtiyar.
\newblock A survey on malware detection with deep learning.
\newblock In Siddika~Berna {\"{O}}rs and Atilla El{\c{c}}i (eds.), \emph{{SIN} 2020: 13th International Conference on Security of Information and Networks, Virtual Event / Istanbul, Turkey, November 4-6, 2020}, pp.\  34:1--34:6. {ACM}, 2020.
\newblock \doi{10.1145/3433174.3433609}.
\newblock URL \url{https://doi.org/10.1145/3433174.3433609}.

\bibitem[Saieva et~al.(2023)Saieva, Chakraborty, and Kaiser]{search2023-1}
Anthony Saieva, Saikat Chakraborty, and Gail~E. Kaiser.
\newblock On contrastive learning of semantic similarity forcode to code search.
\newblock \emph{CoRR}, abs/2305.03843, 2023.
\newblock \doi{10.48550/ARXIV.2305.03843}.
\newblock URL \url{https://doi.org/10.48550/arXiv.2305.03843}.

\bibitem[Sainani et~al.(2020)Sainani, Anish, Joshi, and Ghaisas]{re-ana2020-1}
Abhishek Sainani, Preethu~Rose Anish, Vivek Joshi, and Smita Ghaisas.
\newblock Extracting and classifying requirements from software engineering contracts.
\newblock In Travis~D. Breaux, Andrea Zisman, Samuel Fricker, and Martin Glinz (eds.), \emph{28th {IEEE} International Requirements Engineering Conference, {RE} 2020, Zurich, Switzerland, August 31 - September 4, 2020}, pp.\  147--157. {IEEE}, 2020.
\newblock \doi{10.1109/RE48521.2020.00026}.
\newblock URL \url{https://doi.org/10.1109/RE48521.2020.00026}.

\bibitem[Saini et~al.(2018)Saini, Farmahinifarahani, Lu, Baldi, and Lopes]{clone2018-2}
Vaibhav Saini, Farima Farmahinifarahani, Yadong Lu, Pierre Baldi, and Cristina~V. Lopes.
\newblock Oreo: detection of clones in the twilight zone.
\newblock In Gary~T. Leavens, Alessandro Garcia, and Corina~S. Pasareanu (eds.), \emph{Proceedings of the 2018 {ACM} Joint Meeting on European Software Engineering Conference and Symposium on the Foundations of Software Engineering, {ESEC/SIGSOFT} {FSE} 2018, Lake Buena Vista, FL, USA, November 04-09, 2018}, pp.\  354--365. {ACM}, 2018.
\newblock \doi{10.1145/3236024.3236026}.
\newblock URL \url{https://doi.org/10.1145/3236024.3236026}.

\bibitem[Sajnani et~al.(2016)Sajnani, Saini, Svajlenko, Roy, and Lopes]{clone2015-1}
Hitesh Sajnani, Vaibhav Saini, Jeffrey Svajlenko, Chanchal~K. Roy, and Cristina~V. Lopes.
\newblock Sourcerercc: scaling code clone detection to big-code.
\newblock In Laura~K. Dillon, Willem Visser, and Laurie~A. Williams (eds.), \emph{Proceedings of the 38th International Conference on Software Engineering, {ICSE} 2016, Austin, TX, USA, May 14-22, 2016}, pp.\  1157--1168. {ACM}, 2016.
\newblock \doi{10.1145/2884781.2884877}.
\newblock URL \url{https://doi.org/10.1145/2884781.2884877}.

\bibitem[Salza et~al.(2023)Salza, Schwizer, Gu, and Gall]{retrieval2021-5}
Pasquale Salza, Christoph Schwizer, Jian Gu, and Harald~C. Gall.
\newblock On the effectiveness of transfer learning for code search.
\newblock \emph{{IEEE} Trans. Software Eng.}, 49\penalty0 (4):\penalty0 1804--1822, 2023.
\newblock \doi{10.1109/TSE.2022.3192755}.
\newblock URL \url{https://doi.org/10.1109/TSE.2022.3192755}.

\bibitem[Sandoval et~al.(2023)Sandoval, Pearce, Nys, Karri, Garg, and Dolan{-}Gavitt]{analysis2022-3}
Gustavo Sandoval, Hammond Pearce, Teo Nys, Ramesh Karri, Siddharth Garg, and Brendan Dolan{-}Gavitt.
\newblock Lost at {C:} {A} user study on the security implications of large language model code assistants.
\newblock In Joseph~A. Calandrino and Carmela Troncoso (eds.), \emph{32nd {USENIX} Security Symposium, {USENIX} Security 2023, Anaheim, CA, USA, August 9-11, 2023}, pp.\  2205--2222. {USENIX} Association, 2023.
\newblock URL \url{https://www.usenix.org/conference/usenixsecurity23/presentation/sandoval}.

\bibitem[Sanh et~al.(2022)Sanh, Webson, Raffel, Bach, Sutawika, Alyafeai, Chaffin, Stiegler, Raja, Dey, Bari, Xu, Thakker, Sharma, Szczechla, Kim, Chhablani, Nayak, Datta, Chang, Jiang, Wang, Manica, Shen, Yong, Pandey, Bawden, Wang, Neeraj, Rozen, Sharma, Santilli, F{\'{e}}vry, Fries, Teehan, Scao, Biderman, Gao, Wolf, and Rush]{2021T0}
Victor Sanh, Albert Webson, Colin Raffel, Stephen~H. Bach, Lintang Sutawika, Zaid Alyafeai, Antoine Chaffin, Arnaud Stiegler, Arun Raja, Manan Dey, M~Saiful Bari, Canwen Xu, Urmish Thakker, Shanya~Sharma Sharma, Eliza Szczechla, Taewoon Kim, Gunjan Chhablani, Nihal~V. Nayak, Debajyoti Datta, Jonathan Chang, Mike~Tian{-}Jian Jiang, Han Wang, Matteo Manica, Sheng Shen, Zheng~Xin Yong, Harshit Pandey, Rachel Bawden, Thomas Wang, Trishala Neeraj, Jos Rozen, Abheesht Sharma, Andrea Santilli, Thibault F{\'{e}}vry, Jason~Alan Fries, Ryan Teehan, Teven~Le Scao, Stella Biderman, Leo Gao, Thomas Wolf, and Alexander~M. Rush.
\newblock Multitask prompted training enables zero-shot task generalization.
\newblock In \emph{The Tenth International Conference on Learning Representations, {ICLR} 2022, Virtual Event, April 25-29, 2022}. OpenReview.net, 2022.
\newblock URL \url{https://openreview.net/forum?id=9Vrb9D0WI4}.

\bibitem[Sarker et~al.(2024)Sarker, Downing, Desai, and Bultan]{analysis2024-4}
Laboni Sarker, Mara Downing, Achintya Desai, and Tevfik Bultan.
\newblock Syntactic robustness for llm-based code generation.
\newblock \emph{CoRR}, abs/2404.01535, 2024.
\newblock \doi{10.48550/ARXIV.2404.01535}.
\newblock URL \url{https://doi.org/10.48550/arXiv.2404.01535}.

\bibitem[Scao et~al.(2022)Scao, Fan, Akiki, Pavlick, Ilic, Hesslow, Castagn{\'{e}}, Luccioni, Yvon, Gall{\'{e}}, Tow, Rush, Biderman, Webson, Ammanamanchi, Wang, Sagot, Muennighoff, del Moral, Ruwase, Bawden, Bekman, McMillan{-}Major, Beltagy, Nguyen, Saulnier, Tan, Suarez, Sanh, Lauren{\c{c}}on, Jernite, Launay, Mitchell, Raffel, Gokaslan, Simhi, Soroa, Aji, Alfassy, Rogers, Nitzav, Xu, Mou, Emezue, Klamm, Leong, van Strien, Adelani, and et~al.]{2022BLOOM}
Teven~Le Scao, Angela Fan, Christopher Akiki, Ellie Pavlick, Suzana Ilic, Daniel Hesslow, Roman Castagn{\'{e}}, Alexandra~Sasha Luccioni, Fran{\c{c}}ois Yvon, Matthias Gall{\'{e}}, Jonathan Tow, Alexander~M. Rush, Stella Biderman, Albert Webson, Pawan~Sasanka Ammanamanchi, Thomas Wang, Beno{\^{\i}}t Sagot, Niklas Muennighoff, Albert~Villanova del Moral, Olatunji Ruwase, Rachel Bawden, Stas Bekman, Angelina McMillan{-}Major, Iz~Beltagy, Huu Nguyen, Lucile Saulnier, Samson Tan, Pedro~Ortiz Suarez, Victor Sanh, Hugo Lauren{\c{c}}on, Yacine Jernite, Julien Launay, Margaret Mitchell, Colin Raffel, Aaron Gokaslan, Adi Simhi, Aitor Soroa, Alham~Fikri Aji, Amit Alfassy, Anna Rogers, Ariel~Kreisberg Nitzav, Canwen Xu, Chenghao Mou, Chris Emezue, Christopher Klamm, Colin Leong, Daniel van Strien, David~Ifeoluwa Adelani, and et~al.
\newblock {BLOOM:} {A} 176b-parameter open-access multilingual language model.
\newblock \emph{CoRR}, abs/2211.05100, 2022.
\newblock \doi{10.48550/arXiv.2211.05100}.
\newblock URL \url{https://doi.org/10.48550/arXiv.2211.05100}.

\bibitem[Schick et~al.(2023)Schick, Dwivedi-Yu, Dessi, Raileanu, Lomeli, Hambro, Zettlemoyer, Cancedda, and Scialom]{2023Toolformer}
Timo Schick, Jane Dwivedi-Yu, Roberto Dessi, Roberta Raileanu, Maria Lomeli, Eric Hambro, Luke Zettlemoyer, Nicola Cancedda, and Thomas Scialom.
\newblock Toolformer: Language models can teach themselves to use tools.
\newblock In \emph{Thirty-seventh Conference on Neural Information Processing Systems}, 2023.
\newblock URL \url{https://openreview.net/forum?id=Yacmpz84TH}.

\bibitem[Scholak et~al.(2021)Scholak, Schucher, and Bahdanau]{sql2021-2}
Torsten Scholak, Nathan Schucher, and Dzmitry Bahdanau.
\newblock {PICARD:} parsing incrementally for constrained auto-regressive decoding from language models.
\newblock In Marie{-}Francine Moens, Xuanjing Huang, Lucia Specia, and Scott~Wen{-}tau Yih (eds.), \emph{Proceedings of the 2021 Conference on Empirical Methods in Natural Language Processing, {EMNLP} 2021, Virtual Event / Punta Cana, Dominican Republic, 7-11 November, 2021}, pp.\  9895--9901. Association for Computational Linguistics, 2021.
\newblock \doi{10.18653/v1/2021.emnlp-main.779}.
\newblock URL \url{https://doi.org/10.18653/v1/2021.emnlp-main.779}.

\bibitem[Schulman et~al.(2017)Schulman, Wolski, Dhariwal, Radford, and Klimov]{2017PPO}
John Schulman, Filip Wolski, Prafulla Dhariwal, Alec Radford, and Oleg Klimov.
\newblock Proximal policy optimization algorithms.
\newblock \emph{CoRR}, abs/1707.06347, 2017.
\newblock URL \url{http://arxiv.org/abs/1707.06347}.

\bibitem[Schuster et~al.(2021)Schuster, Song, Tromer, and Shmatikov]{ana-vul2021-1}
Roei Schuster, Congzheng Song, Eran Tromer, and Vitaly Shmatikov.
\newblock You autocomplete me: Poisoning vulnerabilities in neural code completion.
\newblock In Michael~D. Bailey and Rachel Greenstadt (eds.), \emph{30th {USENIX} Security Symposium, {USENIX} Security 2021, August 11-13, 2021}, pp.\  1559--1575. {USENIX} Association, 2021.
\newblock URL \url{https://www.usenix.org/conference/usenixsecurity21/presentation/schuster}.

\bibitem[Schäfer et~al.(2023)Schäfer, Nadi, Eghbali, and Tip]{unit2023-1}
Max Schäfer, Sarah Nadi, Aryaz Eghbali, and Frank Tip.
\newblock An empirical evaluation of using large language models for automated unit test generation, 2023.

\bibitem[Selakovic et~al.(2018)Selakovic, Pradel, Karim, and Tip]{unit2018-1}
Marija Selakovic, Michael Pradel, Rezwana Karim, and Frank Tip.
\newblock Test generation for higher-order functions in dynamic languages.
\newblock \emph{Proc. {ACM} Program. Lang.}, 2\penalty0 ({OOPSLA}):\penalty0 161:1--161:27, 2018.
\newblock \doi{10.1145/3276531}.
\newblock URL \url{https://doi.org/10.1145/3276531}.

\bibitem[Seneviratne et~al.(2022)Seneviratne, Shariffdeen, Rasnayaka, and Kasthuriarachchi]{2022SHERLOCK}
Sachith Seneviratne, Ridwan Shariffdeen, Sanka Rasnayaka, and Nuran Kasthuriarachchi.
\newblock Self-supervised vision transformers for malware detection.
\newblock \emph{{IEEE} Access}, 10:\penalty0 103121--103135, 2022.
\newblock \doi{10.1109/ACCESS.2022.3206445}.
\newblock URL \url{https://doi.org/10.1109/ACCESS.2022.3206445}.

\bibitem[Sennrich et~al.(2016)Sennrich, Haddow, and Birch]{2015BT}
Rico Sennrich, Barry Haddow, and Alexandra Birch.
\newblock Improving neural machine translation models with monolingual data.
\newblock In \emph{Proceedings of the 54th Annual Meeting of the Association for Computational Linguistics, {ACL} 2016, August 7-12, 2016, Berlin, Germany, Volume 1: Long Papers}. The Association for Computer Linguistics, 2016.
\newblock \doi{10.18653/v1/p16-1009}.
\newblock URL \url{https://doi.org/10.18653/v1/p16-1009}.

\bibitem[Shahbazi \& Fard(2023)Shahbazi and Fard]{comment2023-2}
Ramin Shahbazi and Fatemeh~Hendijani Fard.
\newblock Apicontext2com: Code comment generation by incorporating pre-defined {API} documentation.
\newblock In \emph{31st {IEEE/ACM} International Conference on Program Comprehension, {ICPC} 2023, Melbourne, Australia, May 15-16, 2023}, pp.\  13--24. {IEEE}, 2023.
\newblock \doi{10.1109/ICPC58990.2023.00012}.
\newblock URL \url{https://doi.org/10.1109/ICPC58990.2023.00012}.

\bibitem[Shamshiri(2015)]{unit2015-1}
Sina Shamshiri.
\newblock Automated unit test generation for evolving software.
\newblock In Elisabetta~Di Nitto, Mark Harman, and Patrick Heymans (eds.), \emph{Proceedings of the 2015 10th Joint Meeting on Foundations of Software Engineering, {ESEC/FSE} 2015, Bergamo, Italy, August 30 - September 4, 2015}, pp.\  1038--1041. {ACM}, 2015.
\newblock \doi{10.1145/2786805.2803196}.
\newblock URL \url{https://doi.org/10.1145/2786805.2803196}.

\bibitem[Sharma et~al.(2022)Sharma, Chen, and Fard]{comment2022-1}
Rishab Sharma, Fuxiang Chen, and Fatemeh~H. Fard.
\newblock {LAMNER:} code comment generation using character language model and named entity recognition.
\newblock In Ayushi Rastogi, Rosalia Tufano, Gabriele Bavota, Venera Arnaoudova, and Sonia Haiduc (eds.), \emph{Proceedings of the 30th {IEEE/ACM} International Conference on Program Comprehension, {ICPC} 2022, Virtual Event, May 16-17, 2022}, pp.\  48--59. {ACM}, 2022.
\newblock \doi{10.1145/3524610.3527924}.
\newblock URL \url{https://doi.org/10.1145/3524610.3527924}.

\bibitem[Shaw et~al.(2018)Shaw, Uszkoreit, and Vaswani]{2018RelativePE}
Peter Shaw, Jakob Uszkoreit, and Ashish Vaswani.
\newblock Self-attention with relative position representations.
\newblock In Marilyn~A. Walker, Heng Ji, and Amanda Stent (eds.), \emph{Proceedings of the 2018 Conference of the North American Chapter of the Association for Computational Linguistics: Human Language Technologies, NAACL-HLT, New Orleans, Louisiana, USA, June 1-6, 2018, Volume 2 (Short Papers)}, pp.\  464--468. Association for Computational Linguistics, 2018.
\newblock \doi{10.18653/V1/N18-2074}.
\newblock URL \url{https://doi.org/10.18653/v1/n18-2074}.

\bibitem[Shaw et~al.(2021)Shaw, Chang, Pasupat, and Toutanova]{sql2020-4}
Peter Shaw, Ming{-}Wei Chang, Panupong Pasupat, and Kristina Toutanova.
\newblock Compositional generalization and natural language variation: Can a semantic parsing approach handle both?
\newblock In Chengqing Zong, Fei Xia, Wenjie Li, and Roberto Navigli (eds.), \emph{Proceedings of the 59th Annual Meeting of the Association for Computational Linguistics and the 11th International Joint Conference on Natural Language Processing, {ACL/IJCNLP} 2021, (Volume 1: Long Papers), Virtual Event, August 1-6, 2021}, pp.\  922--938. Association for Computational Linguistics, 2021.
\newblock \doi{10.18653/v1/2021.acl-long.75}.
\newblock URL \url{https://doi.org/10.18653/v1/2021.acl-long.75}.

\bibitem[Shazeer(2019)]{2019MultiQuery}
Noam Shazeer.
\newblock Fast transformer decoding: One write-head is all you need.
\newblock \emph{CoRR}, abs/1911.02150, 2019.
\newblock URL \url{http://arxiv.org/abs/1911.02150}.

\bibitem[She et~al.(2019)She, Pei, Epstein, Yang, Ray, and Jana]{fuzz2018-2}
Dongdong She, Kexin Pei, Dave Epstein, Junfeng Yang, Baishakhi Ray, and Suman Jana.
\newblock {NEUZZ:} efficient fuzzing with neural program smoothing.
\newblock In \emph{2019 {IEEE} Symposium on Security and Privacy, {SP} 2019, San Francisco, CA, USA, May 19-23, 2019}, pp.\  803--817. {IEEE}, 2019.
\newblock \doi{10.1109/SP.2019.00052}.
\newblock URL \url{https://doi.org/10.1109/SP.2019.00052}.

\bibitem[She et~al.(2020)She, Krishna, Yan, Jana, and Ray]{fuzz2020-1}
Dongdong She, Rahul Krishna, Lu~Yan, Suman Jana, and Baishakhi Ray.
\newblock Mtfuzz: fuzzing with a multi-task neural network.
\newblock In Prem Devanbu, Myra~B. Cohen, and Thomas Zimmermann (eds.), \emph{{ESEC/FSE} '20: 28th {ACM} Joint European Software Engineering Conference and Symposium on the Foundations of Software Engineering, Virtual Event, USA, November 8-13, 2020}, pp.\  737--749. {ACM}, 2020.
\newblock \doi{10.1145/3368089.3409723}.
\newblock URL \url{https://doi.org/10.1145/3368089.3409723}.

\bibitem[She et~al.(2023)She, Liu, Zhao, He, Li, Tantithamthavorn, Qin, and Wang]{2023survey4}
Xinyu She, Yue Liu, Yanjie Zhao, Yiling He, Li~Li, Chakkrit Tantithamthavorn, Zhan Qin, and Haoyu Wang.
\newblock Pitfalls in language models for code intelligence: {A} taxonomy and survey.
\newblock \emph{CoRR}, abs/2310.17903, 2023.
\newblock \doi{10.48550/ARXIV.2310.17903}.
\newblock URL \url{https://doi.org/10.48550/arXiv.2310.17903}.

\bibitem[Shen et~al.(2023)Shen, Zhang, Chen, Zan, Geng, Fu, Zeng, Yu, Ji, Zhao, Guo, and Wang]{2023Pangu-Coder2}
Bo~Shen, Jiaxin Zhang, Taihong Chen, Daoguang Zan, Bing Geng, An~Fu, Muhan Zeng, Ailun Yu, Jichuan Ji, Jingyang Zhao, Yuenan Guo, and Qianxiang Wang.
\newblock Pangu-coder2: Boosting large language models for code with ranking feedback.
\newblock \emph{CoRR}, abs/2307.14936, 2023.
\newblock \doi{10.48550/arXiv.2307.14936}.
\newblock URL \url{https://doi.org/10.48550/arXiv.2307.14936}.

\bibitem[Shi et~al.(2023{\natexlab{a}})Shi, Cai, Zhao, Gao, Sood, and Xiang]{sum2023-2}
Chaochen Shi, Borui Cai, Yao Zhao, Longxiang Gao, Keshav Sood, and Yong Xiang.
\newblock Coss: Leveraging statement semantics for code summarization.
\newblock \emph{{IEEE} Trans. Software Eng.}, 49\penalty0 (6):\penalty0 3472--3486, 2023{\natexlab{a}}.
\newblock \doi{10.1109/TSE.2023.3256362}.
\newblock URL \url{https://doi.org/10.1109/TSE.2023.3256362}.

\bibitem[Shi et~al.(2022)Shi, Wang, Tao, Du, Zhang, Han, Zhang, and Sun]{commit2022-1}
Ensheng Shi, Yanlin Wang, Wei Tao, Lun Du, Hongyu Zhang, Shi Han, Dongmei Zhang, and Hongbin Sun.
\newblock {RACE:} retrieval-augmented commit message generation.
\newblock In Yoav Goldberg, Zornitsa Kozareva, and Yue Zhang (eds.), \emph{Proceedings of the 2022 Conference on Empirical Methods in Natural Language Processing, {EMNLP} 2022, Abu Dhabi, United Arab Emirates, December 7-11, 2022}, pp.\  5520--5530. Association for Computational Linguistics, 2022.
\newblock \doi{10.18653/V1/2022.EMNLP-MAIN.372}.
\newblock URL \url{https://doi.org/10.18653/v1/2022.emnlp-main.372}.

\bibitem[Shi et~al.(2023{\natexlab{b}})Shi, Wang, Gu, Du, Zhang, Han, Zhang, and Sun]{retrieval2022-4}
Ensheng Shi, Yanlin Wang, Wenchao Gu, Lun Du, Hongyu Zhang, Shi Han, Dongmei Zhang, and Hongbin Sun.
\newblock Cocosoda: Effective contrastive learning for code search.
\newblock In \emph{45th {IEEE/ACM} International Conference on Software Engineering, {ICSE} 2023, Melbourne, Australia, May 14-20, 2023}, pp.\  2198--2210. {IEEE}, 2023{\natexlab{b}}.
\newblock \doi{10.1109/ICSE48619.2023.00185}.
\newblock URL \url{https://doi.org/10.1109/ICSE48619.2023.00185}.

\bibitem[Shi et~al.(2021)Shi, Ng, Wang, Zhu, Li, Wang, dos Santos, and Xiang]{sql2020-5}
Peng Shi, Patrick Ng, Zhiguo Wang, Henghui Zhu, Alexander~Hanbo Li, Jun Wang, C{\'{\i}}cero~Nogueira dos Santos, and Bing Xiang.
\newblock Learning contextual representations for semantic parsing with generation-augmented pre-training.
\newblock In \emph{Thirty-Fifth {AAAI} Conference on Artificial Intelligence, {AAAI} 2021, Thirty-Third Conference on Innovative Applications of Artificial Intelligence, {IAAI} 2021, The Eleventh Symposium on Educational Advances in Artificial Intelligence, {EAAI} 2021, Virtual Event, February 2-9, 2021}, pp.\  13806--13814. {AAAI} Press, 2021.
\newblock \doi{10.1609/AAAI.V35I15.17627}.
\newblock URL \url{https://doi.org/10.1609/aaai.v35i15.17627}.

\bibitem[Shi et~al.(2019)Shi, Li, Lo, Thung, and Huo]{review2019-2}
Shu{-}Ting Shi, Ming Li, David Lo, Ferdian Thung, and Xuan Huo.
\newblock Automatic code review by learning the revision of source code.
\newblock In \emph{The Thirty-Third {AAAI} Conference on Artificial Intelligence, {AAAI} 2019, The Thirty-First Innovative Applications of Artificial Intelligence Conference, {IAAI} 2019, The Ninth {AAAI} Symposium on Educational Advances in Artificial Intelligence, {EAAI} 2019, Honolulu, Hawaii, USA, January 27 - February 1, 2019}, pp.\  4910--4917. {AAAI} Press, 2019.
\newblock \doi{10.1609/aaai.v33i01.33014910}.
\newblock URL \url{https://doi.org/10.1609/aaai.v33i01.33014910}.

\bibitem[Shi et~al.(2020)Shi, Zhao, Boyd{-}Graber, III, and Lee]{sql-data-2020-1}
Tianze Shi, Chen Zhao, Jordan~L. Boyd{-}Graber, Hal~Daum{\'{e}} III, and Lillian Lee.
\newblock On the potential of lexico-logical alignments for semantic parsing to {SQL} queries.
\newblock In Trevor Cohn, Yulan He, and Yang Liu (eds.), \emph{Findings of the Association for Computational Linguistics: {EMNLP} 2020, Online Event, 16-20 November 2020}, volume {EMNLP} 2020 of \emph{Findings of {ACL}}, pp.\  1849--1864. Association for Computational Linguistics, 2020.
\newblock \doi{10.18653/V1/2020.FINDINGS-EMNLP.167}.
\newblock URL \url{https://doi.org/10.18653/v1/2020.findings-emnlp.167}.

\bibitem[Shimmi \& Rahimi(2022)Shimmi and Rahimi]{unit2022-1}
Samiha Shimmi and Mona Rahimi.
\newblock Leveraging code-test co-evolution patterns for automated test case recommendation.
\newblock In \emph{{IEEE/ACM} International Conference on Automation of Software Test, AST@ICSE 2022, Pittsburgh, PA, USA, May 21-22, 2022}, pp.\  65--76. {ACM/IEEE}, 2022.
\newblock \doi{10.1145/3524481.3527222}.
\newblock URL \url{https://doi.org/10.1145/3524481.3527222}.

\bibitem[Shin et~al.(2023)Shin, Hashtroudi, Hemmati, and Wang]{unit2023-8}
Jiho Shin, Sepehr Hashtroudi, Hadi Hemmati, and Song Wang.
\newblock Domain adaptation for deep unit test case generation.
\newblock \emph{CoRR}, abs/2308.08033, 2023.
\newblock \doi{10.48550/ARXIV.2308.08033}.
\newblock URL \url{https://doi.org/10.48550/arXiv.2308.08033}.

\bibitem[Shojaee et~al.(2023)Shojaee, Jain, Tipirneni, and Reddy]{2023PPOCoder}
Parshin Shojaee, Aneesh Jain, Sindhu Tipirneni, and Chandan~K. Reddy.
\newblock Execution-based code generation using deep reinforcement learning.
\newblock \emph{Transactions on Machine Learning Research}, 2023.
\newblock ISSN 2835-8856.
\newblock URL \url{https://openreview.net/forum?id=0XBuaxqEcG}.

\bibitem[Shrivastava et~al.(2023{\natexlab{a}})Shrivastava, Kocetkov, de~Vries, Bahdanau, and Scholak]{repo2023-2}
Disha Shrivastava, Denis Kocetkov, Harm de~Vries, Dzmitry Bahdanau, and Torsten Scholak.
\newblock Repofusion: Training code models to understand your repository.
\newblock \emph{CoRR}, abs/2306.10998, 2023{\natexlab{a}}.
\newblock \doi{10.48550/ARXIV.2306.10998}.
\newblock URL \url{https://doi.org/10.48550/arXiv.2306.10998}.

\bibitem[Shrivastava et~al.(2023{\natexlab{b}})Shrivastava, Larochelle, and Tarlow]{repo2022-1}
Disha Shrivastava, Hugo Larochelle, and Daniel Tarlow.
\newblock Repository-level prompt generation for large language models of code.
\newblock In Andreas Krause, Emma Brunskill, Kyunghyun Cho, Barbara Engelhardt, Sivan Sabato, and Jonathan Scarlett (eds.), \emph{International Conference on Machine Learning, {ICML} 2023, 23-29 July 2023, Honolulu, Hawaii, {USA}}, volume 202 of \emph{Proceedings of Machine Learning Research}, pp.\  31693--31715. {PMLR}, 2023{\natexlab{b}}.
\newblock URL \url{https://proceedings.mlr.press/v202/shrivastava23a.html}.

\bibitem[Shuai et~al.(2020)Shuai, Xu, Liu, Yan, Xia, and Lei]{retrieval2020-0.7}
Jianhang Shuai, Ling Xu, Chao Liu, Meng Yan, Xin Xia, and Yan Lei.
\newblock Improving code search with co-attentive representation learning.
\newblock In \emph{{ICPC} '20: 28th International Conference on Program Comprehension, Seoul, Republic of Korea, July 13-15, 2020}, pp.\  196--207. {ACM}, 2020.
\newblock \doi{10.1145/3387904.3389269}.
\newblock URL \url{https://doi.org/10.1145/3387904.3389269}.

\bibitem[Si et~al.(2024)Si, Zhang, Yang, Liu, and Yang]{UI-data-2024-1}
Chenglei Si, Yanzhe Zhang, Zhengyuan Yang, Ruibo Liu, and Diyi Yang.
\newblock Design2code: How far are we from automating front-end engineering?
\newblock \emph{CoRR}, abs/2403.03163, 2024.
\newblock \doi{10.48550/ARXIV.2403.03163}.
\newblock URL \url{https://doi.org/10.48550/arXiv.2403.03163}.

\bibitem[Singh et~al.(2023)Singh, Cambronero, Gulwani, Le, Negreanu, and Verbruggen]{2023codefusion}
Mukul Singh, José Cambronero, Sumit Gulwani, Vu~Le, Carina Negreanu, and Gust Verbruggen.
\newblock Codefusion: A pre-trained diffusion model for code generation, 2023.

\bibitem[Siow et~al.(2020)Siow, Gao, Fan, Chen, and Liu]{review2019-3}
Jing~Kai Siow, Cuiyun Gao, Lingling Fan, Sen Chen, and Yang Liu.
\newblock {CORE:} automating review recommendation for code changes.
\newblock In Kostas Kontogiannis, Foutse Khomh, Alexander Chatzigeorgiou, Marios{-}Eleftherios Fokaefs, and Minghui Zhou (eds.), \emph{27th {IEEE} International Conference on Software Analysis, Evolution and Reengineering, {SANER} 2020, London, ON, Canada, February 18-21, 2020}, pp.\  284--295. {IEEE}, 2020.
\newblock \doi{10.1109/SANER48275.2020.9054794}.
\newblock URL \url{https://doi.org/10.1109/SANER48275.2020.9054794}.

\bibitem[Sivaraman et~al.(2022)Sivaraman, Abreu, Scott, Akomolede, and Chandra]{idiom2021-1}
Aishwarya Sivaraman, Rui Abreu, Andrew Scott, Tobi Akomolede, and Satish Chandra.
\newblock Mining idioms in the wild.
\newblock In \emph{44th {IEEE/ACM} International Conference on Software Engineering: Software Engineering in Practice, {ICSE} {(SEIP)} 2022, Pittsburgh, PA, USA, May 22-24, 2022}, pp.\  187--196. {IEEE}, 2022.
\newblock \doi{10.1109/ICSE-SEIP55303.2022.9794062}.
\newblock URL \url{https://doi.org/10.1109/ICSE-SEIP55303.2022.9794062}.

\bibitem[Soltan et~al.(2022)Soltan, Ananthakrishnan, FitzGerald, Gupta, Hamza, Khan, Peris, Rawls, Rosenbaum, Rumshisky, Prakash, Sridhar, Triefenbach, Verma, T{\"{u}}r, and Natarajan]{2022AlexaTM}
Saleh Soltan, Shankar Ananthakrishnan, Jack FitzGerald, Rahul Gupta, Wael Hamza, Haidar Khan, Charith Peris, Stephen Rawls, Andy Rosenbaum, Anna Rumshisky, Chandana~Satya Prakash, Mukund Sridhar, Fabian Triefenbach, Apurv Verma, G{\"{o}}khan T{\"{u}}r, and Prem Natarajan.
\newblock Alexatm 20b: Few-shot learning using a large-scale multilingual seq2seq model.
\newblock \emph{CoRR}, abs/2208.01448, 2022.
\newblock \doi{10.48550/arXiv.2208.01448}.
\newblock URL \url{https://doi.org/10.48550/arXiv.2208.01448}.

\bibitem[Sorokin et~al.(2023)Sorokin, Abulkhanov, Nikolenko, and Malykh]{clone2023-0.5}
Nikita Sorokin, Dmitry Abulkhanov, Sergey~I. Nikolenko, and Valentin Malykh.
\newblock Cct-code: Cross-consistency training for multilingual clone detection and code search.
\newblock \emph{CoRR}, abs/2305.11626, 2023.
\newblock \doi{10.48550/ARXIV.2305.11626}.
\newblock URL \url{https://doi.org/10.48550/arXiv.2305.11626}.

\bibitem[Soselia et~al.(2023)Soselia, Saifullah, and Zhou]{UI-2023-4}
Davit Soselia, Khalid Saifullah, and Tianyi Zhou.
\newblock Learning ui-to-code reverse generator using visual critic without rendering.
\newblock \emph{CoRR}, abs/2305.14637, 2023.
\newblock \doi{10.48550/ARXIV.2305.14637}.
\newblock URL \url{https://doi.org/10.48550/arXiv.2305.14637}.

\bibitem[Steenhoek et~al.(2023{\natexlab{a}})Steenhoek, Rahman, Jiles, and Le]{defect-survey-2022-2}
Benjamin Steenhoek, Md~Mahbubur Rahman, Richard Jiles, and Wei Le.
\newblock An empirical study of deep learning models for vulnerability detection.
\newblock In \emph{45th {IEEE/ACM} International Conference on Software Engineering, {ICSE} 2023, Melbourne, Australia, May 14-20, 2023}, pp.\  2237--2248. {IEEE}, 2023{\natexlab{a}}.
\newblock \doi{10.1109/ICSE48619.2023.00188}.
\newblock URL \url{https://doi.org/10.1109/ICSE48619.2023.00188}.

\bibitem[Steenhoek et~al.(2023{\natexlab{b}})Steenhoek, Tufano, Sundaresan, and Svyatkovskiy]{2023RLSQM}
Benjamin Steenhoek, Michele Tufano, Neel Sundaresan, and Alexey Svyatkovskiy.
\newblock Reinforcement learning from automatic feedback for high-quality unit test generation.
\newblock \emph{CoRR}, abs/2310.02368, 2023{\natexlab{b}}.
\newblock \doi{10.48550/ARXIV.2310.02368}.
\newblock URL \url{https://doi.org/10.48550/arXiv.2310.02368}.

\bibitem[Steenhoek et~al.(2024)Steenhoek, Rahman, Roy, Alam, Barr, and Le]{defect2024-1}
Benjamin Steenhoek, Md~Mahbubur Rahman, Monoshi~Kumar Roy, Mirza~Sanjida Alam, Earl~T. Barr, and Wei Le.
\newblock A comprehensive study of the capabilities of large language models for vulnerability detection.
\newblock \emph{CoRR}, abs/2403.17218, 2024.
\newblock \doi{10.48550/ARXIV.2403.17218}.
\newblock URL \url{https://doi.org/10.48550/arXiv.2403.17218}.

\bibitem[Su \& McMillan(2023{\natexlab{a}})Su and McMillan]{sum2023-4}
Chia{-}Yi Su and Collin McMillan.
\newblock Semantic similarity loss for neural source code summarization.
\newblock \emph{CoRR}, abs/2308.07429, 2023{\natexlab{a}}.
\newblock \doi{10.48550/ARXIV.2308.07429}.
\newblock URL \url{https://doi.org/10.48550/arXiv.2308.07429}.

\bibitem[Su \& McMillan(2023{\natexlab{b}})Su and McMillan]{sum2023-5}
Chia{-}Yi Su and Collin McMillan.
\newblock Distilled {GPT} for source code summarization.
\newblock \emph{CoRR}, abs/2308.14731, 2023{\natexlab{b}}.
\newblock \doi{10.48550/ARXIV.2308.14731}.
\newblock URL \url{https://doi.org/10.48550/arXiv.2308.14731}.

\bibitem[Su et~al.(2023{\natexlab{a}})Su, Bansal, Jain, Ghanavati, and McMillan]{2023Jam}
Chia{-}Yi Su, Aakash Bansal, Vijayanta Jain, Sepideh Ghanavati, and Collin McMillan.
\newblock A language model of java methods with train/test deduplication.
\newblock \emph{CoRR}, abs/2305.08286, 2023{\natexlab{a}}.
\newblock \doi{10.48550/ARXIV.2305.08286}.
\newblock URL \url{https://doi.org/10.48550/arXiv.2305.08286}.

\bibitem[Su et~al.(2024)Su, Ahmed, Lu, Pan, Bo, and Liu]{2021RoPE}
Jianlin Su, Murtadha H.~M. Ahmed, Yu~Lu, Shengfeng Pan, Wen Bo, and Yunfeng Liu.
\newblock Roformer: Enhanced transformer with rotary position embedding.
\newblock \emph{Neurocomputing}, 568:\penalty0 127063, 2024.
\newblock \doi{10.1016/J.NEUCOM.2023.127063}.
\newblock URL \url{https://doi.org/10.1016/j.neucom.2023.127063}.

\bibitem[Su et~al.(2023{\natexlab{b}})Su, Wan, Sethi, Lu, Musuvathi, and Nath]{doc2023-1}
Yiming Su, Chengcheng Wan, Utsav Sethi, Shan Lu, Madan Musuvathi, and Suman Nath.
\newblock Hotgpt: How to make software documentation more useful with a large language model?
\newblock In Malte Schwarzkopf, Andrew Baumann, and Natacha Crooks (eds.), \emph{Proceedings of the 19th Workshop on Hot Topics in Operating Systems, {HOTOS} 2023, Providence, RI, USA, June 22-24, 2023}, pp.\  87--93. {ACM}, 2023{\natexlab{b}}.
\newblock \doi{10.1145/3593856.3595910}.
\newblock URL \url{https://doi.org/10.1145/3593856.3595910}.

\bibitem[Suhr et~al.(2018)Suhr, Iyer, and Artzi]{sql2018-1}
Alane Suhr, Srinivasan Iyer, and Yoav Artzi.
\newblock Learning to map context-dependent sentences to executable formal queries.
\newblock In Marilyn~A. Walker, Heng Ji, and Amanda Stent (eds.), \emph{Proceedings of the 2018 Conference of the North American Chapter of the Association for Computational Linguistics: Human Language Technologies, {NAACL-HLT} 2018, New Orleans, Louisiana, USA, June 1-6, 2018, Volume 1 (Long Papers)}, pp.\  2238--2249. Association for Computational Linguistics, 2018.
\newblock \doi{10.18653/v1/n18-1203}.
\newblock URL \url{https://doi.org/10.18653/v1/n18-1203}.

\bibitem[Sun et~al.(2023{\natexlab{a}})Sun, Arik, Nakhost, Dai, Sinha, Yin, and Pfister]{sql2023-1.5}
Ruoxi Sun, Sercan~{\"{O}}. Arik, Hootan Nakhost, Hanjun Dai, Rajarishi Sinha, Pengcheng Yin, and Tomas Pfister.
\newblock Sql-palm: Improved large language model adaptation for text-to-sql.
\newblock \emph{CoRR}, abs/2306.00739, 2023{\natexlab{a}}.
\newblock \doi{10.48550/ARXIV.2306.00739}.
\newblock URL \url{https://doi.org/10.48550/arXiv.2306.00739}.

\bibitem[Sun et~al.(2022)Sun, Fang, Chen, Tao, Han, and Zhang]{retrieval2022-3}
Weisong Sun, Chunrong Fang, Yuchen Chen, Guanhong Tao, Tingxu Han, and Quanjun Zhang.
\newblock Code search based on context-aware code translation.
\newblock In \emph{44th {IEEE/ACM} 44th International Conference on Software Engineering, {ICSE} 2022, Pittsburgh, PA, USA, May 25-27, 2022}, pp.\  388--400. {ACM}, 2022.
\newblock \doi{10.1145/3510003.3510140}.
\newblock URL \url{https://doi.org/10.1145/3510003.3510140}.

\bibitem[Sun et~al.(2023{\natexlab{b}})Sun, Fang, You, Miao, Liu, Li, Deng, Huang, Chen, Zhang, Qian, Liu, and Chen]{sum2023-3}
Weisong Sun, Chunrong Fang, Yudu You, Yun Miao, Yi~Liu, Yuekang Li, Gelei Deng, Shenghan Huang, Yuchen Chen, Quanjun Zhang, Hanwei Qian, Yang Liu, and Zhenyu Chen.
\newblock Automatic code summarization via chatgpt: How far are we?
\newblock \emph{CoRR}, abs/2305.12865, 2023{\natexlab{b}}.
\newblock \doi{10.48550/ARXIV.2305.12865}.
\newblock URL \url{https://doi.org/10.48550/arXiv.2305.12865}.

\bibitem[Sun et~al.(2024)Sun, Wu, Xue, Liu, Ma, Zhang, Shi, and Liu]{2024LLM4Vuln}
Yuqiang Sun, Daoyuan Wu, Yue Xue, Han Liu, Wei Ma, Lyuye Zhang, Miaolei Shi, and Yang Liu.
\newblock Llm4vuln: {A} unified evaluation framework for decoupling and enhancing llms' vulnerability reasoning.
\newblock \emph{CoRR}, abs/2401.16185, 2024.
\newblock \doi{10.48550/ARXIV.2401.16185}.
\newblock URL \url{https://doi.org/10.48550/arXiv.2401.16185}.

\bibitem[Sun et~al.(2020)Sun, Zhu, Xiong, Sun, Mou, and Zhang]{syn2019-2}
Zeyu Sun, Qihao Zhu, Yingfei Xiong, Yican Sun, Lili Mou, and Lu~Zhang.
\newblock Treegen: {A} tree-based transformer architecture for code generation.
\newblock In \emph{The Thirty-Fourth {AAAI} Conference on Artificial Intelligence, {AAAI} 2020, The Thirty-Second Innovative Applications of Artificial Intelligence Conference, {IAAI} 2020, The Tenth {AAAI} Symposium on Educational Advances in Artificial Intelligence, {EAAI} 2020, New York, NY, USA, February 7-12, 2020}, pp.\  8984--8991. {AAAI} Press, 2020.
\newblock \doi{10.1609/AAAI.V34I05.6430}.
\newblock URL \url{https://doi.org/10.1609/aaai.v34i05.6430}.

\bibitem[Sur{\'{\i}}s et~al.(2023)Sur{\'{\i}}s, Menon, and Vondrick]{2023ViperGPT}
D{\'{\i}}dac Sur{\'{\i}}s, Sachit Menon, and Carl Vondrick.
\newblock Vipergpt: Visual inference via python execution for reasoning.
\newblock \emph{CoRR}, abs/2303.08128, 2023.
\newblock \doi{10.48550/ARXIV.2303.08128}.
\newblock URL \url{https://doi.org/10.48550/arXiv.2303.08128}.

\bibitem[Sutskever et~al.(2014)Sutskever, Vinyals, and Le]{2014seq2seq}
Ilya Sutskever, Oriol Vinyals, and Quoc~V. Le.
\newblock Sequence to sequence learning with neural networks.
\newblock In Zoubin Ghahramani, Max Welling, Corinna Cortes, Neil~D. Lawrence, and Kilian~Q. Weinberger (eds.), \emph{Advances in Neural Information Processing Systems 27: Annual Conference on Neural Information Processing Systems 2014, December 8-13 2014, Montreal, Quebec, Canada}, pp.\  3104--3112, 2014.
\newblock URL \url{https://proceedings.neurips.cc/paper/2014/hash/a14ac55a4f27472c5d894ec1c3c743d2-Abstract.html}.

\bibitem[Svajlenko \& Roy(2020)Svajlenko and Roy]{clone-survey-2020-1}
Jeffrey Svajlenko and Chanchal~K. Roy.
\newblock A survey on the evaluation of clone detection performance and benchmarking.
\newblock \emph{CoRR}, abs/2006.15682, 2020.
\newblock URL \url{https://arxiv.org/abs/2006.15682}.

\bibitem[Svajlenko et~al.(2014)Svajlenko, Islam, Keivanloo, Roy, and Mia]{2014BigCloneBench}
Jeffrey Svajlenko, Judith~F. Islam, Iman Keivanloo, Chanchal~Kumar Roy, and Mohammad~Mamun Mia.
\newblock Towards a big data curated benchmark of inter-project code clones.
\newblock In \emph{30th {IEEE} International Conference on Software Maintenance and Evolution, Victoria, BC, Canada, September 29 - October 3, 2014}, pp.\  476--480. {IEEE} Computer Society, 2014.
\newblock \doi{10.1109/ICSME.2014.77}.
\newblock URL \url{https://doi.org/10.1109/ICSME.2014.77}.

\bibitem[Svyatkovskiy et~al.(2019)Svyatkovskiy, Zhao, Fu, and Sundaresan]{completion2019-2}
Alexey Svyatkovskiy, Ying Zhao, Shengyu Fu, and Neel Sundaresan.
\newblock Pythia: Ai-assisted code completion system.
\newblock In Ankur Teredesai, Vipin Kumar, Ying Li, R{\'{o}}mer Rosales, Evimaria Terzi, and George Karypis (eds.), \emph{Proceedings of the 25th {ACM} {SIGKDD} International Conference on Knowledge Discovery {\&} Data Mining, {KDD} 2019, Anchorage, AK, USA, August 4-8, 2019}, pp.\  2727--2735. {ACM}, 2019.
\newblock \doi{10.1145/3292500.3330699}.
\newblock URL \url{https://doi.org/10.1145/3292500.3330699}.

\bibitem[Svyatkovskiy et~al.(2020)Svyatkovskiy, Deng, Fu, and Sundaresan]{2020GPT-C}
Alexey Svyatkovskiy, Shao~Kun Deng, Shengyu Fu, and Neel Sundaresan.
\newblock Intellicode compose: code generation using transformer.
\newblock In Prem Devanbu, Myra~B. Cohen, and Thomas Zimmermann (eds.), \emph{{ESEC/FSE} '20: 28th {ACM} Joint European Software Engineering Conference and Symposium on the Foundations of Software Engineering, Virtual Event, USA, November 8-13, 2020}, pp.\  1433--1443. {ACM}, 2020.
\newblock \doi{10.1145/3368089.3417058}.
\newblock URL \url{https://doi.org/10.1145/3368089.3417058}.

\bibitem[Szafraniec et~al.(2023)Szafraniec, Rozi{\`{e}}re, Leather, Labatut, Charton, and Synnaeve]{2022Transcoder}
Marc Szafraniec, Baptiste Rozi{\`{e}}re, Hugh Leather, Patrick Labatut, Fran{\c{c}}ois Charton, and Gabriel Synnaeve.
\newblock Code translation with compiler representations.
\newblock In \emph{The Eleventh International Conference on Learning Representations, {ICLR} 2023, Kigali, Rwanda, May 1-5, 2023}. OpenReview.net, 2023.
\newblock URL \url{https://openreview.net/pdf?id=XomEU3eNeSQ}.

\bibitem[Tambon et~al.(2024)Tambon, Dakhel, Nikanjam, Khomh, Desmarais, and Antoniol]{analysis2024-1}
Florian Tambon, Arghavan~Moradi Dakhel, Amin Nikanjam, Foutse Khomh, Michel~C. Desmarais, and Giuliano Antoniol.
\newblock Bugs in large language models generated code: An empirical study.
\newblock \emph{CoRR}, abs/2403.08937, 2024.
\newblock \doi{10.48550/ARXIV.2403.08937}.
\newblock URL \url{https://doi.org/10.48550/arXiv.2403.08937}.

\bibitem[Tan et~al.(2024)Tan, Luo, Li, and Zhang]{LLM4Decompile}
Hanzhuo Tan, Qi~Luo, Jing Li, and Yuqun Zhang.
\newblock Llm4decompile: Decompiling binary code with large language models.
\newblock \emph{CoRR}, abs/2403.05286, 2024.
\newblock \doi{10.48550/ARXIV.2403.05286}.
\newblock URL \url{https://doi.org/10.48550/arXiv.2403.05286}.

\bibitem[Tan et~al.(2017)Tan, Yi, Yulis, Mechtaev, and Roychoudhury]{fix-data-2017-1}
Shin~Hwei Tan, Jooyong Yi, Yulis, Sergey Mechtaev, and Abhik Roychoudhury.
\newblock Codeflaws: a programming competition benchmark for evaluating automated program repair tools.
\newblock In Sebasti{\'{a}}n Uchitel, Alessandro Orso, and Martin~P. Robillard (eds.), \emph{Proceedings of the 39th International Conference on Software Engineering, {ICSE} 2017, Buenos Aires, Argentina, May 20-28, 2017 - Companion Volume}, pp.\  180--182. {IEEE} Computer Society, 2017.
\newblock \doi{10.1109/ICSE-C.2017.76}.
\newblock URL \url{https://doi.org/10.1109/ICSE-C.2017.76}.

\bibitem[Tang \& Mooney(2000)Tang and Mooney]{sql-data-2000}
Lappoon~R. Tang and Raymond~J. Mooney.
\newblock Automated construction of database interfaces: Intergrating statistical and relational learning for semantic parsing.
\newblock In Hinrich Sch{\"{u}}tze and Keh{-}Yih Su (eds.), \emph{Joint {SIGDAT} Conference on Empirical Methods in Natural Language Processing and Very Large Corpora, {EMNLP} 2000, Hong Kong, October 7-8, 2000}, pp.\  133--141. Association for Computational Linguistics, 2000.
\newblock \doi{10.3115/1117794.1117811}.
\newblock URL \url{https://aclanthology.org/W00-1317/}.

\bibitem[Tang et~al.(2023{\natexlab{a}})Tang, Liu, Zhou, and Luo]{unit2023-7}
Yutian Tang, Zhijie Liu, Zhichao Zhou, and Xiapu Luo.
\newblock Chatgpt vs {SBST:} {A} comparative assessment of unit test suite generation.
\newblock \emph{CoRR}, abs/2307.00588, 2023{\natexlab{a}}.
\newblock \doi{10.48550/ARXIV.2307.00588}.
\newblock URL \url{https://doi.org/10.48550/arXiv.2307.00588}.

\bibitem[Tang et~al.(2022)Tang, Shen, Li, Ge, Huang, Zhu, and Luo]{sum2022-2}
Ze~Tang, Xiaoyu Shen, Chuanyi Li, Jidong Ge, Liguo Huang, Zheling Zhu, and Bin Luo.
\newblock Ast-trans: Code summarization with efficient tree-structured attention.
\newblock In \emph{44th {IEEE/ACM} 44th International Conference on Software Engineering, {ICSE} 2022, Pittsburgh, PA, USA, May 25-27, 2022}, pp.\  150--162. {ACM}, 2022.
\newblock \doi{10.1145/3510003.3510224}.
\newblock URL \url{https://doi.org/10.1145/3510003.3510224}.

\bibitem[Tang et~al.(2023{\natexlab{b}})Tang, Agarwal, Shypula, Wang, Wijaya, Chen, and Kim]{trans2023-4}
Zilu Tang, Mayank Agarwal, Alex Shypula, Bailin Wang, Derry Wijaya, Jie Chen, and Yoon Kim.
\newblock Explain-then-translate: An analysis on improving program translation with self-generated explanations.
\newblock \emph{CoRR}, abs/2311.07070, 2023{\natexlab{b}}.
\newblock \doi{10.48550/ARXIV.2311.07070}.
\newblock URL \url{https://doi.org/10.48550/arXiv.2311.07070}.

\bibitem[Tao et~al.(2022)Tao, Meng, Chen, Zhu, Liu, Du, Han, Zhao, Wang, and Yang]{log2022-3}
Shimin Tao, Weibin Meng, Yimeng Chen, Yichen Zhu, Ying Liu, Chunning Du, Tao Han, Yongpeng Zhao, Xiangguang Wang, and Hao Yang.
\newblock Logstamp: Automatic online log parsing based on sequence labelling.
\newblock \emph{{SIGMETRICS} Perform. Evaluation Rev.}, 49\penalty0 (4):\penalty0 93--98, 2022.
\newblock \doi{10.1145/3543146.3543168}.
\newblock URL \url{https://doi.org/10.1145/3543146.3543168}.

\bibitem[Tao et~al.(2021)Tao, Wang, Shi, Du, Han, Zhang, Zhang, and Zhang]{commit-survey-2021}
Wei Tao, Yanlin Wang, Ensheng Shi, Lun Du, Shi Han, Hongyu Zhang, Dongmei Zhang, and Wenqiang Zhang.
\newblock On the evaluation of commit message generation models: An experimental study.
\newblock In \emph{{IEEE} International Conference on Software Maintenance and Evolution, {ICSME} 2021, Luxembourg, September 27 - October 1, 2021}, pp.\  126--136. {IEEE}, 2021.
\newblock \doi{10.1109/ICSME52107.2021.00018}.
\newblock URL \url{https://doi.org/10.1109/ICSME52107.2021.00018}.

\bibitem[Tao et~al.(2024)Tao, Zhou, Zhang, and Cheng]{2024MAGIS}
Wei Tao, Yucheng Zhou, Wenqiang Zhang, and Yu~Cheng.
\newblock {MAGIS:} llm-based multi-agent framework for github issue resolution.
\newblock \emph{CoRR}, abs/2403.17927, 2024.
\newblock \doi{10.48550/ARXIV.2403.17927}.
\newblock URL \url{https://doi.org/10.48550/arXiv.2403.17927}.

\bibitem[Taori et~al.(2023)Taori, Gulrajani, Zhang, Dubois, Li, Guestrin, Liang, and Hashimoto]{2023alpaca}
Rohan Taori, Ishaan Gulrajani, Tianyi Zhang, Yann Dubois, Xuechen Li, Carlos Guestrin, Percy Liang, and Tatsunori~B. Hashimoto.
\newblock Stanford alpaca: An instruction-following llama model.
\newblock \url{https://github.com/tatsu-lab/stanford_alpaca}, 2023.

\bibitem[Tay et~al.(2023{\natexlab{a}})Tay, Dehghani, Bahri, and Metzler]{2020EfficientTransformer}
Yi~Tay, Mostafa Dehghani, Dara Bahri, and Donald Metzler.
\newblock Efficient transformers: {A} survey.
\newblock \emph{{ACM} Comput. Surv.}, 55\penalty0 (6):\penalty0 109:1--109:28, 2023{\natexlab{a}}.
\newblock \doi{10.1145/3530811}.
\newblock URL \url{https://doi.org/10.1145/3530811}.

\bibitem[Tay et~al.(2023{\natexlab{b}})Tay, Dehghani, Tran, Garcia, Wei, Wang, Chung, Bahri, Schuster, Zheng, Zhou, Houlsby, and Metzler]{2022UL2}
Yi~Tay, Mostafa Dehghani, Vinh~Q. Tran, Xavier Garcia, Jason Wei, Xuezhi Wang, Hyung~Won Chung, Dara Bahri, Tal Schuster, Huaixiu~Steven Zheng, Denny Zhou, Neil Houlsby, and Donald Metzler.
\newblock {UL2:} unifying language learning paradigms.
\newblock In \emph{The Eleventh International Conference on Learning Representations, {ICLR} 2023, Kigali, Rwanda, May 1-5, 2023}. OpenReview.net, 2023{\natexlab{b}}.
\newblock URL \url{https://openreview.net/pdf?id=6ruVLB727MC}.

\bibitem[Team(2024)]{2024CodeGemma}
CodeGemma Team.
\newblock Codegemma: Open code models based on gemma, 2024.
\newblock URL \url{https://storage.googleapis.com/deepmind-media/gemma/codegemma_report.pdf}.

\bibitem[Thakur et~al.(2023{\natexlab{a}})Thakur, Ahmad, Fan, Pearce, Tan, Karri, Dolan{-}Gavitt, and Garg]{2022Verilog-1}
Shailja Thakur, Baleegh Ahmad, Zhenxing Fan, Hammond Pearce, Benjamin Tan, Ramesh Karri, Brendan Dolan{-}Gavitt, and Siddharth Garg.
\newblock Benchmarking large language models for automated verilog {RTL} code generation.
\newblock In \emph{Design, Automation {\&} Test in Europe Conference {\&} Exhibition, {DATE} 2023, Antwerp, Belgium, April 17-19, 2023}, pp.\  1--6. {IEEE}, 2023{\natexlab{a}}.
\newblock \doi{10.23919/DATE56975.2023.10137086}.
\newblock URL \url{https://doi.org/10.23919/DATE56975.2023.10137086}.

\bibitem[Thakur et~al.(2023{\natexlab{b}})Thakur, Ahmad, Pearce, Tan, Dolan{-}Gavitt, Karri, and Garg]{2023VeriGen}
Shailja Thakur, Baleegh Ahmad, Hammond Pearce, Benjamin Tan, Brendan Dolan{-}Gavitt, Ramesh Karri, and Siddharth Garg.
\newblock Verigen: {A} large language model for verilog code generation.
\newblock \emph{CoRR}, abs/2308.00708, 2023{\natexlab{b}}.
\newblock \doi{10.48550/ARXIV.2308.00708}.
\newblock URL \url{https://doi.org/10.48550/arXiv.2308.00708}.

\bibitem[Thoppilan et~al.(2022)Thoppilan, Freitas, Hall, Shazeer, Kulshreshtha, Cheng, Jin, Bos, Baker, Du, Li, Lee, Zheng, Ghafouri, Menegali, Huang, Krikun, Lepikhin, Qin, Chen, Xu, Chen, Roberts, Bosma, Zhou, Chang, Krivokon, Rusch, Pickett, Meier{-}Hellstern, Morris, Doshi, Santos, Duke, Soraker, Zevenbergen, Prabhakaran, Diaz, Hutchinson, Olson, Molina, Hoffman{-}John, Lee, Aroyo, Rajakumar, Butryna, Lamm, Kuzmina, Fenton, Cohen, Bernstein, Kurzweil, Aguera{-}Arcas, Cui, Croak, Chi, and Le]{2022LaMDA}
Romal Thoppilan, Daniel~De Freitas, Jamie Hall, Noam Shazeer, Apoorv Kulshreshtha, Heng{-}Tze Cheng, Alicia Jin, Taylor Bos, Leslie Baker, Yu~Du, YaGuang Li, Hongrae Lee, Huaixiu~Steven Zheng, Amin Ghafouri, Marcelo Menegali, Yanping Huang, Maxim Krikun, Dmitry Lepikhin, James Qin, Dehao Chen, Yuanzhong Xu, Zhifeng Chen, Adam Roberts, Maarten Bosma, Yanqi Zhou, Chung{-}Ching Chang, Igor Krivokon, Will Rusch, Marc Pickett, Kathleen~S. Meier{-}Hellstern, Meredith~Ringel Morris, Tulsee Doshi, Renelito~Delos Santos, Toju Duke, Johnny Soraker, Ben Zevenbergen, Vinodkumar Prabhakaran, Mark Diaz, Ben Hutchinson, Kristen Olson, Alejandra Molina, Erin Hoffman{-}John, Josh Lee, Lora Aroyo, Ravi Rajakumar, Alena Butryna, Matthew Lamm, Viktoriya Kuzmina, Joe Fenton, Aaron Cohen, Rachel Bernstein, Ray Kurzweil, Blaise Aguera{-}Arcas, Claire Cui, Marian Croak, Ed~H. Chi, and Quoc Le.
\newblock Lamda: Language models for dialog applications.
\newblock \emph{CoRR}, abs/2201.08239, 2022.
\newblock URL \url{https://arxiv.org/abs/2201.08239}.

\bibitem[Thorat et~al.(2023)Thorat, Zhao, Liu, Peng, Xie, Lei, Zhang, and Ding]{2023Verilog-1}
Kiran Thorat, Jiahui Zhao, Yaotian Liu, Hongwu Peng, Xi~Xie, Bin Lei, Jeff Zhang, and Caiwen Ding.
\newblock Advanced large language model (llm)-driven verilog development: Enhancing power, performance, and area optimization in code synthesis.
\newblock \emph{CoRR}, abs/2312.01022, 2023.
\newblock \doi{10.48550/ARXIV.2312.01022}.
\newblock URL \url{https://doi.org/10.48550/arXiv.2312.01022}.

\bibitem[Tian et~al.(2024)Tian, Ye, Qin, Cong, Lin, Pan, Wu, Liu, and Sun]{2024DebugBench}
Runchu Tian, Yining Ye, Yujia Qin, Xin Cong, Yankai Lin, Yinxu Pan, Yesai Wu, Zhiyuan Liu, and Maosong Sun.
\newblock Debugbench: Evaluating debugging capability of large language models.
\newblock \emph{CoRR}, abs/2401.04621, 2024.
\newblock \doi{10.48550/ARXIV.2401.04621}.
\newblock URL \url{https://doi.org/10.48550/arXiv.2401.04621}.

\bibitem[Tiwari et~al.(2019)Tiwari, Ameta, and Banerjee]{re-ana2019-1}
Saurabh Tiwari, Deepti Ameta, and Asim Banerjee.
\newblock An approach to identify use case scenarios from textual requirements specification.
\newblock In Ravindra Naik, Santonu Sarkar, Thomas~T. Hildebrandt, Atul Kumar, and Richa Sharma (eds.), \emph{Proceedings of the 12th Innovations on Software Engineering Conference (formerly known as India Software Engineering Conference), {ISEC} 2019, Pune, India, February 14-16, 2019}, pp.\  5:1--5:11. {ACM}, 2019.
\newblock \doi{10.1145/3299771.3299774}.
\newblock URL \url{https://doi.org/10.1145/3299771.3299774}.

\bibitem[Tomassi et~al.(2019)Tomassi, Dmeiri, Wang, Bhowmick, Liu, Devanbu, Vasilescu, and Rubio{-}Gonz{\'{a}}lez]{fix-data-2019-4}
David~A. Tomassi, Naji Dmeiri, Yichen Wang, Antara Bhowmick, Yen{-}Chuan Liu, Premkumar~T. Devanbu, Bogdan Vasilescu, and Cindy Rubio{-}Gonz{\'{a}}lez.
\newblock Bugswarm: mining and continuously growing a dataset of reproducible failures and fixes.
\newblock In Joanne~M. Atlee, Tevfik Bultan, and Jon Whittle (eds.), \emph{Proceedings of the 41st International Conference on Software Engineering, {ICSE} 2019, Montreal, QC, Canada, May 25-31, 2019}, pp.\  339--349. {IEEE} / {ACM}, 2019.
\newblock \doi{10.1109/ICSE.2019.00048}.
\newblock URL \url{https://doi.org/10.1109/ICSE.2019.00048}.

\bibitem[Touvron et~al.(2023{\natexlab{a}})Touvron, Lavril, Izacard, Martinet, Lachaux, Lacroix, Rozi{\`{e}}re, Goyal, Hambro, Azhar, Rodriguez, Joulin, Grave, and Lample]{2023LLaMA}
Hugo Touvron, Thibaut Lavril, Gautier Izacard, Xavier Martinet, Marie{-}Anne Lachaux, Timoth{\'{e}}e Lacroix, Baptiste Rozi{\`{e}}re, Naman Goyal, Eric Hambro, Faisal Azhar, Aur{\'{e}}lien Rodriguez, Armand Joulin, Edouard Grave, and Guillaume Lample.
\newblock Llama: Open and efficient foundation language models.
\newblock \emph{CoRR}, abs/2302.13971, 2023{\natexlab{a}}.
\newblock \doi{10.48550/arXiv.2302.13971}.
\newblock URL \url{https://doi.org/10.48550/arXiv.2302.13971}.

\bibitem[Touvron et~al.(2023{\natexlab{b}})Touvron, Martin, Stone, Albert, Almahairi, Babaei, Bashlykov, Batra, Bhargava, Bhosale, Bikel, Blecher, Canton{-}Ferrer, Chen, Cucurull, Esiobu, Fernandes, Fu, Fu, Fuller, Gao, Goswami, Goyal, Hartshorn, Hosseini, Hou, Inan, Kardas, Kerkez, Khabsa, Kloumann, Korenev, Koura, Lachaux, Lavril, Lee, Liskovich, Lu, Mao, Martinet, Mihaylov, Mishra, Molybog, Nie, Poulton, Reizenstein, Rungta, Saladi, Schelten, Silva, Smith, Subramanian, Tan, Tang, Taylor, Williams, Kuan, Xu, Yan, Zarov, Zhang, Fan, Kambadur, Narang, Rodriguez, Stojnic, Edunov, and Scialom]{2023LLaMA2}
Hugo Touvron, Louis Martin, Kevin Stone, Peter Albert, Amjad Almahairi, Yasmine Babaei, Nikolay Bashlykov, Soumya Batra, Prajjwal Bhargava, Shruti Bhosale, Dan Bikel, Lukas Blecher, Cristian Canton{-}Ferrer, Moya Chen, Guillem Cucurull, David Esiobu, Jude Fernandes, Jeremy Fu, Wenyin Fu, Brian Fuller, Cynthia Gao, Vedanuj Goswami, Naman Goyal, Anthony Hartshorn, Saghar Hosseini, Rui Hou, Hakan Inan, Marcin Kardas, Viktor Kerkez, Madian Khabsa, Isabel Kloumann, Artem Korenev, Punit~Singh Koura, Marie{-}Anne Lachaux, Thibaut Lavril, Jenya Lee, Diana Liskovich, Yinghai Lu, Yuning Mao, Xavier Martinet, Todor Mihaylov, Pushkar Mishra, Igor Molybog, Yixin Nie, Andrew Poulton, Jeremy Reizenstein, Rashi Rungta, Kalyan Saladi, Alan Schelten, Ruan Silva, Eric~Michael Smith, Ranjan Subramanian, Xiaoqing~Ellen Tan, Binh Tang, Ross Taylor, Adina Williams, Jian~Xiang Kuan, Puxin Xu, Zheng Yan, Iliyan Zarov, Yuchen Zhang, Angela Fan, Melanie Kambadur, Sharan Narang, Aur{\'{e}}lien Rodriguez, Robert Stojnic, Sergey Edunov,
  and Thomas Scialom.
\newblock Llama 2: Open foundation and fine-tuned chat models.
\newblock \emph{CoRR}, abs/2307.09288, 2023{\natexlab{b}}.
\newblock \doi{10.48550/arXiv.2307.09288}.
\newblock URL \url{https://doi.org/10.48550/arXiv.2307.09288}.

\bibitem[Tran et~al.(2019)Tran, Tran, Nguyen, Nguyen, and Nguyen]{ob2019-1}
Hieu Tran, Ngoc~M. Tran, Son Nguyen, Hoan Nguyen, and Tien~N. Nguyen.
\newblock Recovering variable names for minified code with usage contexts.
\newblock In Joanne~M. Atlee, Tevfik Bultan, and Jon Whittle (eds.), \emph{Proceedings of the 41st International Conference on Software Engineering, {ICSE} 2019, Montreal, QC, Canada, May 25-31, 2019}, pp.\  1165--1175. {IEEE} / {ACM}, 2019.
\newblock \doi{10.1109/ICSE.2019.00119}.
\newblock URL \url{https://doi.org/10.1109/ICSE.2019.00119}.

\bibitem[Trummer(2022)]{sql2022-1}
Immanuel Trummer.
\newblock Codexdb: Synthesizing code for query processing from natural language instructions using {GPT-3} codex.
\newblock \emph{Proc. {VLDB} Endow.}, 15\penalty0 (11):\penalty0 2921--2928, 2022.
\newblock URL \url{https://www.vldb.org/pvldb/vol15/p2921-trummer.pdf}.

\bibitem[Tsai et~al.(2023)Tsai, Liu, and Ren]{2023RTLFixer}
Yun{-}Da Tsai, Mingjie Liu, and Haoxing Ren.
\newblock Rtlfixer: Automatically fixing {RTL} syntax errors with large language models.
\newblock \emph{CoRR}, abs/2311.16543, 2023.
\newblock \doi{10.48550/ARXIV.2311.16543}.
\newblock URL \url{https://doi.org/10.48550/arXiv.2311.16543}.

\bibitem[Tsfaty \& Fire(2022)Tsfaty and Fire]{2022MSDT}
Chen Tsfaty and Michael Fire.
\newblock Malicious source code detection using transformer.
\newblock \emph{CoRR}, abs/2209.07957, 2022.
\newblock \doi{10.48550/ARXIV.2209.07957}.
\newblock URL \url{https://doi.org/10.48550/arXiv.2209.07957}.

\bibitem[Tu et~al.(2014)Tu, Su, and Devanbu]{completion2014-3}
Zhaopeng Tu, Zhendong Su, and Premkumar~T. Devanbu.
\newblock On the localness of software.
\newblock In Shing{-}Chi Cheung, Alessandro Orso, and Margaret{-}Anne~D. Storey (eds.), \emph{Proceedings of the 22nd {ACM} {SIGSOFT} International Symposium on Foundations of Software Engineering, (FSE-22), Hong Kong, China, November 16 - 22, 2014}, pp.\  269--280. {ACM}, 2014.
\newblock \doi{10.1145/2635868.2635875}.
\newblock URL \url{https://doi.org/10.1145/2635868.2635875}.

\bibitem[Tufano et~al.(2019{\natexlab{a}})Tufano, Pantiuchina, Watson, Bavota, and Poshyvanyk]{fix2019-1}
Michele Tufano, Jevgenija Pantiuchina, Cody Watson, Gabriele Bavota, and Denys Poshyvanyk.
\newblock On learning meaningful code changes via neural machine translation.
\newblock In Joanne~M. Atlee, Tevfik Bultan, and Jon Whittle (eds.), \emph{Proceedings of the 41st International Conference on Software Engineering, {ICSE} 2019, Montreal, QC, Canada, May 25-31, 2019}, pp.\  25--36. {IEEE} / {ACM}, 2019{\natexlab{a}}.
\newblock \doi{10.1109/ICSE.2019.00021}.
\newblock URL \url{https://doi.org/10.1109/ICSE.2019.00021}.

\bibitem[Tufano et~al.(2019{\natexlab{b}})Tufano, Watson, Bavota, Penta, White, and Poshyvanyk]{fix2018-3}
Michele Tufano, Cody Watson, Gabriele Bavota, Massimiliano~Di Penta, Martin White, and Denys Poshyvanyk.
\newblock An empirical study on learning bug-fixing patches in the wild via neural machine translation.
\newblock \emph{{ACM} Trans. Softw. Eng. Methodol.}, 28\penalty0 (4):\penalty0 19:1--19:29, 2019{\natexlab{b}}.
\newblock \doi{10.1145/3340544}.
\newblock URL \url{https://doi.org/10.1145/3340544}.

\bibitem[Tufano et~al.(2019{\natexlab{c}})Tufano, Watson, Bavota, Penta, White, and Poshyvanyk]{mutant2018-2}
Michele Tufano, Cody Watson, Gabriele Bavota, Massimiliano~Di Penta, Martin White, and Denys Poshyvanyk.
\newblock Learning how to mutate source code from bug-fixes.
\newblock In \emph{2019 {IEEE} International Conference on Software Maintenance and Evolution, {ICSME} 2019, Cleveland, OH, USA, September 29 - October 4, 2019}, pp.\  301--312. {IEEE}, 2019{\natexlab{c}}.
\newblock \doi{10.1109/ICSME.2019.00046}.
\newblock URL \url{https://doi.org/10.1109/ICSME.2019.00046}.

\bibitem[Tufano et~al.(2020)Tufano, Kimko, Wang, Watson, Bavota, Penta, and Poshyvanyk]{mutant2020-1}
Michele Tufano, Jason Kimko, Shiya Wang, Cody Watson, Gabriele Bavota, Massimiliano~Di Penta, and Denys Poshyvanyk.
\newblock Deepmutation: a neural mutation tool.
\newblock In Gregg Rothermel and Doo{-}Hwan Bae (eds.), \emph{{ICSE} '20: 42nd International Conference on Software Engineering, Companion Volume, Seoul, South Korea, 27 June - 19 July, 2020}, pp.\  29--32. {ACM}, 2020.
\newblock \doi{10.1145/3377812.3382146}.
\newblock URL \url{https://doi.org/10.1145/3377812.3382146}.

\bibitem[Tufano et~al.(2021{\natexlab{a}})Tufano, Drain, Svyatkovskiy, Deng, and Sundaresan]{unit2020-1}
Michele Tufano, Dawn Drain, Alexey Svyatkovskiy, Shao~Kun Deng, and Neel Sundaresan.
\newblock Unit test case generation with transformers and focal context, 2021{\natexlab{a}}.

\bibitem[Tufano et~al.(2022{\natexlab{a}})Tufano, Drain, Svyatkovskiy, and Sundaresan]{assert2020-2}
Michele Tufano, Dawn Drain, Alexey Svyatkovskiy, and Neel Sundaresan.
\newblock Generating accurate assert statements for unit test cases using pretrained transformers.
\newblock In \emph{{IEEE/ACM} International Conference on Automation of Software Test, AST@ICSE 2022, Pittsburgh, PA, USA, May 21-22, 2022}, pp.\  54--64. {ACM/IEEE}, 2022{\natexlab{a}}.
\newblock \doi{10.1145/3524481.3527220}.
\newblock URL \url{https://doi.org/10.1145/3524481.3527220}.

\bibitem[Tufano et~al.(2021{\natexlab{b}})Tufano, Pascarella, Tufano, Poshyvanyk, and Bavota]{review2021-1}
Rosalia Tufano, Luca Pascarella, Michele Tufano, Denys Poshyvanyk, and Gabriele Bavota.
\newblock Towards automating code review activities.
\newblock In \emph{43rd {IEEE/ACM} International Conference on Software Engineering, {ICSE} 2021, Madrid, Spain, 22-30 May 2021}, pp.\  163--174. {IEEE}, 2021{\natexlab{b}}.
\newblock \doi{10.1109/ICSE43902.2021.00027}.
\newblock URL \url{https://doi.org/10.1109/ICSE43902.2021.00027}.

\bibitem[Tufano et~al.(2022{\natexlab{b}})Tufano, Masiero, Mastropaolo, Pascarella, Poshyvanyk, and Bavota]{review2022-1}
Rosalia Tufano, Simone Masiero, Antonio Mastropaolo, Luca Pascarella, Denys Poshyvanyk, and Gabriele Bavota.
\newblock Using pre-trained models to boost code review automation.
\newblock In \emph{44th {IEEE/ACM} 44th International Conference on Software Engineering, {ICSE} 2022, Pittsburgh, PA, USA, May 25-27, 2022}, pp.\  2291--2302. {ACM}, 2022{\natexlab{b}}.
\newblock \doi{10.1145/3510003.3510621}.
\newblock URL \url{https://doi.org/10.1145/3510003.3510621}.

\bibitem[Tunstall et~al.(2022)Tunstall, von Werra, and Wolf]{2022TransformersBook}
Lewis Tunstall, Leandro von Werra, and Thomas Wolf.
\newblock \emph{Natural Language Processing with Transformers: Building Language Applications with Hugging Face}.
\newblock O'Reilly Media, Incorporated, 2022.
\newblock ISBN 1098103246.
\newblock URL \url{https://books.google.ch/books?id=7hhyzgEACAAJ}.

\bibitem[Ullah et~al.(2023)Ullah, Han, Pujar, Pearce, Coskun, and Stringhini]{defect2023-5}
Saad Ullah, Mingji Han, Saurabh Pujar, Hammond Pearce, Ayse~K. Coskun, and Gianluca Stringhini.
\newblock Can large language models identify and reason about security vulnerabilities? not yet.
\newblock \emph{CoRR}, abs/2312.12575, 2023.
\newblock \doi{10.48550/ARXIV.2312.12575}.
\newblock URL \url{https://doi.org/10.48550/arXiv.2312.12575}.

\bibitem[van Dam et~al.(2024)van Dam, van~der Heijden, de~Bekker, Nieuwschepen, Otten, and Izadi]{2024Haskel}
Tim van Dam, Frank van~der Heijden, Philippe de~Bekker, Berend Nieuwschepen, Marc Otten, and Maliheh Izadi.
\newblock Investigating the performance of language models for completing code in functional programming languages: a haskell case study.
\newblock \emph{CoRR}, abs/2403.15185, 2024.
\newblock \doi{10.48550/ARXIV.2403.15185}.
\newblock URL \url{https://doi.org/10.48550/arXiv.2403.15185}.

\bibitem[Vasic et~al.(2019)Vasic, Kanade, Maniatis, Bieber, and Singh]{fix2019-2}
Marko Vasic, Aditya Kanade, Petros Maniatis, David Bieber, and Rishabh Singh.
\newblock Neural program repair by jointly learning to localize and repair.
\newblock In \emph{7th International Conference on Learning Representations, {ICLR} 2019, New Orleans, LA, USA, May 6-9, 2019}. OpenReview.net, 2019.
\newblock URL \url{https://openreview.net/forum?id=ByloJ20qtm}.

\bibitem[Vasilescu et~al.(2017)Vasilescu, Casalnuovo, and Devanbu]{ob2017-1}
Bogdan Vasilescu, Casey Casalnuovo, and Premkumar~T. Devanbu.
\newblock Recovering clear, natural identifiers from obfuscated {JS} names.
\newblock In Eric Bodden, Wilhelm Sch{\"{a}}fer, Arie van Deursen, and Andrea Zisman (eds.), \emph{Proceedings of the 2017 11th Joint Meeting on Foundations of Software Engineering, {ESEC/FSE} 2017, Paderborn, Germany, September 4-8, 2017}, pp.\  683--693. {ACM}, 2017.
\newblock \doi{10.1145/3106237.3106289}.
\newblock URL \url{https://doi.org/10.1145/3106237.3106289}.

\bibitem[V{\'{a}}squez et~al.(2015)V{\'{a}}squez, Cortes{-}Coy, Aponte, and Poshyvanyk]{commit2015-1}
Mario~Linares V{\'{a}}squez, Luis~Fernando Cortes{-}Coy, Jairo Aponte, and Denys Poshyvanyk.
\newblock Changescribe: {A} tool for automatically generating commit messages.
\newblock In Antonia Bertolino, Gerardo Canfora, and Sebastian~G. Elbaum (eds.), \emph{37th {IEEE/ACM} International Conference on Software Engineering, {ICSE} 2015, Florence, Italy, May 16-24, 2015, Volume 2}, pp.\  709--712. {IEEE} Computer Society, 2015.
\newblock \doi{10.1109/ICSE.2015.229}.
\newblock URL \url{https://doi.org/10.1109/ICSE.2015.229}.

\bibitem[Vaswani et~al.(2017)Vaswani, Shazeer, Parmar, Uszkoreit, Jones, Gomez, Kaiser, and Polosukhin]{2017Transformer}
Ashish Vaswani, Noam Shazeer, Niki Parmar, Jakob Uszkoreit, Llion Jones, Aidan~N. Gomez, Lukasz Kaiser, and Illia Polosukhin.
\newblock Attention is all you need.
\newblock In Isabelle Guyon, Ulrike von Luxburg, Samy Bengio, Hanna~M. Wallach, Rob Fergus, S.~V.~N. Vishwanathan, and Roman Garnett (eds.), \emph{Advances in Neural Information Processing Systems 30: Annual Conference on Neural Information Processing Systems 2017, December 4-9, 2017, Long Beach, CA, {USA}}, pp.\  5998--6008, 2017.
\newblock URL \url{https://proceedings.neurips.cc/paper/2017/hash/3f5ee243547dee91fbd053c1c4a845aa-Abstract.html}.

\bibitem[Vikram et~al.(2023)Vikram, Lemieux, and Padhye]{2023PBT-GPT}
Vasudev Vikram, Caroline Lemieux, and Rohan Padhye.
\newblock Can large language models write good property-based tests?
\newblock \emph{CoRR}, abs/2307.04346, 2023.
\newblock \doi{10.48550/ARXIV.2307.04346}.
\newblock URL \url{https://doi.org/10.48550/arXiv.2307.04346}.

\bibitem[Villmow et~al.(2021)Villmow, Depoix, and Ulges]{assert-data-2021-1}
Johannes Villmow, Jonas Depoix, and Adrian Ulges.
\newblock {C}on{T}est: A unit test completion benchmark featuring context.
\newblock In Royi Lachmy, Ziyu Yao, Greg Durrett, Milos Gligoric, Junyi~Jessy Li, Ray Mooney, Graham Neubig, Yu~Su, Huan Sun, and Reut Tsarfaty (eds.), \emph{Proceedings of the 1st Workshop on Natural Language Processing for Programming (NLP4Prog 2021)}, pp.\  17--25, Online, August 2021. Association for Computational Linguistics.
\newblock \doi{10.18653/v1/2021.nlp4prog-1.2}.
\newblock URL \url{https://aclanthology.org/2021.nlp4prog-1.2}.

\bibitem[Wan et~al.(2018)Wan, Zhao, Yang, Xu, Ying, Wu, and Yu]{sum2018-3}
Yao Wan, Zhou Zhao, Min Yang, Guandong Xu, Haochao Ying, Jian Wu, and Philip~S. Yu.
\newblock Improving automatic source code summarization via deep reinforcement learning.
\newblock In Marianne Huchard, Christian K{\"{a}}stner, and Gordon Fraser (eds.), \emph{Proceedings of the 33rd {ACM/IEEE} International Conference on Automated Software Engineering, {ASE} 2018, Montpellier, France, September 3-7, 2018}, pp.\  397--407. {ACM}, 2018.
\newblock \doi{10.1145/3238147.3238206}.
\newblock URL \url{https://doi.org/10.1145/3238147.3238206}.

\bibitem[Wan et~al.(2024)Wan, Wan, Zhang, Zhang, Sui, Zhou, Jin, and Sun]{ana-leak2024-1}
Yao Wan, Guanghua Wan, Shijie Zhang, Hongyu Zhang, Yulei Sui, Pan Zhou, Hai Jin, and Lichao Sun.
\newblock Does your neural code completion model use my code? a membership inference approach.
\newblock \emph{CoRR}, abs/2404.14296, 2024.
\newblock \doi{10.48550/ARXIV.2404.14296}.
\newblock URL \url{https://doi.org/10.48550/arXiv.2404.14296}.

\bibitem[Wang et~al.(2018{\natexlab{a}})Wang, Singh, Michael, Hill, Levy, and Bowman]{2018GLUE}
Alex Wang, Amanpreet Singh, Julian Michael, Felix Hill, Omer Levy, and Samuel~R. Bowman.
\newblock {GLUE:} {A} multi-task benchmark and analysis platform for natural language understanding.
\newblock In Tal Linzen, Grzegorz Chrupala, and Afra Alishahi (eds.), \emph{Proceedings of the Workshop: Analyzing and Interpreting Neural Networks for NLP, BlackboxNLP@EMNLP 2018, Brussels, Belgium, November 1, 2018}, pp.\  353--355. Association for Computational Linguistics, 2018{\natexlab{a}}.
\newblock \doi{10.18653/v1/w18-5446}.
\newblock URL \url{https://doi.org/10.18653/v1/w18-5446}.

\bibitem[Wang et~al.(2019)Wang, Pruksachatkun, Nangia, Singh, Michael, Hill, Levy, and Bowman]{2019SuperGLUE}
Alex Wang, Yada Pruksachatkun, Nikita Nangia, Amanpreet Singh, Julian Michael, Felix Hill, Omer Levy, and Samuel~R. Bowman.
\newblock Superglue: {A} stickier benchmark for general-purpose language understanding systems.
\newblock In Hanna~M. Wallach, Hugo Larochelle, Alina Beygelzimer, Florence d'Alch{\'{e}}{-}Buc, Emily~B. Fox, and Roman Garnett (eds.), \emph{Advances in Neural Information Processing Systems 32: Annual Conference on Neural Information Processing Systems 2019, NeurIPS 2019, December 8-14, 2019, Vancouver, BC, Canada}, pp.\  3261--3275, 2019.
\newblock URL \url{https://proceedings.neurips.cc/paper/2019/hash/4496bf24afe7fab6f046bf4923da8de6-Abstract.html}.

\bibitem[Wang et~al.(2020{\natexlab{a}})Wang, Shin, Liu, Polozov, and Richardson]{sql2019-7}
Bailin Wang, Richard Shin, Xiaodong Liu, Oleksandr Polozov, and Matthew Richardson.
\newblock {RAT-SQL:} relation-aware schema encoding and linking for text-to-sql parsers.
\newblock In Dan Jurafsky, Joyce Chai, Natalie Schluter, and Joel~R. Tetreault (eds.), \emph{Proceedings of the 58th Annual Meeting of the Association for Computational Linguistics, {ACL} 2020, Online, July 5-10, 2020}, pp.\  7567--7578. Association for Computational Linguistics, 2020{\natexlab{a}}.
\newblock \doi{10.18653/v1/2020.acl-main.677}.
\newblock URL \url{https://doi.org/10.18653/v1/2020.acl-main.677}.

\bibitem[Wang et~al.(2021{\natexlab{a}})Wang, Yin, Lin, and Xiong]{sql2021-0.7}
Bailin Wang, Wenpeng Yin, Xi~Victoria Lin, and Caiming Xiong.
\newblock Learning to synthesize data for semantic parsing.
\newblock In Kristina Toutanova, Anna Rumshisky, Luke Zettlemoyer, Dilek Hakkani{-}T{\"{u}}r, Iz~Beltagy, Steven Bethard, Ryan Cotterell, Tanmoy Chakraborty, and Yichao Zhou (eds.), \emph{Proceedings of the 2021 Conference of the North American Chapter of the Association for Computational Linguistics: Human Language Technologies, {NAACL-HLT} 2021, Online, June 6-11, 2021}, pp.\  2760--2766. Association for Computational Linguistics, 2021{\natexlab{a}}.
\newblock \doi{10.18653/V1/2021.NAACL-MAIN.220}.
\newblock URL \url{https://doi.org/10.18653/v1/2021.naacl-main.220}.

\bibitem[Wang \& Komatsuzaki(2021)Wang and Komatsuzaki]{2021GPT-J}
Ben Wang and Aran Komatsuzaki.
\newblock {GPT-J-6B: A 6 Billion Parameter Autoregressive Language Model}.
\newblock \url{https://github.com/kingoflolz/mesh-transformer-jax}, May 2021.

\bibitem[Wang et~al.(2021{\natexlab{b}})Wang, Li, Zhou, Chen, Grossman, and Li]{UI2021-2}
Bryan Wang, Gang Li, Xin Zhou, Zhourong Chen, Tovi Grossman, and Yang Li.
\newblock Screen2words: Automatic mobile {UI} summarization with multimodal learning.
\newblock In Jeffrey Nichols, Ranjitha Kumar, and Michael Nebeling (eds.), \emph{{UIST} '21: The 34th Annual {ACM} Symposium on User Interface Software and Technology, Virtual Event, USA, October 10-14, 2021}, pp.\  498--510. {ACM}, 2021{\natexlab{b}}.
\newblock \doi{10.1145/3472749.3474765}.
\newblock URL \url{https://doi.org/10.1145/3472749.3474765}.

\bibitem[Wang et~al.(2023{\natexlab{a}})Wang, Li, and Li]{UI2022-4}
Bryan Wang, Gang Li, and Yang Li.
\newblock Enabling conversational interaction with mobile {UI} using large language models.
\newblock In Albrecht Schmidt, Kaisa V{\"{a}}{\"{a}}n{\"{a}}nen, Tesh Goyal, Per~Ola Kristensson, Anicia Peters, Stefanie Mueller, Julie~R. Williamson, and Max~L. Wilson (eds.), \emph{Proceedings of the 2023 {CHI} Conference on Human Factors in Computing Systems, {CHI} 2023, Hamburg, Germany, April 23-28, 2023}, pp.\  432:1--432:17. {ACM}, 2023{\natexlab{a}}.
\newblock \doi{10.1145/3544548.3580895}.
\newblock URL \url{https://doi.org/10.1145/3544548.3580895}.

\bibitem[Wang et~al.(2018{\natexlab{b}})Wang, Pastore, and Briand]{re-ana2018-1}
Chunhui Wang, Fabrizio Pastore, and Lionel~C. Briand.
\newblock Automated generation of constraints from use case specifications to support system testing.
\newblock In \emph{11th {IEEE} International Conference on Software Testing, Verification and Validation, {ICST} 2018, V{\"{a}}ster{\aa}s, Sweden, April 9-13, 2018}, pp.\  23--33. {IEEE} Computer Society, 2018{\natexlab{b}}.
\newblock \doi{10.1109/ICST.2018.00013}.
\newblock URL \url{https://doi.ieeecomputersociety.org/10.1109/ICST.2018.00013}.

\bibitem[Wang et~al.(2022{\natexlab{a}})Wang, Jia, Li, Yu, Xiong, Dong, and Liao]{retrieval2021-7}
Deze Wang, Zhouyang Jia, Shanshan Li, Yue Yu, Yun Xiong, Wei Dong, and Xiangke Liao.
\newblock Bridging pre-trained models and downstream tasks for source code understanding.
\newblock In \emph{44th {IEEE/ACM} 44th International Conference on Software Engineering, {ICSE} 2022, Pittsburgh, PA, USA, May 25-27, 2022}, pp.\  287--298. {ACM}, 2022{\natexlab{a}}.
\newblock \doi{10.1145/3510003.3510062}.
\newblock URL \url{https://doi.org/10.1145/3510003.3510062}.

\bibitem[Wang et~al.(2021{\natexlab{c}})Wang, Xia, Lo, He, Wang, and Grundy]{commit-data-2021-1}
Haoye Wang, Xin Xia, David Lo, Qiang He, Xinyu Wang, and John Grundy.
\newblock Context-aware retrieval-based deep commit message generation.
\newblock \emph{{ACM} Trans. Softw. Eng. Methodol.}, 30\penalty0 (4):\penalty0 56:1--56:30, 2021{\natexlab{c}}.
\newblock \doi{10.1145/3464689}.
\newblock URL \url{https://doi.org/10.1145/3464689}.

\bibitem[Wang et~al.(2023{\natexlab{b}})Wang, Huang, Chen, Liu, Wang, and Wang]{test-survey-2023}
Junjie Wang, Yuchao Huang, Chunyang Chen, Zhe Liu, Song Wang, and Qing Wang.
\newblock Software testing with large language models: Survey, landscape, and vision.
\newblock \emph{CoRR}, abs/2307.07221, 2023{\natexlab{b}}.
\newblock \doi{10.48550/ARXIV.2307.07221}.
\newblock URL \url{https://doi.org/10.48550/arXiv.2307.07221}.

\bibitem[Wang \& Su(2020)Wang and Su]{2019LiGer}
Ke~Wang and Zhendong Su.
\newblock Blended, precise semantic program embeddings.
\newblock In \emph{Proceedings of the 41st ACM SIGPLAN Conference on Programming Language Design and Implementation}, PLDI 2020, pp.\  121–134, New York, NY, USA, 2020. Association for Computing Machinery.
\newblock ISBN 9781450376136.
\newblock \doi{10.1145/3385412.3385999}.
\newblock URL \url{https://doi.org/10.1145/3385412.3385999}.

\bibitem[Wang et~al.(2023{\natexlab{c}})Wang, Tang, He, Ren, Shi, Yan, and Li]{commit-data-2023-1}
Liran Wang, Xunzhu Tang, Yichen He, Changyu Ren, Shuhua Shi, Chaoran Yan, and Zhoujun Li.
\newblock Delving into commit-issue correlation to enhance commit message generation models.
\newblock In \emph{38th {IEEE/ACM} International Conference on Automated Software Engineering, {ASE} 2023, Luxembourg, September 11-15, 2023}, pp.\  710--722. {IEEE}, 2023{\natexlab{c}}.
\newblock \doi{10.1109/ASE56229.2023.00050}.
\newblock URL \url{https://doi.org/10.1109/ASE56229.2023.00050}.

\bibitem[Wang et~al.(2018{\natexlab{c}})Wang, Svajlenko, Wu, Xu, and Roy]{clone2018-1}
Pengcheng Wang, Jeffrey Svajlenko, Yanzhao Wu, Yun Xu, and Chanchal~K. Roy.
\newblock Ccaligner: a token based large-gap clone detector.
\newblock In Michel Chaudron, Ivica Crnkovic, Marsha Chechik, and Mark Harman (eds.), \emph{Proceedings of the 40th International Conference on Software Engineering, {ICSE} 2018, Gothenburg, Sweden, May 27 - June 03, 2018}, pp.\  1066--1077. {ACM}, 2018{\natexlab{c}}.
\newblock \doi{10.1145/3180155.3180179}.
\newblock URL \url{https://doi.org/10.1145/3180155.3180179}.

\bibitem[Wang et~al.(2020{\natexlab{b}})Wang, Shi, and Reddy]{sql-data-2019-1}
Ping Wang, Tian Shi, and Chandan~K. Reddy.
\newblock Text-to-sql generation for question answering on electronic medical records.
\newblock In Yennun Huang, Irwin King, Tie{-}Yan Liu, and Maarten van Steen (eds.), \emph{{WWW} '20: The Web Conference 2020, Taipei, Taiwan, April 20-24, 2020}, pp.\  350--361. {ACM} / {IW3C2}, 2020{\natexlab{b}}.
\newblock \doi{10.1145/3366423.3380120}.
\newblock URL \url{https://doi.org/10.1145/3366423.3380120}.

\bibitem[Wang et~al.(2023{\natexlab{d}})Wang, Madaio, Kane, Kapania, Terry, and Wilcox]{UI-2023-2}
Qiaosi Wang, Michael Madaio, Shaun~K. Kane, Shivani Kapania, Michael Terry, and Lauren Wilcox.
\newblock Designing responsible {AI:} adaptations of {UX} practice to meet responsible {AI} challenges.
\newblock In Albrecht Schmidt, Kaisa V{\"{a}}{\"{a}}n{\"{a}}nen, Tesh Goyal, Per~Ola Kristensson, Anicia Peters, Stefanie Mueller, Julie~R. Williamson, and Max~L. Wilson (eds.), \emph{Proceedings of the 2023 {CHI} Conference on Human Factors in Computing Systems, {CHI} 2023, Hamburg, Germany, April 23-28, 2023}, pp.\  249:1--249:16. {ACM}, 2023{\natexlab{d}}.
\newblock \doi{10.1145/3544548.3581278}.
\newblock URL \url{https://doi.org/10.1145/3544548.3581278}.

\bibitem[Wang et~al.(2022{\natexlab{b}})Wang, Fang, Ravula, Feng, Quan, and Liu]{ui2022-2}
Qifan Wang, Yi~Fang, Anirudh Ravula, Fuli Feng, Xiaojun Quan, and Dongfang Liu.
\newblock Webformer: The web-page transformer for structure information extraction.
\newblock In Fr{\'{e}}d{\'{e}}rique Laforest, Rapha{\"{e}}l Troncy, Elena Simperl, Deepak Agarwal, Aristides Gionis, Ivan Herman, and Lionel M{\'{e}}dini (eds.), \emph{{WWW} '22: The {ACM} Web Conference 2022, Virtual Event, Lyon, France, April 25 - 29, 2022}, pp.\  3124--3133. {ACM}, 2022{\natexlab{b}}.
\newblock \doi{10.1145/3485447.3512032}.
\newblock URL \url{https://doi.org/10.1145/3485447.3512032}.

\bibitem[Wang et~al.(2023{\natexlab{e}})Wang, Wen, Lin, Liu, Bissyand\'{e}, and Mao]{id2023-1}
Shangwen Wang, Ming Wen, Bo~Lin, Yepang Liu, Tegawend\'{e}~F. Bissyand\'{e}, and Xiaoguang Mao.
\newblock Pre-implementation method name prediction for object-oriented programming.
\newblock \emph{ACM Trans. Softw. Eng. Methodol.}, 32\penalty0 (6), sep 2023{\natexlab{e}}.
\newblock ISSN 1049-331X.
\newblock \doi{10.1145/3597203}.
\newblock URL \url{https://doi.org/10.1145/3597203}.

\bibitem[Wang et~al.(2020{\natexlab{c}})Wang, Li, Khabsa, Fang, and Ma]{2020Linformer}
Sinong Wang, Belinda~Z. Li, Madian Khabsa, Han Fang, and Hao Ma.
\newblock Linformer: Self-attention with linear complexity.
\newblock \emph{CoRR}, abs/2006.04768, 2020{\natexlab{c}}.
\newblock URL \url{https://arxiv.org/abs/2006.04768}.

\bibitem[Wang et~al.(2016{\natexlab{a}})Wang, Chollak, Movshovitz{-}Attias, and Tan]{defect2016-2}
Song Wang, Devin Chollak, Dana Movshovitz{-}Attias, and Lin Tan.
\newblock Bugram: bug detection with n-gram language models.
\newblock In David Lo, Sven Apel, and Sarfraz Khurshid (eds.), \emph{Proceedings of the 31st {IEEE/ACM} International Conference on Automated Software Engineering, {ASE} 2016, Singapore, September 3-7, 2016}, pp.\  708--719. {ACM}, 2016{\natexlab{a}}.
\newblock \doi{10.1145/2970276.2970341}.
\newblock URL \url{https://doi.org/10.1145/2970276.2970341}.

\bibitem[Wang et~al.(2016{\natexlab{b}})Wang, Liu, and Tan]{defect2016-1}
Song Wang, Taiyue Liu, and Lin Tan.
\newblock Automatically learning semantic features for defect prediction.
\newblock In Laura~K. Dillon, Willem Visser, and Laurie~A. Williams (eds.), \emph{Proceedings of the 38th International Conference on Software Engineering, {ICSE} 2016, Austin, TX, USA, May 14-22, 2016}, pp.\  297--308. {ACM}, 2016{\natexlab{b}}.
\newblock \doi{10.1145/2884781.2884804}.
\newblock URL \url{https://doi.org/10.1145/2884781.2884804}.

\bibitem[Wang et~al.(2022{\natexlab{c}})Wang, Roberts, Hesslow, Scao, Chung, Beltagy, Launay, and Raffel]{20220-shot}
Thomas Wang, Adam Roberts, Daniel Hesslow, Teven~Le Scao, Hyung~Won Chung, Iz~Beltagy, Julien Launay, and Colin Raffel.
\newblock What language model architecture and pretraining objective works best for zero-shot generalization?
\newblock In Kamalika Chaudhuri, Stefanie Jegelka, Le~Song, Csaba Szepesv{\'{a}}ri, Gang Niu, and Sivan Sabato (eds.), \emph{International Conference on Machine Learning, {ICML} 2022, 17-23 July 2022, Baltimore, Maryland, {USA}}, volume 162 of \emph{Proceedings of Machine Learning Research}, pp.\  22964--22984. {PMLR}, 2022{\natexlab{c}}.
\newblock URL \url{https://proceedings.mlr.press/v162/wang22u.html}.

\bibitem[Wang et~al.(2020{\natexlab{d}})Wang, Li, Ma, Xia, and Jin]{clone2020-1}
Wenhan Wang, Ge~Li, Bo~Ma, Xin Xia, and Zhi Jin.
\newblock Detecting code clones with graph neural network and flow-augmented abstract syntax tree.
\newblock In Kostas Kontogiannis, Foutse Khomh, Alexander Chatzigeorgiou, Marios{-}Eleftherios Fokaefs, and Minghui Zhou (eds.), \emph{27th {IEEE} International Conference on Software Analysis, Evolution and Reengineering, {SANER} 2020, London, ON, Canada, February 18-21, 2020}, pp.\  261--271. {IEEE}, 2020{\natexlab{d}}.
\newblock \doi{10.1109/SANER48275.2020.9054857}.
\newblock URL \url{https://doi.org/10.1109/SANER48275.2020.9054857}.

\bibitem[Wang et~al.(2020{\natexlab{e}})Wang, Zhang, Zeng, and Xu]{2020Trans3}
Wenhua Wang, Yuqun Zhang, Zhengran Zeng, and Guandong Xu.
\newblock Trans{\^{}}3: {A} transformer-based framework for unifying code summarization and code search.
\newblock \emph{CoRR}, abs/2003.03238, 2020{\natexlab{e}}.
\newblock URL \url{https://arxiv.org/abs/2003.03238}.

\bibitem[Wang et~al.(2021{\natexlab{d}})Wang, Wang, Mi, Zhou, Wan, Liu, Li, Wu, Liu, and Jiang]{2021SynCoBERT}
Xin Wang, Yasheng Wang, Fei Mi, Pingyi Zhou, Yao Wan, Xiao Liu, Li~Li, Hao Wu, Jin Liu, and Xin Jiang.
\newblock Syncobert: Syntax-guided multi-modal contrastive pre-training for code representation, 2021{\natexlab{d}}.

\bibitem[Wang et~al.(2022{\natexlab{d}})Wang, Wang, Wan, Mi, Li, Zhou, Liu, Wu, Jiang, and Liu]{2022CompCoder}
Xin Wang, Yasheng Wang, Yao Wan, Fei Mi, Yitong Li, Pingyi Zhou, Jin Liu, Hao Wu, Xin Jiang, and Qun Liu.
\newblock Compilable neural code generation with compiler feedback.
\newblock In Smaranda Muresan, Preslav Nakov, and Aline Villavicencio (eds.), \emph{Findings of the Association for Computational Linguistics: {ACL} 2022, Dublin, Ireland, May 22-27, 2022}, pp.\  9--19. Association for Computational Linguistics, 2022{\natexlab{d}}.
\newblock \doi{10.18653/V1/2022.FINDINGS-ACL.2}.
\newblock URL \url{https://doi.org/10.18653/v1/2022.findings-acl.2}.

\bibitem[Wang et~al.(2022{\natexlab{e}})Wang, Wang, Wan, Wang, Zhou, Li, Wu, and Liu]{2022Code-MVP}
Xin Wang, Yasheng Wang, Yao Wan, Jiawei Wang, Pingyi Zhou, Li~Li, Hao Wu, and Jin Liu.
\newblock {CODE-MVP:} learning to represent source code from multiple views with contrastive pre-training.
\newblock In Marine Carpuat, Marie{-}Catherine de~Marneffe, and Iv{\'{a}}n Vladimir~Meza Ru{\'{\i}}z (eds.), \emph{Findings of the Association for Computational Linguistics: {NAACL} 2022, Seattle, WA, United States, July 10-15, 2022}, pp.\  1066--1077. Association for Computational Linguistics, 2022{\natexlab{e}}.
\newblock \doi{10.18653/v1/2022.findings-naacl.80}.
\newblock URL \url{https://doi.org/10.18653/v1/2022.findings-naacl.80}.

\bibitem[Wang et~al.(2023{\natexlab{f}})Wang, Peng, Jabbarvand, and Ji]{2023LeTI}
Xingyao Wang, Hao Peng, Reyhaneh Jabbarvand, and Heng Ji.
\newblock Leti: Learning to generate from textual interactions.
\newblock \emph{CoRR}, abs/2305.10314, 2023{\natexlab{f}}.
\newblock \doi{10.48550/ARXIV.2305.10314}.
\newblock URL \url{https://doi.org/10.48550/arXiv.2305.10314}.

\bibitem[Wang et~al.(2022{\natexlab{f}})Wang, Zhang, Li, He, Zhang, Liu, Zheng, Kang, Lin, Dang, Rajmohan, and Zhang]{log2022-2}
Xuheng Wang, Xu~Zhang, Liqun Li, Shilin He, Hongyu Zhang, Yudong Liu, Lingling Zheng, Yu~Kang, Qingwei Lin, Yingnong Dang, Saravanakumar Rajmohan, and Dongmei Zhang.
\newblock {SPINE:} a scalable log parser with feedback guidance.
\newblock In Abhik Roychoudhury, Cristian Cadar, and Miryung Kim (eds.), \emph{Proceedings of the 30th {ACM} Joint European Software Engineering Conference and Symposium on the Foundations of Software Engineering, {ESEC/FSE} 2022, Singapore, Singapore, November 14-18, 2022}, pp.\  1198--1208. {ACM}, 2022{\natexlab{f}}.
\newblock \doi{10.1145/3540250.3549176}.
\newblock URL \url{https://doi.org/10.1145/3540250.3549176}.

\bibitem[Wang et~al.(2021{\natexlab{e}})Wang, Shi, Du, Yang, Hu, Han, Zhang, and Zhang]{sum2021-1}
Yanlin Wang, Ensheng Shi, Lun Du, Xiaodi Yang, Yuxuan Hu, Shi Han, Hongyu Zhang, and Dongmei Zhang.
\newblock Cocosum: Contextual code summarization with multi-relational graph neural network.
\newblock \emph{CoRR}, abs/2107.01933, 2021{\natexlab{e}}.
\newblock URL \url{https://arxiv.org/abs/2107.01933}.

\bibitem[Wang et~al.(2022{\natexlab{g}})Wang, Mishra, Alipoormolabashi, Kordi, Mirzaei, Naik, Ashok, Dhanasekaran, Arunkumar, Stap, Pathak, Karamanolakis, Lai, Purohit, Mondal, Anderson, Kuznia, Doshi, Pal, Patel, Moradshahi, Parmar, Purohit, Varshney, Kaza, Verma, Puri, Karia, Doshi, Sampat, Mishra, A, Patro, Dixit, and Shen]{2022SuperNatural}
Yizhong Wang, Swaroop Mishra, Pegah Alipoormolabashi, Yeganeh Kordi, Amirreza Mirzaei, Atharva Naik, Arjun Ashok, Arut~Selvan Dhanasekaran, Anjana Arunkumar, David Stap, Eshaan Pathak, Giannis Karamanolakis, Haizhi~Gary Lai, Ishan Purohit, Ishani Mondal, Jacob Anderson, Kirby Kuznia, Krima Doshi, Kuntal~Kumar Pal, Maitreya Patel, Mehrad Moradshahi, Mihir Parmar, Mirali Purohit, Neeraj Varshney, Phani~Rohitha Kaza, Pulkit Verma, Ravsehaj~Singh Puri, Rushang Karia, Savan Doshi, Shailaja~Keyur Sampat, Siddhartha Mishra, Sujan~Reddy A, Sumanta Patro, Tanay Dixit, and Xudong Shen.
\newblock Super-naturalinstructions: Generalization via declarative instructions on 1600+ {NLP} tasks.
\newblock In Yoav Goldberg, Zornitsa Kozareva, and Yue Zhang (eds.), \emph{Proceedings of the 2022 Conference on Empirical Methods in Natural Language Processing, {EMNLP} 2022, Abu Dhabi, United Arab Emirates, December 7-11, 2022}, pp.\  5085--5109. Association for Computational Linguistics, 2022{\natexlab{g}}.
\newblock URL \url{https://aclanthology.org/2022.emnlp-main.340}.

\bibitem[Wang et~al.(2023{\natexlab{g}})Wang, Kordi, Mishra, Liu, Smith, Khashabi, and Hajishirzi]{2022Self-Instruct}
Yizhong Wang, Yeganeh Kordi, Swaroop Mishra, Alisa Liu, Noah~A. Smith, Daniel Khashabi, and Hannaneh Hajishirzi.
\newblock Self-instruct: Aligning language models with self-generated instructions.
\newblock In Anna Rogers, Jordan~L. Boyd{-}Graber, and Naoaki Okazaki (eds.), \emph{Proceedings of the 61st Annual Meeting of the Association for Computational Linguistics (Volume 1: Long Papers), {ACL} 2023, Toronto, Canada, July 9-14, 2023}, pp.\  13484--13508. Association for Computational Linguistics, 2023{\natexlab{g}}.
\newblock \doi{10.18653/v1/2023.acl-long.754}.
\newblock URL \url{https://doi.org/10.18653/v1/2023.acl-long.754}.

\bibitem[Wang et~al.(2021{\natexlab{f}})Wang, Wang, Joty, and Hoi]{2021CodeT5}
Yue Wang, Weishi Wang, Shafiq~R. Joty, and Steven C.~H. Hoi.
\newblock Codet5: Identifier-aware unified pre-trained encoder-decoder models for code understanding and generation.
\newblock In Marie{-}Francine Moens, Xuanjing Huang, Lucia Specia, and Scott~Wen{-}tau Yih (eds.), \emph{Proceedings of the 2021 Conference on Empirical Methods in Natural Language Processing, {EMNLP} 2021, Virtual Event / Punta Cana, Dominican Republic, 7-11 November, 2021}, pp.\  8696--8708. Association for Computational Linguistics, 2021{\natexlab{f}}.
\newblock \doi{10.18653/v1/2021.emnlp-main.685}.
\newblock URL \url{https://doi.org/10.18653/v1/2021.emnlp-main.685}.

\bibitem[Wang et~al.(2023{\natexlab{h}})Wang, Le, Gotmare, Bui, Li, and Hoi]{2023CodeT5+}
Yue Wang, Hung Le, Akhilesh~Deepak Gotmare, Nghi D.~Q. Bui, Junnan Li, and Steven C.~H. Hoi.
\newblock Codet5+: Open code large language models for code understanding and generation.
\newblock \emph{CoRR}, abs/2305.07922, 2023{\natexlab{h}}.
\newblock \doi{10.48550/arXiv.2305.07922}.
\newblock URL \url{https://doi.org/10.48550/arXiv.2305.07922}.

\bibitem[Wang et~al.(2020{\natexlab{f}})Wang, Yan, Chen, Liu, and Zhang]{fuzz2020-3}
Zan Wang, Ming Yan, Junjie Chen, Shuang Liu, and Dongdi Zhang.
\newblock Deep learning library testing via effective model generation.
\newblock In Prem Devanbu, Myra~B. Cohen, and Thomas Zimmermann (eds.), \emph{{ESEC/FSE} '20: 28th {ACM} Joint European Software Engineering Conference and Symposium on the Foundations of Software Engineering, Virtual Event, USA, November 8-13, 2020}, pp.\  788--799. {ACM}, 2020{\natexlab{f}}.
\newblock \doi{10.1145/3368089.3409761}.
\newblock URL \url{https://doi.org/10.1145/3368089.3409761}.

\bibitem[Wang \& O'Boyle(2018)Wang and O'Boyle]{comp-opt-survey2018-1}
Zheng Wang and Michael F.~P. O'Boyle.
\newblock Machine learning in compiler optimisation.
\newblock \emph{CoRR}, abs/1805.03441, 2018.
\newblock URL \url{http://arxiv.org/abs/1805.03441}.

\bibitem[Wang et~al.(2022{\natexlab{h}})Wang, Zhou, Fried, and Neubig]{2022ODEX}
Zhiruo Wang, Shuyan Zhou, Daniel Fried, and Graham Neubig.
\newblock Execution-based evaluation for open-domain code generation.
\newblock \emph{CoRR}, abs/2212.10481, 2022{\natexlab{h}}.
\newblock \doi{10.48550/ARXIV.2212.10481}.
\newblock URL \url{https://doi.org/10.48550/arXiv.2212.10481}.

\bibitem[Wang et~al.(2023{\natexlab{i}})Wang, Cuenca, Zhou, Xu, and Neubig]{2022MCoNaLa}
Zhiruo Wang, Grace Cuenca, Shuyan Zhou, Frank~F. Xu, and Graham Neubig.
\newblock Mconala: {A} benchmark for code generation from multiple natural languages.
\newblock In Andreas Vlachos and Isabelle Augenstein (eds.), \emph{Findings of the Association for Computational Linguistics: {EACL} 2023, Dubrovnik, Croatia, May 2-6, 2023}, pp.\  265--273. Association for Computational Linguistics, 2023{\natexlab{i}}.
\newblock \doi{10.18653/V1/2023.FINDINGS-EACL.20}.
\newblock URL \url{https://doi.org/10.18653/v1/2023.findings-eacl.20}.

\bibitem[Watson et~al.(2020)Watson, Tufano, Moran, Bavota, and Poshyvanyk]{assert2020-1}
Cody Watson, Michele Tufano, Kevin Moran, Gabriele Bavota, and Denys Poshyvanyk.
\newblock On learning meaningful assert statements for unit test cases.
\newblock In Gregg Rothermel and Doo{-}Hwan Bae (eds.), \emph{{ICSE} '20: 42nd International Conference on Software Engineering, Seoul, South Korea, 27 June - 19 July, 2020}, pp.\  1398--1409. {ACM}, 2020.
\newblock \doi{10.1145/3377811.3380429}.
\newblock URL \url{https://doi.org/10.1145/3377811.3380429}.

\bibitem[Wei et~al.(2022{\natexlab{a}})Wei, Deng, Yang, and Zhang]{fuzz2022-1}
Anjiang Wei, Yinlin Deng, Chenyuan Yang, and Lingming Zhang.
\newblock Free lunch for testing: Fuzzing deep-learning libraries from open source.
\newblock In \emph{44th {IEEE/ACM} 44th International Conference on Software Engineering, {ICSE} 2022, Pittsburgh, PA, USA, May 25-27, 2022}, pp.\  995--1007. {ACM}, 2022{\natexlab{a}}.
\newblock \doi{10.1145/3510003.3510041}.
\newblock URL \url{https://doi.org/10.1145/3510003.3510041}.

\bibitem[Wei et~al.(2019)Wei, Li, Xia, Fu, and Jin]{sum2019-1}
Bolin Wei, Ge~Li, Xin Xia, Zhiyi Fu, and Zhi Jin.
\newblock Code generation as a dual task of code summarization.
\newblock In Hanna~M. Wallach, Hugo Larochelle, Alina Beygelzimer, Florence d'Alch{\'{e}}{-}Buc, Emily~B. Fox, and Roman Garnett (eds.), \emph{Advances in Neural Information Processing Systems 32: Annual Conference on Neural Information Processing Systems 2019, NeurIPS 2019, December 8-14, 2019, Vancouver, BC, Canada}, pp.\  6559--6569, 2019.
\newblock URL \url{https://proceedings.neurips.cc/paper/2019/hash/e52ad5c9f751f599492b4f087ed7ecfc-Abstract.html}.

\bibitem[Wei \& Li(2017)Wei and Li]{clone2017-1}
Huihui Wei and Ming Li.
\newblock Supervised deep features for software functional clone detection by exploiting lexical and syntactical information in source code.
\newblock In Carles Sierra (ed.), \emph{Proceedings of the Twenty-Sixth International Joint Conference on Artificial Intelligence, {IJCAI} 2017, Melbourne, Australia, August 19-25, 2017}, pp.\  3034--3040. ijcai.org, 2017.
\newblock \doi{10.24963/ijcai.2017/423}.
\newblock URL \url{https://doi.org/10.24963/ijcai.2017/423}.

\bibitem[Wei et~al.(2022{\natexlab{b}})Wei, Bosma, Zhao, Guu, Yu, Lester, Du, Dai, and Le]{2021FLAN}
Jason Wei, Maarten Bosma, Vincent~Y. Zhao, Kelvin Guu, Adams~Wei Yu, Brian Lester, Nan Du, Andrew~M. Dai, and Quoc~V. Le.
\newblock Finetuned language models are zero-shot learners.
\newblock In \emph{The Tenth International Conference on Learning Representations, {ICLR} 2022, Virtual Event, April 25-29, 2022}. OpenReview.net, 2022{\natexlab{b}}.
\newblock URL \url{https://openreview.net/forum?id=gEZrGCozdqR}.

\bibitem[Wei et~al.(2022{\natexlab{c}})Wei, Tay, Bommasani, Raffel, Zoph, Borgeaud, Yogatama, Bosma, Zhou, Metzler, Chi, Hashimoto, Vinyals, Liang, Dean, and Fedus]{2022emergent}
Jason Wei, Yi~Tay, Rishi Bommasani, Colin Raffel, Barret Zoph, Sebastian Borgeaud, Dani Yogatama, Maarten Bosma, Denny Zhou, Donald Metzler, Ed~H. Chi, Tatsunori Hashimoto, Oriol Vinyals, Percy Liang, Jeff Dean, and William Fedus.
\newblock Emergent abilities of large language models.
\newblock \emph{Transactions on Machine Learning Research}, 2022{\natexlab{c}}.
\newblock ISSN 2835-8856.
\newblock URL \url{https://openreview.net/forum?id=yzkSU5zdwD}.
\newblock Survey Certification.

\bibitem[Wei et~al.(2022{\natexlab{d}})Wei, Wang, Schuurmans, Bosma, Ichter, Xia, Chi, Le, and Zhou]{2022CoT}
Jason Wei, Xuezhi Wang, Dale Schuurmans, Maarten Bosma, Brian Ichter, Fei Xia, Ed~H. Chi, Quoc~V. Le, and Denny Zhou.
\newblock Chain-of-thought prompting elicits reasoning in large language models.
\newblock In \emph{NeurIPS}, 2022{\natexlab{d}}.
\newblock URL \url{http://papers.nips.cc/paper\_files/paper/2022/hash/9d5609613524ecf4f15af0f7b31abca4-Abstract-Conference.html}.

\bibitem[Wei et~al.(2020)Wei, Goyal, Durrett, and Dillig]{type2020-3}
Jiayi Wei, Maruth Goyal, Greg Durrett, and Isil Dillig.
\newblock Lambdanet: Probabilistic type inference using graph neural networks.
\newblock In \emph{8th International Conference on Learning Representations, {ICLR} 2020, Addis Ababa, Ethiopia, April 26-30, 2020}. OpenReview.net, 2020.
\newblock URL \url{https://openreview.net/forum?id=Hkx6hANtwH}.

\bibitem[Wei et~al.(2023)Wei, Durrett, and Dillig]{type2023-2}
Jiayi Wei, Greg Durrett, and Isil Dillig.
\newblock Typet5: Seq2seq type inference using static analysis.
\newblock In \emph{The Eleventh International Conference on Learning Representations, {ICLR} 2023, Kigali, Rwanda, May 1-5, 2023}. OpenReview.net, 2023.
\newblock URL \url{https://openreview.net/pdf?id=4TyNEhI2GdN}.

\bibitem[Wen et~al.(2022)Wen, Guo, Fu, Li, Xu, Tang, Zhao, Hu, Du, Li, Wang, Zhou, and Chen]{trans2022-1}
Yuanbo Wen, Qi~Guo, Qiang Fu, Xiaqing Li, Jianxing Xu, Yanlin Tang, Yongwei Zhao, Xing Hu, Zidong Du, Ling Li, Chao Wang, Xuehai Zhou, and Yunji Chen.
\newblock Babeltower: Learning to auto-parallelized program translation.
\newblock In Kamalika Chaudhuri, Stefanie Jegelka, Le~Song, Csaba Szepesv{\'{a}}ri, Gang Niu, and Sivan Sabato (eds.), \emph{International Conference on Machine Learning, {ICML} 2022, 17-23 July 2022, Baltimore, Maryland, {USA}}, volume 162 of \emph{Proceedings of Machine Learning Research}, pp.\  23685--23700. {PMLR}, 2022.
\newblock URL \url{https://proceedings.mlr.press/v162/wen22b.html}.

\bibitem[Weyssow et~al.(2022)Weyssow, Sahraoui, and Syriani]{re-model2021-2}
Martin Weyssow, Houari~A. Sahraoui, and Eugene Syriani.
\newblock Recommending metamodel concepts during modeling activities with pre-trained language models.
\newblock \emph{Softw. Syst. Model.}, 21\penalty0 (3):\penalty0 1071--1089, 2022.
\newblock \doi{10.1007/S10270-022-00975-5}.
\newblock URL \url{https://doi.org/10.1007/s10270-022-00975-5}.

\bibitem[White et~al.(2023)White, Hays, Fu, Spencer{-}Smith, and Schmidt]{re-ana2023-1}
Jules White, Sam Hays, Quchen Fu, Jesse Spencer{-}Smith, and Douglas~C. Schmidt.
\newblock Chatgpt prompt patterns for improving code quality, refactoring, requirements elicitation, and software design.
\newblock \emph{CoRR}, abs/2303.07839, 2023.
\newblock \doi{10.48550/ARXIV.2303.07839}.
\newblock URL \url{https://doi.org/10.48550/arXiv.2303.07839}.

\bibitem[White et~al.(2015)White, Vendome, V{\'{a}}squez, and Poshyvanyk]{completion2015-1}
Martin White, Christopher Vendome, Mario~Linares V{\'{a}}squez, and Denys Poshyvanyk.
\newblock Toward deep learning software repositories.
\newblock In Massimiliano~Di Penta, Martin Pinzger, and Romain Robbes (eds.), \emph{12th {IEEE/ACM} Working Conference on Mining Software Repositories, {MSR} 2015, Florence, Italy, May 16-17, 2015}, pp.\  334--345. {IEEE} Computer Society, 2015.
\newblock \doi{10.1109/MSR.2015.38}.
\newblock URL \url{https://doi.org/10.1109/MSR.2015.38}.

\bibitem[White et~al.(2016)White, Tufano, Vendome, and Poshyvanyk]{clone2016-1}
Martin White, Michele Tufano, Christopher Vendome, and Denys Poshyvanyk.
\newblock Deep learning code fragments for code clone detection.
\newblock In David Lo, Sven Apel, and Sarfraz Khurshid (eds.), \emph{Proceedings of the 31st {IEEE/ACM} International Conference on Automated Software Engineering, {ASE} 2016, Singapore, September 3-7, 2016}, pp.\  87--98. {ACM}, 2016.
\newblock \doi{10.1145/2970276.2970326}.
\newblock URL \url{https://doi.org/10.1145/2970276.2970326}.

\bibitem[Widyasari et~al.(2020)Widyasari, Sim, Lok, Qi, Phan, Tay, Tan, Wee, Tan, Yieh, Goh, Thung, Kang, Hoang, Lo, and Ouh]{fix-data-2020-1}
Ratnadira Widyasari, Sheng~Qin Sim, Camellia Lok, Haodi Qi, Jack Phan, Qijin Tay, Constance Tan, Fiona Wee, Jodie~Ethelda Tan, Yuheng Yieh, Brian Goh, Ferdian Thung, Hong~Jin Kang, Thong Hoang, David Lo, and Eng~Lieh Ouh.
\newblock Bugsinpy: a database of existing bugs in python programs to enable controlled testing and debugging studies.
\newblock In Prem Devanbu, Myra~B. Cohen, and Thomas Zimmermann (eds.), \emph{{ESEC/FSE} '20: 28th {ACM} Joint European Software Engineering Conference and Symposium on the Foundations of Software Engineering, Virtual Event, USA, November 8-13, 2020}, pp.\  1556--1560. {ACM}, 2020.
\newblock \doi{10.1145/3368089.3417943}.
\newblock URL \url{https://doi.org/10.1145/3368089.3417943}.

\bibitem[Williams(1992)]{1992REINFORCE}
Ronald~J. Williams.
\newblock Simple statistical gradient-following algorithms for connectionist reinforcement learning.
\newblock \emph{Mach. Learn.}, 8:\penalty0 229--256, 1992.
\newblock \doi{10.1007/BF00992696}.
\newblock URL \url{https://doi.org/10.1007/BF00992696}.

\bibitem[Wong et~al.(2013)Wong, Yang, and Tan]{comment2013-1}
Edmund Wong, Jinqiu Yang, and Lin Tan.
\newblock Autocomment: Mining question and answer sites for automatic comment generation.
\newblock In Ewen Denney, Tevfik Bultan, and Andreas Zeller (eds.), \emph{2013 28th {IEEE/ACM} International Conference on Automated Software Engineering, {ASE} 2013, Silicon Valley, CA, USA, November 11-15, 2013}, pp.\  562--567. {IEEE}, 2013.
\newblock \doi{10.1109/ASE.2013.6693113}.
\newblock URL \url{https://doi.org/10.1109/ASE.2013.6693113}.

\bibitem[Wong et~al.(2015)Wong, Liu, and Tan]{sum2015-1}
Edmund Wong, Taiyue Liu, and Lin Tan.
\newblock Clocom: Mining existing source code for automatic comment generation.
\newblock In Yann{-}Ga{\"{e}}l Gu{\'{e}}h{\'{e}}neuc, Bram Adams, and Alexander Serebrenik (eds.), \emph{22nd {IEEE} International Conference on Software Analysis, Evolution, and Reengineering, {SANER} 2015, Montreal, QC, Canada, March 2-6, 2015}, pp.\  380--389. {IEEE} Computer Society, 2015.
\newblock \doi{10.1109/SANER.2015.7081848}.
\newblock URL \url{https://doi.org/10.1109/SANER.2015.7081848}.

\bibitem[Wong et~al.(2023)Wong, Wang, Li, Liu, Wang, Tang, Nie, and Wu]{2023DecGPT}
Wai~Kin Wong, Huaijin Wang, Zongjie Li, Zhibo Liu, Shuai Wang, Qiyi Tang, Sen Nie, and Shi Wu.
\newblock Refining decompiled {C} code with large language models.
\newblock \emph{CoRR}, abs/2310.06530, 2023.
\newblock \doi{10.48550/ARXIV.2310.06530}.
\newblock URL \url{https://doi.org/10.48550/arXiv.2310.06530}.

\bibitem[Wu et~al.(2024{\natexlab{a}})Wu, Ge, Guo, Wang, Liang, Lu, Shan, and Luo]{2024Plot2Code}
Chengyue Wu, Yixiao Ge, Qiushan Guo, Jiahao Wang, Zhixuan Liang, Zeyu Lu, Ying Shan, and Ping Luo.
\newblock Plot2code: A comprehensive benchmark for evaluating multi-modal large language models in code generation from scientific plots.
\newblock 2024{\natexlab{a}}.
\newblock URL \url{https://doi.org/10.48550/arXiv.2405.07990}.

\bibitem[Wu et~al.(2024{\natexlab{b}})Wu, Ahmad, Zhang, Ramanathan, and Ma]{2024Repoformer}
Di~Wu, Wasi~Uddin Ahmad, Dejiao Zhang, Murali~Krishna Ramanathan, and Xiaofei Ma.
\newblock Repoformer: Selective retrieval for repository-level code completion.
\newblock \emph{CoRR}, abs/2403.10059, 2024{\natexlab{b}}.
\newblock \doi{10.48550/ARXIV.2403.10059}.
\newblock URL \url{https://doi.org/10.48550/arXiv.2403.10059}.

\bibitem[Wu et~al.(2021)Wu, Zhao, and Zhang]{sum2020-3}
Hongqiu Wu, Hai Zhao, and Min Zhang.
\newblock Code summarization with structure-induced transformer.
\newblock In Chengqing Zong, Fei Xia, Wenjie Li, and Roberto Navigli (eds.), \emph{Findings of the Association for Computational Linguistics: {ACL/IJCNLP} 2021, Online Event, August 1-6, 2021}, volume {ACL/IJCNLP} 2021 of \emph{Findings of {ACL}}, pp.\  1078--1090. Association for Computational Linguistics, 2021.
\newblock \doi{10.18653/v1/2021.findings-acl.93}.
\newblock URL \url{https://doi.org/10.18653/v1/2021.findings-acl.93}.

\bibitem[Wu \& Fard(2024)Wu and Fard]{2024communication}
Jie~JW Wu and Fatemeh~H Fard.
\newblock Benchmarking the communication competence of code generation for llms and llm agent.
\newblock \emph{arXiv preprint arXiv:2406.00215}, 2024.
\newblock URL \url{https://doi.org/10.48550/arXiv.2406.00215}.

\bibitem[Wu et~al.(2020)Wu, Wang, Yin, Cheng, Xu, and Roy]{clone2019-3}
Ming Wu, Pengcheng Wang, Kangqi Yin, Haoyu Cheng, Yun Xu, and Chanchal~K. Roy.
\newblock Lvmapper: {A} large-variance clone detector using sequencing alignment approach.
\newblock \emph{{IEEE} Access}, 8:\penalty0 27986--27997, 2020.
\newblock \doi{10.1109/ACCESS.2020.2971545}.
\newblock URL \url{https://doi.org/10.1109/ACCESS.2020.2971545}.

\bibitem[Wu et~al.(2022)Wu, Jiang, Xiang, Zhang, Yang, Ma, Nie, Wu, Cui, and Zhang]{fuzz2022-4}
Mingyuan Wu, Ling Jiang, Jiahong Xiang, Yuqun Zhang, Guowei Yang, Huixin Ma, Sen Nie, Shi Wu, Heming Cui, and Lingming Zhang.
\newblock Evaluating and improving neural program-smoothing-based fuzzing.
\newblock In \emph{44th {IEEE/ACM} 44th International Conference on Software Engineering, {ICSE} 2022, Pittsburgh, PA, USA, May 25-27, 2022}, pp.\  847--858. {ACM}, 2022.
\newblock \doi{10.1145/3510003.3510089}.
\newblock URL \url{https://doi.org/10.1145/3510003.3510089}.

\bibitem[Xia \& Zhang(2022)Xia and Zhang]{fix2022-2}
Chunqiu~Steven Xia and Lingming Zhang.
\newblock Less training, more repairing please: revisiting automated program repair via zero-shot learning.
\newblock In Abhik Roychoudhury, Cristian Cadar, and Miryung Kim (eds.), \emph{Proceedings of the 30th {ACM} Joint European Software Engineering Conference and Symposium on the Foundations of Software Engineering, {ESEC/FSE} 2022, Singapore, Singapore, November 14-18, 2022}, pp.\  959--971. {ACM}, 2022.
\newblock \doi{10.1145/3540250.3549101}.
\newblock URL \url{https://doi.org/10.1145/3540250.3549101}.

\bibitem[Xia \& Zhang(2023)Xia and Zhang]{fix2023-0.5}
Chunqiu~Steven Xia and Lingming Zhang.
\newblock Conversational automated program repair.
\newblock \emph{CoRR}, abs/2301.13246, 2023.
\newblock \doi{10.48550/ARXIV.2301.13246}.
\newblock URL \url{https://doi.org/10.48550/arXiv.2301.13246}.

\bibitem[Xia et~al.(2023{\natexlab{a}})Xia, Paltenghi, Tian, Pradel, and Zhang]{2023Fuzz4All}
Chunqiu~Steven Xia, Matteo Paltenghi, Jia~Le Tian, Michael Pradel, and Lingming Zhang.
\newblock Fuzz4all: Universal fuzzing with large language models.
\newblock \emph{CoRR}, abs/2308.04748, 2023{\natexlab{a}}.
\newblock \doi{10.48550/ARXIV.2308.04748}.
\newblock URL \url{https://doi.org/10.48550/arXiv.2308.04748}.

\bibitem[Xia et~al.(2023{\natexlab{b}})Xia, Wei, and Zhang]{fix2022-3}
Chunqiu~Steven Xia, Yuxiang Wei, and Lingming Zhang.
\newblock Automated program repair in the era of large pre-trained language models.
\newblock In \emph{45th {IEEE/ACM} International Conference on Software Engineering, {ICSE} 2023, Melbourne, Australia, May 14-20, 2023}, pp.\  1482--1494. {IEEE}, 2023{\natexlab{b}}.
\newblock \doi{10.1109/ICSE48619.2023.00129}.
\newblock URL \url{https://doi.org/10.1109/ICSE48619.2023.00129}.

\bibitem[Xie et~al.(2021{\natexlab{a}})Xie, Huang, Liang, Huang, and Xiao]{UI2021-4}
Chenhao Xie, Wenhao Huang, Jiaqing Liang, Chengsong Huang, and Yanghua Xiao.
\newblock Webke: Knowledge extraction from semi-structured web with pre-trained markup language model.
\newblock In Gianluca Demartini, Guido Zuccon, J.~Shane Culpepper, Zi~Huang, and Hanghang Tong (eds.), \emph{{CIKM} '21: The 30th {ACM} International Conference on Information and Knowledge Management, Virtual Event, Queensland, Australia, November 1 - 5, 2021}, pp.\  2211--2220. {ACM}, 2021{\natexlab{a}}.
\newblock \doi{10.1145/3459637.3482491}.
\newblock URL \url{https://doi.org/10.1145/3459637.3482491}.

\bibitem[Xie et~al.(2022{\natexlab{a}})Xie, Li, Kim, Pham, Tan, Zhang, and Godfrey]{fuzz2021-1}
Danning Xie, Yitong Li, Mijung Kim, Hung~Viet Pham, Lin Tan, Xiangyu Zhang, and Michael~W. Godfrey.
\newblock Docter: documentation-guided fuzzing for testing deep learning {API} functions.
\newblock In Sukyoung Ryu and Yannis Smaragdakis (eds.), \emph{{ISSTA} '22: 31st {ACM} {SIGSOFT} International Symposium on Software Testing and Analysis, Virtual Event, South Korea, July 18 - 22, 2022}, pp.\  176--188. {ACM}, 2022{\natexlab{a}}.
\newblock \doi{10.1145/3533767.3534220}.
\newblock URL \url{https://doi.org/10.1145/3533767.3534220}.

\bibitem[Xie et~al.(2021{\natexlab{b}})Xie, Ye, Sun, and Zhang]{id2021-2}
Rui Xie, Wei Ye, Jinan Sun, and Shikun Zhang.
\newblock Exploiting method names to improve code summarization: {A} deliberation multi-task learning approach.
\newblock In \emph{29th {IEEE/ACM} International Conference on Program Comprehension, {ICPC} 2021, Madrid, Spain, May 20-21, 2021}, pp.\  138--148. {IEEE}, 2021{\natexlab{b}}.
\newblock \doi{10.1109/ICPC52881.2021.00022}.
\newblock URL \url{https://doi.org/10.1109/ICPC52881.2021.00022}.

\bibitem[Xie et~al.(2024)Xie, Zeng, Yu, Gao, Zhang, and Ye]{2024CodeShell}
Rui Xie, Zhengran Zeng, Zhuohao Yu, Chang Gao, Shikun Zhang, and Wei Ye.
\newblock Codeshell technical report.
\newblock \emph{CoRR}, abs/2403.15747, 2024.
\newblock \doi{10.48550/ARXIV.2403.15747}.
\newblock URL \url{https://doi.org/10.48550/arXiv.2403.15747}.

\bibitem[Xie et~al.(2022{\natexlab{b}})Xie, Wu, Shi, Zhong, Scholak, Yasunaga, Wu, Zhong, Yin, Wang, Zhong, Wang, Li, Boyle, Ni, Yao, Radev, Xiong, Kong, Zhang, Smith, Zettlemoyer, and Yu]{2022UnifiedSKG}
Tianbao Xie, Chen~Henry Wu, Peng Shi, Ruiqi Zhong, Torsten Scholak, Michihiro Yasunaga, Chien{-}Sheng Wu, Ming Zhong, Pengcheng Yin, Sida~I. Wang, Victor Zhong, Bailin Wang, Chengzu Li, Connor Boyle, Ansong Ni, Ziyu Yao, Dragomir Radev, Caiming Xiong, Lingpeng Kong, Rui Zhang, Noah~A. Smith, Luke Zettlemoyer, and Tao Yu.
\newblock Unifiedskg: Unifying and multi-tasking structured knowledge grounding with text-to-text language models.
\newblock In Yoav Goldberg, Zornitsa Kozareva, and Yue Zhang (eds.), \emph{Proceedings of the 2022 Conference on Empirical Methods in Natural Language Processing, {EMNLP} 2022, Abu Dhabi, United Arab Emirates, December 7-11, 2022}, pp.\  602--631. Association for Computational Linguistics, 2022{\natexlab{b}}.
\newblock \doi{10.18653/v1/2022.emnlp-main.39}.
\newblock URL \url{https://doi.org/10.18653/v1/2022.emnlp-main.39}.

\bibitem[Xie et~al.(2023{\natexlab{a}})Xie, Lin, Dong, Zhang, and Wu]{retrieval-survey-2023-1}
Yutao Xie, Jiayi Lin, Hande Dong, Lei Zhang, and Zhonghai Wu.
\newblock A survey of deep code search.
\newblock \emph{CoRR}, abs/2305.05959, 2023{\natexlab{a}}.
\newblock \doi{10.48550/arXiv.2305.05959}.
\newblock URL \url{https://doi.org/10.48550/arXiv.2305.05959}.

\bibitem[Xie et~al.(2023{\natexlab{b}})Xie, Chen, Zhi, Deng, and Yin]{unit2023-6}
Zhuokui Xie, Yinghao Chen, Chen Zhi, Shuiguang Deng, and Jianwei Yin.
\newblock Chatunitest: a chatgpt-based automated unit test generation tool.
\newblock \emph{CoRR}, abs/2305.04764, 2023{\natexlab{b}}.
\newblock \doi{10.48550/ARXIV.2305.04764}.
\newblock URL \url{https://doi.org/10.48550/arXiv.2305.04764}.

\bibitem[Xu et~al.(2024)Xu, Sun, Zheng, Geng, Zhao, Feng, Tao, Lin, and Jiang]{2023WizardLM}
Can Xu, Qingfeng Sun, Kai Zheng, Xiubo Geng, Pu~Zhao, Jiazhan Feng, Chongyang Tao, Qingwei Lin, and Daxin Jiang.
\newblock Wizard{LM}: Empowering large pre-trained language models to follow complex instructions.
\newblock In \emph{The Twelfth International Conference on Learning Representations}, 2024.
\newblock URL \url{https://openreview.net/forum?id=CfXh93NDgH}.

\bibitem[Xu et~al.(2022)Xu, Alon, Neubig, and Hellendoorn]{2022PolyCoder}
Frank~F. Xu, Uri Alon, Graham Neubig, and Vincent~Josua Hellendoorn.
\newblock A systematic evaluation of large language models of code.
\newblock In Swarat Chaudhuri and Charles Sutton (eds.), \emph{MAPS@PLDI 2022: 6th {ACM} {SIGPLAN} International Symposium on Machine Programming, San Diego, CA, USA, 13 June 2022}, pp.\  1--10. {ACM}, 2022.
\newblock \doi{10.1145/3520312.3534862}.
\newblock URL \url{https://doi.org/10.1145/3520312.3534862}.

\bibitem[Xu et~al.(2021{\natexlab{a}})Xu, Yang, Liu, Shuai, Yan, Lei, and Xu]{retrieval2021-1}
Ling Xu, Huanhuan Yang, Chao Liu, Jianhang Shuai, Meng Yan, Yan Lei, and Zhou Xu.
\newblock Two-stage attention-based model for code search with textual and structural features.
\newblock In \emph{28th {IEEE} International Conference on Software Analysis, Evolution and Reengineering, {SANER} 2021, Honolulu, HI, USA, March 9-12, 2021}, pp.\  342--353. {IEEE}, 2021{\natexlab{a}}.
\newblock \doi{10.1109/SANER50967.2021.00039}.
\newblock URL \url{https://doi.org/10.1109/SANER50967.2021.00039}.

\bibitem[Xu et~al.(2019{\natexlab{a}})Xu, Yao, Xu, Gu, Tong, and Lu]{commit2019-2}
Shengbin Xu, Yuan Yao, Feng Xu, Tianxiao Gu, Hanghang Tong, and Jian Lu.
\newblock Commit message generation for source code changes.
\newblock In Sarit Kraus (ed.), \emph{Proceedings of the Twenty-Eighth International Joint Conference on Artificial Intelligence, {IJCAI} 2019, Macao, China, August 10-16, 2019}, pp.\  3975--3981. ijcai.org, 2019{\natexlab{a}}.
\newblock \doi{10.24963/IJCAI.2019/552}.
\newblock URL \url{https://doi.org/10.24963/ijcai.2019/552}.

\bibitem[Xu et~al.(2019{\natexlab{b}})Xu, Zhang, Wang, Cao, Guo, and Xu]{id2019-1}
Sihan Xu, Sen Zhang, Weijing Wang, Xinya Cao, Chenkai Guo, and Jing Xu.
\newblock Method name suggestion with hierarchical attention networks.
\newblock In Manuel~V. Hermenegildo and Atsushi Igarashi (eds.), \emph{Proceedings of the 2019 {ACM} {SIGPLAN} Workshop on Partial Evaluation and Program Manipulation, PEPM@POPL 2019, Cascais, Portugal, January 14-15, 2019}, pp.\  10--21. {ACM}, 2019{\natexlab{b}}.
\newblock \doi{10.1145/3294032.3294079}.
\newblock URL \url{https://doi.org/10.1145/3294032.3294079}.

\bibitem[Xu et~al.(2023)Xu, Zhang, Feng, Ye, Su, Jiang, Cheng, Tan, and Zhang]{ob2023-2}
Xiangzhe Xu, Zhuo Zhang, Shiwei Feng, Yapeng Ye, Zian Su, Nan Jiang, Siyuan Cheng, Lin Tan, and Xiangyu Zhang.
\newblock Lmpa: Improving decompilation by synergy of large language model and program analysis.
\newblock \emph{CoRR}, abs/2306.02546, 2023.
\newblock \doi{10.48550/ARXIV.2306.02546}.
\newblock URL \url{https://doi.org/10.48550/arXiv.2306.02546}.

\bibitem[Xu et~al.(2017)Xu, Liu, and Song]{sql2017-2}
Xiaojun Xu, Chang Liu, and Dawn Song.
\newblock Sqlnet: Generating structured queries from natural language without reinforcement learning.
\newblock \emph{CoRR}, abs/1711.04436, 2017.
\newblock URL \url{http://arxiv.org/abs/1711.04436}.

\bibitem[Xu \& Zhu(2022)Xu and Zhu]{2022survey2}
Yichen Xu and Yanqiao Zhu.
\newblock A survey on pretrained language models for neural code intelligence.
\newblock \emph{CoRR}, abs/2212.10079, 2022.
\newblock \doi{10.48550/arXiv.2212.10079}.
\newblock URL \url{https://doi.org/10.48550/arXiv.2212.10079}.

\bibitem[Xu et~al.(2016)Xu, Zhang, Chen, Pei, and Xu]{type2016-1}
Zhaogui Xu, Xiangyu Zhang, Lin Chen, Kexin Pei, and Baowen Xu.
\newblock Python probabilistic type inference with natural language support.
\newblock In Thomas Zimmermann, Jane Cleland{-}Huang, and Zhendong Su (eds.), \emph{Proceedings of the 24th {ACM} {SIGSOFT} International Symposium on Foundations of Software Engineering, {FSE} 2016, Seattle, WA, USA, November 13-18, 2016}, pp.\  607--618. {ACM}, 2016.
\newblock \doi{10.1145/2950290.2950343}.
\newblock URL \url{https://doi.org/10.1145/2950290.2950343}.

\bibitem[Xu et~al.(2021{\natexlab{b}})Xu, Fang, and Yang]{2021Malbert}
Zhifeng Xu, Xianjin Fang, and Gaoming Yang.
\newblock Malbert: {A} novel pre-training method for malware detection.
\newblock \emph{Comput. Secur.}, 111:\penalty0 102458, 2021{\natexlab{b}}.
\newblock \doi{10.1016/J.COSE.2021.102458}.
\newblock URL \url{https://doi.org/10.1016/j.cose.2021.102458}.

\bibitem[Xue et~al.(2021)Xue, Constant, Roberts, Kale, Al{-}Rfou, Siddhant, Barua, and Raffel]{2020mT5}
Linting Xue, Noah Constant, Adam Roberts, Mihir Kale, Rami Al{-}Rfou, Aditya Siddhant, Aditya Barua, and Colin Raffel.
\newblock mt5: {A} massively multilingual pre-trained text-to-text transformer.
\newblock In Kristina Toutanova, Anna Rumshisky, Luke Zettlemoyer, Dilek Hakkani{-}T{\"{u}}r, Iz~Beltagy, Steven Bethard, Ryan Cotterell, Tanmoy Chakraborty, and Yichao Zhou (eds.), \emph{Proceedings of the 2021 Conference of the North American Chapter of the Association for Computational Linguistics: Human Language Technologies, {NAACL-HLT} 2021, Online, June 6-11, 2021}, pp.\  483--498. Association for Computational Linguistics, 2021.
\newblock \doi{10.18653/v1/2021.naacl-main.41}.
\newblock URL \url{https://doi.org/10.18653/v1/2021.naacl-main.41}.

\bibitem[Yaghmazadeh et~al.(2017)Yaghmazadeh, Wang, Dillig, and Dillig]{sql-data-2017-1}
Navid Yaghmazadeh, Yuepeng Wang, Isil Dillig, and Thomas Dillig.
\newblock Sqlizer: query synthesis from natural language.
\newblock \emph{Proc. {ACM} Program. Lang.}, 1\penalty0 ({OOPSLA}):\penalty0 63:1--63:26, 2017.
\newblock \doi{10.1145/3133887}.
\newblock URL \url{https://doi.org/10.1145/3133887}.

\bibitem[Yahya \& Kim(2022)Yahya and Kim]{clone2022-1}
Mohammad~A. Yahya and Dae{-}Kyoo Kim.
\newblock Cross-language source code clone detection using deep learning with infercode.
\newblock \emph{CoRR}, abs/2205.04913, 2022.
\newblock \doi{10.48550/arXiv.2205.04913}.
\newblock URL \url{https://doi.org/10.48550/arXiv.2205.04913}.

\bibitem[Yan et~al.(2020)Yan, Yu, Chen, Shen, and Jiang]{retrieval-data-2020-1}
Shuhan Yan, Hang Yu, Yuting Chen, Beijun Shen, and Lingxiao Jiang.
\newblock Are the code snippets what we are searching for? {A} benchmark and an empirical study on code search with natural-language queries.
\newblock In Kostas Kontogiannis, Foutse Khomh, Alexander Chatzigeorgiou, Marios{-}Eleftherios Fokaefs, and Minghui Zhou (eds.), \emph{27th {IEEE} International Conference on Software Analysis, Evolution and Reengineering, {SANER} 2020, London, ON, Canada, February 18-21, 2020}, pp.\  344--354. {IEEE}, 2020.
\newblock \doi{10.1109/SANER48275.2020.9054840}.
\newblock URL \url{https://doi.org/10.1109/SANER48275.2020.9054840}.

\bibitem[Yan et~al.(2023)Yan, Tian, Li, Chen, and Wang]{trans-data-2023-1}
Weixiang Yan, Yuchen Tian, Yunzhe Li, Qian Chen, and Wen Wang.
\newblock Codetransocean: {A} comprehensive multilingual benchmark for code translation.
\newblock \emph{CoRR}, abs/2310.04951, 2023.
\newblock \doi{10.48550/ARXIV.2310.04951}.
\newblock URL \url{https://doi.org/10.48550/arXiv.2310.04951}.

\bibitem[Yang et~al.(2023{\natexlab{a}})Yang, Xiao, Wang, Zhang, Bian, Yin, Lv, Pan, Wang, Yan, Yang, Deng, Wang, Liu, Ai, Dong, Zhao, Xu, Sun, Zhang, Liu, Ji, Xie, Dai, Fang, Su, Song, Liu, Ru, Ma, Wang, Liu, Lin, Nie, Guo, Sun, Zhang, Li, Li, Cheng, Chen, Zeng, Wang, Chen, Men, Yu, Pan, Shen, Wang, Li, Jiang, Gao, Zhang, Zhou, and Wu]{2023Baichuan2}
Aiyuan Yang, Bin Xiao, Bingning Wang, Borong Zhang, Ce~Bian, Chao Yin, Chenxu Lv, Da~Pan, Dian Wang, Dong Yan, Fan Yang, Fei Deng, Feng Wang, Feng Liu, Guangwei Ai, Guosheng Dong, Haizhou Zhao, Hang Xu, Haoze Sun, Hongda Zhang, Hui Liu, Jiaming Ji, Jian Xie, Juntao Dai, Kun Fang, Lei Su, Liang Song, Lifeng Liu, Liyun Ru, Luyao Ma, Mang Wang, Mickel Liu, MingAn Lin, Nuolan Nie, Peidong Guo, Ruiyang Sun, Tao Zhang, Tianpeng Li, Tianyu Li, Wei Cheng, Weipeng Chen, Xiangrong Zeng, Xiaochuan Wang, Xiaoxi Chen, Xin Men, Xin Yu, Xuehai Pan, Yanjun Shen, Yiding Wang, Yiyu Li, Youxin Jiang, Yuchen Gao, Yupeng Zhang, Zenan Zhou, and Zhiying Wu.
\newblock Baichuan 2: Open large-scale language models.
\newblock \emph{CoRR}, abs/2309.10305, 2023{\natexlab{a}}.
\newblock \doi{10.48550/arXiv.2309.10305}.
\newblock URL \url{https://doi.org/10.48550/arXiv.2309.10305}.

\bibitem[Yang et~al.(2024{\natexlab{a}})Yang, Chen, Lin, Zhou, and Wang]{2024TELPA}
Chen Yang, Junjie Chen, Bin Lin, Jianyi Zhou, and Ziqi Wang.
\newblock Enhancing llm-based test generation for hard-to-cover branches via program analysis.
\newblock \emph{CoRR}, abs/2404.04966, 2024{\natexlab{a}}.
\newblock \doi{10.48550/ARXIV.2404.04966}.
\newblock URL \url{https://doi.org/10.48550/arXiv.2404.04966}.

\bibitem[Yang et~al.(2023{\natexlab{b}})Yang, Deng, Lu, Yao, Liu, Jabbarvand, and Zhang]{fuzz2023-2}
Chenyuan Yang, Yinlin Deng, Runyu Lu, Jiayi Yao, Jiawei Liu, Reyhaneh Jabbarvand, and Lingming Zhang.
\newblock White-box compiler fuzzing empowered by large language models.
\newblock \emph{CoRR}, abs/2310.15991, 2023{\natexlab{b}}.
\newblock \doi{10.48550/ARXIV.2310.15991}.
\newblock URL \url{https://doi.org/10.48550/arXiv.2310.15991}.

\bibitem[Yang et~al.(2023{\natexlab{c}})Yang, Deng, Yao, Tu, Li, and Zhang]{fuzz2023-1}
Chenyuan Yang, Yinlin Deng, Jiayi Yao, Yuxing Tu, Hanchi Li, and Lingming Zhang.
\newblock Fuzzing automatic differentiation in deep-learning libraries.
\newblock In \emph{45th {IEEE/ACM} International Conference on Software Engineering, {ICSE} 2023, Melbourne, Australia, May 14-20, 2023}, pp.\  1174--1186. {IEEE}, 2023{\natexlab{c}}.
\newblock \doi{10.1109/ICSE48619.2023.00105}.
\newblock URL \url{https://doi.org/10.1109/ICSE48619.2023.00105}.

\bibitem[Yang et~al.(2023{\natexlab{d}})Yang, Prabhakar, Narasimhan, and Yao]{2023InterCode}
John Yang, Akshara Prabhakar, Karthik Narasimhan, and Shunyu Yao.
\newblock Intercode: Standardizing and benchmarking interactive coding with execution feedback.
\newblock In Alice Oh, Tristan Naumann, Amir Globerson, Kate Saenko, Moritz Hardt, and Sergey Levine (eds.), \emph{Advances in Neural Information Processing Systems 36: Annual Conference on Neural Information Processing Systems 2023, NeurIPS 2023, New Orleans, LA, USA, December 10 - 16, 2023}, 2023{\natexlab{d}}.
\newblock URL \url{http://papers.nips.cc/paper\_files/paper/2023/hash/4b175d846fb008d540d233c188379ff9-Abstract-Datasets\_and\_Benchmarks.html}.

\bibitem[Yang et~al.(2024{\natexlab{b}})Yang, Jimenez, Wettig, Lieret, Yao, Narasimhan, and Press]{2024SWE-agent}
John Yang, Carlos~E. Jimenez, Alexander Wettig, Kilian Lieret, Shunyu Yao, Karthik Narasimhan, and Ofir Press.
\newblock Swe-agent: Agent-computer interfaces enable automated software engineering.
\newblock \emph{CoRR}, abs/2405.15793, 2024{\natexlab{b}}.
\newblock \doi{10.48550/ARXIV.2405.15793}.
\newblock URL \url{https://doi.org/10.48550/arXiv.2405.15793}.

\bibitem[Yang et~al.(2021)Yang, Xu, and Cao]{sql2021-3}
Wei Yang, Peng Xu, and Yanshuai Cao.
\newblock Hierarchical neural data synthesis for semantic parsing.
\newblock \emph{CoRR}, abs/2112.02212, 2021.
\newblock URL \url{https://arxiv.org/abs/2112.02212}.

\bibitem[Yang et~al.(2023{\natexlab{e}})Yang, Zhang, Chen, Petzold, Wang, and Cheng]{2023detect}
Xianjun Yang, Kexun Zhang, Haifeng Chen, Linda~R. Petzold, William~Yang Wang, and Wei Cheng.
\newblock Zero-shot detection of machine-generated codes.
\newblock \emph{CoRR}, abs/2310.05103, 2023{\natexlab{e}}.
\newblock \doi{10.48550/ARXIV.2310.05103}.
\newblock URL \url{https://doi.org/10.48550/arXiv.2310.05103}.

\bibitem[Yang et~al.(2022)Yang, Xu, Yan, Xu, and Deng]{id2022-3}
Yanping Yang, Ling Xu, Meng Yan, Zhou Xu, and Zhongyang Deng.
\newblock A naming pattern based approach for method name recommendation.
\newblock In \emph{{IEEE} 33rd International Symposium on Software Reliability Engineering, {ISSRE} 2022, Charlotte, NC, USA, October 31 - Nov. 3, 2022}, pp.\  344--354. {IEEE}, 2022.
\newblock \doi{10.1109/ISSRE55969.2022.00041}.
\newblock URL \url{https://doi.org/10.1109/ISSRE55969.2022.00041}.

\bibitem[Yao et~al.(2018)Yao, Weld, Chen, and Sun]{retrieval-data-2018-1}
Ziyu Yao, Daniel~S. Weld, Wei{-}Peng Chen, and Huan Sun.
\newblock Staqc: {A} systematically mined question-code dataset from stack overflow.
\newblock In Pierre{-}Antoine Champin, Fabien Gandon, Mounia Lalmas, and Panagiotis~G. Ipeirotis (eds.), \emph{Proceedings of the 2018 World Wide Web Conference on World Wide Web, {WWW} 2018, Lyon, France, April 23-27, 2018}, pp.\  1693--1703. {ACM}, 2018.
\newblock \doi{10.1145/3178876.3186081}.
\newblock URL \url{https://doi.org/10.1145/3178876.3186081}.

\bibitem[Yasunaga \& Liang(2020)Yasunaga and Liang]{fix2020-1}
Michihiro Yasunaga and Percy Liang.
\newblock Graph-based, self-supervised program repair from diagnostic feedback.
\newblock In \emph{Proceedings of the 37th International Conference on Machine Learning, {ICML} 2020, 13-18 July 2020, Virtual Event}, volume 119 of \emph{Proceedings of Machine Learning Research}, pp.\  10799--10808. {PMLR}, 2020.
\newblock URL \url{http://proceedings.mlr.press/v119/yasunaga20a.html}.

\bibitem[Yasunaga \& Liang(2021)Yasunaga and Liang]{fix2021-3}
Michihiro Yasunaga and Percy Liang.
\newblock Break-it-fix-it: Unsupervised learning for program repair.
\newblock In Marina Meila and Tong Zhang (eds.), \emph{Proceedings of the 38th International Conference on Machine Learning, {ICML} 2021, 18-24 July 2021, Virtual Event}, volume 139 of \emph{Proceedings of Machine Learning Research}, pp.\  11941--11952. {PMLR}, 2021.
\newblock URL \url{http://proceedings.mlr.press/v139/yasunaga21a.html}.

\bibitem[Yee \& Guha(2023)Yee and Guha]{type2023-1}
Ming{-}Ho Yee and Arjun Guha.
\newblock Do machine learning models produce typescript types that type check?
\newblock In Karim Ali and Guido Salvaneschi (eds.), \emph{37th European Conference on Object-Oriented Programming, {ECOOP} 2023, July 17-21, 2023, Seattle, Washington, United States}, volume 263 of \emph{LIPIcs}, pp.\  37:1--37:28. Schloss Dagstuhl - Leibniz-Zentrum f{\"{u}}r Informatik, 2023.
\newblock \doi{10.4230/LIPIcs.ECOOP.2023.37}.
\newblock URL \url{https://doi.org/10.4230/LIPIcs.ECOOP.2023.37}.

\bibitem[Yetistiren et~al.(2023)Yetistiren, {\"{O}}zsoy, Ayerdem, and T{\"{u}}z{\"{u}}n]{analysis2023-2}
Burak Yetistiren, Isik {\"{O}}zsoy, Miray Ayerdem, and Eray T{\"{u}}z{\"{u}}n.
\newblock Evaluating the code quality of ai-assisted code generation tools: An empirical study on github copilot, amazon codewhisperer, and chatgpt.
\newblock \emph{CoRR}, abs/2304.10778, 2023.
\newblock \doi{10.48550/ARXIV.2304.10778}.
\newblock URL \url{https://doi.org/10.48550/arXiv.2304.10778}.

\bibitem[Yin \& Neubig(2017)Yin and Neubig]{syn2017-2}
Pengcheng Yin and Graham Neubig.
\newblock A syntactic neural model for general-purpose code generation.
\newblock In Regina Barzilay and Min{-}Yen Kan (eds.), \emph{Proceedings of the 55th Annual Meeting of the Association for Computational Linguistics, {ACL} 2017, Vancouver, Canada, July 30 - August 4, Volume 1: Long Papers}, pp.\  440--450. Association for Computational Linguistics, 2017.
\newblock \doi{10.18653/v1/P17-1041}.
\newblock URL \url{https://doi.org/10.18653/v1/P17-1041}.

\bibitem[Yin et~al.(2018)Yin, Deng, Chen, Vasilescu, and Neubig]{retrieval2018-2}
Pengcheng Yin, Bowen Deng, Edgar Chen, Bogdan Vasilescu, and Graham Neubig.
\newblock Learning to mine aligned code and natural language pairs from stack overflow.
\newblock In Andy Zaidman, Yasutaka Kamei, and Emily Hill (eds.), \emph{Proceedings of the 15th International Conference on Mining Software Repositories, {MSR} 2018, Gothenburg, Sweden, May 28-29, 2018}, pp.\  476--486. {ACM}, 2018.
\newblock \doi{10.1145/3196398.3196408}.
\newblock URL \url{https://doi.org/10.1145/3196398.3196408}.

\bibitem[Yin et~al.(2020)Yin, Neubig, Yih, and Riedel]{sql2020-2.2}
Pengcheng Yin, Graham Neubig, Wen{-}tau Yih, and Sebastian Riedel.
\newblock Tabert: Pretraining for joint understanding of textual and tabular data.
\newblock In Dan Jurafsky, Joyce Chai, Natalie Schluter, and Joel~R. Tetreault (eds.), \emph{Proceedings of the 58th Annual Meeting of the Association for Computational Linguistics, {ACL} 2020, Online, July 5-10, 2020}, pp.\  8413--8426. Association for Computational Linguistics, 2020.
\newblock \doi{10.18653/V1/2020.ACL-MAIN.745}.
\newblock URL \url{https://doi.org/10.18653/v1/2020.acl-main.745}.

\bibitem[Yin et~al.(2023)Yin, Zhao, Sun, and Chen]{review2023-1}
Ying Yin, Yuhai Zhao, Yiming Sun, and Chen Chen.
\newblock Automatic code review by learning the structure information of code graph.
\newblock \emph{Sensors}, 23\penalty0 (5):\penalty0 2551, 2023.
\newblock \doi{10.3390/s23052551}.
\newblock URL \url{https://doi.org/10.3390/s23052551}.

\bibitem[Young et~al.(2024)Young, Chen, Li, Huang, Zhang, Zhang, Li, Zhu, Chen, Chang, Yu, Liu, Liu, Yue, Yang, Yang, Yu, Xie, Huang, Hu, Ren, Niu, Nie, Xu, Liu, Wang, Cai, Gu, Liu, and Dai]{2024Yi}
Alex Young, Bei Chen, Chao Li, Chengen Huang, Ge~Zhang, Guanwei Zhang, Heng Li, Jiangcheng Zhu, Jianqun Chen, Jing Chang, Kaidong Yu, Peng Liu, Qiang Liu, Shawn Yue, Senbin Yang, Shiming Yang, Tao Yu, Wen Xie, Wenhao Huang, Xiaohui Hu, Xiaoyi Ren, Xinyao Niu, Pengcheng Nie, Yuchi Xu, Yudong Liu, Yue Wang, Yuxuan Cai, Zhenyu Gu, Zhiyuan Liu, and Zonghong Dai.
\newblock Yi: Open foundation models by 01.ai.
\newblock \emph{CoRR}, abs/2403.04652, 2024.
\newblock \doi{10.48550/ARXIV.2403.04652}.
\newblock URL \url{https://doi.org/10.48550/arXiv.2403.04652}.

\bibitem[Yu et~al.(2019{\natexlab{a}})Yu, Lam, Chen, Li, Xie, and Wang]{clone2019-2}
Hao Yu, Wing Lam, Long Chen, Ge~Li, Tao Xie, and Qianxiang Wang.
\newblock Neural detection of semantic code clones via tree-based convolution.
\newblock In Yann{-}Ga{\"{e}}l Gu{\'{e}}h{\'{e}}neuc, Foutse Khomh, and Federica Sarro (eds.), \emph{Proceedings of the 27th International Conference on Program Comprehension, {ICPC} 2019, Montreal, QC, Canada, May 25-31, 2019}, pp.\  70--80. {IEEE} / {ACM}, 2019{\natexlab{a}}.
\newblock \doi{10.1109/ICPC.2019.00021}.
\newblock URL \url{https://doi.org/10.1109/ICPC.2019.00021}.

\bibitem[Yu et~al.(2023{\natexlab{a}})Yu, Shen, Ran, Zhang, Zhang, Ma, Liang, Li, Xie, and Wang]{2023CoderEval}
Hao Yu, Bo~Shen, Dezhi Ran, Jiaxin Zhang, Qi~Zhang, Yuchi Ma, Guangtai Liang, Ying Li, Tao Xie, and Qianxiang Wang.
\newblock Codereval: {A} benchmark of pragmatic code generation with generative pre-trained models.
\newblock \emph{CoRR}, abs/2302.00288, 2023{\natexlab{a}}.
\newblock \doi{10.48550/arXiv.2302.00288}.
\newblock URL \url{https://doi.org/10.48550/arXiv.2302.00288}.

\bibitem[Yu et~al.(2023{\natexlab{b}})Yu, Chen, Wu, and Dou]{log2023-3}
Siyu Yu, Ningjiang Chen, Yifan Wu, and Wensheng Dou.
\newblock Self-supervised log parsing using semantic contribution difference.
\newblock \emph{J. Syst. Softw.}, 200:\penalty0 111646, 2023{\natexlab{b}}.
\newblock \doi{10.1016/J.JSS.2023.111646}.
\newblock URL \url{https://doi.org/10.1016/j.jss.2023.111646}.

\bibitem[Yu et~al.(2018{\natexlab{a}})Yu, Li, Zhang, Zhang, and Radev]{sql2018-2}
Tao Yu, Zifan Li, Zilin Zhang, Rui Zhang, and Dragomir~R. Radev.
\newblock Typesql: Knowledge-based type-aware neural text-to-sql generation.
\newblock In Marilyn~A. Walker, Heng Ji, and Amanda Stent (eds.), \emph{Proceedings of the 2018 Conference of the North American Chapter of the Association for Computational Linguistics: Human Language Technologies, NAACL-HLT, New Orleans, Louisiana, USA, June 1-6, 2018, Volume 2 (Short Papers)}, pp.\  588--594. Association for Computational Linguistics, 2018{\natexlab{a}}.
\newblock \doi{10.18653/v1/n18-2093}.
\newblock URL \url{https://doi.org/10.18653/v1/n18-2093}.

\bibitem[Yu et~al.(2018{\natexlab{b}})Yu, Yasunaga, Yang, Zhang, Wang, Li, and Radev]{sql2018-6}
Tao Yu, Michihiro Yasunaga, Kai Yang, Rui Zhang, Dongxu Wang, Zifan Li, and Dragomir~R. Radev.
\newblock Syntaxsqlnet: Syntax tree networks for complex and cross-domain text-to-sql task.
\newblock In Ellen Riloff, David Chiang, Julia Hockenmaier, and Jun'ichi Tsujii (eds.), \emph{Proceedings of the 2018 Conference on Empirical Methods in Natural Language Processing, Brussels, Belgium, October 31 - November 4, 2018}, pp.\  1653--1663. Association for Computational Linguistics, 2018{\natexlab{b}}.
\newblock \doi{10.18653/v1/d18-1193}.
\newblock URL \url{https://doi.org/10.18653/v1/d18-1193}.

\bibitem[Yu et~al.(2018{\natexlab{c}})Yu, Zhang, Yang, Yasunaga, Wang, Li, Ma, Li, Yao, Roman, Zhang, and Radev]{sql2018-5}
Tao Yu, Rui Zhang, Kai Yang, Michihiro Yasunaga, Dongxu Wang, Zifan Li, James Ma, Irene Li, Qingning Yao, Shanelle Roman, Zilin Zhang, and Dragomir~R. Radev.
\newblock Spider: {A} large-scale human-labeled dataset for complex and cross-domain semantic parsing and text-to-sql task.
\newblock In Ellen Riloff, David Chiang, Julia Hockenmaier, and Jun'ichi Tsujii (eds.), \emph{Proceedings of the 2018 Conference on Empirical Methods in Natural Language Processing, Brussels, Belgium, October 31 - November 4, 2018}, pp.\  3911--3921. Association for Computational Linguistics, 2018{\natexlab{c}}.
\newblock \doi{10.18653/v1/d18-1425}.
\newblock URL \url{https://doi.org/10.18653/v1/d18-1425}.

\bibitem[Yu et~al.(2019{\natexlab{b}})Yu, Zhang, Er, Li, Xue, Pang, Lin, Tan, Shi, Li, Jiang, Yasunaga, Shim, Chen, Fabbri, Li, Chen, Zhang, Dixit, Zhang, Xiong, Socher, Lasecki, and Radev]{sql2019-6}
Tao Yu, Rui Zhang, Heyang Er, Suyi Li, Eric Xue, Bo~Pang, Xi~Victoria Lin, Yi~Chern Tan, Tianze Shi, Zihan Li, Youxuan Jiang, Michihiro Yasunaga, Sungrok Shim, Tao Chen, Alexander~R. Fabbri, Zifan Li, Luyao Chen, Yuwen Zhang, Shreya Dixit, Vincent Zhang, Caiming Xiong, Richard Socher, Walter~S. Lasecki, and Dragomir~R. Radev.
\newblock Cosql: {A} conversational text-to-sql challenge towards cross-domain natural language interfaces to databases.
\newblock In Kentaro Inui, Jing Jiang, Vincent Ng, and Xiaojun Wan (eds.), \emph{Proceedings of the 2019 Conference on Empirical Methods in Natural Language Processing and the 9th International Joint Conference on Natural Language Processing, {EMNLP-IJCNLP} 2019, Hong Kong, China, November 3-7, 2019}, pp.\  1962--1979. Association for Computational Linguistics, 2019{\natexlab{b}}.
\newblock \doi{10.18653/v1/D19-1204}.
\newblock URL \url{https://doi.org/10.18653/v1/D19-1204}.

\bibitem[Yu et~al.(2019{\natexlab{c}})Yu, Zhang, Yasunaga, Tan, Lin, Li, Er, Li, Pang, Chen, Ji, Dixit, Proctor, Shim, Kraft, Zhang, Xiong, Socher, and Radev]{sql2019-4}
Tao Yu, Rui Zhang, Michihiro Yasunaga, Yi~Chern Tan, Xi~Victoria Lin, Suyi Li, Heyang Er, Irene Li, Bo~Pang, Tao Chen, Emily Ji, Shreya Dixit, David Proctor, Sungrok Shim, Jonathan Kraft, Vincent Zhang, Caiming Xiong, Richard Socher, and Dragomir~R. Radev.
\newblock Sparc: Cross-domain semantic parsing in context.
\newblock In Anna Korhonen, David~R. Traum, and Llu{\'{\i}}s M{\`{a}}rquez (eds.), \emph{Proceedings of the 57th Conference of the Association for Computational Linguistics, {ACL} 2019, Florence, Italy, July 28- August 2, 2019, Volume 1: Long Papers}, pp.\  4511--4523. Association for Computational Linguistics, 2019{\natexlab{c}}.
\newblock \doi{10.18653/v1/p19-1443}.
\newblock URL \url{https://doi.org/10.18653/v1/p19-1443}.

\bibitem[Yu et~al.(2021)Yu, Wu, Lin, Wang, Tan, Yang, Radev, Socher, and Xiong]{sql2020-2.7}
Tao Yu, Chien{-}Sheng Wu, Xi~Victoria Lin, Bailin Wang, Yi~Chern Tan, Xinyi Yang, Dragomir~R. Radev, Richard Socher, and Caiming Xiong.
\newblock Grappa: Grammar-augmented pre-training for table semantic parsing.
\newblock In \emph{9th International Conference on Learning Representations, {ICLR} 2021, Virtual Event, Austria, May 3-7, 2021}. OpenReview.net, 2021.
\newblock URL \url{https://openreview.net/forum?id=kyaIeYj4zZ}.

\bibitem[Yu et~al.(2020)Yu, Chen, Yu, Li, Yang, Jiang, and Jiang]{sql-data-2020-2}
Xiaojing Yu, Tianlong Chen, Zhengjie Yu, Huiyu Li, Yang Yang, Xiaoqian Jiang, and Anxiao Jiang.
\newblock Dataset and enhanced model for eligibility criteria-to-sql semantic parsing.
\newblock In Nicoletta Calzolari, Fr{\'{e}}d{\'{e}}ric B{\'{e}}chet, Philippe Blache, Khalid Choukri, Christopher Cieri, Thierry Declerck, Sara Goggi, Hitoshi Isahara, Bente Maegaard, Joseph Mariani, H{\'{e}}l{\`{e}}ne Mazo, Asunci{\'{o}}n Moreno, Jan Odijk, and Stelios Piperidis (eds.), \emph{Proceedings of The 12th Language Resources and Evaluation Conference, {LREC} 2020, Marseille, France, May 11-16, 2020}, pp.\  5829--5837. European Language Resources Association, 2020.
\newblock URL \url{https://aclanthology.org/2020.lrec-1.714/}.

\bibitem[Yu et~al.(2023{\natexlab{c}})Yu, Zhang, Shang, Huang, Xu, Zhao, Hu, and Yin]{2023WaveCoder}
Zhaojian Yu, Xin Zhang, Ning Shang, Yangyu Huang, Can Xu, Yishujie Zhao, Wenxiang Hu, and Qiufeng Yin.
\newblock Wavecoder: Widespread and versatile enhanced instruction tuning with refined data generation.
\newblock \emph{CoRR}, abs/2312.14187, 2023{\natexlab{c}}.
\newblock \doi{10.48550/ARXIV.2312.14187}.
\newblock URL \url{https://doi.org/10.48550/arXiv.2312.14187}.

\bibitem[Yuan et~al.(2023{\natexlab{a}})Yuan, Liu, Zi, Liu, Peng, and Lou]{defect2023-1}
Zhiqiang Yuan, Junwei Liu, Qiancheng Zi, Mingwei Liu, Xin Peng, and Yiling Lou.
\newblock Evaluating instruction-tuned large language models on code comprehension and generation.
\newblock \emph{CoRR}, abs/2308.01240, 2023{\natexlab{a}}.
\newblock \doi{10.48550/arXiv.2308.01240}.
\newblock URL \url{https://doi.org/10.48550/arXiv.2308.01240}.

\bibitem[Yuan et~al.(2023{\natexlab{b}})Yuan, Lou, Liu, Ding, Wang, Chen, and Peng]{unit2023-5}
Zhiqiang Yuan, Yiling Lou, Mingwei Liu, Shiji Ding, Kaixin Wang, Yixuan Chen, and Xin Peng.
\newblock No more manual tests? evaluating and improving chatgpt for unit test generation.
\newblock \emph{CoRR}, abs/2305.04207, 2023{\natexlab{b}}.
\newblock \doi{10.48550/ARXIV.2305.04207}.
\newblock URL \url{https://doi.org/10.48550/arXiv.2305.04207}.

\bibitem[Zahan et~al.(2024)Zahan, Burckhardt, Lysenko, Aboukhadijeh, and Williams]{2024SocketAIScanner}
Nusrat Zahan, Philipp Burckhardt, Mikola Lysenko, Feross Aboukhadijeh, and Laurie~A. Williams.
\newblock Shifting the lens: Detecting malware in npm ecosystem with large language models.
\newblock \emph{CoRR}, abs/2403.12196, 2024.
\newblock \doi{10.48550/ARXIV.2403.12196}.
\newblock URL \url{https://doi.org/10.48550/arXiv.2403.12196}.

\bibitem[Zan et~al.(2022)Zan, Chen, Yang, Lin, Kim, Guan, Wang, Chen, and Lou]{2022PyCodeGPT}
Daoguang Zan, Bei Chen, Dejian Yang, Zeqi Lin, Minsu Kim, Bei Guan, Yongji Wang, Weizhu Chen, and Jian{-}Guang Lou.
\newblock {CERT:} continual pre-training on sketches for library-oriented code generation.
\newblock In Luc~De Raedt (ed.), \emph{Proceedings of the Thirty-First International Joint Conference on Artificial Intelligence, {IJCAI} 2022, Vienna, Austria, 23-29 July 2022}, pp.\  2369--2375. ijcai.org, 2022.
\newblock \doi{10.24963/ijcai.2022/329}.
\newblock URL \url{https://doi.org/10.24963/ijcai.2022/329}.

\bibitem[Zan et~al.(2023)Zan, Chen, Zhang, Lu, Wu, Guan, Wang, and Lou]{2022survey1}
Daoguang Zan, Bei Chen, Fengji Zhang, Dianjie Lu, Bingchao Wu, Bei Guan, Yongji Wang, and Jian{-}Guang Lou.
\newblock Large language models meet nl2code: {A} survey.
\newblock In Anna Rogers, Jordan~L. Boyd{-}Graber, and Naoaki Okazaki (eds.), \emph{Proceedings of the 61st Annual Meeting of the Association for Computational Linguistics (Volume 1: Long Papers), {ACL} 2023, Toronto, Canada, July 9-14, 2023}, pp.\  7443--7464. Association for Computational Linguistics, 2023.
\newblock \doi{10.18653/v1/2023.acl-long.411}.
\newblock URL \url{https://doi.org/10.18653/v1/2023.acl-long.411}.

\bibitem[Zan et~al.(2024)Zan, Yu, Liu, Chen, Shen, Li, Yao, Gong, Chen, Guan, Yang, Wang, Wang, and Cui]{2024CodeS}
Daoguang Zan, Ailun Yu, Wei Liu, Dong Chen, Bo~Shen, Wei Li, Yafen Yao, Yongshun Gong, Xiaolin Chen, Bei Guan, Zhiguang Yang, Yongji Wang, Qianxiang Wang, and Lizhen Cui.
\newblock Codes: Natural language to code repository via multi-layer sketch.
\newblock \emph{CoRR}, abs/2403.16443, 2024.
\newblock \doi{10.48550/ARXIV.2403.16443}.
\newblock URL \url{https://doi.org/10.48550/arXiv.2403.16443}.

\bibitem[Zelle \& Mooney(1996)Zelle and Mooney]{sql-data-1996}
John~M. Zelle and Raymond~J. Mooney.
\newblock Learning to parse database queries using inductive logic programming.
\newblock In William~J. Clancey and Daniel~S. Weld (eds.), \emph{Proceedings of the Thirteenth National Conference on Artificial Intelligence and Eighth Innovative Applications of Artificial Intelligence Conference, {AAAI} 96, {IAAI} 96, Portland, Oregon, USA, August 4-8, 1996, Volume 2}, pp.\  1050--1055. {AAAI} Press / The {MIT} Press, 1996.
\newblock URL \url{http://www.aaai.org/Library/AAAI/1996/aaai96-156.php}.

\bibitem[Zeng et~al.(2020)Zeng, Lin, Hoi, Socher, Xiong, Lyu, and King]{sql2020-2.3}
Jichuan Zeng, Xi~Victoria Lin, Steven C.~H. Hoi, Richard Socher, Caiming Xiong, Michael~R. Lyu, and Irwin King.
\newblock Photon: {A} robust cross-domain text-to-sql system.
\newblock In Asli Celikyilmaz and Tsung{-}Hsien Wen (eds.), \emph{Proceedings of the 58th Annual Meeting of the Association for Computational Linguistics: System Demonstrations, {ACL} 2020, Online, July 5-10, 2020}, pp.\  204--214. Association for Computational Linguistics, 2020.
\newblock \doi{10.18653/V1/2020.ACL-DEMOS.24}.
\newblock URL \url{https://doi.org/10.18653/v1/2020.acl-demos.24}.

\bibitem[Zhang et~al.(2023{\natexlab{a}})Zhang, Liu, Zeng, Yang, Li, and Li]{defect2023-2}
Chenyuan Zhang, Hao Liu, Jiutian Zeng, Kejing Yang, Yuhong Li, and Hui Li.
\newblock Prompt-enhanced software vulnerability detection using chatgpt.
\newblock \emph{CoRR}, abs/2308.12697, 2023{\natexlab{a}}.
\newblock \doi{10.48550/ARXIV.2308.12697}.
\newblock URL \url{https://doi.org/10.48550/arXiv.2308.12697}.

\bibitem[Zhang et~al.(2022)Zhang, Wang, Zhou, Xu, Tang, Gui, and Liu]{sum-survey-2022}
Chunyan Zhang, Junchao Wang, Qinglei Zhou, Ting Xu, Ke~Tang, Hairen Gui, and Fudong Liu.
\newblock A survey of automatic source code summarization.
\newblock \emph{Symmetry}, 14\penalty0 (3):\penalty0 471, 2022.
\newblock \doi{10.3390/SYM14030471}.
\newblock URL \url{https://doi.org/10.3390/sym14030471}.

\bibitem[Zhang et~al.(2024{\natexlab{a}})Zhang, Ahmad, Tan, Ding, Nallapati, Roth, Ma, and Xiang]{2024CodeSage}
Dejiao Zhang, Wasi~Uddin Ahmad, Ming Tan, Hantian Ding, Ramesh Nallapati, Dan Roth, Xiaofei Ma, and Bing Xiang.
\newblock Code representation learning at scale.
\newblock \emph{CoRR}, abs/2402.01935, 2024{\natexlab{a}}.
\newblock \doi{10.48550/ARXIV.2402.01935}.
\newblock URL \url{https://doi.org/10.48550/arXiv.2402.01935}.

\bibitem[Zhang et~al.(2023{\natexlab{b}})Zhang, Chen, Zhang, Liu, Zan, Mao, Lou, and Chen]{repo2023-1}
Fengji Zhang, Bei Chen, Yue Zhang, Jin Liu, Daoguang Zan, Yi~Mao, Jian{-}Guang Lou, and Weizhu Chen.
\newblock Repocoder: Repository-level code completion through iterative retrieval and generation.
\newblock \emph{CoRR}, abs/2303.12570, 2023{\natexlab{b}}.
\newblock \doi{10.48550/ARXIV.2303.12570}.
\newblock URL \url{https://doi.org/10.48550/arXiv.2303.12570}.

\bibitem[Zhang \& Sakurai(2021)Zhang and Sakurai]{clone-survey-2021-1}
Haibo Zhang and Kouichi Sakurai.
\newblock A survey of software clone detection from security perspective.
\newblock \emph{{IEEE} Access}, 9:\penalty0 48157--48173, 2021.
\newblock \doi{10.1109/ACCESS.2021.3065872}.
\newblock URL \url{https://doi.org/10.1109/ACCESS.2021.3065872}.

\bibitem[Zhang et~al.(2019{\natexlab{a}})Zhang, Wang, Zhang, Sun, Wang, and Liu]{clone2019-1}
Jian Zhang, Xu~Wang, Hongyu Zhang, Hailong Sun, Kaixuan Wang, and Xudong Liu.
\newblock A novel neural source code representation based on abstract syntax tree.
\newblock In Joanne~M. Atlee, Tevfik Bultan, and Jon Whittle (eds.), \emph{Proceedings of the 41st International Conference on Software Engineering, {ICSE} 2019, Montreal, QC, Canada, May 25-31, 2019}, pp.\  783--794. {IEEE} / {ACM}, 2019{\natexlab{a}}.
\newblock \doi{10.1109/ICSE.2019.00086}.
\newblock URL \url{https://doi.org/10.1109/ICSE.2019.00086}.

\bibitem[Zhang \& El-Gohary(2017)Zhang and El-Gohary]{re-ana2016-2}
Jiansong Zhang and Nora~M. El-Gohary.
\newblock Integrating semantic nlp and logic reasoning into a unified system for fully-automated code checking.
\newblock \emph{Automation in Construction}, 73:\penalty0 45--57, 2017.
\newblock ISSN 0926-5805.
\newblock \doi{https://doi.org/10.1016/j.autcon.2016.08.027}.
\newblock URL \url{https://www.sciencedirect.com/science/article/pii/S0926580516301819}.

\bibitem[Zhang et~al.(2018)Zhang, Jiang, Ren, and Chen]{api2017-4}
Jingxuan Zhang, He~Jiang, Zhilei Ren, and Xin Chen.
\newblock Recommending apis for {API} related questions in stack overflow.
\newblock \emph{{IEEE} Access}, 6:\penalty0 6205--6219, 2018.
\newblock \doi{10.1109/ACCESS.2017.2777845}.
\newblock URL \url{https://doi.org/10.1109/ACCESS.2017.2777845}.

\bibitem[Zhang et~al.(2023{\natexlab{c}})Zhang, Li, Li, Li, and Jin]{2023Self-Edit}
Kechi Zhang, Zhuo Li, Jia Li, Ge~Li, and Zhi Jin.
\newblock Self-edit: Fault-aware code editor for code generation.
\newblock In Anna Rogers, Jordan~L. Boyd{-}Graber, and Naoaki Okazaki (eds.), \emph{Proceedings of the 61st Annual Meeting of the Association for Computational Linguistics (Volume 1: Long Papers), {ACL} 2023, Toronto, Canada, July 9-14, 2023}, pp.\  769--787. Association for Computational Linguistics, 2023{\natexlab{c}}.
\newblock \doi{10.18653/V1/2023.ACL-LONG.45}.
\newblock URL \url{https://doi.org/10.18653/v1/2023.acl-long.45}.

\bibitem[Zhang et~al.(2023{\natexlab{d}})Zhang, Fang, Ma, Sun, and Chen]{fix-survey-2023}
Quanjun Zhang, Chunrong Fang, Yuxiang Ma, Weisong Sun, and Zhenyu Chen.
\newblock A survey of learning-based automated program repair.
\newblock \emph{CoRR}, abs/2301.03270, 2023{\natexlab{d}}.
\newblock \doi{10.48550/arXiv.2301.03270}.
\newblock URL \url{https://doi.org/10.48550/arXiv.2301.03270}.

\bibitem[Zhang et~al.(2019{\natexlab{b}})Zhang, Yu, Er, Shim, Xue, Lin, Shi, Xiong, Socher, and Radev]{sql2019-5}
Rui Zhang, Tao Yu, Heyang Er, Sungrok Shim, Eric Xue, Xi~Victoria Lin, Tianze Shi, Caiming Xiong, Richard Socher, and Dragomir~R. Radev.
\newblock Editing-based {SQL} query generation for cross-domain context-dependent questions.
\newblock In Kentaro Inui, Jing Jiang, Vincent Ng, and Xiaojun Wan (eds.), \emph{Proceedings of the 2019 Conference on Empirical Methods in Natural Language Processing and the 9th International Joint Conference on Natural Language Processing, {EMNLP-IJCNLP} 2019, Hong Kong, China, November 3-7, 2019}, pp.\  5337--5348. Association for Computational Linguistics, 2019{\natexlab{b}}.
\newblock \doi{10.18653/v1/D19-1537}.
\newblock URL \url{https://doi.org/10.18653/v1/D19-1537}.

\bibitem[Zhang et~al.(2024{\natexlab{b}})Zhang, Zhao, Liu, Zheng, Qi, Gu, Zhang, Dong, and Tang]{2024NaturalCodeBench}
Shudan Zhang, Hanlin Zhao, Xiao Liu, Qinkai Zheng, Zehan Qi, Xiaotao Gu, Xiaohan Zhang, Yuxiao Dong, and Jie Tang.
\newblock Naturalcodebench: Examining coding performance mismatch on humaneval and natural user prompts.
\newblock 2024{\natexlab{b}}.
\newblock URL \url{https://doi.org/10.48550/arXiv.2405.04520}.

\bibitem[Zhang et~al.(2023{\natexlab{e}})Zhang, Qiu, Castellano, Rifai, Chen, and Pianese]{log-survey-2022-1}
Tianzhu Zhang, Han Qiu, Gabriele Castellano, Myriana Rifai, Chung~Shue Chen, and Fabio Pianese.
\newblock System log parsing: {A} survey.
\newblock \emph{{IEEE} Trans. Knowl. Data Eng.}, 35\penalty0 (8):\penalty0 8596--8614, 2023{\natexlab{e}}.
\newblock \doi{10.1109/TKDE.2022.3222417}.
\newblock URL \url{https://doi.org/10.1109/TKDE.2022.3222417}.

\bibitem[Zhang et~al.(2024{\natexlab{c}})Zhang, Muralee, Cherupattamoolayil, and Machiry]{workflow2024-1}
Xinyu Zhang, Siddharth Muralee, Sourag Cherupattamoolayil, and Aravind Machiry.
\newblock On the effectiveness of large language models for github workflows.
\newblock \emph{CoRR}, abs/2403.12446, 2024{\natexlab{c}}.
\newblock \doi{10.48550/ARXIV.2403.12446}.
\newblock URL \url{https://doi.org/10.48550/arXiv.2403.12446}.

\bibitem[Zhang et~al.(2019{\natexlab{c}})Zhang, Xu, Lin, Qiao, Zhang, Dang, Xie, Yang, Cheng, Li, Chen, He, Yao, Lou, Chintalapati, Shen, and Zhang]{log2019-2}
Xu~Zhang, Yong Xu, Qingwei Lin, Bo~Qiao, Hongyu Zhang, Yingnong Dang, Chunyu Xie, Xinsheng Yang, Qian Cheng, Ze~Li, Junjie Chen, Xiaoting He, Randolph Yao, Jian{-}Guang Lou, Murali Chintalapati, Furao Shen, and Dongmei Zhang.
\newblock Robust log-based anomaly detection on unstable log data.
\newblock In Marlon Dumas, Dietmar Pfahl, Sven Apel, and Alessandra Russo (eds.), \emph{Proceedings of the {ACM} Joint Meeting on European Software Engineering Conference and Symposium on the Foundations of Software Engineering, {ESEC/SIGSOFT} {FSE} 2019, Tallinn, Estonia, August 26-30, 2019}, pp.\  807--817. {ACM}, 2019{\natexlab{c}}.
\newblock \doi{10.1145/3338906.3338931}.
\newblock URL \url{https://doi.org/10.1145/3338906.3338931}.

\bibitem[Zhang et~al.(2023{\natexlab{f}})Zhang, Song, Ji, Yao, and Meng]{unit2023-9}
Ying Zhang, Wenjia Song, Zhengjie Ji, Danfeng Yao, and Na~Meng.
\newblock How well does {LLM} generate security tests?
\newblock \emph{CoRR}, abs/2310.00710, 2023{\natexlab{f}}.
\newblock \doi{10.48550/ARXIV.2310.00710}.
\newblock URL \url{https://doi.org/10.48550/arXiv.2310.00710}.

\bibitem[Zhang et~al.(2023{\natexlab{g}})Zhang, Liu, Huang, Mao, Wang, and Hu]{2023MELA}
Ziyin Zhang, Yikang Liu, Weifang Huang, Junyu Mao, Rui Wang, and Hai Hu.
\newblock {MELA:} multilingual evaluation of linguistic acceptability.
\newblock \emph{CoRR}, abs/2311.09033, 2023{\natexlab{g}}.
\newblock \doi{10.48550/ARXIV.2311.09033}.
\newblock URL \url{https://doi.org/10.48550/arXiv.2311.09033}.

\bibitem[Zhang et~al.(2024{\natexlab{d}})Zhang, Xu, Jiang, Hao, and Wang]{2024PythonIO}
Ziyin Zhang, Lizhen Xu, Zhaokun Jiang, Hongkun Hao, and Rui Wang.
\newblock Multiple-choice questions are efficient and robust llm evaluators.
\newblock 2024{\natexlab{d}}.
\newblock URL \url{https://doi.org/10.48550/arXiv.2405.11966}.

\bibitem[Zhao \& Huang(2018)Zhao and Huang]{clone2018-3}
Gang Zhao and Jeff Huang.
\newblock Deepsim: deep learning code functional similarity.
\newblock In Gary~T. Leavens, Alessandro Garcia, and Corina~S. Pasareanu (eds.), \emph{Proceedings of the 2018 {ACM} Joint Meeting on European Software Engineering Conference and Symposium on the Foundations of Software Engineering, {ESEC/SIGSOFT} {FSE} 2018, Lake Buena Vista, FL, USA, November 04-09, 2018}, pp.\  141--151. {ACM}, 2018.
\newblock \doi{10.1145/3236024.3236068}.
\newblock URL \url{https://doi.org/10.1145/3236024.3236068}.

\bibitem[Zhao et~al.(2021)Zhao, Cao, and Zhao]{sql2021-0.5}
Liang Zhao, Hexin Cao, and Yunsong Zhao.
\newblock {GP:} context-free grammar pre-training for text-to-sql parsers.
\newblock \emph{CoRR}, abs/2101.09901, 2021.
\newblock URL \url{https://arxiv.org/abs/2101.09901}.

\bibitem[Zhao et~al.(2023)Zhao, Feng, Feng, Qin, and Liu]{2023PESurvey}
Liang Zhao, Xiaocheng Feng, Xiachong Feng, Bing Qin, and Ting Liu.
\newblock Length extrapolation of transformers: {A} survey from the perspective of position encoding.
\newblock \emph{CoRR}, abs/2312.17044, 2023.
\newblock \doi{10.48550/ARXIV.2312.17044}.
\newblock URL \url{https://doi.org/10.48550/arXiv.2312.17044}.

\bibitem[Zhao et~al.(2020)Zhao, Alhoshan, Ferrari, Letsholo, Ajagbe, Chioasca, and Batista{-}Navarro]{re-ana-survey-2020-1}
Liping Zhao, Waad Alhoshan, Alessio Ferrari, Keletso~J. Letsholo, Muideen~A. Ajagbe, Erol{-}Valeriu Chioasca, and Riza~Theresa Batista{-}Navarro.
\newblock Natural language processing {(NLP)} for requirements engineering: {A} systematic mapping study.
\newblock \emph{CoRR}, abs/2004.01099, 2020.
\newblock URL \url{https://arxiv.org/abs/2004.01099}.

\bibitem[Zheng et~al.(2023{\natexlab{a}})Zheng, Xia, Zou, Dong, Wang, Xue, Shen, Wang, Wang, Li, Su, Yang, and Tang]{2023CodeGeeX}
Qinkai Zheng, Xiao Xia, Xu~Zou, Yuxiao Dong, Shan Wang, Yufei Xue, Lei Shen, Zihan Wang, Andi Wang, Yang Li, Teng Su, Zhilin Yang, and Jie Tang.
\newblock Codegeex: A pre-trained model for code generation with multilingual benchmarking on humaneval-x.
\newblock In \emph{Proceedings of the 29th ACM SIGKDD Conference on Knowledge Discovery and Data Mining}, KDD '23, pp.\  5673–5684, New York, NY, USA, 2023{\natexlab{a}}. Association for Computing Machinery.
\newblock ISBN 9798400701030.
\newblock \doi{10.1145/3580305.3599790}.
\newblock URL \url{https://doi.org/10.1145/3580305.3599790}.

\bibitem[Zheng et~al.(2024)Zheng, Zhang, Shen, Liu, Lin, Fu, Chen, and Yue]{2024OpenCodeInterpreter}
Tianyu Zheng, Ge~Zhang, Tianhao Shen, Xueling Liu, Bill~Yuchen Lin, Jie Fu, Wenhu Chen, and Xiang Yue.
\newblock Opencodeinterpreter: Integrating code generation with execution and refinement.
\newblock \emph{CoRR}, abs/2402.14658, 2024.
\newblock \doi{10.48550/ARXIV.2402.14658}.
\newblock URL \url{https://doi.org/10.48550/arXiv.2402.14658}.

\bibitem[Zheng et~al.(2021)Zheng, Pujar, Lewis, Buratti, Epstein, Yang, Laredo, Morari, and Su]{defect-data-2021-1}
Yunhui Zheng, Saurabh Pujar, Burn~L. Lewis, Luca Buratti, Edward~A. Epstein, Bo~Yang, Jim Laredo, Alessandro Morari, and Zhong Su.
\newblock {D2A:} {A} dataset built for ai-based vulnerability detection methods using differential analysis.
\newblock In \emph{43rd {IEEE/ACM} International Conference on Software Engineering: Software Engineering in Practice, {ICSE} {(SEIP)} 2021, Madrid, Spain, May 25-28, 2021}, pp.\  111--120. {IEEE}, 2021.
\newblock \doi{10.1109/ICSE-SEIP52600.2021.00020}.
\newblock URL \url{https://doi.org/10.1109/ICSE-SEIP52600.2021.00020}.

\bibitem[Zheng et~al.(2023{\natexlab{b}})Zheng, Ning, Chen, Wang, Chen, Guo, and Wang]{2023survey3}
Zibin Zheng, Kaiwen Ning, Jiachi Chen, Yanlin Wang, Wenqing Chen, Lianghong Guo, and Weicheng Wang.
\newblock Towards an understanding of large language models in software engineering tasks.
\newblock \emph{CoRR}, abs/2308.11396, 2023{\natexlab{b}}.
\newblock \doi{10.48550/ARXIV.2308.11396}.
\newblock URL \url{https://doi.org/10.48550/arXiv.2308.11396}.

\bibitem[Zhong et~al.(2017)Zhong, Xiong, and Socher]{sql2017-1}
Victor Zhong, Caiming Xiong, and Richard Socher.
\newblock Seq2sql: Generating structured queries from natural language using reinforcement learning.
\newblock \emph{CoRR}, abs/1709.00103, 2017.
\newblock URL \url{http://arxiv.org/abs/1709.00103}.

\bibitem[Zhong et~al.(2020)Zhong, Lewis, Wang, and Zettlemoyer]{sql2020-2.5}
Victor Zhong, Mike Lewis, Sida~I. Wang, and Luke Zettlemoyer.
\newblock Grounded adaptation for zero-shot executable semantic parsing.
\newblock In Bonnie Webber, Trevor Cohn, Yulan He, and Yang Liu (eds.), \emph{Proceedings of the 2020 Conference on Empirical Methods in Natural Language Processing, {EMNLP} 2020, Online, November 16-20, 2020}, pp.\  6869--6882. Association for Computational Linguistics, 2020.
\newblock \doi{10.18653/V1/2020.EMNLP-MAIN.558}.
\newblock URL \url{https://doi.org/10.18653/v1/2020.emnlp-main.558}.

\bibitem[Zhong et~al.(2022)Zhong, Li, Ge, and Luo]{fix-survey-2022}
Wenkang Zhong, Chuanyi Li, Jidong Ge, and Bin Luo.
\newblock Neural program repair : Systems, challenges and solutions.
\newblock In \emph{Internetware 2022: 13th Asia-Pacific Symposium on Internetware, Hohhot, China, June 11 - 12, 2022}, pp.\  96--106. {ACM}, 2022.
\newblock \doi{10.1145/3545258.3545268}.
\newblock URL \url{https://doi.org/10.1145/3545258.3545268}.

\bibitem[Zhou et~al.(2023{\natexlab{a}})Zhou, Wang, Lu, Shi, Luo, Qin, Lu, Jia, Song, Zhan, and Li]{2023interpreter}
Aojun Zhou, Ke~Wang, Zimu Lu, Weikang Shi, Sichun Luo, Zipeng Qin, Shaoqing Lu, Anya Jia, Linqi Song, Mingjie Zhan, and Hongsheng Li.
\newblock Solving challenging math word problems using {GPT-4} code interpreter with code-based self-verification.
\newblock \emph{CoRR}, abs/2308.07921, 2023{\natexlab{a}}.
\newblock \doi{10.48550/ARXIV.2308.07921}.
\newblock URL \url{https://doi.org/10.48550/arXiv.2308.07921}.

\bibitem[Zhou et~al.(2023{\natexlab{b}})Zhou, Wang, Xu, Yao, Pan, Xu, and Ma]{api2023-1}
Bingzhe Zhou, Xinying Wang, Shengbin Xu, Yuan Yao, Minxue Pan, Feng Xu, and Xiaoxing Ma.
\newblock Hybrid {API} migration: {A} marriage of small {API} mapping models and large language models.
\newblock In Hong Mei, Jian Lv, Zhi Jin, Xuandong Li, Xiaohu Yang, and Xin Xia (eds.), \emph{Proceedings of the 14th Asia-Pacific Symposium on Internetware, Internetware 2023, Hangzhou, China, August 4-6, 2023}, pp.\  12--21. {ACM}, 2023{\natexlab{b}}.
\newblock \doi{10.1145/3609437.3609466}.
\newblock URL \url{https://doi.org/10.1145/3609437.3609466}.

\bibitem[Zhou et~al.(2024)Zhou, Cao, Sun, and Lo]{defect-survey-2024-1}
Xin Zhou, Sicong Cao, Xiaobing Sun, and David Lo.
\newblock Large language model for vulnerability detection and repair: Literature review and the road ahead.
\newblock \emph{CoRR}, abs/2404.02525, 2024.
\newblock \doi{10.48550/ARXIV.2404.02525}.
\newblock URL \url{https://doi.org/10.48550/arXiv.2404.02525}.

\bibitem[Zhou et~al.(2019)Zhou, Liu, Siow, Du, and Liu]{2019Devign}
Yaqin Zhou, Shangqing Liu, Jing~Kai Siow, Xiaoning Du, and Yang Liu.
\newblock Devign: Effective vulnerability identification by learning comprehensive program semantics via graph neural networks.
\newblock In Hanna~M. Wallach, Hugo Larochelle, Alina Beygelzimer, Florence d'Alch{\'{e}}{-}Buc, Emily~B. Fox, and Roman Garnett (eds.), \emph{Advances in Neural Information Processing Systems 32: Annual Conference on Neural Information Processing Systems 2019, NeurIPS 2019, December 8-14, 2019, Vancouver, BC, Canada}, pp.\  10197--10207, 2019.
\newblock URL \url{https://proceedings.neurips.cc/paper/2019/hash/49265d2447bc3bbfe9e76306ce40a31f-Abstract.html}.

\bibitem[Zhou et~al.(2023{\natexlab{c}})Zhou, Yu, Fan, Huang, and Yang]{sum2023-1}
Ziyi Zhou, Huiqun Yu, Guisheng Fan, Zijie Huang, and Kang Yang.
\newblock Towards retrieval-based neural code summarization: {A} meta-learning approach.
\newblock \emph{{IEEE} Trans. Software Eng.}, 49\penalty0 (4):\penalty0 3008--3031, 2023{\natexlab{c}}.
\newblock \doi{10.1109/TSE.2023.3238161}.
\newblock URL \url{https://doi.org/10.1109/TSE.2023.3238161}.

\bibitem[Zhu et~al.(2019)Zhu, He, Liu, He, Xie, Zheng, and Lyu]{log-survey-2018}
Jieming Zhu, Shilin He, Jinyang Liu, Pinjia He, Qi~Xie, Zibin Zheng, and Michael~R. Lyu.
\newblock Tools and benchmarks for automated log parsing.
\newblock In Helen Sharp and Mike Whalen (eds.), \emph{Proceedings of the 41st International Conference on Software Engineering: Software Engineering in Practice, {ICSE} {(SEIP)} 2019, Montreal, QC, Canada, May 25-31, 2019}, pp.\  121--130. {IEEE} / {ACM}, 2019.
\newblock \doi{10.1109/ICSE-SEIP.2019.00021}.
\newblock URL \url{https://doi.org/10.1109/ICSE-SEIP.2019.00021}.

\bibitem[Zhu et~al.(2022{\natexlab{a}})Zhu, Jain, Suresh, Ravindran, Tipirneni, and Reddy]{trans-data-2022-1}
Ming Zhu, Aneesh Jain, Karthik Suresh, Roshan Ravindran, Sindhu Tipirneni, and Chandan~K. Reddy.
\newblock Xlcost: {A} benchmark dataset for cross-lingual code intelligence.
\newblock \emph{CoRR}, abs/2206.08474, 2022{\natexlab{a}}.
\newblock \doi{10.48550/ARXIV.2206.08474}.
\newblock URL \url{https://doi.org/10.48550/arXiv.2206.08474}.

\bibitem[Zhu et~al.(2022{\natexlab{b}})Zhu, Suresh, and Reddy]{trans-data-2022-2}
Ming Zhu, Karthik Suresh, and Chandan~K. Reddy.
\newblock Multilingual code snippets training for program translation.
\newblock In \emph{Thirty-Sixth {AAAI} Conference on Artificial Intelligence, {AAAI} 2022, Thirty-Fourth Conference on Innovative Applications of Artificial Intelligence, {IAAI} 2022, The Twelveth Symposium on Educational Advances in Artificial Intelligence, {EAAI} 2022 Virtual Event, February 22 - March 1, 2022}, pp.\  11783--11790. {AAAI} Press, 2022{\natexlab{b}}.
\newblock \doi{10.1609/AAAI.V36I10.21434}.
\newblock URL \url{https://doi.org/10.1609/aaai.v36i10.21434}.

\bibitem[Zhu et~al.(2021)Zhu, Sun, Xiao, Zhang, Yuan, Xiong, and Zhang]{fix2021-4}
Qihao Zhu, Zeyu Sun, Yuan{-}an Xiao, Wenjie Zhang, Kang Yuan, Yingfei Xiong, and Lu~Zhang.
\newblock A syntax-guided edit decoder for neural program repair.
\newblock In Diomidis Spinellis, Georgios Gousios, Marsha Chechik, and Massimiliano~Di Penta (eds.), \emph{{ESEC/FSE} '21: 29th {ACM} Joint European Software Engineering Conference and Symposium on the Foundations of Software Engineering, Athens, Greece, August 23-28, 2021}, pp.\  341--353. {ACM}, 2021.
\newblock \doi{10.1145/3468264.3468544}.
\newblock URL \url{https://doi.org/10.1145/3468264.3468544}.

\bibitem[Zhu et~al.(2024)Zhu, Guo, Shao, Yang, Wang, Xu, Wu, Li, Gao, Ma, et~al.]{2024DeepSeekCoder-V2}
Qihao Zhu, Daya Guo, Zhihong Shao, Dejian Yang, Peiyi Wang, Runxin Xu, Y~Wu, Yukun Li, Huazuo Gao, Shirong Ma, et~al.
\newblock Deepseek-coder-v2: Breaking the barrier of closed-source models in code intelligence.
\newblock \emph{arXiv preprint arXiv:2406.11931}, 2024.
\newblock URL \url{https://doi.org/10.48550/arXiv.2406.11931}.

\bibitem[Zhuo et~al.(2024)Zhuo, Zebaze, Suppattarachai, von Werra, de~Vries, Liu, and Muennighoff]{2024Astraios}
Terry~Yue Zhuo, Armel Zebaze, Nitchakarn Suppattarachai, Leandro von Werra, Harm de~Vries, Qian Liu, and Niklas Muennighoff.
\newblock Astraios: Parameter-efficient instruction tuning code large language models.
\newblock \emph{CoRR}, abs/2401.00788, 2024.
\newblock \doi{10.48550/ARXIV.2401.00788}.
\newblock URL \url{https://doi.org/10.48550/arXiv.2401.00788}.

\bibitem[Zou et~al.(2021)Zou, Wang, Xu, Li, and Jin]{defect-data-2020-3}
Deqing Zou, Sujuan Wang, Shouhuai Xu, Zhen Li, and Hai Jin.
\newblock {\textdollar}{\textbackslash}mu{\textdollar}{\(\mu\)}vuldeepecker: {A} deep learning-based system for multiclass vulnerability detection.
\newblock \emph{{IEEE} Trans. Dependable Secur. Comput.}, 18\penalty0 (5):\penalty0 2224--2236, 2021.
\newblock \doi{10.1109/TDSC.2019.2942930}.
\newblock URL \url{https://doi.org/10.1109/TDSC.2019.2942930}.

\end{thebibliography}
